\documentclass{article}
\usepackage{preprint,times}

\usepackage{amsmath,amsfonts,bm}

\def\eqref#1{equation~\ref{#1}}

\def\1{\bm{1}}

\DeclareMathAlphabet{\mathsfit}{\encodingdefault}{\sfdefault}{m}{sl}
\SetMathAlphabet{\mathsfit}{bold}{\encodingdefault}{\sfdefault}{bx}{n}

\definecolor{mydarkblue}{rgb}{0,0.08,0.45}
\definecolor{mygray}{gray}{0.4}
\newcommand{\rulesep}{\unskip\ \vrule\ }

\usepackage[colorlinks=true,
    linkcolor=mydarkblue,
    citecolor=mydarkblue,
    filecolor=mydarkblue,
    urlcolor=mydarkblue]{hyperref}

\usepackage{caption}
\usepackage{subcaption}
\usepackage{graphicx}
\usepackage{url}
\usepackage{wrapfig}
\renewcommand{\v}[1]{\mathbf{#1}}

\iclrfinalcopy

\title{From Points to Functions: \\ Infinite-dimensional Representations in \\Diffusion Models}

\author{Sarthak Mittal\thanks{Correspondence author \href{mailto:sarthmit@gmail.com}{sarthmit@gmail.com}}$^{\;\,1,2}$\;, Guillaume Lajoie$^{1,2}$, Stefan Bauer$^{5}$, Arash Mehrjou$^{3,4}$\\
$^1$Mila, $^2$Université de Montréal, $^3$MPI-IS, $^4$ETH Zurich, $^5$KTH Stockholm\\
}

\newcommand{\norm}[1]{\left\lVert#1\right\rVert}

\begin{document}

\maketitle

\begin{abstract}
Diffusion-based generative models learn to iteratively transfer unstructured noise to a complex target distribution as opposed to Generative Adversarial Networks (GANs) or the decoder of Variational Autoencoders (VAEs) which produce samples from the target distribution in a single step. Thus, in diffusion models every sample is naturally connected to a random trajectory which is a solution to a learned stochastic differential equation (SDE). Generative models are only concerned with the final state of this trajectory that delivers samples from the desired distribution. \cite{abstreiter2021diffusion} showed that these stochastic trajectories can be seen as continuous filters that wash out information along the way. Consequently, it is reasonable to ask if there is an intermediate time step at which the preserved information is optimal for a given downstream task. In this work, we show that a combination of information content from different time steps gives a strictly better representation for the downstream task. We introduce an attention and recurrence based modules that ``learn to mix'' information content of various time-steps such that the resultant representation leads to superior performance in downstream tasks.\footnote{Open-sourced implementation is available at \href{https://github.com/sarthmit/traj_drl}{https://github.com/sarthmit/traj\_drl}}
\end{abstract}
\vspace{-4mm}
\section{Introduction}
\label{sec:introduction}
A lot of the progress in Machine Learning hinges on learning good representations of the data, whether in supervised or unsupervised fashion. Typically in the absence of label information, learning a good representation is often guided by reconstruction of the input, as is the case with autoencoders and generative models like variational autoencoders \citep{vincent2010stacked,kingma2013auto,rezende2014stochastic}; or by some notion of invariance to certain transformations like in Contrastive Learning and similar approaches \citep{chen2020simple,chen2020improved,grill2020bootstrap}. In this work, we analyze a novel way of representation learning which was introduced in \cite{abstreiter2021diffusion} with a denoising objective using diffusion based models to obtain unbounded representations.

Diffusion-based models~\citep{sohldickstein2015deep,song2020denoising,song2021scorebased,sajjadi2018tempered,niu2020permutation,cai2020learning,chen2020wavegrad,saremi2018deep,dhariwal2021diffusion,luhman2021knowledge,ho2021cascaded,mehrjou2017annealed} are generative models that leverage step-wise perturbations to the samples of the data distribution (eg. CIFAR10), modeled via a Stochastic Differential Equation (SDE), until convergence to an unstructured distribution (eg. $\mathcal{N}\left(\v{0}, \v{I}\right)$) called, in this context, the prior distribution. In contrast to this diffusion process, a ``score model" is learned to approximate the reverse process that iteratively converges to the data distribution starting from the prior distribution.
Beyond the generative modelling capacity of score-based models, we instead use the additionally encoded representations to perform inference tasks, such as classification. 

In this work, we revisit the formulation provided by \cite{abstreiter2021diffusion,preechakul2021diffusion} which augments such diffusion-based systems with an encoder for performing representation learning which can be used for downstream tasks. In particular, we look at the infinite-dimensional representation learning methodology from \cite{abstreiter2021diffusion} and perform a deeper dive into (a) the benefits of utilizing the trajectory or multiple points on it as opposed to choosing just a single point, and (b) the kind of information encoded at different points. Using trained attention mechanisms over diffusion trajectories, we ask about similarity and differences of representations across diffusion processes. Do they encode certain interpretable features at different points, or is it redundant to look at the whole trajectory?

Our findings can be summarized as follows:
\begin{itemize}
    \item We propose using the trajectory-based representation combined with sequential architectures like Recurrent Neural Networks (RNNs) and Transformers to perform downstream predictions using multiple points as it leads to better performance than just finding one-best point on the trajectory for downstream predictions \citep{abstreiter2021diffusion}.
    \item We analyze the representations obtained at different parts of the trajectory through Mutual Information and Attention-based relevance to downstream tasks to showcase the differences in information contained along the trajectory.
    \item We also provide insights into the benefits of using more points on the trajectory, with saturating benefits as our discretization becomes finer. We further show that finer discretizations lead to even more performance benefits when the latent space is severely restricted, eg. just a 2-dimensional output from the encoder.
\end{itemize}
\begin{figure}
    \centering
    \includegraphics[width=\linewidth]{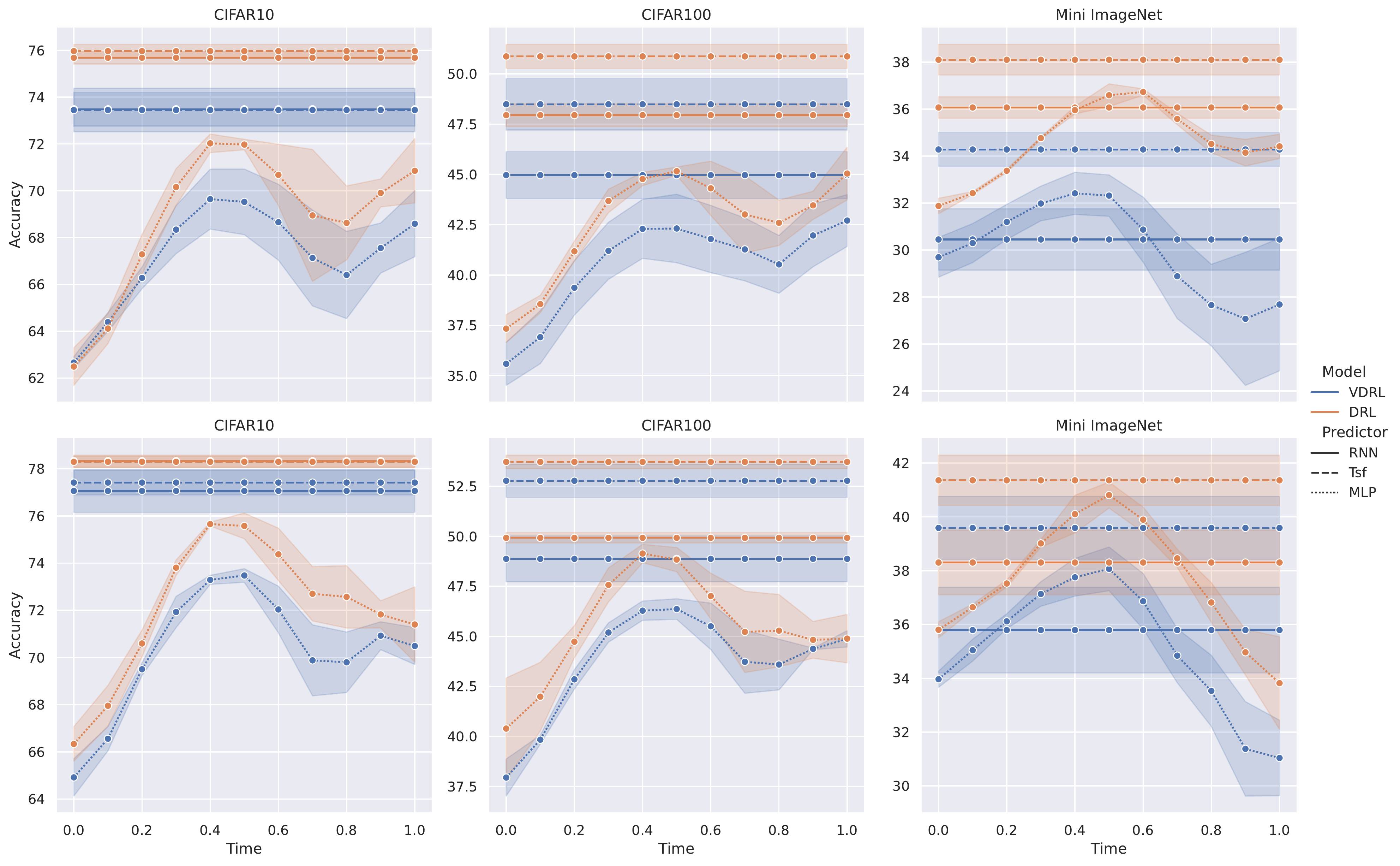}
    \caption{Downstream performance of single point based representations (MLP) and full trajectory based representations (RNN and Tsf) on different datasets for both types of learned encoders: probabilistic (VDRL) and deterministic (DRL) using a 64-dimensional latent space (\textit{Top}) and a 128-dimensional latent space (\textit{Bottom}).}
    \label{fig:perf}
    \vspace{-5mm}
\end{figure}

\section{Beyond Fixed Representations}
We first outline how diffusion-based representation learning systems are trained.
Given some example $\v{x}_0 \in \mathbb{R}^d$ which is sampled from the target distribution $p_0$, the diffusion process constructs the trajectory $(\v{x}_t)_{t \in [0,1]}$ through the application of an SDE. In this work, we consider the Variance Exploding SDE~\citep{song2021scorebased} for this diffusion process, defined as
\begin{align}
    d\v{x} &= f(\v{x}, t) + g(t)d\v{w} := \sqrt{\frac{d[\sigma^2(t)]}{dt}}d\v{w}
\end{align}
where $\v{w}$ is the standard Wiener process and $\sigma^2(\cdot)$ the noise variance of the diffusion process. This leads to a closed form distribution of $\v{x}_t$ conditional on $\v{x}_0$ as $p_{0t}(\v{x}_t | \v{x}_0) = \mathcal{N}(\v{x}_t; \v{x}_0, [\sigma^2(t) - \sigma^2(0)]\v{I})$. Given this diffusion process modeled through the Variance Exploding SDE, the reverse SDE takes a similar form but requires the knowledge about the score function, i.e. $\nabla_{\v{x}} \log p_t(\v{x})$ for all $t \in [0,1]$. A common way to obtain this score function is through the Explicit Score Matching~\citep{hyvarinen2005estimation} objective,
\begin{align}
    \mathbb{E}_{\v{x}_t}\left[\norm{s_\theta(\v{x}_t, t) - \nabla_{\v{x}_t} \log p_t(\v{x}_t)}^2\right]
\end{align}
which suffers from just one hiccup, which is that data about the ground-truth score function is not available. To solve this problem, Denoising Score Matching~\citep{vincent} was proposed, 
\begin{align}
    \mathbb{E}_{\v{x}_0} \left[ \mathbb{E}_{\v{x}_t | \v{x}_0} \left[ \norm{s_\theta(\v{x}_t, t) - \nabla_{\v{x}_t} \log p_{0t}(\v{x}_t | \v{x}_0)}^2 \right]\right]
\end{align}
where the term $\log p_{0t}(\v{x}_t | \v{x}_0)$ is available due to its closed-form structure.
Given that the above objective cannot be reduced to 0, \cite{abstreiter2021diffusion} proposes the objective
\begin{align}
\label{eq:obj}
    \mathbb{E}_{\v{x}_0} \left[ \mathbb{E}_{\v{x}_t | \v{x}_0} \left[ \norm{s_\theta(\v{x}_t, E_\phi(\v{x}_0, t), t) - \nabla_{\v{x}_t} \log p_{0t}(\v{x}_t | \v{x}_0)}^2 \right]\right]
\end{align}
where the additional input $E_\phi(\v{x}_0, t)$ to the score function is obtained from a learned encoder. It provides information about the unperturbed sample that might be useful for denoising data at time step $t$ in the diffusion process. Training this system can lead to the objective being reduced to 0, thereby providing incentive to the encoder $E_\phi(\cdot, t)$ to learn meaningful representations for each time $t$. From this, we obtain a trajectory-based representation $(E_\phi(\v{x}_0, t))_{t \in [0,1]}$ for each sample $\v{x}_0$, as opposed to finite sized representations obtained from typical Autoencoder~\citep{bengio2013representation,DBLP:journals/corr/VinyalsBLKW16,kingma2013auto,rezende2014stochastic} and Contrastive Learning~\citep{DBLP:journals/corr/abs-2006-10029,grill2020bootstrap,caron2021unsupervised,articlesiamese,chen2020exploring} approaches.

Following the setup in~\cite{abstreiter2021diffusion}, we consider two different versions of the encoder $E_\phi(\cdot, \cdot)$, (a) the VDRL setup, where the output of $E_\phi(\cdot, \cdot)$ represents a distribution from which a sample is used, and the distribution is regularized using a KL-Divergence term with the standard Normal distribution $\mathcal{N}(0, \text{I})$, and (b) the DRL setup, where the output  of the encoder is deterministic and regularized using an $L_1$ distance metric to be as close to $0$ as possible. Typically in all our experiments, we see that not only the trends hold with multiple seeds but also across these two types of encoders, substantiating the statistical significance of the trends.

\begin{figure}
\begin{subfigure}[c]{0.32\columnwidth}
  \centering
  \includegraphics[width=\columnwidth]{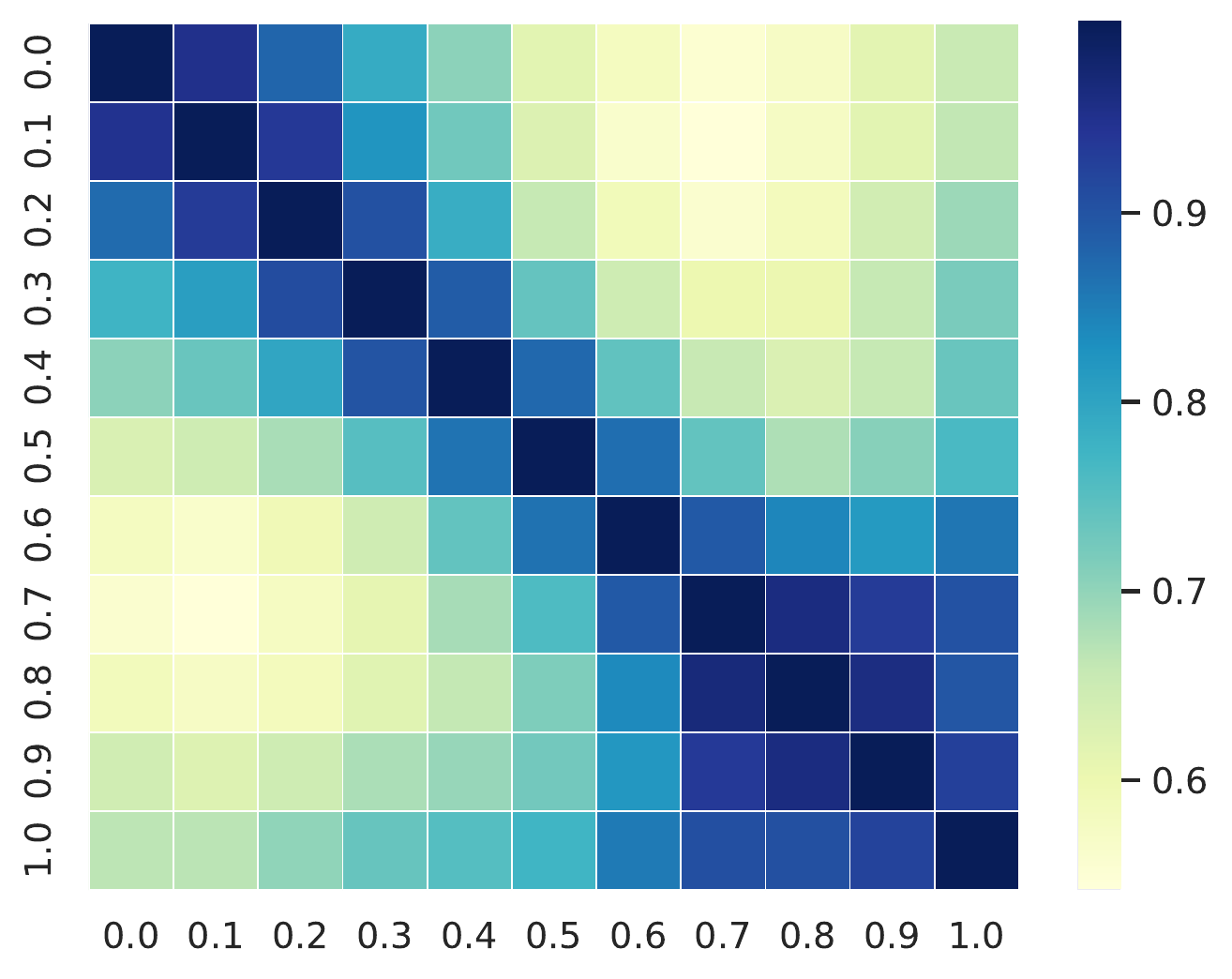}  \vspace{-6mm}
  \subcaption{\scriptsize VDRL $|$ CIFAR10}
\end{subfigure}
\begin{subfigure}[c]{0.32\columnwidth}
  \centering
  \includegraphics[width=\columnwidth]{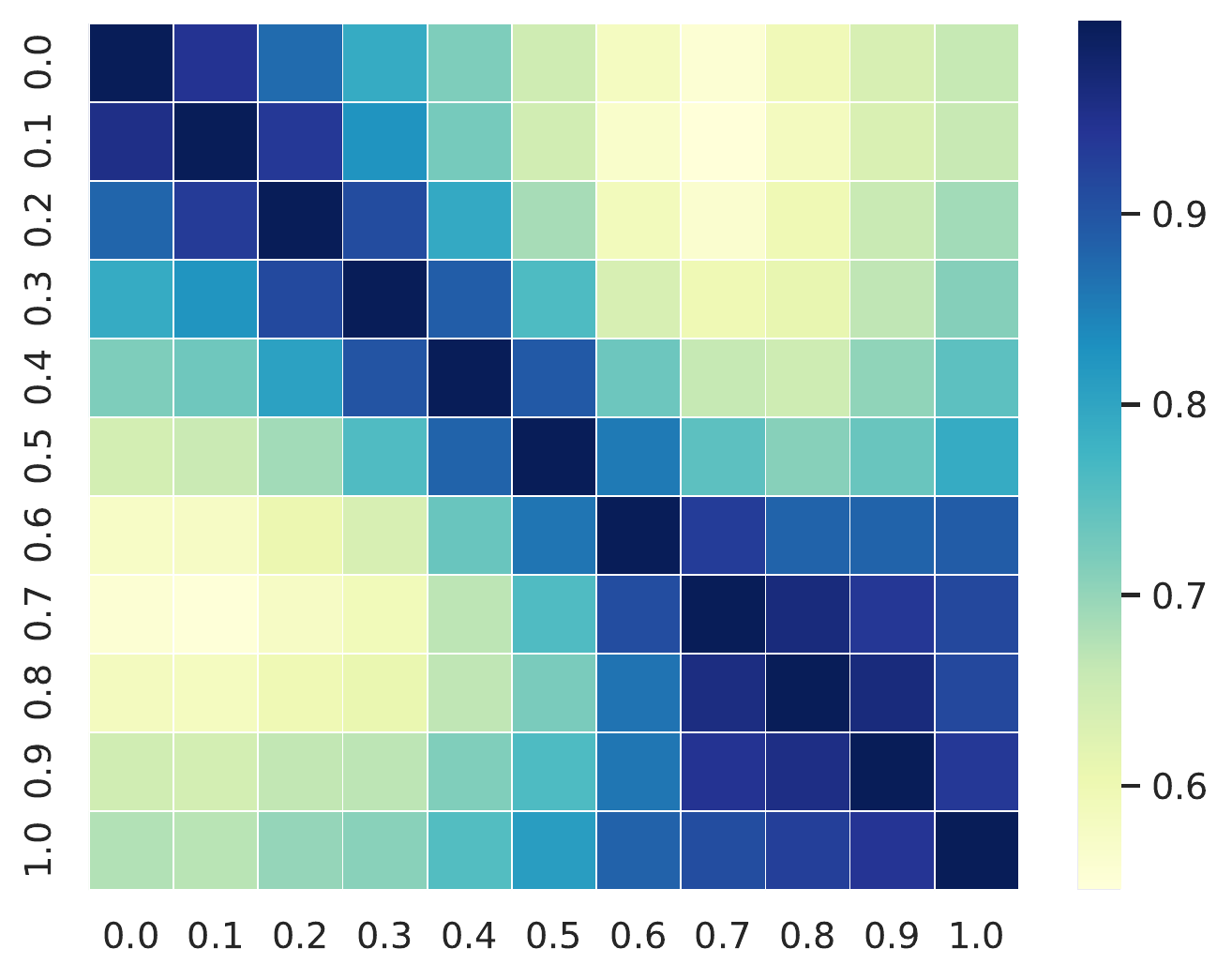}  
 \vspace{-6mm}
  \subcaption{\scriptsize VDRL $|$ CIFAR100}
\end{subfigure}
\begin{subfigure}[c]{0.32\columnwidth}
  \centering
  \includegraphics[width=\columnwidth]{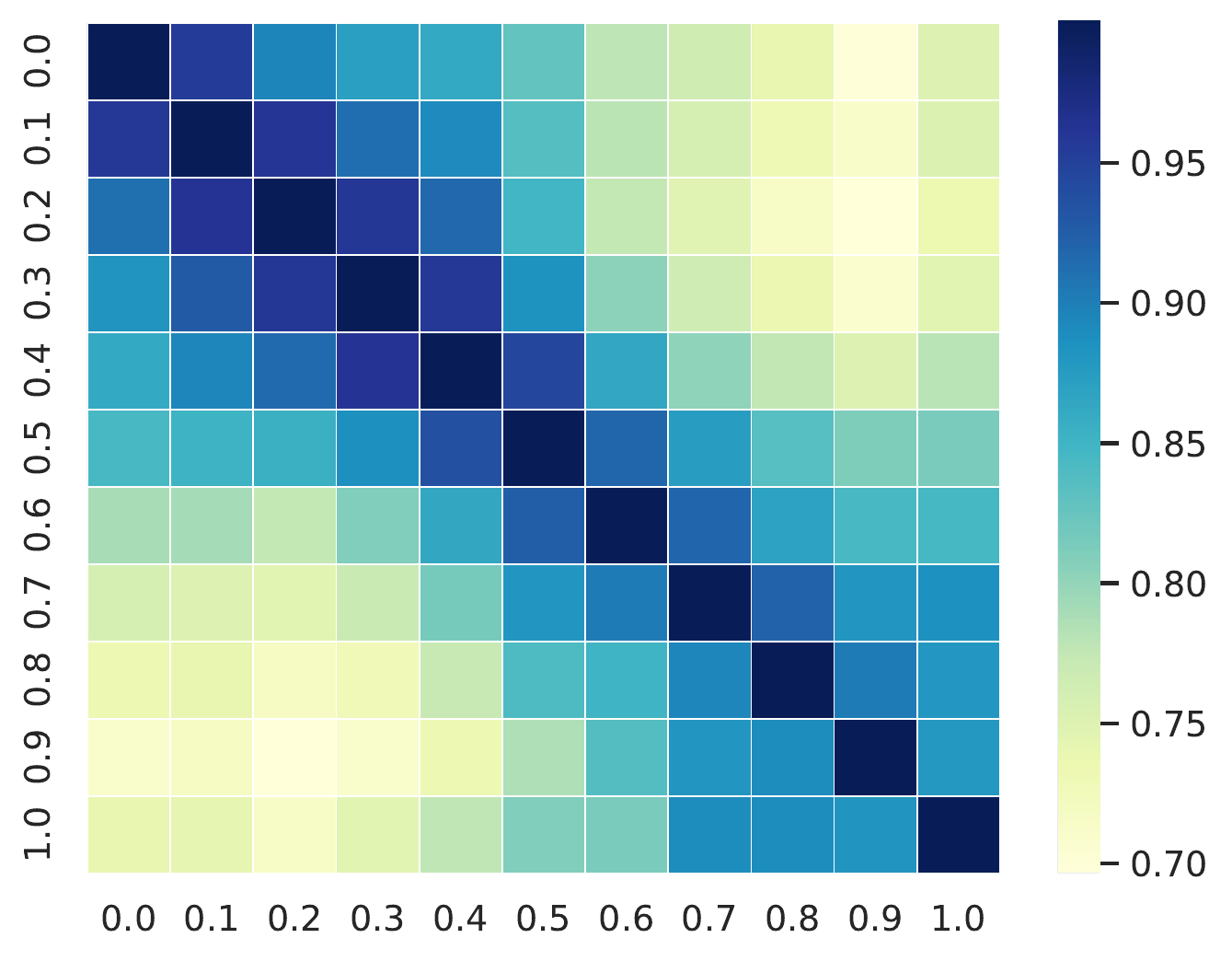}
 \vspace{-6mm}
  \subcaption{\scriptsize VDRL $|$ Mini-ImageNet}
\end{subfigure} \\
\begin{subfigure}[c]{0.32\columnwidth}
  \centering
  \includegraphics[width=\columnwidth]{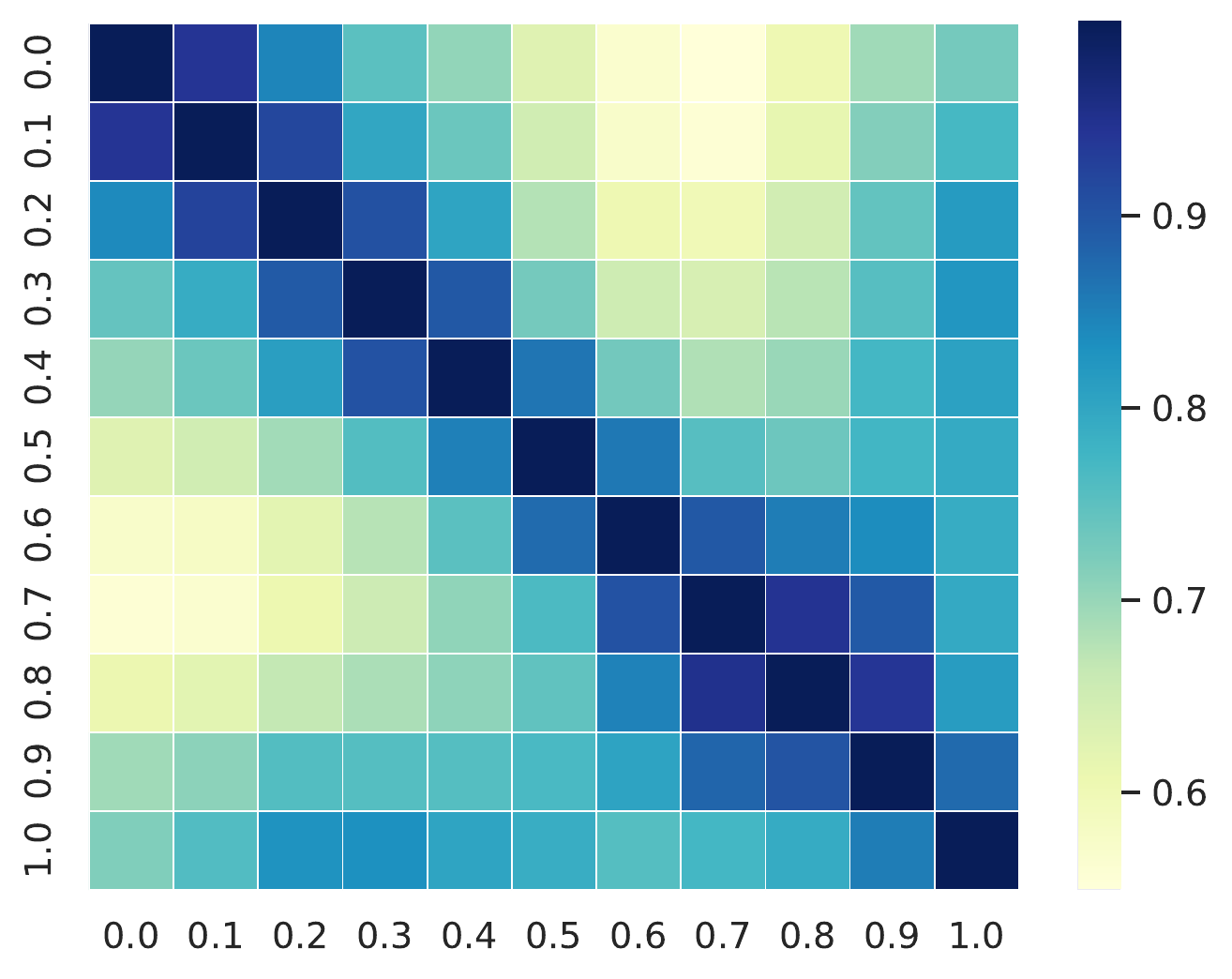}
 \vspace{-6mm}
  \subcaption{\scriptsize DRL $|$ CIFAR10}
\end{subfigure}
\begin{subfigure}[c]{0.32\columnwidth}
  \centering
  \includegraphics[width=\columnwidth]{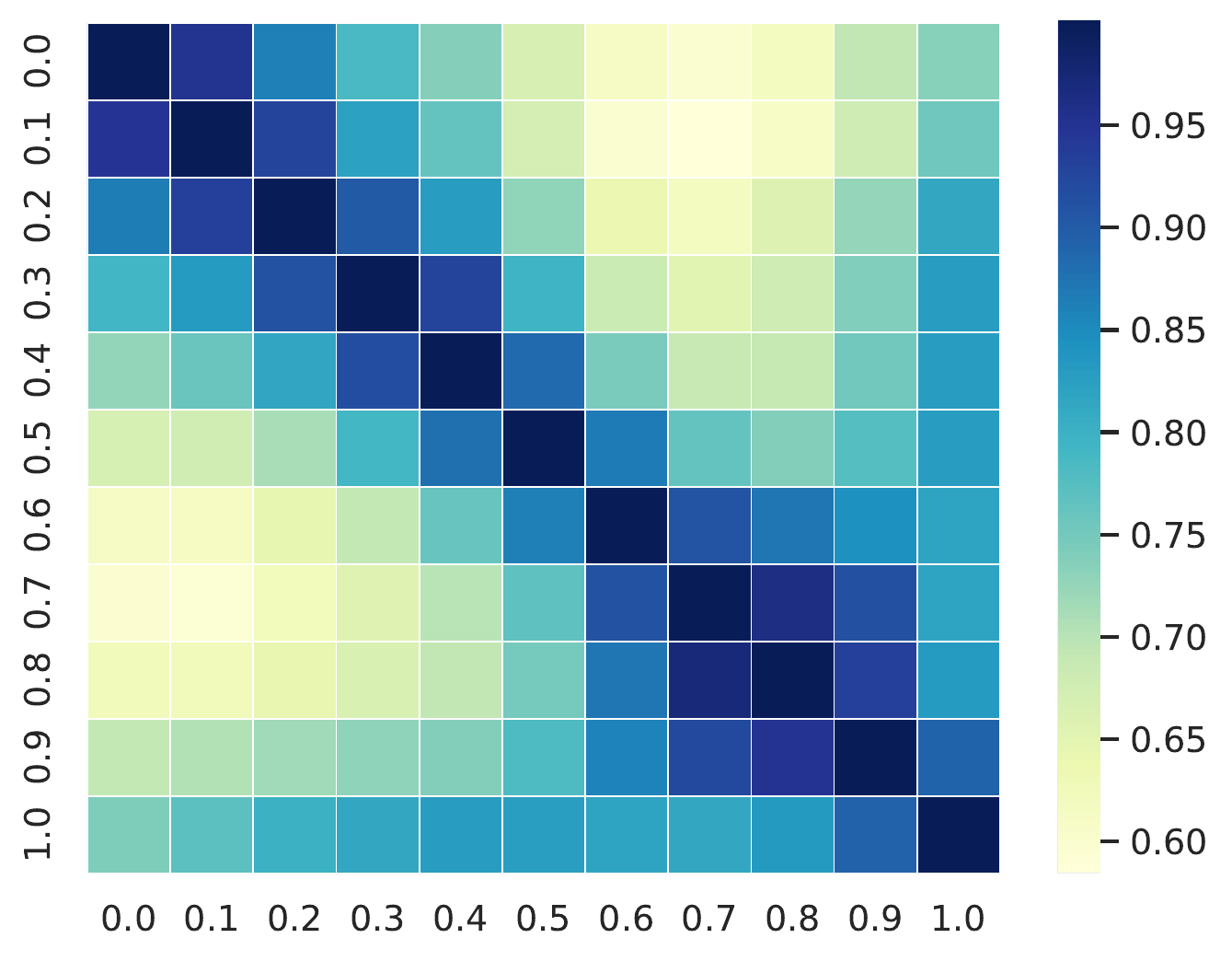}
   \vspace{-6mm}
    \subcaption{\scriptsize DRL $|$ CIFAR100}
\end{subfigure}
\begin{subfigure}[c]{0.32\columnwidth}
  \centering
  \includegraphics[width=\columnwidth]{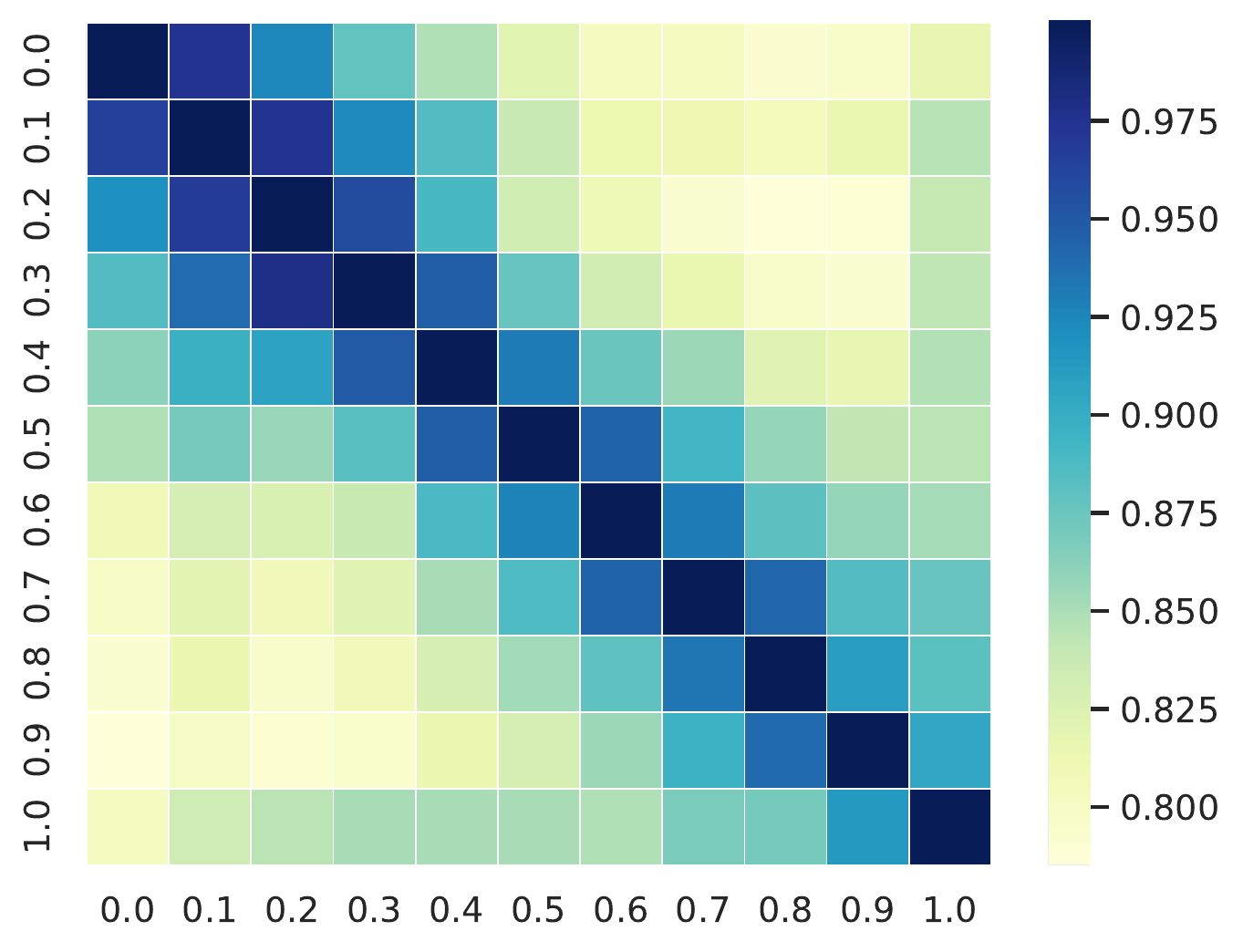}
 \vspace{-6mm}
  \subcaption{\scriptsize DRL $|$ Mini-ImageNet}
\end{subfigure}
\caption{Normalized Mutual Information between different points on the trajectory. Cell ($i,j$) demonstrates the normalized mutual information, estimated with the MINE algorithm, between the representations at time $t=i$ and $t=j$.}
\label{fig:nmi-mult}
\vspace{-5mm}
\end{figure}
It is important to note that our goal here is strictly representation learning, and thus we use the representations obtained for downstream (multitask-) image classification. This should not be confused with generative modelling as the provided mechanism augments a generative model for representation learning, but is not a generative model on its own. Since this representation learning paradigm can be augmented with a time-conditioned encoder model, this leads to a natural extension to trajectory-based (unbounded) representation, in contrast to typical bounded representation learning models like Autoencoders. Thus, this representation learning paradigm constructs a functional map from the input space to a curve / trajectory in $\mathbb{R}^d$, where we refer to $d$ as the dimensionality of this encoded space.

\vspace{-2mm}
\subsection{Infinite-dimensional representation of finite-dimensional data}
\vspace{-1mm}
Normally in autoencoders or other \emph{static} representation learning methods, the input data $\v{x}_0\in \mathbb{R}^d$ is mapped to a single point $\v{z}\in \mathbb{R}^c$ in the code space. However, our proposed algorithm learns a richer representation where the input $\v{x}_0$ is mapped to a curve in $\mathbb{R}^c$ instead of a single point through the encoder $E_\phi(\cdot, t)$. Hence, the learned code is produced by the map $\v{x}_0\to (E_\phi(\v{x}_0, t))_{t\in[0, 1]}$ where the infinite-dimensional object $(E_\phi(\v{x}_0, t))_{t\in[0, 1]}$ is the encoding for $\v{x}_0$. 

The learned code is at least as good as static codes in terms of separation induced among the codes. Consider two input samples $\v{x}_0$ and $\v{x}'_0$, hence we have: 
\begin{equation}
    \lVert E_\phi(\v{x}_0, 0) - E_\phi(\v{x}'_0, 0) \rVert \leq \sup_{t\in [0,1]} \lVert E_\phi(\v{x}_0, t) - E_\phi(\v{x}'_0, t) \rVert
\end{equation}
which implies that the downstream task can at least recover the separation provided by finite-dimensional codes from the infinite-dimensional code by looking for the maximum separation along the representation trajectory.

A downstream task can leverage this rich encoding in various ways. Consider the classification task where we want to find a mapping $f:\mathbb{R}^d\to \{0, 1\}$ from input data to the label space. Instead of giving $\v{x}_0$ as the input to $f$, we define $f:\mathcal{H}\to \{0, 1\}$ where the input to the classifier is the whole trajectory $(E_\phi(\v{x}_0, t))_{t\in[0, 1]}$.
Thus, the classifier can now use RNN and Transformer models to make use of the information content of the entire trajectories.
\begin{figure}
\begin{subfigure}[c]{0.32\columnwidth}
  \centering
  \includegraphics[width=\columnwidth]{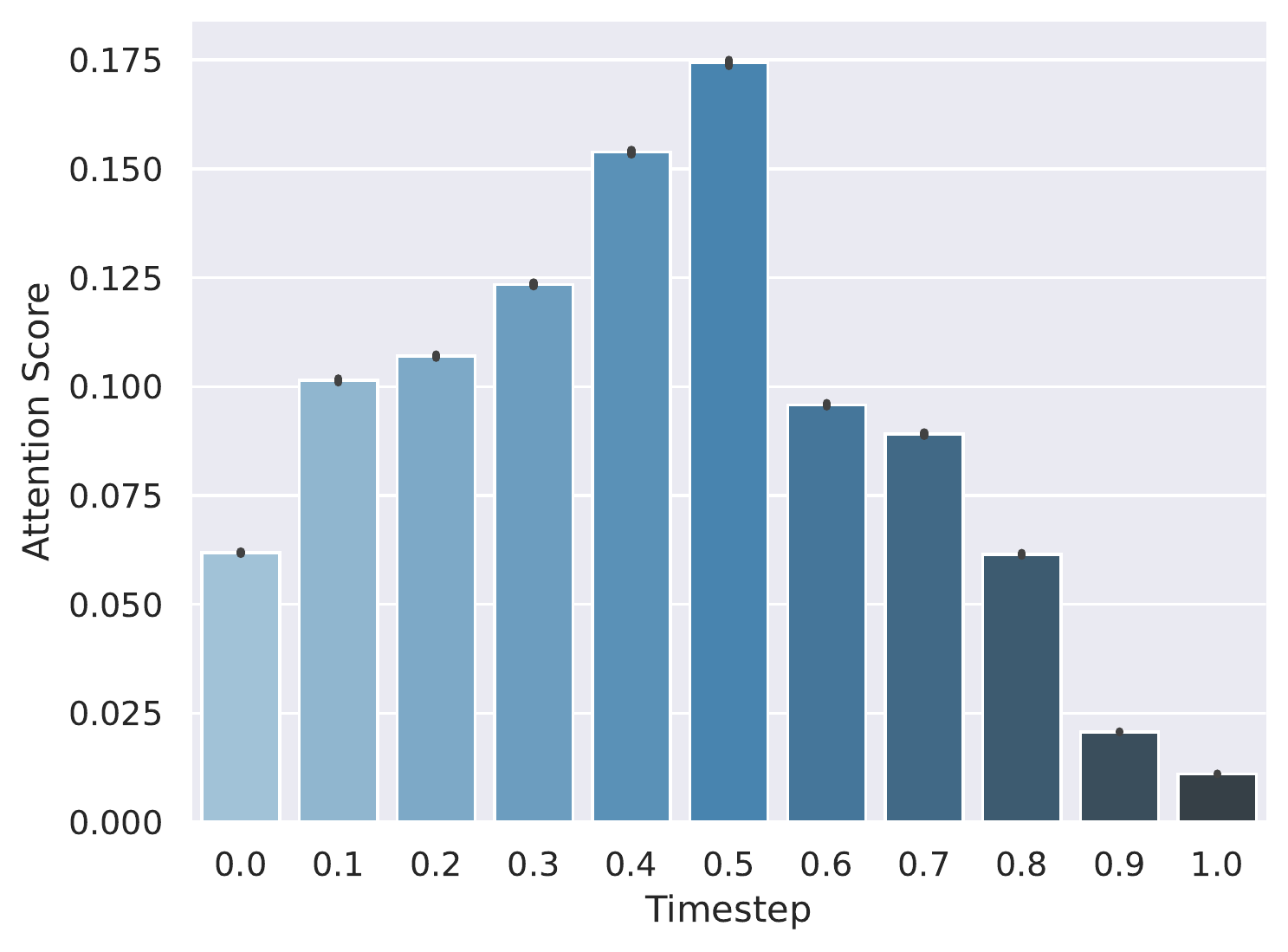}
  \vspace{-6mm}
  \subcaption{\scriptsize VDRL $|$ CIFAR10}
\end{subfigure}
\begin{subfigure}[c]{0.32\columnwidth}
  \centering
  \includegraphics[width=\columnwidth]{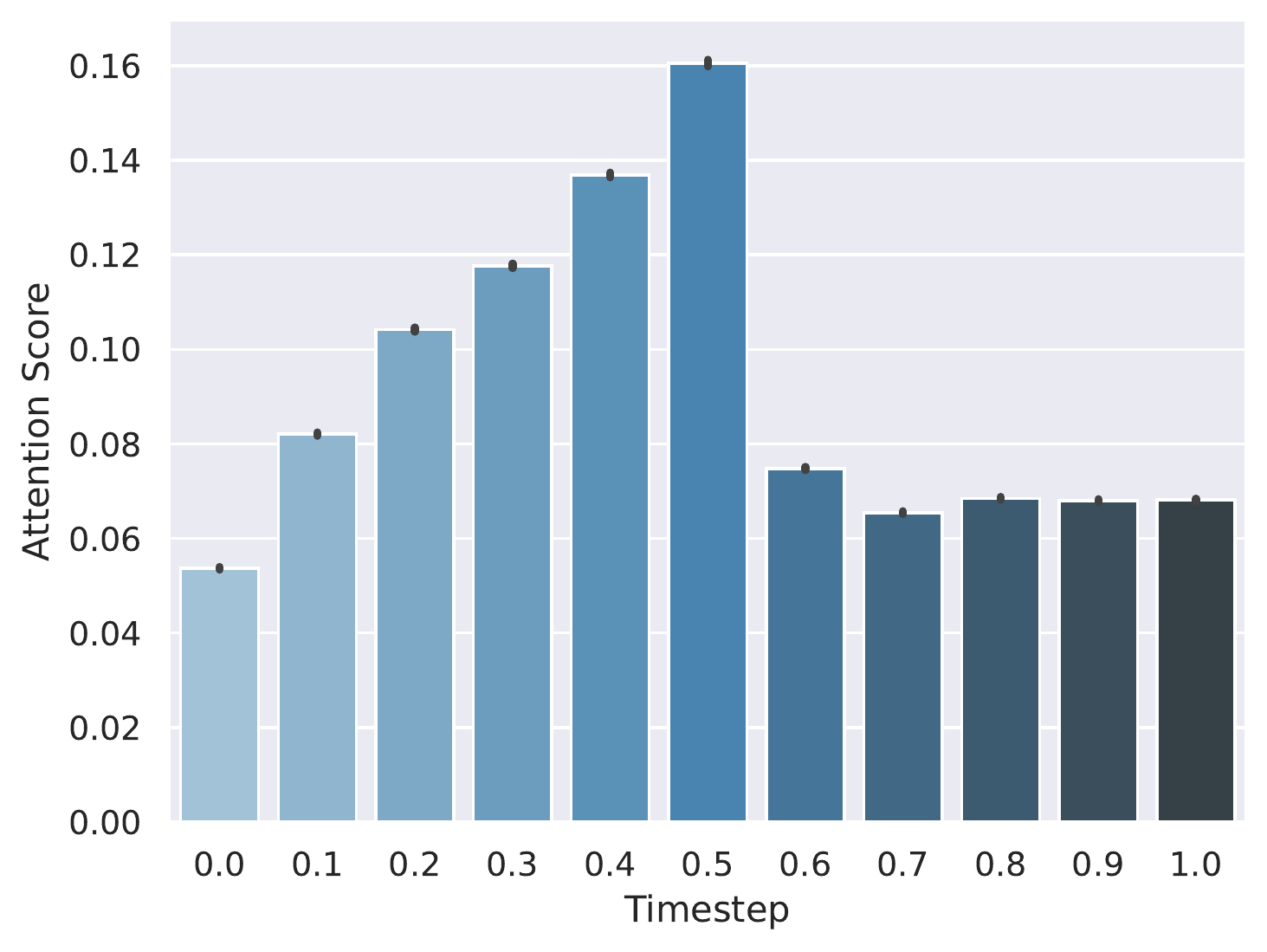}
  \vspace{-6mm}
  \subcaption{\scriptsize VDRL $|$ CIFAR100}
\end{subfigure}
\begin{subfigure}[c]{0.32\columnwidth}
  \centering
  \includegraphics[width=\columnwidth]{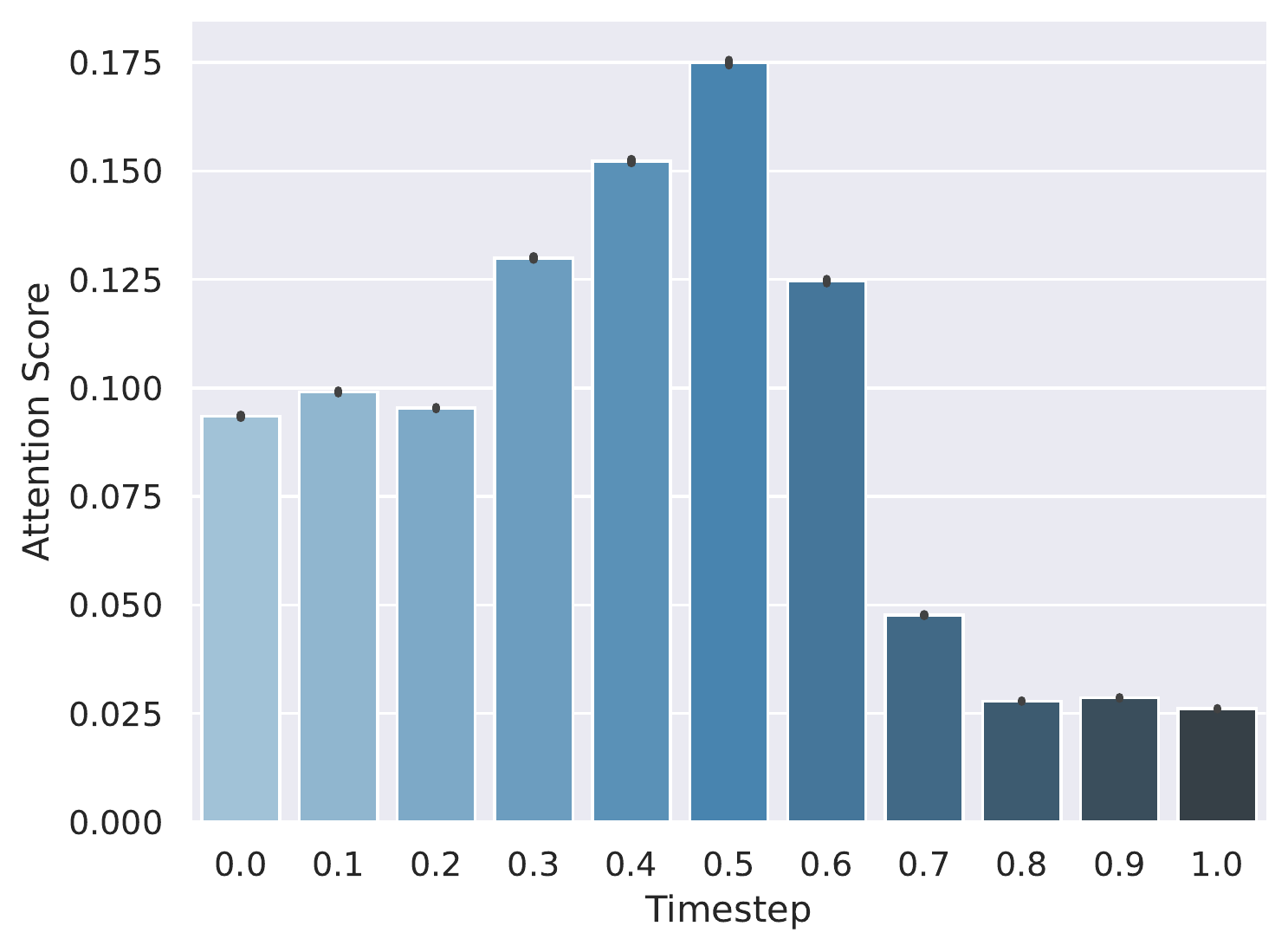}
  \vspace{-6mm}
  \subcaption{\scriptsize VDRL $|$ Mini-ImageNet}
\end{subfigure} \\
\begin{subfigure}[c]{0.32\columnwidth}
  \centering
  \includegraphics[width=\columnwidth]{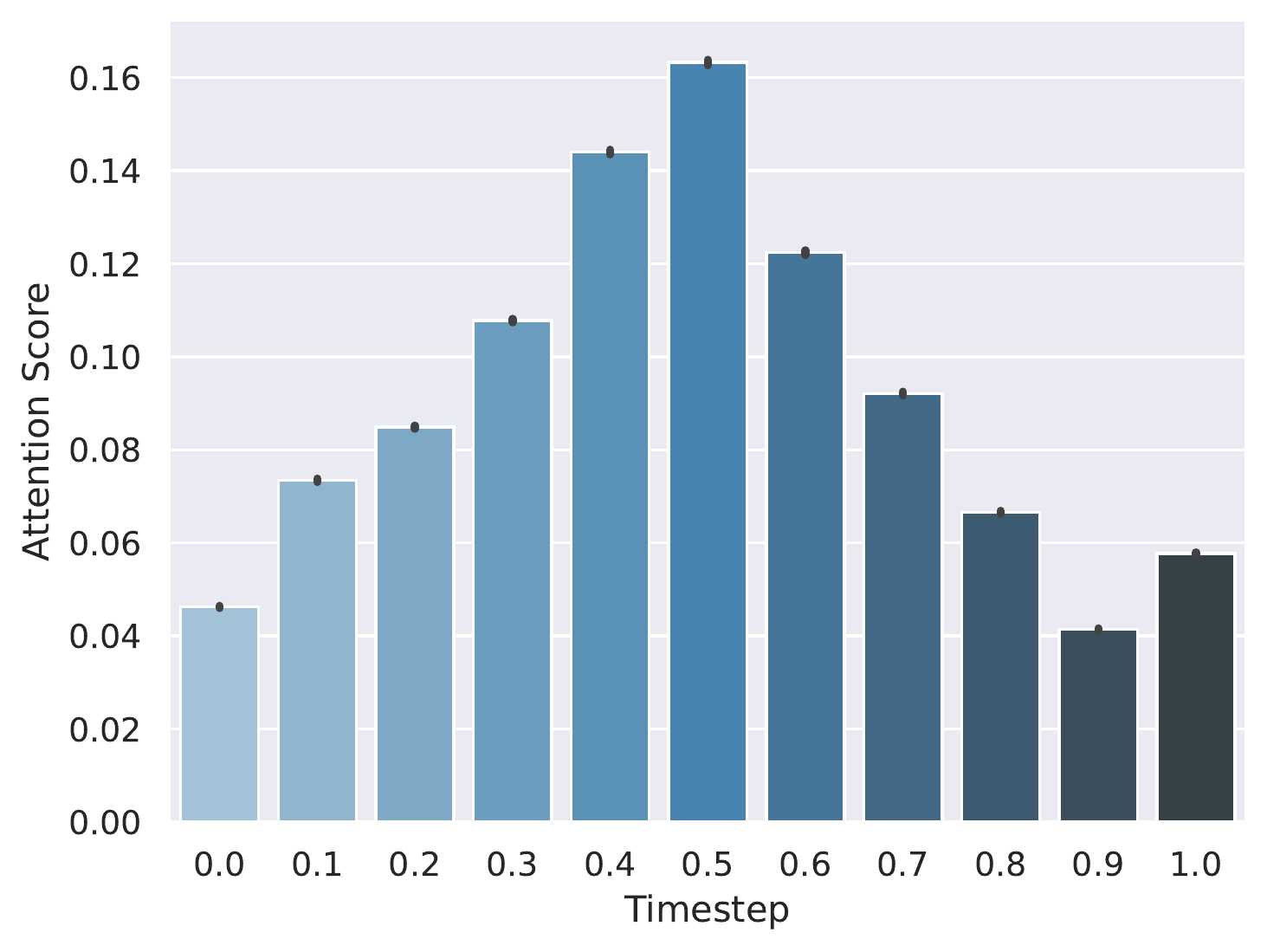}
  \vspace{-6mm}
  \subcaption{\scriptsize DRL $|$ CIFAR10}
\end{subfigure}
\begin{subfigure}[c]{0.32\columnwidth}
  \centering
  \includegraphics[width=\columnwidth]{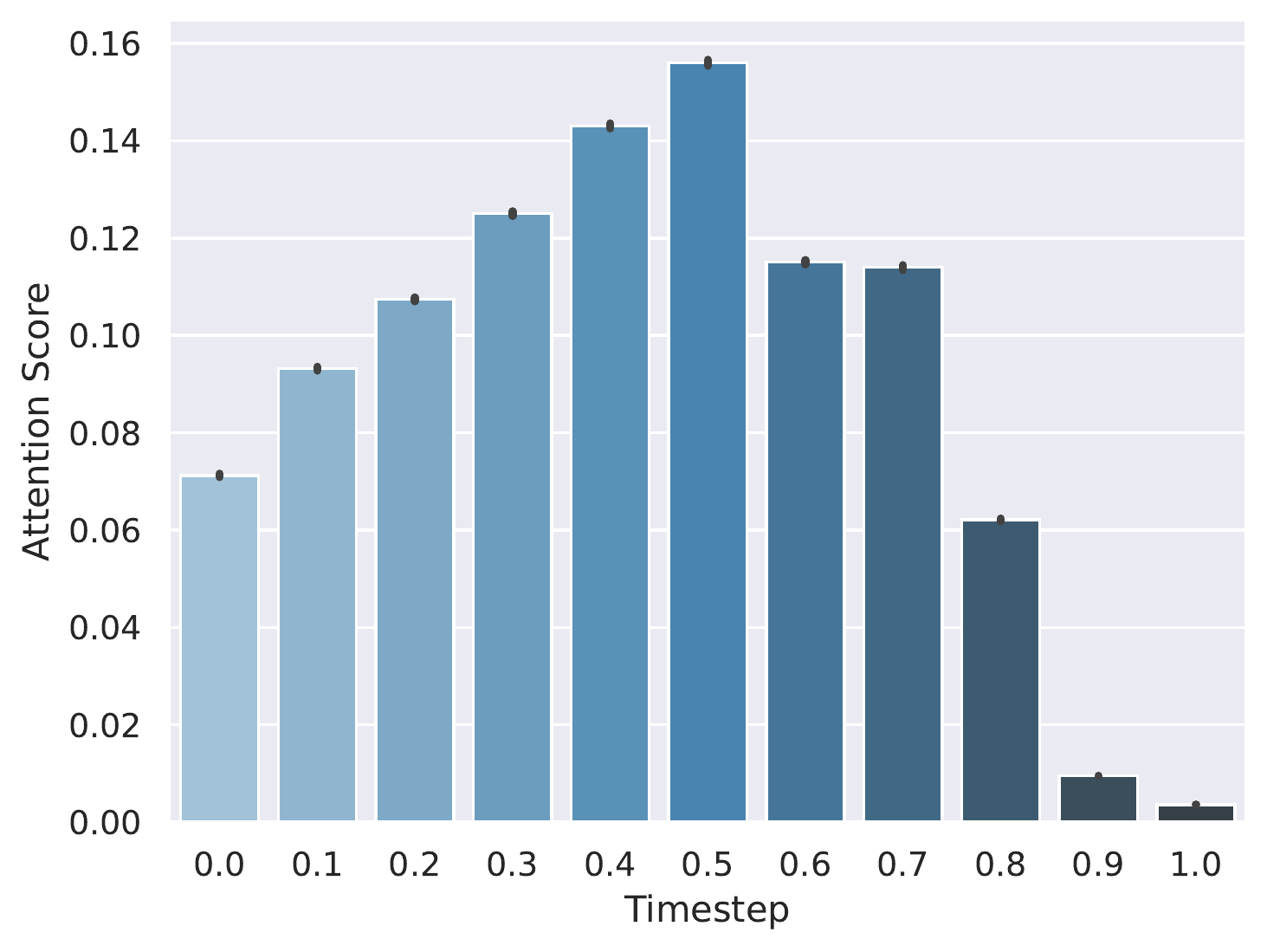}
  \vspace{-6mm}
  \subcaption{\scriptsize DRL $|$ CIFAR100}
\end{subfigure}
\begin{subfigure}[c]{0.32\columnwidth}
  \centering
  \includegraphics[width=\columnwidth]{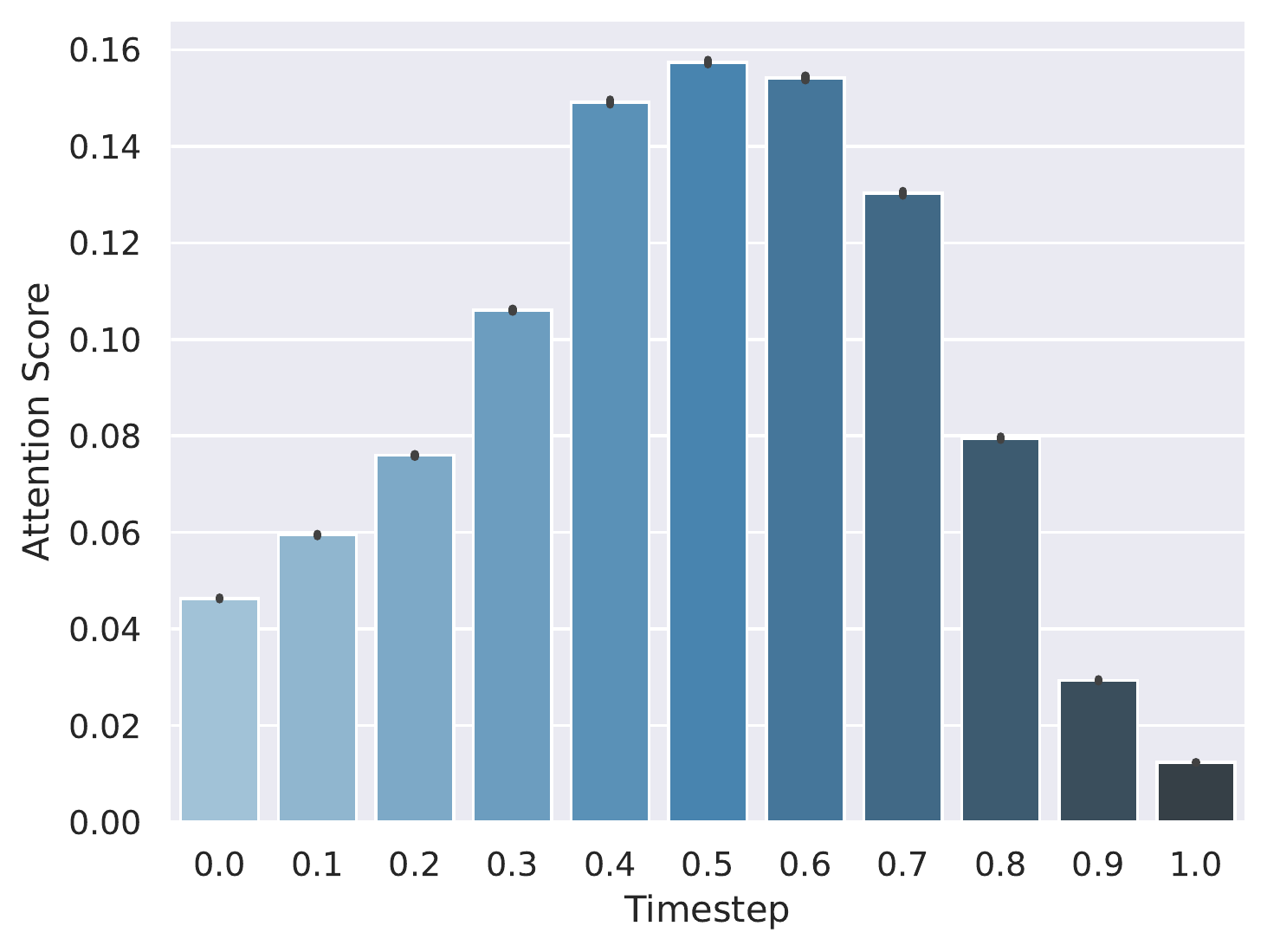}
  \vspace{-6mm}
  \subcaption{\scriptsize DRL $|$ Mini-ImageNet}
\end{subfigure}
\caption{Attention Scores provided to different points on the trajectories, which are obtained from diffusion based representation learning systems with probabilistic encoders (VDRL; top row) and with deterministic encoders (DRL; bottom row) across the following datasets (i) Left: CIFAR10, (ii) Middle: CIFAR100, and (iii) Right: Mini-Imagenet.}
\label{fig:att-ablate}
\vspace{-5mm}
\end{figure}

\vspace{-2mm}
\section{Experiments}
\vspace{-1mm}
We first train two kinds of diffusion-based generative model as outlined in~\cite{abstreiter2021diffusion}, based on probabilistic (VDRL) and deterministic (DRL) encoders respectively. After training, the encoder model is kept fixed. For all our downstream experiments, we use this trained encoder to obtain the trajectory based representation for each of the samples. While the trajectories lie in a continuous domain $[0,1]$, we sample it at regular intervals with length $0.1$, unless specified otherwise. This leads to a discretization of the trajectory, which is then used for various analysis as outlined below. Further, we consider the dimensionality of the latent space, that is, the output of the encoder, as 128 unless otherwise specified. Additional details about the architectures used, the optimization strategy and other implementation details can be found in Appendix \ref{apdx:impl}.

\vspace{-2mm}
\subsection{Downstream Performance Reveals Benefits of Trajectory Information}
\label{sec:downstream}
\vspace{-1mm}
To understand the benefits of utilizing the trajectory-based representations, we train standard Multi-Layer Perceptron (MLP) models at different points on the trajectory and compare it with Recurrent Neural Network (RNN)~\citep{hochreiter1997long,cho2014properties} and Transformer~\citep{vaswani2017attention} based models that are able to aggregate information from different parts of the trajectory.

We evaluate the MLP, RNN and Transformer based downstream models on diffusion systems with both probabilistic encoders (VDRL) and also non-probabilistic ones (DRL). In Figure~\ref{fig:perf}, we see the performance of these different setups for the following datasets: CIFAR10~\citep{cifar10}, CIFAR100~\citep{cifar100} and Mini-ImageNet~\citep{DBLP:journals/corr/VinyalsBLKW16}. Note that in contrast to MLP implementations, RNN and Transformer use the entire trajectory and the obtained performance is plotted across all time points for visual comparison. We typically see that RNN and Transformer based models perform better than even the peaks obtained by the MLP systems. This shows that there is no single point on the trajectory that encapsulates all the information necessary for optimal classification, and thus utilizing the whole trajectory as opposed to individual points leads to improvements in performance.

We further do this performance analysis for different dimensionality of the latent spaces, that is, when the trajectory representation is embedded in a 64-dimensional Euclidean space (Figure \ref{fig:perf}: Top) and when it is emebdded in a 128-dimensional Euclidean space (Figure \ref{fig:perf}: Bottom). We see similar trends across the two settings, thus highlighting consistent benefits when using a discretization of the whole trajectory.

\begin{figure}
\centering
\begin{subfigure}[c]{0.24\linewidth}
    \includegraphics[width=\linewidth]{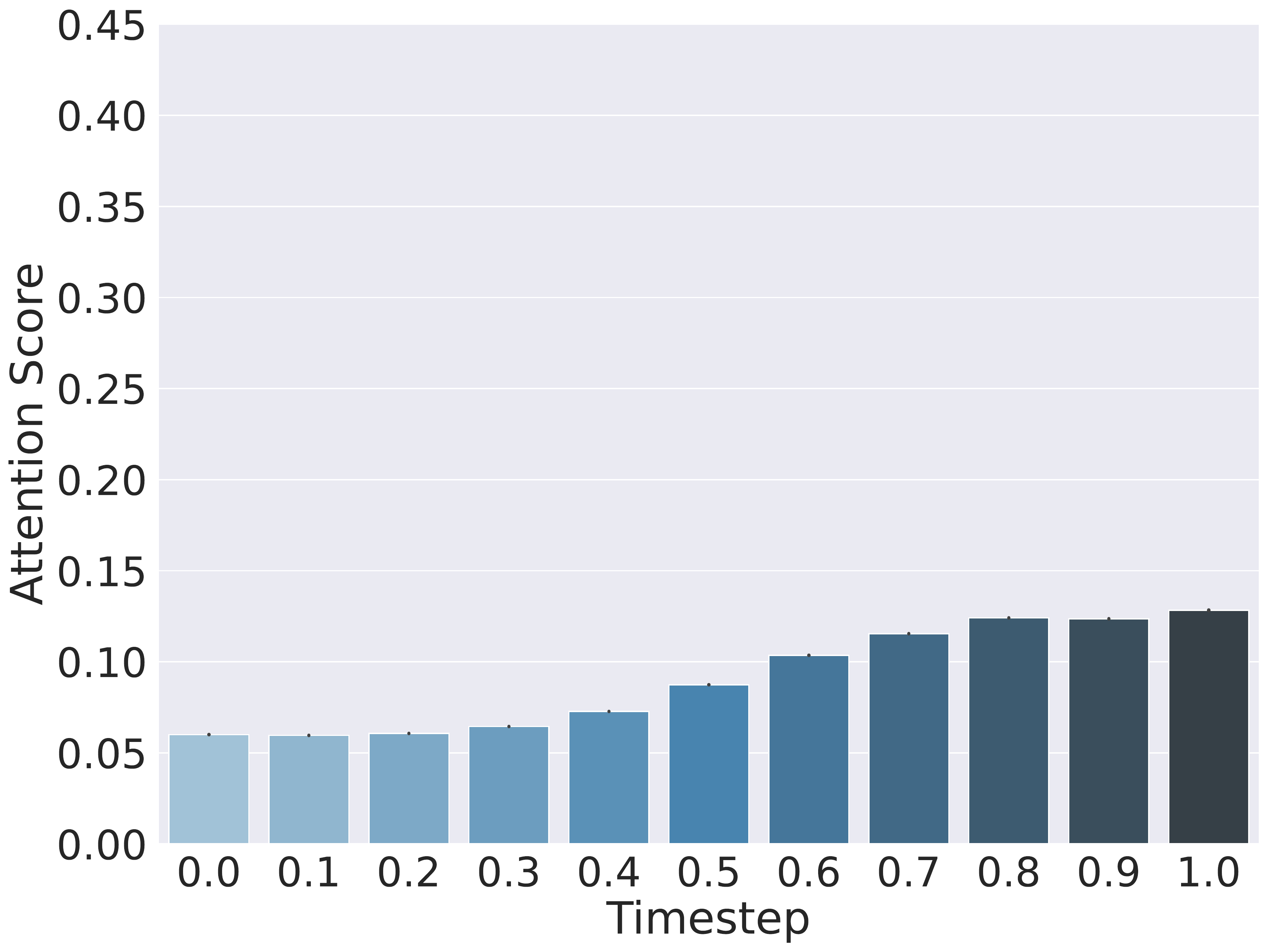}
    \vspace{-6mm}
    \subcaption{\scriptsize VDRL $|$ Background Color}
\end{subfigure}
\begin{subfigure}[c]{0.24\linewidth}
    \includegraphics[width=\linewidth]{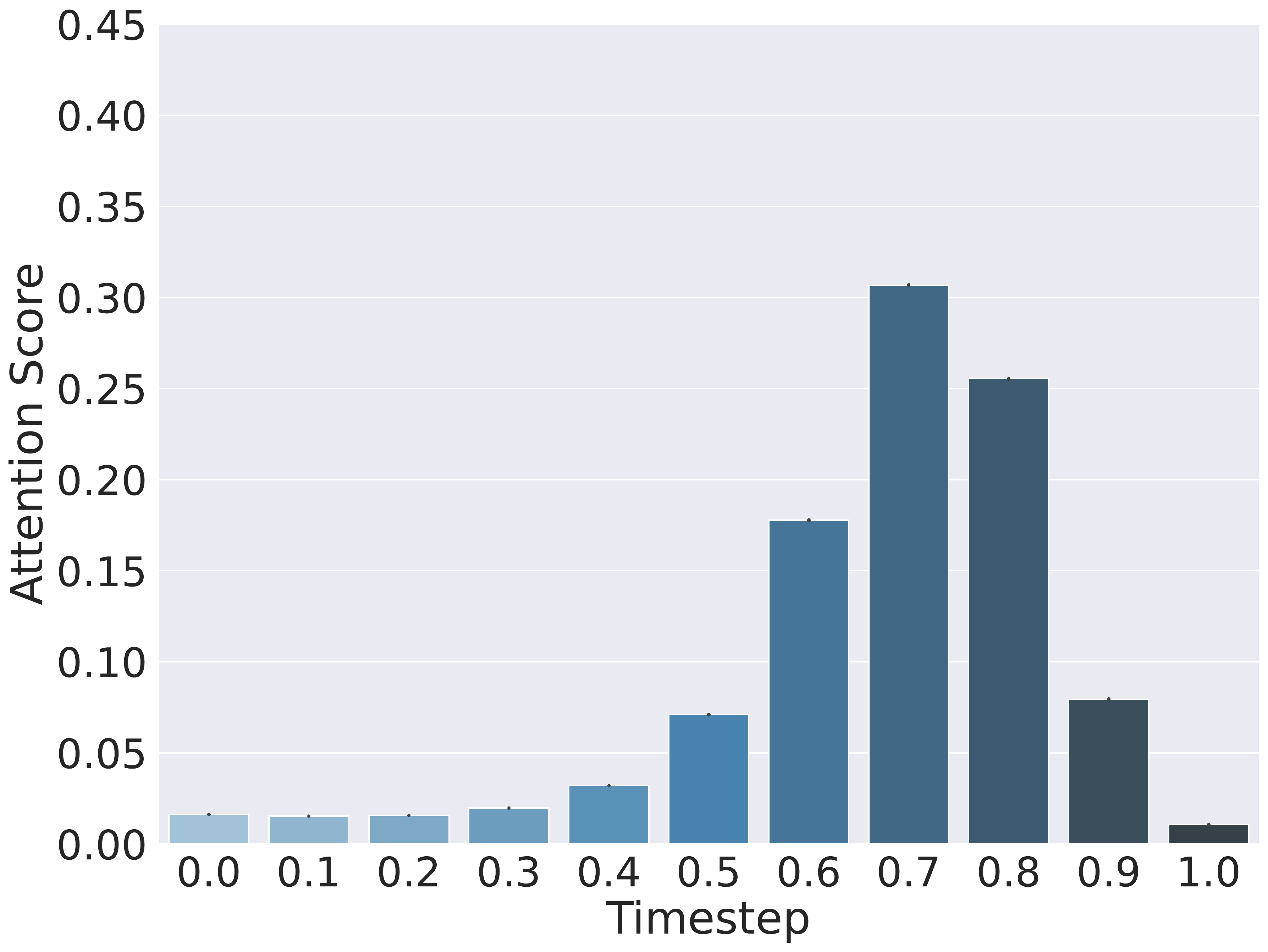}
    \vspace{-6mm}
    \subcaption{\scriptsize VDRL $|$ Foreground Color}
\end{subfigure}
\begin{subfigure}[c]{0.24\linewidth}
    \includegraphics[width=\linewidth]{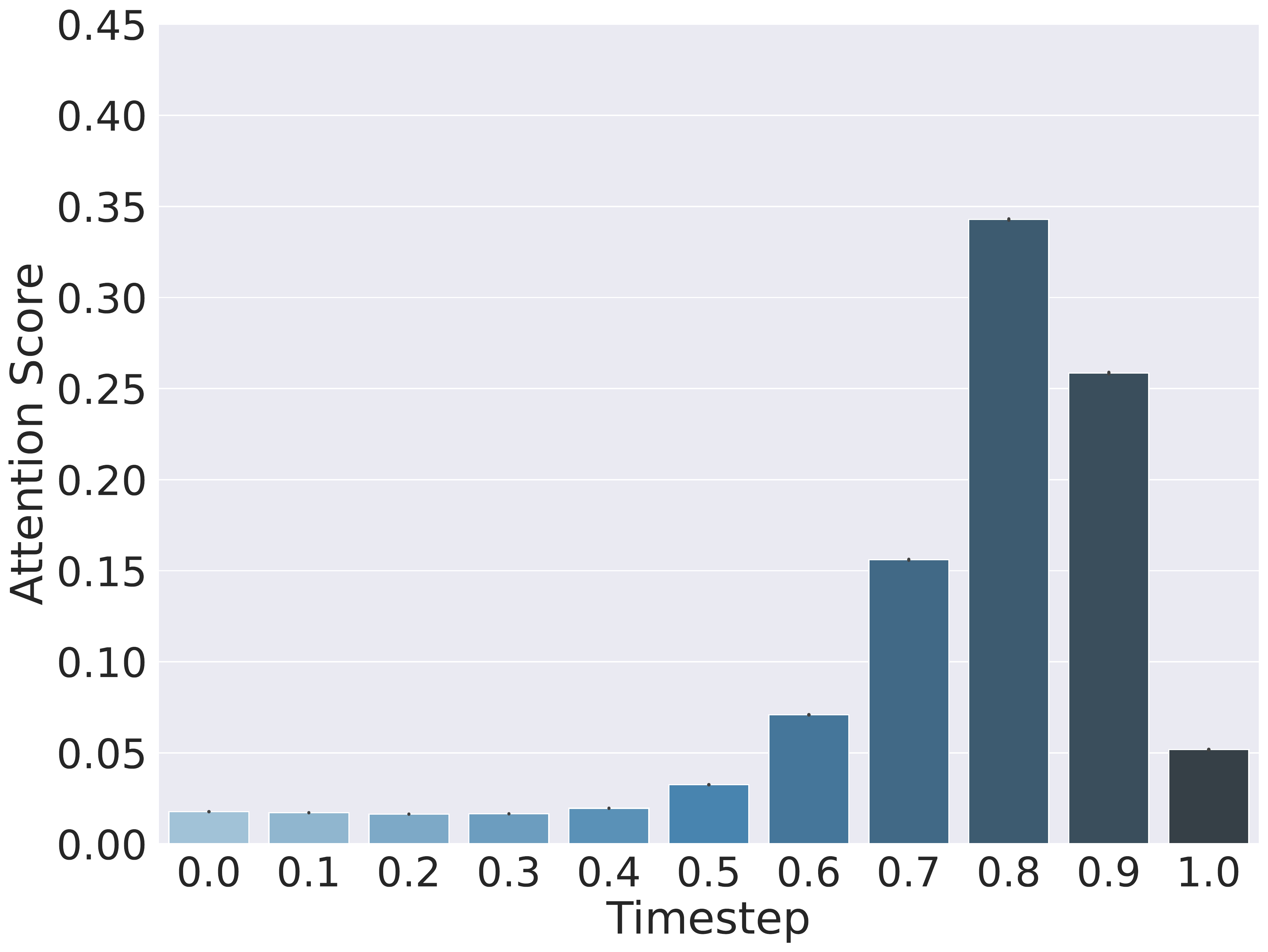}
    \vspace{-6mm}
    \subcaption{\scriptsize VDRL $|$ Location}
\end{subfigure}
\begin{subfigure}[c]{0.24\linewidth}
    \includegraphics[width=\linewidth]{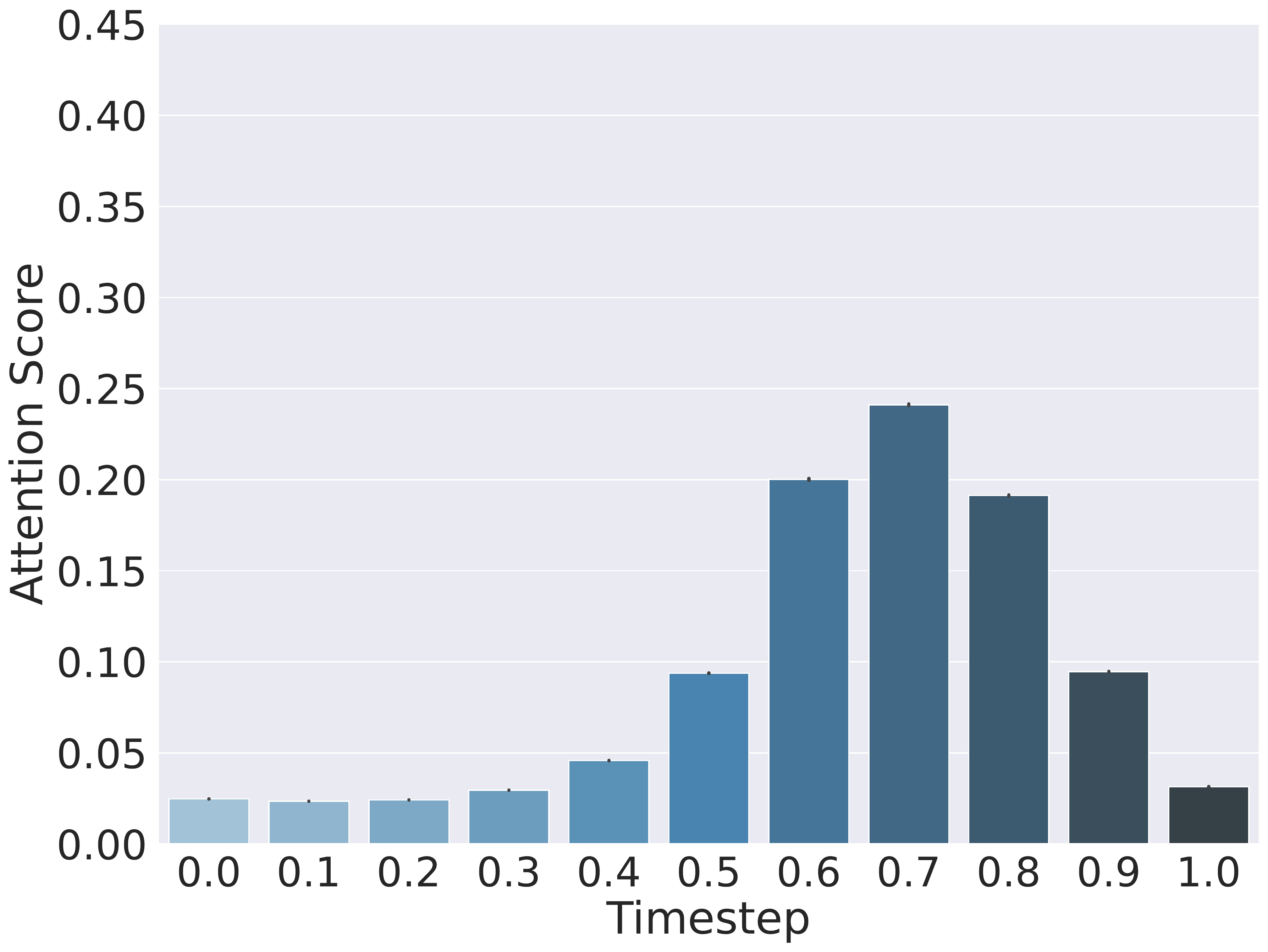}
    \vspace{-6mm}
    \subcaption{\scriptsize VDRL $|$ Object Shape}
\end{subfigure} \\
\begin{subfigure}[c]{0.24\linewidth}
    \includegraphics[width=\linewidth]{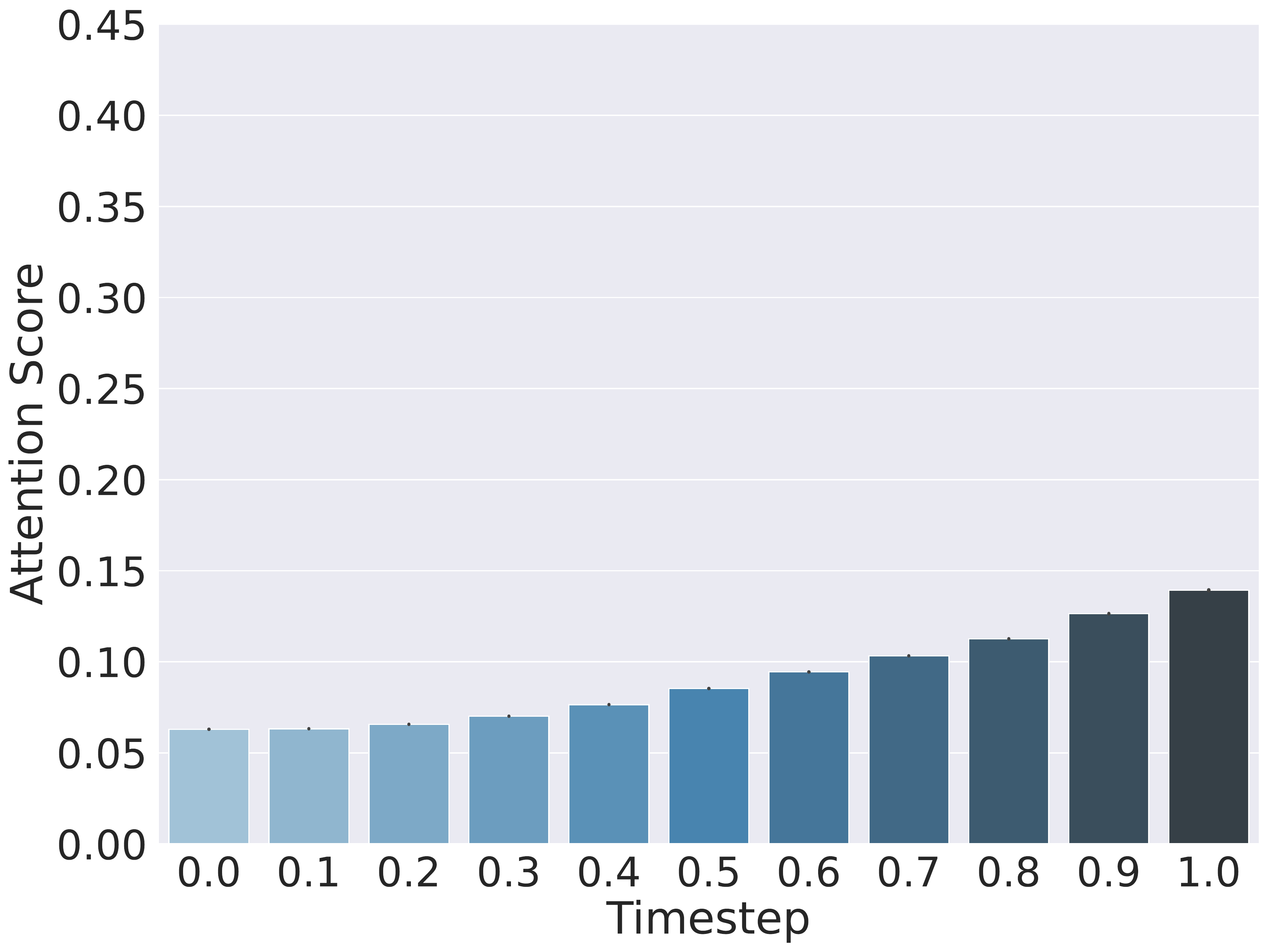}
    \vspace{-6mm}
    \subcaption{\scriptsize DRL $|$ Background Color}
\end{subfigure}
\begin{subfigure}[c]{0.24\linewidth}
    \includegraphics[width=\linewidth]{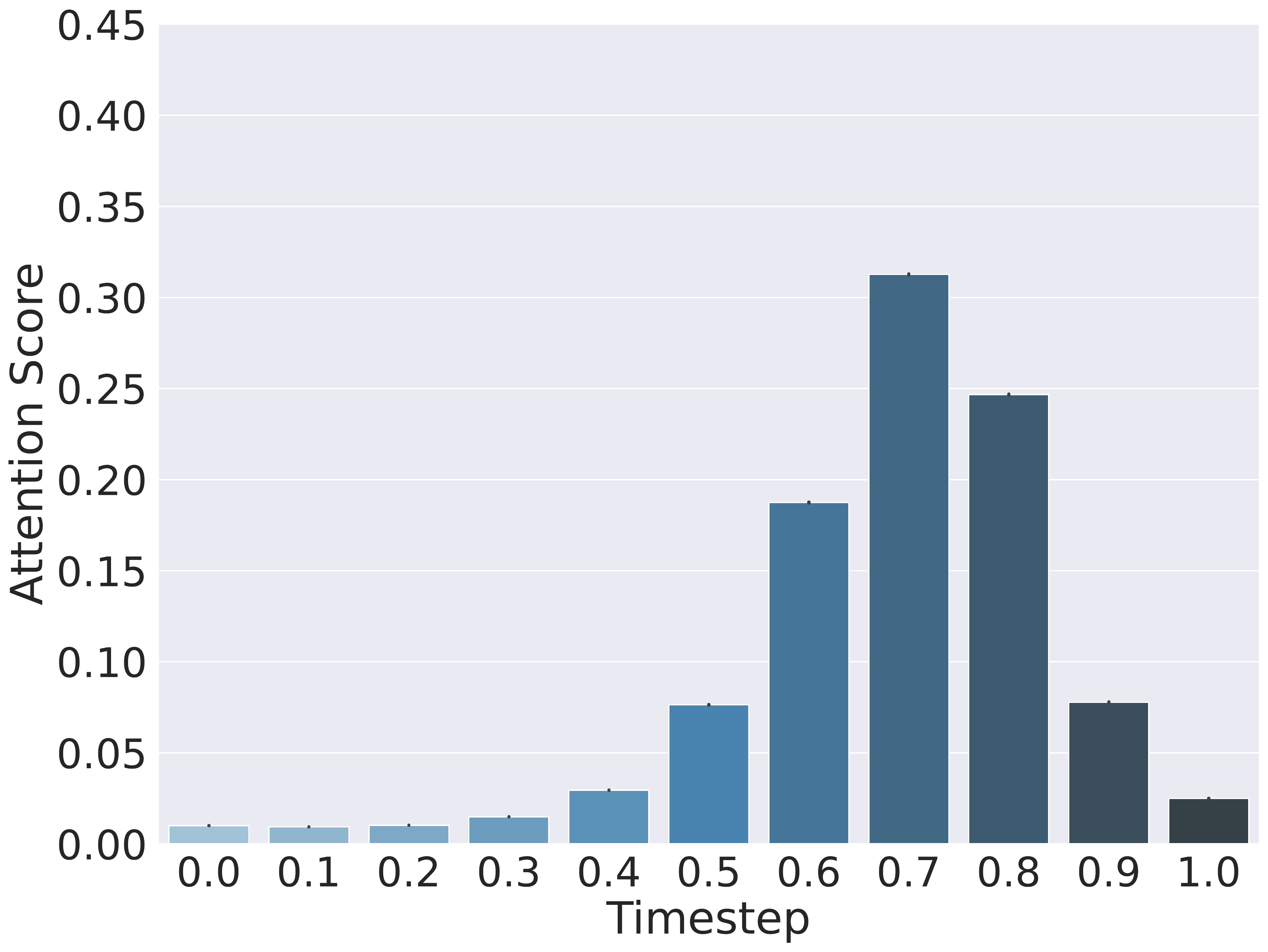}
    \vspace{-6mm}
    \subcaption{\scriptsize DRL $|$ Foreground Color}
\end{subfigure}
\begin{subfigure}[c]{0.24\linewidth}
    \includegraphics[width=\linewidth]{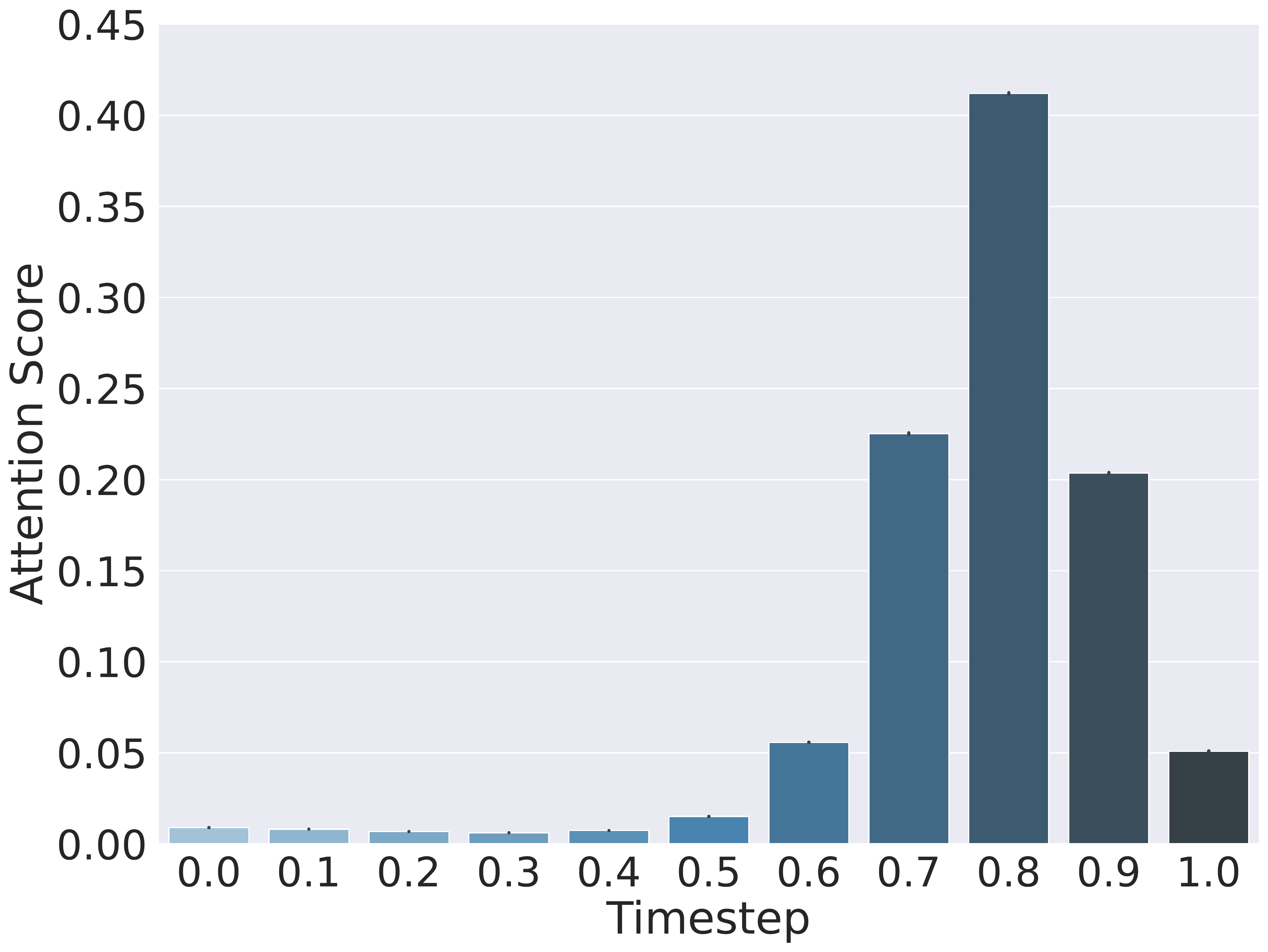}
    \vspace{-6mm}
    \subcaption{\scriptsize DRL $|$ Location}
\end{subfigure}
\begin{subfigure}[c]{0.24\linewidth}
    \includegraphics[width=\linewidth]{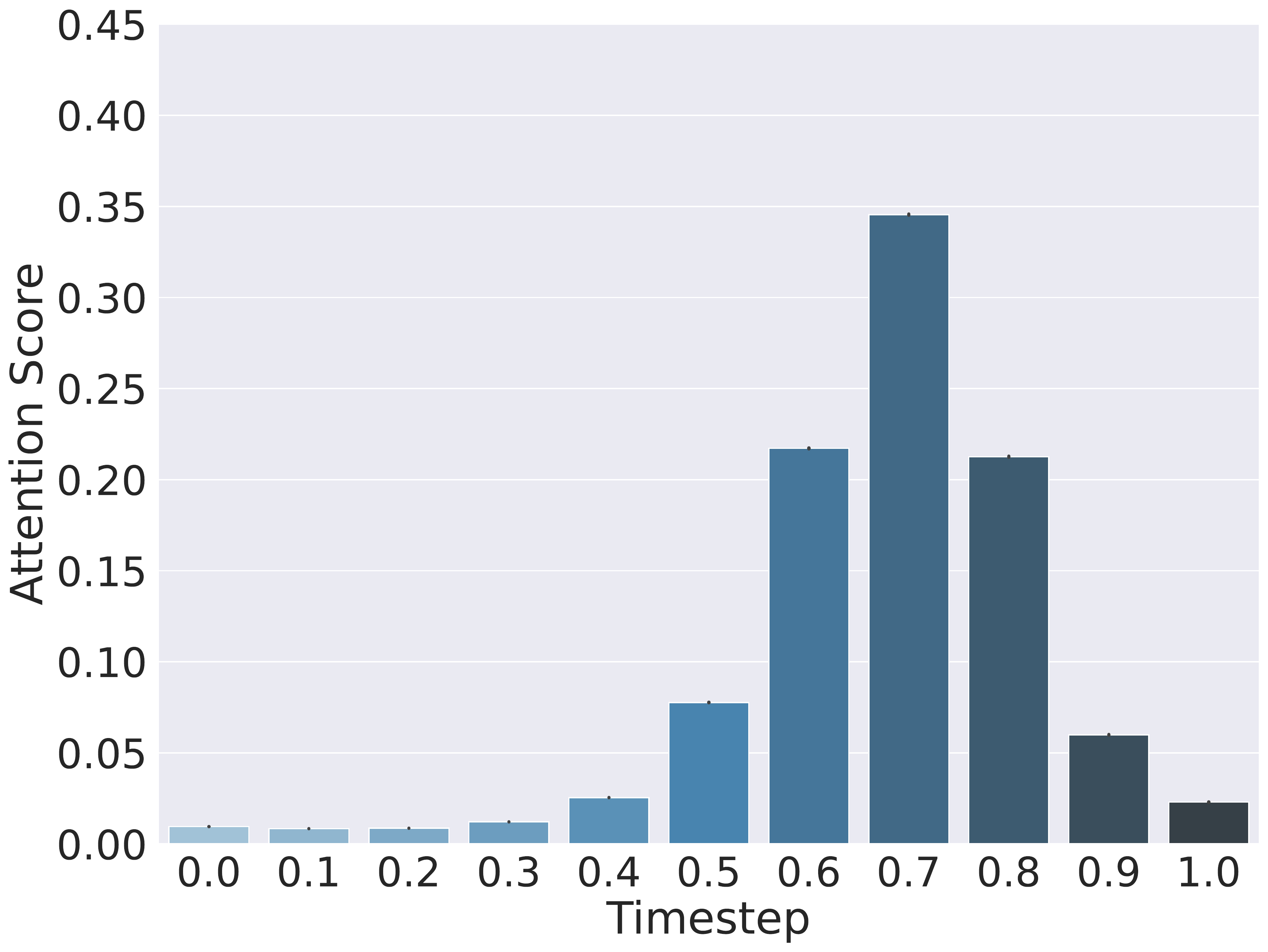}
    \vspace{-6mm}
    \subcaption{\scriptsize DRL $|$ Object Shape}
\end{subfigure}
\caption{Attention score profiles for different tasks under the \textit{Synthetic} dataset, when using the VDRL framework (top) and the DRL framework (bottom). The scores reveal that almost all points in the trajectory store similar amounts of information about the background color, while the latter part of the trajectory encodes more information about the foreground object. In particular, information about the location is most heavily found near the end of the trajectory.}
\label{fig:syn}
\vspace{-5mm}
\end{figure}
\vspace{-2mm}
\subsection{Mutual Information Reveals Differences Along the Trajectory}
\label{sec:mi}
In an effort to understand whether different parts of the trajectory based representation actually contain different types of information about the sample, we evaluate the mutual information between the representations at various points in the trajectory. We use the MINE algorithm~\citep{belghazi2018mutual} to estimate the mutual information between the representations at any two different points in the trajectory. Through this algorithm, we compute and analyse a normalized version of the mutual information, defined as
$\text{NMI}(\v{X};\v{Y}) := \text{I}(\v{X};\v{Y}) / \sqrt{\text{H}(\v{X})\text{H}(\v{Y})}$
where $\text{I}(\cdot \;;\, \cdot)$ is the standard Mutual Information function~\citep{cover1999elements} and $\text{H}(\cdot)$ is the entropy function.

Figure~\ref{fig:nmi-mult} illustrates the normalized mutual information between representations at different parts of the trajectory across three different datasets: CIFAR10, CIFAR100 and Mini-ImageNet as well as two different types of models: VDRL and DRL, where the former uses a probabilistic encoder and the latter doesn't. We see large normalized mutual information values near the principal diagonal and small values that are away from it, demonstrating that nearby representations on the trajectory are similar whereas distant points in the trajectory are considerably different. This shows that different parts of the trajectory learn to encode different kinds of information.

\begin{figure}
    \centering
\begin{subfigure}[c]{0.32\linewidth}
    \includegraphics[width=\linewidth]{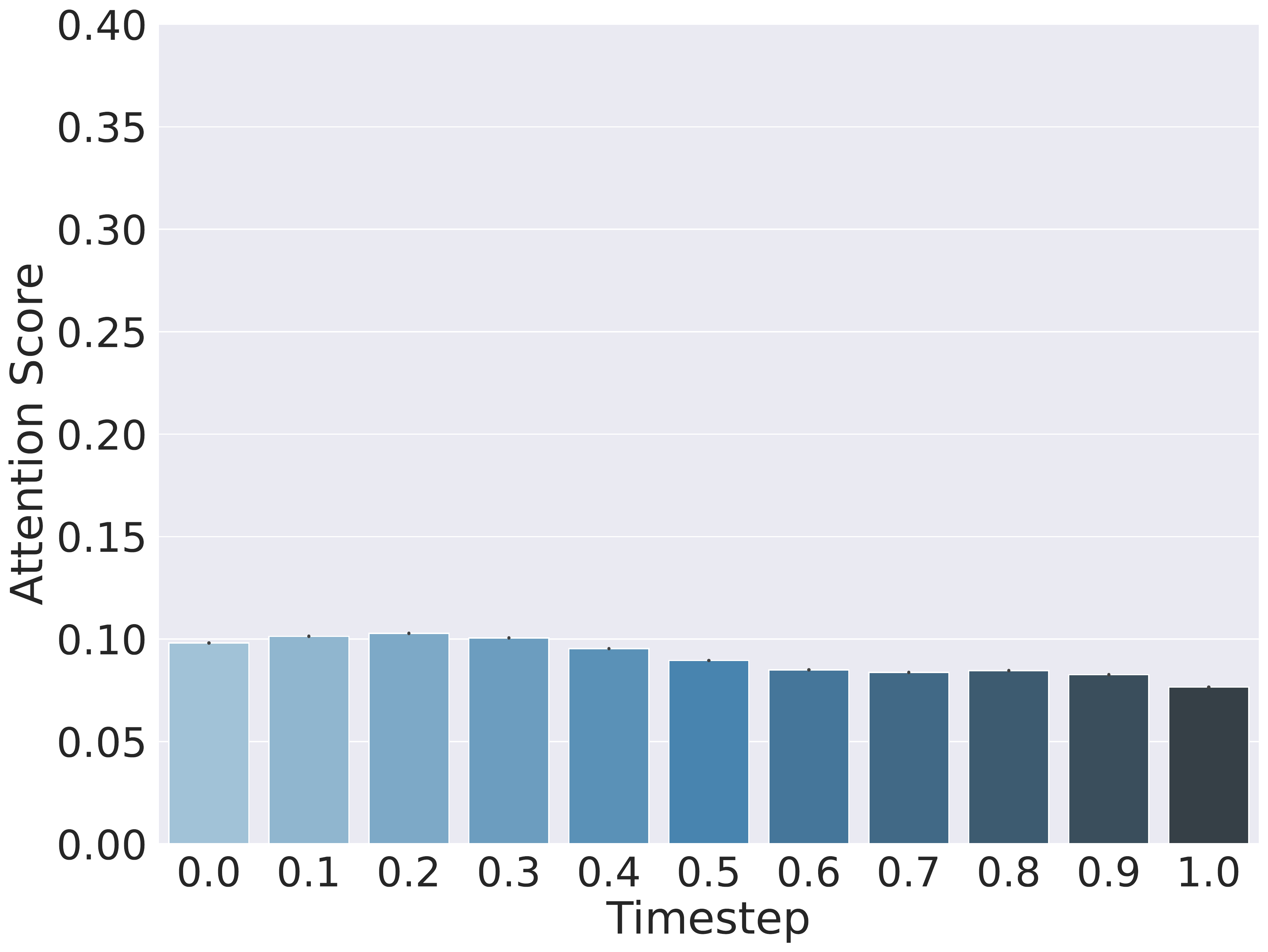}
    \vspace{-6mm}
    \subcaption{\scriptsize VDRL $|$ Background Color}
\end{subfigure}
\begin{subfigure}[c]{0.32\linewidth}
    \includegraphics[width=\linewidth]{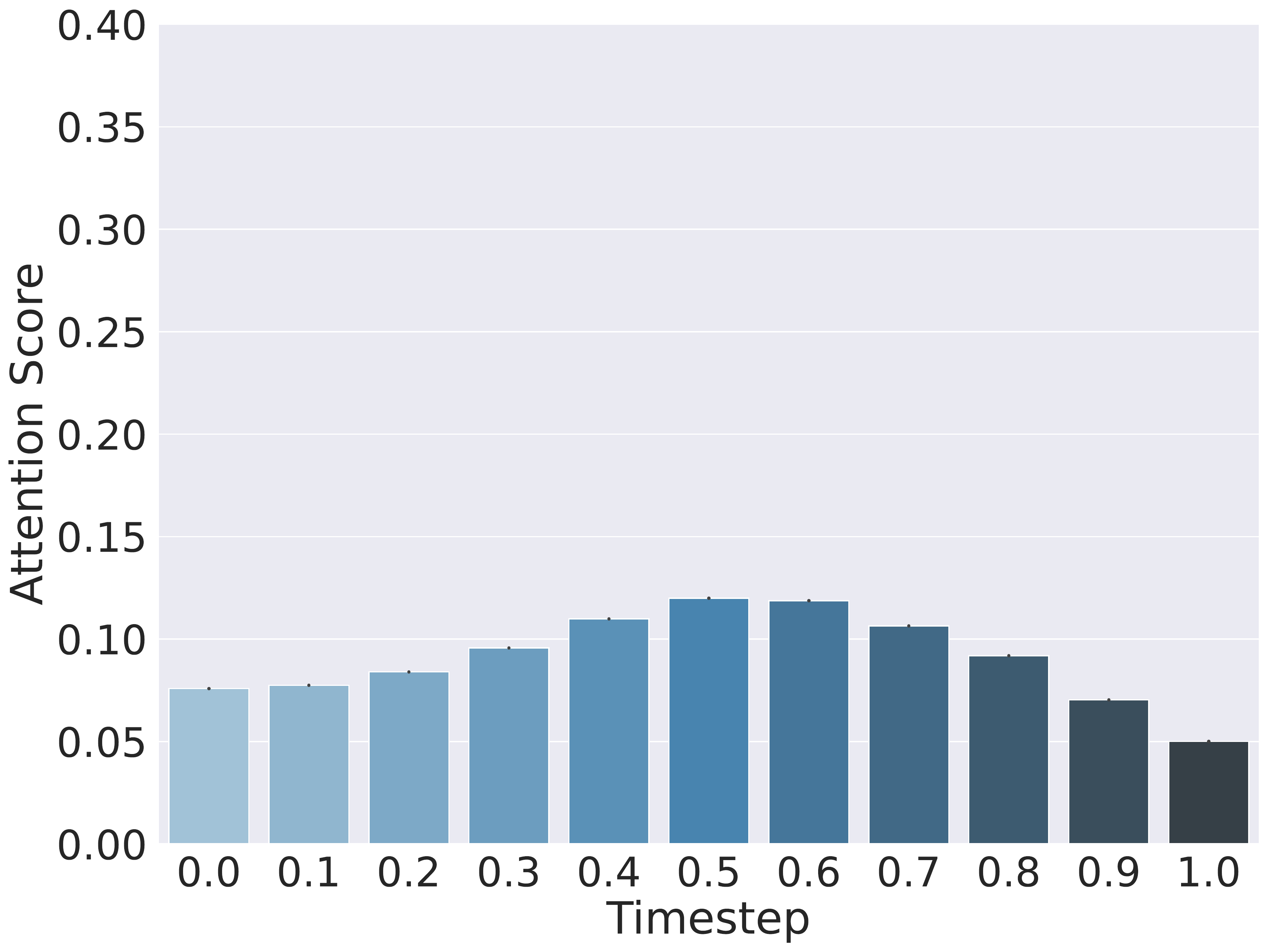}
    \vspace{-6mm}
    \subcaption{\scriptsize VDRL $|$ Foreground Color}
\end{subfigure}
\begin{subfigure}[c]{0.32\linewidth}
    \includegraphics[width=\linewidth]{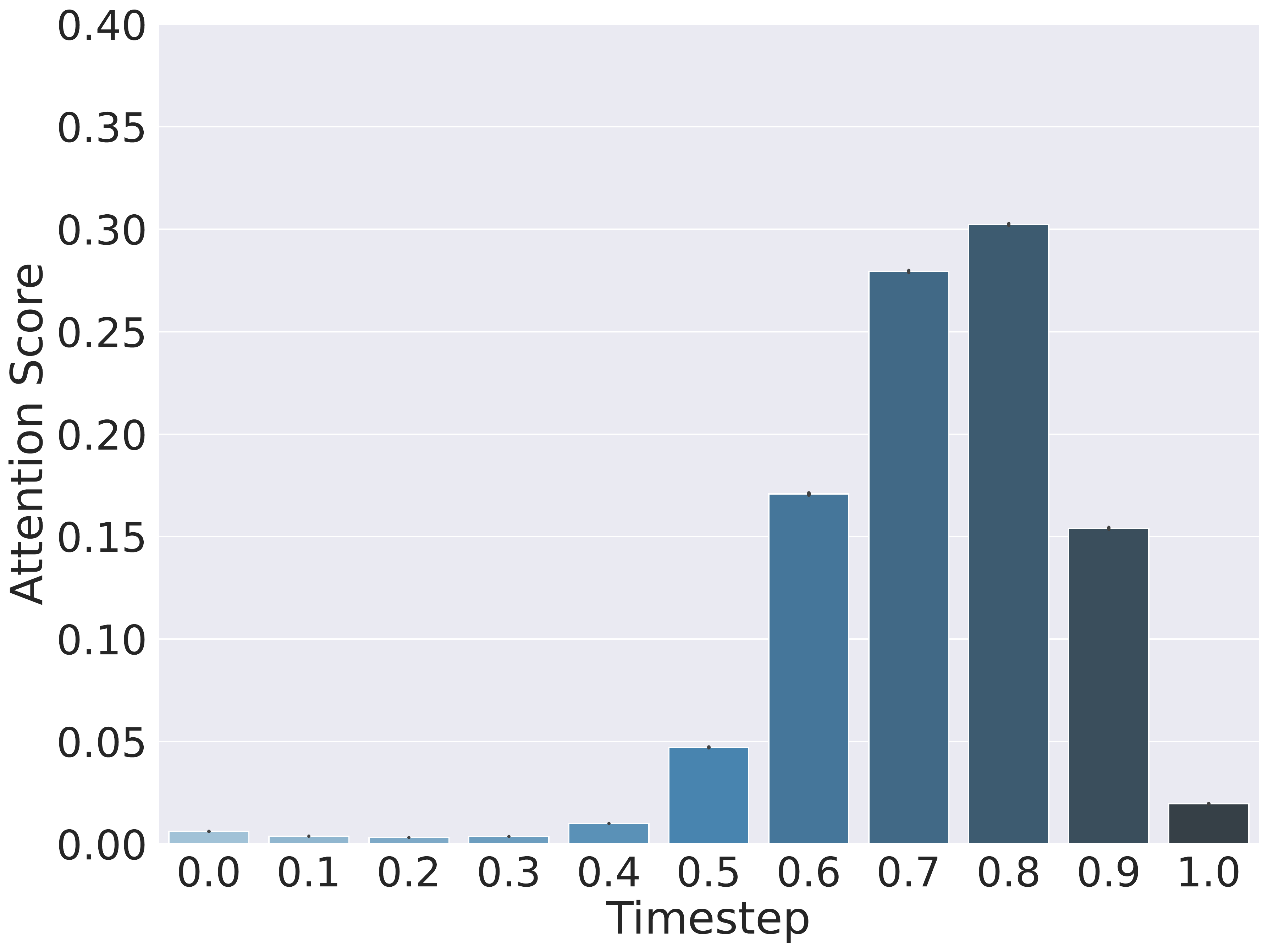}
    \vspace{-6mm}
    \subcaption{\scriptsize VDRL $|$ Digit Identity}
\end{subfigure} \\
\begin{subfigure}[c]{0.32\linewidth}
    \includegraphics[width=\linewidth]{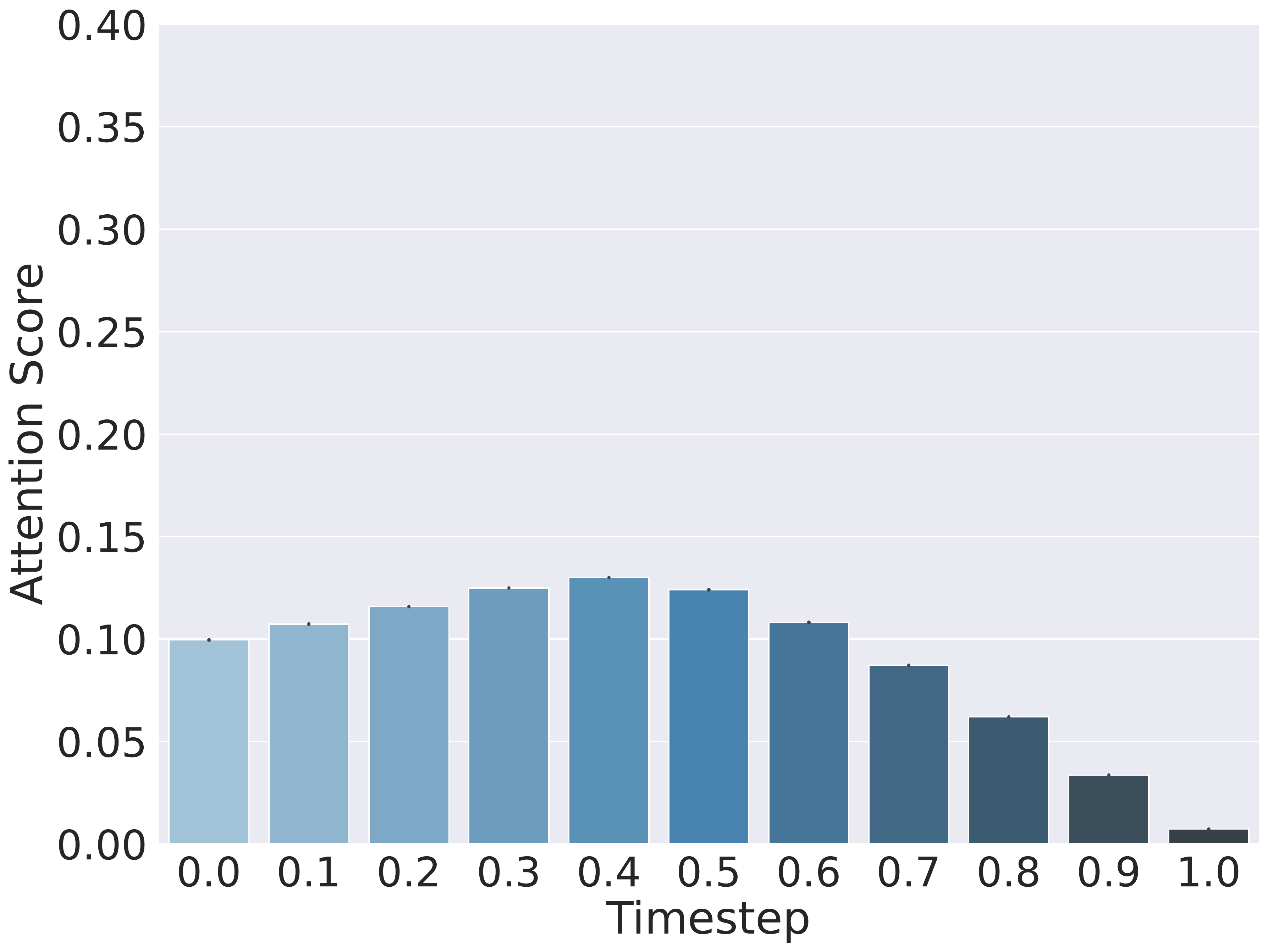}
    \vspace{-6mm}
    \subcaption{\scriptsize DRL $|$ Background Color}
\end{subfigure}
\begin{subfigure}[c]{0.32\linewidth}
    \includegraphics[width=\linewidth]{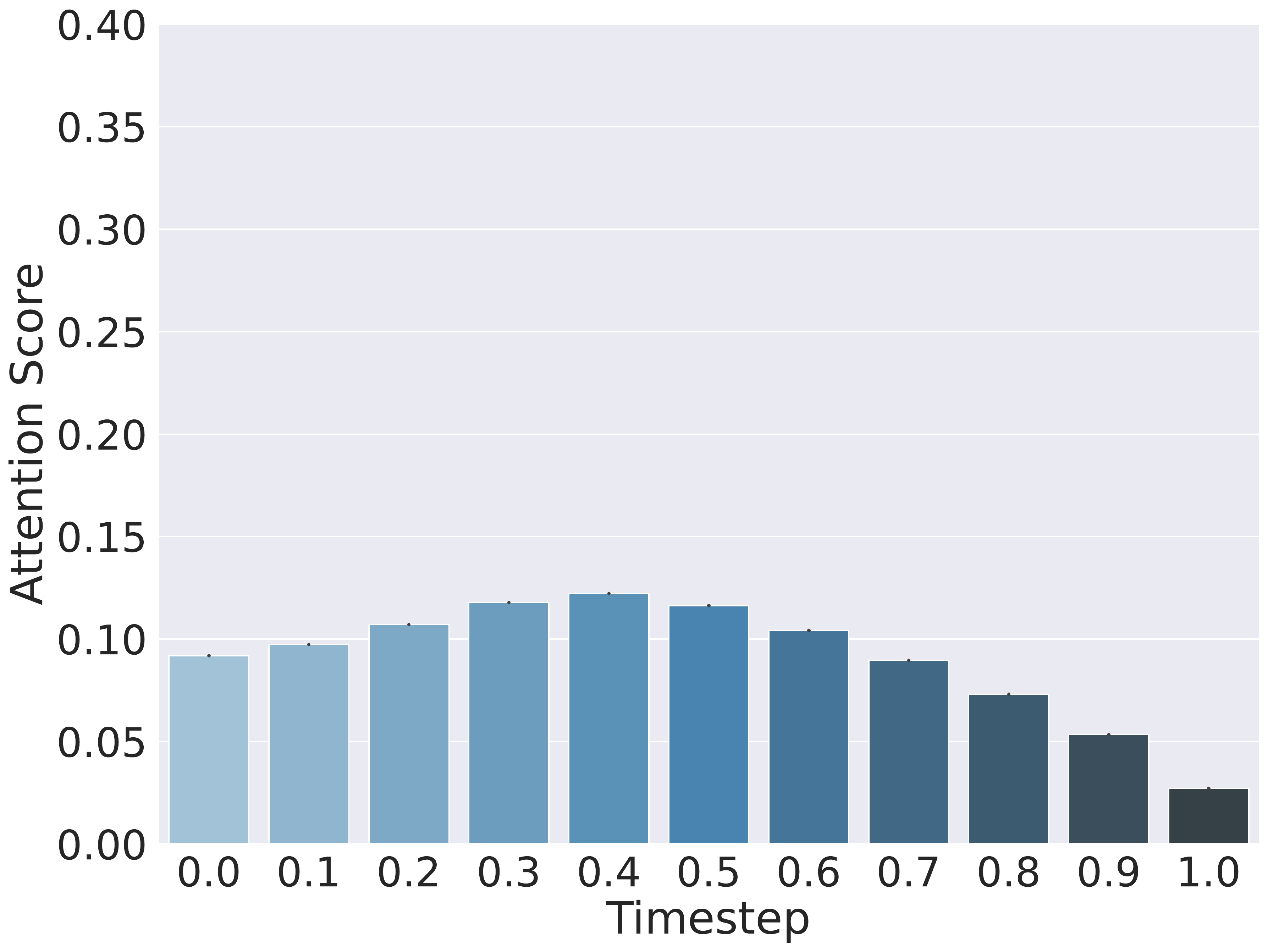}
    \vspace{-6mm}
    \subcaption{\scriptsize DRL $|$ Foreground Color}
\end{subfigure}
\begin{subfigure}[c]{0.32\linewidth}
    \includegraphics[width=\linewidth]{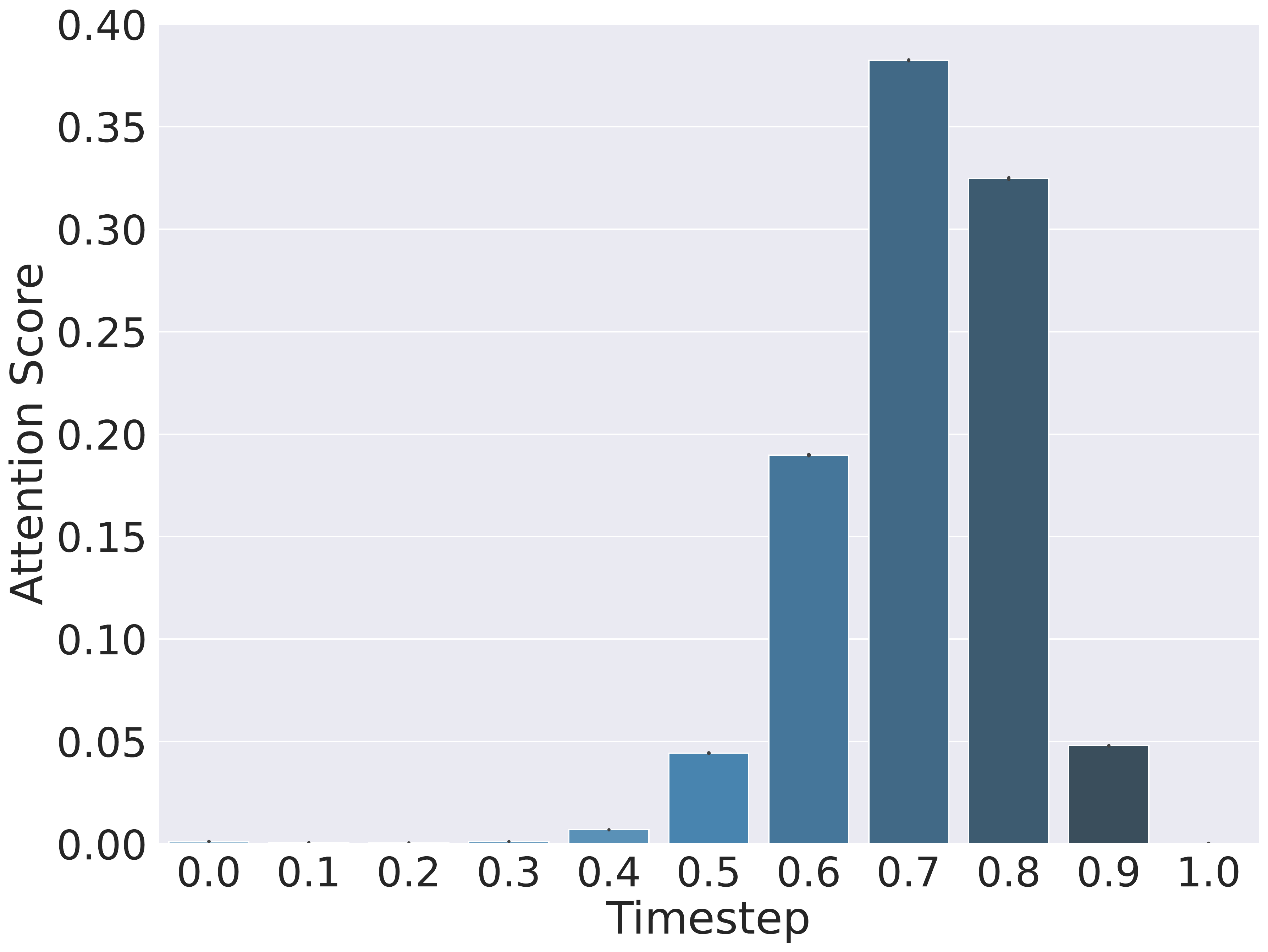}
    \vspace{-6mm}
    \subcaption{\scriptsize DRL $|$ Digit Identity}
\end{subfigure}
\caption{Attention score profiles for different tasks under the \textit{Colored-MNIST} dataset, when using the VDRL framework (top) and the DRL framework (bottom). The scores reveal that almost all points in the trajectory store similar amounts of information about the background as well as the foreground color, while the latter part of the trajectory encodes more information about the identity of the digit.}
    \label{fig:cmnist}
    \vspace{-5mm}
\end{figure}
\vspace{-2mm}
\subsection{Attention Reveals Relevance of Different Parts of the Trajectory}
\vspace{-2mm}
\label{sec:att-ablate}
To complement the analysis in Sections~\ref{sec:downstream} and~\ref{sec:mi}, we train a single-layered Transformer model for downstream prediction, which comes from a learned embedding that queries information from different parts of the trajectory. Through the analysis of the attention scores at different points in the trajectory, we realize that the middle parts of the trajectory are the most important, as illustrated in the high attention scores around $t=0.5$ in Figure~\ref{fig:att-ablate}. 

This is in line with the performance results in Figure~\ref{fig:perf} which also shows that amongst the single-point MLP-based systems, the best downstream performance is reached near the middle of the trajectory. Attention score for any point in the trajectory, in a single-layered Transformer network, can be understood as the weight or importance of that point in the whole trajectory for the task in consideration. For the three image-classification datasets that we experiment on, we see that the attention patterns are quite similar. However, in later sections we will provide analysis with more controlled settings and see that the attention score profiles show varied behaviour for different features, indicating and strengthening the claim from Section \ref{sec:mi} that the trajectory indeed encodes different information at different points.
\vspace{-2mm}
\subsection{Parsing Semantic Information Encoded along the Trajectory}
\vspace{-2mm}
To better understand the different kind of information encoded in different parts of the trajectory, we expand our analysis into multi-task domains where each task relies on information from different features in the input. We consider three different datasets for this fine-grained analysis: \textit{Synthetic}, \textit{Colored-MNIST} and \textit{CelebA}.

For all the analysis performed here, we train the diffusion model with the time-dependent encoder using Equation \ref{eq:obj}, and then keep it frozen. We then perform inference to obtain the trajectory representations for different granularities for each data point. Then, for each task in the dataset, we train a different single-layered transformer model and obtain attention scores over the trajectory corresponding to that particular task. This attention score over the points on the trajectory encodes the relevance of that area of the trajectory for the task, thereby providing insights on how much information about a particular feature is encoded in which part of the trajectory.

\begin{wrapfigure}{r}{0.3\textwidth}
\vspace{-4mm}
\includegraphics[width=0.3\textwidth]{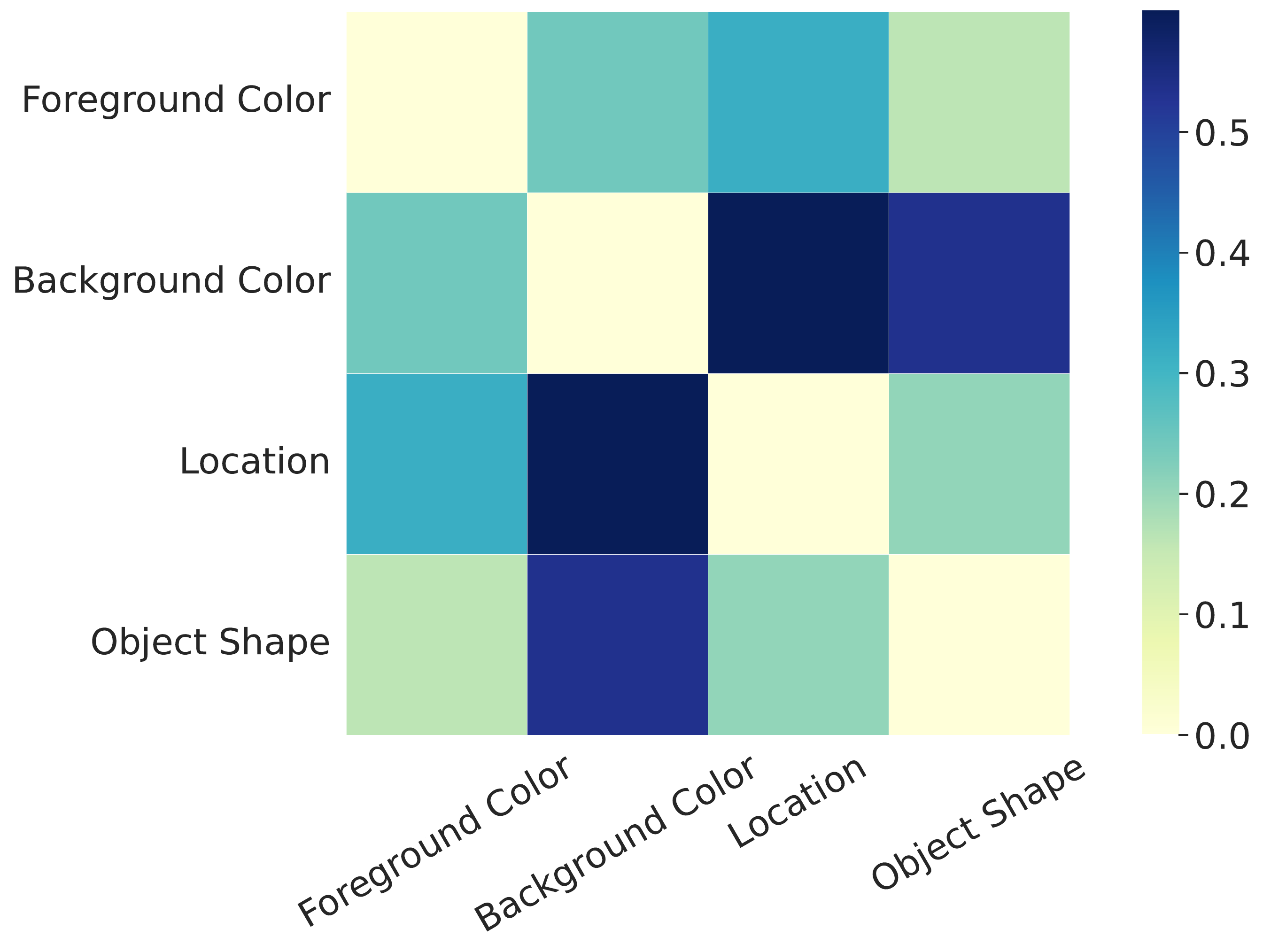}
\caption{Jensen Shannon Divergence plot for the attention profiles obtained for any pair of features for the \textit{Synthetic} dataset using VDRL encoder.}
\label{fig:syn_jsd}
\vspace{-7mm}
\end{wrapfigure}
\textbf{Synthetic}. Synthetic dataset consists of an object in a scene. The scene consists of a distinct background color and the object is associated with a distinct foreground, location as well as the object shape. The system consists of four tasks; determining the (a) background color, (b) foreground color, (c) object location, and (d) object shape.

Figure \ref{fig:syn} highlights the relevance of different parts of the trajectories for the different tasks. We see that while the background color information is more or less uniformly diffused over the whole trajectory, foreground information like object color, location and shape have a much more peaky distribution.

We also refer the readers to Figure \ref{fig:syn_jsd} which highlights the differences in attention distributions over the trajectory. Cell ($i,j$) in the figure refers to the Jensen-Shannon divergence (JSD) between the distribution over trajectory obtained for task $i$ with that obtained for task $j$.

\begin{wrapfigure}{r}{0.3\textwidth}
\vspace{-4mm}
\includegraphics[width=0.3\textwidth]{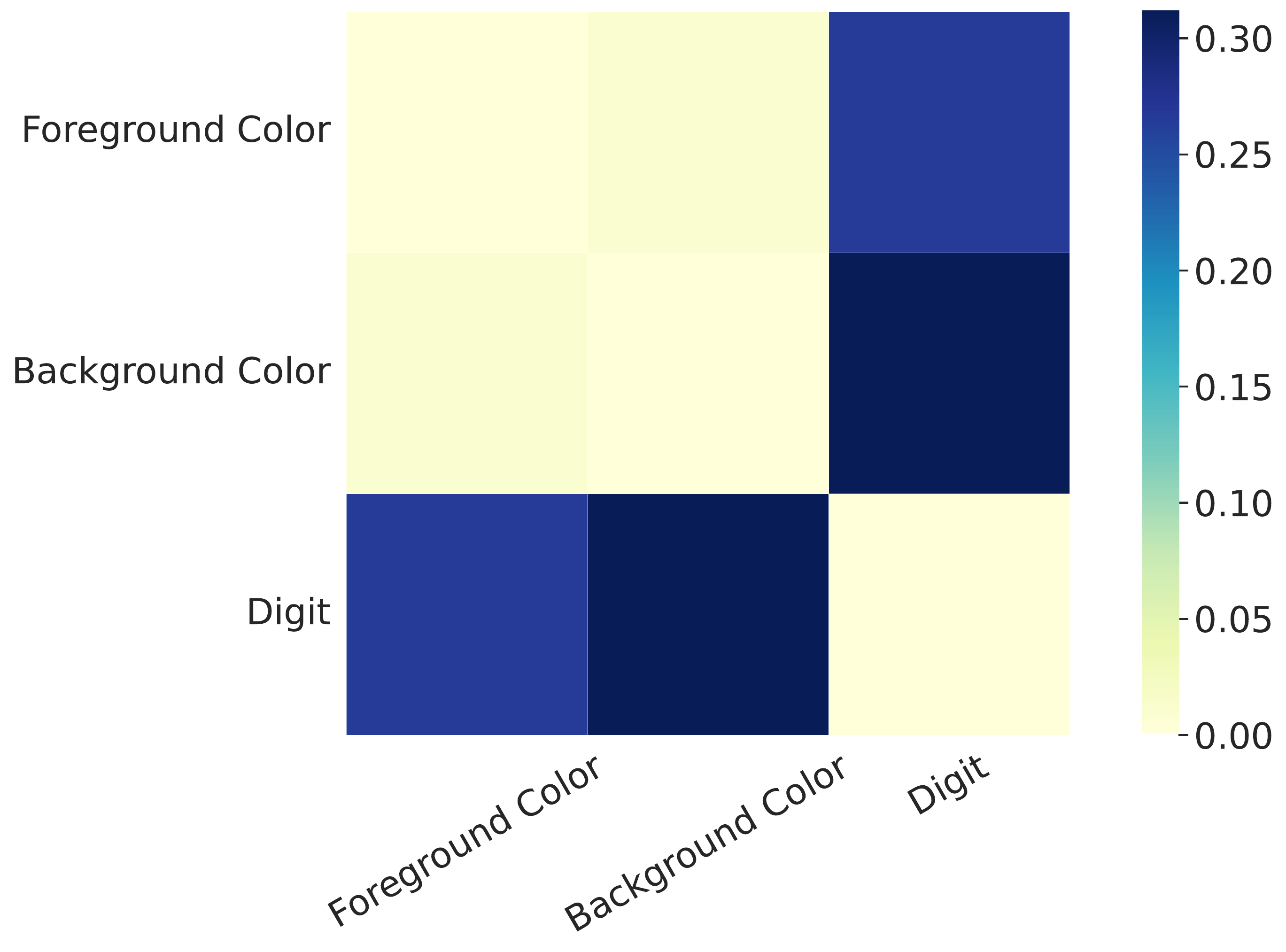}
\caption{Jensen Shannon Divergence plot for the attention profiles for any pair of features for the \textit{Colored-MNIST} dataset using VDRL encoder.}
\label{fig:cmnist_jsd}
\vspace{-7mm}
\end{wrapfigure}
For additional details about the analysis as well as additional ablations and results using different granularities and latent dimension sizes, please check out Appendix \ref{apdx:syn}. We also provide examples of samples from this setup in the Appendix.

\begin{figure}
\begin{minipage}[c]{0.42\textwidth}
    \includegraphics[width=\textwidth]{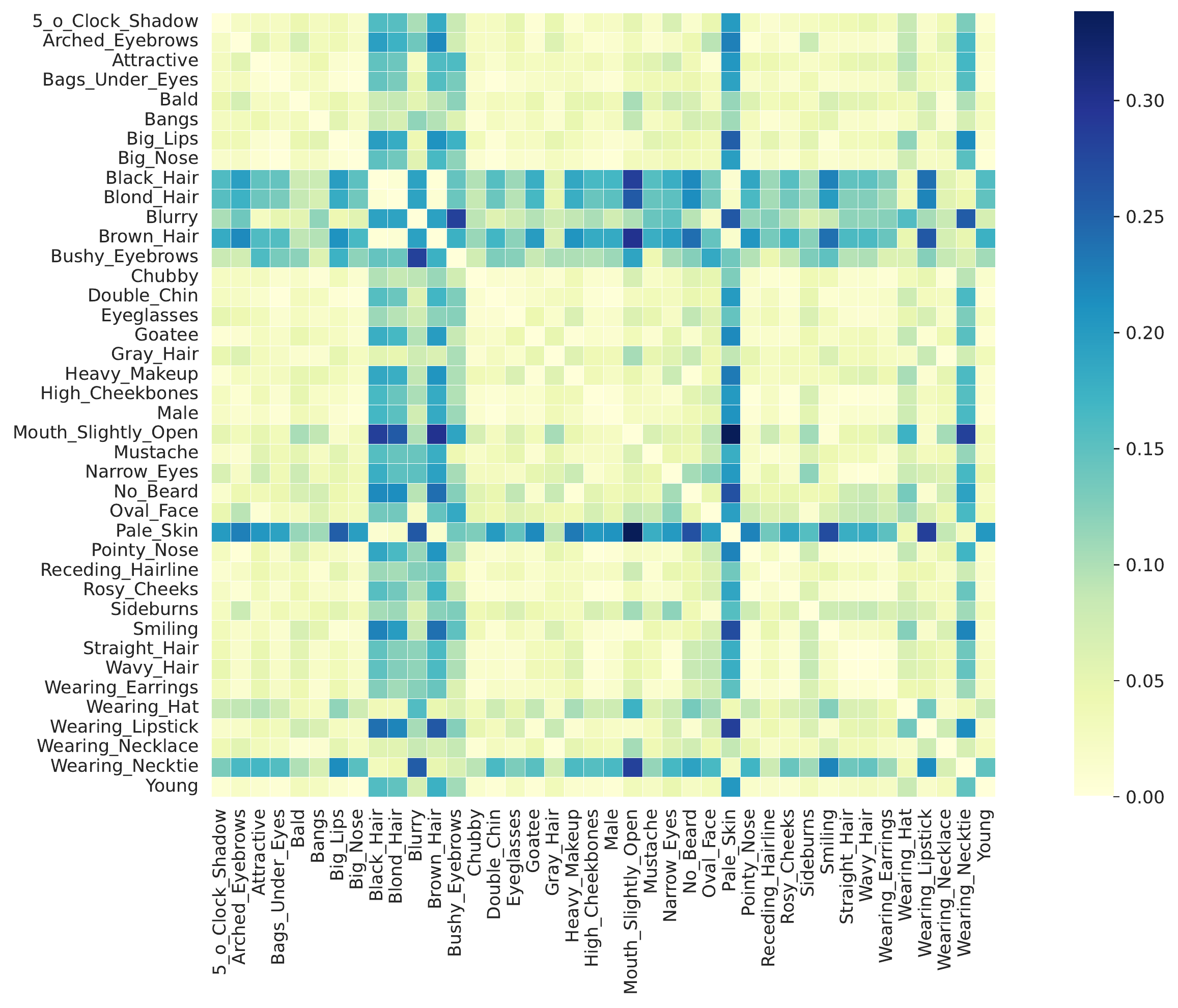}
\end{minipage}
\begin{minipage}{0.57\textwidth}
\begin{subfigure}[c]{0.49\textwidth}
\includegraphics[width=\textwidth]{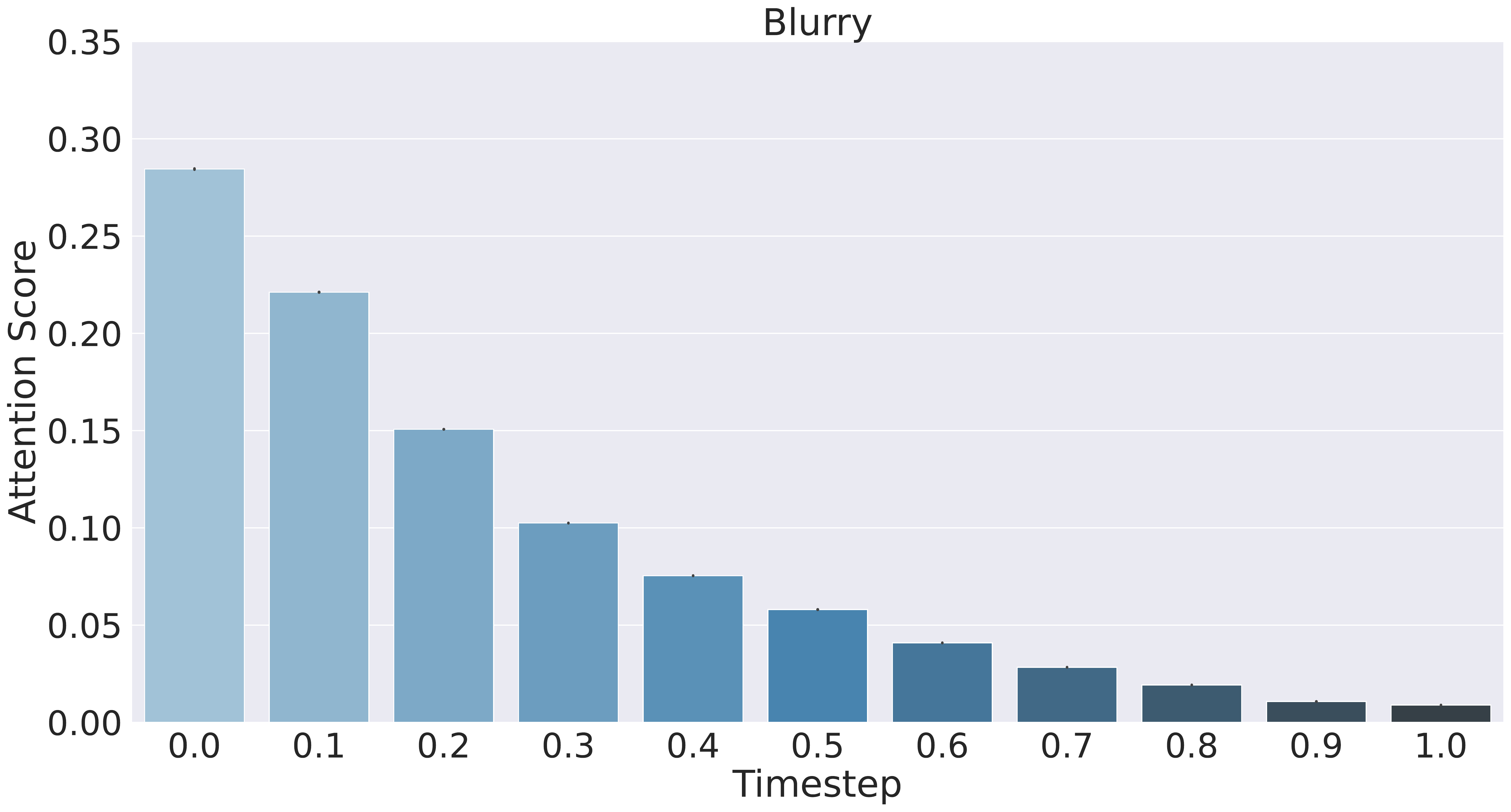}
\vspace{-5mm}
\subcaption{\tiny VDRL $|$ Blurry}
\end{subfigure}
\begin{subfigure}[c]{0.49\textwidth}
    \includegraphics[width=\textwidth]{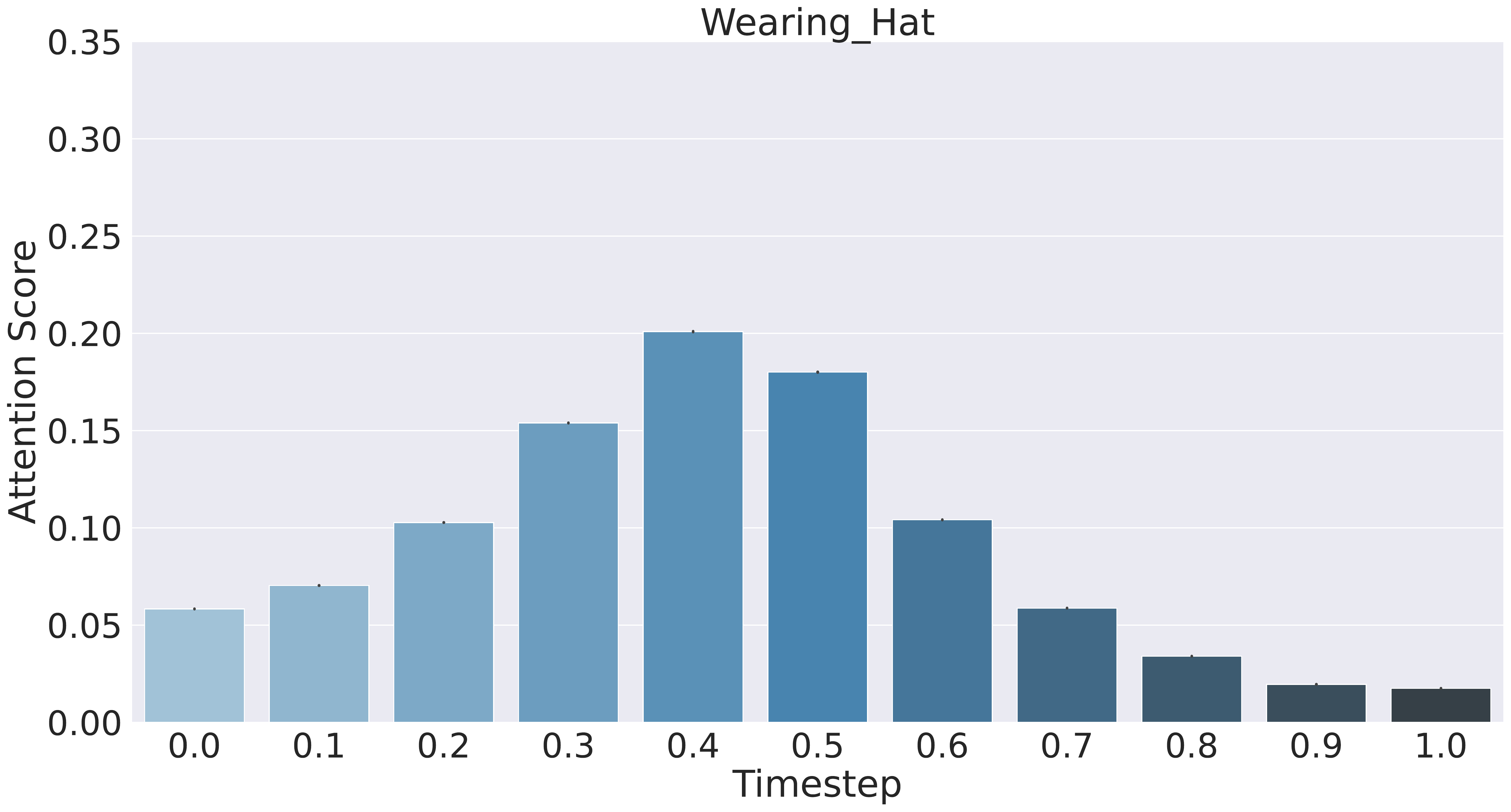}
\vspace{-5mm}
\subcaption{\tiny VDRL $|$ Wearing Hat}
\end{subfigure} \\
\begin{subfigure}[c]{0.49\textwidth}
    \includegraphics[width=\textwidth]{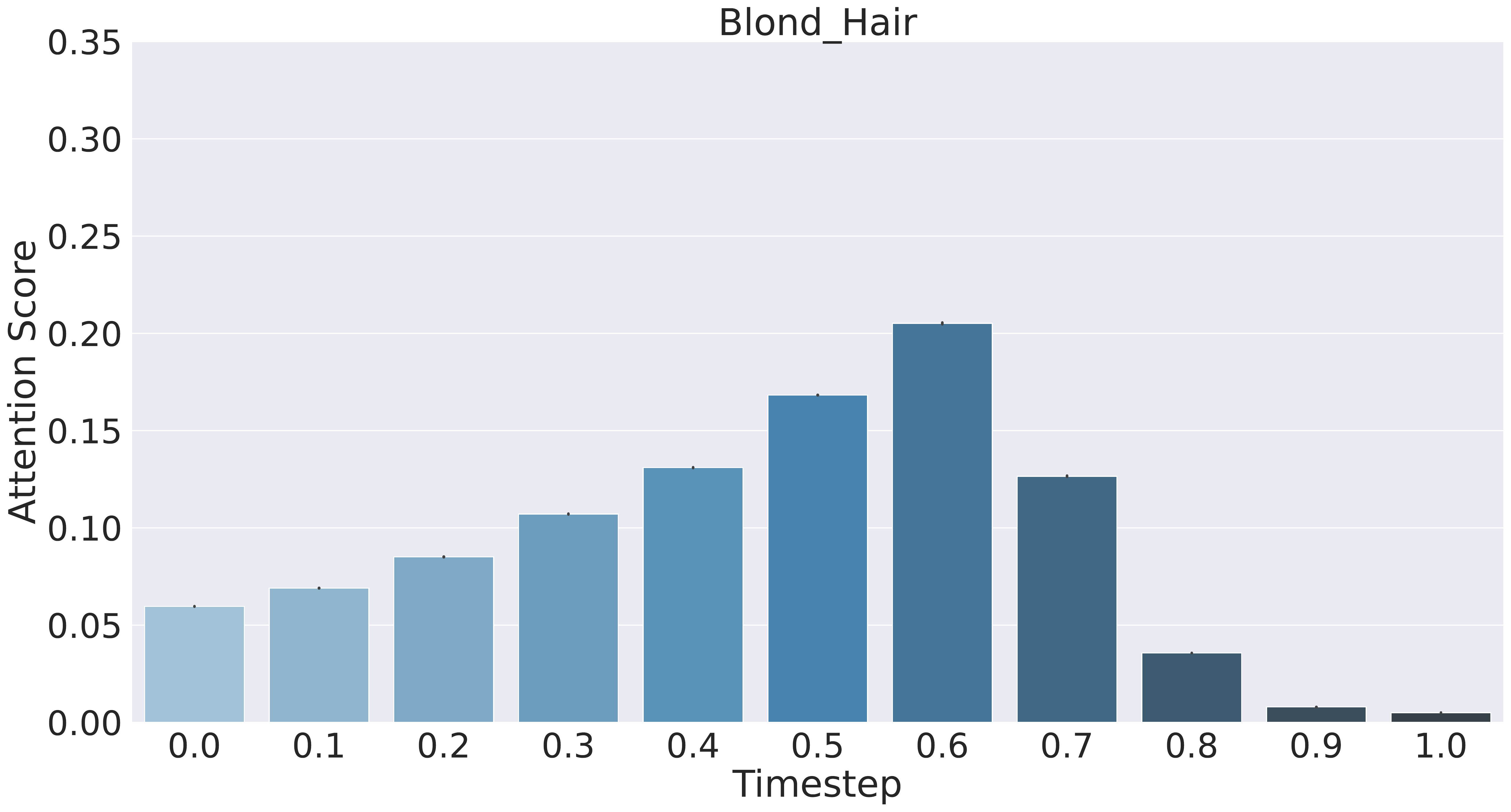}
\vspace{-5mm}
\subcaption{\tiny VDRL $|$ Blond Hair}
\end{subfigure}
\begin{subfigure}[c]{0.49\textwidth}
    \includegraphics[width=\textwidth]{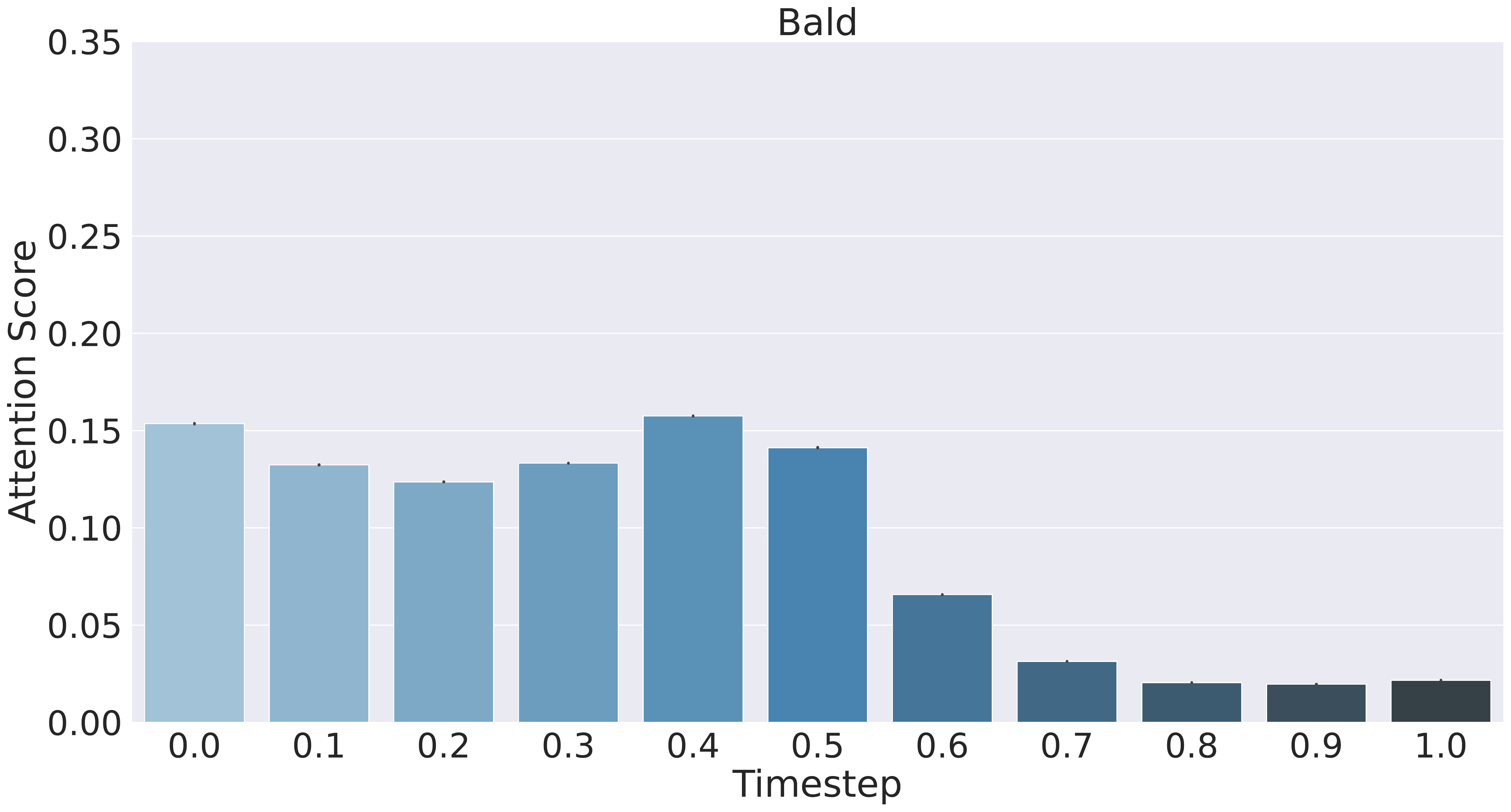}
\vspace{-5mm}
\subcaption{\tiny VDRL $|$ Bald}
\end{subfigure}
\end{minipage}
\caption{\textbf{Left}: Jensen Shannon Divergence plot for the attention profiles for any pair of features in the \textit{CelebA} dataset; \textbf{Right}: Attention Profile plots for a subset of features in the \textit{CelebA} dataset.}
\label{fig:celeba_abla}
\vspace{-9mm}
\end{figure}

\textbf{Colored-MNIST}. We then extend our analysis to a slightly more complex setting, where each sample consists of a digit with a distinct foreground color and digit identity (0-9), along with a background color on which the digit is embedded. The multi-task setting here consists of determining the (a) background color, (b) foreground color, and (c) digit identity.

Figure \ref{fig:cmnist} shows which parts of the trajectory are important for which task, and thus encode what kind of information. In this setting, we see that both the background color and the foreground color information is quite diffused over the trajectory, while the information about the digit identity is heavily present in the later parts of the trajectory.

We also refer the readers to Figure \ref{fig:cmnist_jsd} for a birds-eye view on the differences in attention distributions over the trajectory for different tasks. Each cell ($i,j$) in the figure refers to the JSD between the distribution over trajectory obtained for task $i$ with that obtained for task $j$. A high divergence implies that the information corresponding to the two tasks is not present together simultaneously.

For additional details about the analysis as well as additional ablations and results using different granularities and latent dimension sizes, please check out Appendix \ref{apdx:cmnist}. We also provide examples of samples from this setup in the Appendix.

\textbf{CelebA}. Finally, we conduct experiments on the CelebA dataset \citep{liu2015faceattributes}, which is a large-scale dataset that consists of images of celebrities as well as multiple binary labels for each corresponding to the different attributes; eg. whether the celebrity in the image has brown hair or not? We consider experimentation on CelebA to understand if we can semantically understand what kind of features, (in a more real world setting), are encoded in different parts of the trajectory. For a full list of the attributes in consideration in this dataset, we refer the readers to Appendix \ref{apdx:celeba}.

We refer the readers to Figure \ref{fig:celeba_abla} for a similar analysis of JSD between distributions over trajectories encoded by different features (or more formally; tasks corresponding to different features). While we do see some clustering (eg. blond hair, black hair and brown hair all have similar attention profiles), there is also a lot of uniformity in the divergences. We believe that scaling and extending this setting to richer and more diverse multi-task, multi-feature domains would allow for a much richer semantic separation between features.

We also highlight the individual attention profiles in Figure \ref{fig:celeba_abla} for a subset of the features. It shows that the distributions learned for different features are actually different, implying presence of complementary information along the trajectory (Figures \ref{fig:celeba_activ_vdrl} and \ref{fig:celeba_activ_drl} in the Appendix).

Overall, we highlight how information about different features is encoded in different regions of the trajectory. This can be leveraged by learning an automated task-conditioned system that learns to ``look in more detail" at certain parts of the trajectory while ignoring the others. For additional details about the analysis as well as additional ablations, please check out Appendix \ref{apdx:celeba}.

\begin{figure}
    \centering
    \includegraphics[width=\textwidth]{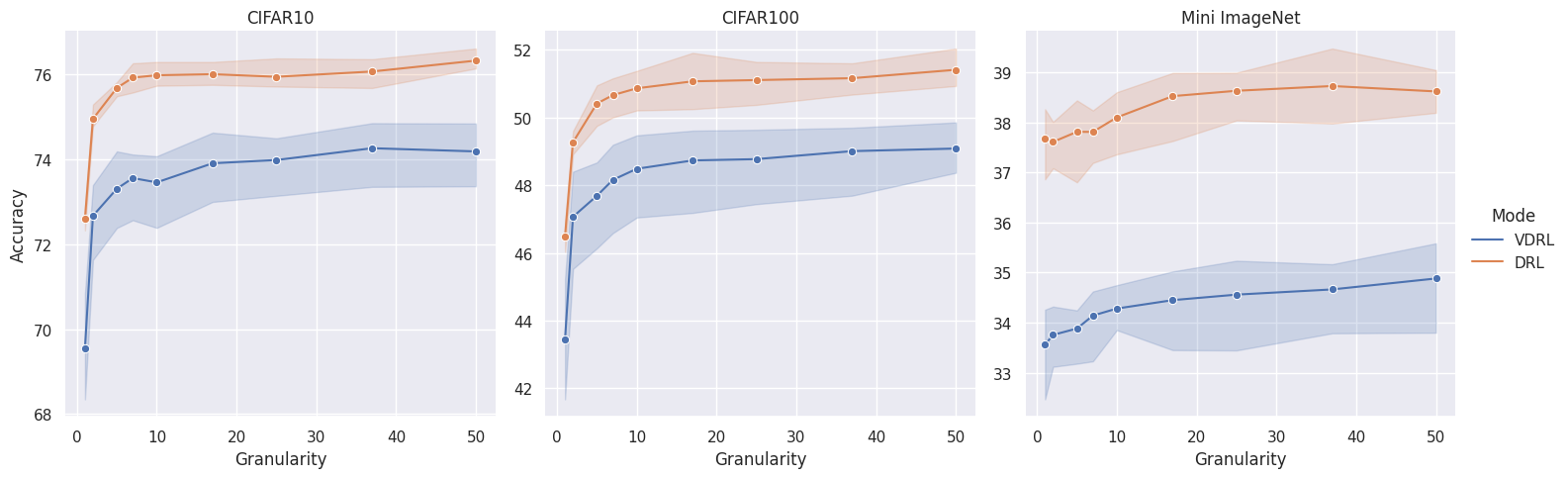}
    \caption{Performance on different datasets evaluated at different granularities of the trajectory representation. Granularity-$k$ on the X-axis implies that the discretization of the trajectory representation (obtained from the same frozen diffusion model), is done at $k$ uniform points, which are then fed to a Transformer model for downstream classification.}
    \label{fig:granularity}
    \vspace{-6mm}
\end{figure}
\vspace{-3mm}
\subsection{Benefits of using more Points in the Trajectory}
\vspace{-2mm}
\label{sec:gran}
We now extend our analysis to understand the benefits of having more points in the trajectory, that is, of moving closer to the continuous-time domain. We do this by considering more and more fine-grained discretizations of the trajectories, which we denote as granularity. A granularity of $k$ discretizes each trajectory representation by uniformly querying it at $(k+1)$ different points, in $[0,1]$. Thus, a granularity level of $2$ indicates using the points $\{0.0, 0.5, 1.0\}$ for downstream predictions.

We refer the readers to Figure \ref{fig:granularity} which illustrates the benefits of having more points in the trajectory. In particular, even though a granularity level of $2$ has access to the mid-point of the trajectory, it doesn't do as well as when using a larger granularity. However, we do notice that the benefits to having more points in the trajectory do start to saturate beyond a certain point.

We believe that the dimensionality of the latent code might have a significant effect on the saturating point. To this end, we perform experimentation on low-dimensional trajectories on the synthetic datasets in the next section.

\subsection{Interplay between Code-Dimensionality and Granularity}

To provide additional analysis into the benefits of having this infinite-dimensional representation, we consider the \textit{Synthetic} and \textit{Colored-MNIST} datasets and use a very small dimensionality for the latent space, which is the output of the encoder. In particular, we consider the output of the encoder to be a 2-dimensional code, but unbounded in the time domain.

We discretize this 2-dimensional trajectory representation at different uniform points, similar to the analysis done in Section \ref{sec:gran} to see how well a limited capacity (in latent bits per point on trajectory) code can do and how much does it benefit from the increase in dimensions through the time dimension?

\begin{figure}
\centering
\begin{minipage}[c]{0.45\textwidth}
    \includegraphics[width=\textwidth]{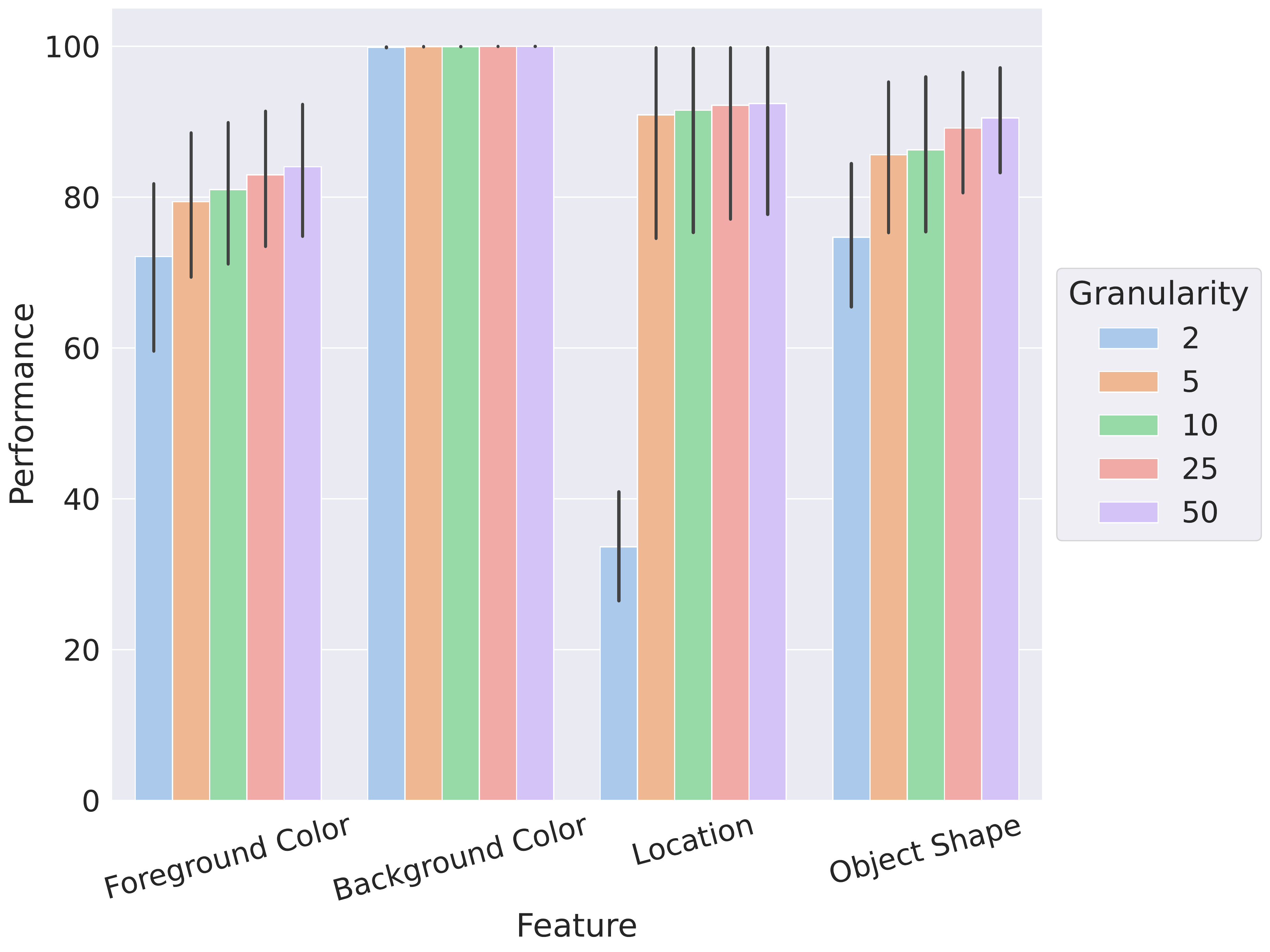}
\end{minipage}
\hspace{0.02\textwidth}
\rulesep
\hspace{0.02\textwidth}
\begin{minipage}[c]{0.45\textwidth}
    \includegraphics[width=\textwidth]{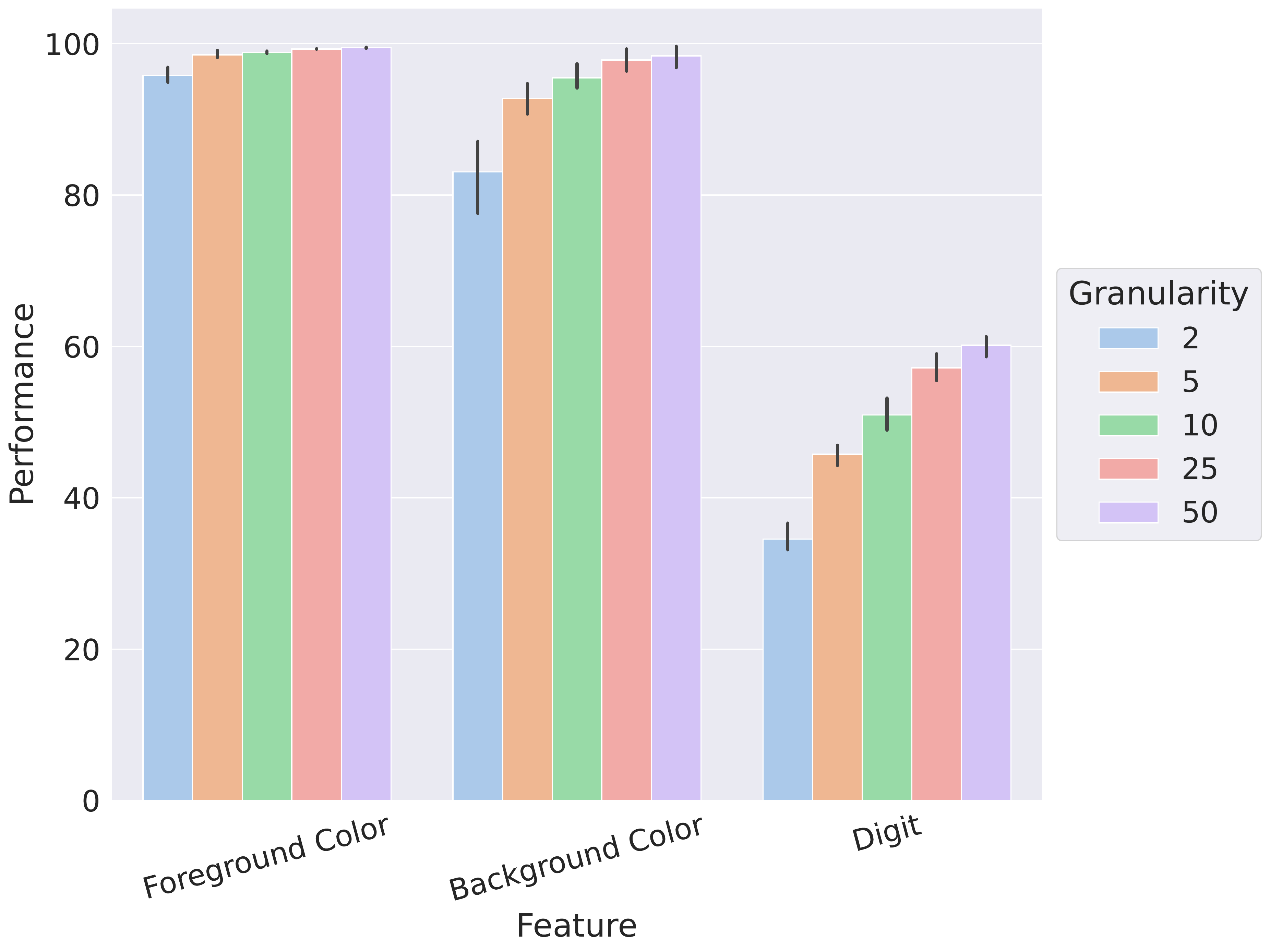}
\end{minipage}
\caption{Improvements of increased granularity (finer discretization schedule) on different features for the Synthetic Dataset (\textit{Left}) and the Colored-MNIST Dataset (\textit{Right}). The dimensionality of the latent space is heavily restricted to 2, and hence we see that with the coarsest discretization, performance on a number of features suffers.}
\label{fig:interp}
\vspace{-4mm}
\end{figure}

Figure \ref{fig:interp} highlights that for both the datasets, we see substantial improvement when using more points on the trajectory when the latent code is severely restricted, further signifying the benefit of the temporal unbounded-ness of the trajectory, and the monotonic improvement in performance with increase in granularity.

\vspace{-2mm}
\section{Conclusion}
\vspace{-2mm}
Through our analysis, we realize that the encoder $E_\phi(\cdot, t)$ actually learns different kinds of information at different time-steps $t$. Typically the mid-points of the trajectory are the most important for downstream classification tasks but we uncover that using as many points on the whole trajectory, i.e. increasingly finer discretization of an infinite-dimensional object, is much better than just singular points on it. What kind of semantic information is encoded in the different parts of the trajectories? Can we show some benefits of the unboundedness of the trajectory? Through our analysis, we provide insights into the differences of information stored along the trajectory, as well as the benefits of its unbouneded structure especially in the domain of restricted latent dimensionality.

While we highlight some interesting properties of these trajectory-based representations as well as the diversity of information over it, we believe that an important next step is to automate and learn the discretization process as opposed to the heuristic based uniform discretization. We believe this could lead to a variable computation system, where the downstream model would learn on its own which part of the trajectory should it sample more finely than others, when conditioned on the task/feature used.

This kind of task-conditioned discretization process would not only be able to use the whole trajectory information without heuristics but would also be able to leverage the structure which we show in our analysis in a more efficient and improved manner. We believe that this is an important direction to obtaining task and feature centric representations which are more general than the finite-sized representations afforded by contemporary representation learning models.
\clearpage
\section*{Acknowledgements}SM would like to acknowledge the support of UNIQUE and IVADO towards his research. GL acknowledges the support from Canada CIFAR AI Chair Program, Samsung SAIT, and NSERC Discovery Grant [RGPIN-2018-04821]. SB would like to thank the Berzelius cluster and the Swedish National Supercomputer Center for providing resources for the experiments done in the work.

\section*{Ethics Statement}
We do not foresee any negative or unethical implications of this work, which is in addition to the general impacts of advancement of Machine Learning and Representation Learning.

\section*{Reproducibility Statement}
We perform all the experiments with multiple seeds, ranging from 3 to 10 depending on the experiments. For each run of the score-based diffusion model, we also perform the downstream experiments with multiple seeds to obtain statistically significant results. We refer the readers to the implementation details outlined in Appendix \ref{apdx:impl} and we will be open-sourcing our code for ease of reproducibility.
\clearpage

\clearpage
\bibliography{iclr2022_conference}

\begin{thebibliography}{37}
\providecommand{\natexlab}[1]{#1}
\providecommand{\url}[1]{\texttt{#1}}
\expandafter\ifx\csname urlstyle\endcsname\relax
  \providecommand{\doi}[1]{doi: #1}\else
  \providecommand{\doi}{doi: \begingroup \urlstyle{rm}\Url}\fi

\bibitem[Abstreiter et~al.(2021)Abstreiter, Bauer, Sch{\"o}lkopf, and
  Mehrjou]{abstreiter2021diffusion}
Korbinian Abstreiter, Stefan Bauer, Bernhard Sch{\"o}lkopf, and Arash Mehrjou.
\newblock Diffusion-based representation learning.
\newblock \emph{arXiv preprint arXiv:2105.14257}, 2021.

\bibitem[Belghazi et~al.(2018)Belghazi, Baratin, Rajeshwar, Ozair, Bengio,
  Courville, and Hjelm]{belghazi2018mutual}
Mohamed~Ishmael Belghazi, Aristide Baratin, Sai Rajeshwar, Sherjil Ozair,
  Yoshua Bengio, Aaron Courville, and Devon Hjelm.
\newblock Mutual information neural estimation.
\newblock In \emph{International conference on machine learning}, pp.\
  531--540. PMLR, 2018.

\bibitem[Bengio et~al.(2013)Bengio, Courville, and
  Vincent]{bengio2013representation}
Yoshua Bengio, Aaron Courville, and Pascal Vincent.
\newblock Representation learning: A review and new perspectives.
\newblock \emph{IEEE transactions on pattern analysis and machine
  intelligence}, 35\penalty0 (8):\penalty0 1798--1828, 2013.

\bibitem[Bromley et~al.(1993)Bromley, Bentz, Bottou, Guyon, Lecun, Moore,
  Sackinger, and Shah]{articlesiamese}
Jane Bromley, James Bentz, Leon Bottou, Isabelle Guyon, Yann Lecun, Cliff
  Moore, Eduard Sackinger, and Rookpak Shah.
\newblock Signature verification using a "siamese" time delay neural network.
\newblock \emph{International Journal of Pattern Recognition and Artificial
  Intelligence}, 7:\penalty0 25, 08 1993.
\newblock \doi{10.1142/S0218001493000339}.

\bibitem[Cai et~al.(2020)Cai, Yang, Averbuch-Elor, Hao, Belongie, Snavely, and
  Hariharan]{cai2020learning}
Ruojin Cai, Guandao Yang, Hadar Averbuch-Elor, Zekun Hao, Serge Belongie, Noah
  Snavely, and Bharath Hariharan.
\newblock Learning gradient fields for shape generation, 2020.

\bibitem[Caron et~al.(2021)Caron, Misra, Mairal, Goyal, Bojanowski, and
  Joulin]{caron2021unsupervised}
Mathilde Caron, Ishan Misra, Julien Mairal, Priya Goyal, Piotr Bojanowski, and
  Armand Joulin.
\newblock Unsupervised learning of visual features by contrasting cluster
  assignments, 2021.

\bibitem[Chen et~al.(2020{\natexlab{a}})Chen, Zhang, Zen, Weiss, Norouzi, and
  Chan]{chen2020wavegrad}
Nanxin Chen, Yu~Zhang, Heiga Zen, Ron~J. Weiss, Mohammad Norouzi, and William
  Chan.
\newblock Wavegrad: Estimating gradients for waveform generation,
  2020{\natexlab{a}}.

\bibitem[Chen et~al.(2020{\natexlab{b}})Chen, Kornblith, Norouzi, and
  Hinton]{chen2020simple}
Ting Chen, Simon Kornblith, Mohammad Norouzi, and Geoffrey Hinton.
\newblock A simple framework for contrastive learning of visual
  representations.
\newblock In \emph{International conference on machine learning}, pp.\
  1597--1607. PMLR, 2020{\natexlab{b}}.

\bibitem[Chen et~al.(2020{\natexlab{c}})Chen, Kornblith, Swersky, Norouzi, and
  Hinton]{DBLP:journals/corr/abs-2006-10029}
Ting Chen, Simon Kornblith, Kevin Swersky, Mohammad Norouzi, and Geoffrey~E.
  Hinton.
\newblock Big self-supervised models are strong semi-supervised learners.
\newblock \emph{CoRR}, abs/2006.10029, 2020{\natexlab{c}}.
\newblock URL \url{https://arxiv.org/abs/2006.10029}.

\bibitem[Chen \& He(2020)Chen and He]{chen2020exploring}
Xinlei Chen and Kaiming He.
\newblock Exploring simple siamese representation learning, 2020.

\bibitem[Chen et~al.(2020{\natexlab{d}})Chen, Fan, Girshick, and
  He]{chen2020improved}
Xinlei Chen, Haoqi Fan, Ross Girshick, and Kaiming He.
\newblock Improved baselines with momentum contrastive learning.
\newblock \emph{arXiv preprint arXiv:2003.04297}, 2020{\natexlab{d}}.

\bibitem[Cho et~al.(2014)Cho, Van~Merri{\"e}nboer, Bahdanau, and
  Bengio]{cho2014properties}
Kyunghyun Cho, Bart Van~Merri{\"e}nboer, Dzmitry Bahdanau, and Yoshua Bengio.
\newblock On the properties of neural machine translation: Encoder-decoder
  approaches.
\newblock \emph{arXiv preprint arXiv:1409.1259}, 2014.

\bibitem[Cover(1999)]{cover1999elements}
Thomas~M Cover.
\newblock \emph{Elements of information theory}.
\newblock John Wiley \& Sons, 1999.

\bibitem[Dhariwal \& Nichol(2021)Dhariwal and Nichol]{dhariwal2021diffusion}
Prafulla Dhariwal and Alex Nichol.
\newblock Diffusion models beat gans on image synthesis, 2021.

\bibitem[Grill et~al.(2020)Grill, Strub, Altché, Tallec, Richemond,
  Buchatskaya, Doersch, Pires, Guo, Azar, Piot, Kavukcuoglu, Munos, and
  Valko]{grill2020bootstrap}
Jean-Bastien Grill, Florian Strub, Florent Altché, Corentin Tallec, Pierre~H.
  Richemond, Elena Buchatskaya, Carl Doersch, Bernardo~Avila Pires,
  Zhaohan~Daniel Guo, Mohammad~Gheshlaghi Azar, Bilal Piot, Koray Kavukcuoglu,
  Rémi Munos, and Michal Valko.
\newblock Bootstrap your own latent: A new approach to self-supervised
  learning, 2020.

\bibitem[Ho et~al.(2021)Ho, Saharia, Chan, Fleet, Norouzi, and
  Salimans]{ho2021cascaded}
Jonathan Ho, Chitwan Saharia, William Chan, David~J Fleet, Mohammad Norouzi,
  and Tim Salimans.
\newblock Cascaded diffusion models for high fidelity image generation.
\newblock \emph{arXiv preprint arXiv:2106.15282}, 2021.

\bibitem[Hochreiter \& Schmidhuber(1997)Hochreiter and
  Schmidhuber]{hochreiter1997long}
Sepp Hochreiter and J{\"u}rgen Schmidhuber.
\newblock Long short-term memory.
\newblock \emph{Neural computation}, 9\penalty0 (8):\penalty0 1735--1780, 1997.

\bibitem[Hyv{\"a}rinen \& Dayan(2005)Hyv{\"a}rinen and
  Dayan]{hyvarinen2005estimation}
Aapo Hyv{\"a}rinen and Peter Dayan.
\newblock Estimation of non-normalized statistical models by score matching.
\newblock \emph{Journal of Machine Learning Research}, 6\penalty0 (4), 2005.

\bibitem[Kingma \& Welling(2013)Kingma and Welling]{kingma2013auto}
Diederik~P Kingma and Max Welling.
\newblock Auto-encoding variational bayes.
\newblock \emph{arXiv preprint arXiv:1312.6114}, 2013.

\bibitem[Krizhevsky et~al.({\natexlab{a}})Krizhevsky, Nair, and
  Hinton]{cifar10}
Alex Krizhevsky, Vinod Nair, and Geoffrey Hinton.
\newblock Cifar-10 (canadian institute for advanced research).
\newblock {\natexlab{a}}.
\newblock URL \url{http://www.cs.toronto.edu/~kriz/cifar.html}.

\bibitem[Krizhevsky et~al.({\natexlab{b}})Krizhevsky, Nair, and
  Hinton]{cifar100}
Alex Krizhevsky, Vinod Nair, and Geoffrey Hinton.
\newblock Cifar-100 (canadian institute for advanced research).
\newblock {\natexlab{b}}.
\newblock URL \url{http://www.cs.toronto.edu/~kriz/cifar.html}.

\bibitem[Liu et~al.(2015)Liu, Luo, Wang, and Tang]{liu2015faceattributes}
Ziwei Liu, Ping Luo, Xiaogang Wang, and Xiaoou Tang.
\newblock Deep learning face attributes in the wild.
\newblock In \emph{Proceedings of International Conference on Computer Vision
  (ICCV)}, December 2015.

\bibitem[Luhman \& Luhman(2021)Luhman and Luhman]{luhman2021knowledge}
Eric Luhman and Troy Luhman.
\newblock Knowledge distillation in iterative generative models for improved
  sampling speed, 2021.

\bibitem[Mehrjou et~al.(2017)Mehrjou, Sch{\"o}lkopf, and
  Saremi]{mehrjou2017annealed}
Arash Mehrjou, Bernhard Sch{\"o}lkopf, and Saeed Saremi.
\newblock Annealed generative adversarial networks.
\newblock \emph{arXiv preprint arXiv:1705.07505}, 2017.

\bibitem[Niu et~al.(2020)Niu, Song, Song, Zhao, Grover, and
  Ermon]{niu2020permutation}
Chenhao Niu, Yang Song, Jiaming Song, Shengjia Zhao, Aditya Grover, and Stefano
  Ermon.
\newblock Permutation invariant graph generation via score-based generative
  modeling, 2020.

\bibitem[Preechakul et~al.(2022)Preechakul, Chatthee, Wizadwongsa, and
  Suwajanakorn]{preechakul2021diffusion}
Konpat Preechakul, Nattanat Chatthee, Suttisak Wizadwongsa, and Supasorn
  Suwajanakorn.
\newblock Diffusion autoencoders: Toward a meaningful and decodable
  representation.
\newblock In \emph{IEEE Conference on Computer Vision and Pattern Recognition
  (CVPR)}, 2022.

\bibitem[Rezende et~al.(2014)Rezende, Mohamed, and
  Wierstra]{rezende2014stochastic}
Danilo~Jimenez Rezende, Shakir Mohamed, and Daan Wierstra.
\newblock Stochastic backpropagation and approximate inference in deep
  generative models.
\newblock In \emph{International conference on machine learning}, pp.\
  1278--1286. PMLR, 2014.

\bibitem[Sajjadi et~al.(2018)Sajjadi, Parascandolo, Mehrjou, and
  Sch{\"o}lkopf]{sajjadi2018tempered}
Mehdi~SM Sajjadi, Giambattista Parascandolo, Arash Mehrjou, and Bernhard
  Sch{\"o}lkopf.
\newblock Tempered adversarial networks.
\newblock In \emph{International Conference on Machine Learning}, pp.\
  4451--4459. PMLR, 2018.

\bibitem[Saremi et~al.(2018)Saremi, Mehrjou, Sch{\"o}lkopf, and
  Hyv{\"a}rinen]{saremi2018deep}
Saeed Saremi, Arash Mehrjou, Bernhard Sch{\"o}lkopf, and Aapo Hyv{\"a}rinen.
\newblock Deep energy estimator networks.
\newblock \emph{arXiv preprint arXiv:1805.08306}, 2018.

\bibitem[Sohl-Dickstein et~al.(2015)Sohl-Dickstein, Weiss, Maheswaranathan, and
  Ganguli]{sohldickstein2015deep}
Jascha Sohl-Dickstein, Eric~A. Weiss, Niru Maheswaranathan, and Surya Ganguli.
\newblock Deep unsupervised learning using nonequilibrium thermodynamics, 2015.

\bibitem[Song et~al.(2020)Song, Meng, and Ermon]{song2020denoising}
Jiaming Song, Chenlin Meng, and Stefano Ermon.
\newblock Denoising diffusion implicit models, 2020.

\bibitem[Song et~al.(2021)Song, Sohl-Dickstein, Kingma, Kumar, Ermon, and
  Poole]{song2021scorebased}
Yang Song, Jascha Sohl-Dickstein, Diederik~P. Kingma, Abhishek Kumar, Stefano
  Ermon, and Ben Poole.
\newblock Score-based generative modeling through stochastic differential
  equations, 2021.

\bibitem[Vaswani et~al.(2017)Vaswani, Shazeer, Parmar, Uszkoreit, Jones, Gomez,
  Kaiser, and Polosukhin]{vaswani2017attention}
Ashish Vaswani, Noam Shazeer, Niki Parmar, Jakob Uszkoreit, Llion Jones,
  Aidan~N Gomez, {\L}ukasz Kaiser, and Illia Polosukhin.
\newblock Attention is all you need.
\newblock \emph{Advances in neural information processing systems}, 30, 2017.

\bibitem[Vincent(2011)]{vincent}
Pascal Vincent.
\newblock A connection between score matching and denoising autoencoders.
\newblock \emph{Neural Computation}, 23\penalty0 (7):\penalty0 1661--1674,
  2011.
\newblock \doi{10.1162/NECO_a_00142}.

\bibitem[Vincent et~al.(2010)Vincent, Larochelle, Lajoie, Bengio, Manzagol, and
  Bottou]{vincent2010stacked}
Pascal Vincent, Hugo Larochelle, Isabelle Lajoie, Yoshua Bengio, Pierre-Antoine
  Manzagol, and L{\'e}on Bottou.
\newblock Stacked denoising autoencoders: Learning useful representations in a
  deep network with a local denoising criterion.
\newblock \emph{Journal of machine learning research}, 11\penalty0 (12), 2010.

\bibitem[Vinyals et~al.(2016)Vinyals, Blundell, Lillicrap, Kavukcuoglu, and
  Wierstra]{DBLP:journals/corr/VinyalsBLKW16}
Oriol Vinyals, Charles Blundell, Timothy~P. Lillicrap, Koray Kavukcuoglu, and
  Daan Wierstra.
\newblock Matching networks for one shot learning.
\newblock \emph{CoRR}, abs/1606.04080, 2016.
\newblock URL \url{http://arxiv.org/abs/1606.04080}.

\bibitem[Zagoruyko \& Komodakis(2016)Zagoruyko and
  Komodakis]{zagoruyko2016wide}
Sergey Zagoruyko and Nikos Komodakis.
\newblock Wide residual networks.
\newblock \emph{arXiv preprint arXiv:1605.07146}, 2016.

\end{thebibliography}
\bibliographystyle{iclr2022_conference}

\clearpage
\appendix
\section*{Appendix}

\section{Implementation Details}
\label{apdx:impl}

\textbf{Score Model:} We use the implementation of the score model from \cite{song2021scorebased} using the variance-exploding configuration, in particular the CIFAR10 configuration provided on their codebase. We augment the score-model with an Encoder $E_\phi$ which is implemented as the Wide-ResNet architecture \citep{zagoruyko2016wide} that maps the input with time embeddings to a vector in $\mathbb{R}^d$, where $d$ is the dimensionality of the latent space and is set to 128 unless otherwise specified. The time embeddings for the encoder model are implemented in the exact same way as for the score inputs, as outlined in \cite{song2021scorebased}. We use the learning rate of $2 \times 10^{-4}$ to optimize the score network.

\textbf{Downstream Model:} For Multi-Layer-Perceptron (MLP) based Classification model, we consider a network with a single hidden layer, ReLU activation function, and 512 neurons. For the Recurrent Neural Network (RNN) model, we use a GRU with 256 hidden units and for the transformer system, we use a Multi-Head attention system with 4 heads and two layers, with weight sharing between the layers. For our attention profile based analysis settings, we consider the same Multi-Head attention system but only use a single layer instead of two, as it allows to make the score (averaged over heads) more interpretable.

We train all the downstream models with dropout of 0.25 and perform hyperparameter optimization for the learning rate over the set \{$0.001, 0.00075, 0.0005, 0.00025, 0.0001, 0.00005$\}. In particular, we found the hyperparameter optimization important when considering the granularity analysis.

\section{CIFAR10, CIFAR100 and Mini-ImageNet}
\label{apdx:cifar10}
We train the score model for 70,000 iterations and then the downstream models for 100 epochs. For the performance of the models, we use a $2-$layered Transformer model while for attention score profiles, we use a single layered Transformer model.

\section{Synthetic}
\label{apdx:syn}
\begin{wrapfigure}{r}{0.2\textwidth}
    \vspace{-5mm}
    \includegraphics[width=0.2\textwidth]{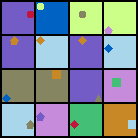}
    \caption{Samples from Synthetic Dataset}
    \label{fig:syn_samples}
    \vspace{-10mm}
\end{wrapfigure}
We train the score model for 250,000 iterations and then the downstream models for 1500 epochs. Figure \ref{fig:syn_samples} shows some samples obtained from this dataset, showcasing the different features present as well as the diversity of these different features.

We additionally perform the Jensen-Shannon Divergence analysis between different features for different granularities, as well as visualize the attention score profiles for the different granularities as well. Furthermore, we do the same analysis with both the types of encoders; VDRL and DRL.

\begin{figure}
\begin{subfigure}[c]{\textwidth}
\begin{subfigure}[c]{0.24\textwidth}
\includegraphics[width=\textwidth]{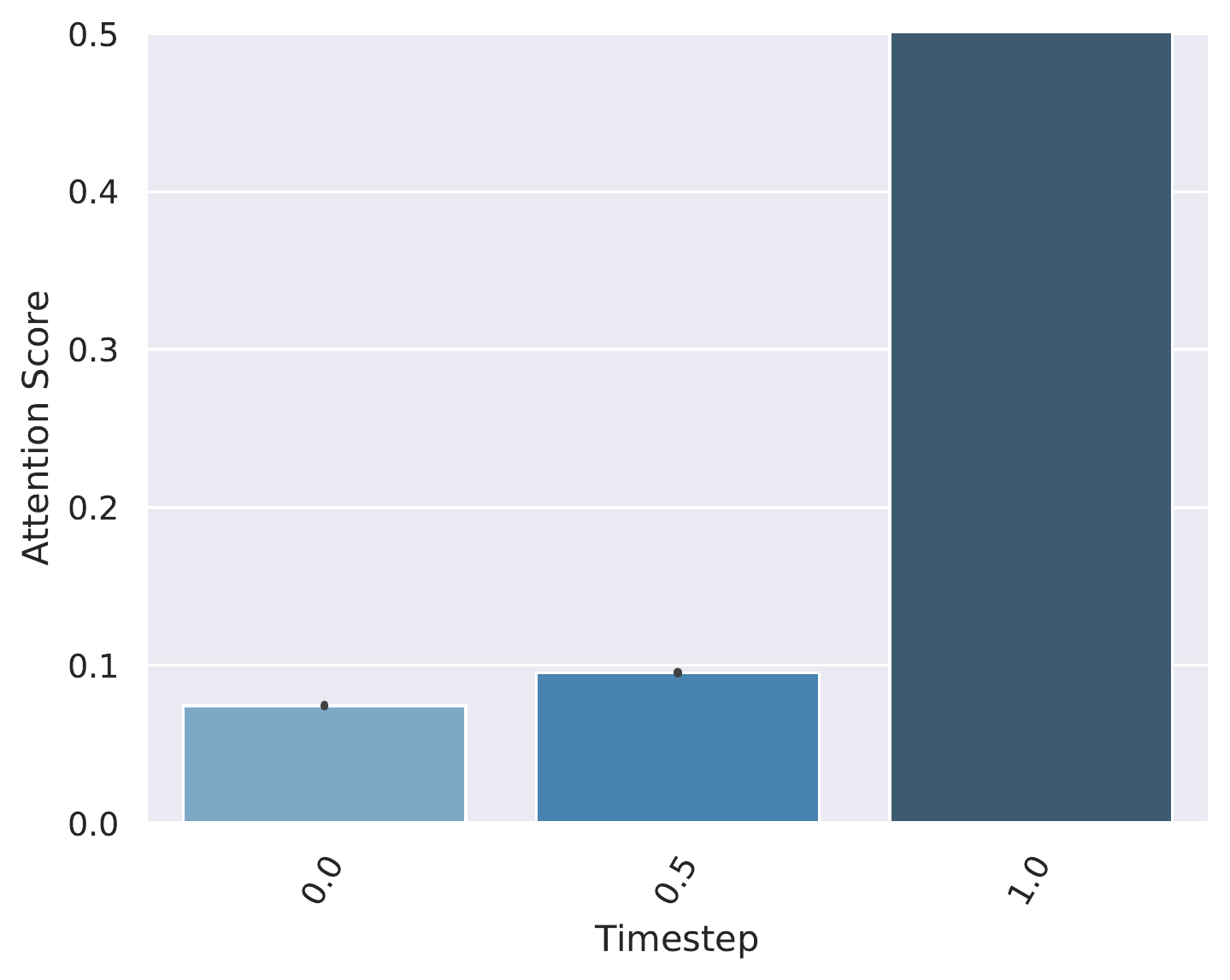}
\vspace{-6mm}
\subcaption*{\scriptsize Background Color}
\end{subfigure}
\begin{subfigure}[c]{0.24\textwidth}
\includegraphics[width=\textwidth]{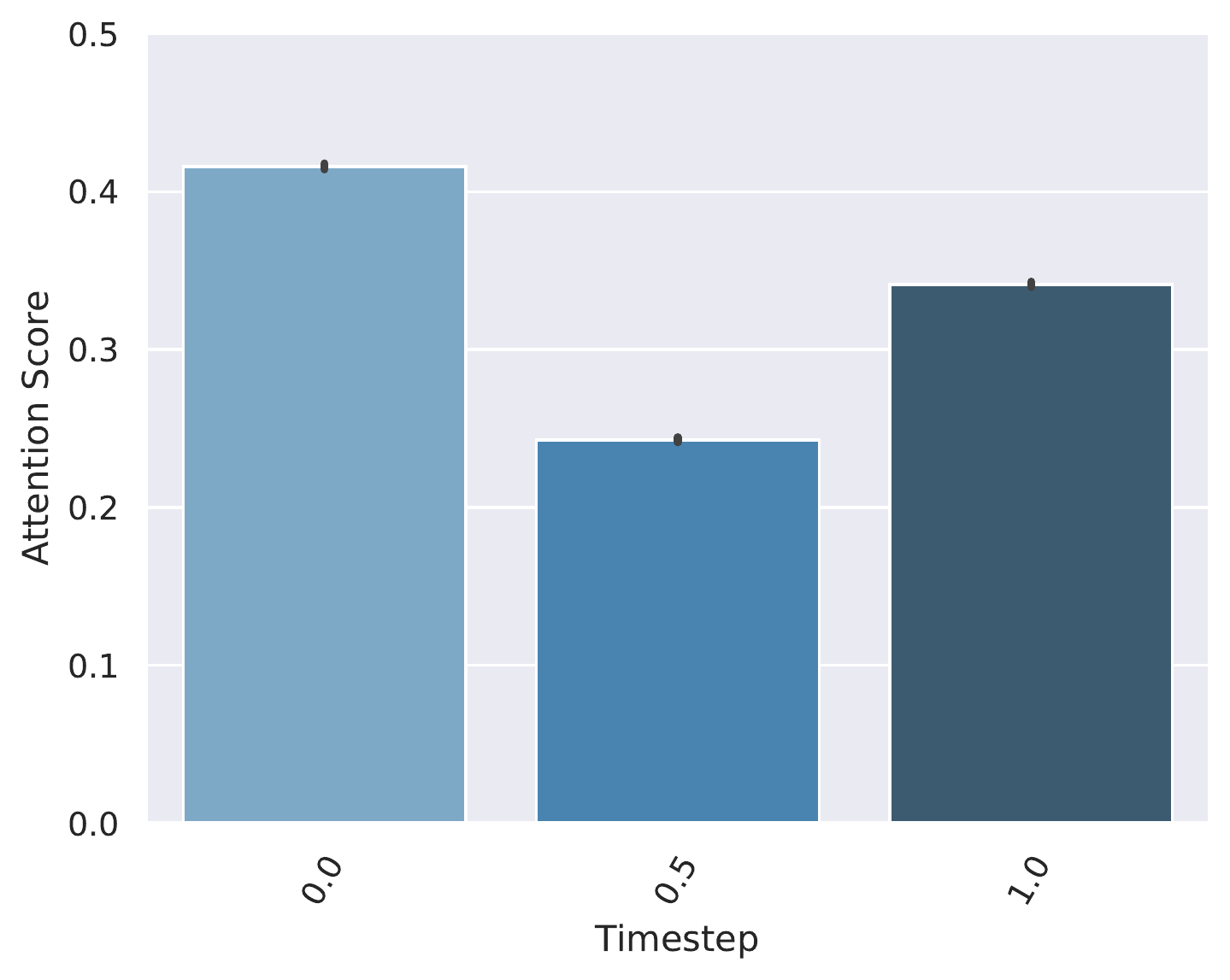}
\vspace{-6mm}
\subcaption*{\scriptsize Foreground Color}
\end{subfigure}
\begin{subfigure}[c]{0.24\textwidth}
\includegraphics[width=\textwidth]{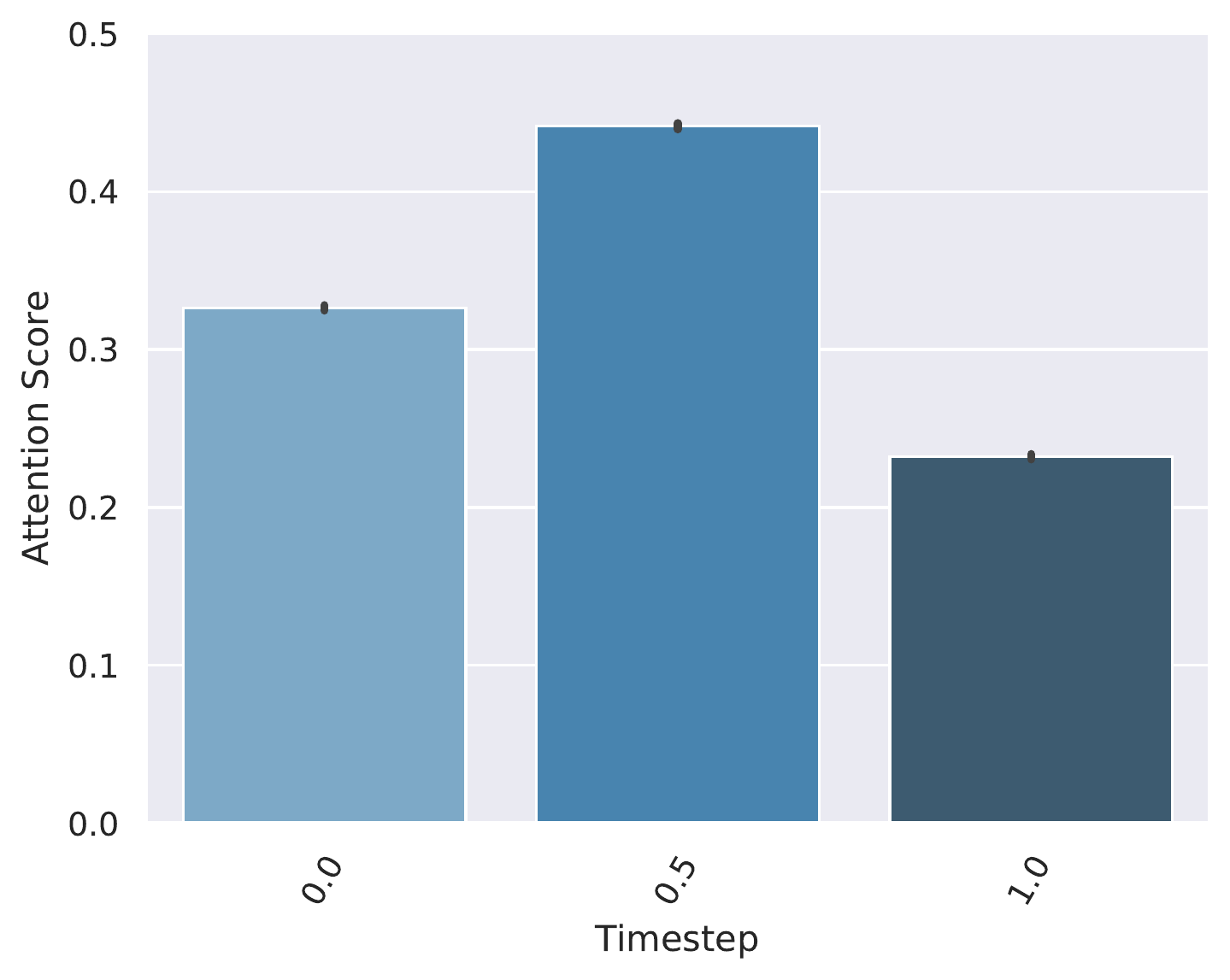}
\vspace{-6mm}
\subcaption*{\scriptsize Location}
\end{subfigure}
\begin{subfigure}[c]{0.24\textwidth}
\includegraphics[width=\textwidth]{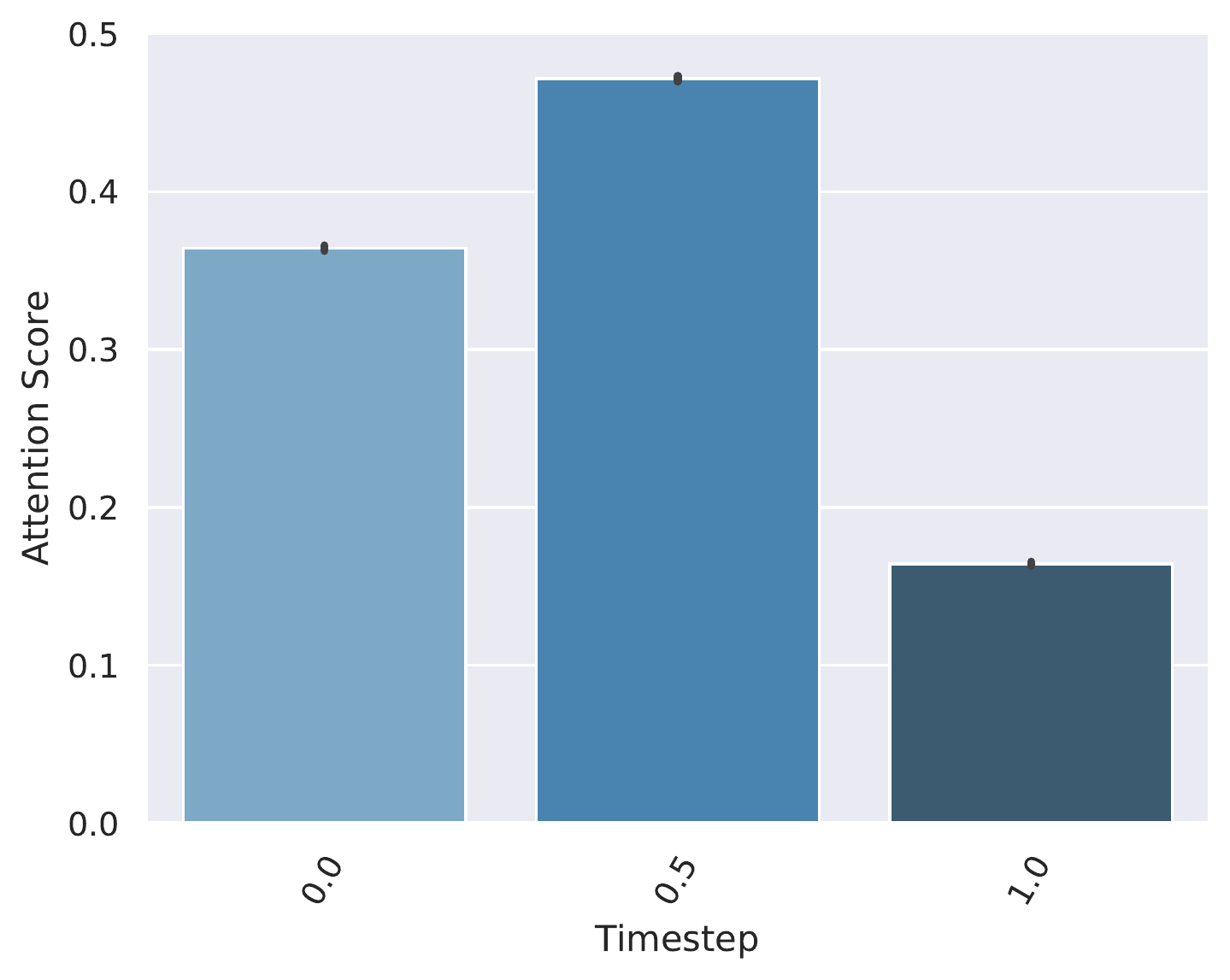}
\vspace{-6mm}
\subcaption*{\scriptsize Object Shape}
\end{subfigure}
\subcaption*{Granularity: 2}
\end{subfigure} \\
\begin{subfigure}[c]{\textwidth}
\begin{subfigure}[c]{0.24\textwidth}
\includegraphics[width=\textwidth]{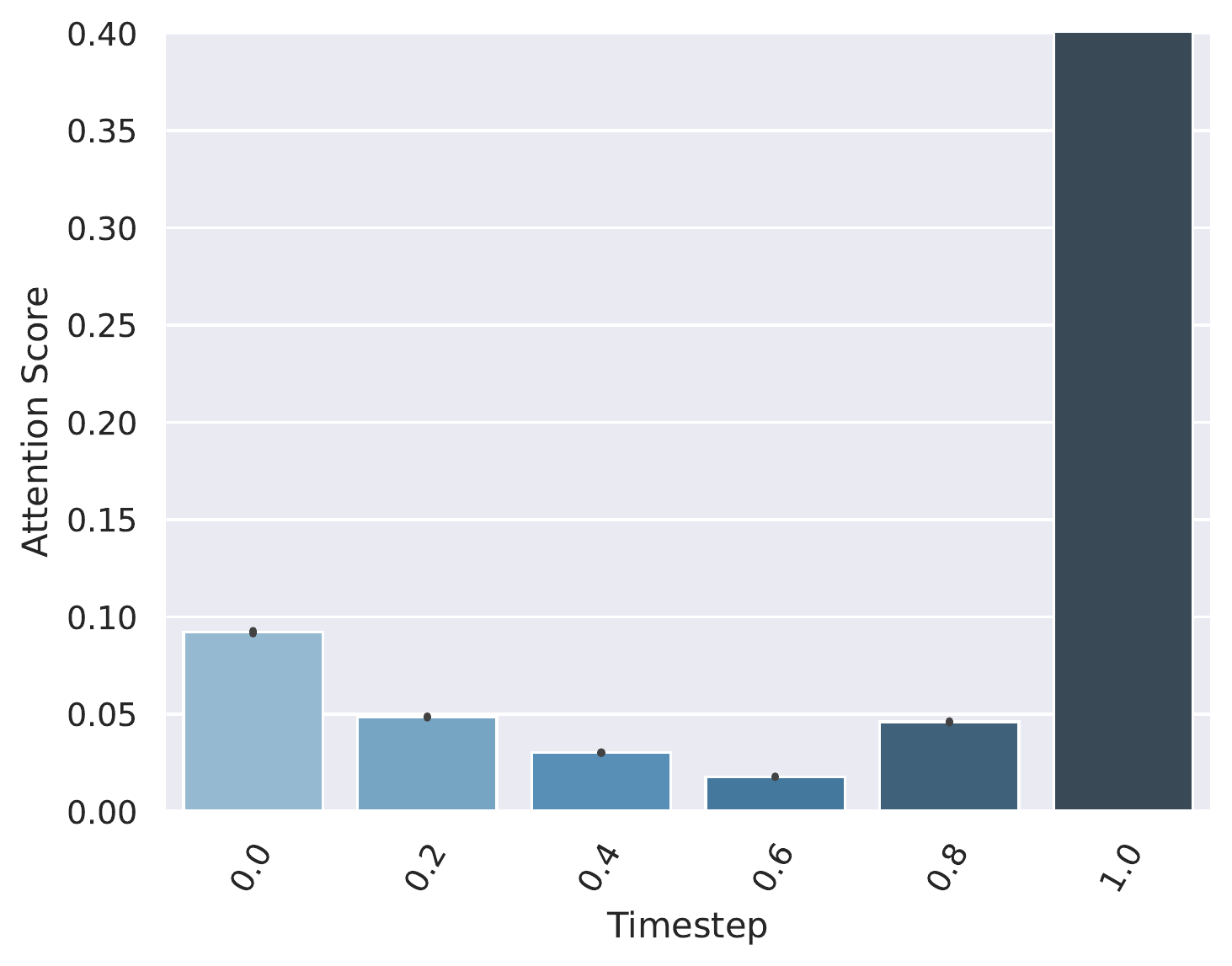}
\vspace{-6mm}
\subcaption*{\scriptsize Background Color}
\end{subfigure}
\begin{subfigure}[c]{0.24\textwidth}
\includegraphics[width=\textwidth]{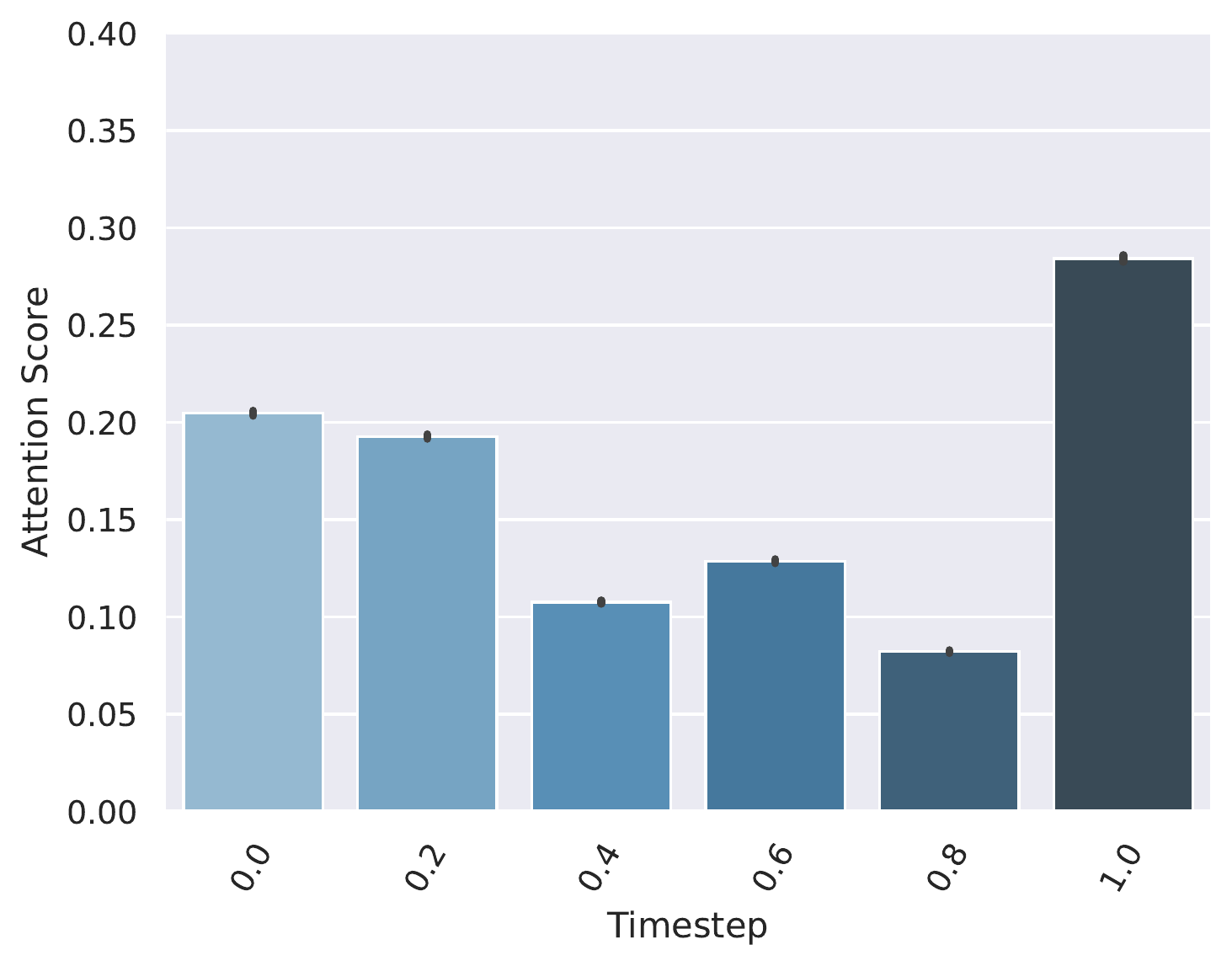}
\vspace{-6mm}
\subcaption*{\scriptsize Foreground Color}
\end{subfigure}
\begin{subfigure}[c]{0.24\textwidth}
\includegraphics[width=\textwidth]{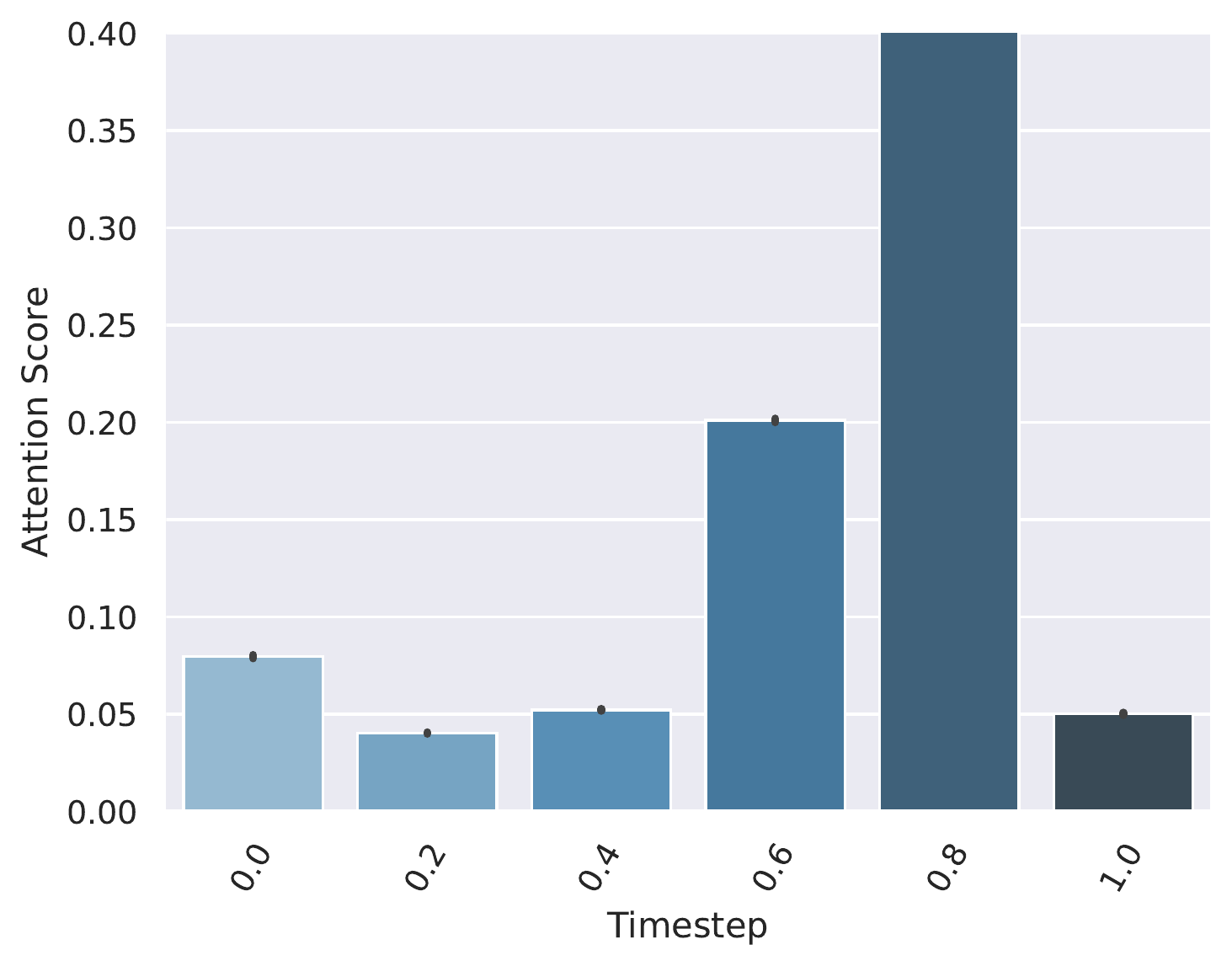}
\vspace{-6mm}
\subcaption*{\scriptsize Location}
\end{subfigure}
\begin{subfigure}[c]{0.24\textwidth}
\includegraphics[width=\textwidth]{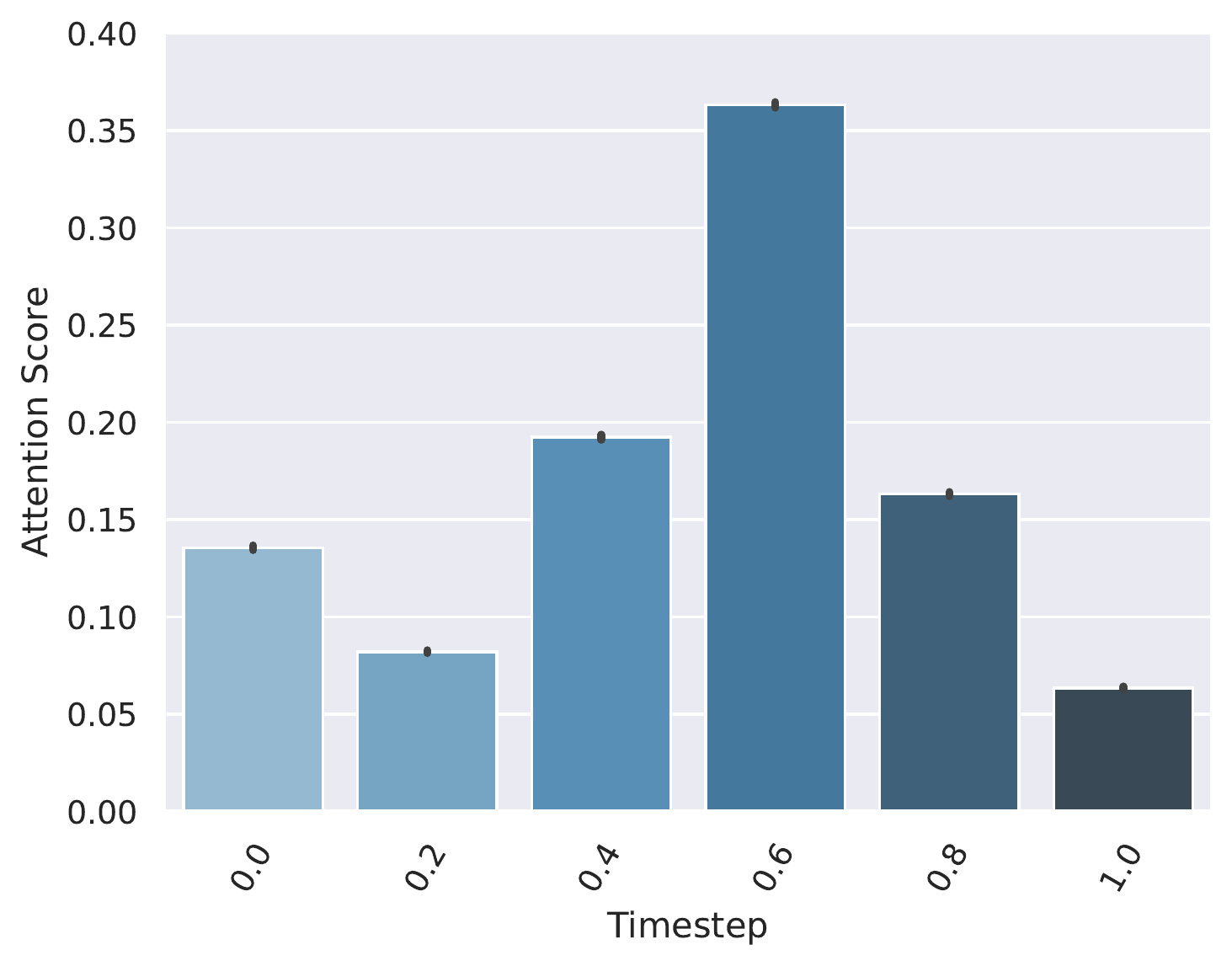}
\vspace{-6mm}
\subcaption*{\scriptsize Object Shape}
\end{subfigure}
\subcaption*{Granularity: 5}
\end{subfigure} \\
\begin{subfigure}[c]{\textwidth}
\begin{subfigure}[c]{0.24\textwidth}
\includegraphics[width=\textwidth]{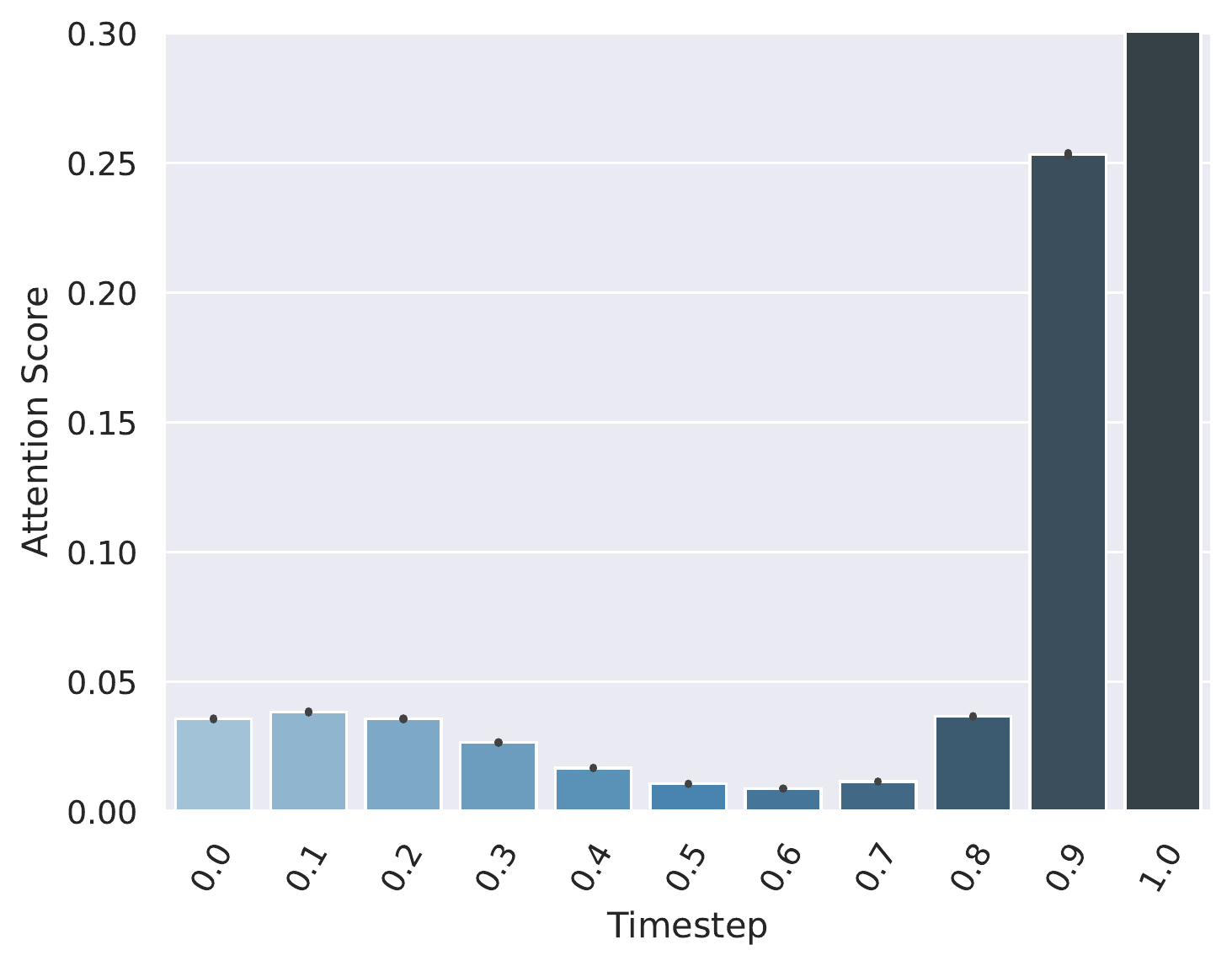}
\vspace{-6mm}
\subcaption*{\scriptsize Background Color}
\end{subfigure}
\begin{subfigure}[c]{0.24\textwidth}
\includegraphics[width=\textwidth]{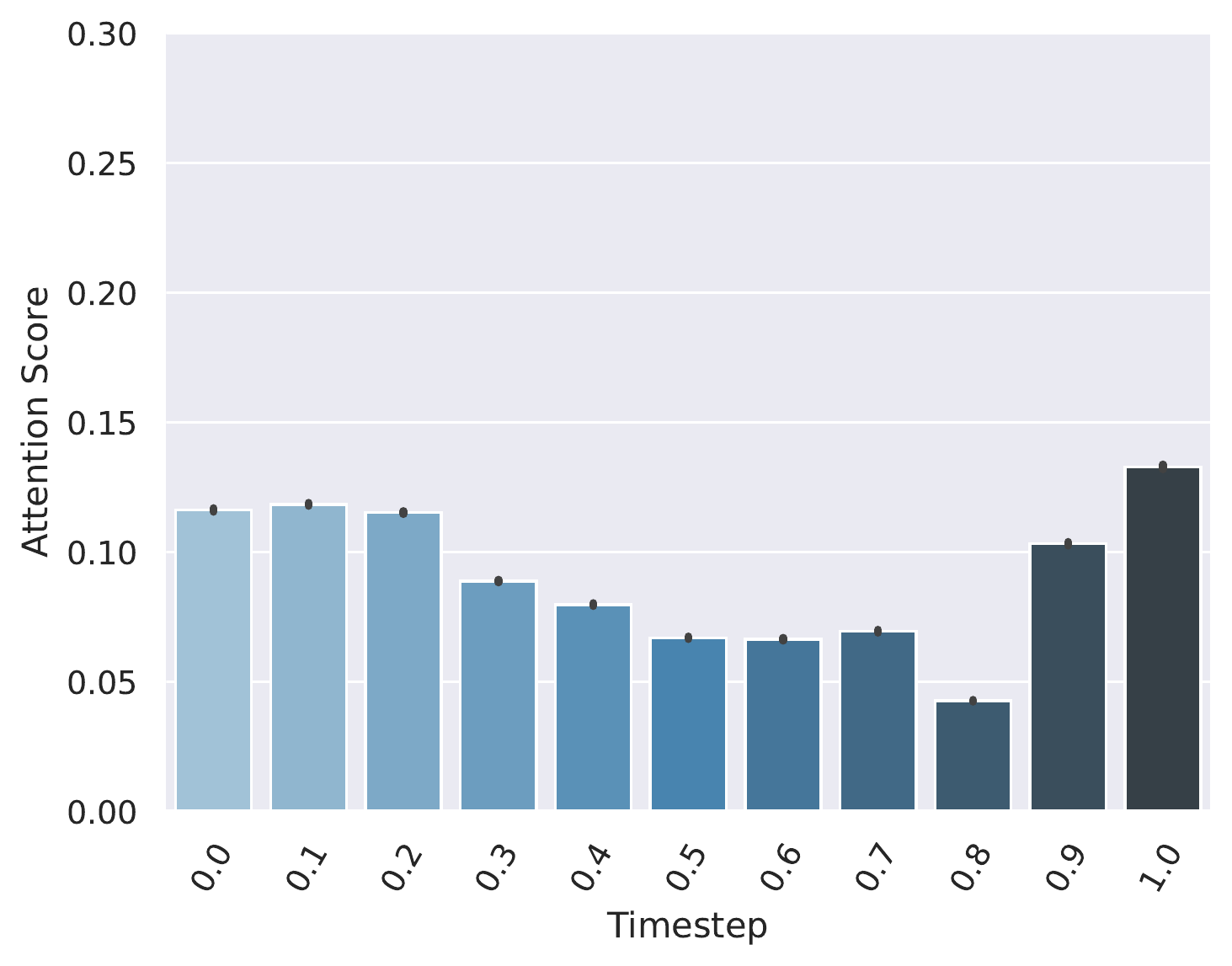}
\vspace{-6mm}
\subcaption*{\scriptsize Foreground Color}
\end{subfigure}
\begin{subfigure}[c]{0.24\textwidth}
\includegraphics[width=\textwidth]{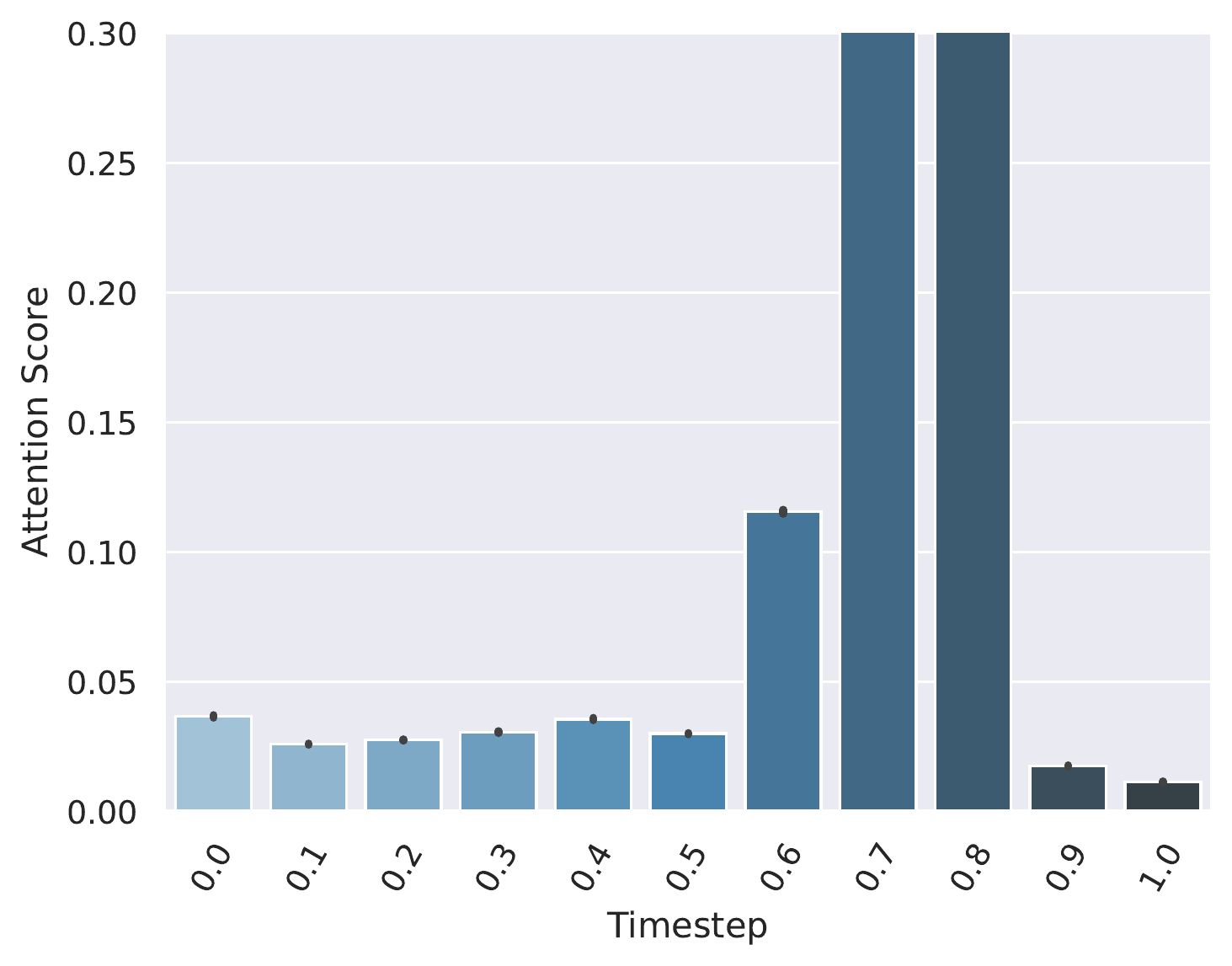}
\vspace{-6mm}
\subcaption*{\scriptsize Location}
\end{subfigure}
\begin{subfigure}[c]{0.24\textwidth}
\includegraphics[width=\textwidth]{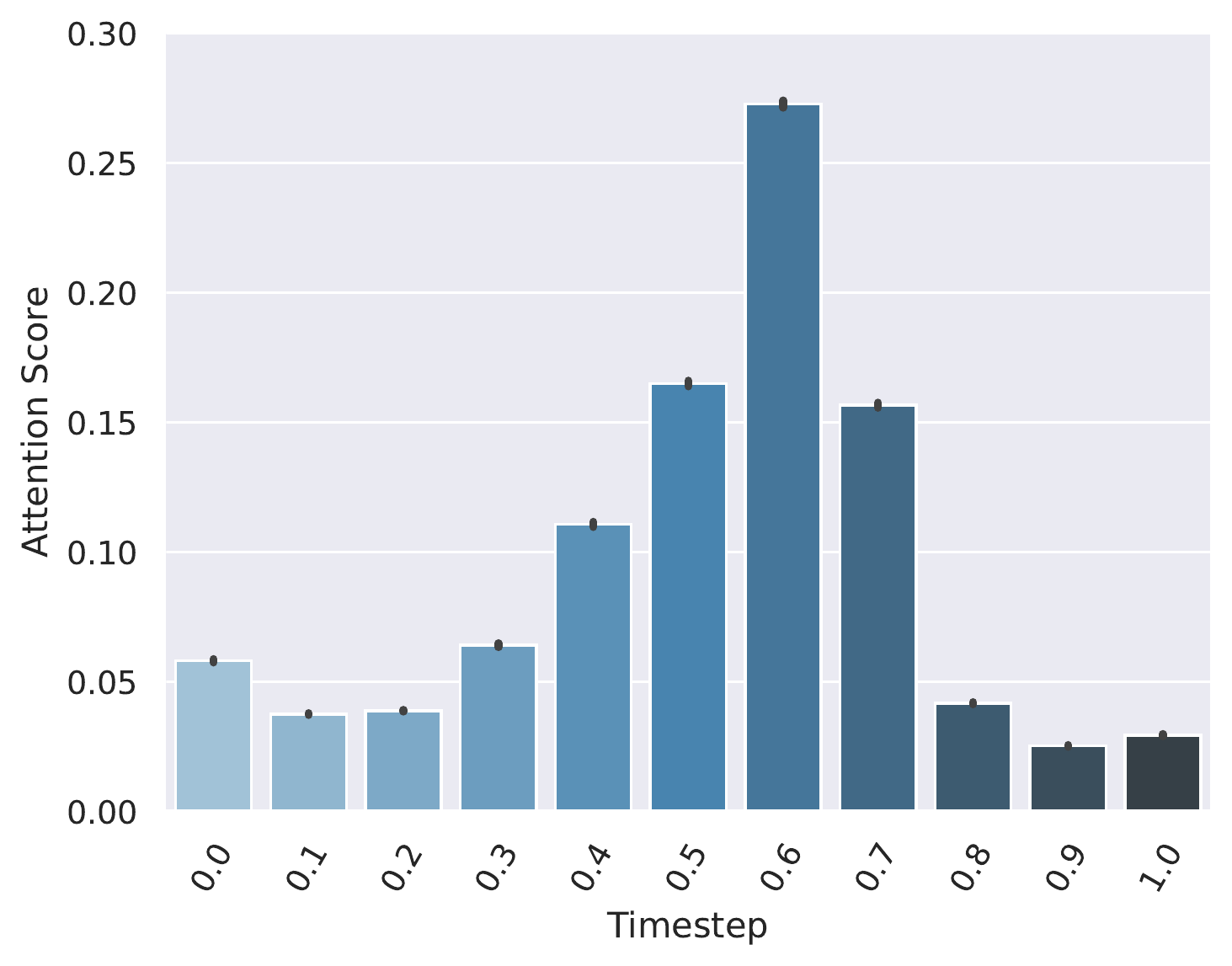}
\vspace{-6mm}
\subcaption*{\scriptsize Object Shape}
\end{subfigure}
\subcaption*{Granularity: 10}
\end{subfigure} \\
\begin{subfigure}[c]{\textwidth}
\begin{subfigure}[c]{0.24\textwidth}
\includegraphics[width=\textwidth]{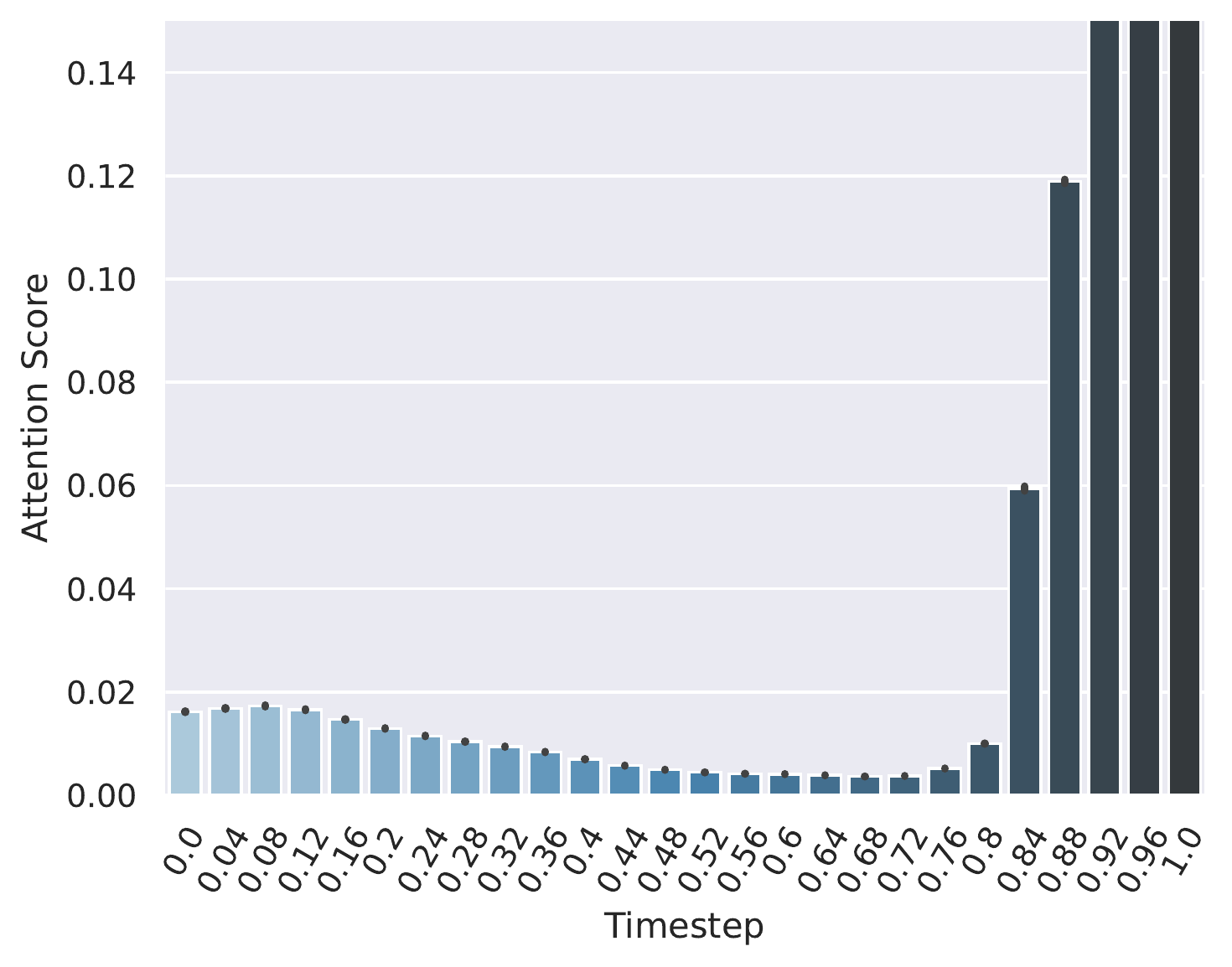}
\vspace{-6mm}
\subcaption*{\scriptsize Background Color}
\end{subfigure}
\begin{subfigure}[c]{0.24\textwidth}
\includegraphics[width=\textwidth]{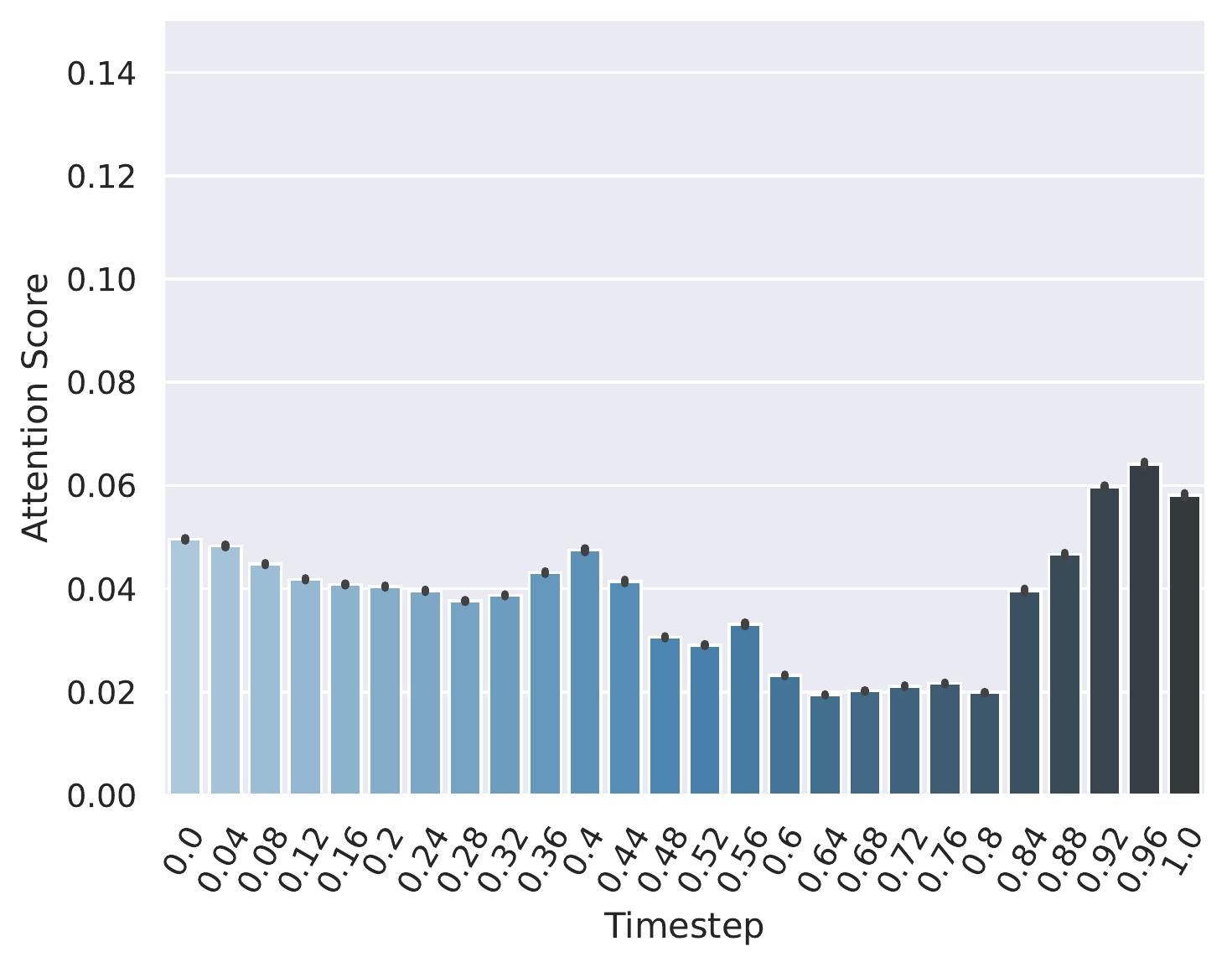}
\vspace{-6mm}
\subcaption*{\scriptsize Foreground Color}
\end{subfigure}
\begin{subfigure}[c]{0.24\textwidth}
\includegraphics[width=\textwidth]{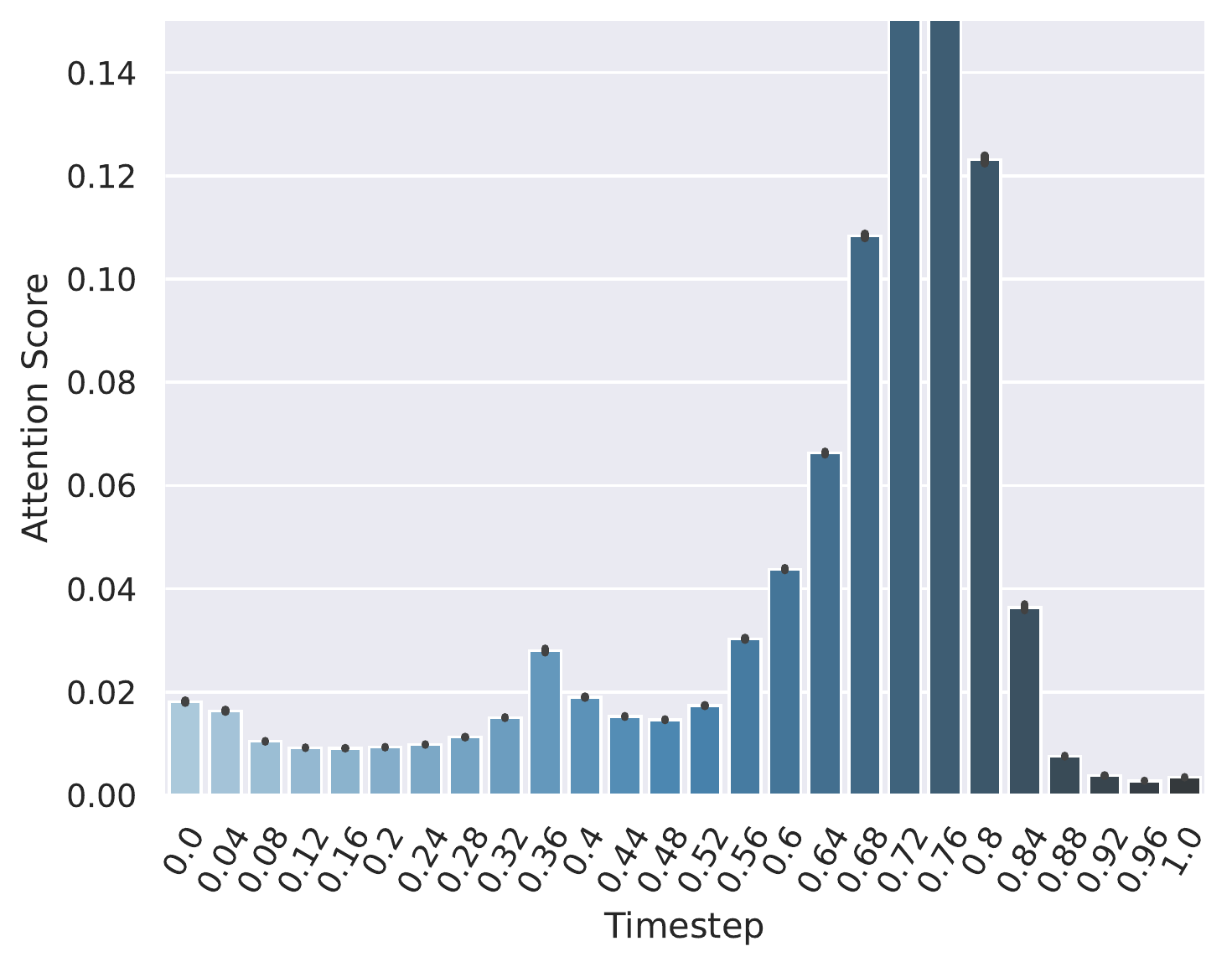}
\vspace{-6mm}
\subcaption*{\scriptsize Location}
\end{subfigure}
\begin{subfigure}[c]{0.24\textwidth}
\includegraphics[width=\textwidth]{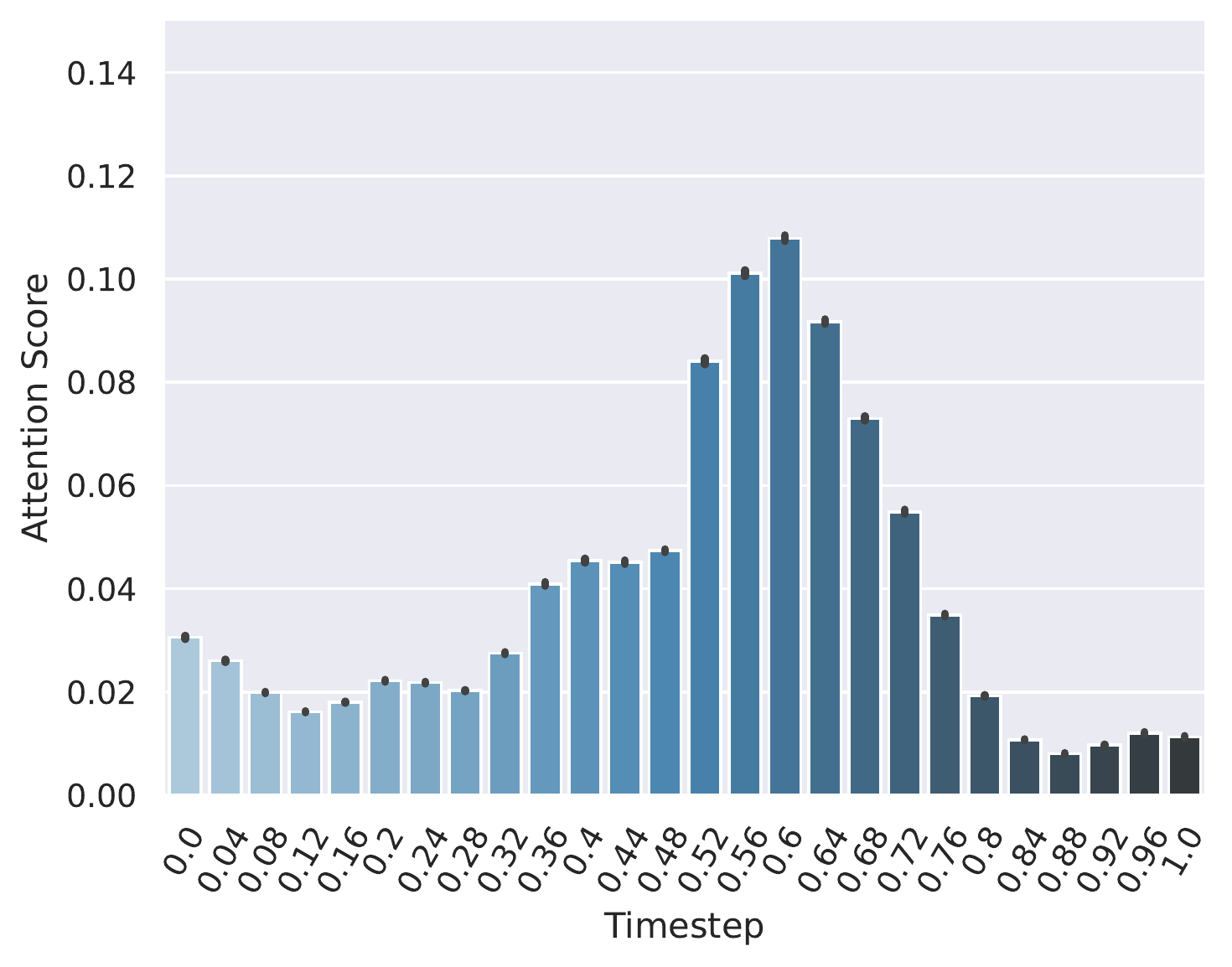}
\vspace{-6mm}
\subcaption*{\scriptsize Object Shape}
\end{subfigure}
\subcaption*{Granularity: 25}
\end{subfigure} \\
\caption{Attention score profiles for the synthetic dataset on the different features, using different granularities, with the dimensionality of the latent space as 2 and the VDRL encoder.}
\label{fig:syn_VDRL_2}
\end{figure}

\begin{figure}
\begin{subfigure}[c]{\textwidth}
\begin{subfigure}[c]{0.24\textwidth}
\includegraphics[width=\textwidth]{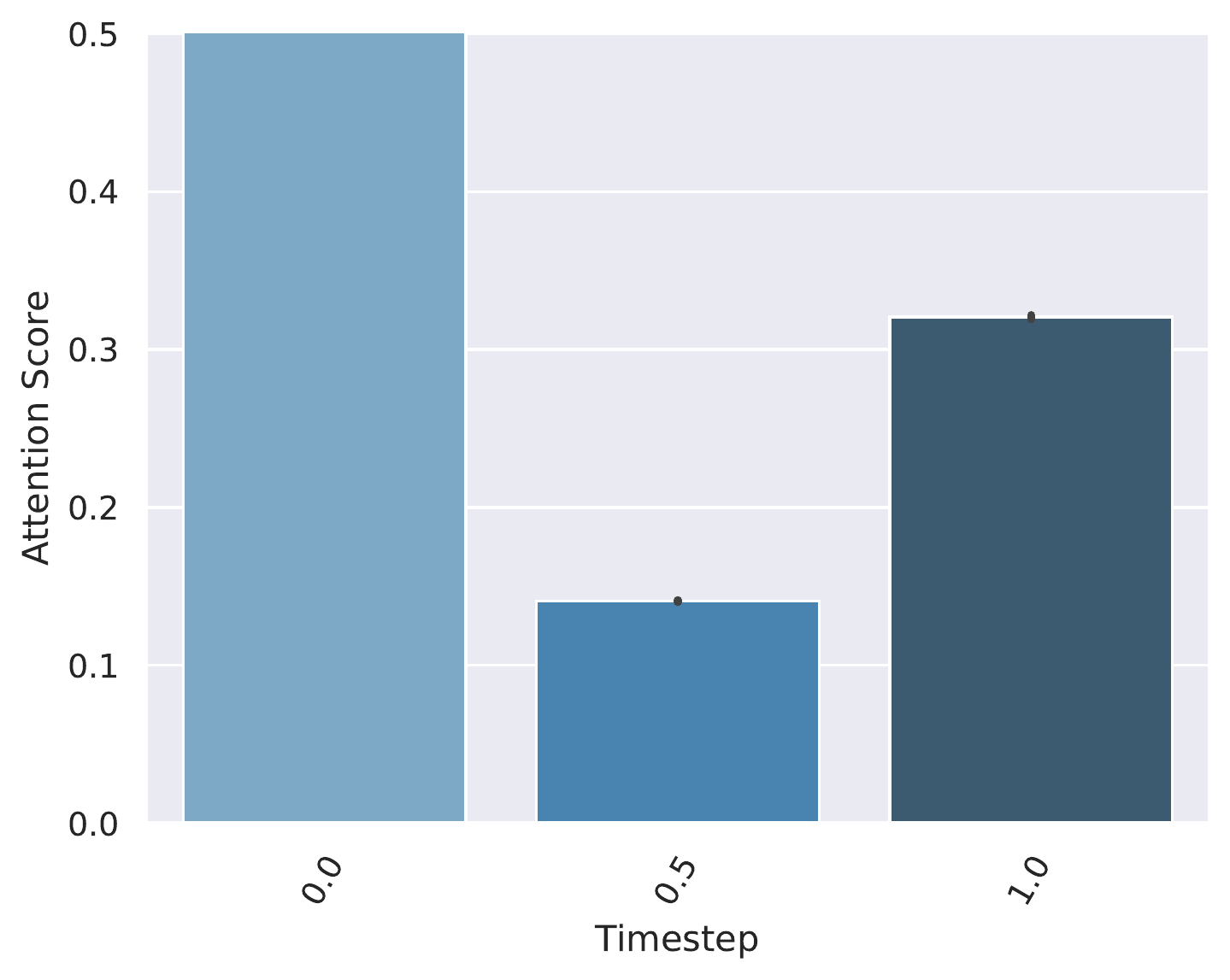}
\vspace{-6mm}
\subcaption*{\scriptsize Background Color}
\end{subfigure}
\begin{subfigure}[c]{0.24\textwidth}
\includegraphics[width=\textwidth]{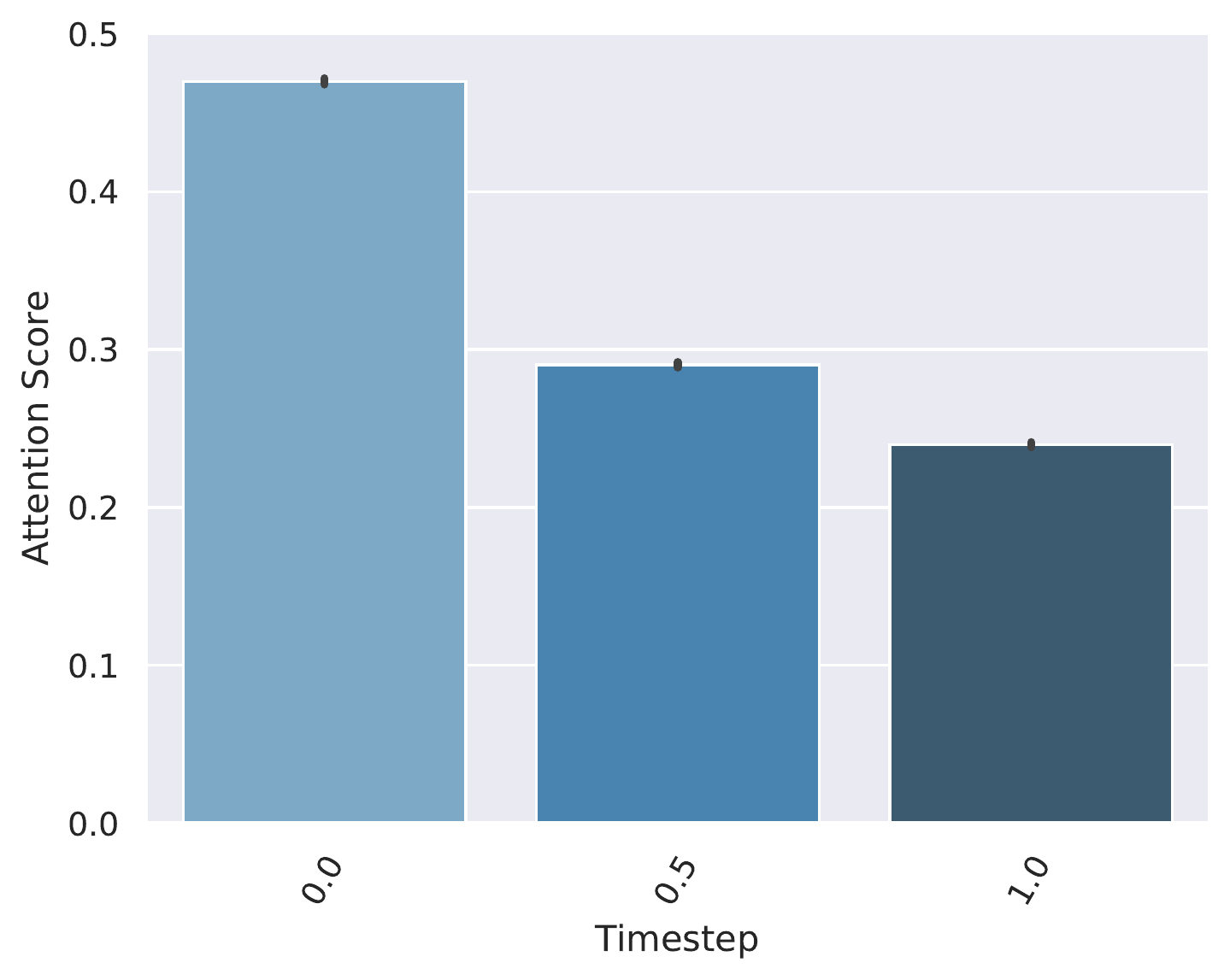}
\vspace{-6mm}
\subcaption*{\scriptsize Foreground Color}
\end{subfigure}
\begin{subfigure}[c]{0.24\textwidth}
\includegraphics[width=\textwidth]{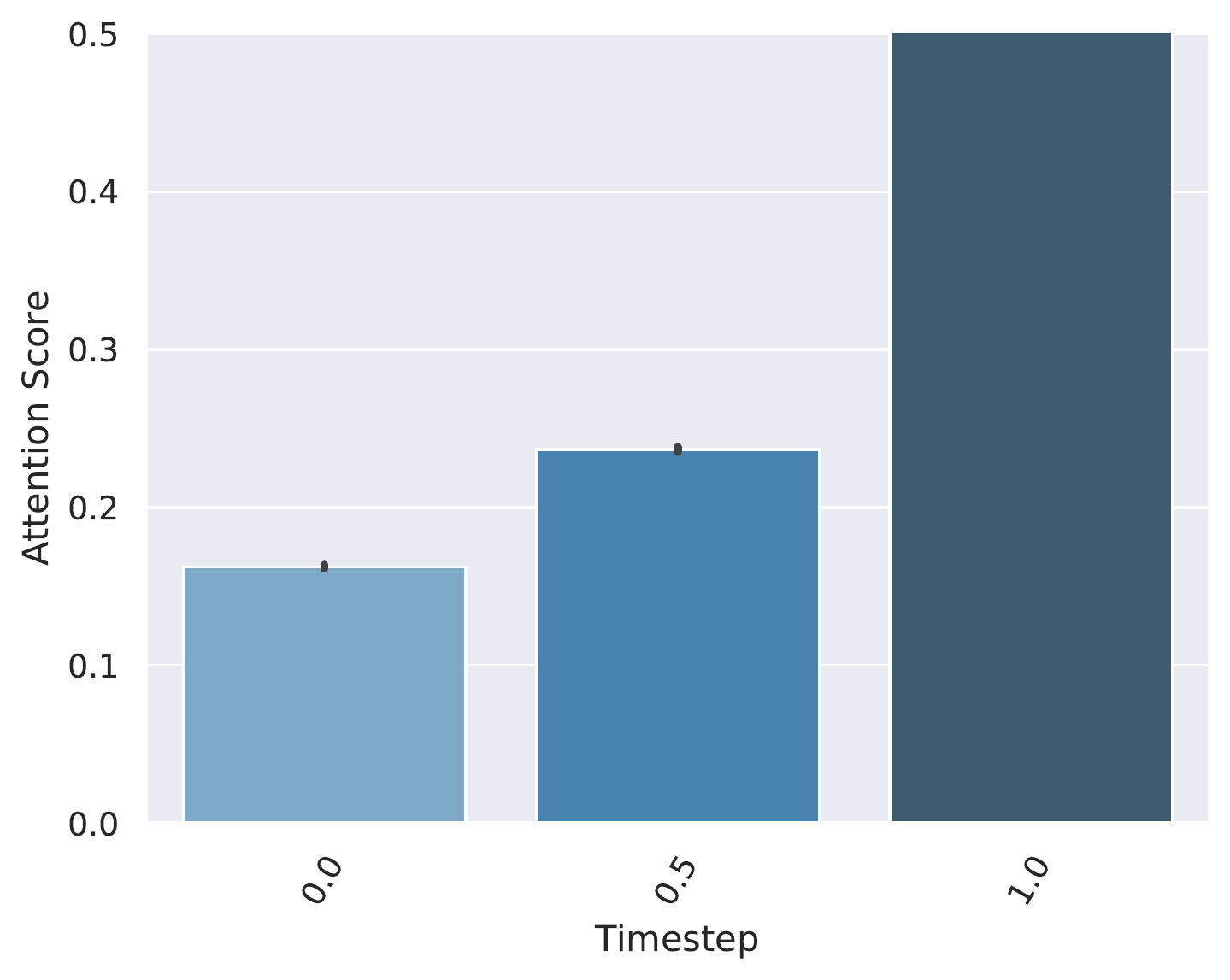}
\vspace{-6mm}
\subcaption*{\scriptsize Location}
\end{subfigure}
\begin{subfigure}[c]{0.24\textwidth}
\includegraphics[width=\textwidth]{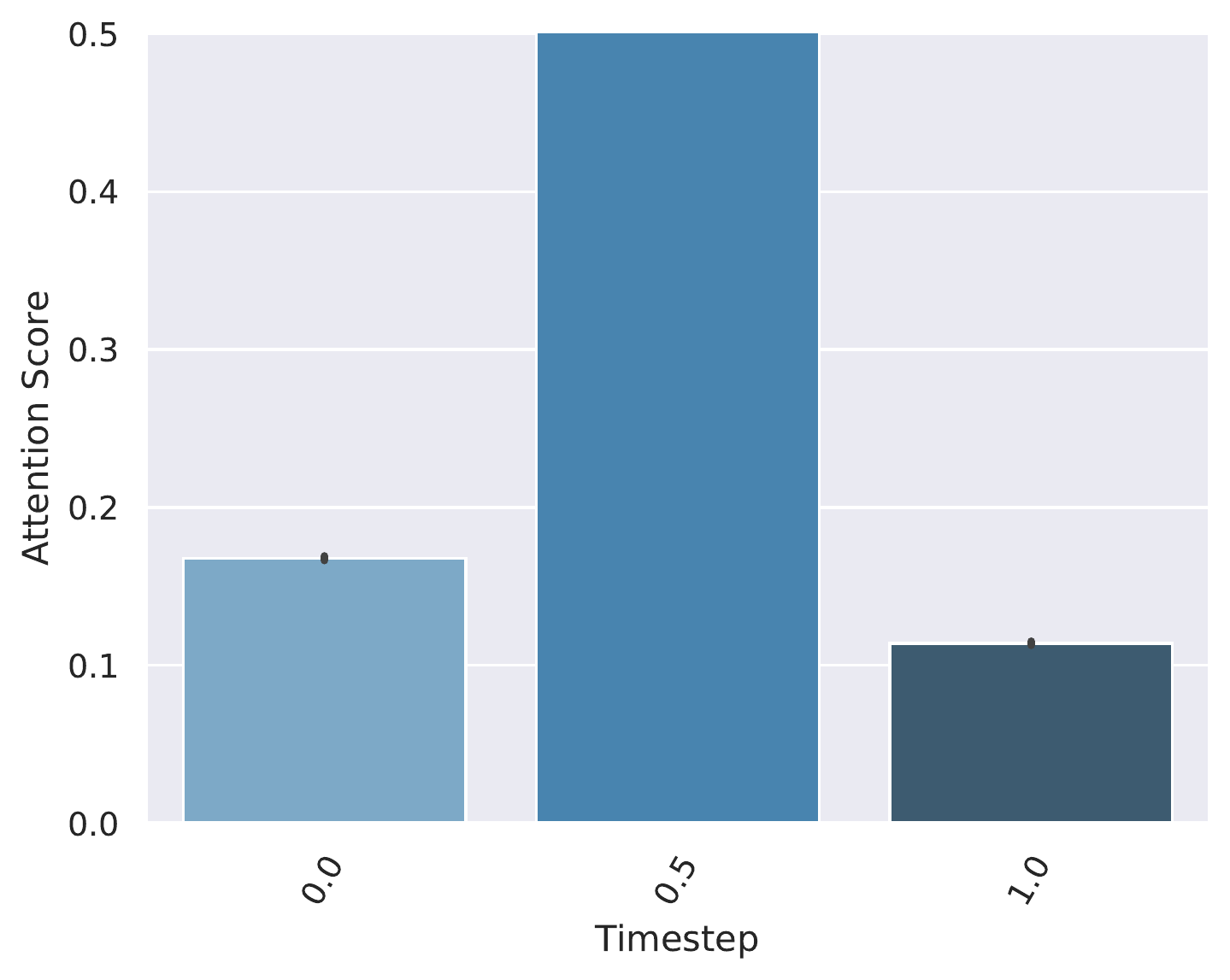}
\vspace{-6mm}
\subcaption*{\scriptsize Object Shape}
\end{subfigure}
\subcaption*{Granularity: 2}
\end{subfigure} \\
\begin{subfigure}[c]{\textwidth}
\begin{subfigure}[c]{0.24\textwidth}
\includegraphics[width=\textwidth]{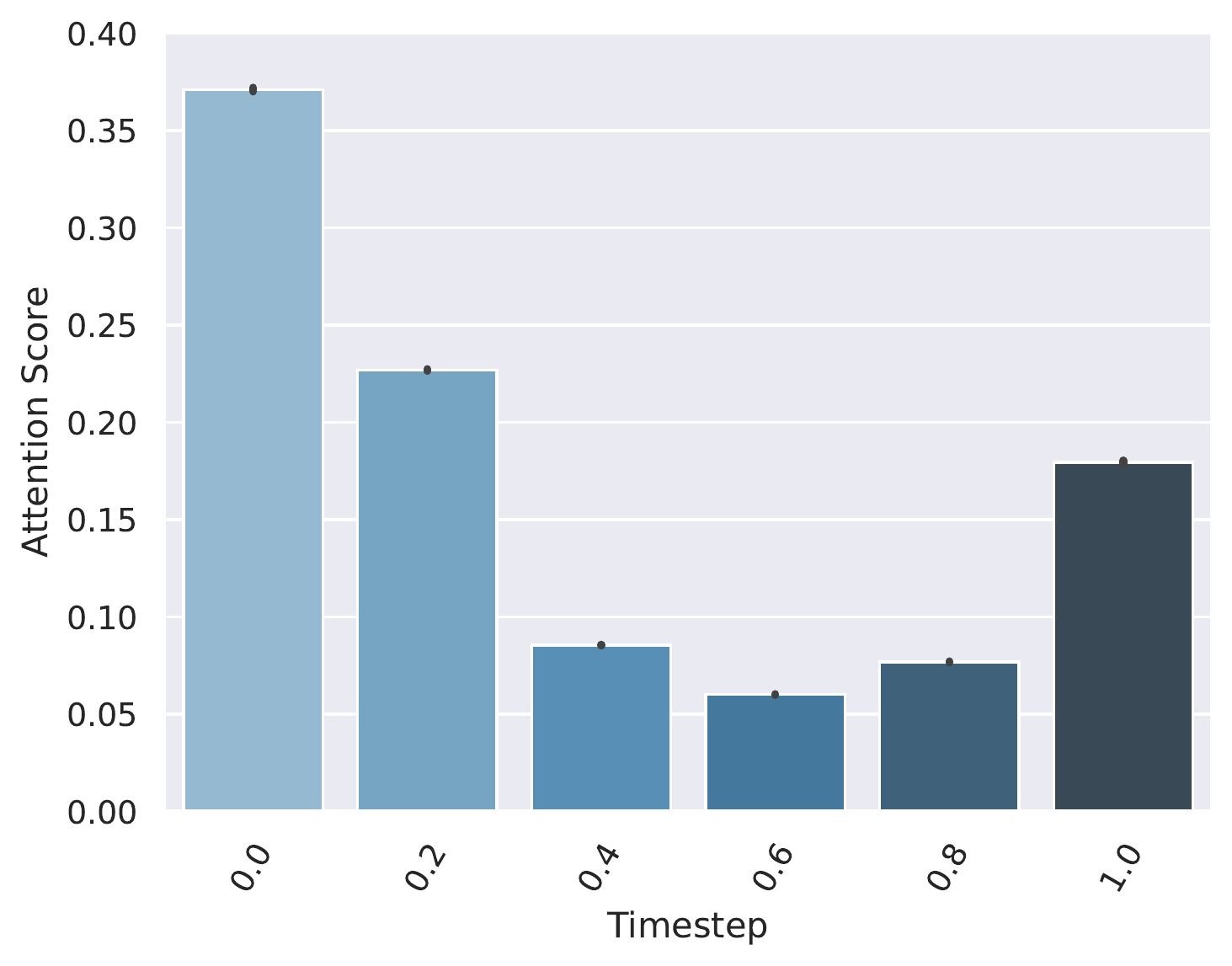}
\vspace{-6mm}
\subcaption*{\scriptsize Background Color}
\end{subfigure}
\begin{subfigure}[c]{0.24\textwidth}
\includegraphics[width=\textwidth]{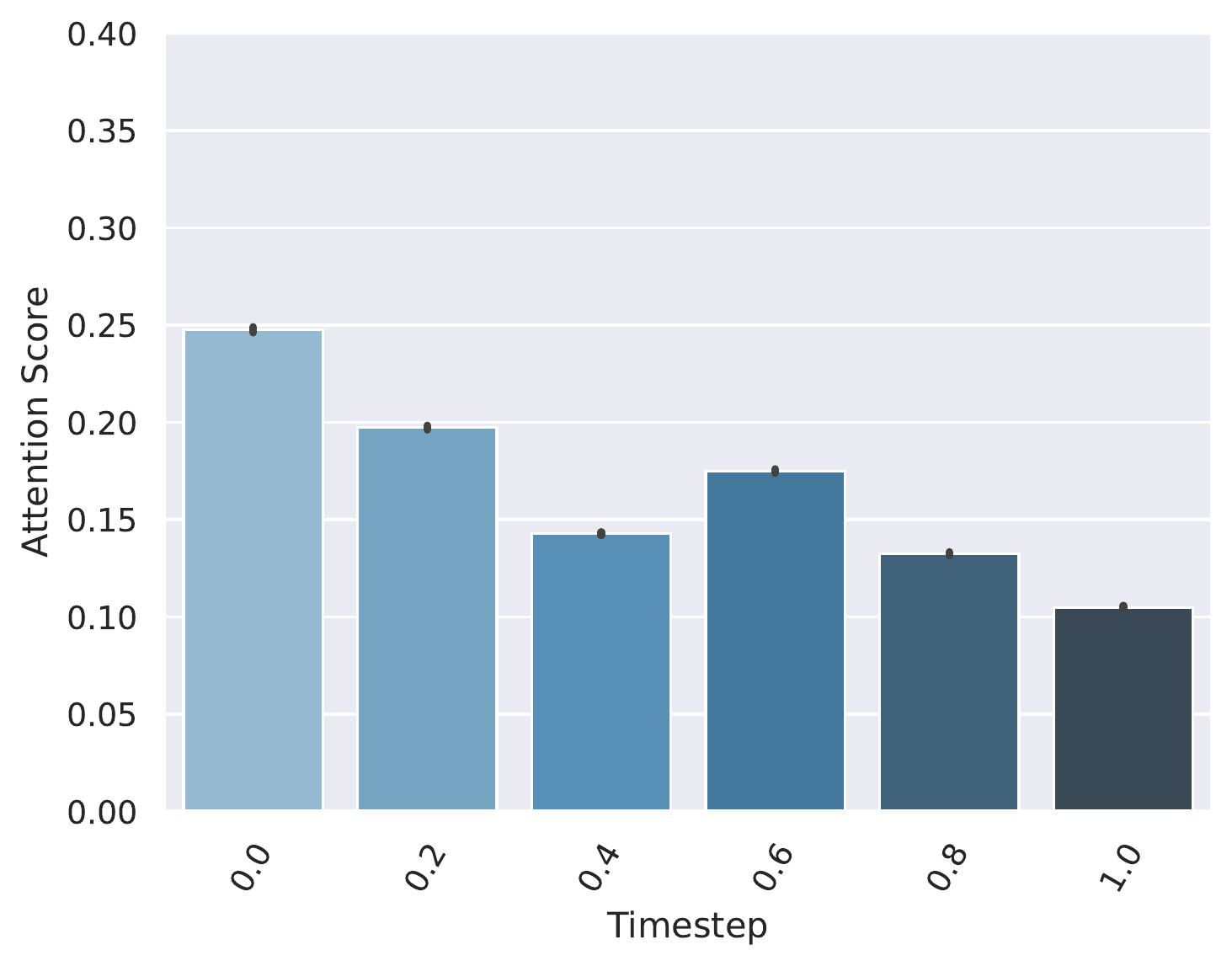}
\vspace{-6mm}
\subcaption*{\scriptsize Foreground Color}
\end{subfigure}
\begin{subfigure}[c]{0.24\textwidth}
\includegraphics[width=\textwidth]{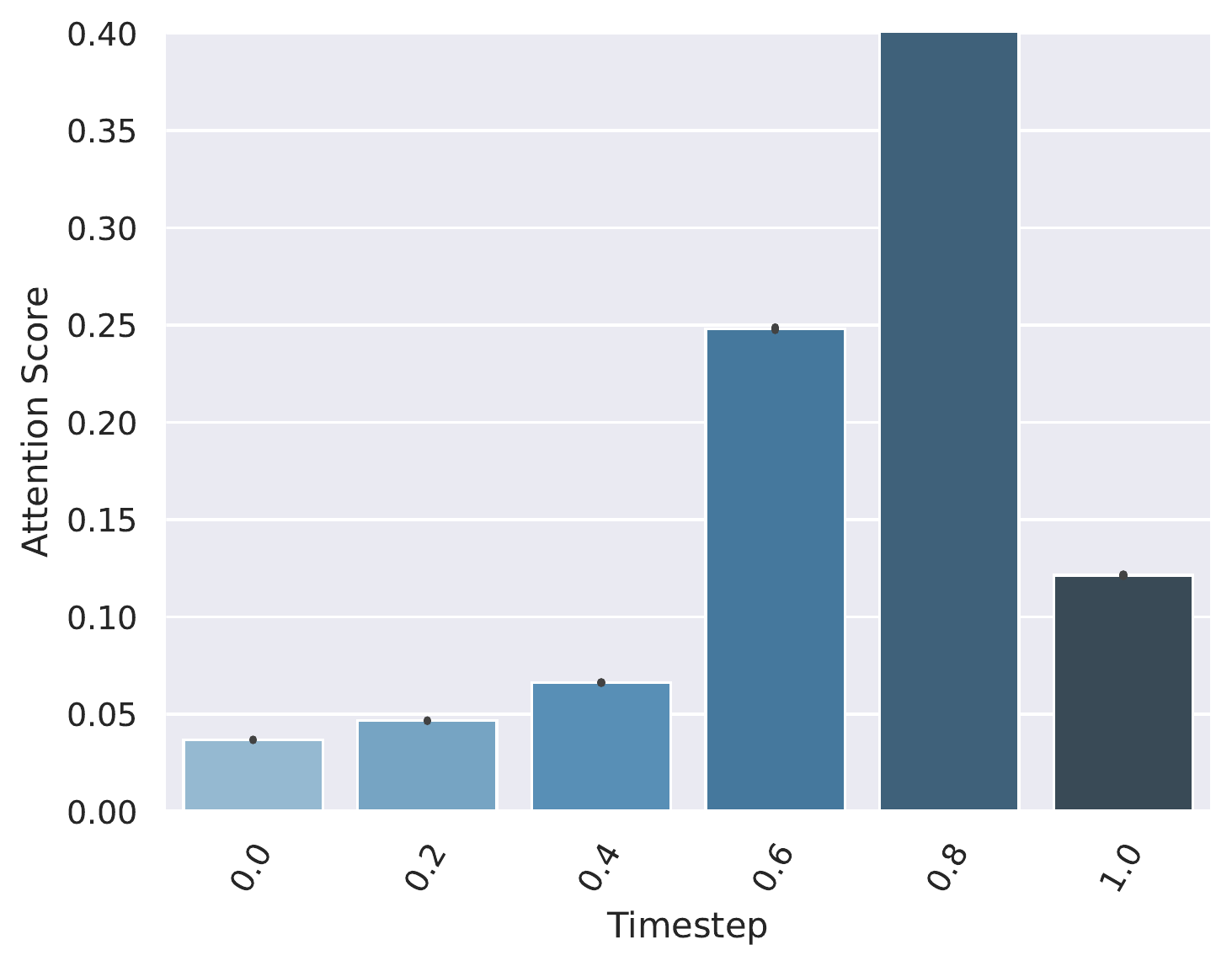}
\vspace{-6mm}
\subcaption*{\scriptsize Location}
\end{subfigure}
\begin{subfigure}[c]{0.24\textwidth}
\includegraphics[width=\textwidth]{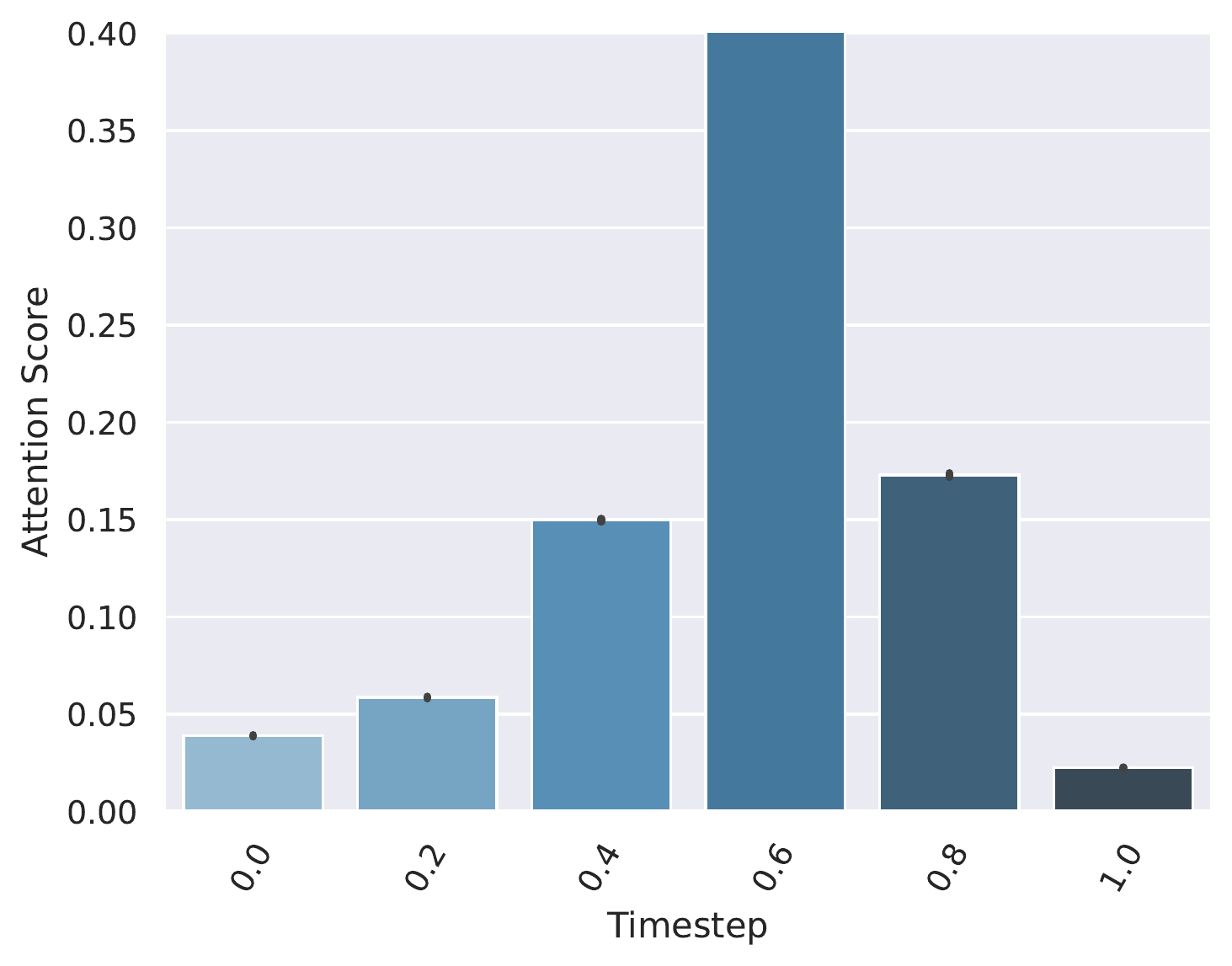}
\vspace{-6mm}
\subcaption*{\scriptsize Object Shape}
\end{subfigure}
\subcaption*{Granularity: 5}
\end{subfigure} \\
\begin{subfigure}[c]{\textwidth}
\begin{subfigure}[c]{0.24\textwidth}
\includegraphics[width=\textwidth]{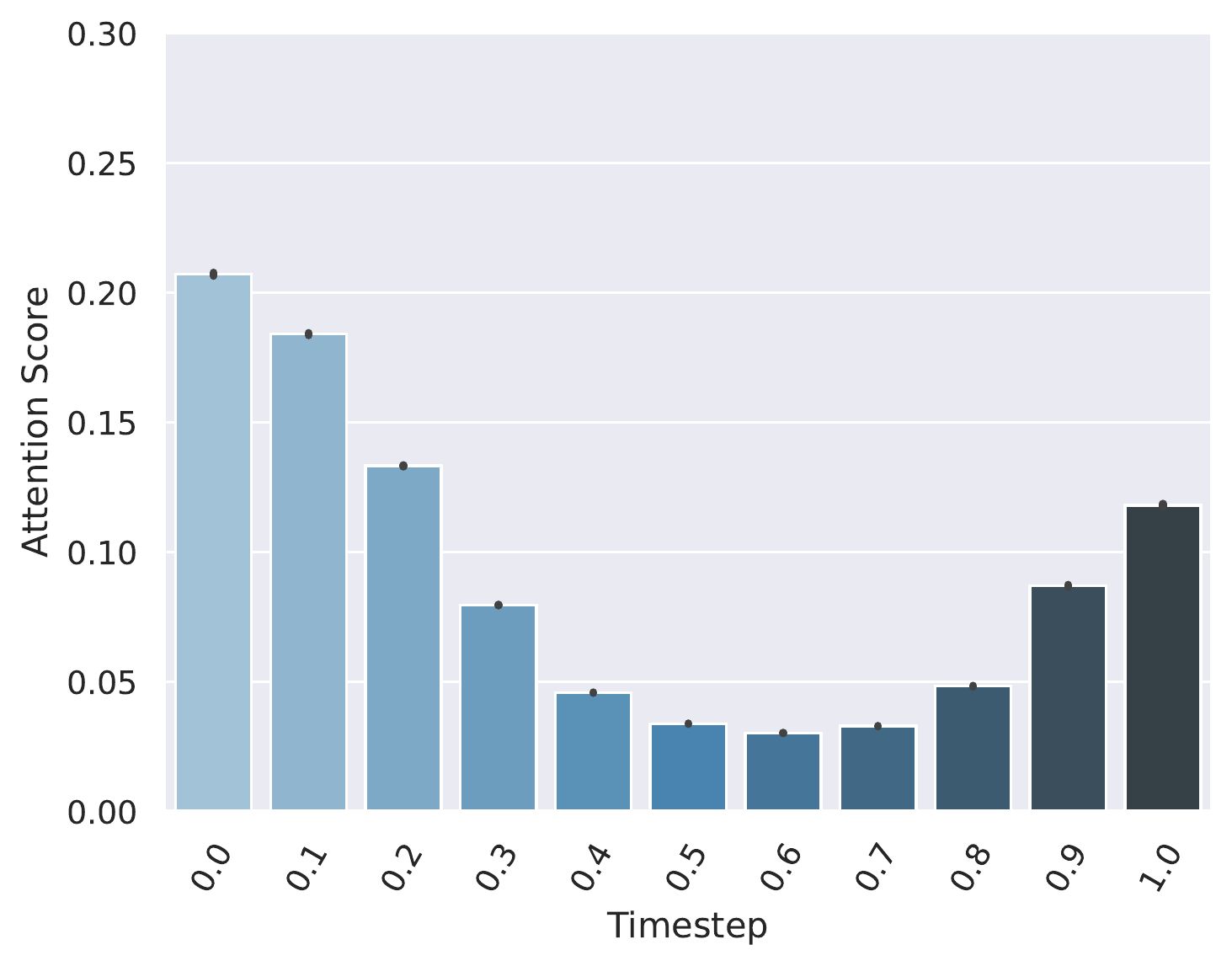}
\vspace{-6mm}
\subcaption*{\scriptsize Background Color}
\end{subfigure}
\begin{subfigure}[c]{0.24\textwidth}
\includegraphics[width=\textwidth]{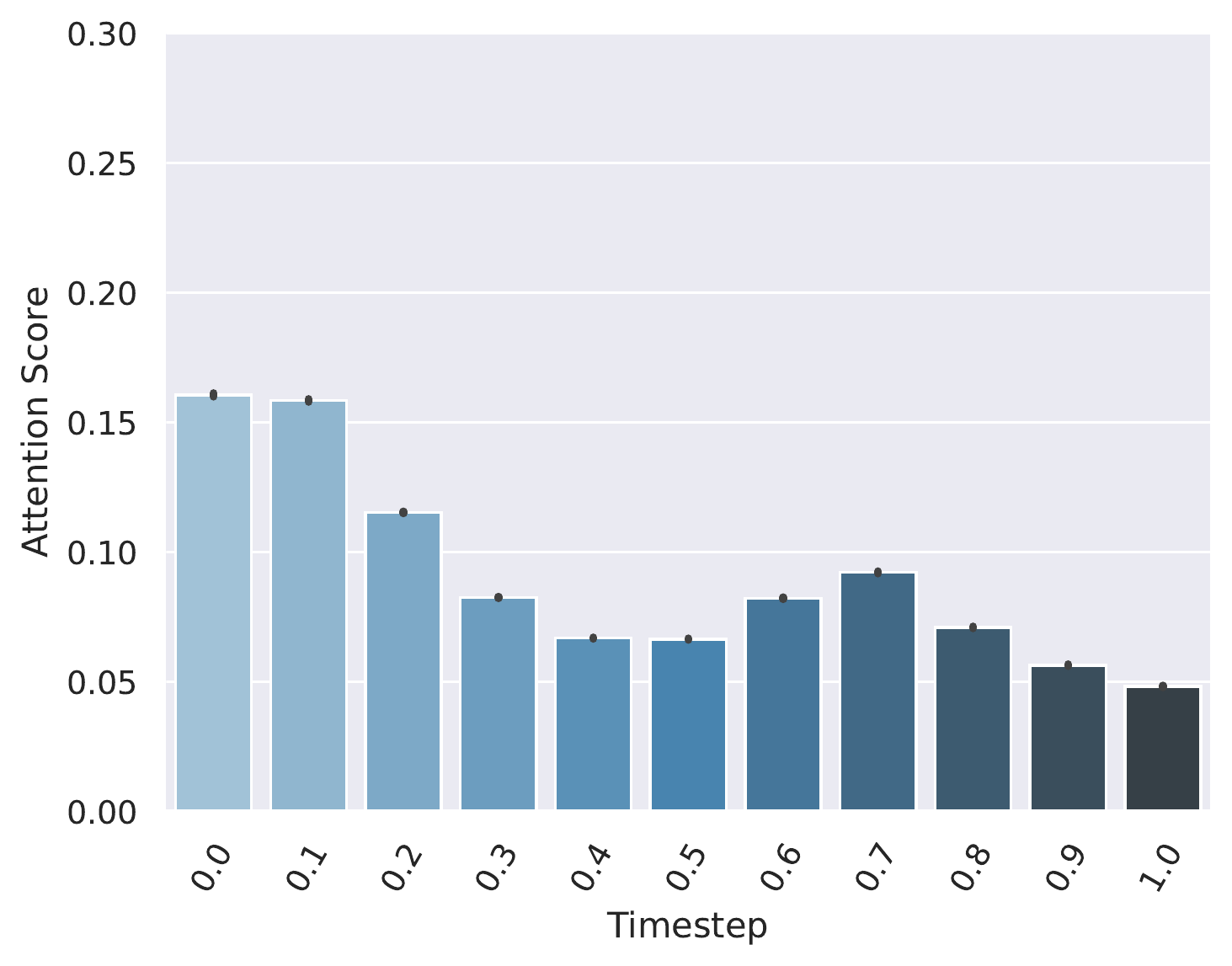}
\vspace{-6mm}
\subcaption*{\scriptsize Foreground Color}
\end{subfigure}
\begin{subfigure}[c]{0.24\textwidth}
\includegraphics[width=\textwidth]{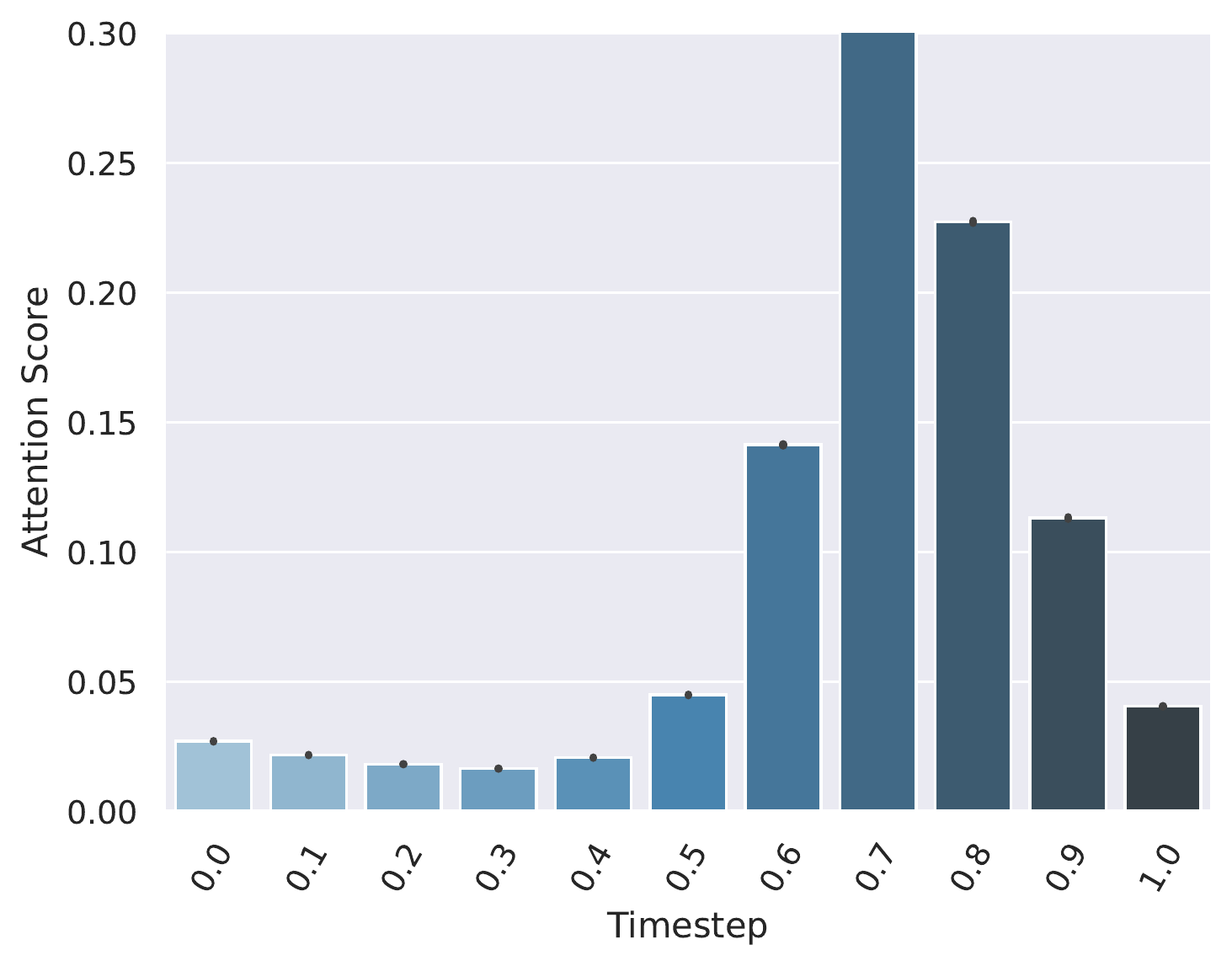}
\vspace{-6mm}
\subcaption*{\scriptsize Location}
\end{subfigure}
\begin{subfigure}[c]{0.24\textwidth}
\includegraphics[width=\textwidth]{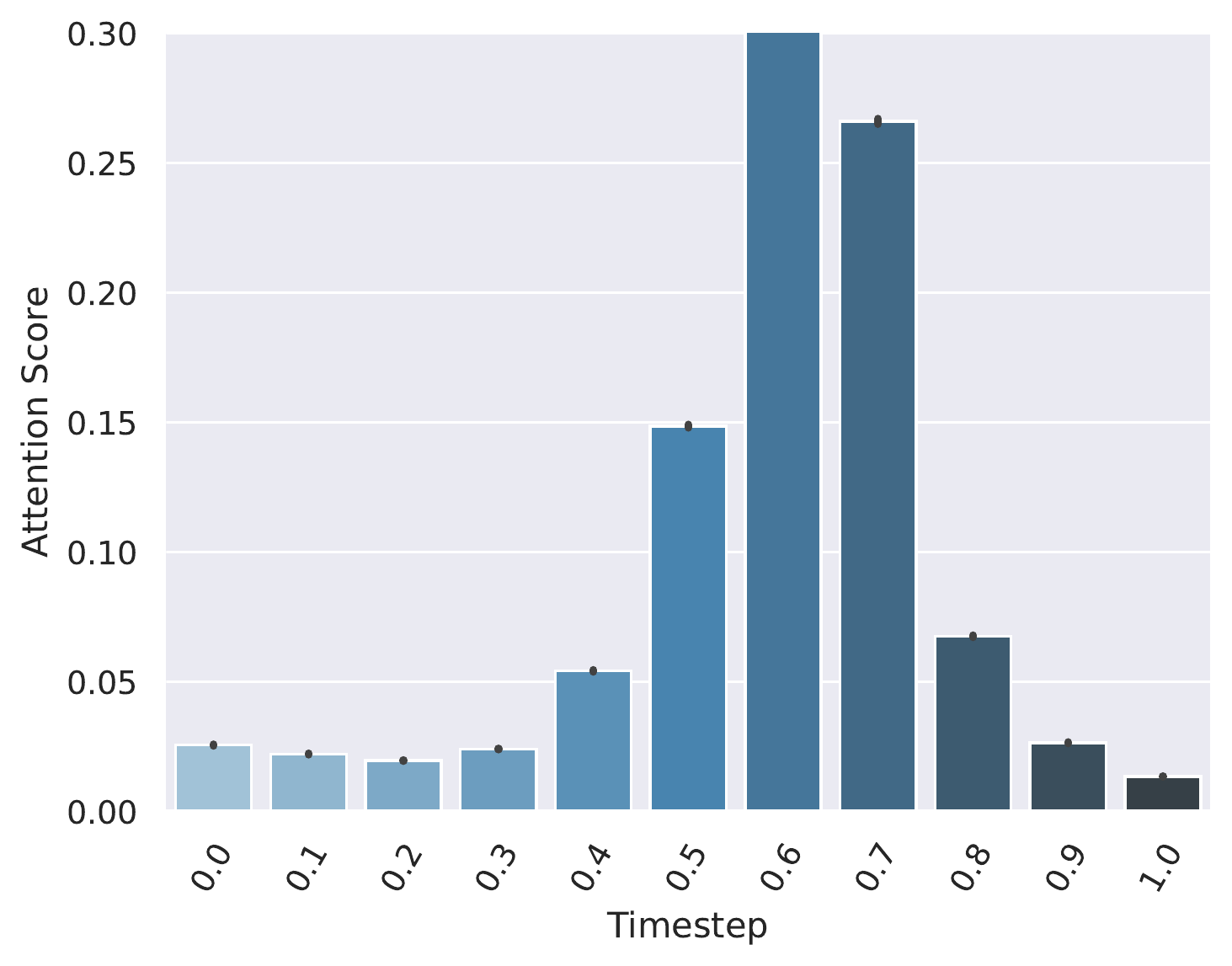}
\vspace{-6mm}
\subcaption*{\scriptsize Object Shape}
\end{subfigure}
\subcaption*{Granularity: 10}
\end{subfigure} \\
\begin{subfigure}[c]{\textwidth}
\begin{subfigure}[c]{0.24\textwidth}
\includegraphics[width=\textwidth]{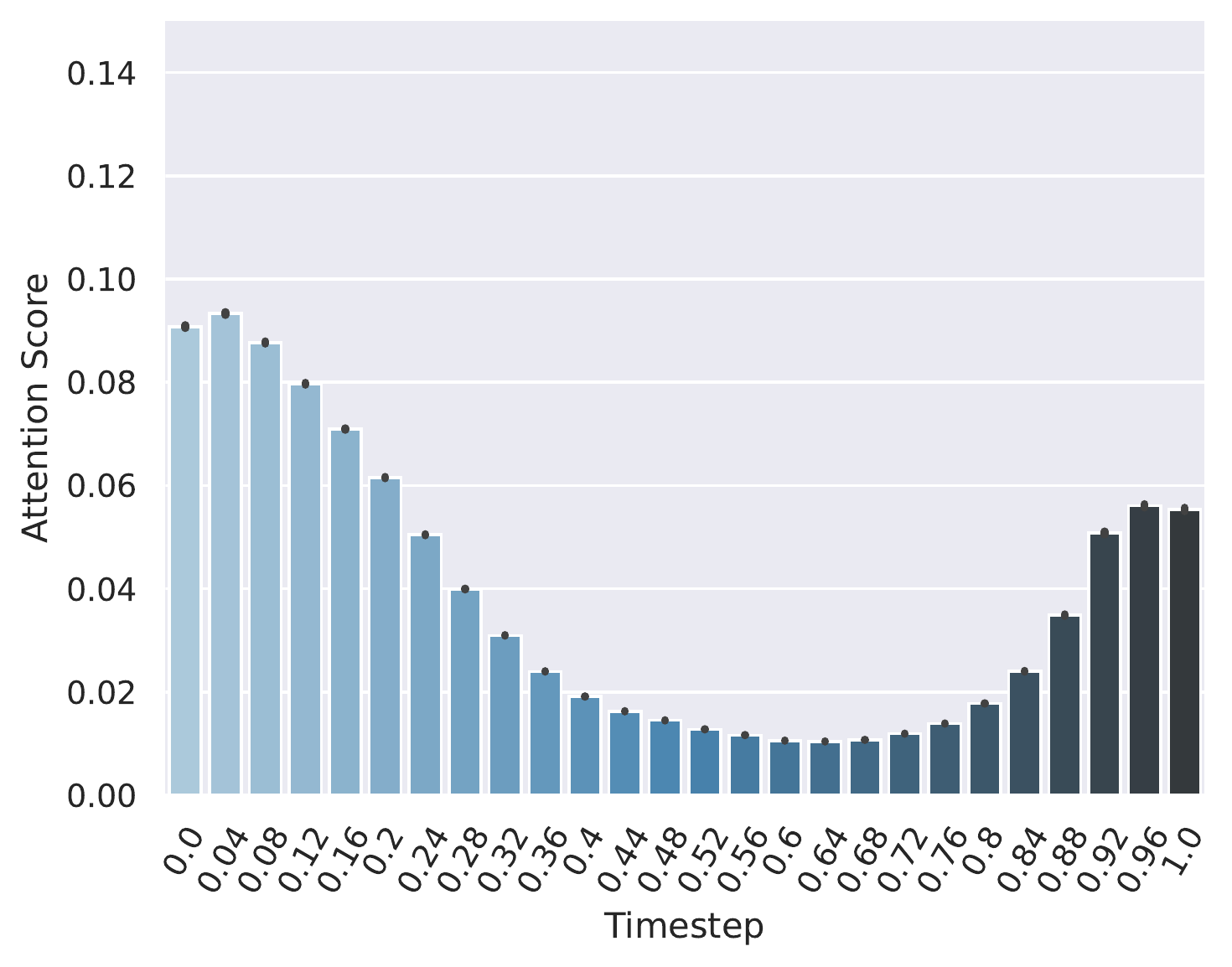}
\vspace{-6mm}
\subcaption*{\scriptsize Background Color}
\end{subfigure}
\begin{subfigure}[c]{0.24\textwidth}
\includegraphics[width=\textwidth]{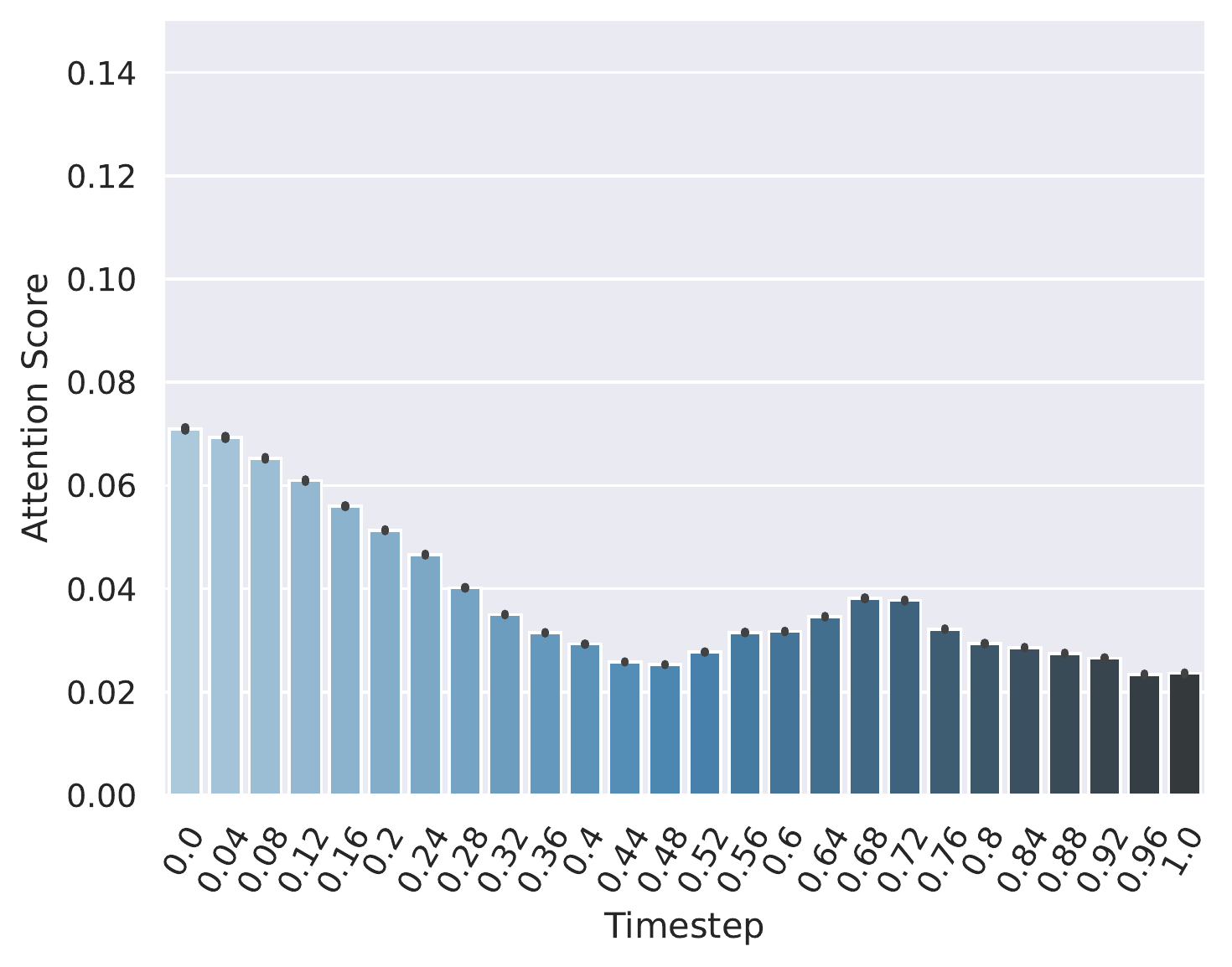}
\vspace{-6mm}
\subcaption*{\scriptsize Foreground Color}
\end{subfigure}
\begin{subfigure}[c]{0.24\textwidth}
\includegraphics[width=\textwidth]{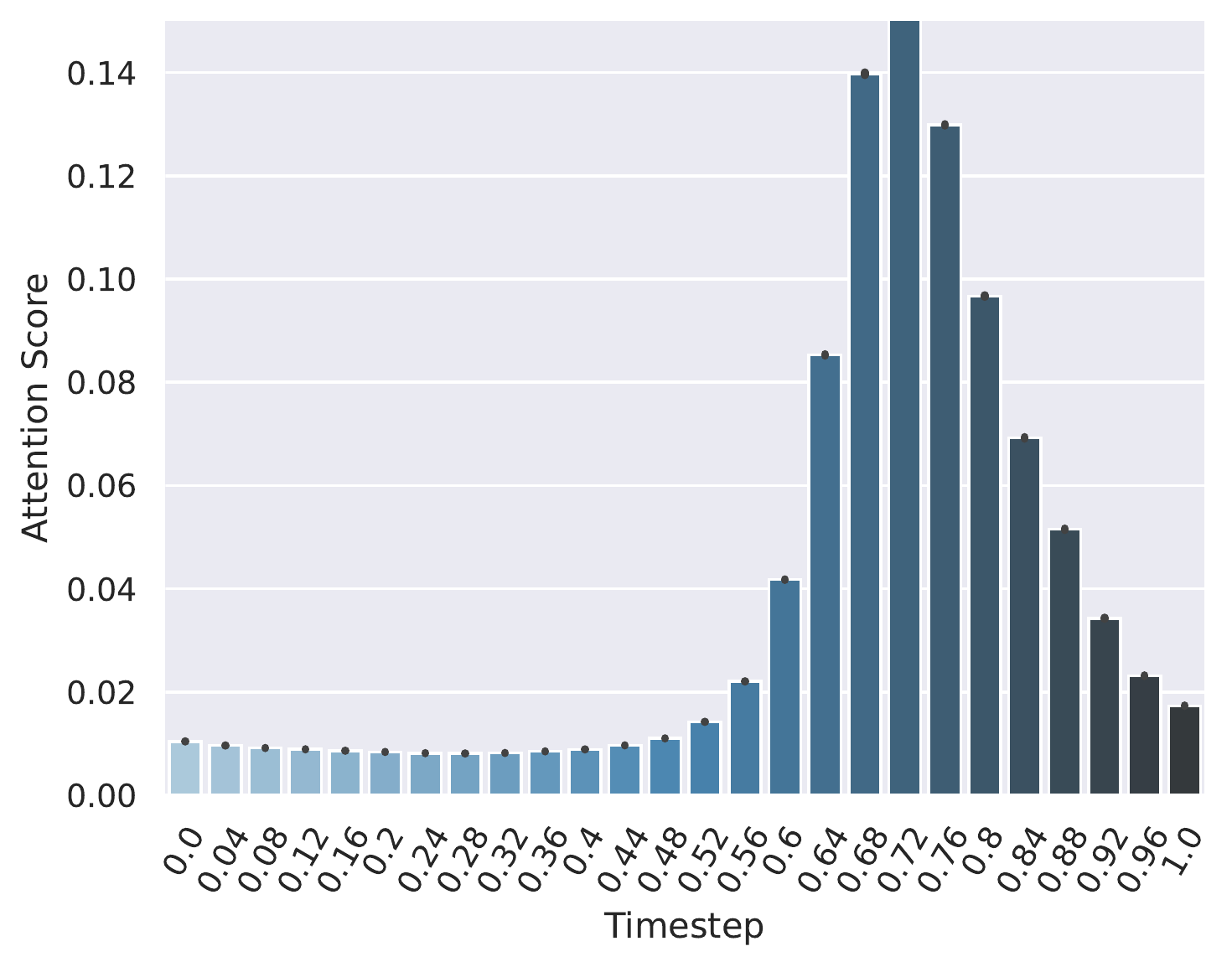}
\vspace{-6mm}
\subcaption*{\scriptsize Location}
\end{subfigure}
\begin{subfigure}[c]{0.24\textwidth}
\includegraphics[width=\textwidth]{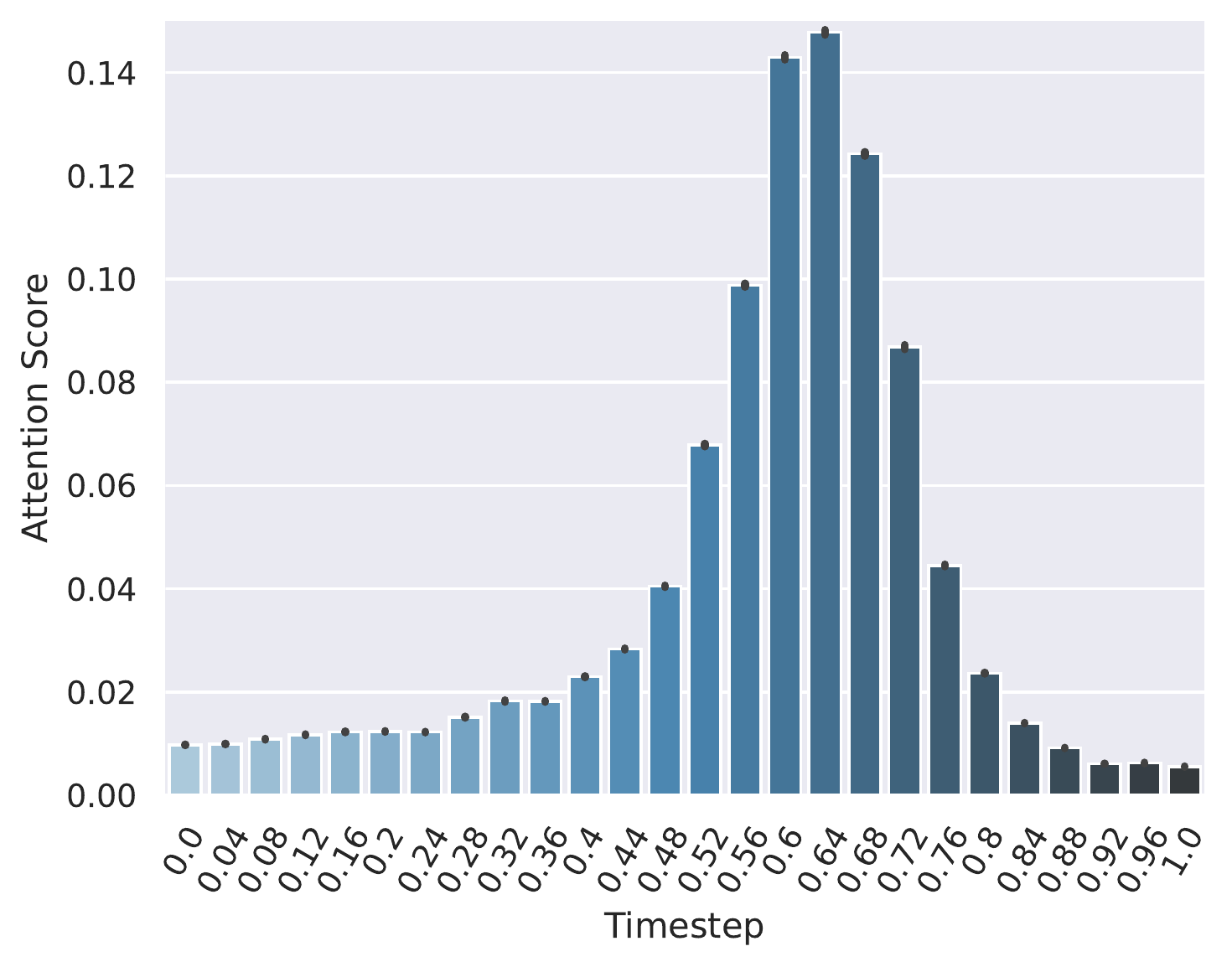}
\vspace{-6mm}
\subcaption*{\scriptsize Object Shape}
\end{subfigure}
\subcaption*{Granularity: 25}
\end{subfigure} \\
\caption{Attention score profiles for the synthetic dataset on the different features, using different granularities, with the dimensionality of the latent space as 16 and the VDRL encoder.}
\label{fig:syn_VDRL_16}
\end{figure}

\begin{figure}
\begin{subfigure}[c]{\textwidth}
\begin{subfigure}[c]{0.24\textwidth}
\includegraphics[width=\textwidth]{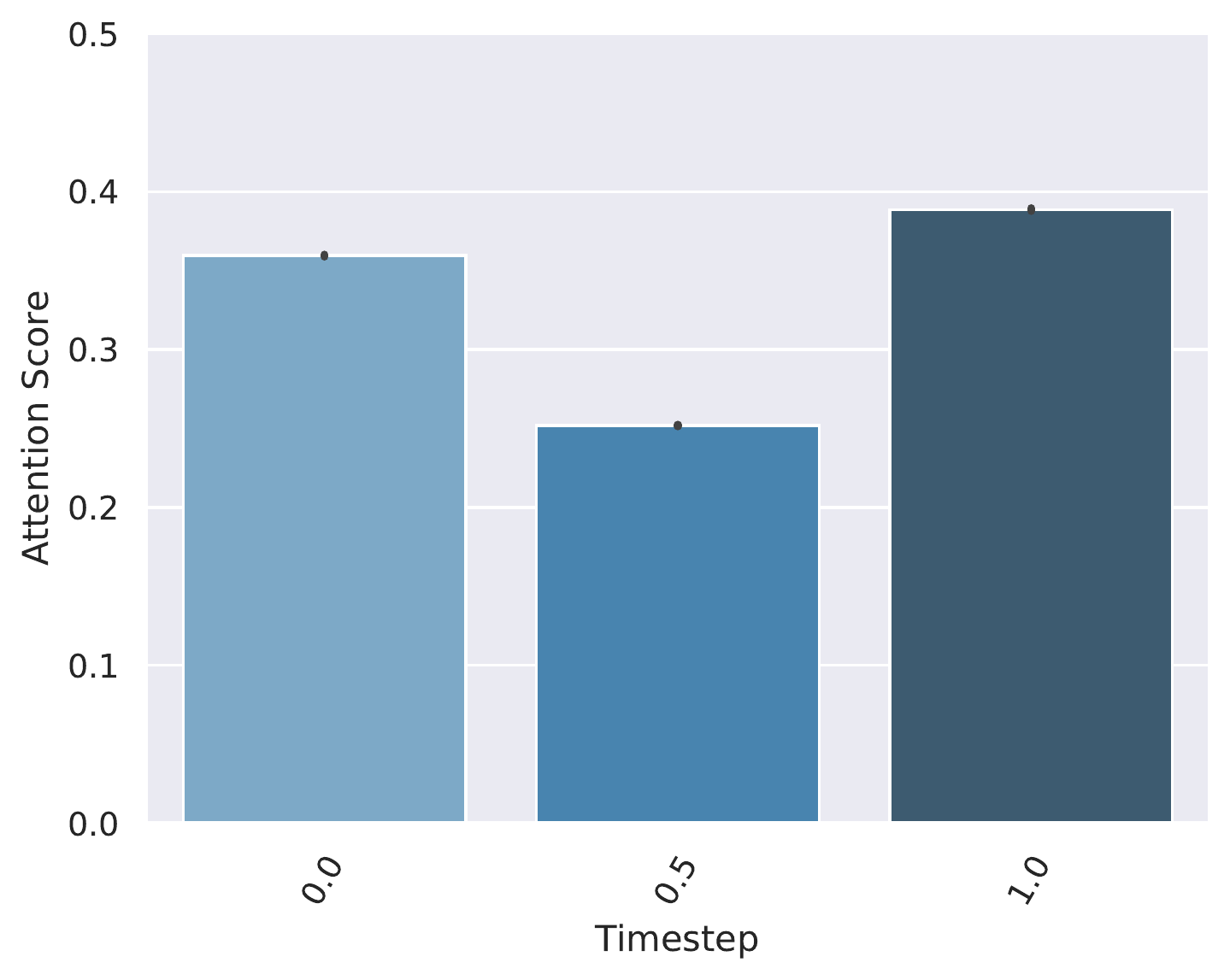}
\vspace{-6mm}
\subcaption*{\scriptsize Background Color}
\end{subfigure}
\begin{subfigure}[c]{0.24\textwidth}
\includegraphics[width=\textwidth]{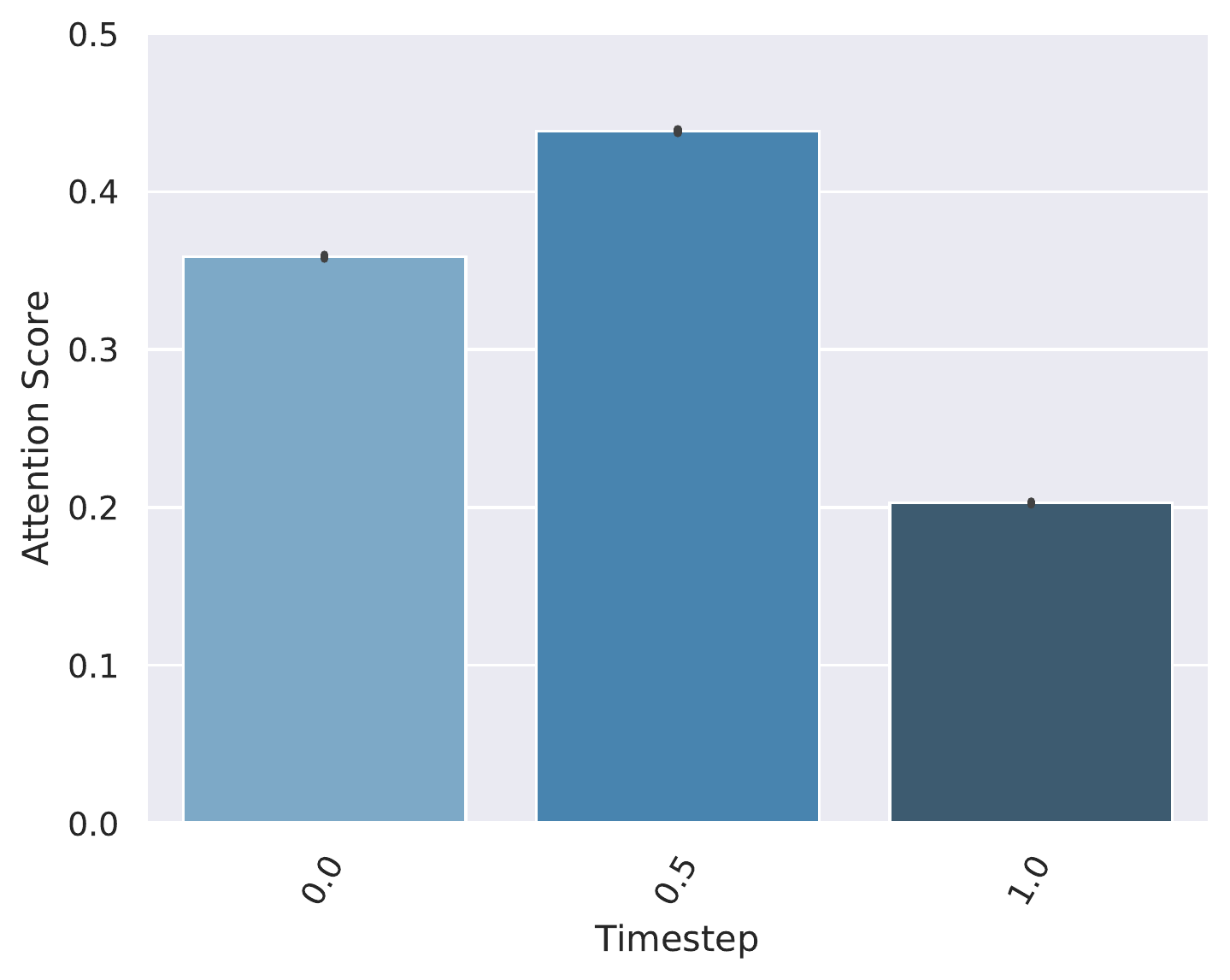}
\vspace{-6mm}
\subcaption*{\scriptsize Foreground Color}
\end{subfigure}
\begin{subfigure}[c]{0.24\textwidth}
\includegraphics[width=\textwidth]{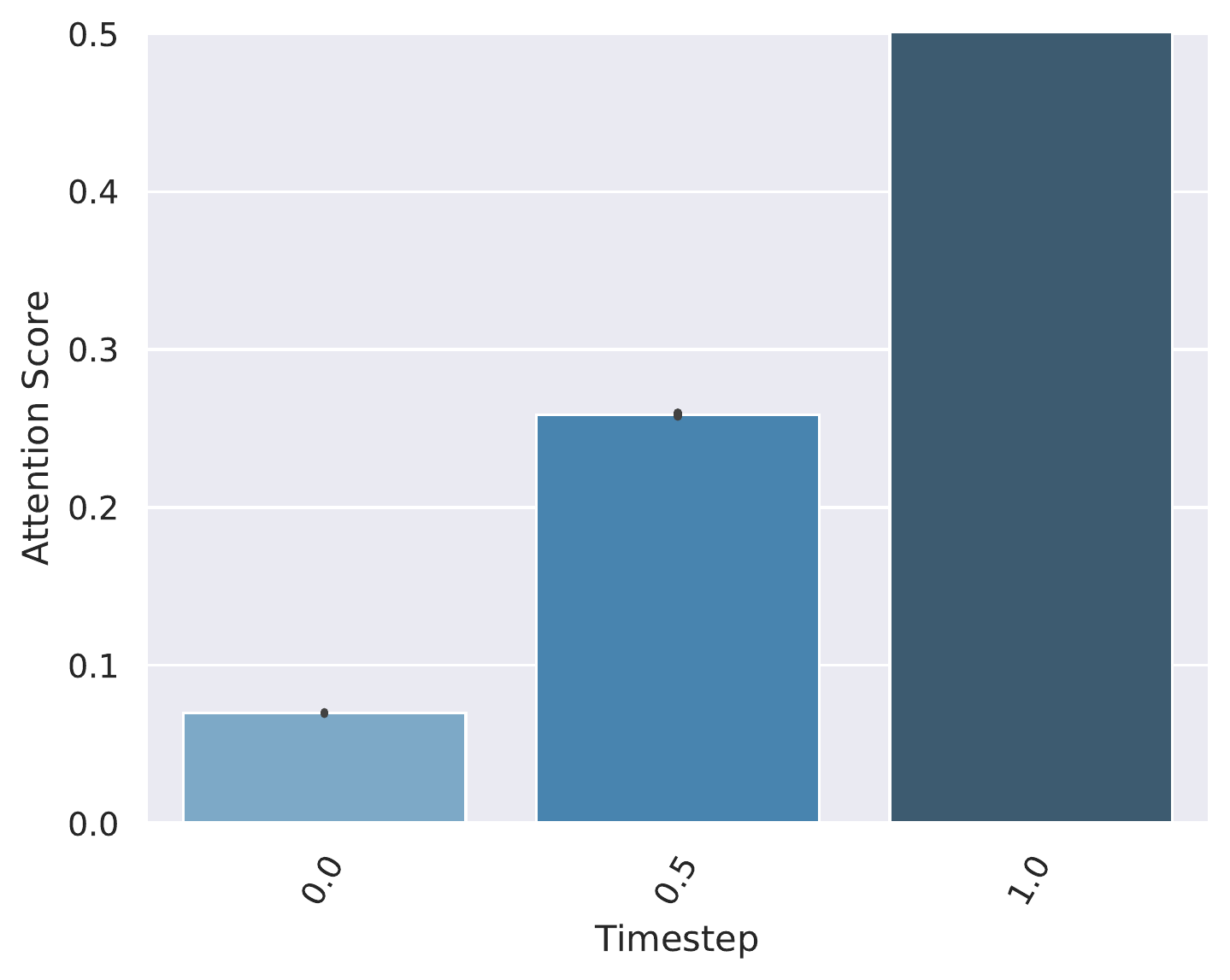}
\vspace{-6mm}
\subcaption*{\scriptsize Location}
\end{subfigure}
\begin{subfigure}[c]{0.24\textwidth}
\includegraphics[width=\textwidth]{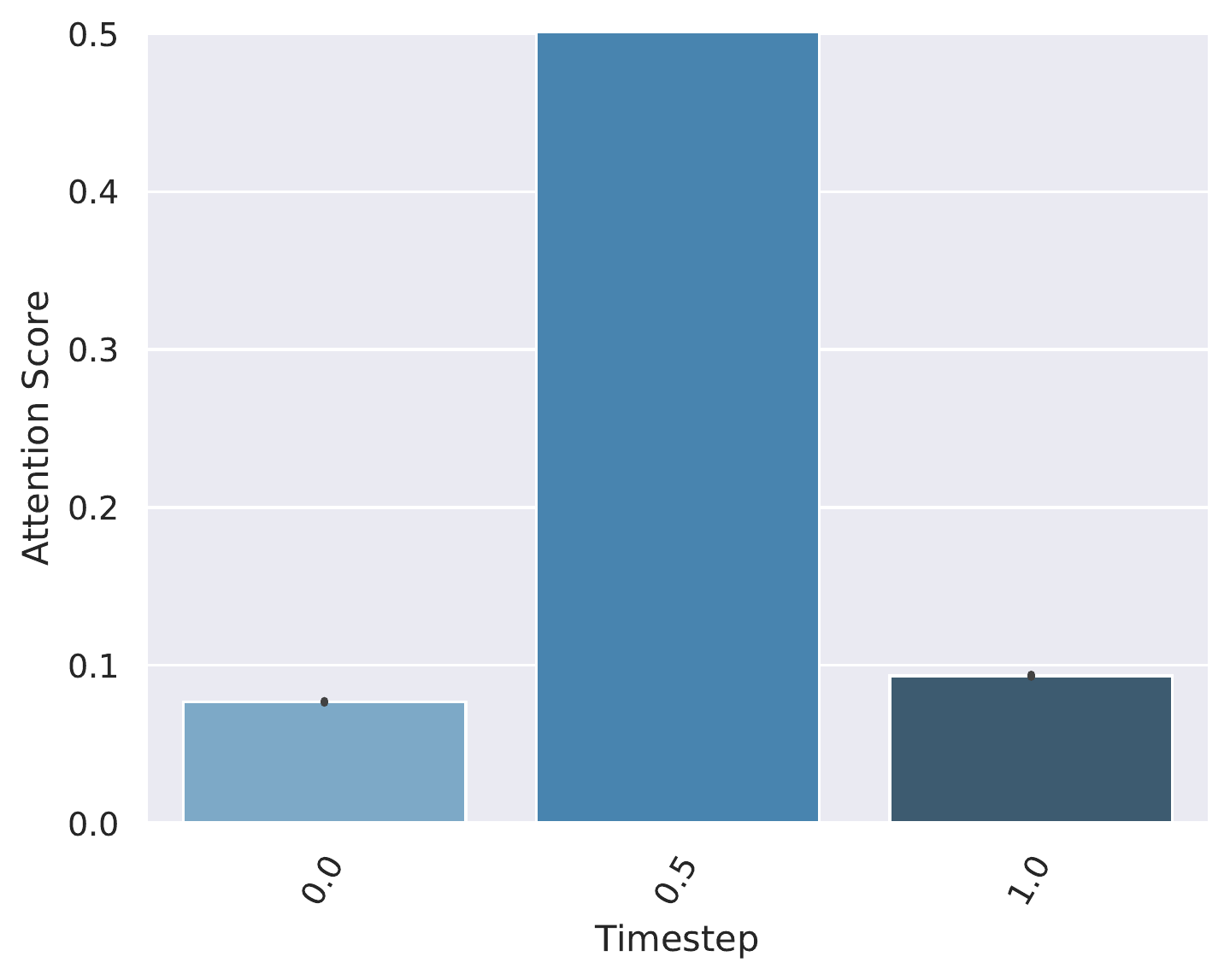}
\vspace{-6mm}
\subcaption*{\scriptsize Object Shape}
\end{subfigure}
\subcaption*{Granularity: 2}
\end{subfigure} \\
\begin{subfigure}[c]{\textwidth}
\begin{subfigure}[c]{0.24\textwidth}
\includegraphics[width=\textwidth]{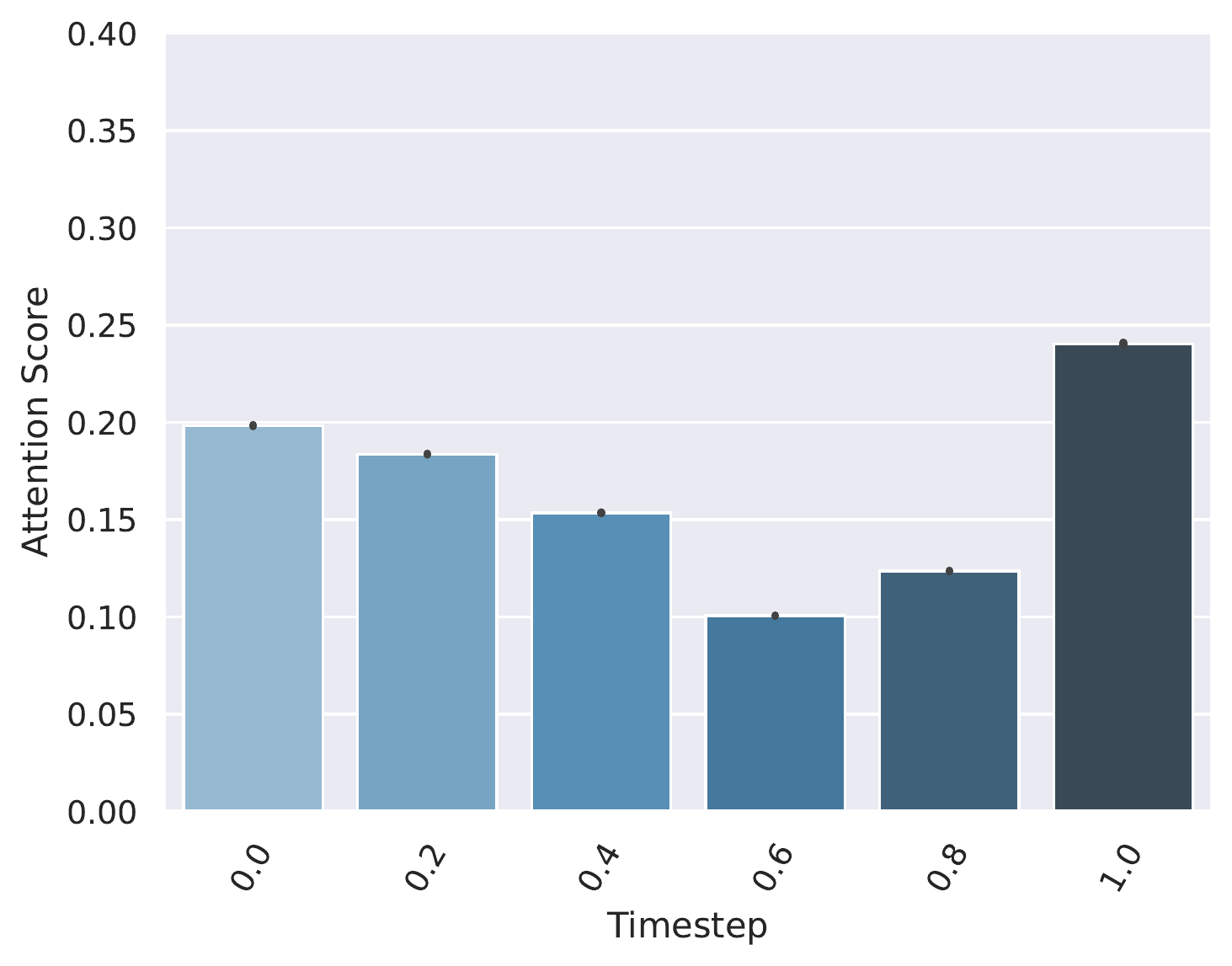}
\vspace{-6mm}
\subcaption*{\scriptsize Background Color}
\end{subfigure}
\begin{subfigure}[c]{0.24\textwidth}
\includegraphics[width=\textwidth]{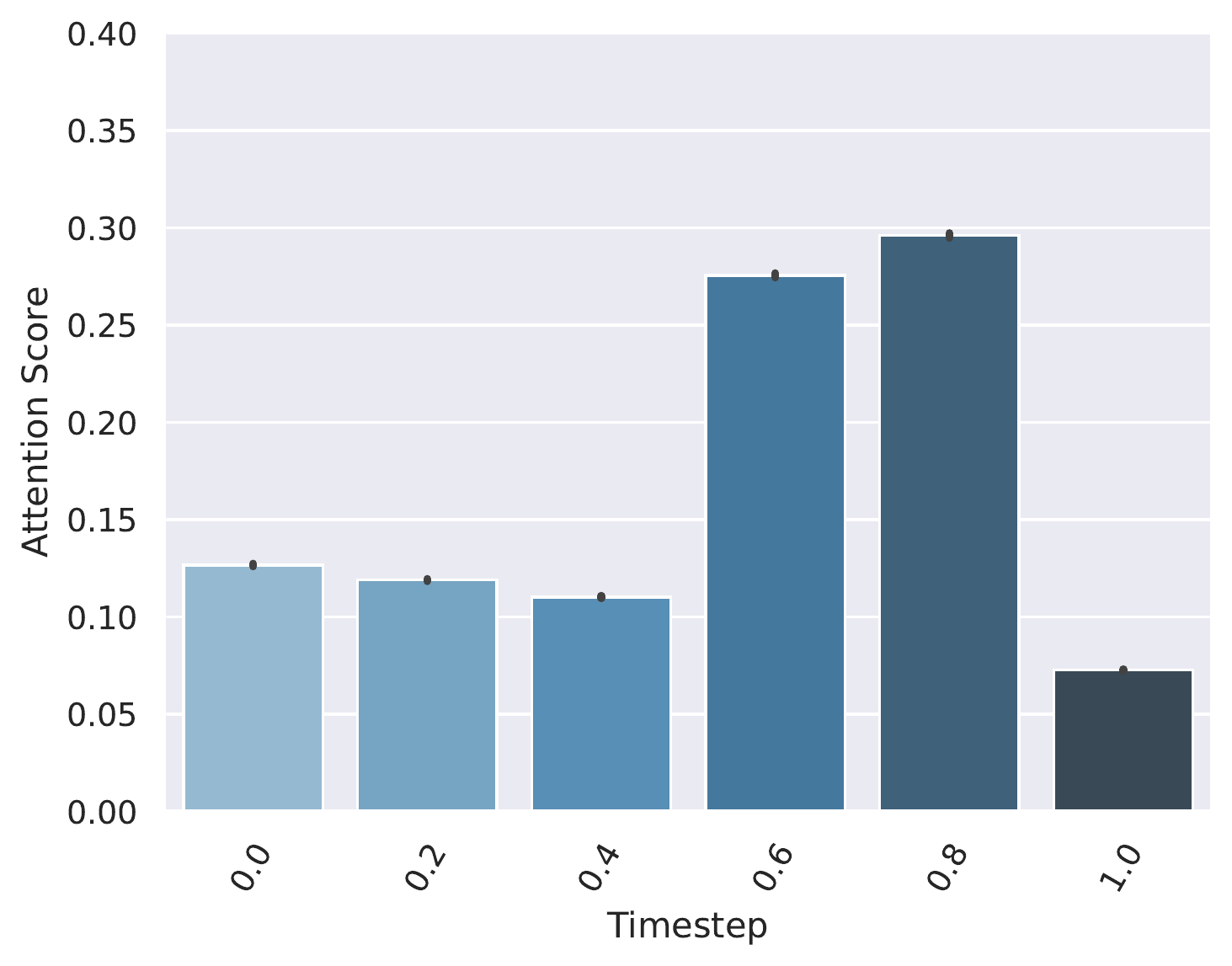}
\vspace{-6mm}
\subcaption*{\scriptsize Foreground Color}
\end{subfigure}
\begin{subfigure}[c]{0.24\textwidth}
\includegraphics[width=\textwidth]{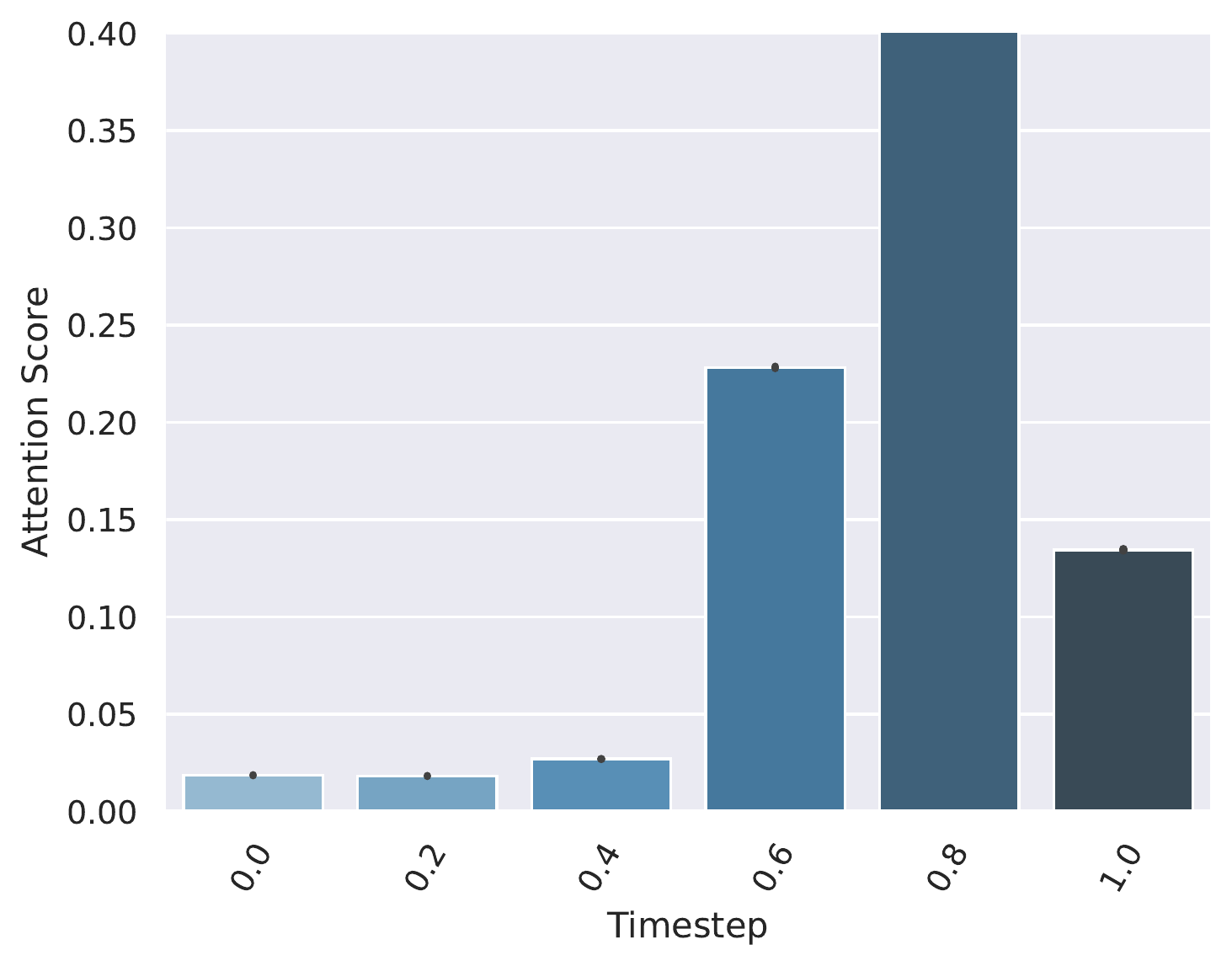}
\vspace{-6mm}
\subcaption*{\scriptsize Location}
\end{subfigure}
\begin{subfigure}[c]{0.24\textwidth}
\includegraphics[width=\textwidth]{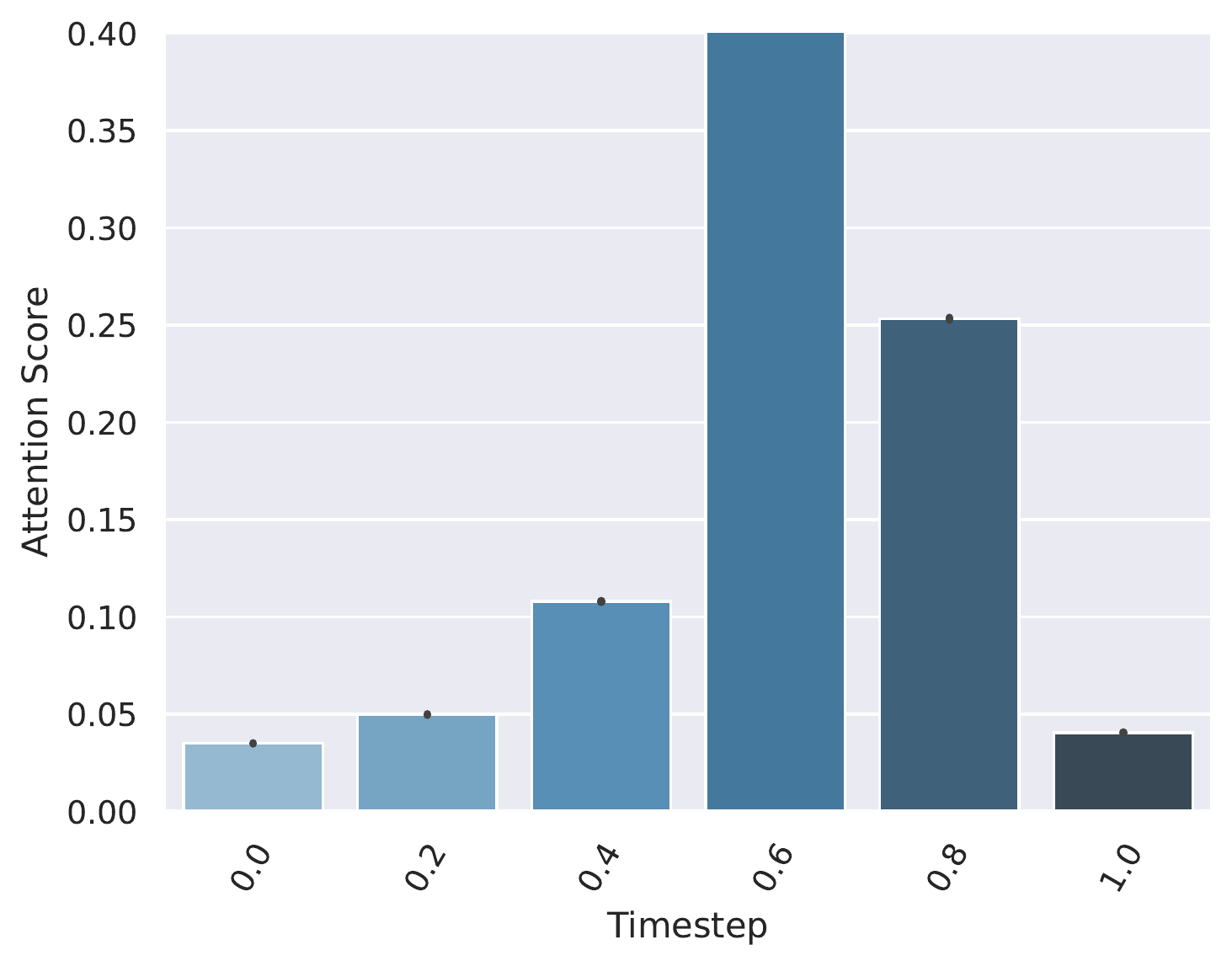}
\vspace{-6mm}
\subcaption*{\scriptsize Object Shape}
\end{subfigure}
\subcaption*{Granularity: 5}
\end{subfigure} \\
\begin{subfigure}[c]{\textwidth}
\begin{subfigure}[c]{0.24\textwidth}
\includegraphics[width=\textwidth]{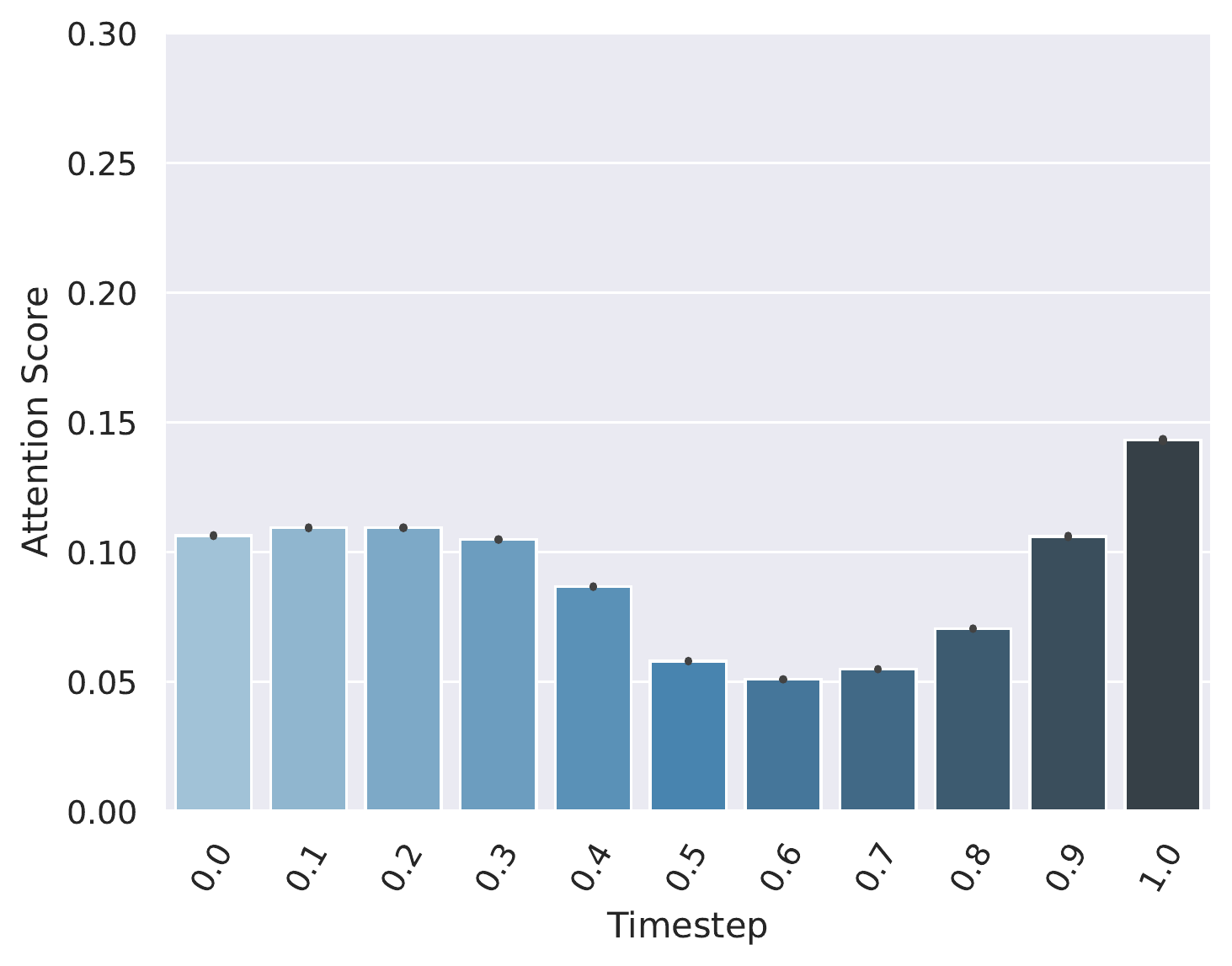}
\vspace{-6mm}
\subcaption*{\scriptsize Background Color}
\end{subfigure}
\begin{subfigure}[c]{0.24\textwidth}
\includegraphics[width=\textwidth]{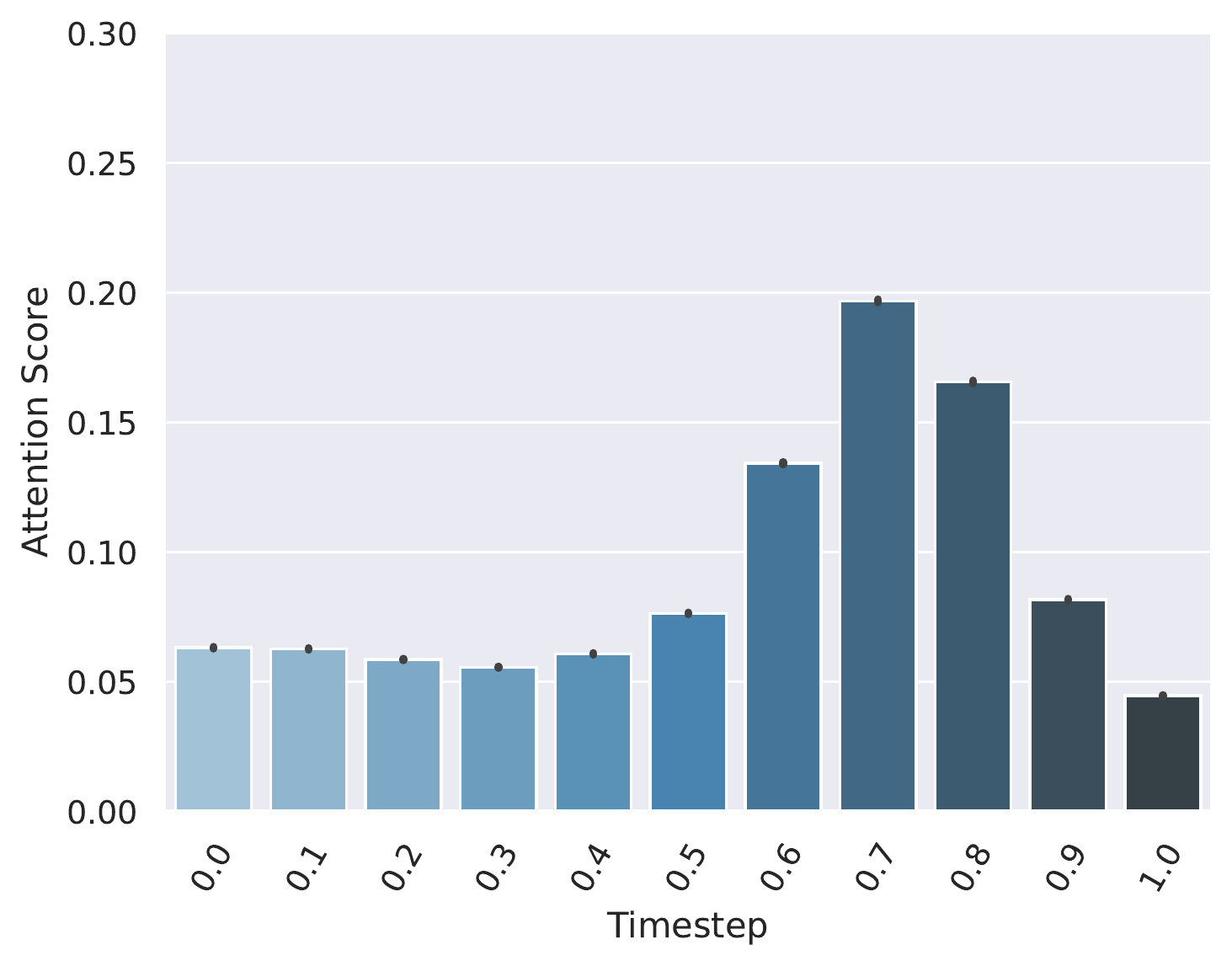}
\vspace{-6mm}
\subcaption*{\scriptsize Foreground Color}
\end{subfigure}
\begin{subfigure}[c]{0.24\textwidth}
\includegraphics[width=\textwidth]{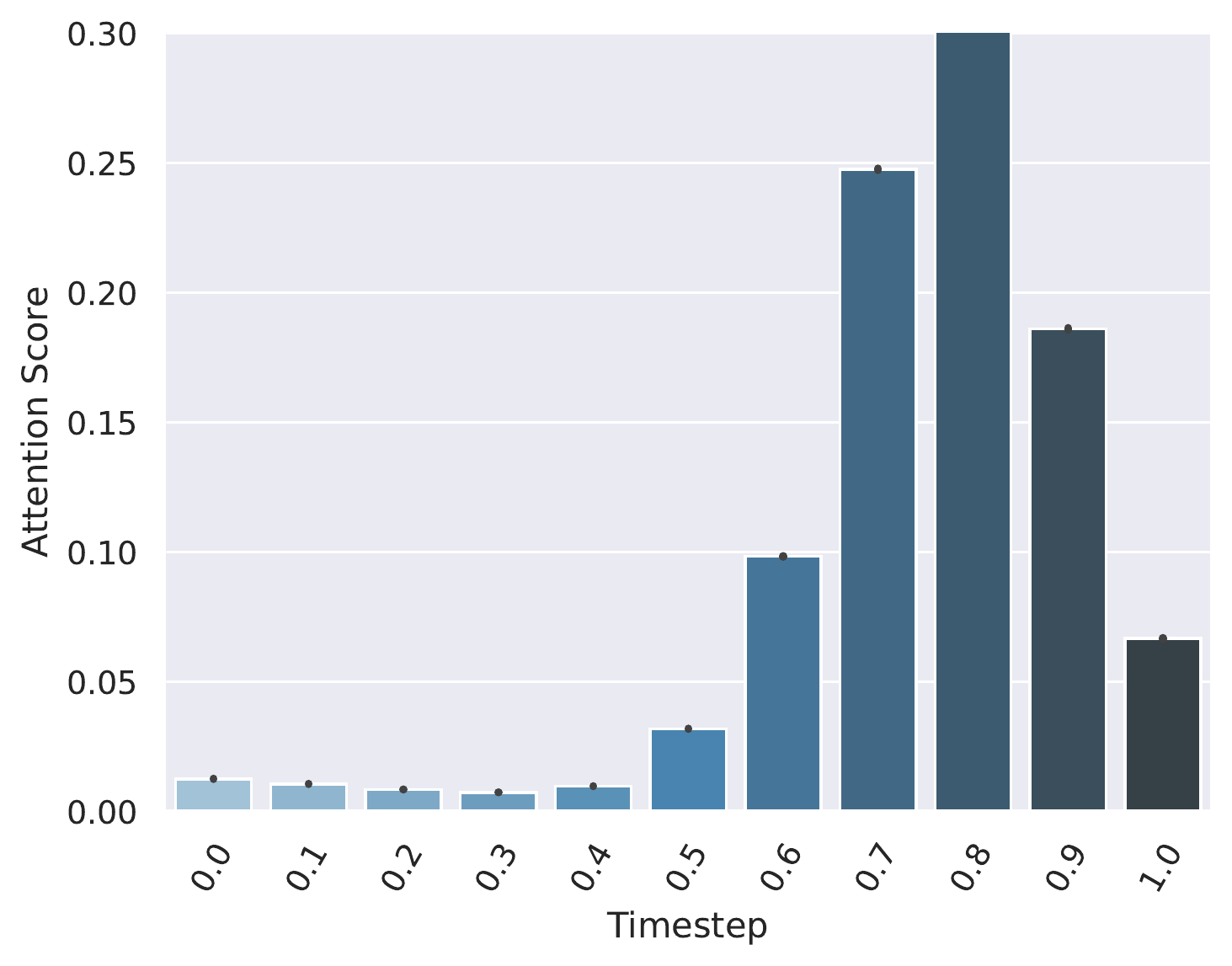}
\vspace{-6mm}
\subcaption*{\scriptsize Location}
\end{subfigure}
\begin{subfigure}[c]{0.24\textwidth}
\includegraphics[width=\textwidth]{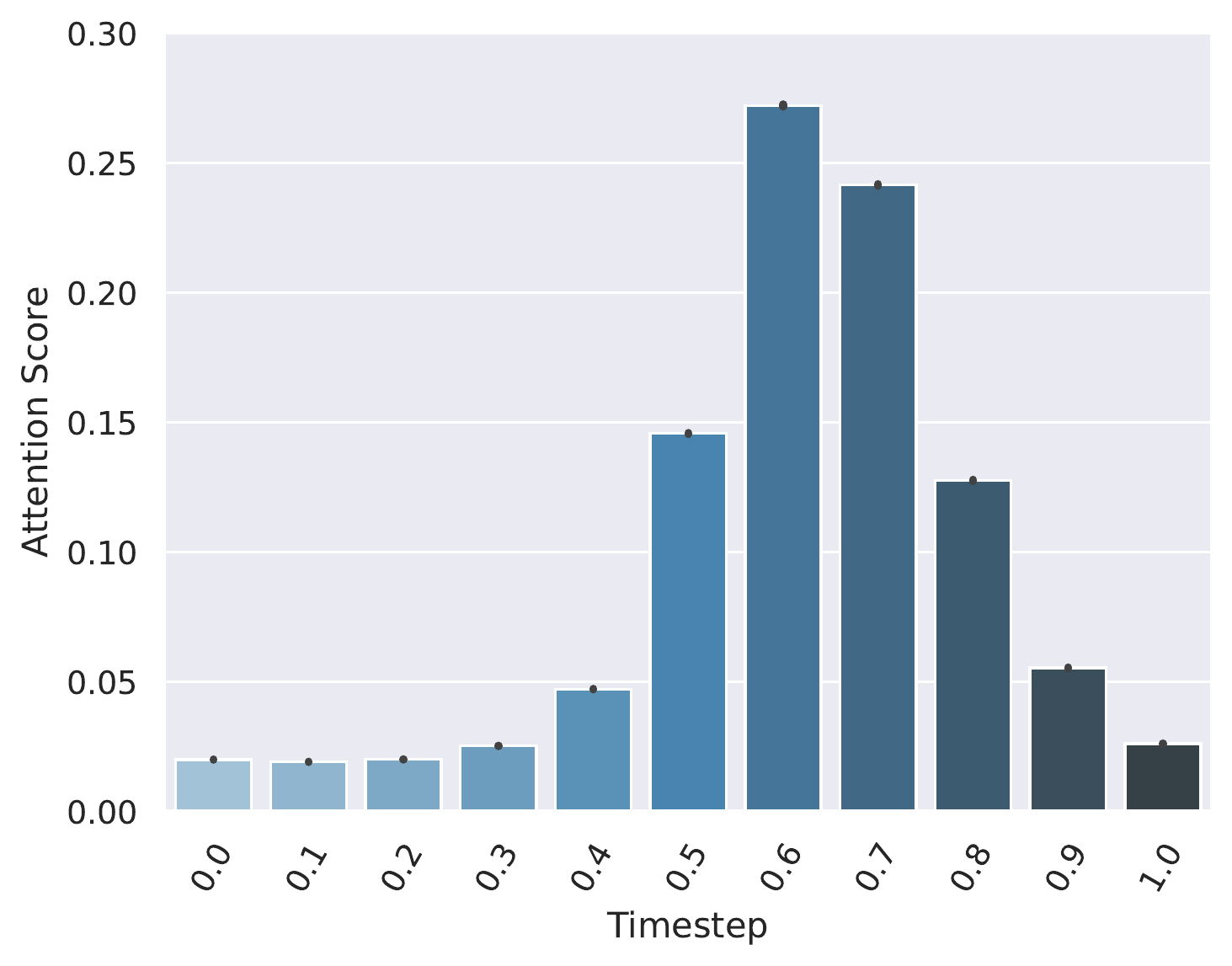}
\vspace{-6mm}
\subcaption*{\scriptsize Object Shape}
\end{subfigure}
\subcaption*{Granularity: 10}
\end{subfigure} \\
\begin{subfigure}[c]{\textwidth}
\begin{subfigure}[c]{0.24\textwidth}
\includegraphics[width=\textwidth]{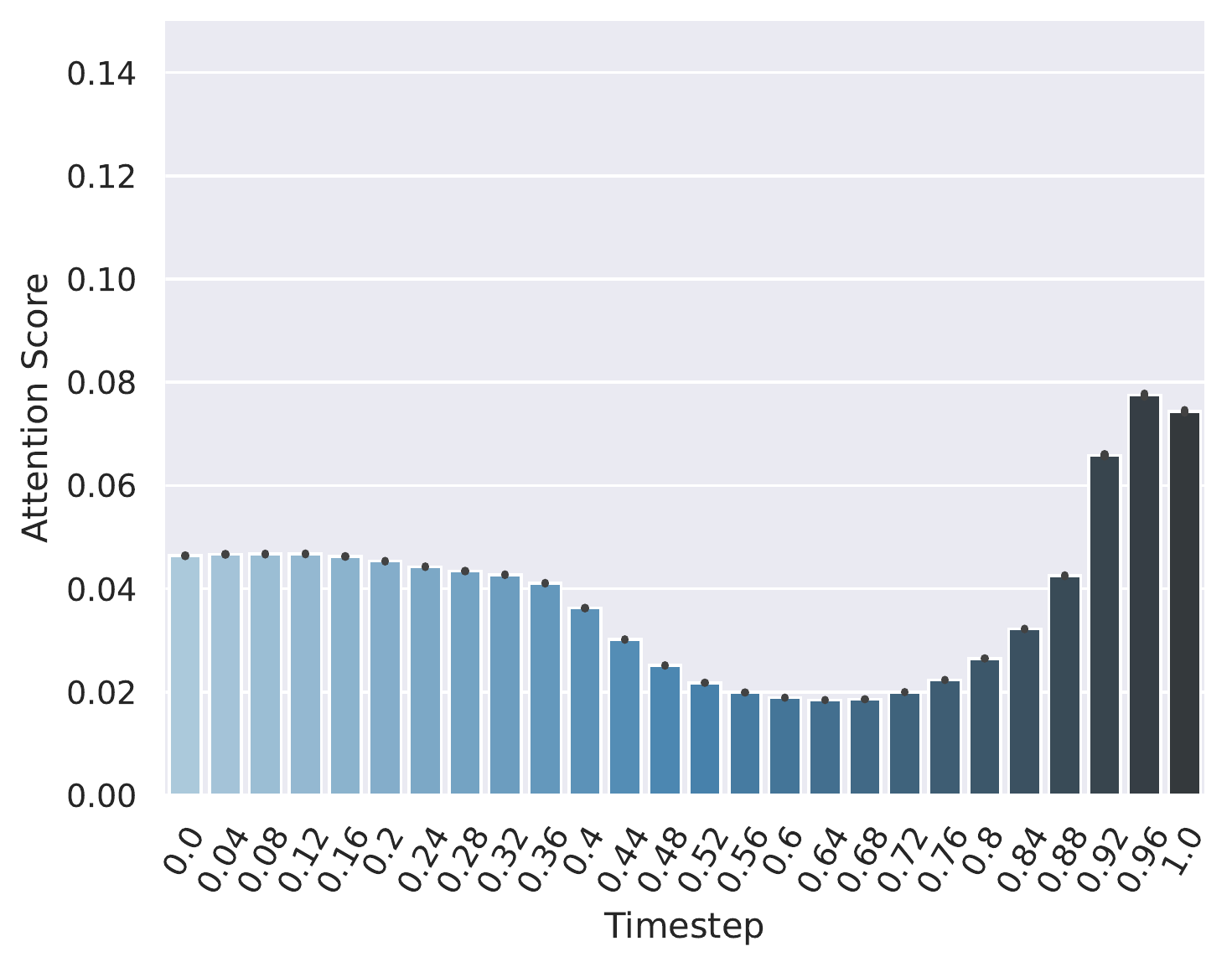}
\vspace{-6mm}
\subcaption*{\scriptsize Background Color}
\end{subfigure}
\begin{subfigure}[c]{0.24\textwidth}
\includegraphics[width=\textwidth]{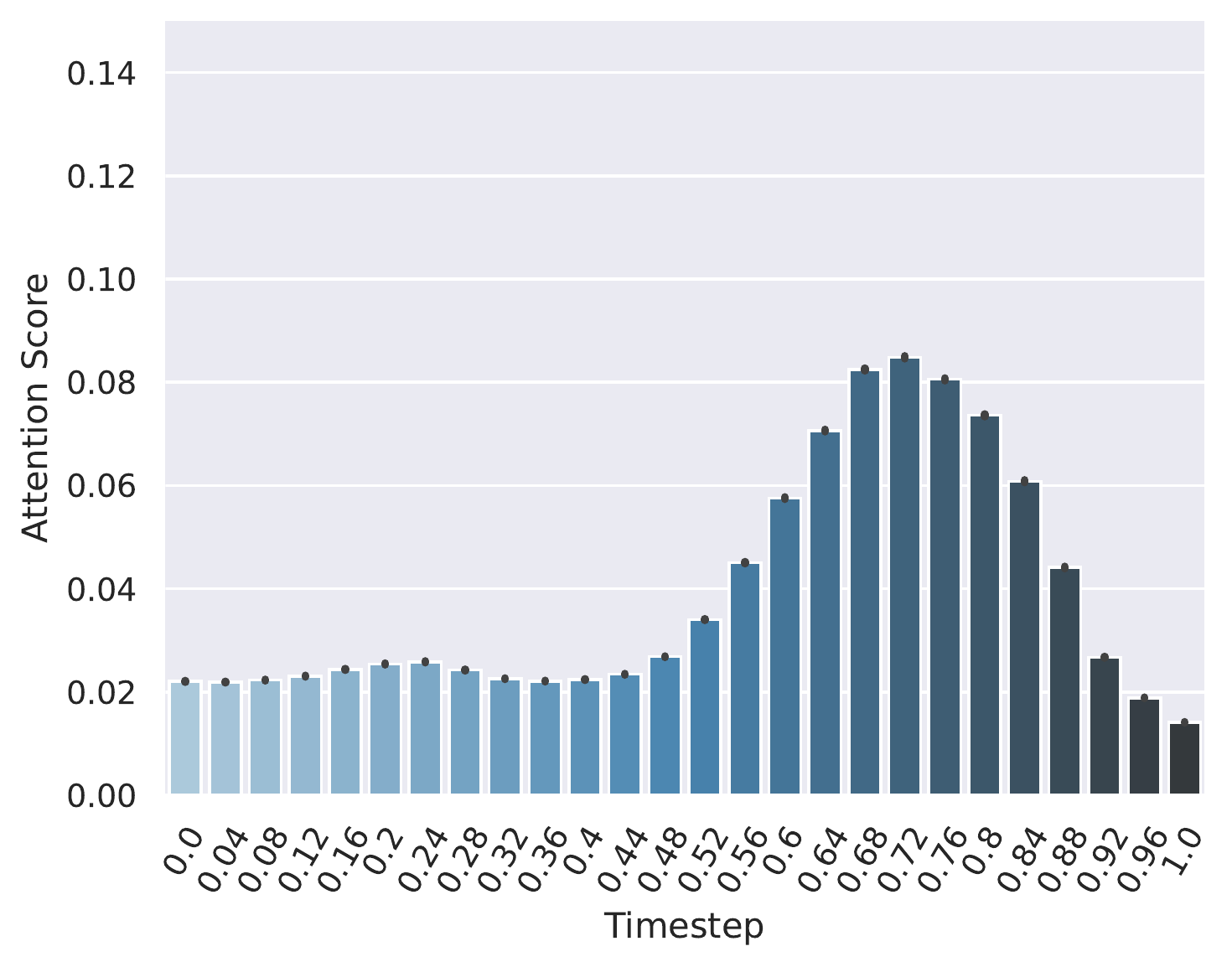}
\vspace{-6mm}
\subcaption*{\scriptsize Foreground Color}
\end{subfigure}
\begin{subfigure}[c]{0.24\textwidth}
\includegraphics[width=\textwidth]{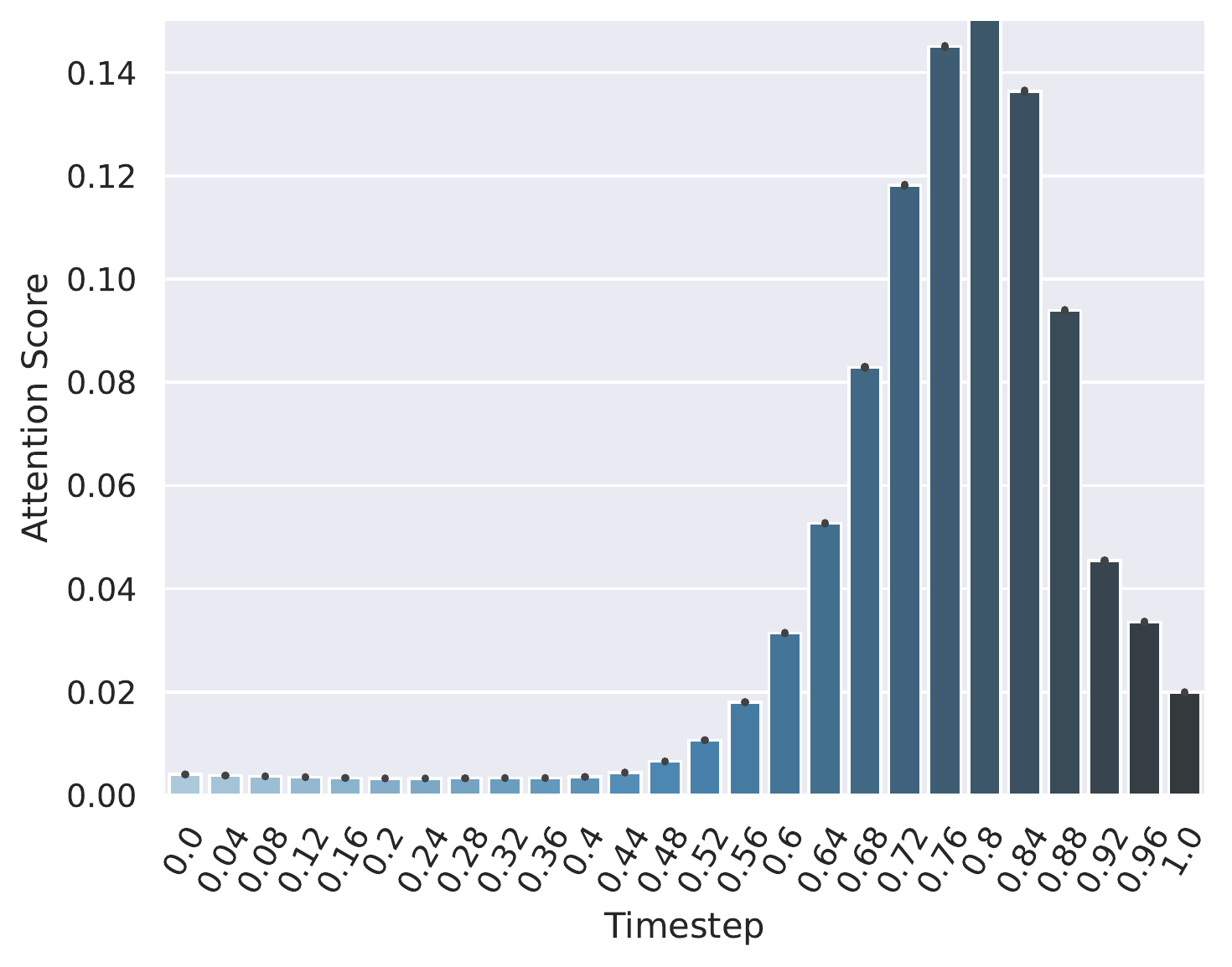}
\vspace{-6mm}
\subcaption*{\scriptsize Location}
\end{subfigure}
\begin{subfigure}[c]{0.24\textwidth}
\includegraphics[width=\textwidth]{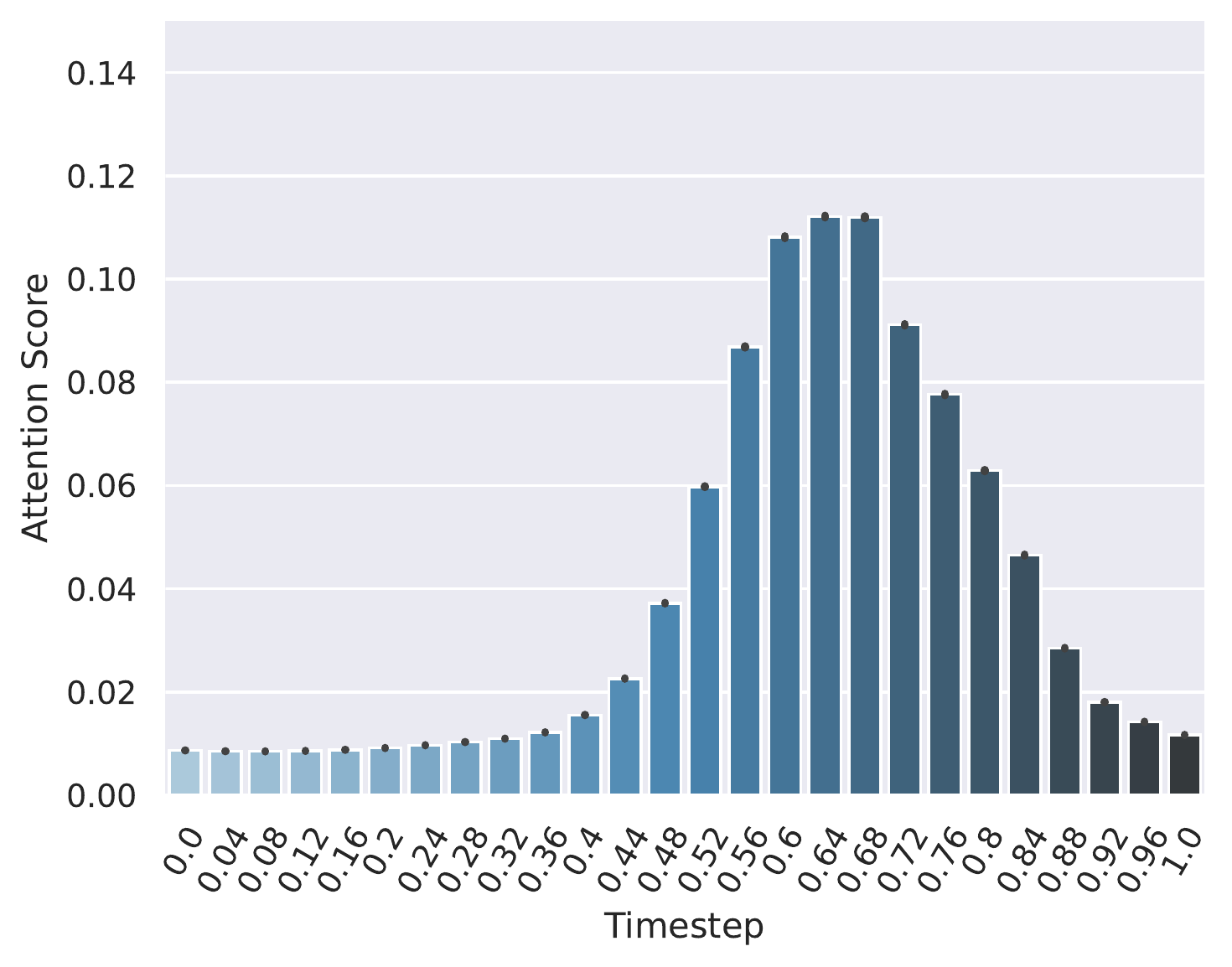}
\vspace{-6mm}
\subcaption*{\scriptsize Object Shape}
\end{subfigure}
\subcaption*{Granularity: 25}
\end{subfigure} \\
\caption{Attention score profiles for the synthetic dataset on the different features, using different granularities, with the dimensionality of the latent space as 32 and the VDRL encoder.}
\label{fig:syn_VDRL_32}
\end{figure}

\begin{figure}
\begin{subfigure}[c]{\textwidth}
\begin{subfigure}[c]{0.24\textwidth}
\includegraphics[width=\textwidth]{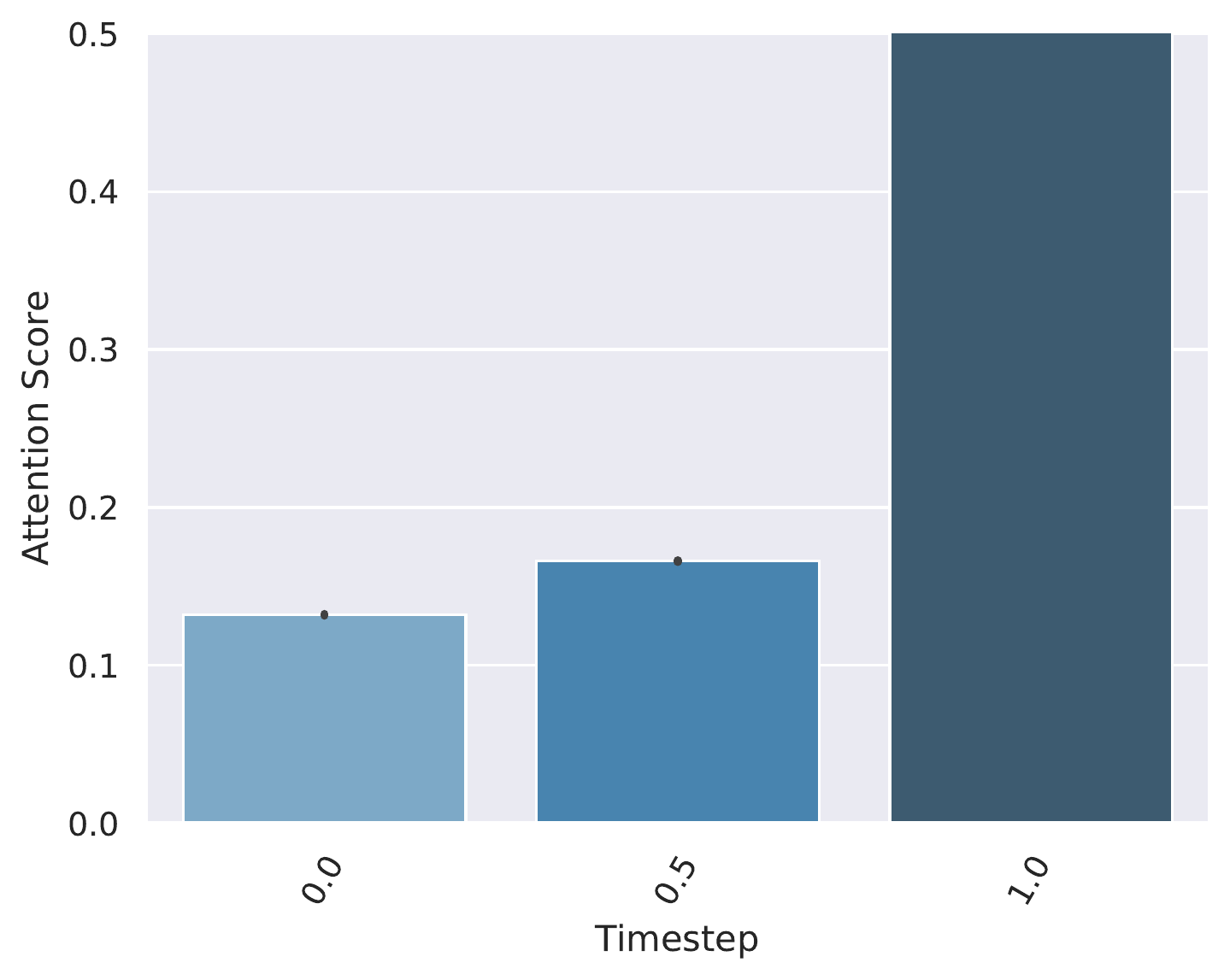}
\vspace{-6mm}
\subcaption*{\scriptsize Background Color}
\end{subfigure}
\begin{subfigure}[c]{0.24\textwidth}
\includegraphics[width=\textwidth]{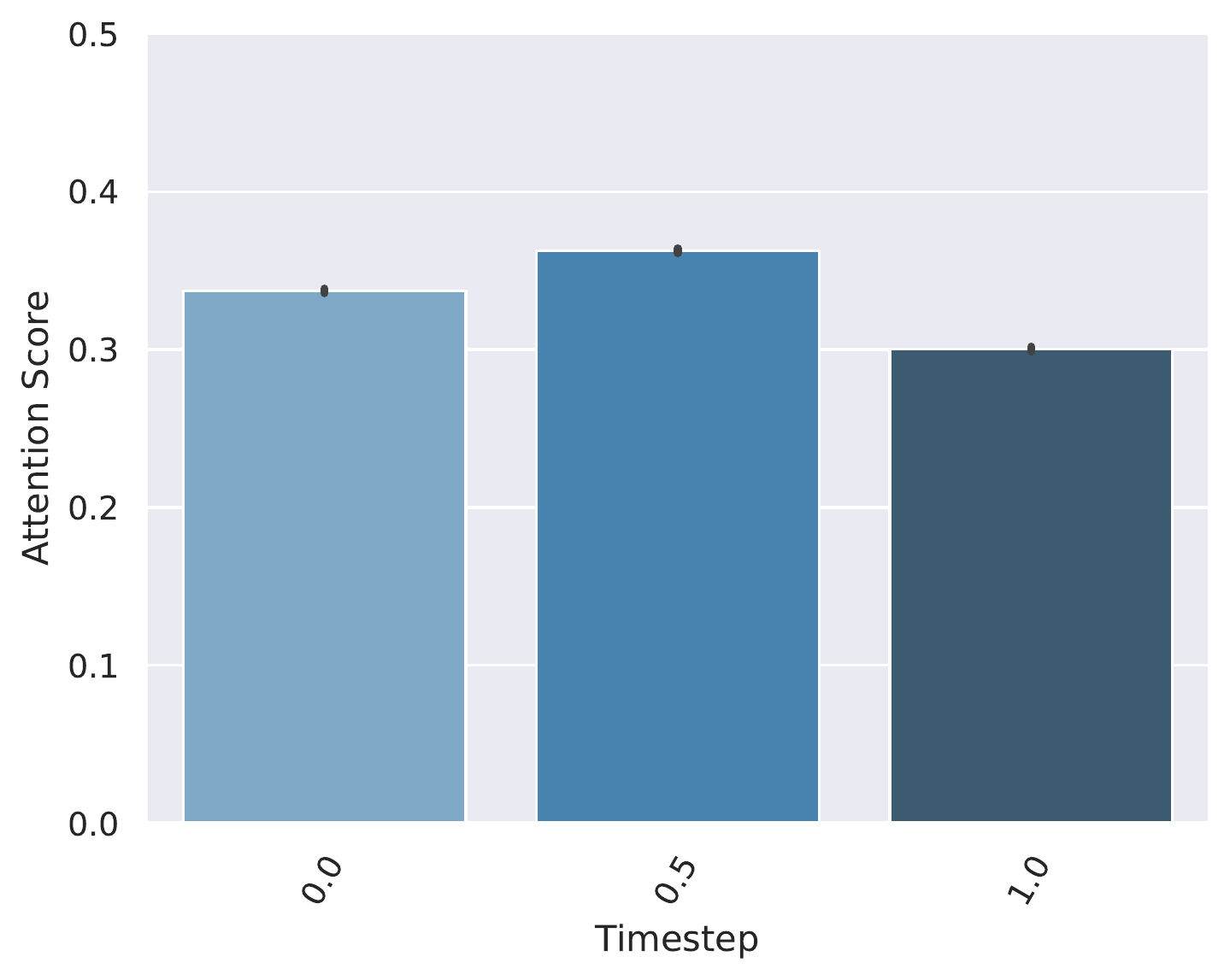}
\vspace{-6mm}
\subcaption*{\scriptsize Foreground Color}
\end{subfigure}
\begin{subfigure}[c]{0.24\textwidth}
\includegraphics[width=\textwidth]{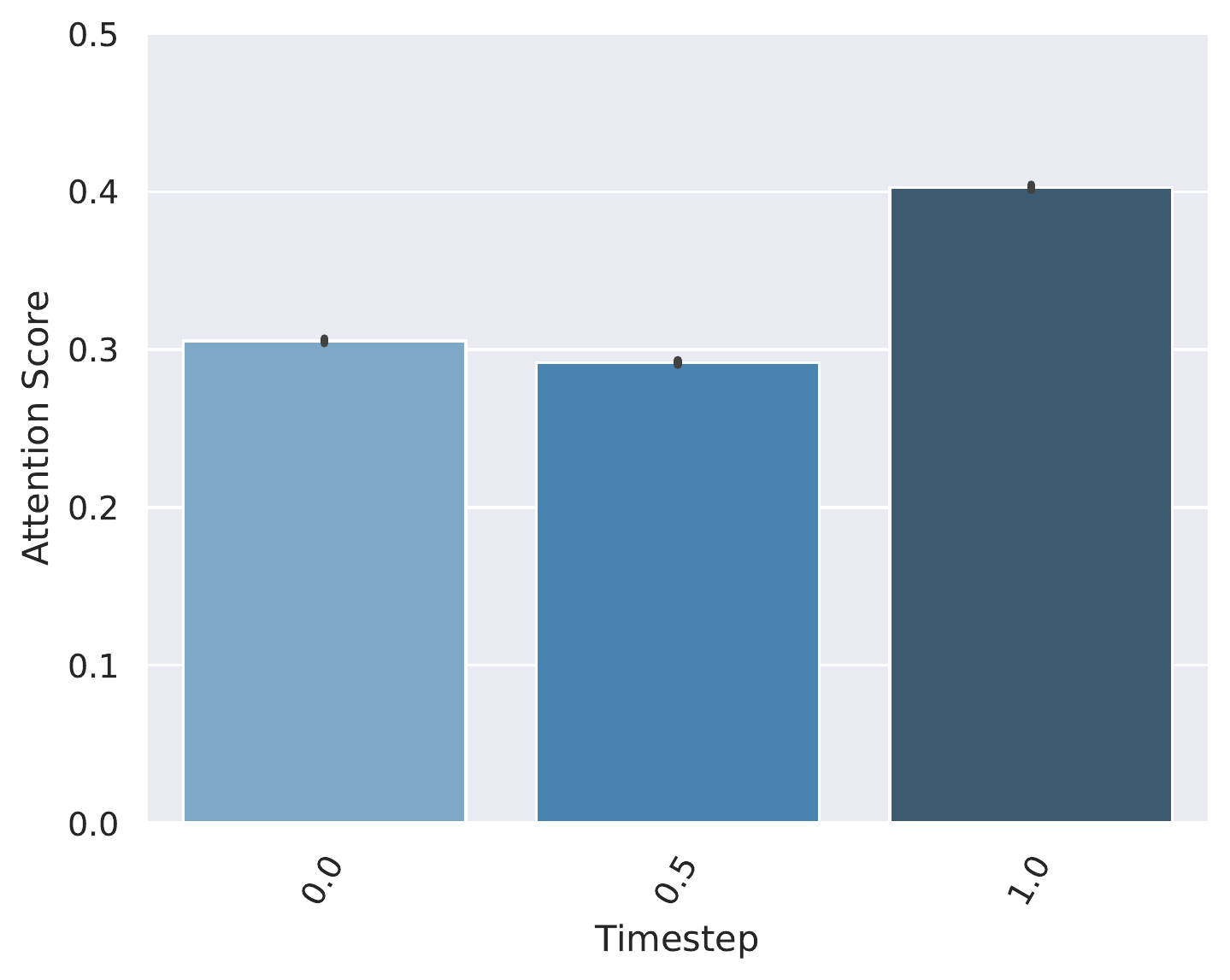}
\vspace{-6mm}
\subcaption*{\scriptsize Location}
\end{subfigure}
\begin{subfigure}[c]{0.24\textwidth}
\includegraphics[width=\textwidth]{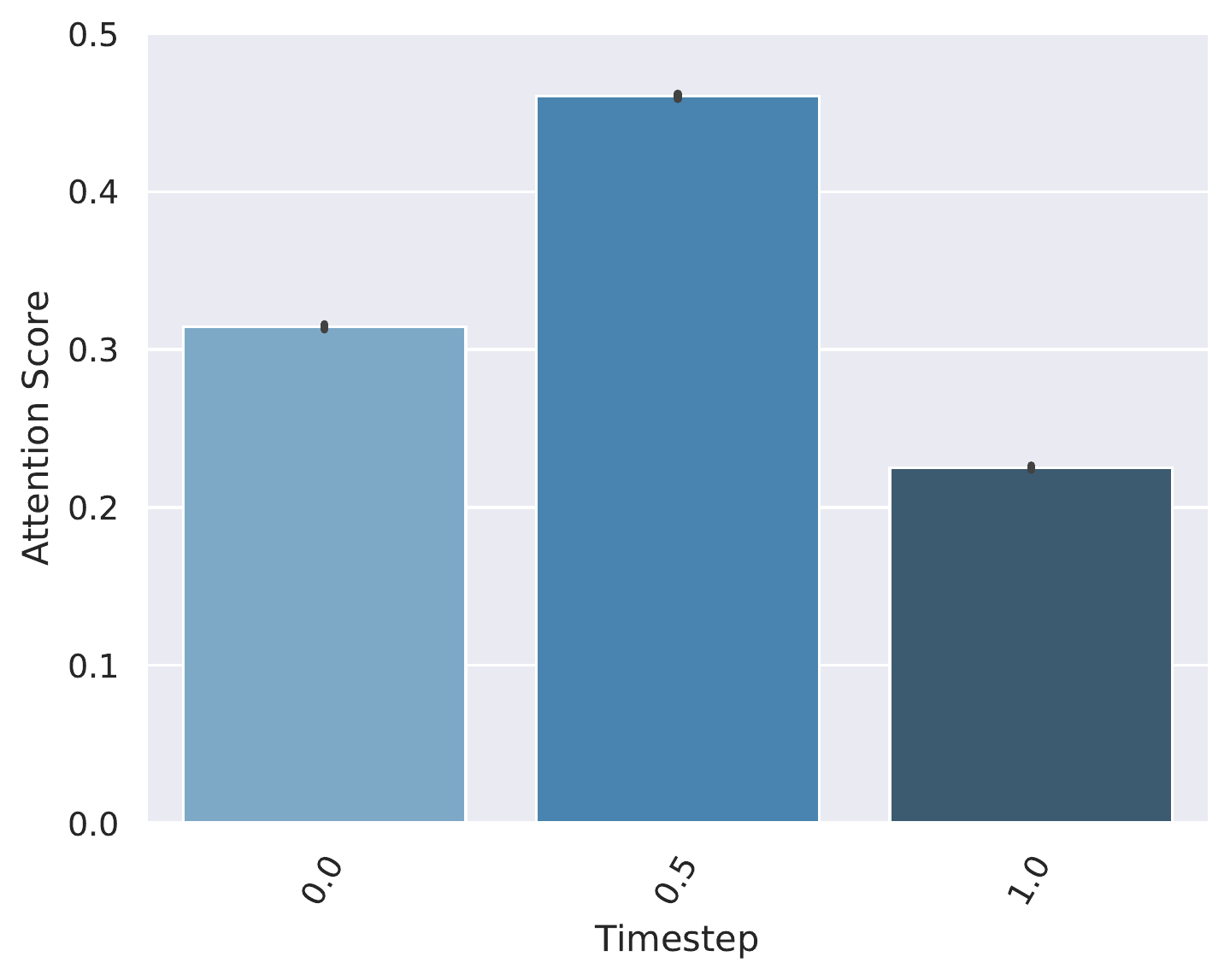}
\vspace{-6mm}
\subcaption*{\scriptsize Object Shape}
\end{subfigure}
\subcaption*{Granularity: 2}
\end{subfigure} \\
\begin{subfigure}[c]{\textwidth}
\begin{subfigure}[c]{0.24\textwidth}
\includegraphics[width=\textwidth]{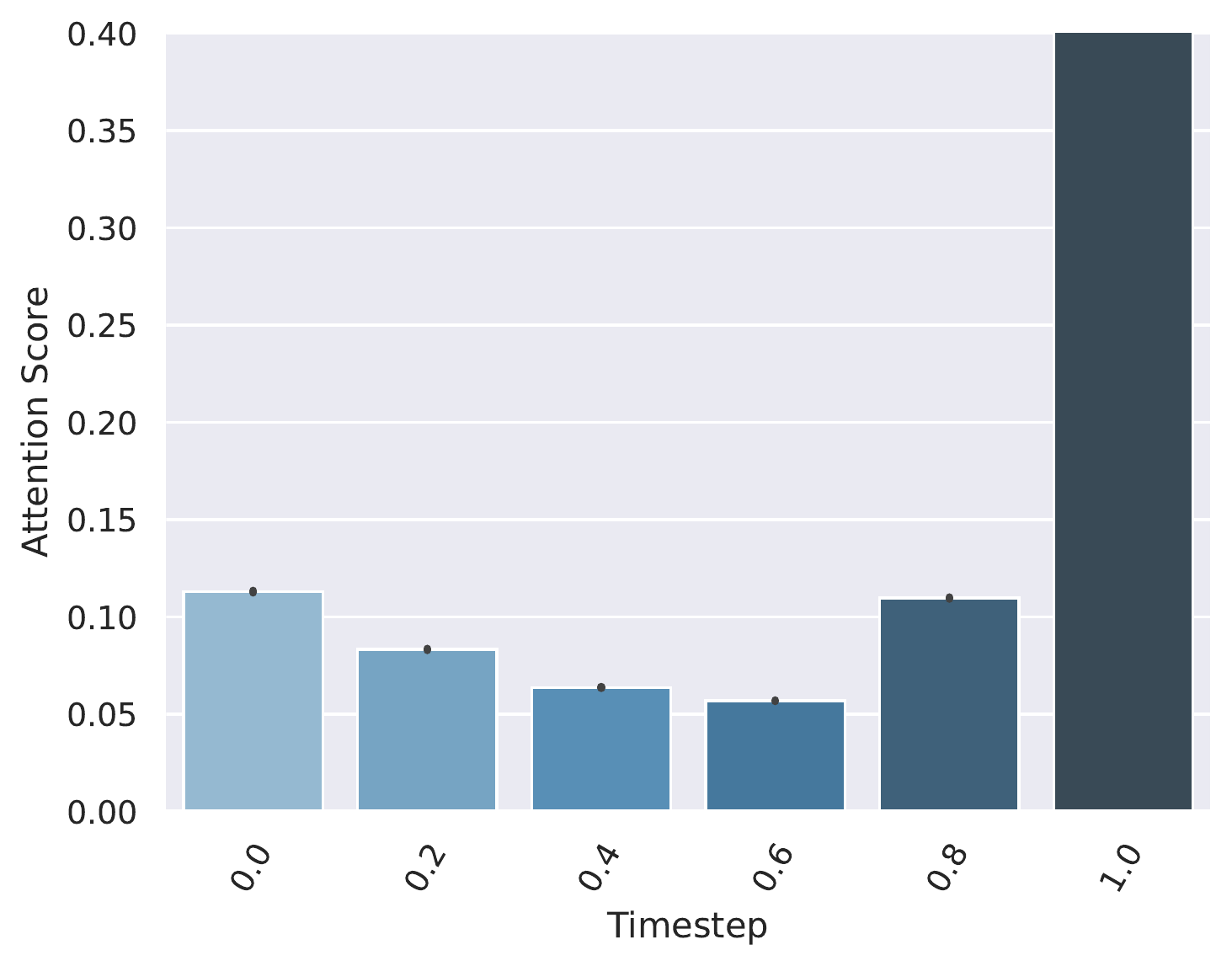}
\vspace{-6mm}
\subcaption*{\scriptsize Background Color}
\end{subfigure}
\begin{subfigure}[c]{0.24\textwidth}
\includegraphics[width=\textwidth]{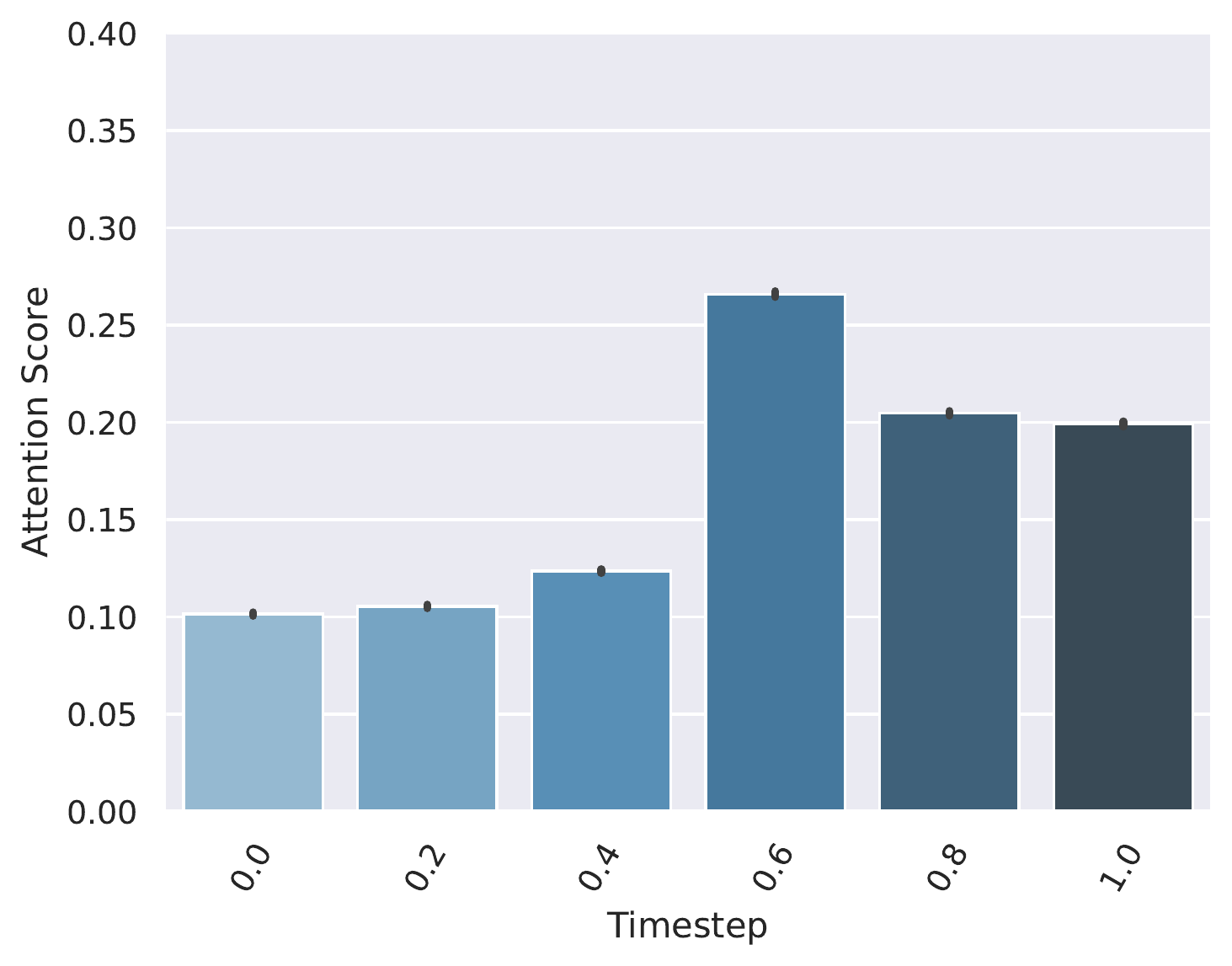}
\vspace{-6mm}
\subcaption*{\scriptsize Foreground Color}
\end{subfigure}
\begin{subfigure}[c]{0.24\textwidth}
\includegraphics[width=\textwidth]{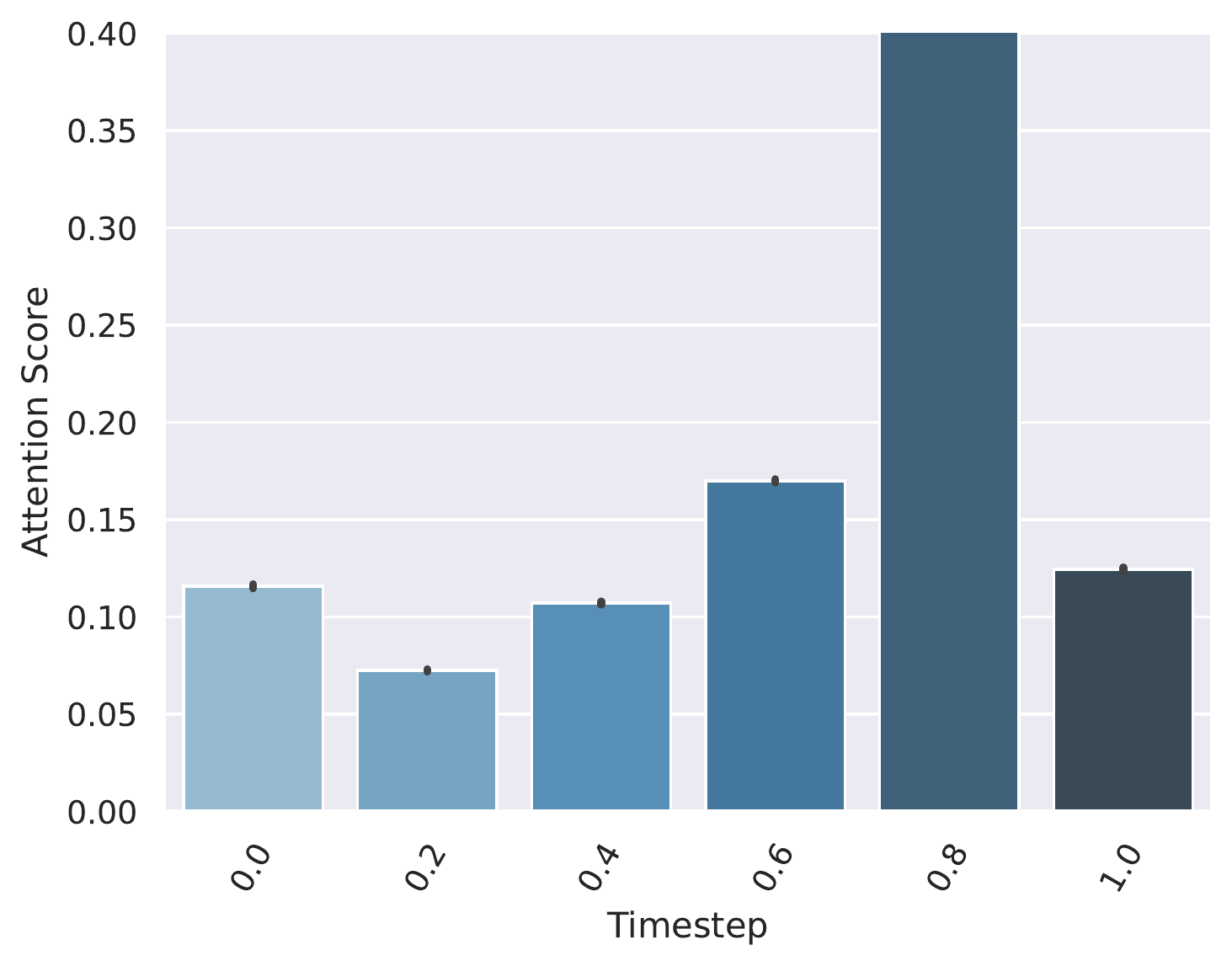}
\vspace{-6mm}
\subcaption*{\scriptsize Location}
\end{subfigure}
\begin{subfigure}[c]{0.24\textwidth}
\includegraphics[width=\textwidth]{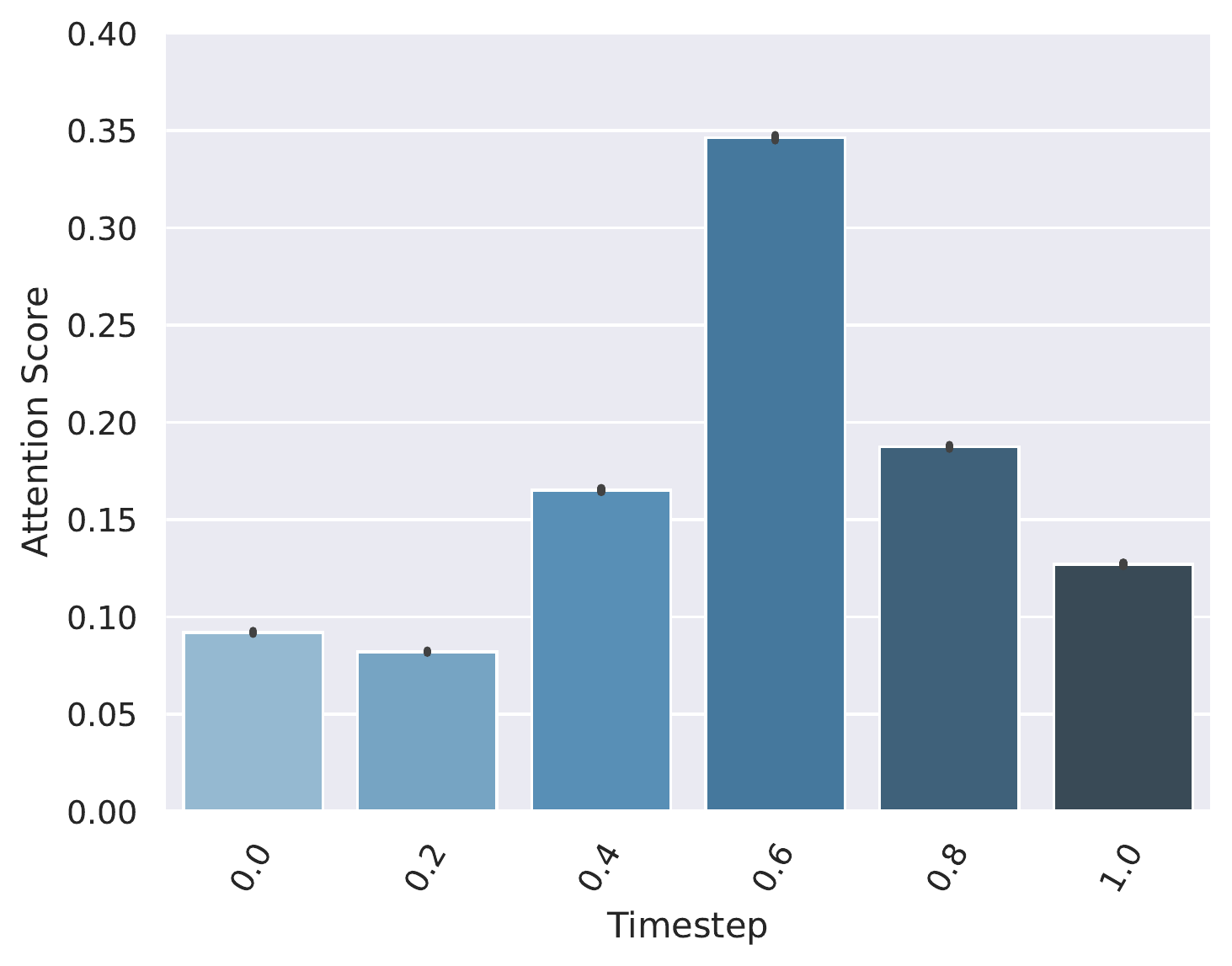}
\vspace{-6mm}
\subcaption*{\scriptsize Object Shape}
\end{subfigure}
\subcaption*{Granularity: 5}
\end{subfigure} \\
\begin{subfigure}[c]{\textwidth}
\begin{subfigure}[c]{0.24\textwidth}
\includegraphics[width=\textwidth]{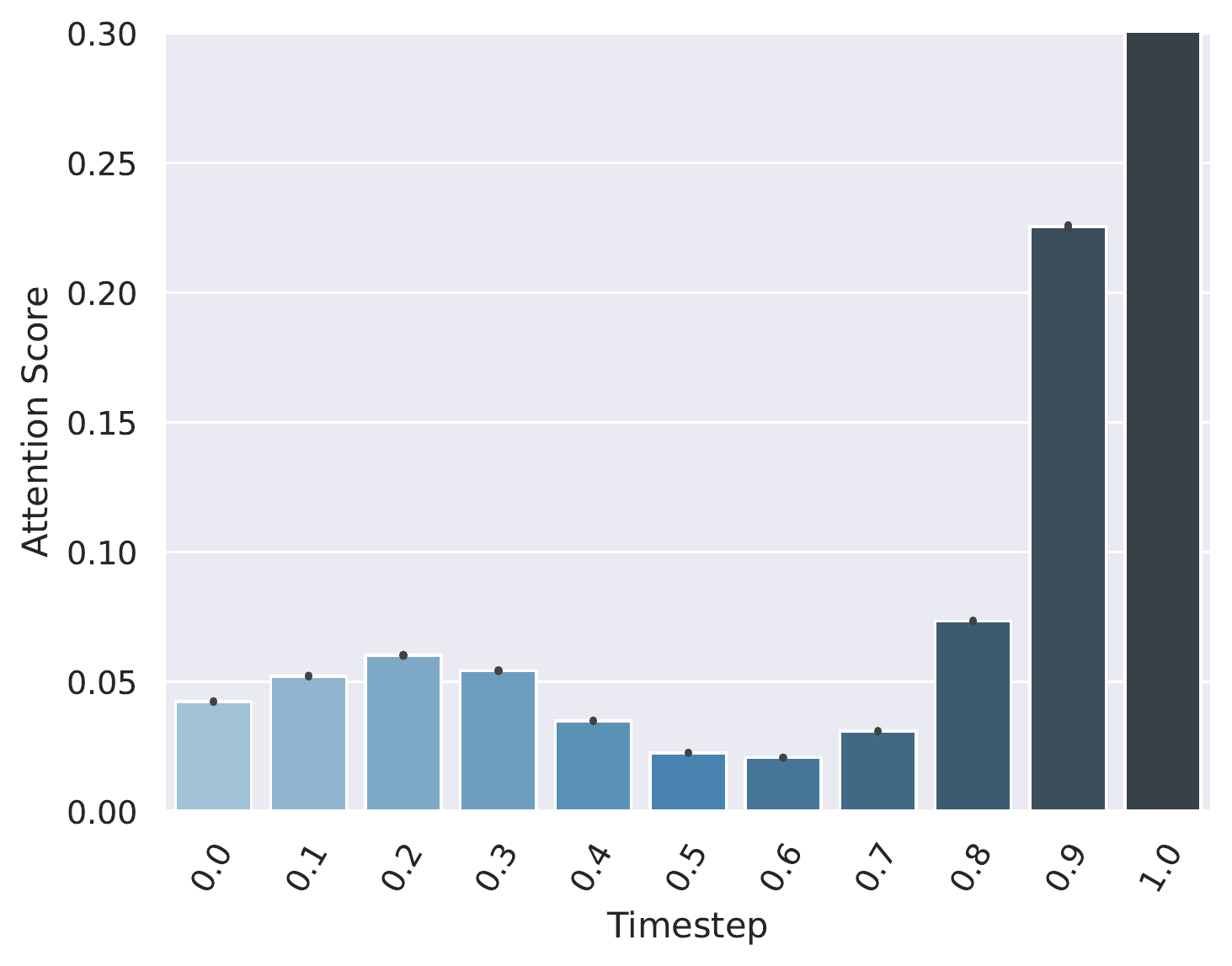}
\vspace{-6mm}
\subcaption*{\scriptsize Background Color}
\end{subfigure}
\begin{subfigure}[c]{0.24\textwidth}
\includegraphics[width=\textwidth]{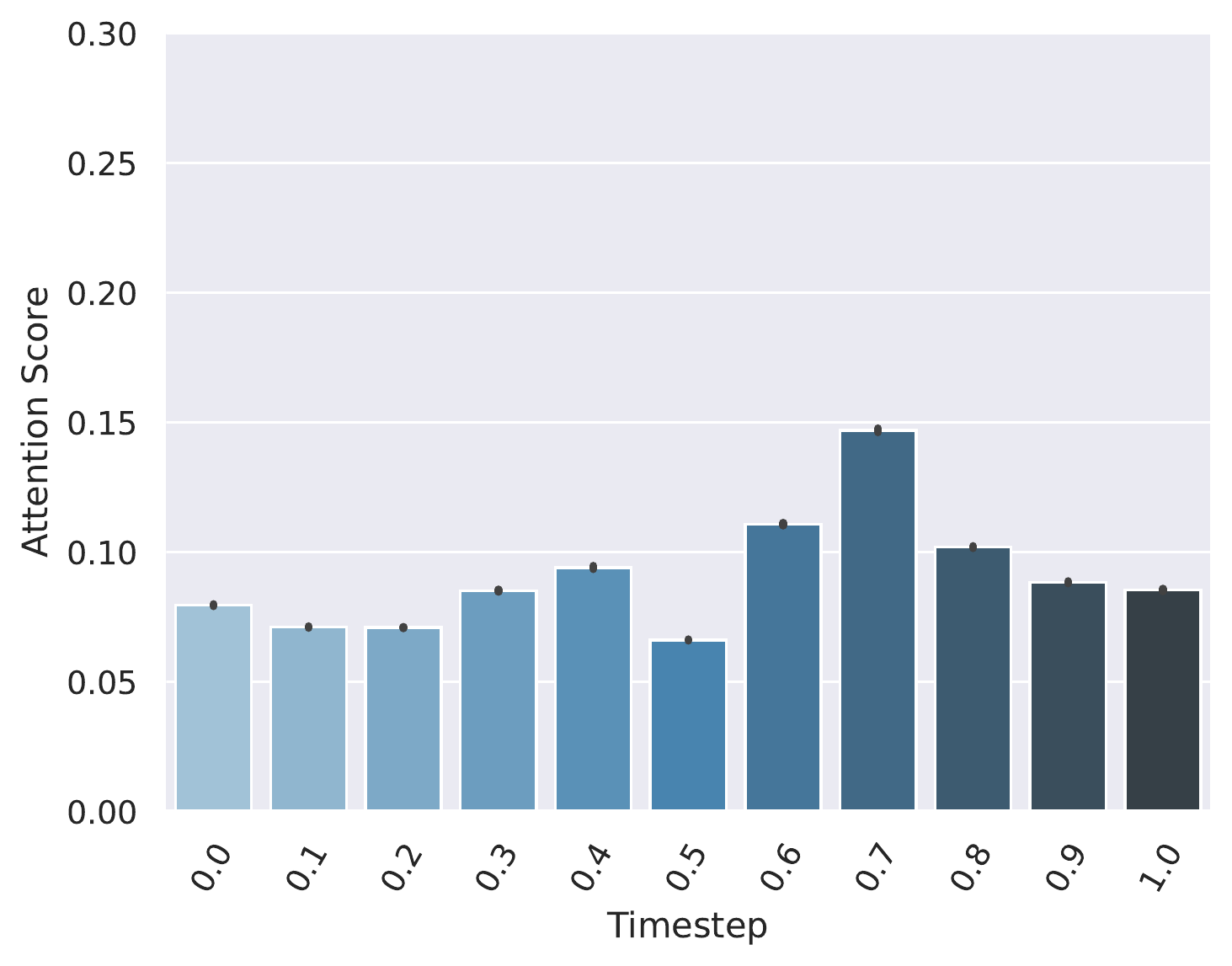}
\vspace{-6mm}
\subcaption*{\scriptsize Foreground Color}
\end{subfigure}
\begin{subfigure}[c]{0.24\textwidth}
\includegraphics[width=\textwidth]{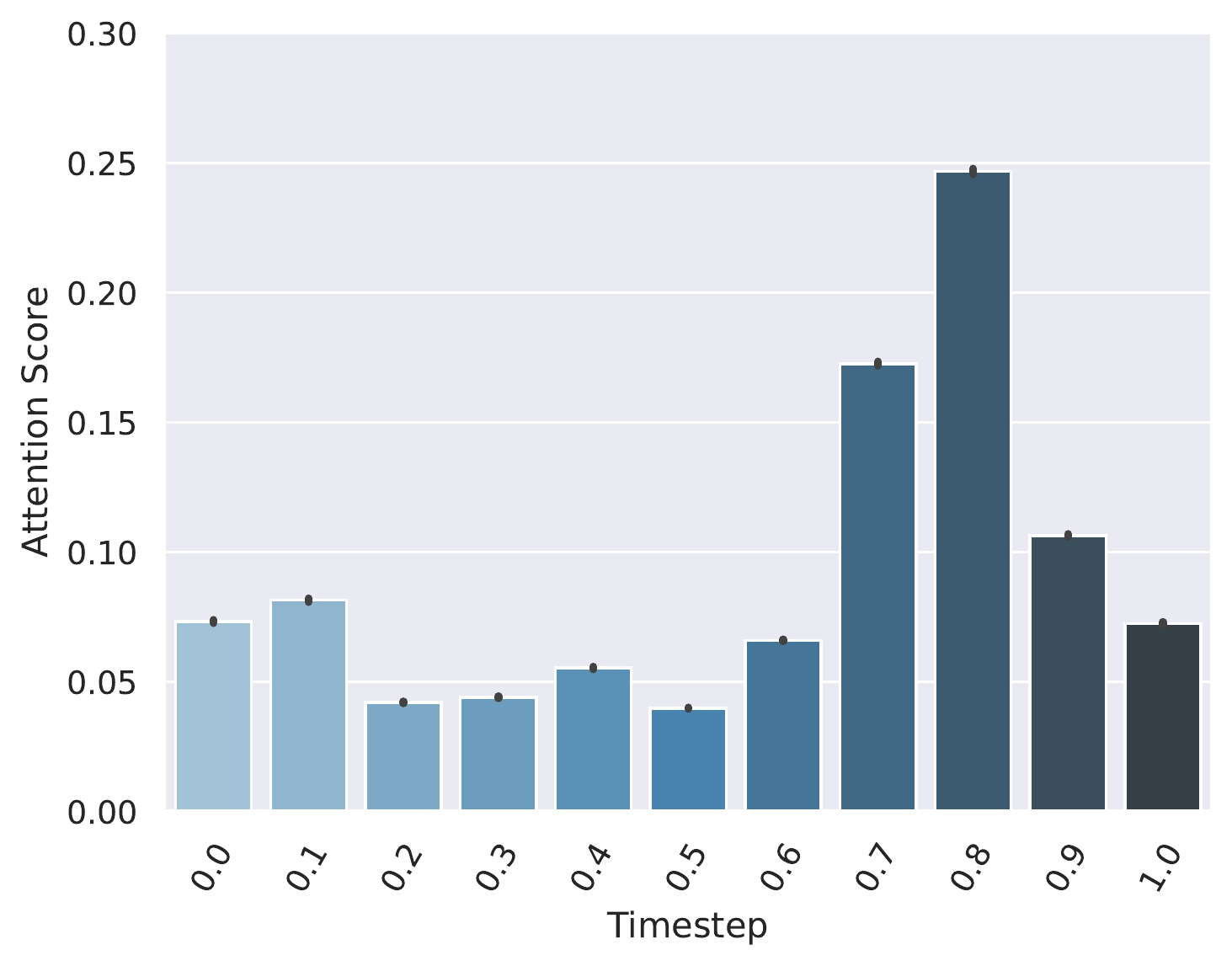}
\vspace{-6mm}
\subcaption*{\scriptsize Location}
\end{subfigure}
\begin{subfigure}[c]{0.24\textwidth}
\includegraphics[width=\textwidth]{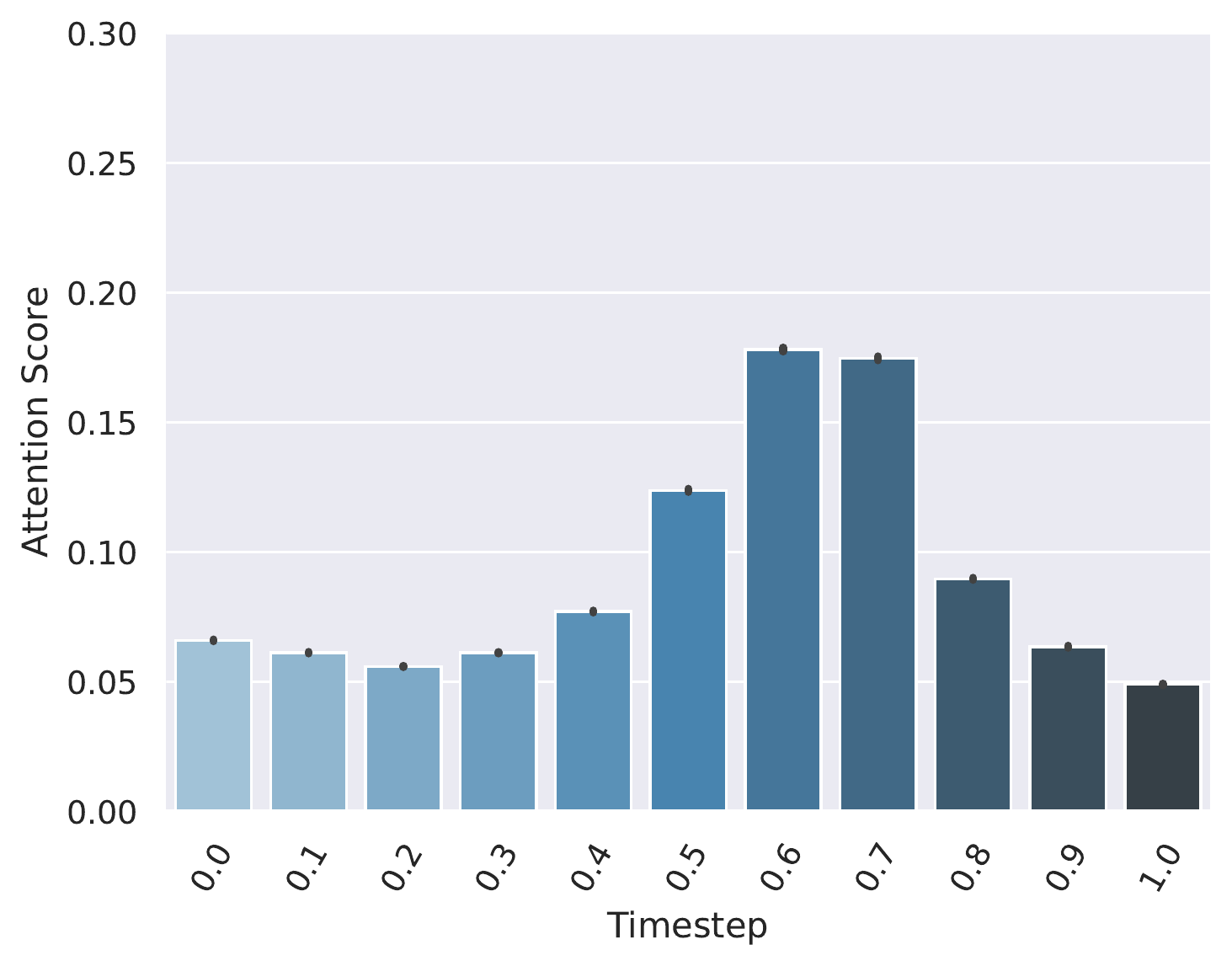}
\vspace{-6mm}
\subcaption*{\scriptsize Object Shape}
\end{subfigure}
\subcaption*{Granularity: 10}
\end{subfigure} \\
\begin{subfigure}[c]{\textwidth}
\begin{subfigure}[c]{0.24\textwidth}
\includegraphics[width=\textwidth]{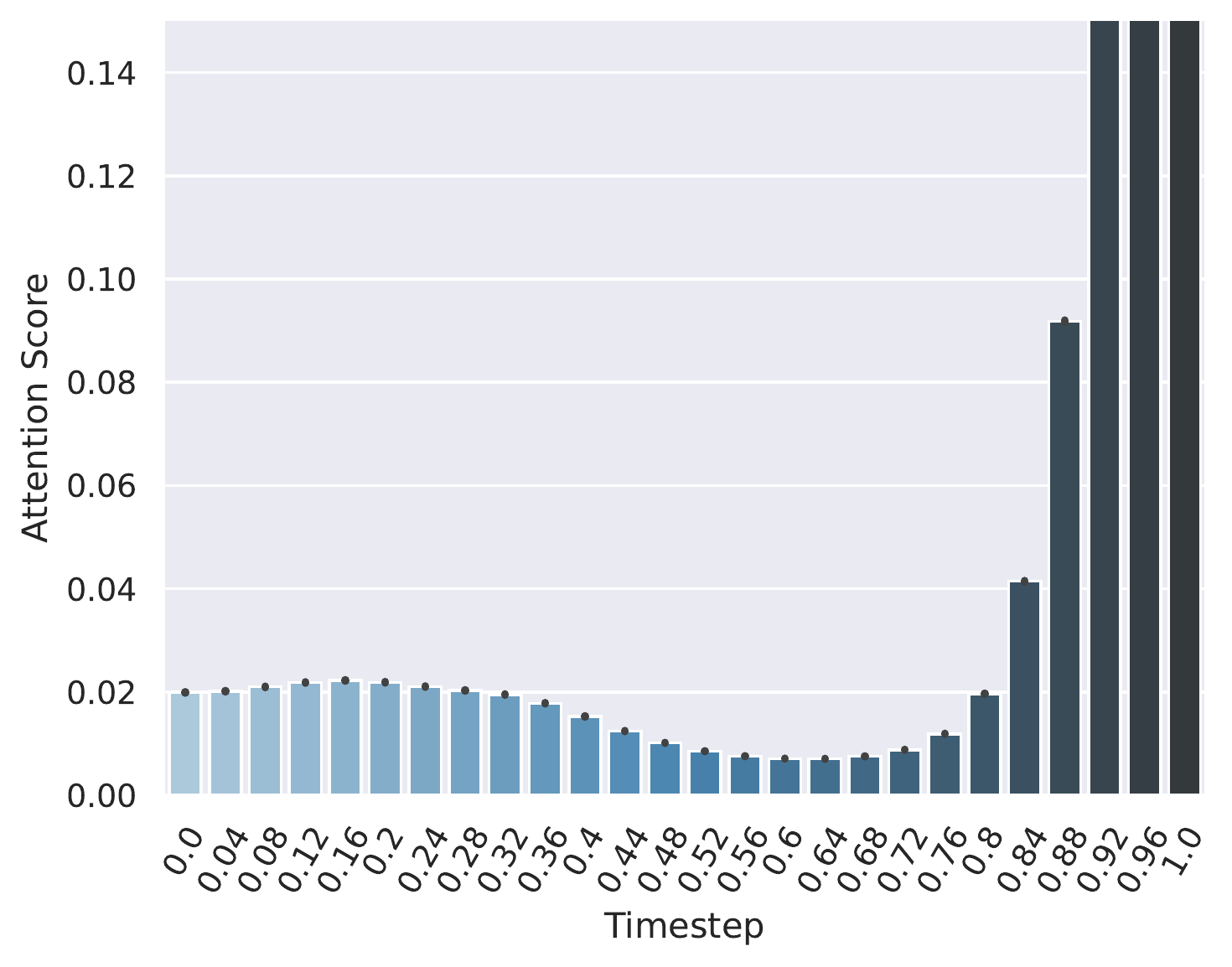}
\vspace{-6mm}
\subcaption*{\scriptsize Background Color}
\end{subfigure}
\begin{subfigure}[c]{0.24\textwidth}
\includegraphics[width=\textwidth]{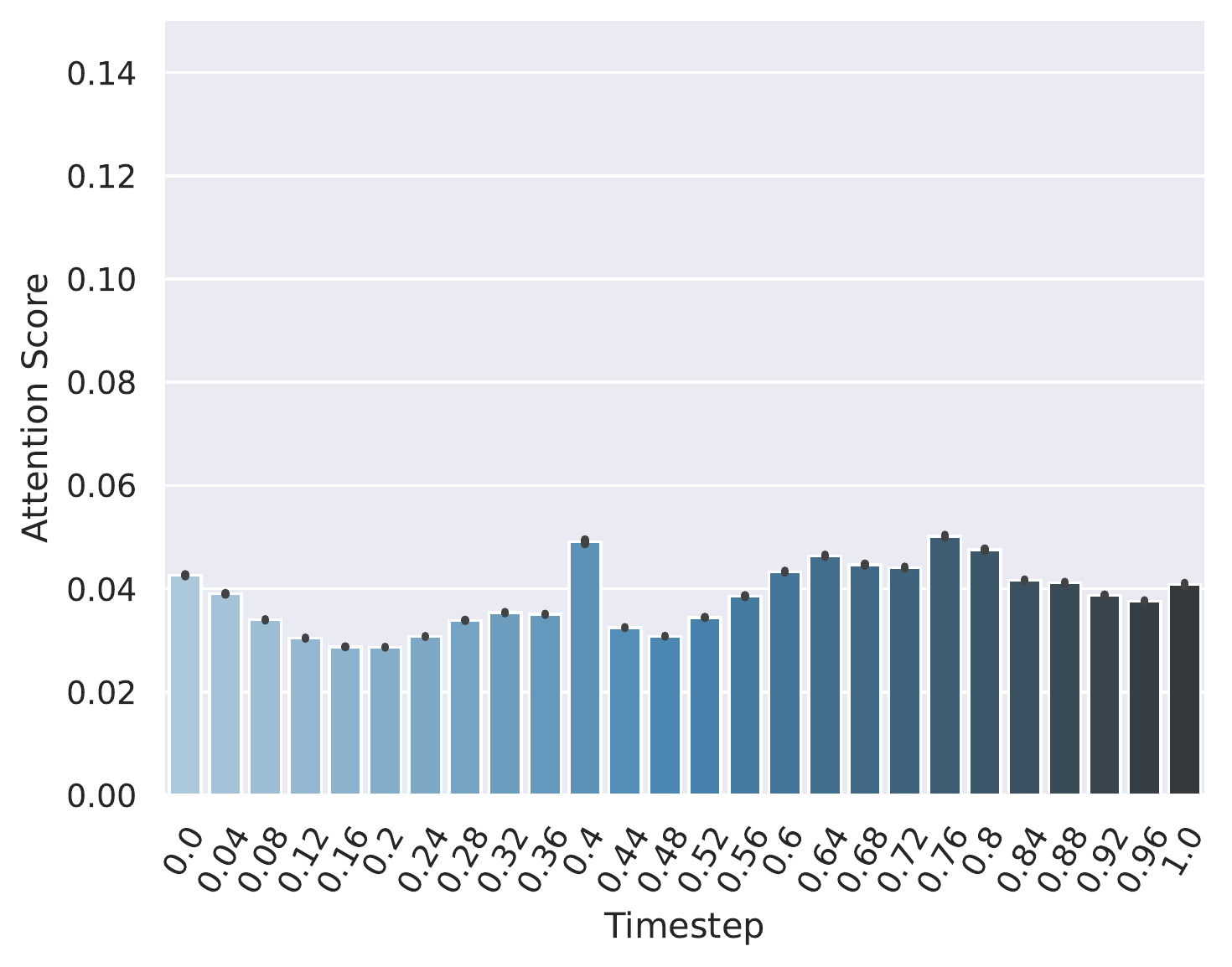}
\vspace{-6mm}
\subcaption*{\scriptsize Foreground Color}
\end{subfigure}
\begin{subfigure}[c]{0.24\textwidth}
\includegraphics[width=\textwidth]{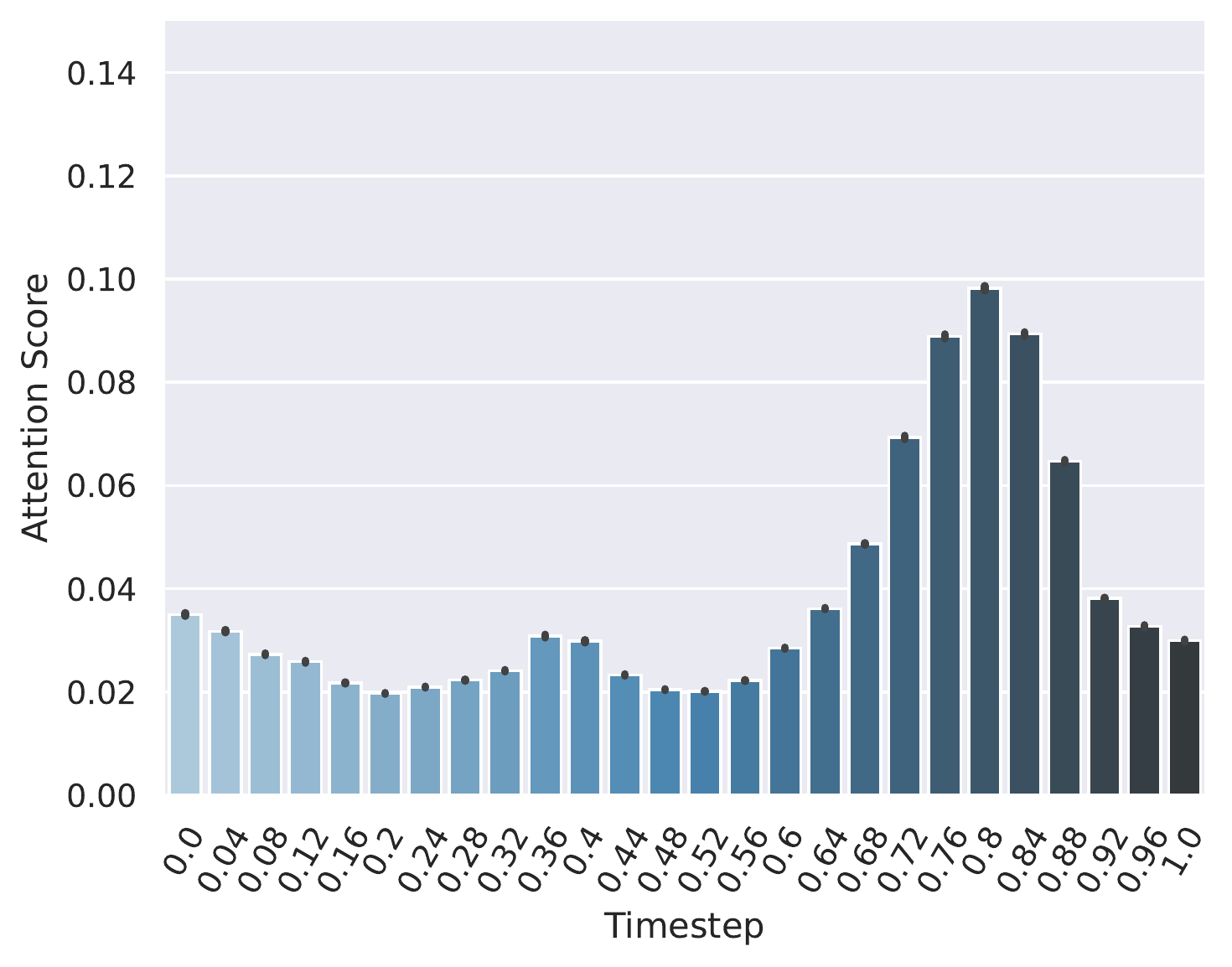}
\vspace{-6mm}
\subcaption*{\scriptsize Location}
\end{subfigure}
\begin{subfigure}[c]{0.24\textwidth}
\includegraphics[width=\textwidth]{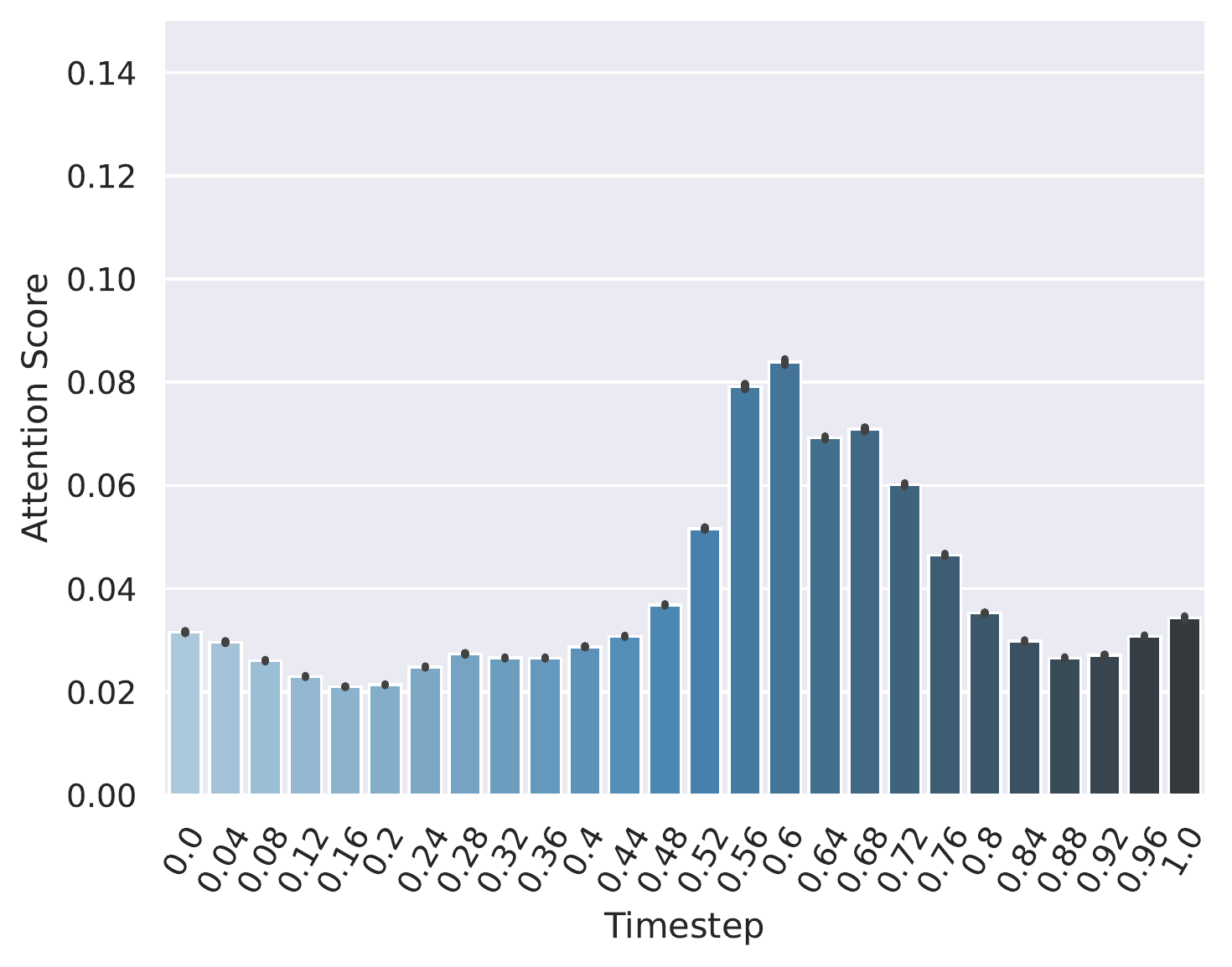}
\vspace{-6mm}
\subcaption*{\scriptsize Object Shape}
\end{subfigure}
\subcaption*{Granularity: 25}
\end{subfigure} \\
\caption{Attention score profiles for the synthetic dataset on the different features, using different granularities, with the dimensionality of the latent space as 2 and the DRL encoder.}
\label{fig:syn_DRL_2}
\end{figure}

\begin{figure}
\begin{subfigure}[c]{\textwidth}
\begin{subfigure}[c]{0.24\textwidth}
\includegraphics[width=\textwidth]{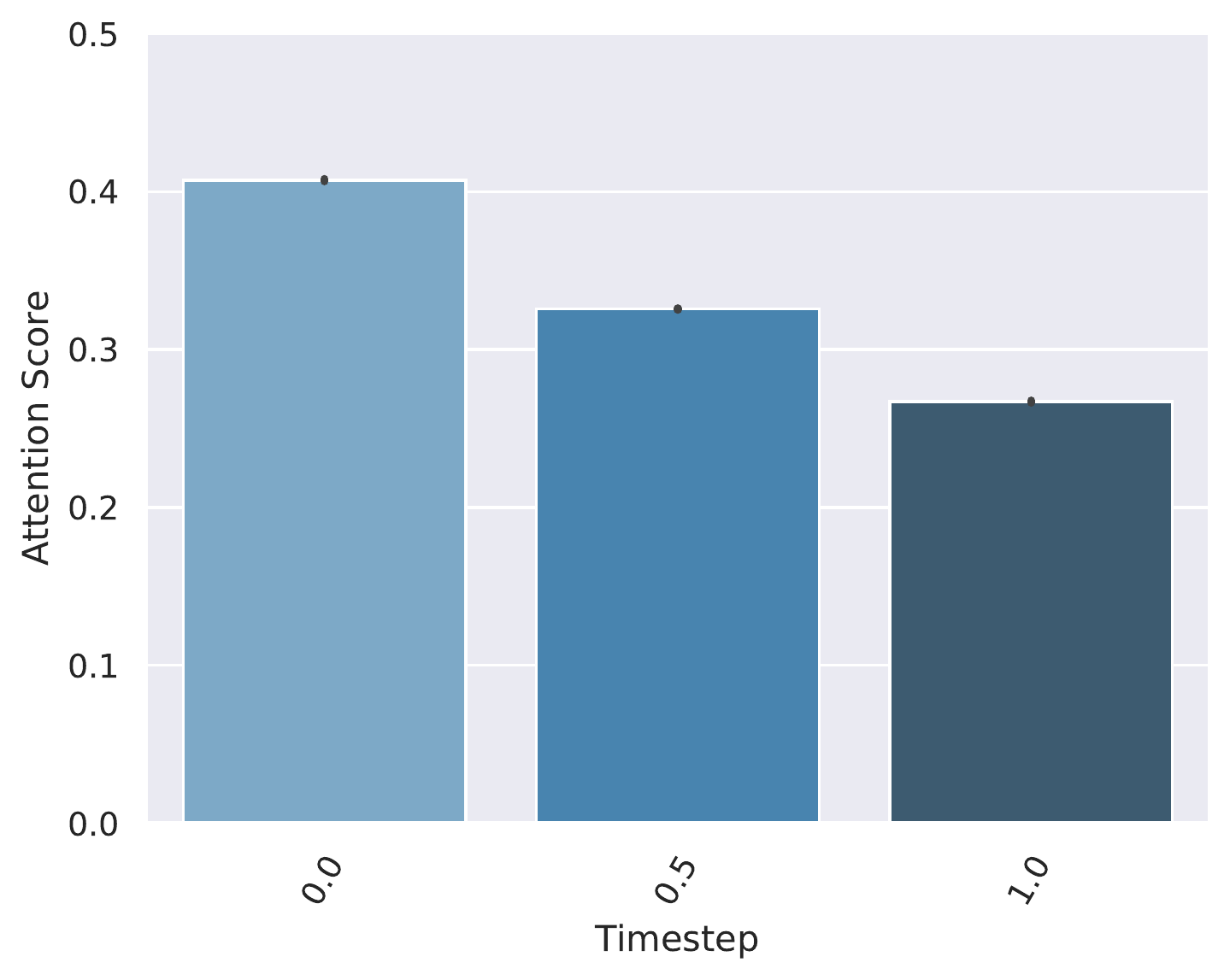}
\vspace{-6mm}
\subcaption*{\scriptsize Background Color}
\end{subfigure}
\begin{subfigure}[c]{0.24\textwidth}
\includegraphics[width=\textwidth]{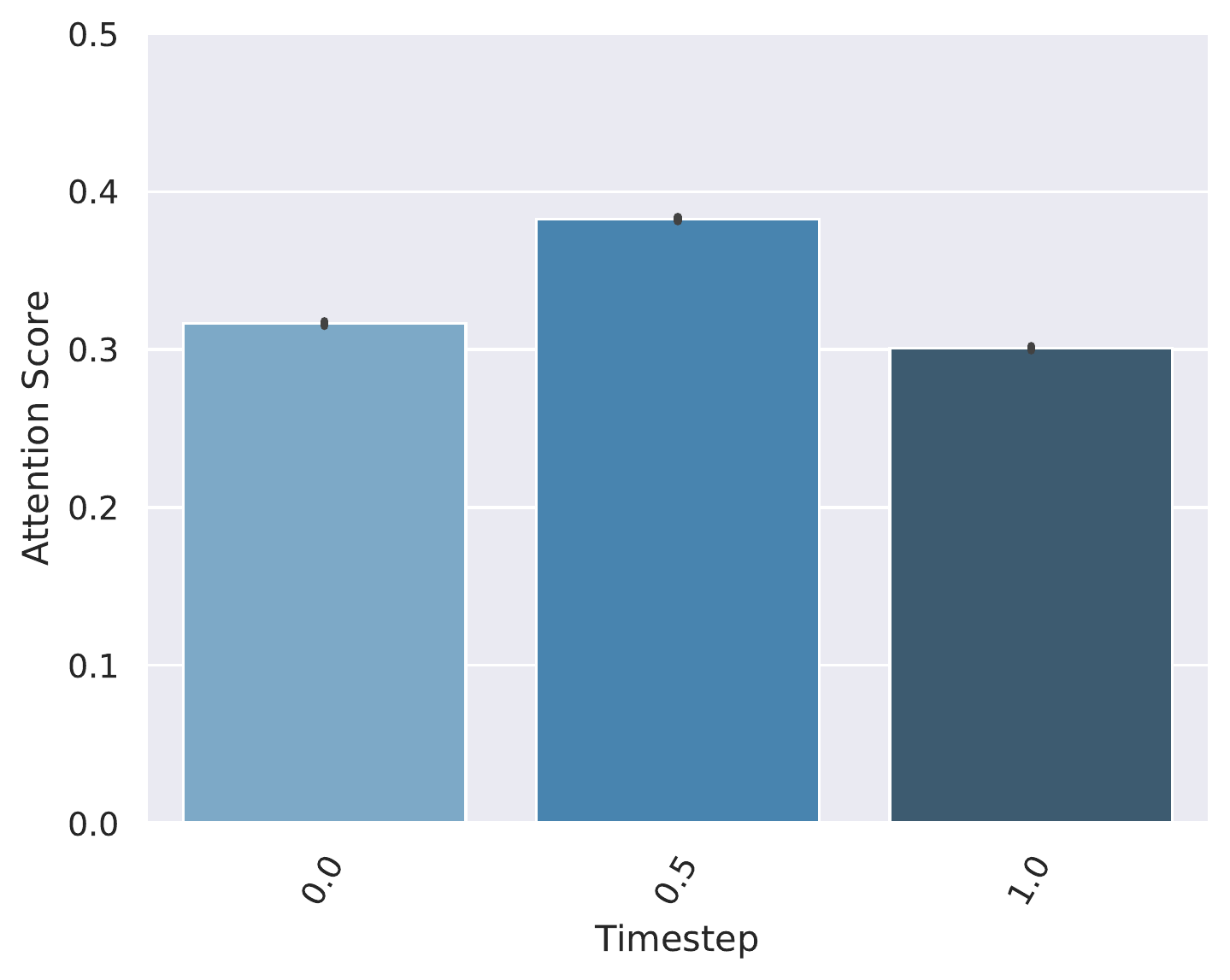}
\vspace{-6mm}
\subcaption*{\scriptsize Foreground Color}
\end{subfigure}
\begin{subfigure}[c]{0.24\textwidth}
\includegraphics[width=\textwidth]{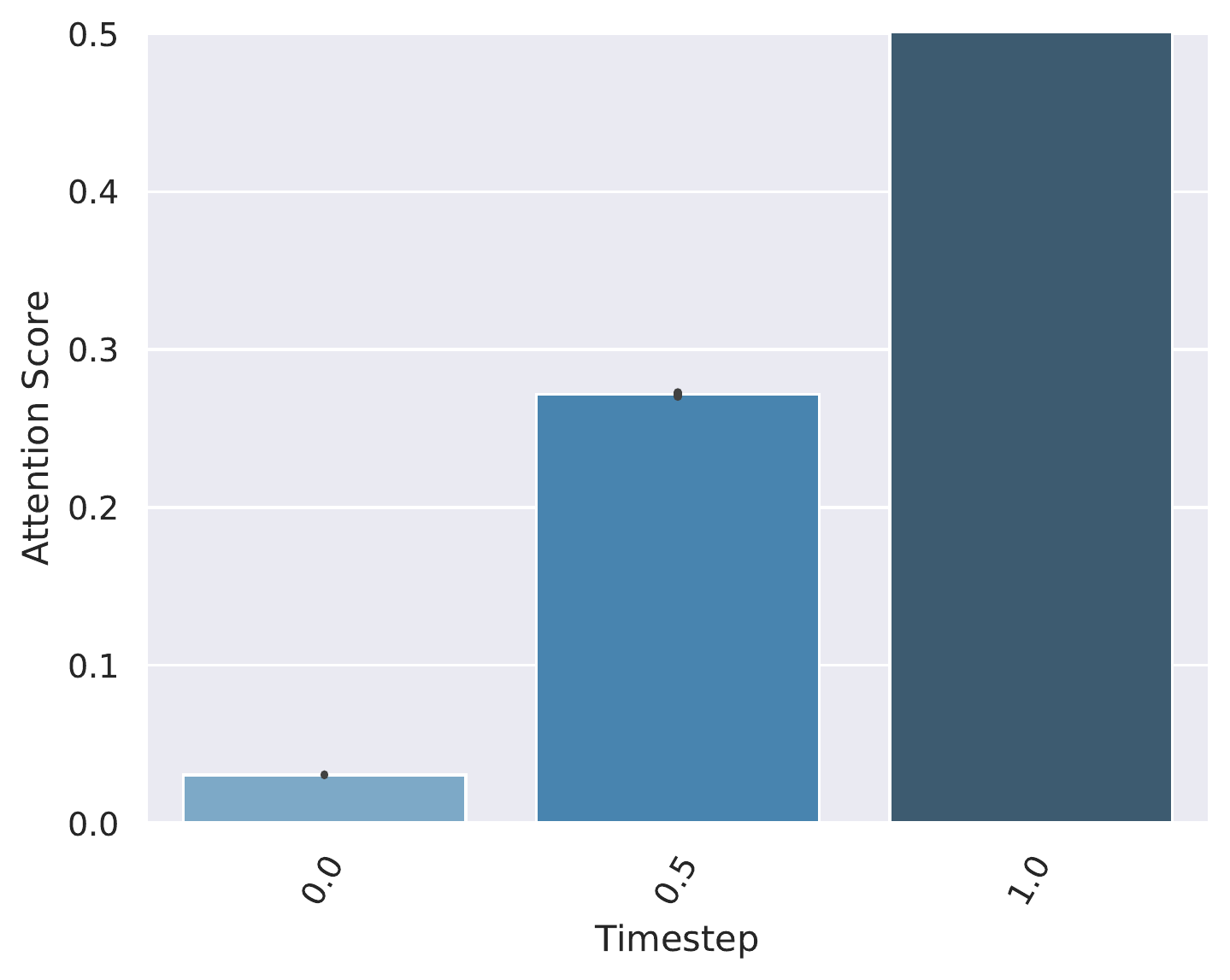}
\vspace{-6mm}
\subcaption*{\scriptsize Location}
\end{subfigure}
\begin{subfigure}[c]{0.24\textwidth}
\includegraphics[width=\textwidth]{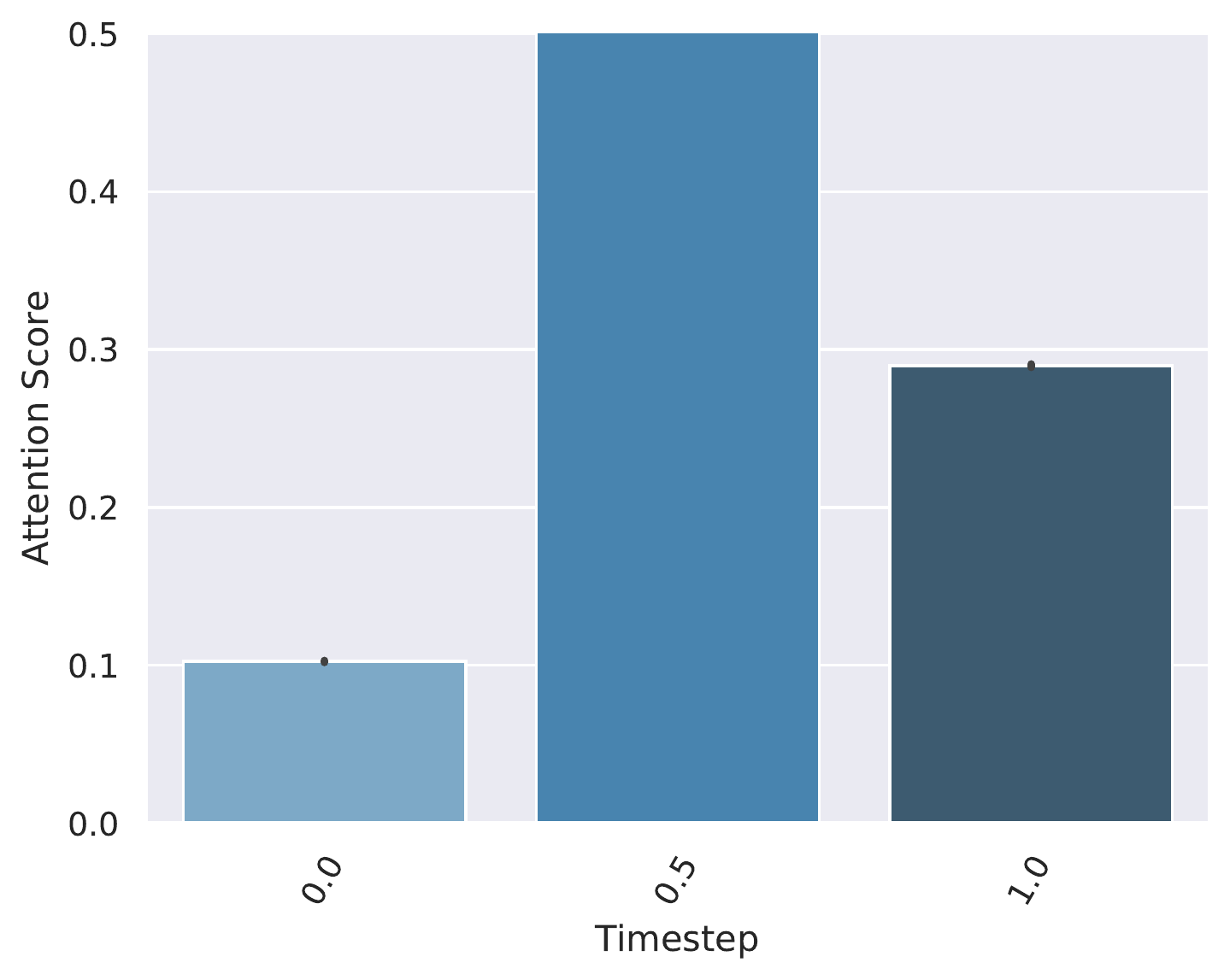}
\vspace{-6mm}
\subcaption*{\scriptsize Object Shape}
\end{subfigure}
\subcaption*{Granularity: 2}
\end{subfigure} \\
\begin{subfigure}[c]{\textwidth}
\begin{subfigure}[c]{0.24\textwidth}
\includegraphics[width=\textwidth]{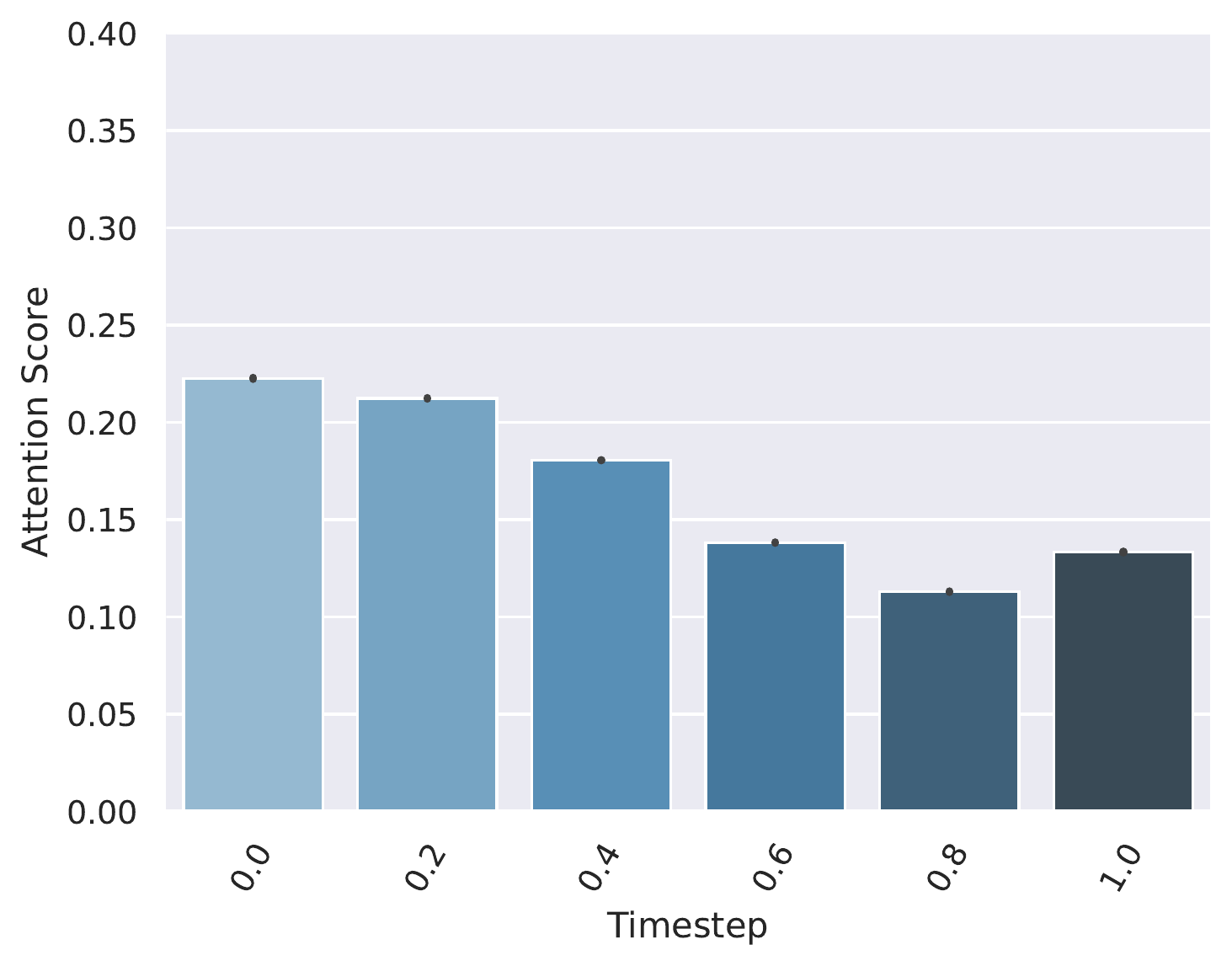}
\vspace{-6mm}
\subcaption*{\scriptsize Background Color}
\end{subfigure}
\begin{subfigure}[c]{0.24\textwidth}
\includegraphics[width=\textwidth]{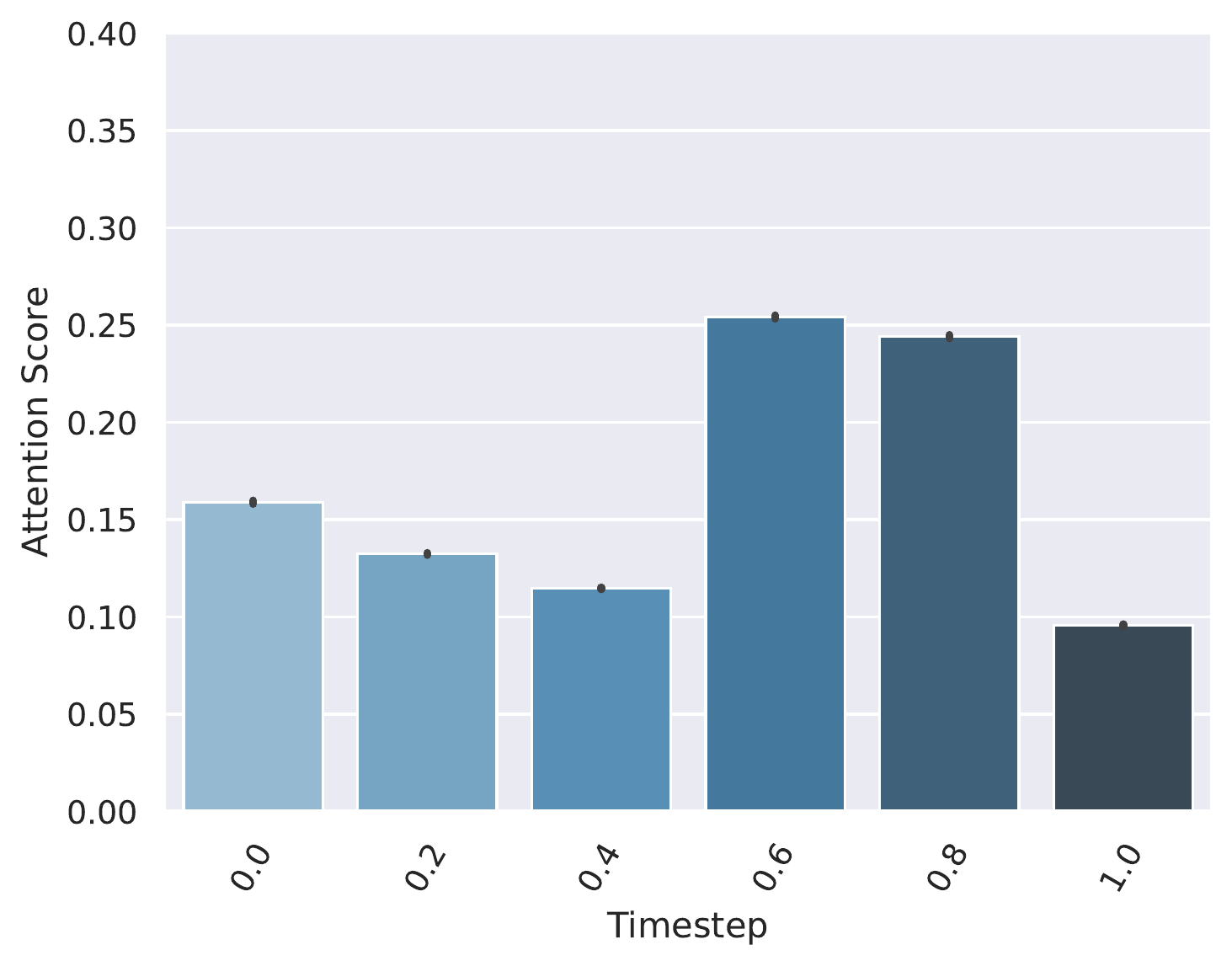}
\vspace{-6mm}
\subcaption*{\scriptsize Foreground Color}
\end{subfigure}
\begin{subfigure}[c]{0.24\textwidth}
\includegraphics[width=\textwidth]{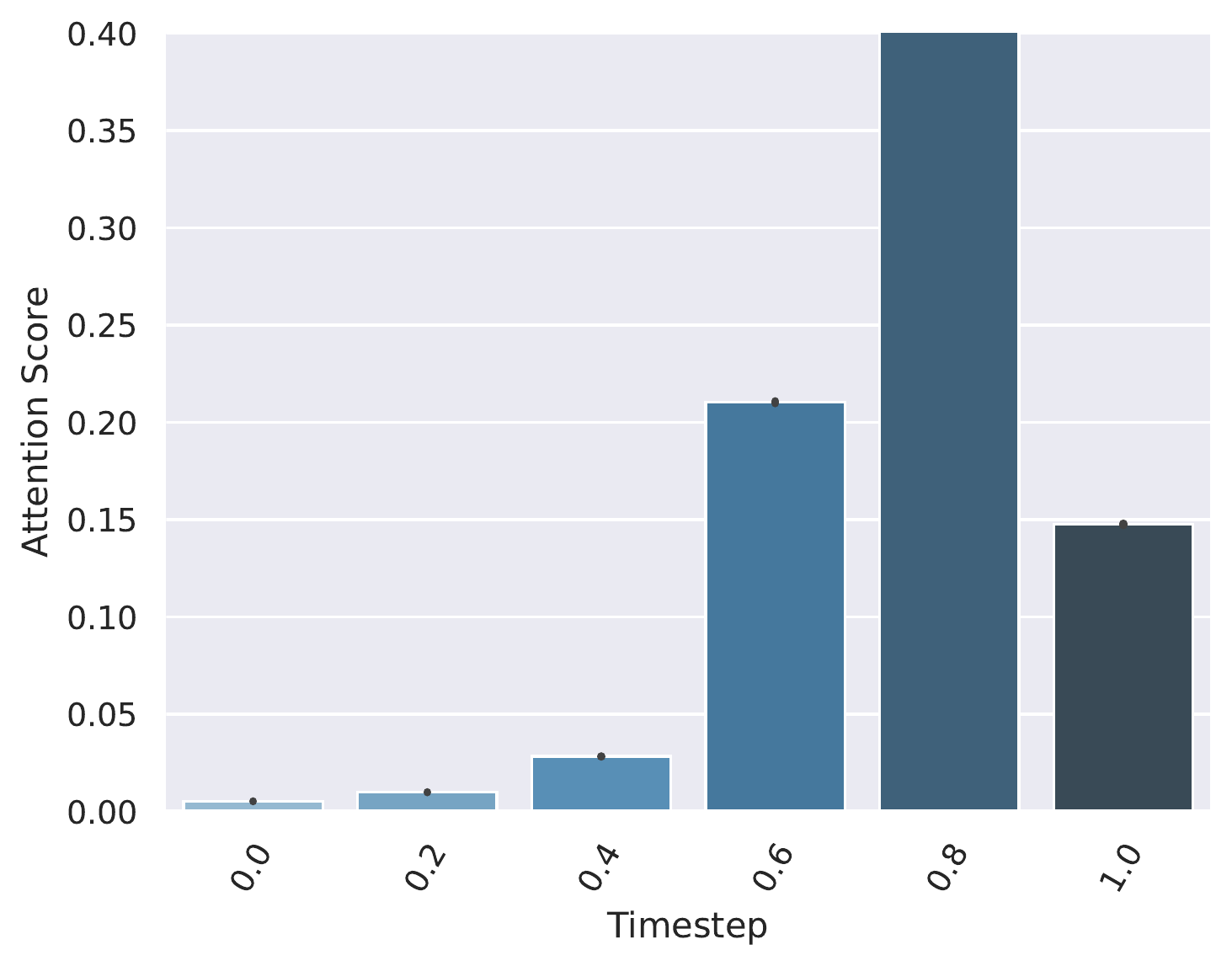}
\vspace{-6mm}
\subcaption*{\scriptsize Location}
\end{subfigure}
\begin{subfigure}[c]{0.24\textwidth}
\includegraphics[width=\textwidth]{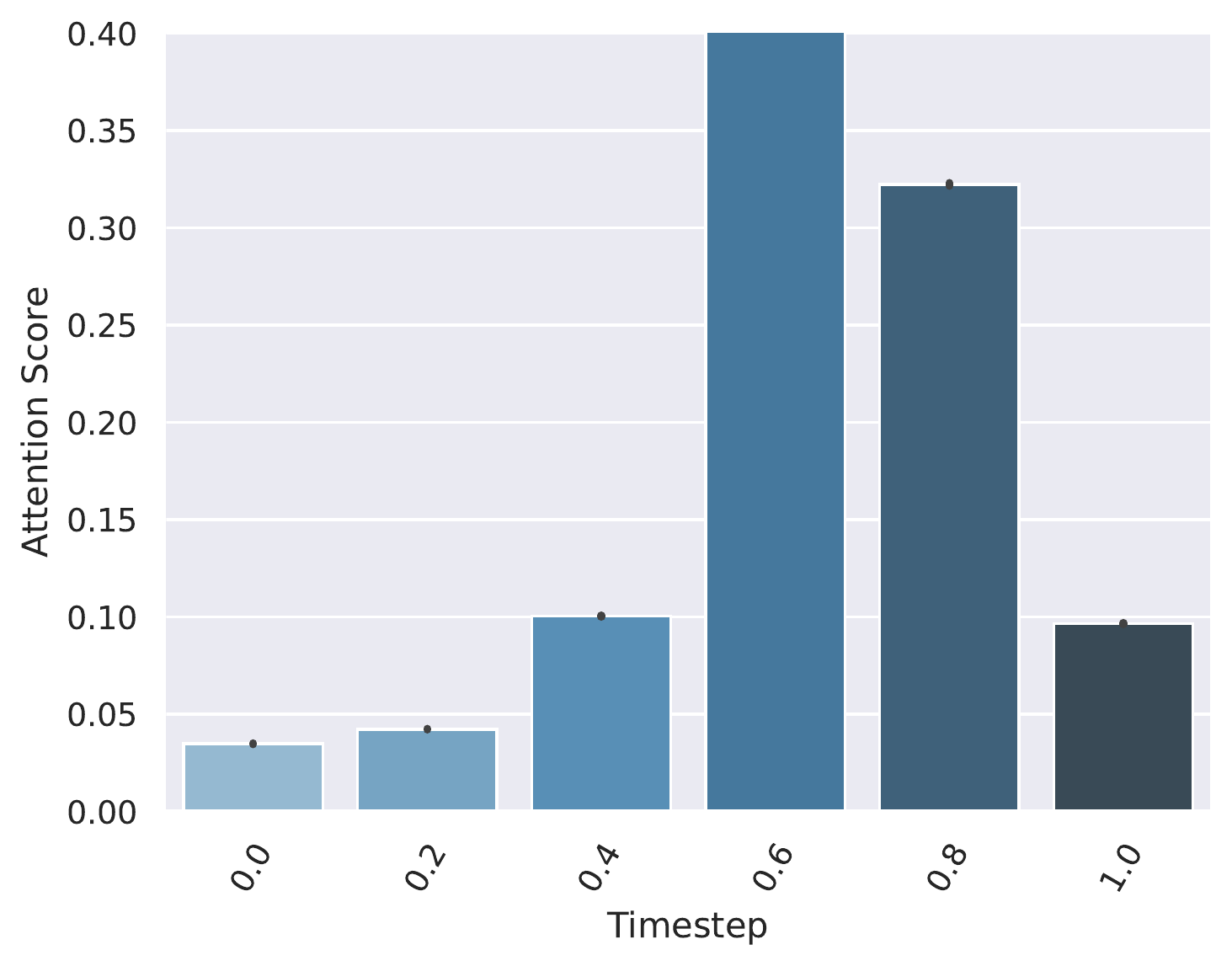}
\vspace{-6mm}
\subcaption*{\scriptsize Object Shape}
\end{subfigure}
\subcaption*{Granularity: 5}
\end{subfigure} \\
\begin{subfigure}[c]{\textwidth}
\begin{subfigure}[c]{0.24\textwidth}
\includegraphics[width=\textwidth]{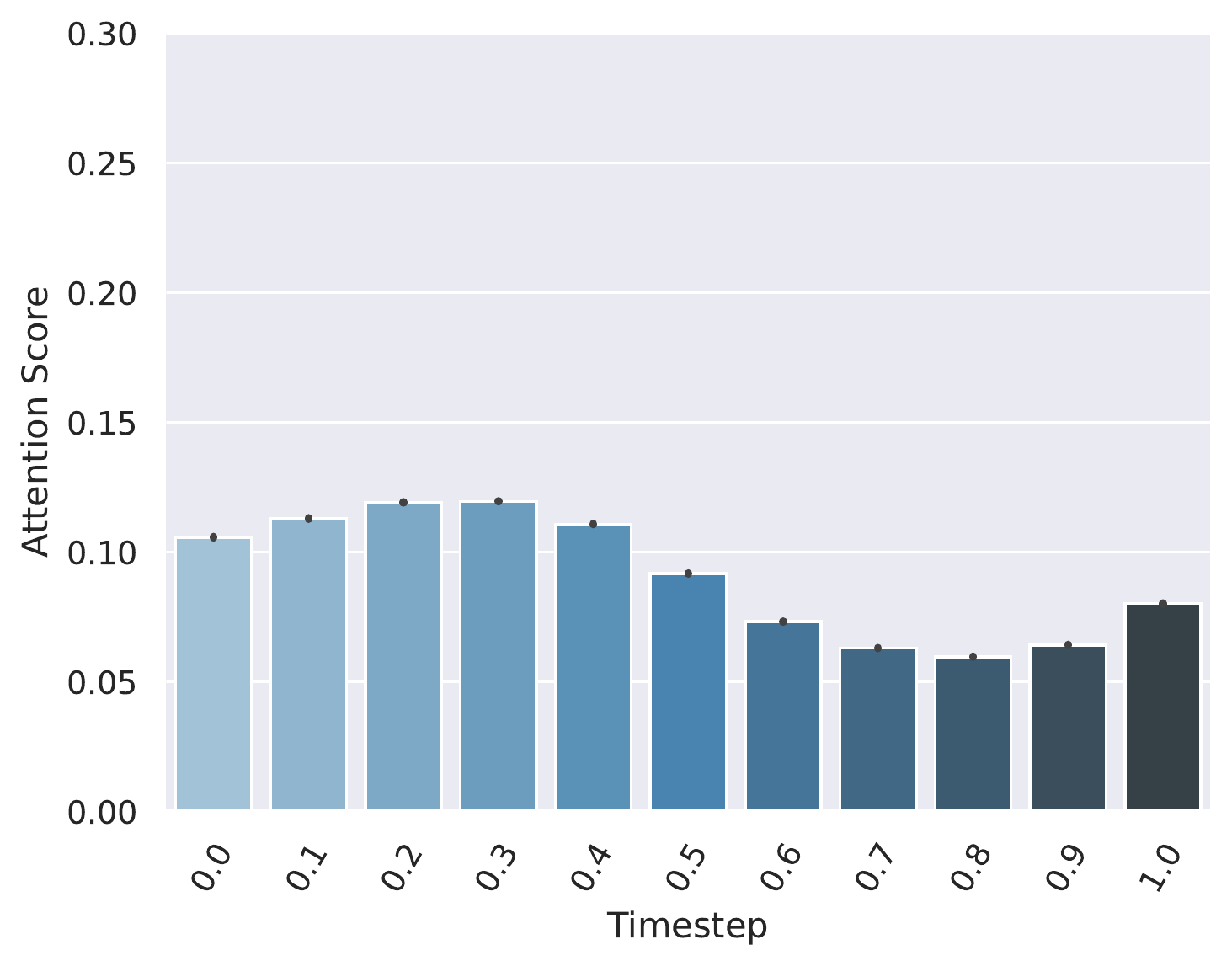}
\vspace{-6mm}
\subcaption*{\scriptsize Background Color}
\end{subfigure}
\begin{subfigure}[c]{0.24\textwidth}
\includegraphics[width=\textwidth]{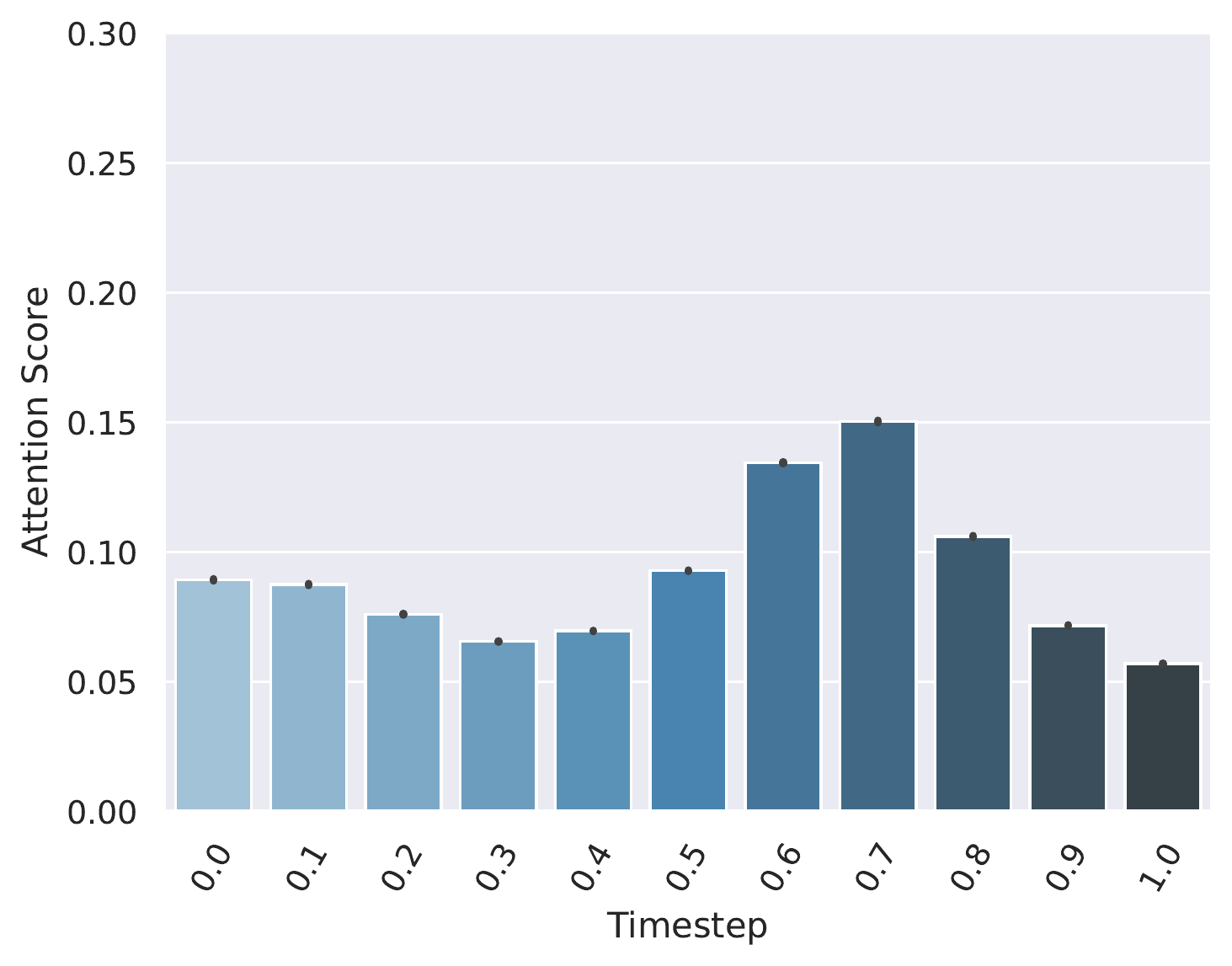}
\vspace{-6mm}
\subcaption*{\scriptsize Foreground Color}
\end{subfigure}
\begin{subfigure}[c]{0.24\textwidth}
\includegraphics[width=\textwidth]{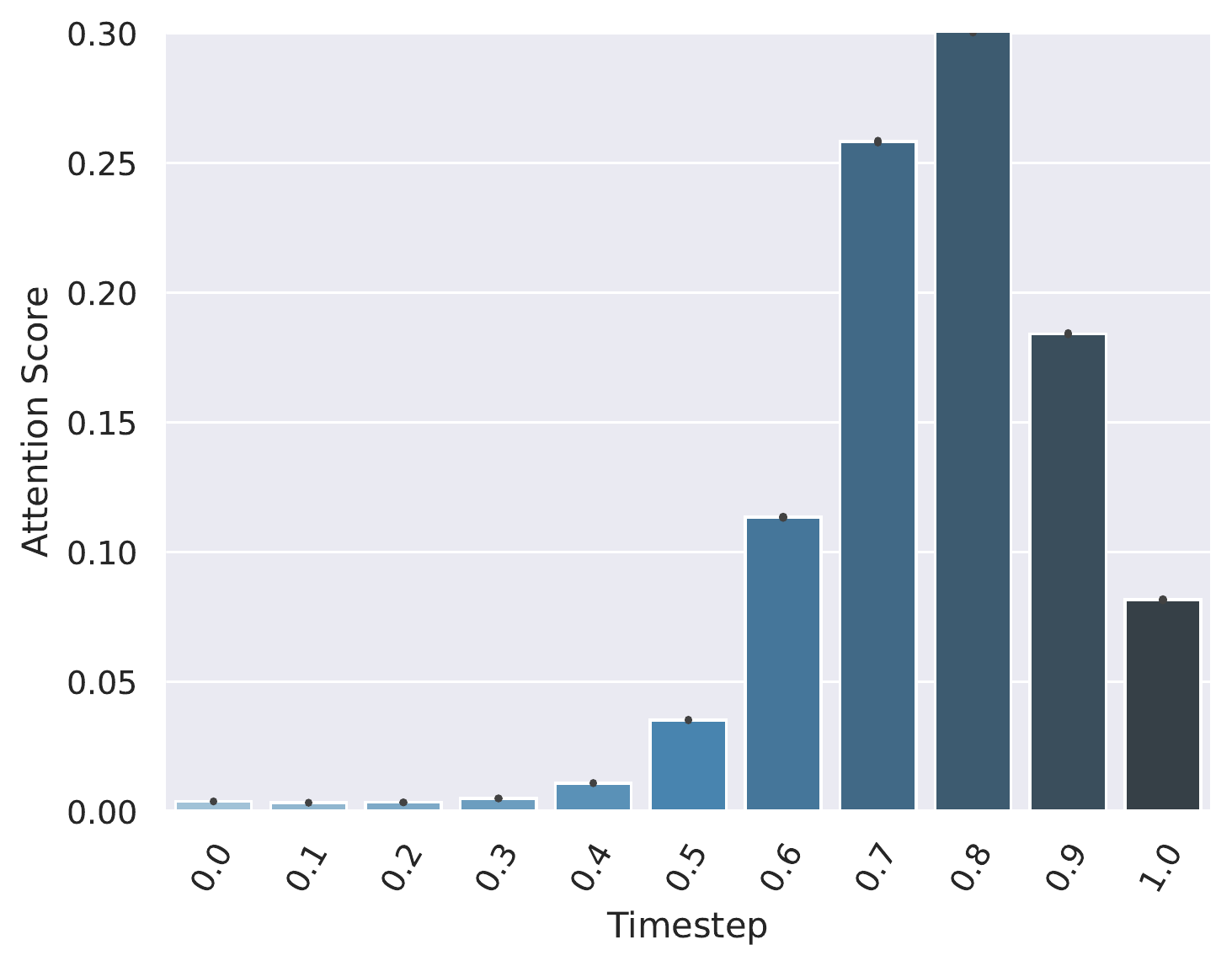}
\vspace{-6mm}
\subcaption*{\scriptsize Location}
\end{subfigure}
\begin{subfigure}[c]{0.24\textwidth}
\includegraphics[width=\textwidth]{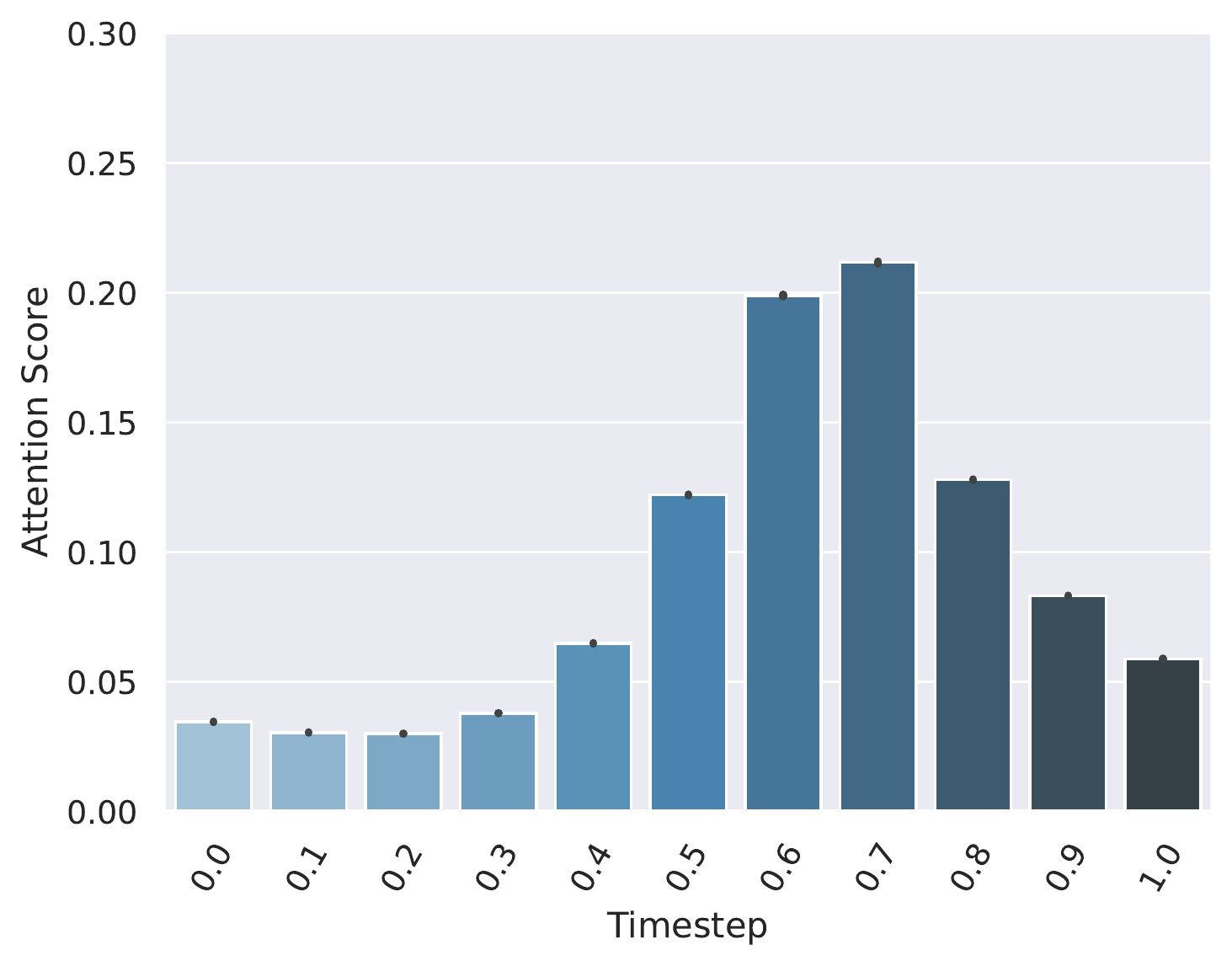}
\vspace{-6mm}
\subcaption*{\scriptsize Object Shape}
\end{subfigure}
\subcaption*{Granularity: 10}
\end{subfigure} \\
\begin{subfigure}[c]{\textwidth}
\begin{subfigure}[c]{0.24\textwidth}
\includegraphics[width=\textwidth]{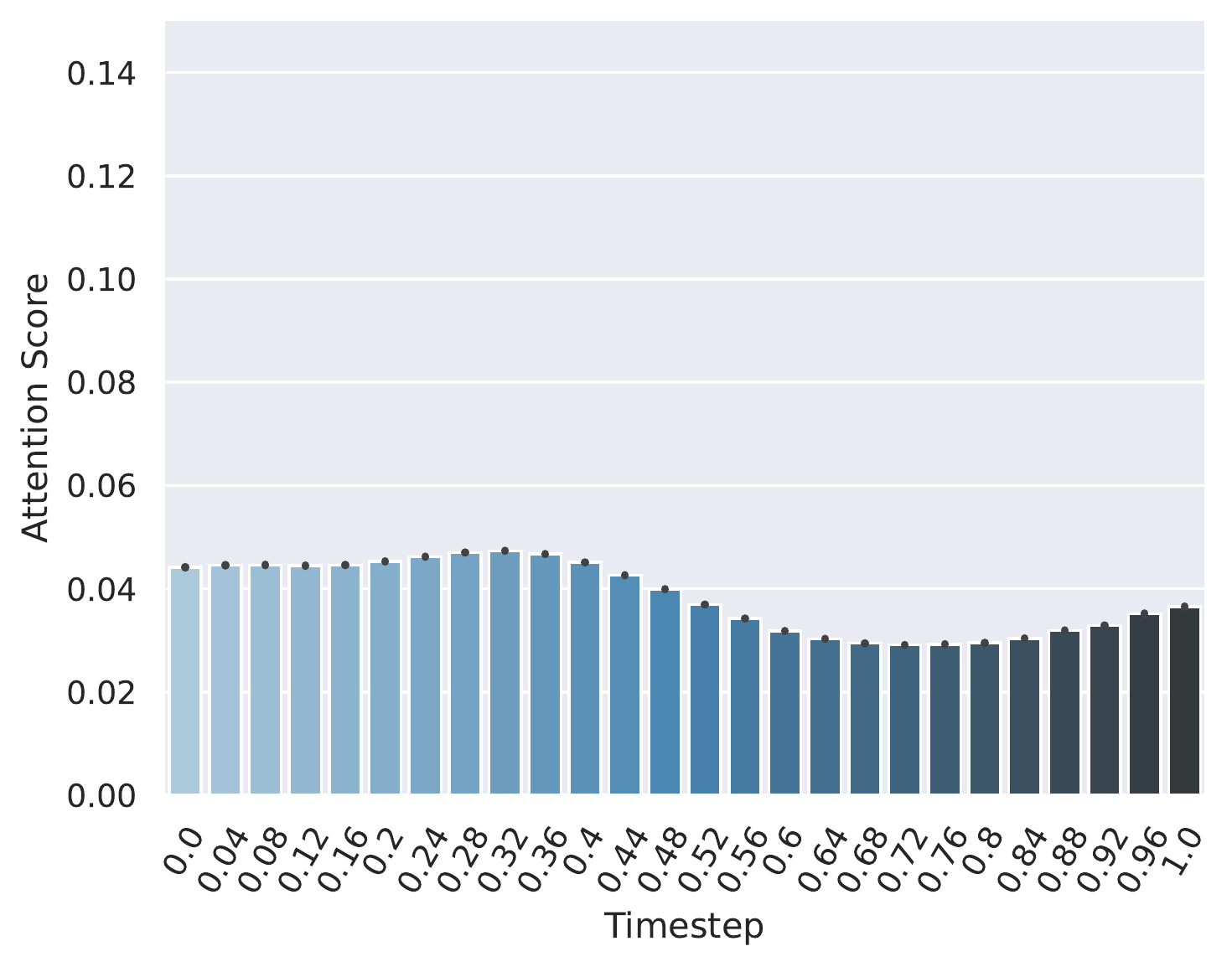}
\vspace{-6mm}
\subcaption*{\scriptsize Background Color}
\end{subfigure}
\begin{subfigure}[c]{0.24\textwidth}
\includegraphics[width=\textwidth]{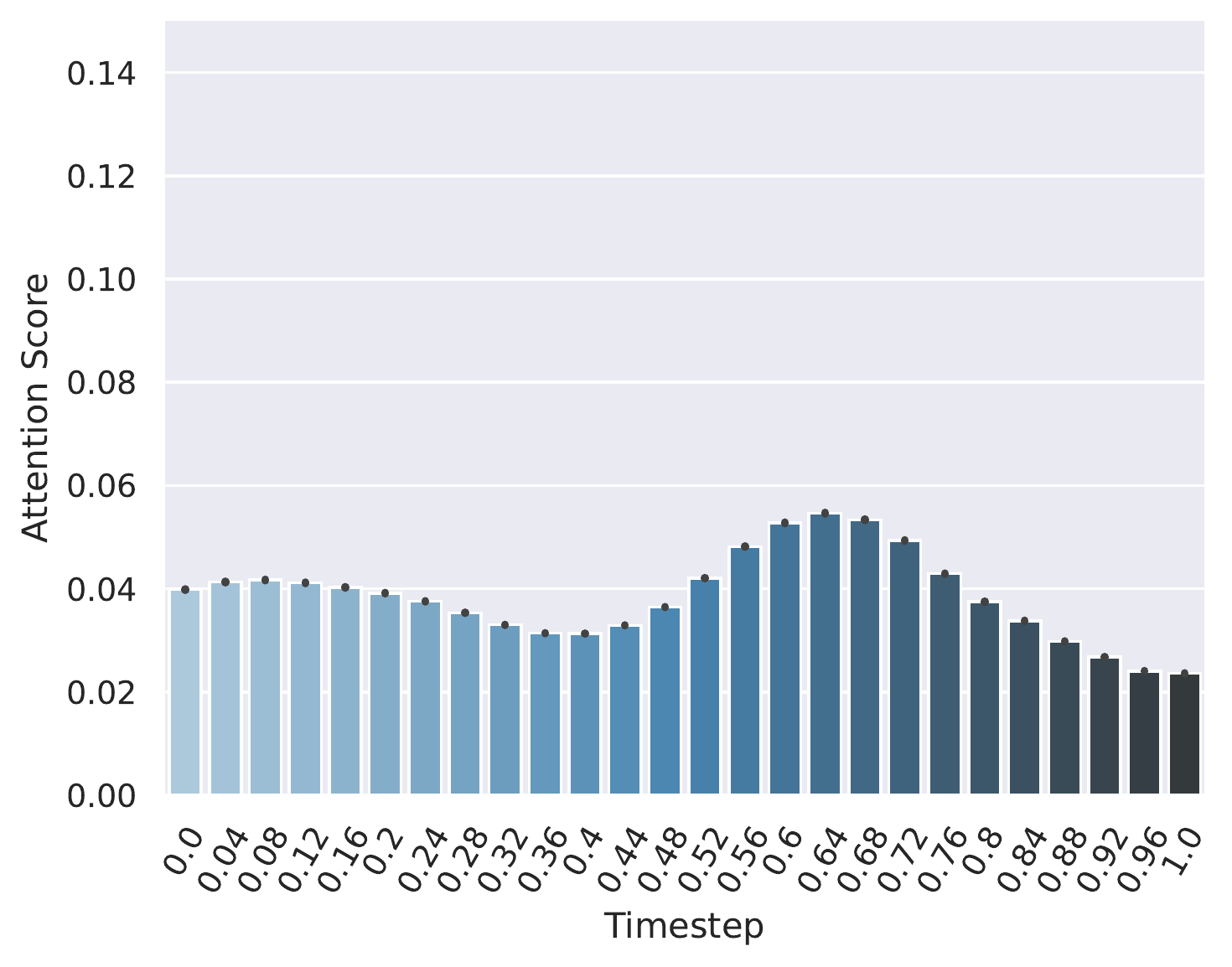}
\vspace{-6mm}
\subcaption*{\scriptsize Foreground Color}
\end{subfigure}
\begin{subfigure}[c]{0.24\textwidth}
\includegraphics[width=\textwidth]{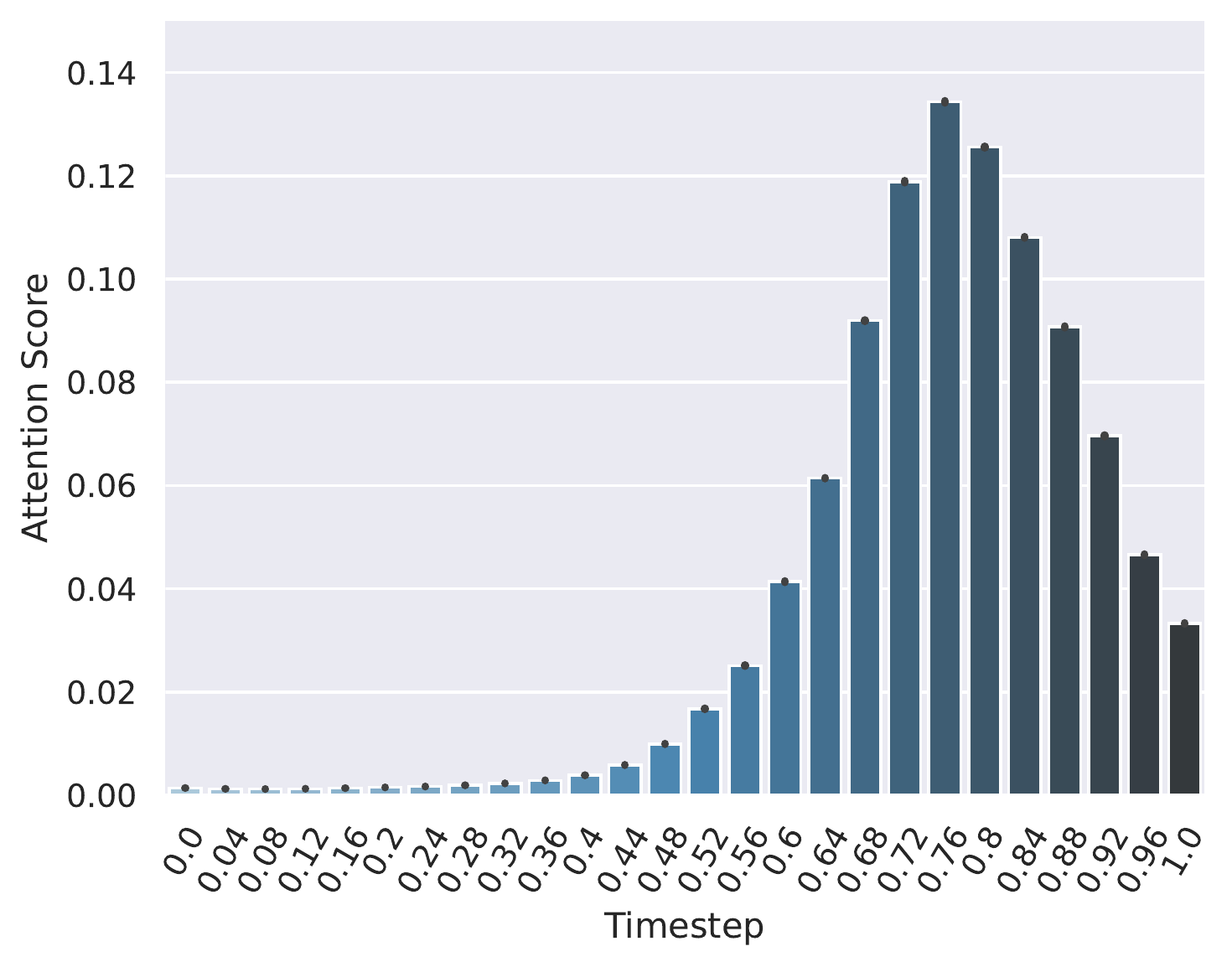}
\vspace{-6mm}
\subcaption*{\scriptsize Location}
\end{subfigure}
\begin{subfigure}[c]{0.24\textwidth}
\includegraphics[width=\textwidth]{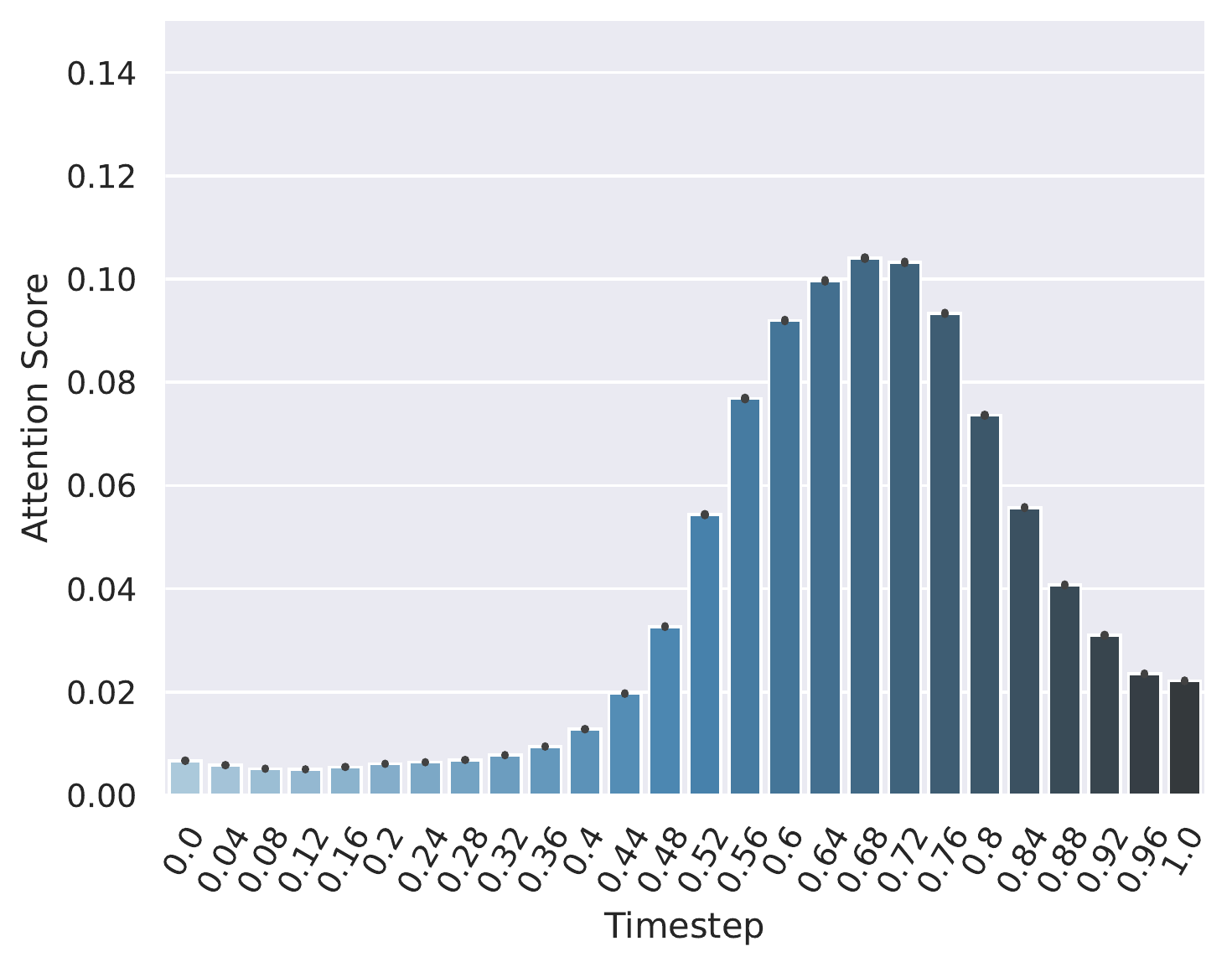}
\vspace{-6mm}
\subcaption*{\scriptsize Object Shape}
\end{subfigure}
\subcaption*{Granularity: 25}
\end{subfigure} \\
\caption{Attention score profiles for the synthetic dataset on the different features, using different granularities, with the dimensionality of the latent space as 16 and the DRL encoder.}
\label{fig:syn_DRL_16}
\end{figure}

\begin{figure}
\begin{subfigure}[c]{\textwidth}
\begin{subfigure}[c]{0.24\textwidth}
\includegraphics[width=\textwidth]{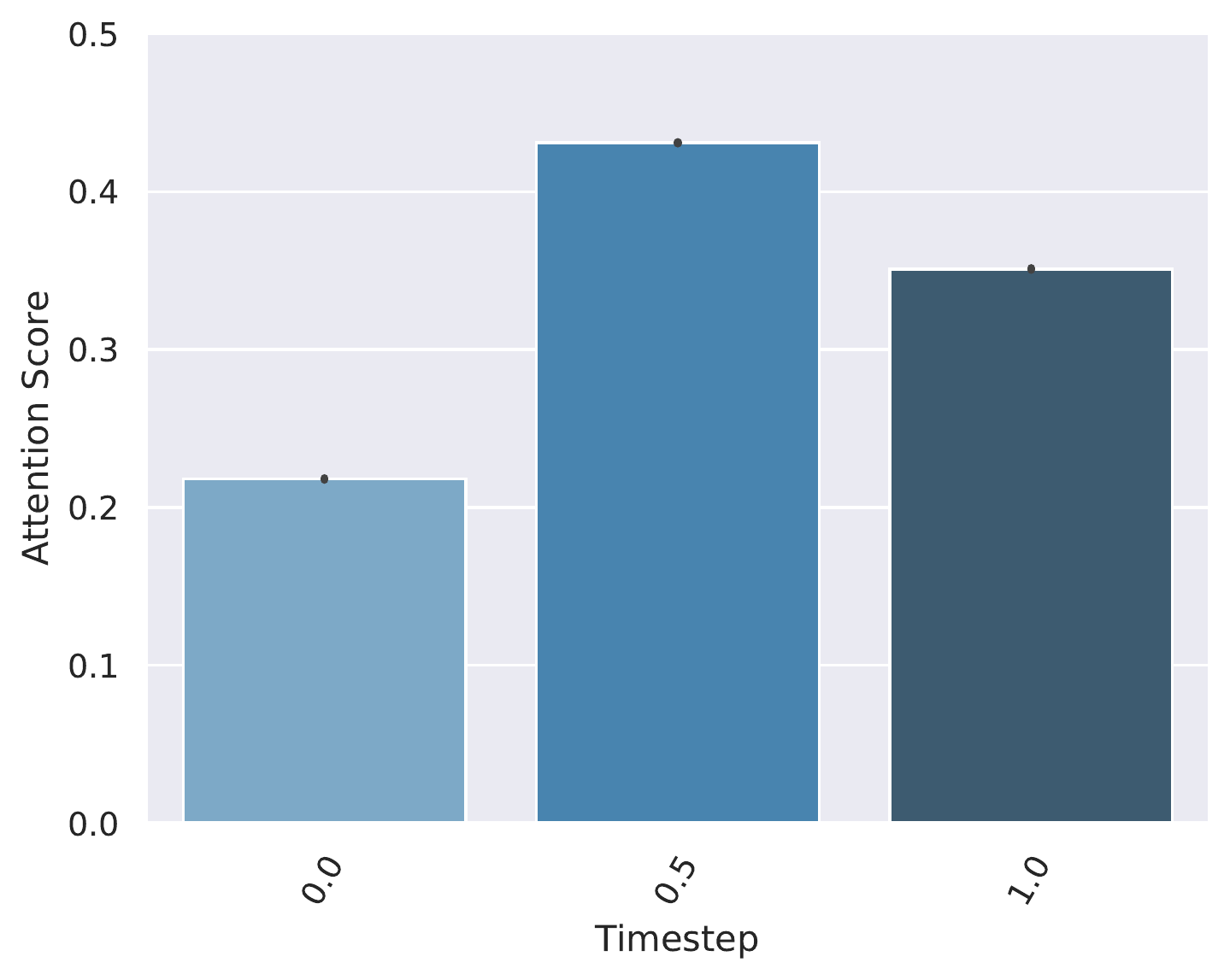}
\vspace{-6mm}
\subcaption*{\scriptsize Background Color}
\end{subfigure}
\begin{subfigure}[c]{0.24\textwidth}
\includegraphics[width=\textwidth]{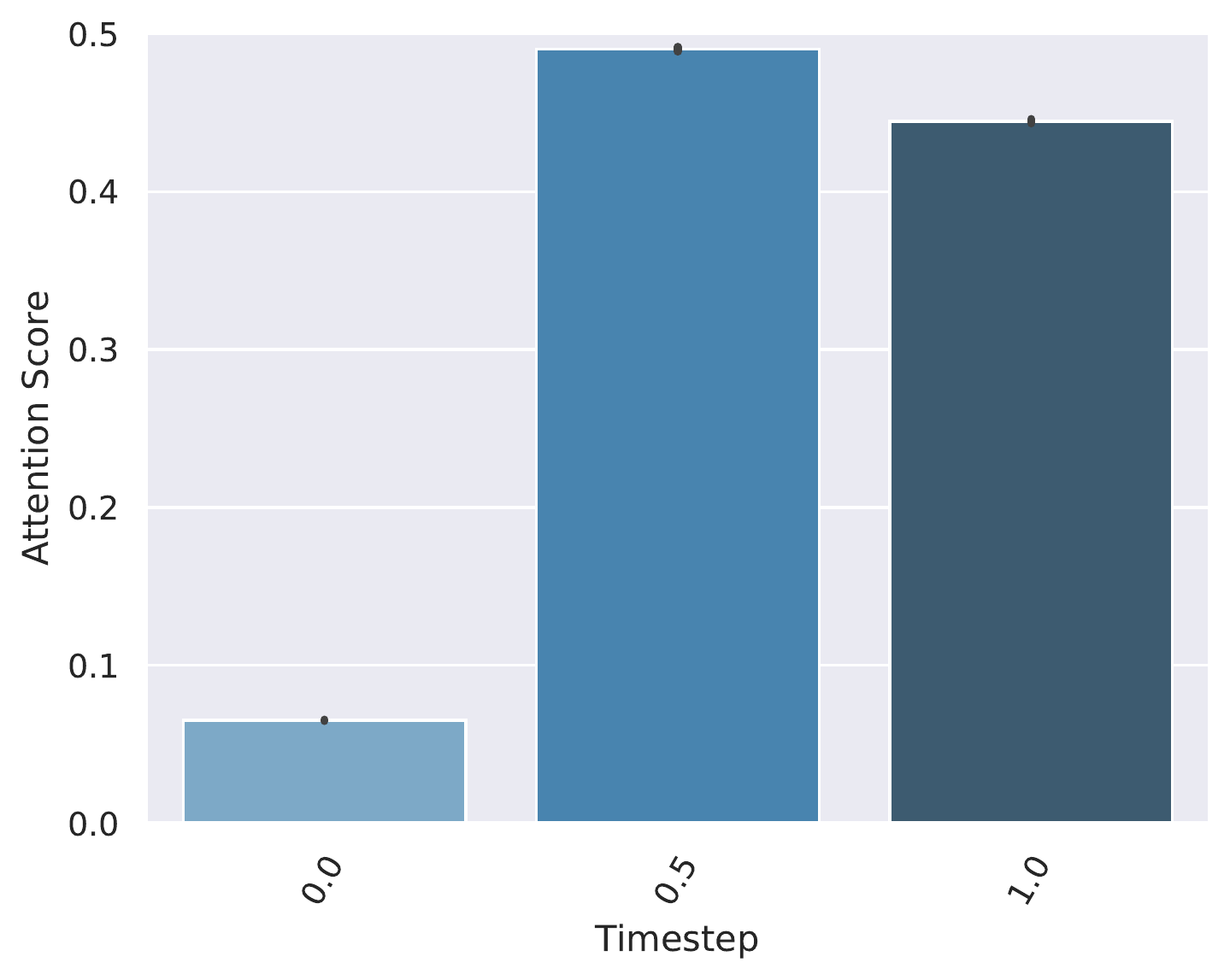}
\vspace{-6mm}
\subcaption*{\scriptsize Foreground Color}
\end{subfigure}
\begin{subfigure}[c]{0.24\textwidth}
\includegraphics[width=\textwidth]{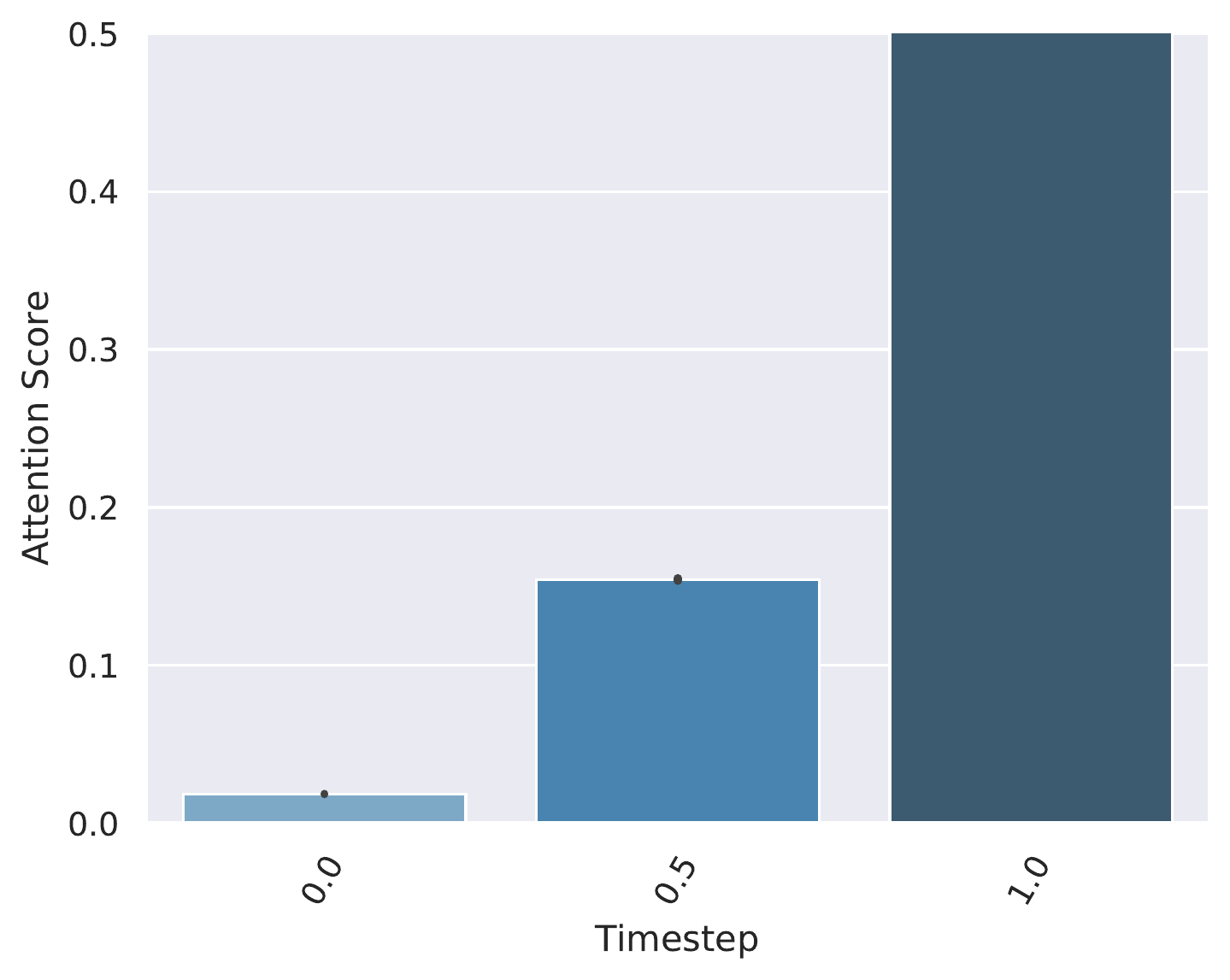}
\vspace{-6mm}
\subcaption*{\scriptsize Location}
\end{subfigure}
\begin{subfigure}[c]{0.24\textwidth}
\includegraphics[width=\textwidth]{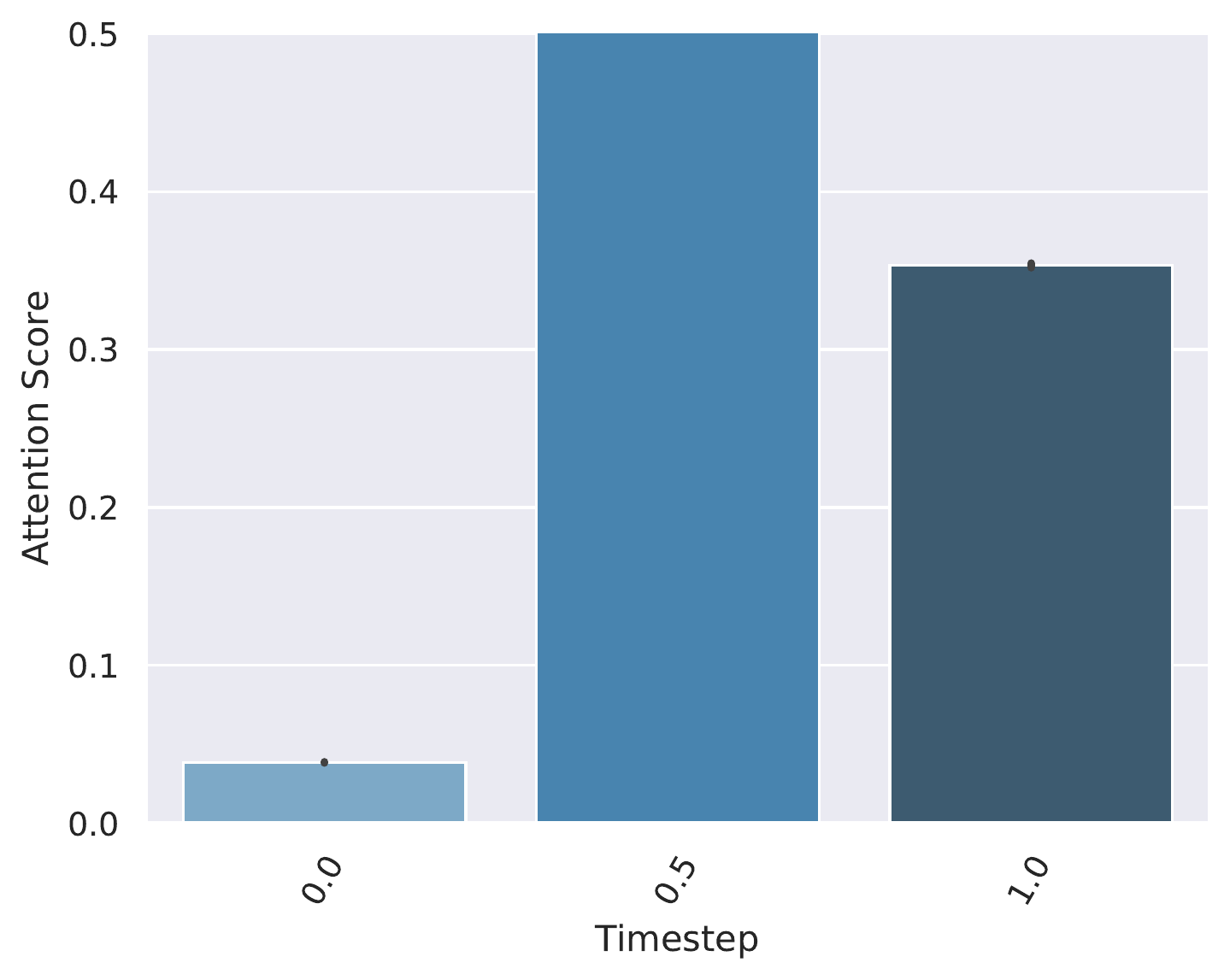}
\vspace{-6mm}
\subcaption*{\scriptsize Object Shape}
\end{subfigure}
\subcaption*{Granularity: 2}
\end{subfigure} \\
\begin{subfigure}[c]{\textwidth}
\begin{subfigure}[c]{0.24\textwidth}
\includegraphics[width=\textwidth]{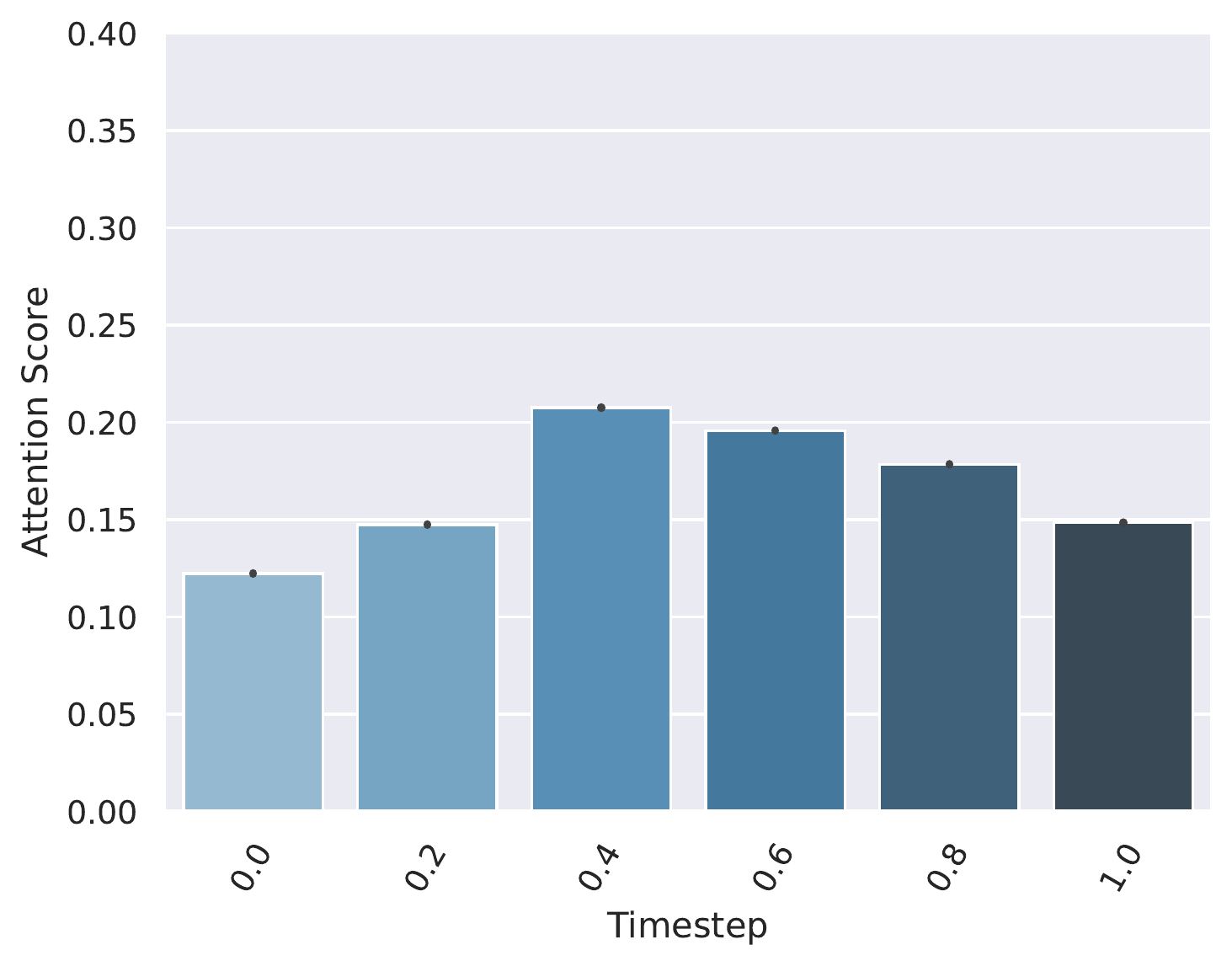}
\vspace{-6mm}
\subcaption*{\scriptsize Background Color}
\end{subfigure}
\begin{subfigure}[c]{0.24\textwidth}
\includegraphics[width=\textwidth]{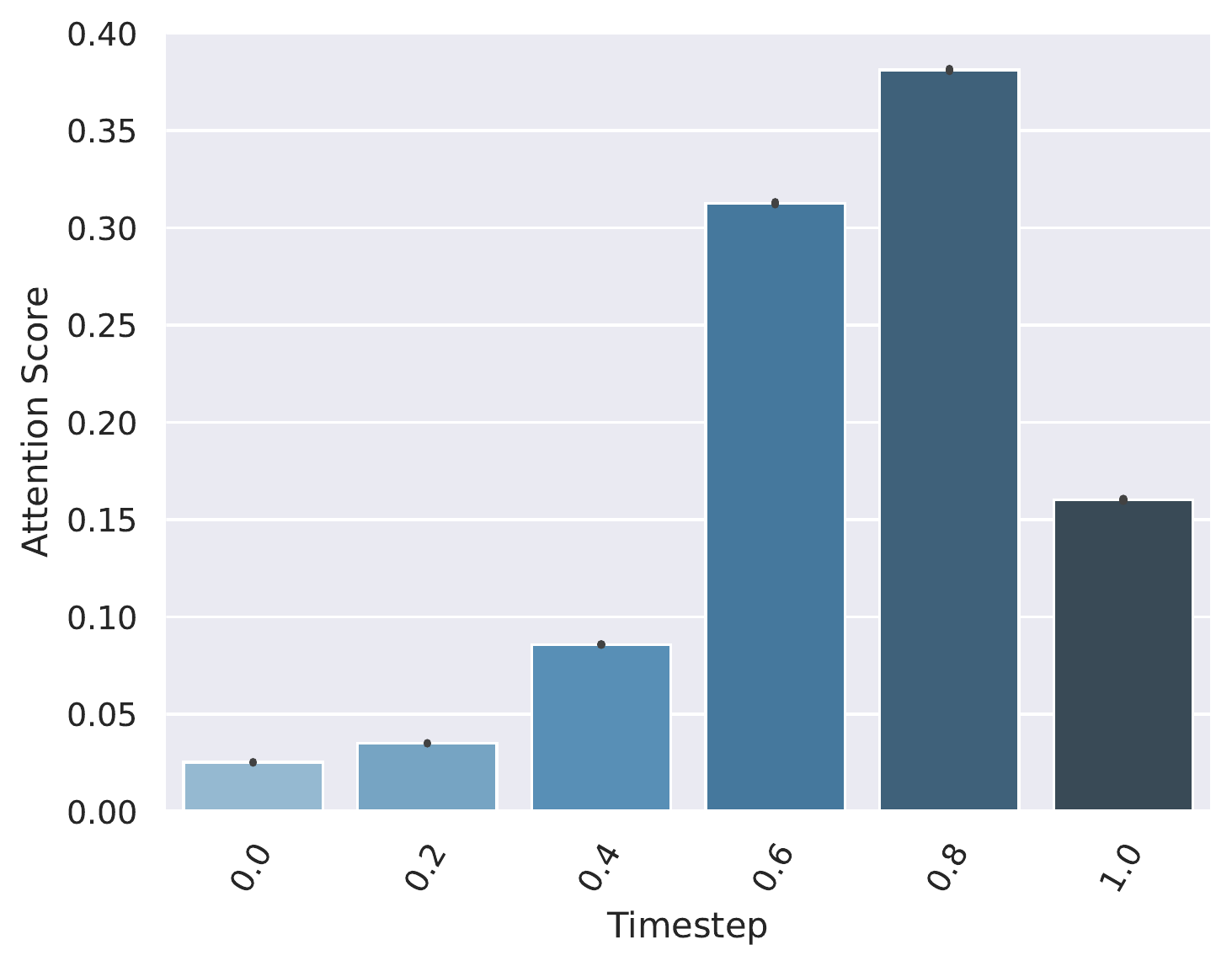}
\vspace{-6mm}
\subcaption*{\scriptsize Foreground Color}
\end{subfigure}
\begin{subfigure}[c]{0.24\textwidth}
\includegraphics[width=\textwidth]{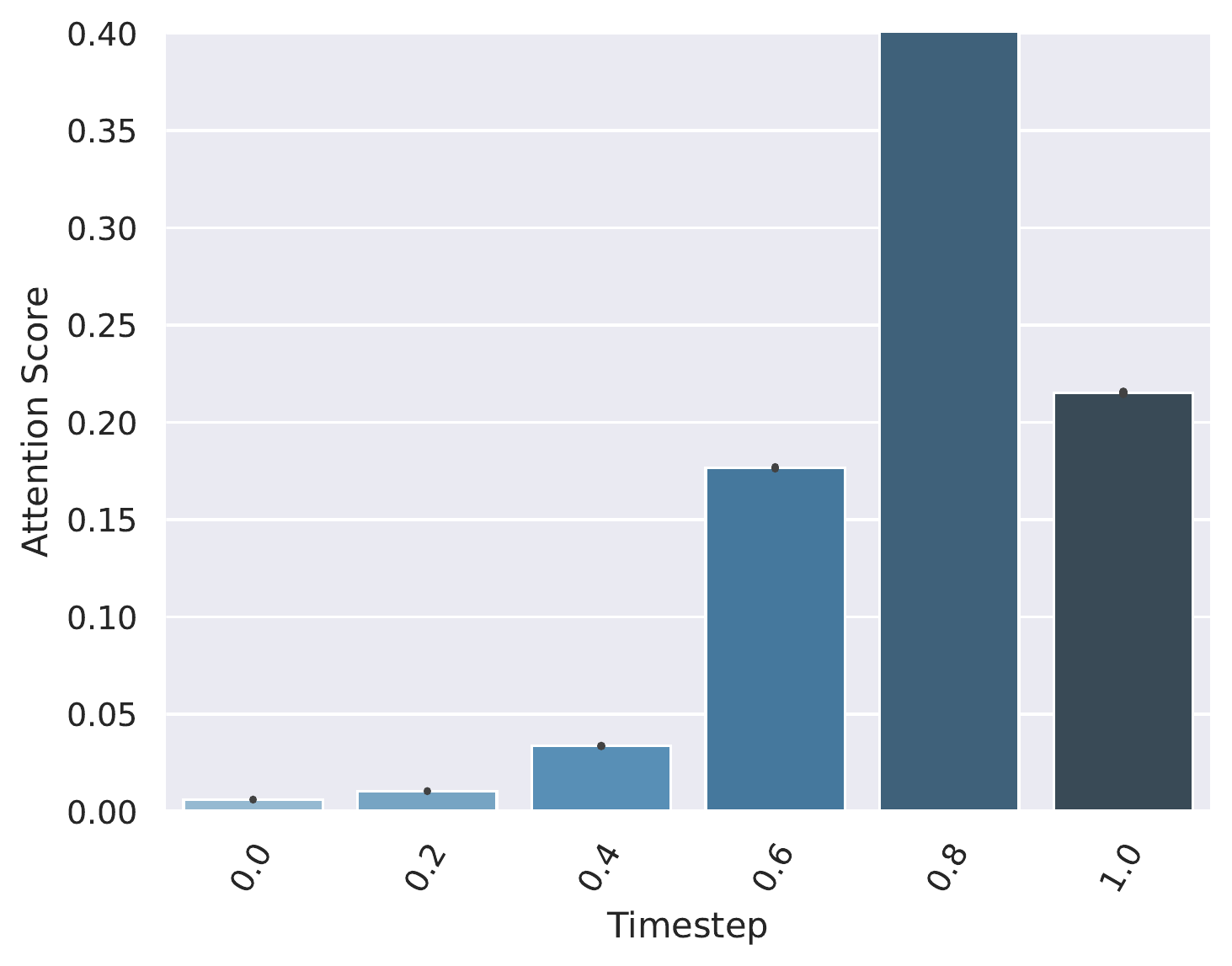}
\vspace{-6mm}
\subcaption*{\scriptsize Location}
\end{subfigure}
\begin{subfigure}[c]{0.24\textwidth}
\includegraphics[width=\textwidth]{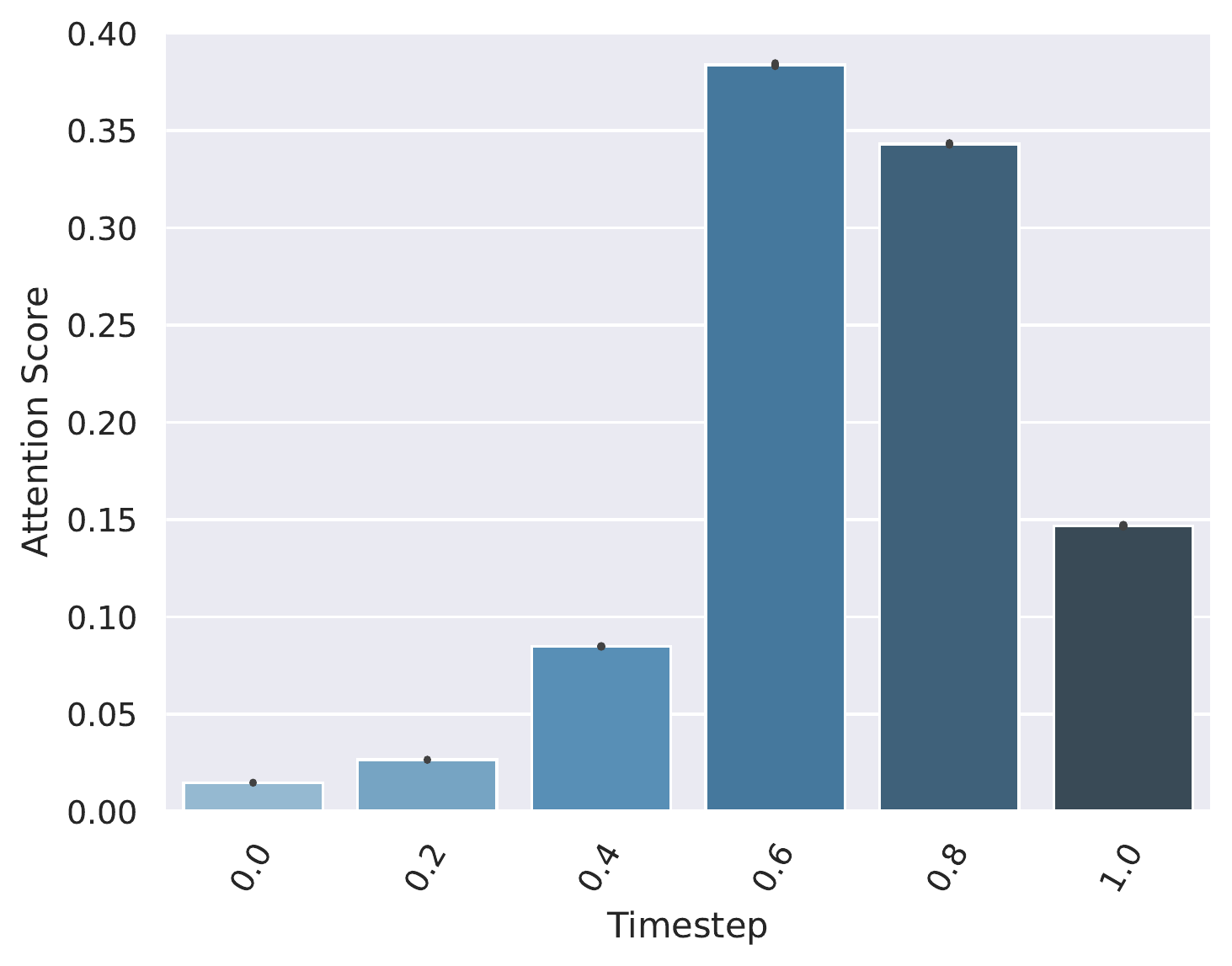}
\vspace{-6mm}
\subcaption*{\scriptsize Object Shape}
\end{subfigure}
\subcaption*{Granularity: 5}
\end{subfigure} \\
\begin{subfigure}[c]{\textwidth}
\begin{subfigure}[c]{0.24\textwidth}
\includegraphics[width=\textwidth]{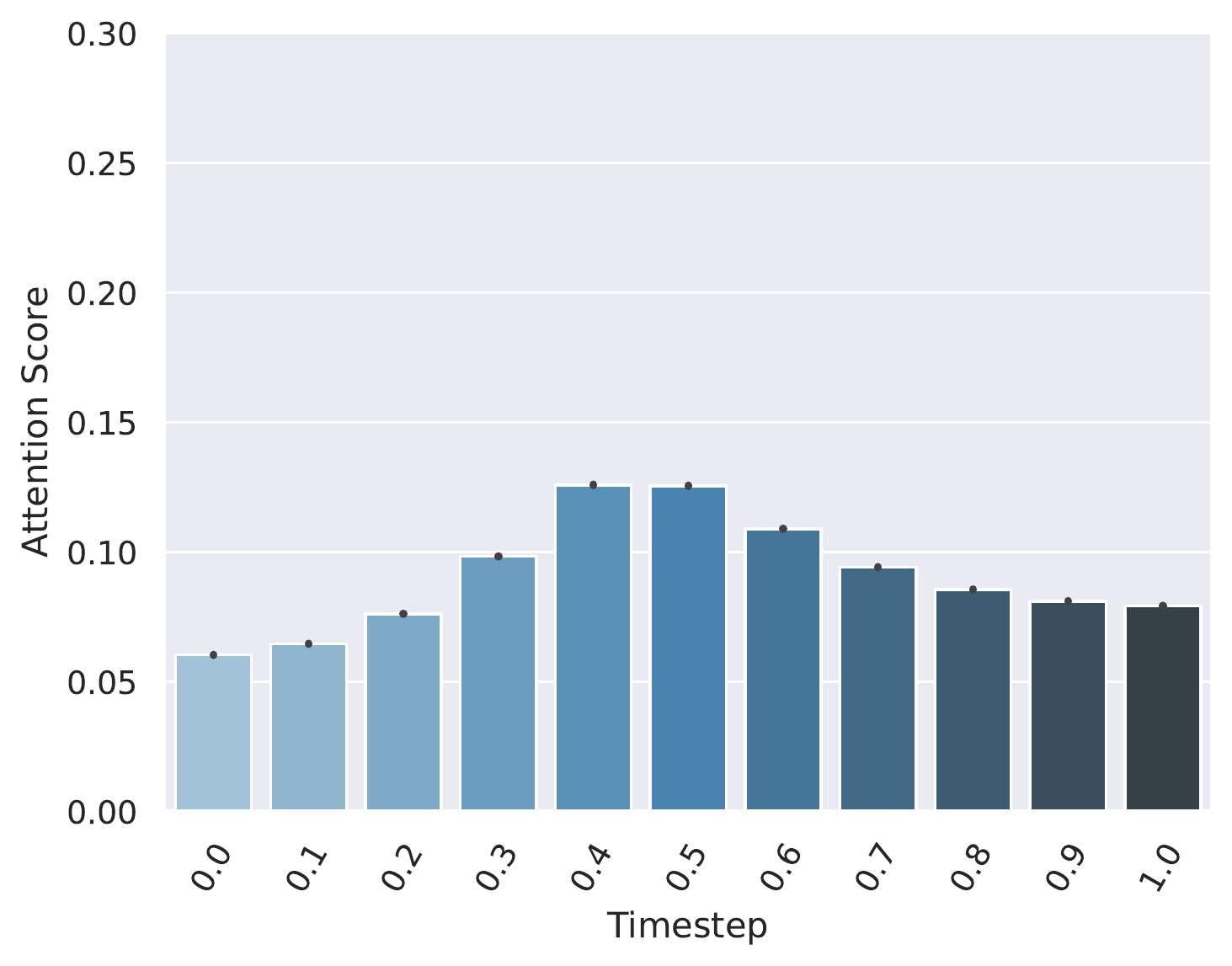}
\vspace{-6mm}
\subcaption*{\scriptsize Background Color}
\end{subfigure}
\begin{subfigure}[c]{0.24\textwidth}
\includegraphics[width=\textwidth]{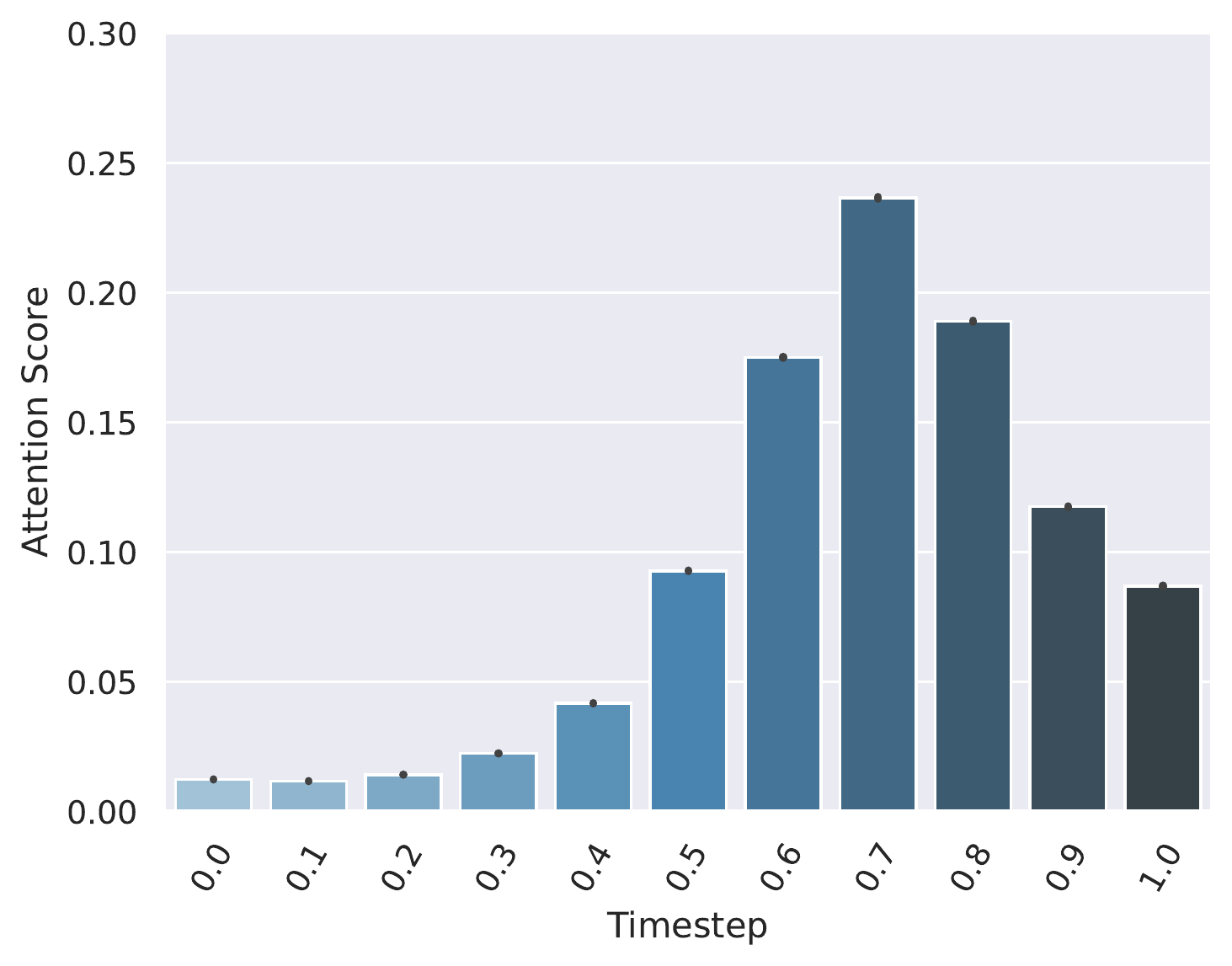}
\vspace{-6mm}
\subcaption*{\scriptsize Foreground Color}
\end{subfigure}
\begin{subfigure}[c]{0.24\textwidth}
\includegraphics[width=\textwidth]{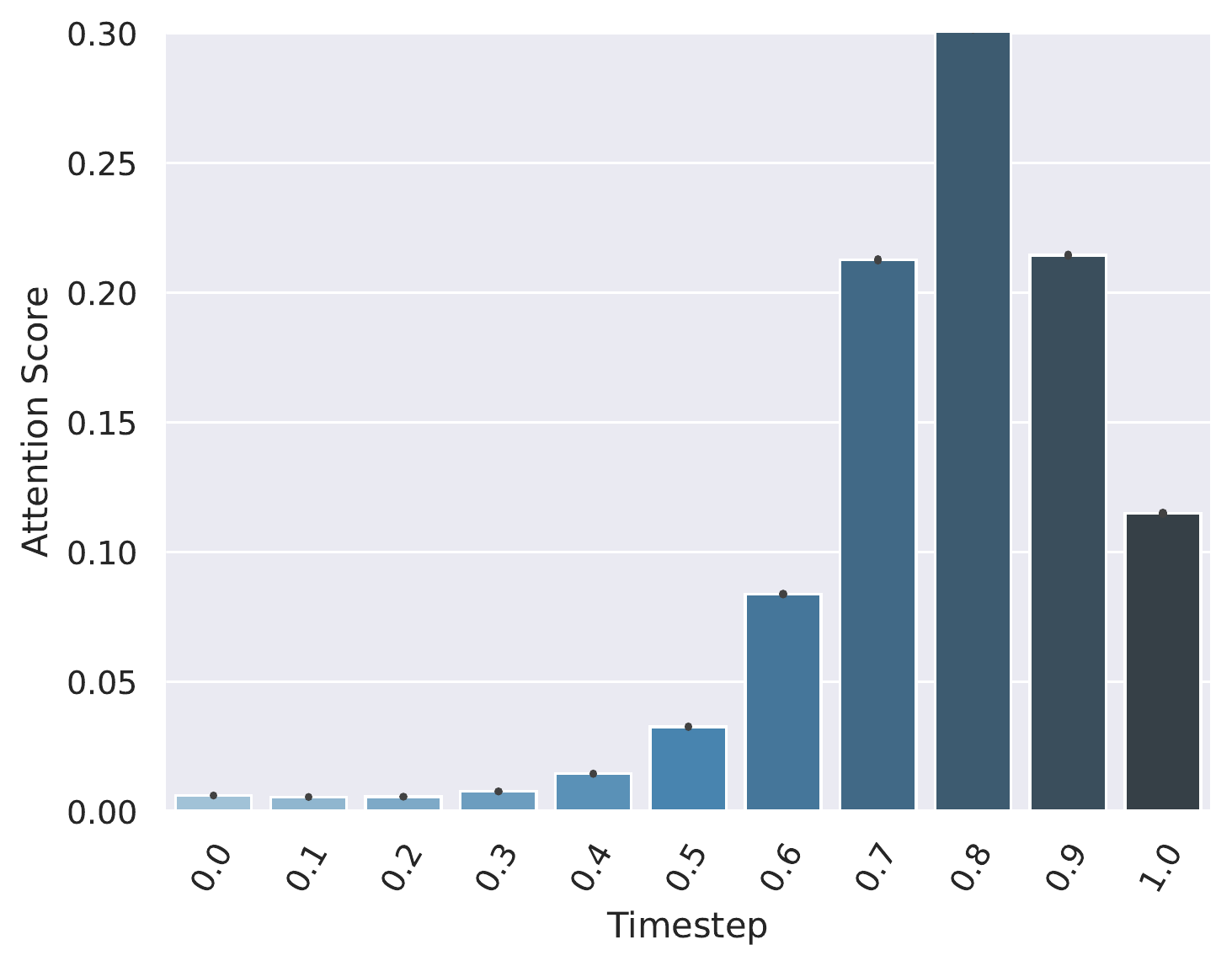}
\vspace{-6mm}
\subcaption*{\scriptsize Location}
\end{subfigure}
\begin{subfigure}[c]{0.24\textwidth}
\includegraphics[width=\textwidth]{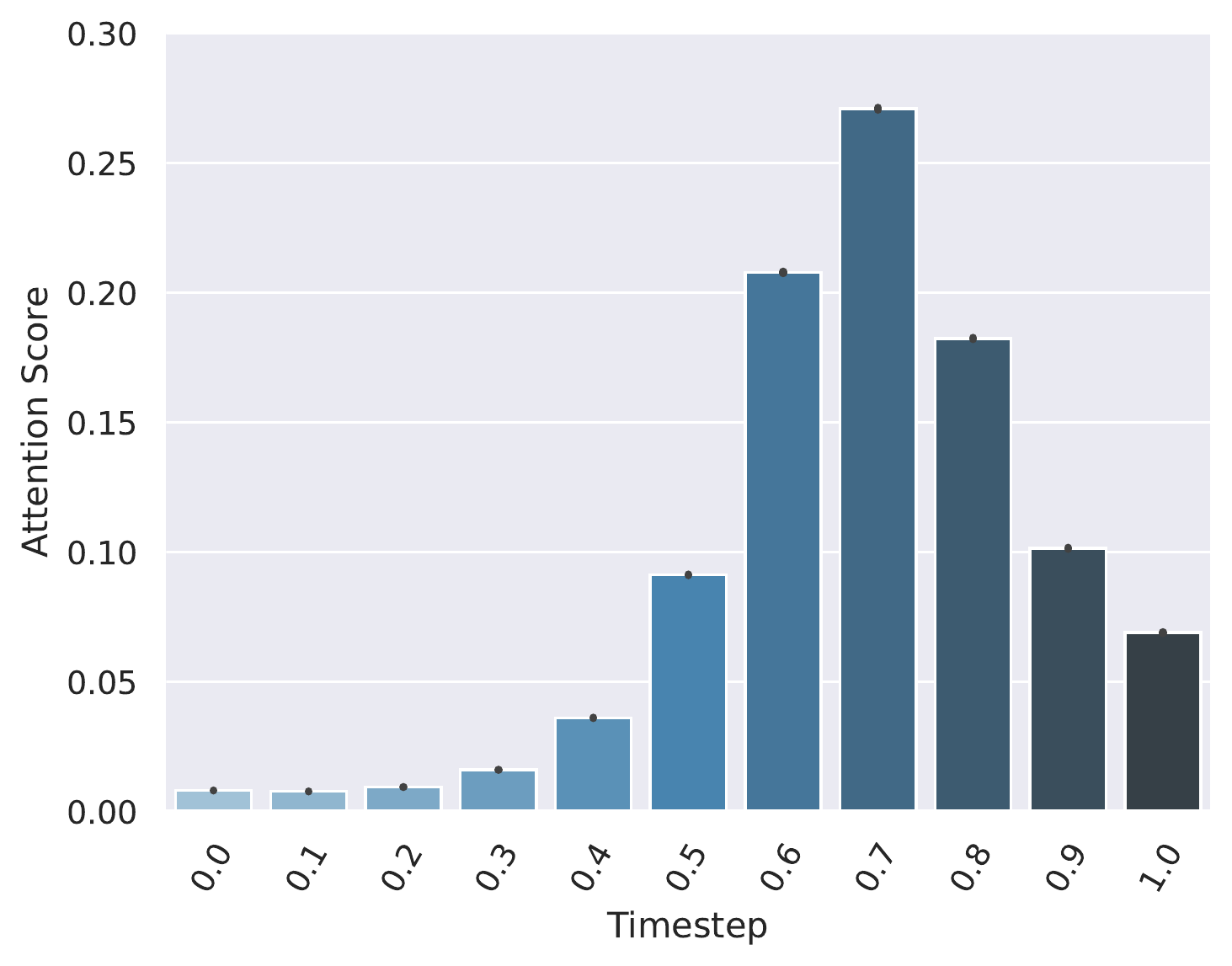}
\vspace{-6mm}
\subcaption*{\scriptsize Object Shape}
\end{subfigure}
\subcaption*{Granularity: 10}
\end{subfigure} \\
\begin{subfigure}[c]{\textwidth}
\begin{subfigure}[c]{0.24\textwidth}
\includegraphics[width=\textwidth]{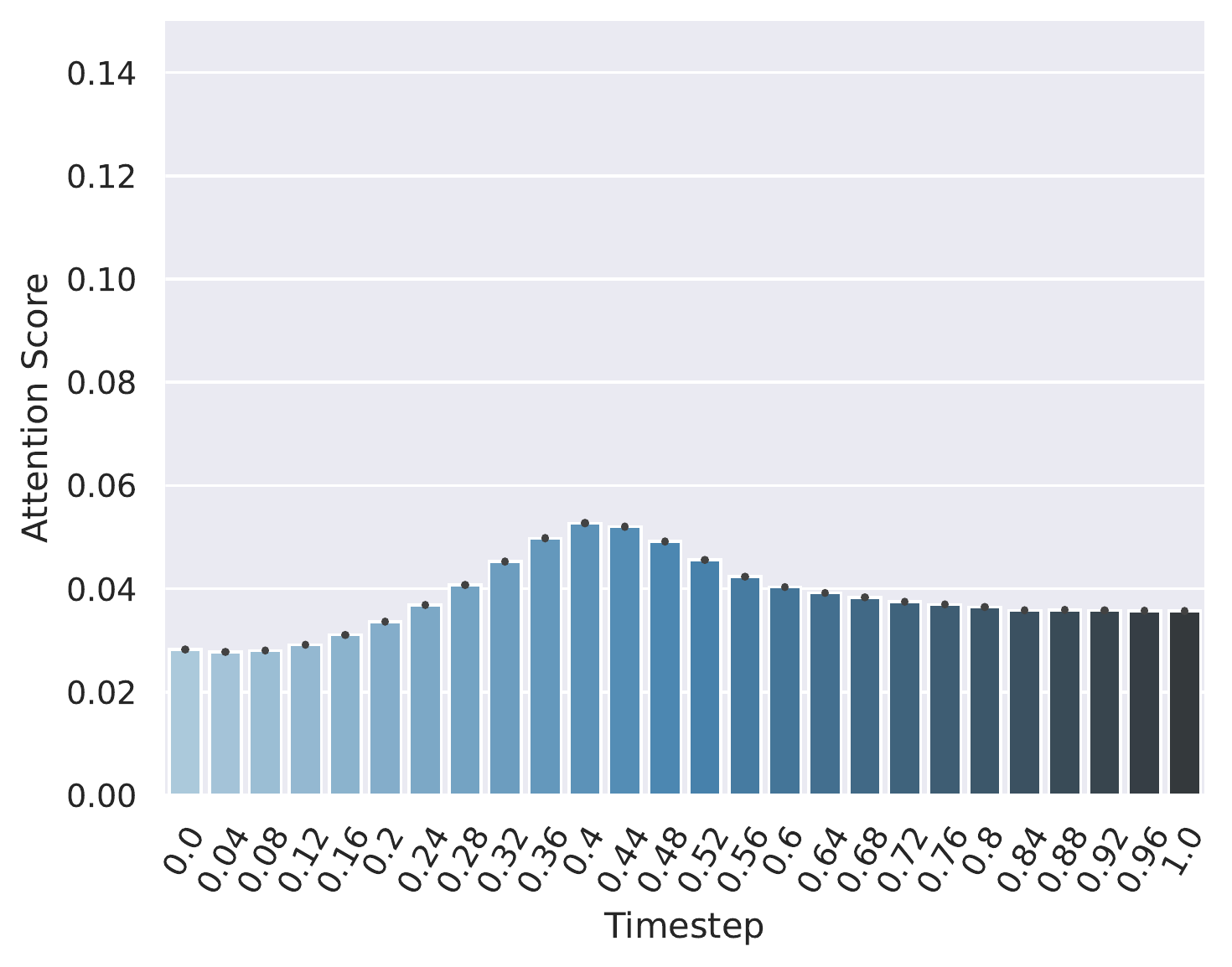}
\vspace{-6mm}
\subcaption*{\scriptsize Background Color}
\end{subfigure}
\begin{subfigure}[c]{0.24\textwidth}
\includegraphics[width=\textwidth]{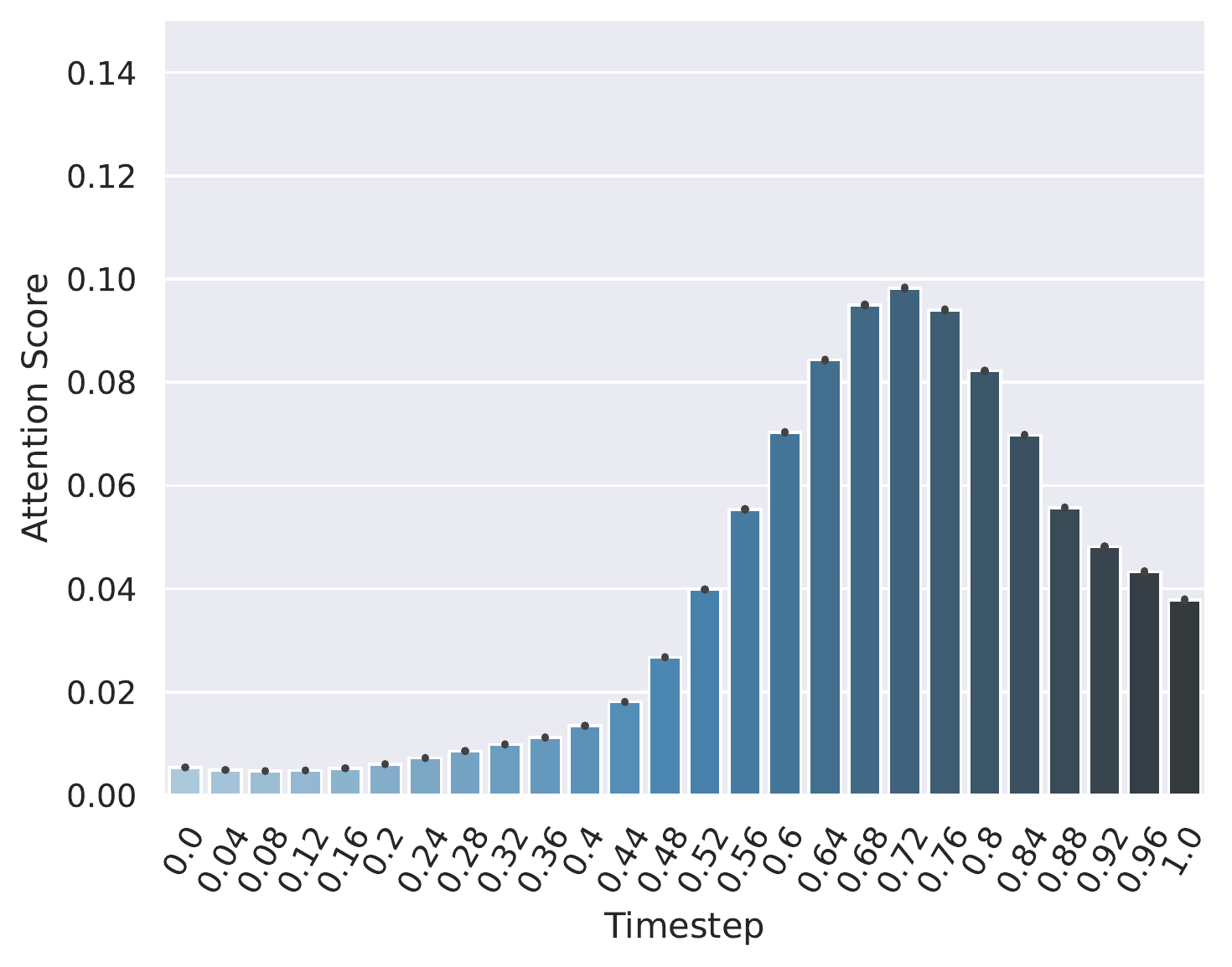}
\vspace{-6mm}
\subcaption*{\scriptsize Foreground Color}
\end{subfigure}
\begin{subfigure}[c]{0.24\textwidth}
\includegraphics[width=\textwidth]{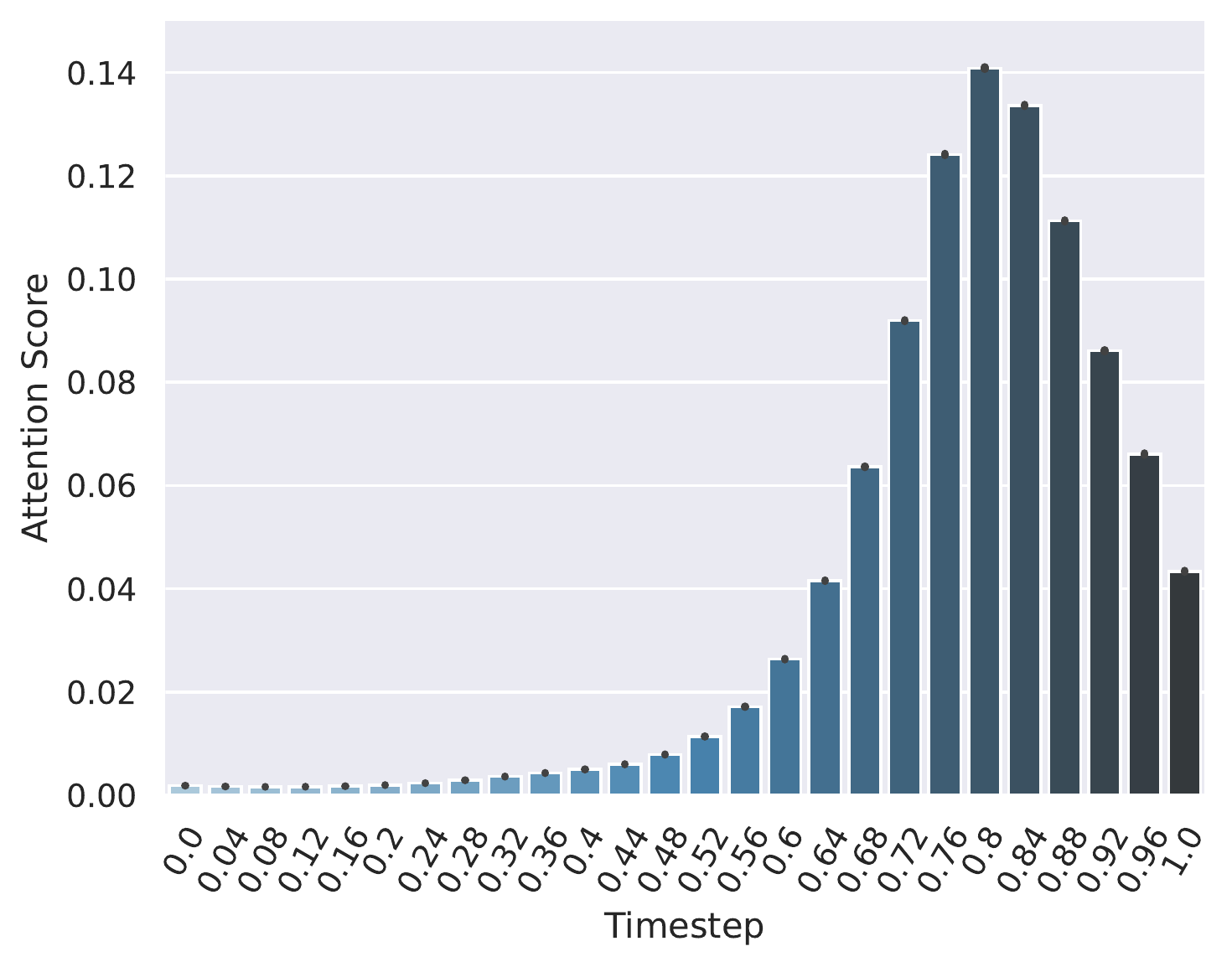}
\vspace{-6mm}
\subcaption*{\scriptsize Location}
\end{subfigure}
\begin{subfigure}[c]{0.24\textwidth}
\includegraphics[width=\textwidth]{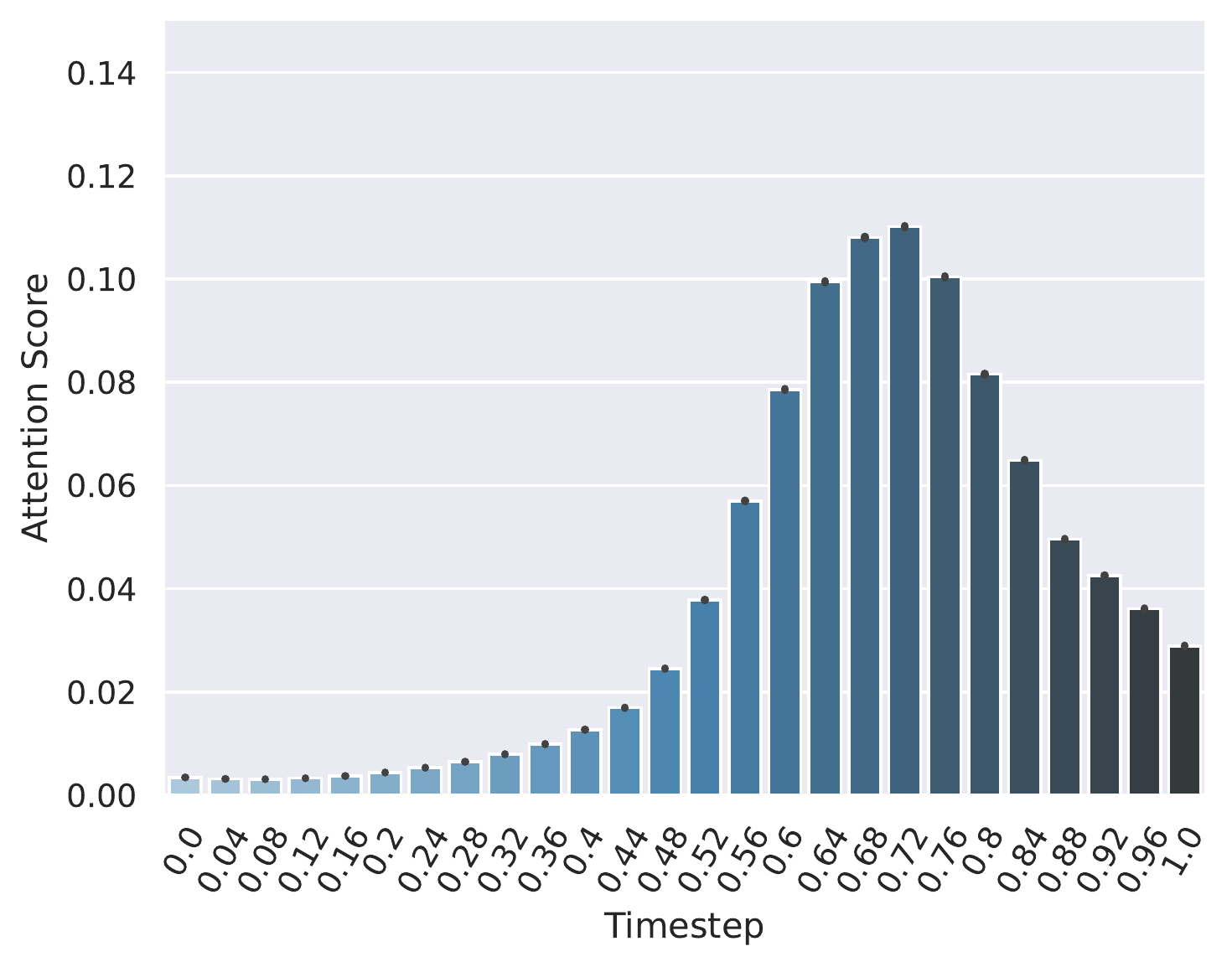}
\vspace{-6mm}
\subcaption*{\scriptsize Object Shape}
\end{subfigure}
\subcaption*{Granularity: 25}
\end{subfigure} \\
\caption{Attention score profiles for the synthetic dataset on the different features, using different granularities, with the dimensionality of the latent space as 32 and the DRL encoder.}
\label{fig:syn_DRL_32}
\end{figure}

\begin{figure}
    \centering
    \begin{subfigure}[c]{\linewidth}
    \includegraphics[width=\textwidth]{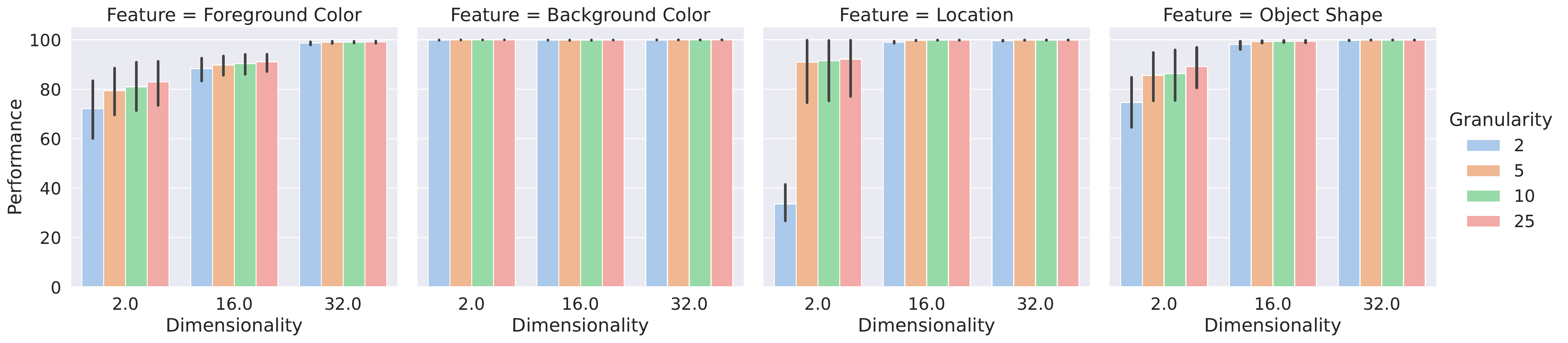}
    \subcaption{VDRL Encoder}
    \end{subfigure}
    \begin{subfigure}[c]{\linewidth}
    \includegraphics[width=\textwidth]{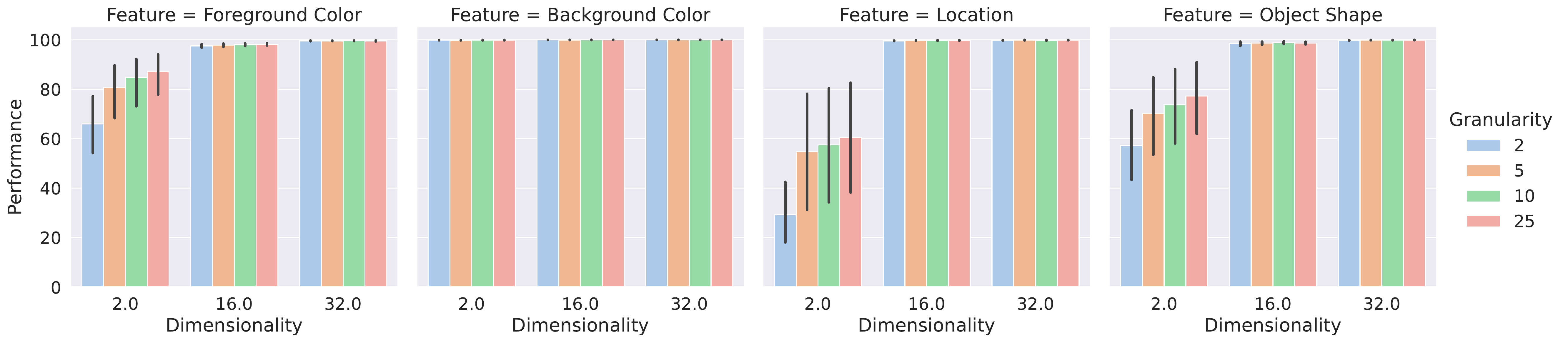}
    \subcaption{DRL Encoder}
    \end{subfigure}
    \caption{Downstream performance plots for the Synthetic Dataset for different features, when the score model is trained with different latent dimensionality and the downstream models are trained with different granularities for discretization.}
    \label{fig:syn_abla}
\end{figure}

The corresponding plots for the attention score profiles are present in Figures \ref{fig:syn_VDRL_2} - \ref{fig:syn_DRL_32} for different latent space dimensionalities, different granularities and the different types of encoding schemes (VDRL and DRL). Further analysis into the performance on different features with different granularities and dimensionalities can be found in Figure \ref{fig:syn_abla}.


\begin{figure}
\begin{subfigure}[c]{\textwidth}
\begin{subfigure}[c]{0.32\textwidth}
\includegraphics[width=\textwidth]{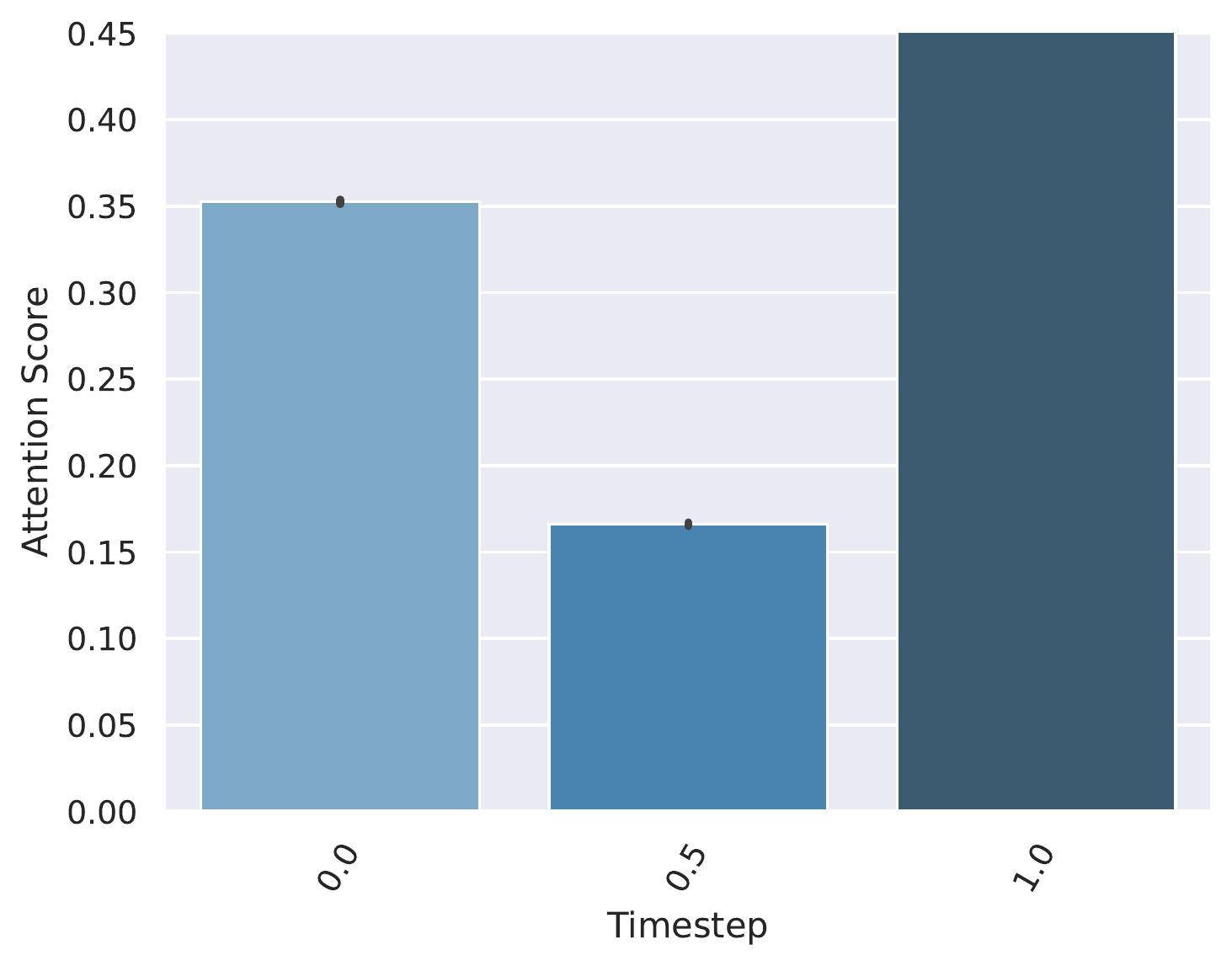}
\vspace{-6mm}
\subcaption*{\scriptsize Background Color}
\end{subfigure}
\begin{subfigure}[c]{0.32\textwidth}
\includegraphics[width=\textwidth]{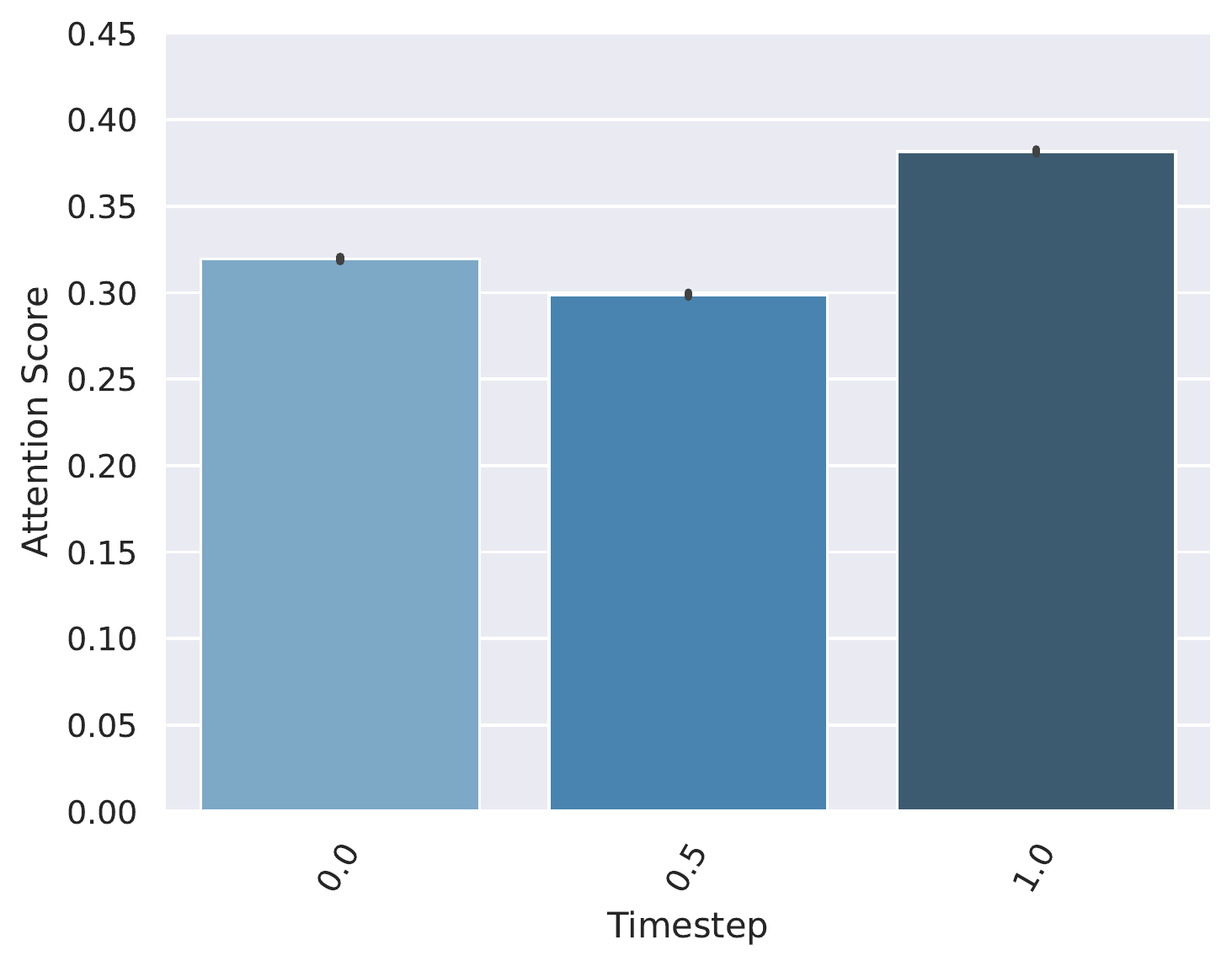}
\vspace{-6mm}
\subcaption*{\scriptsize Foreground Color}
\end{subfigure}
\begin{subfigure}[c]{0.32\textwidth}
\includegraphics[width=\textwidth]{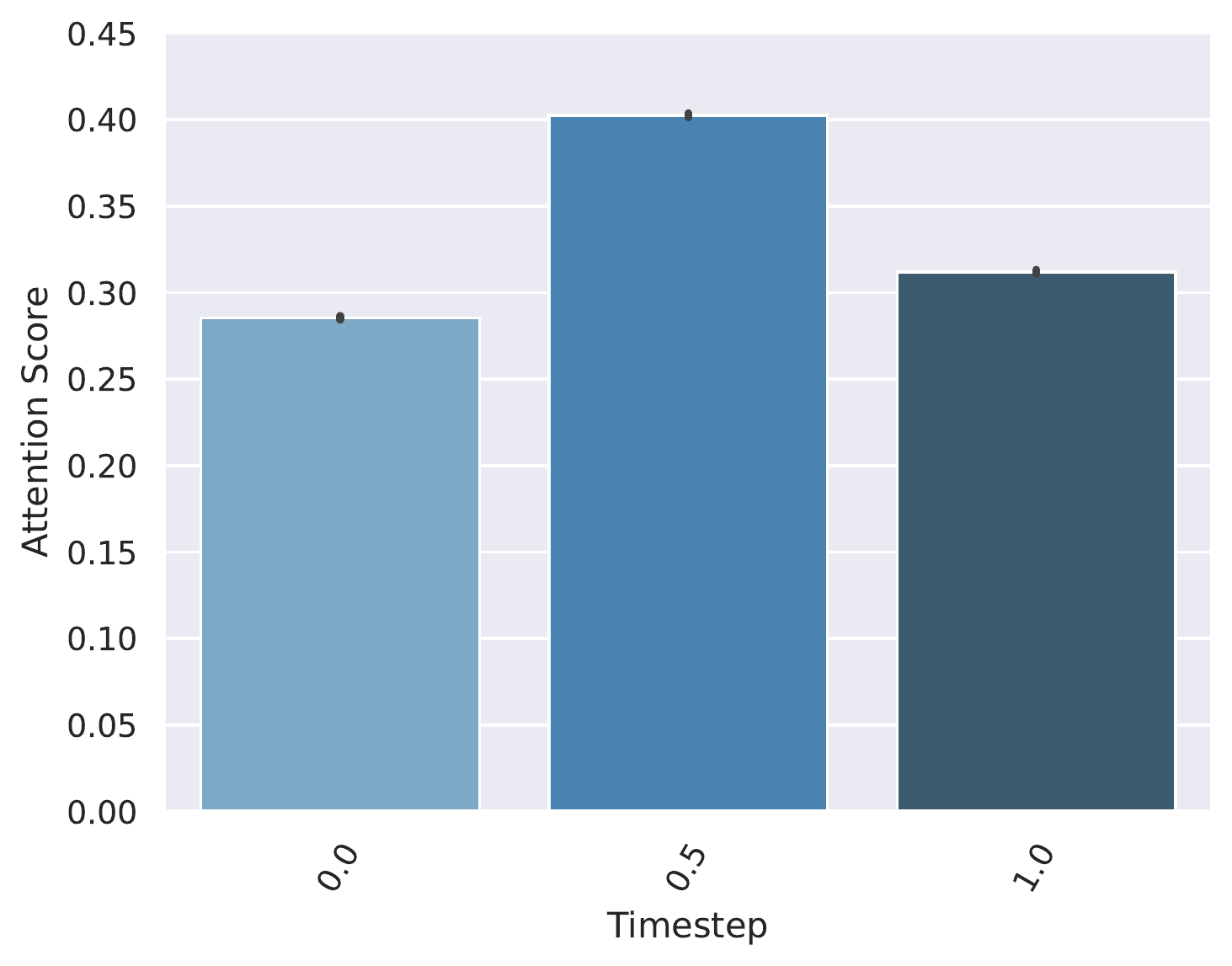}
\vspace{-6mm}
\subcaption*{\scriptsize Digit}
\end{subfigure}
\subcaption*{Granularity: 2}
\end{subfigure} \\
\begin{subfigure}[c]{\textwidth}
\begin{subfigure}[c]{0.32\textwidth}
\includegraphics[width=\textwidth]{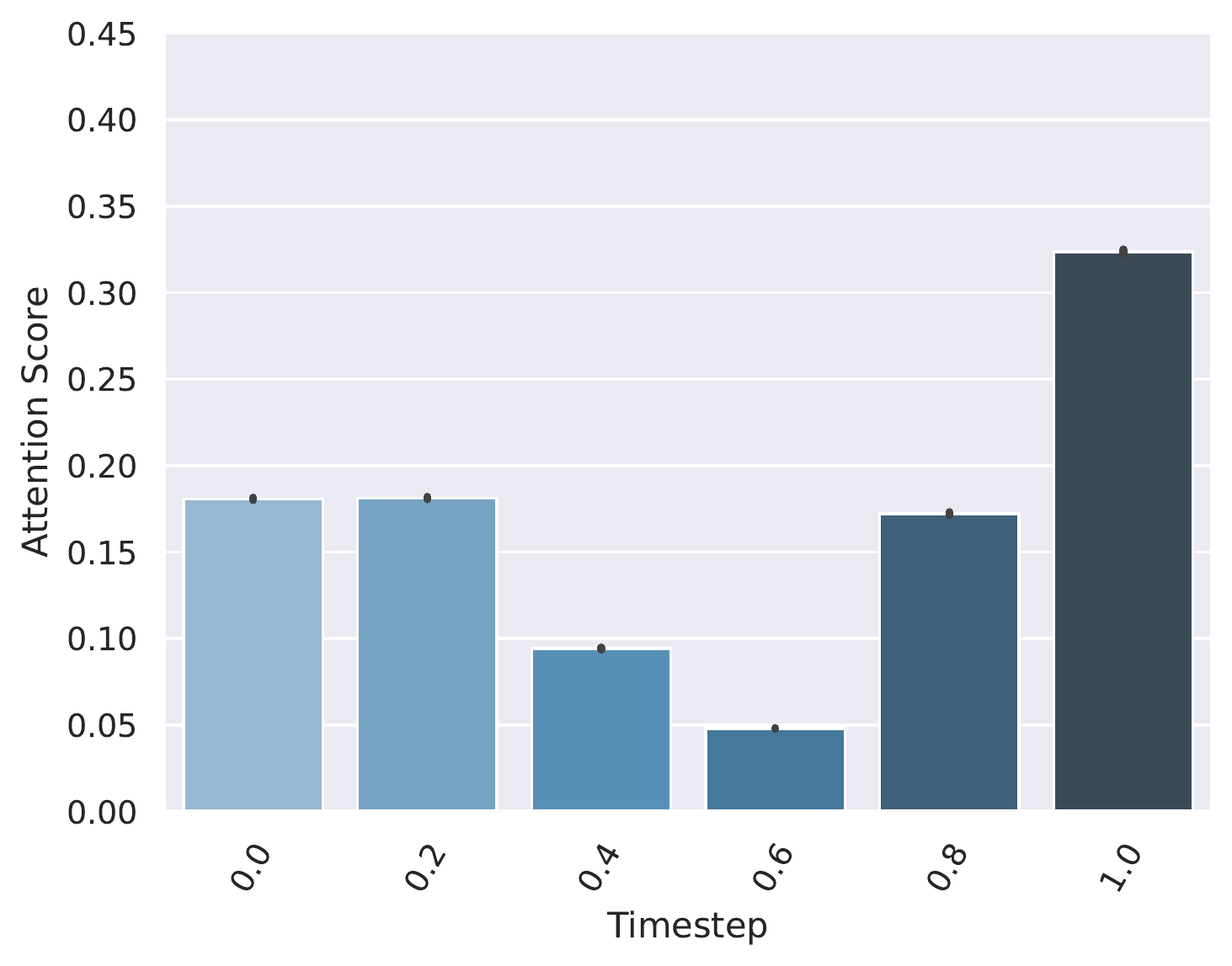}
\vspace{-6mm}
\subcaption*{\scriptsize Background Color}
\end{subfigure}
\begin{subfigure}[c]{0.32\textwidth}
\includegraphics[width=\textwidth]{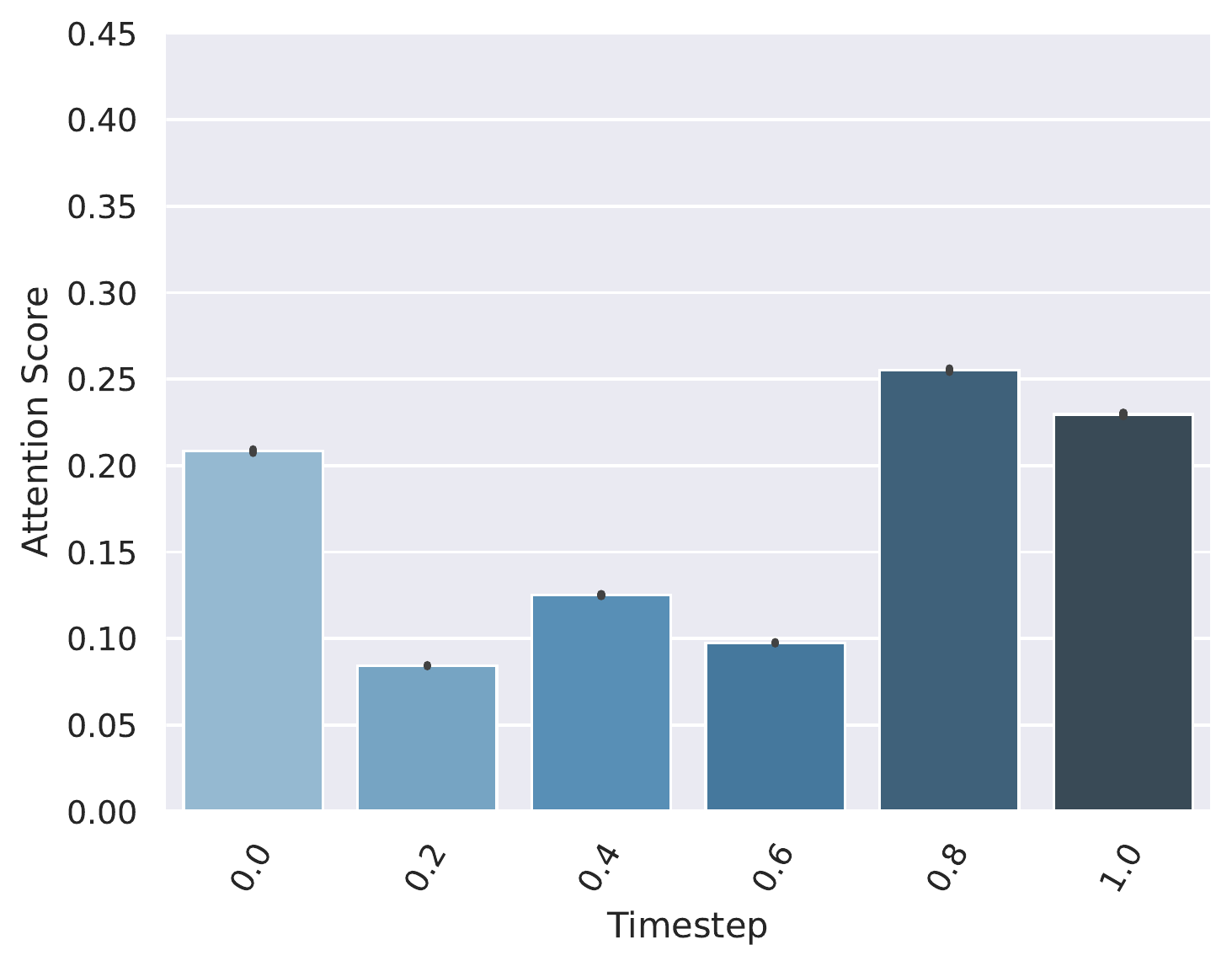}
\vspace{-6mm}
\subcaption*{\scriptsize Foreground Color}
\end{subfigure}
\begin{subfigure}[c]{0.32\textwidth}
\includegraphics[width=\textwidth]{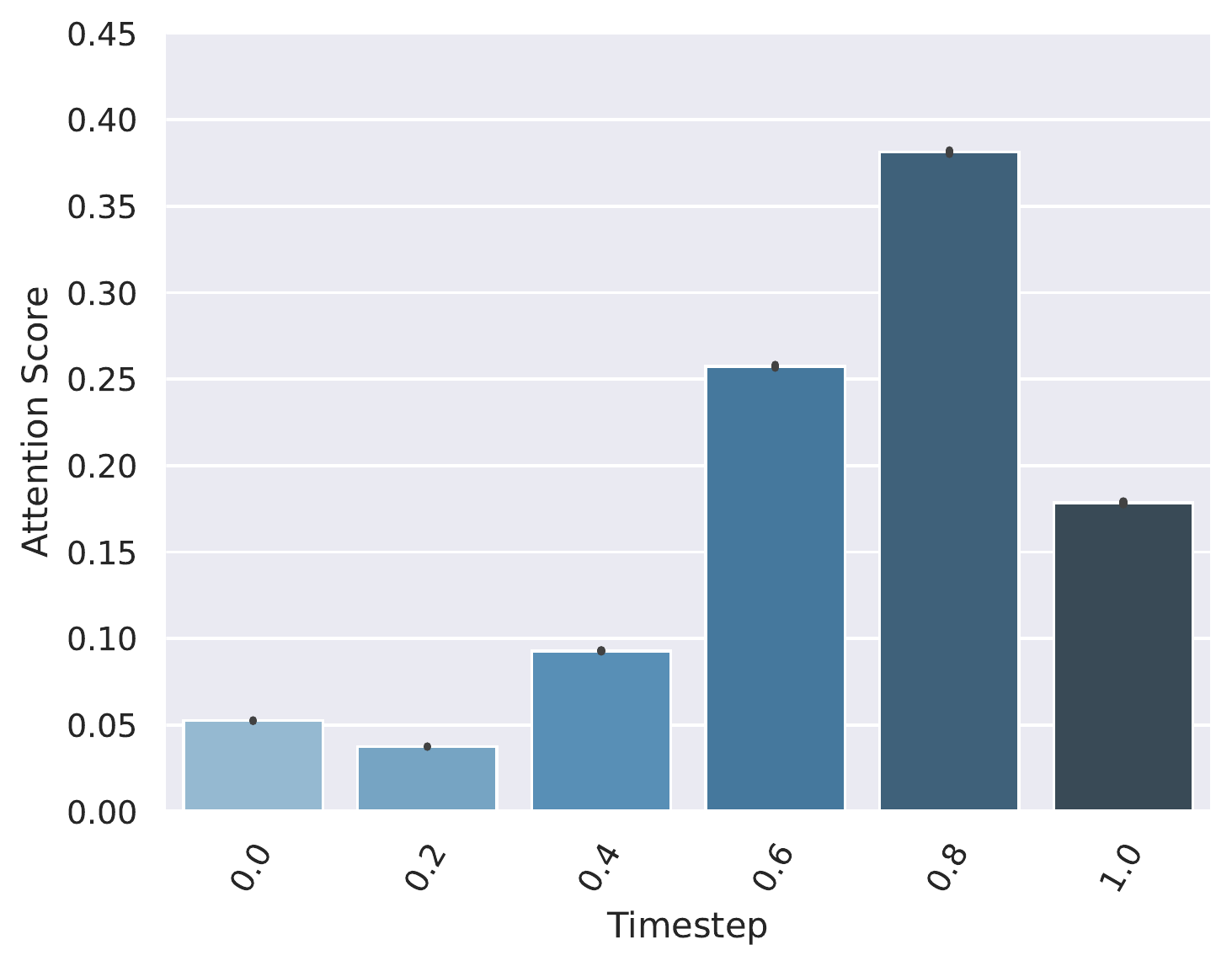}
\vspace{-6mm}
\subcaption*{\scriptsize Digit}
\end{subfigure}
\subcaption*{Granularity: 5}
\end{subfigure} \\
\begin{subfigure}[c]{\textwidth}
\begin{subfigure}[c]{0.32\textwidth}
\includegraphics[width=\textwidth]{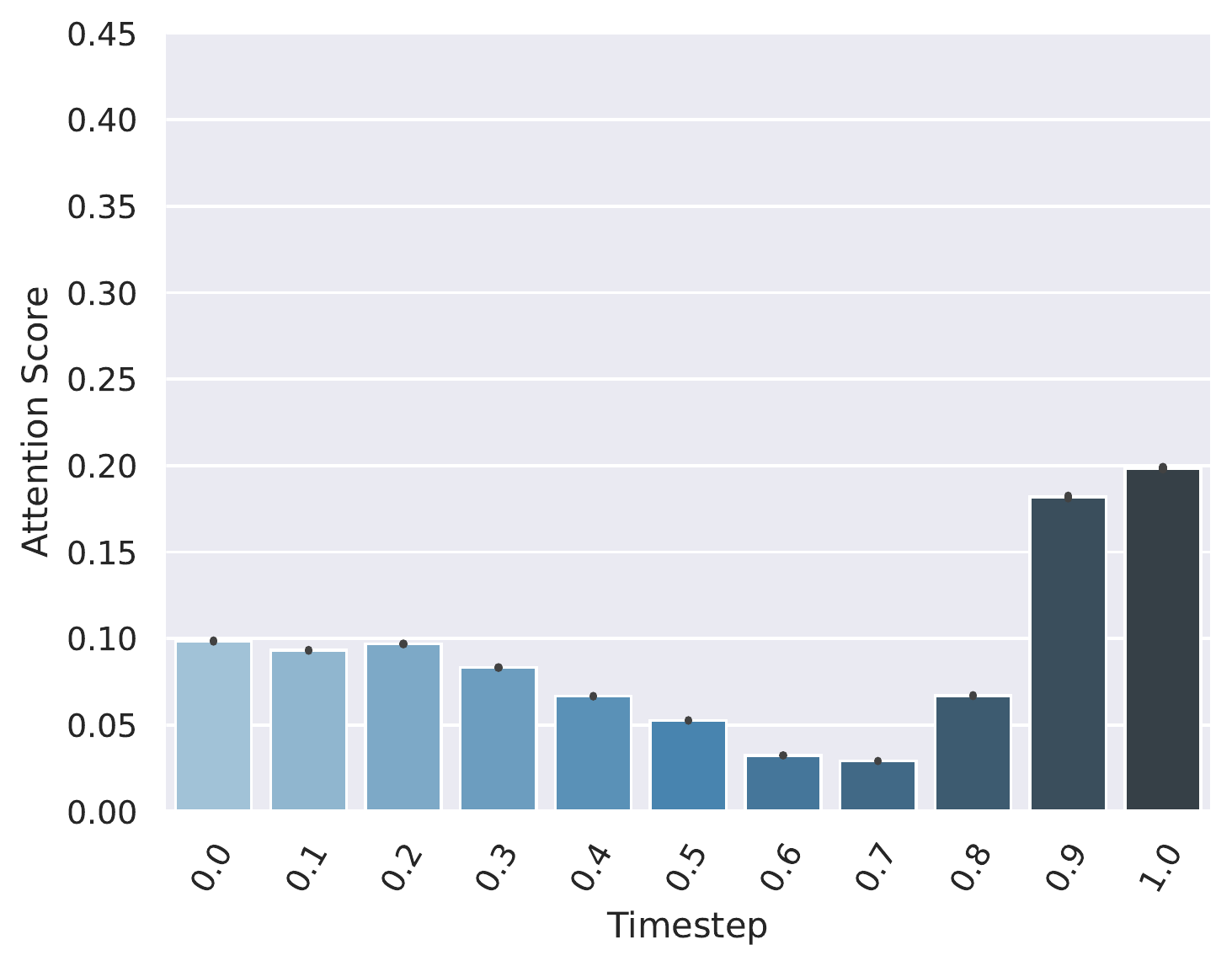}
\vspace{-6mm}
\subcaption*{\scriptsize Background Color}
\end{subfigure}
\begin{subfigure}[c]{0.32\textwidth}
\includegraphics[width=\textwidth]{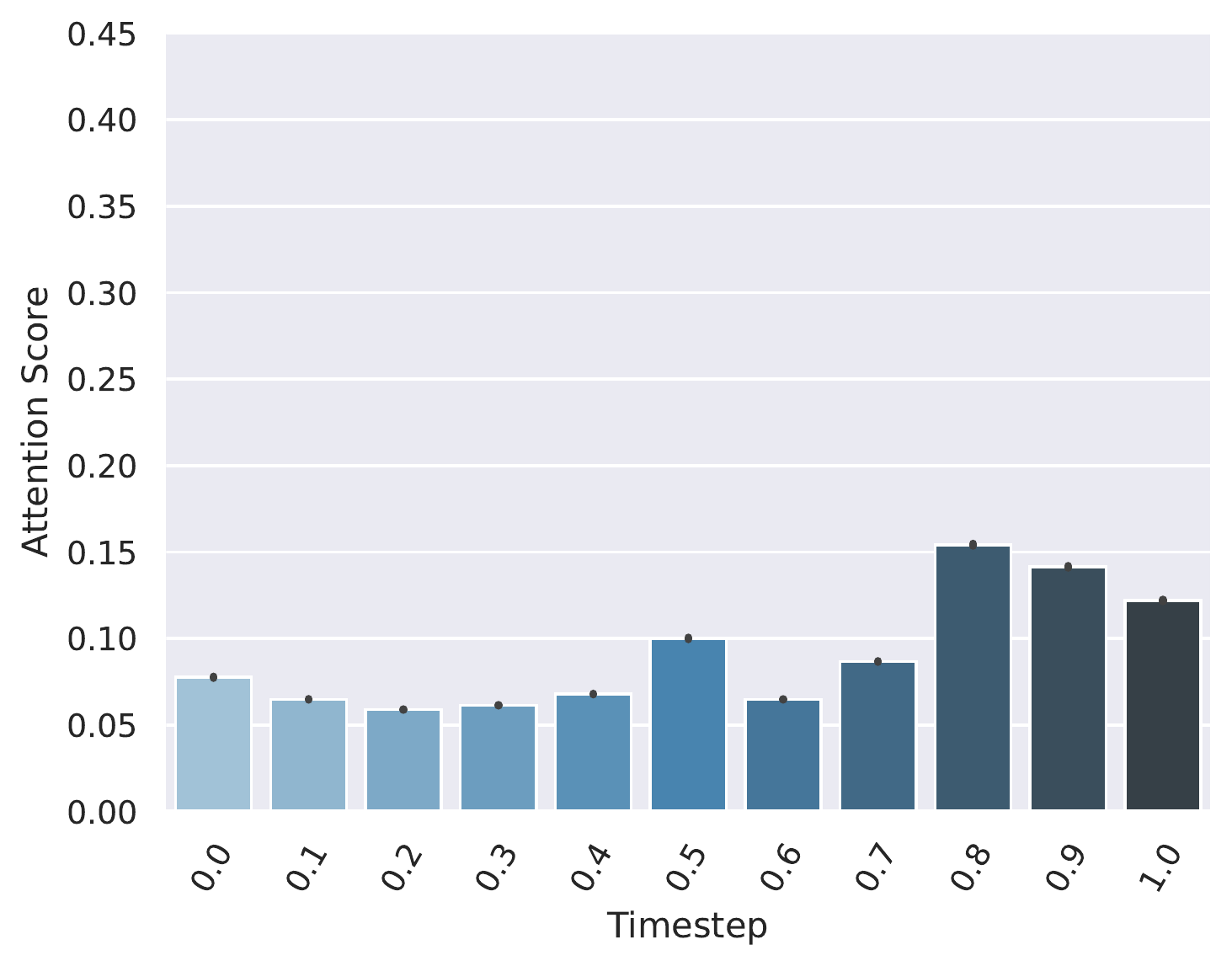}
\vspace{-6mm}
\subcaption*{\scriptsize Foreground Color}
\end{subfigure}
\begin{subfigure}[c]{0.32\textwidth}
\includegraphics[width=\textwidth]{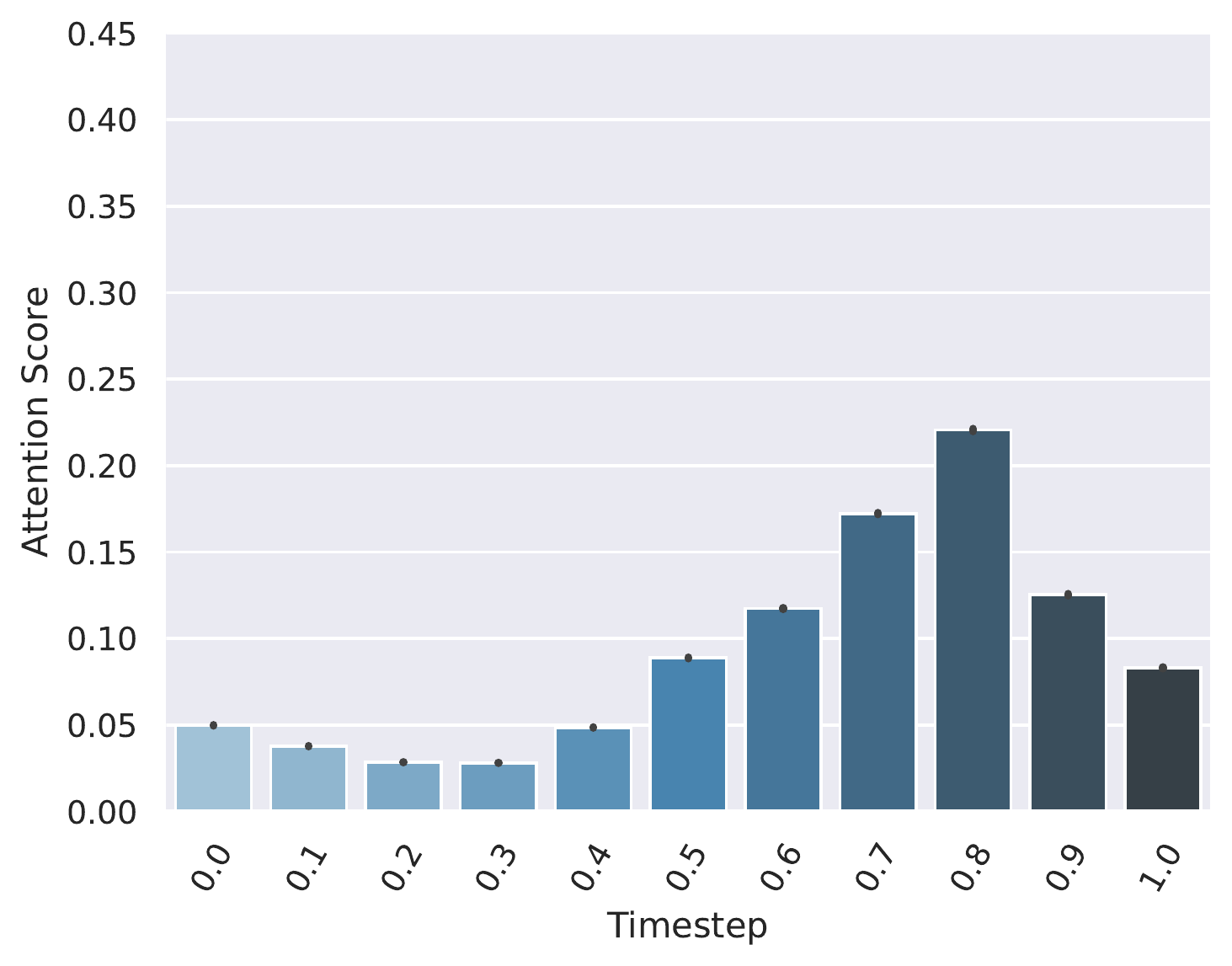}
\vspace{-6mm}
\subcaption*{\scriptsize Digit}
\end{subfigure}
\subcaption*{Granularity: 10}
\end{subfigure} \\
\begin{subfigure}[c]{\textwidth}
\begin{subfigure}[c]{0.32\textwidth}
\includegraphics[width=\textwidth]{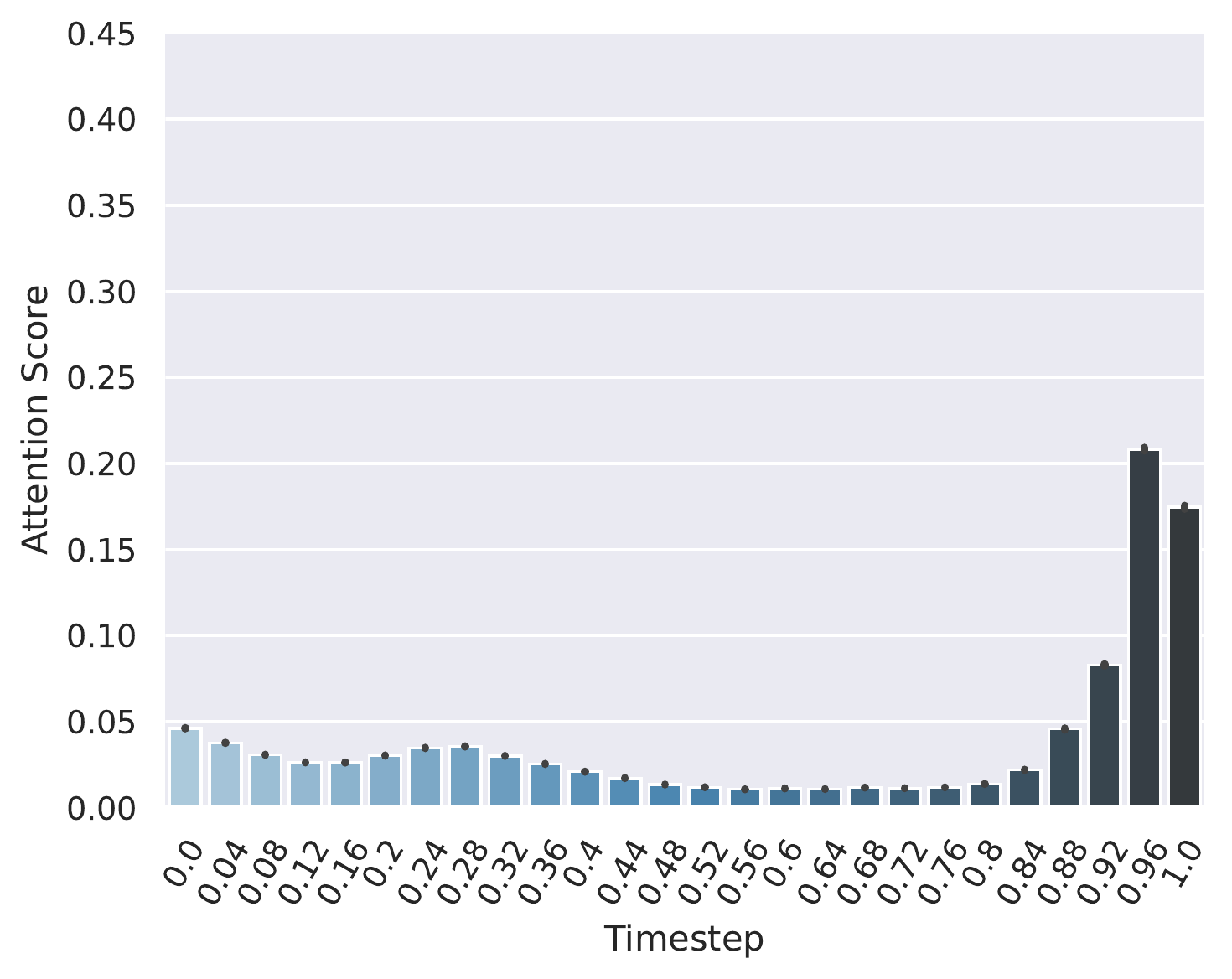}
\vspace{-6mm}
\subcaption*{\scriptsize Background Color}
\end{subfigure}
\begin{subfigure}[c]{0.32\textwidth}
\includegraphics[width=\textwidth]{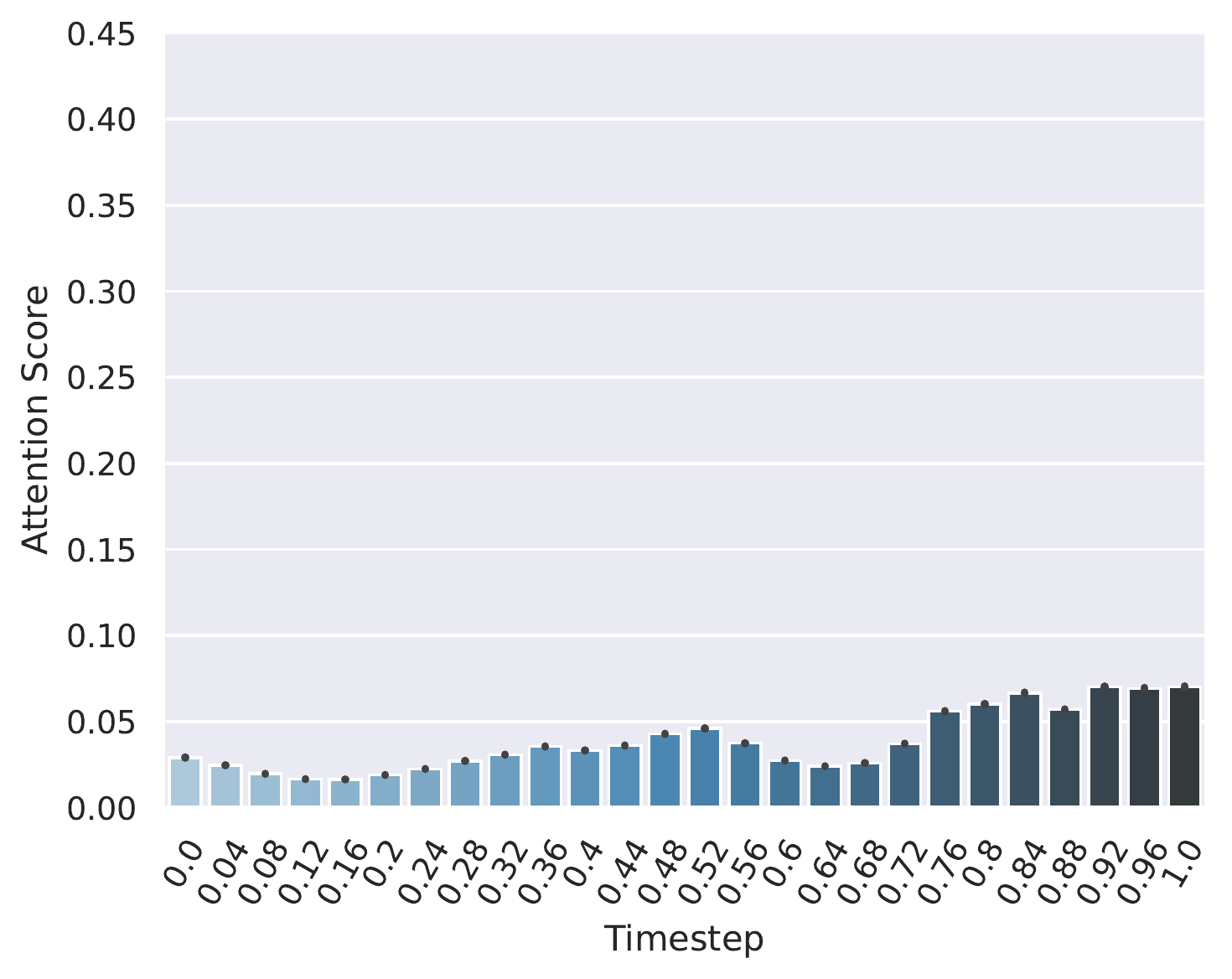}
\vspace{-6mm}
\subcaption*{\scriptsize Foreground Color}
\end{subfigure}
\begin{subfigure}[c]{0.32\textwidth}
\includegraphics[width=\textwidth]{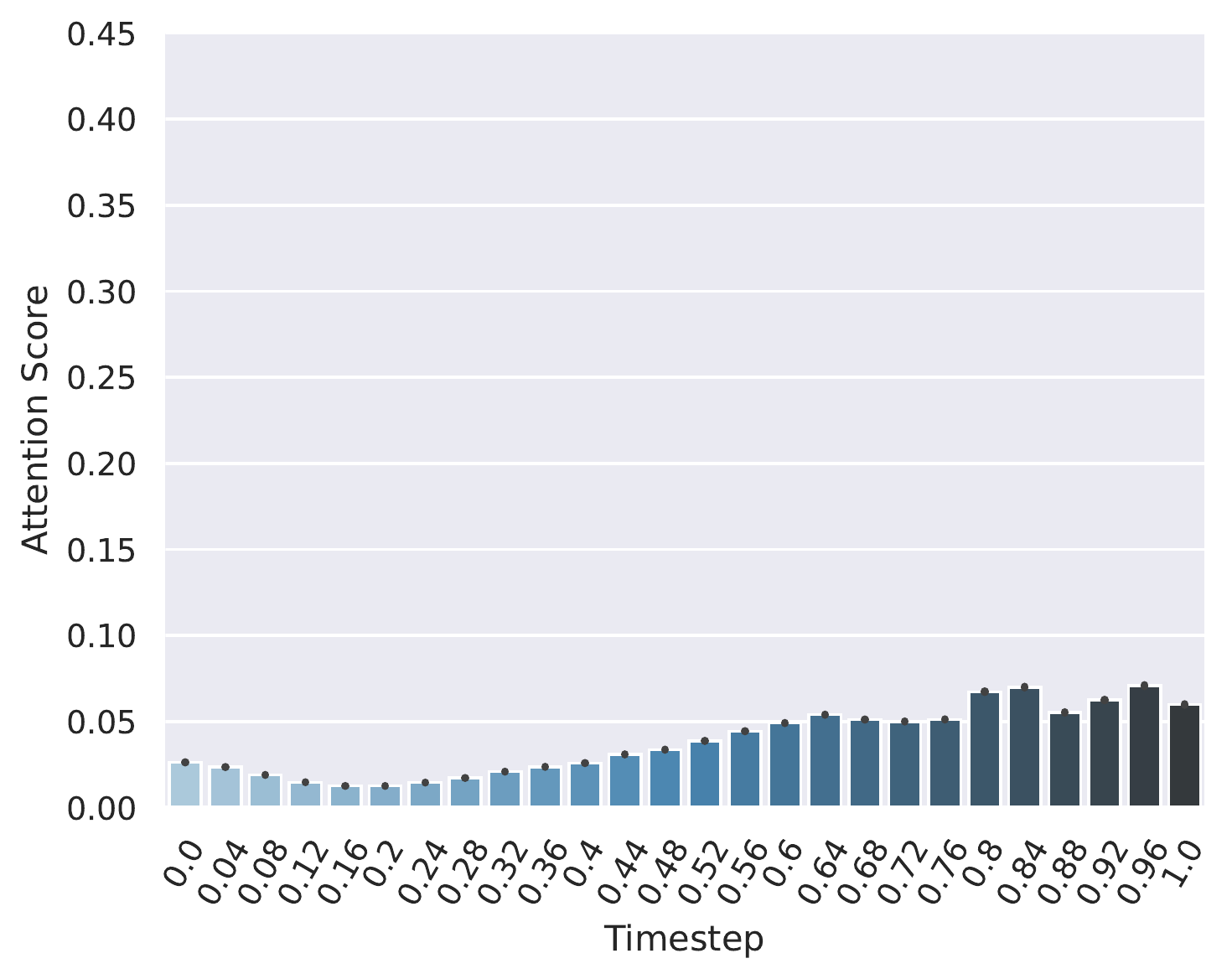}
\vspace{-6mm}
\subcaption*{\scriptsize Digit}
\end{subfigure}
\subcaption*{Granularity: 25}
\end{subfigure} \\
\caption{Attention score profiles for the Colored-MNIST dataset on the different features, using different granularities, with the dimensionality of the latent space as 2 and the VDRL encoder.}
\label{fig:cm_VDRL_2}
\end{figure}
\begin{figure}
\begin{subfigure}[c]{\textwidth}
\begin{subfigure}[c]{0.32\textwidth}
\includegraphics[width=\textwidth]{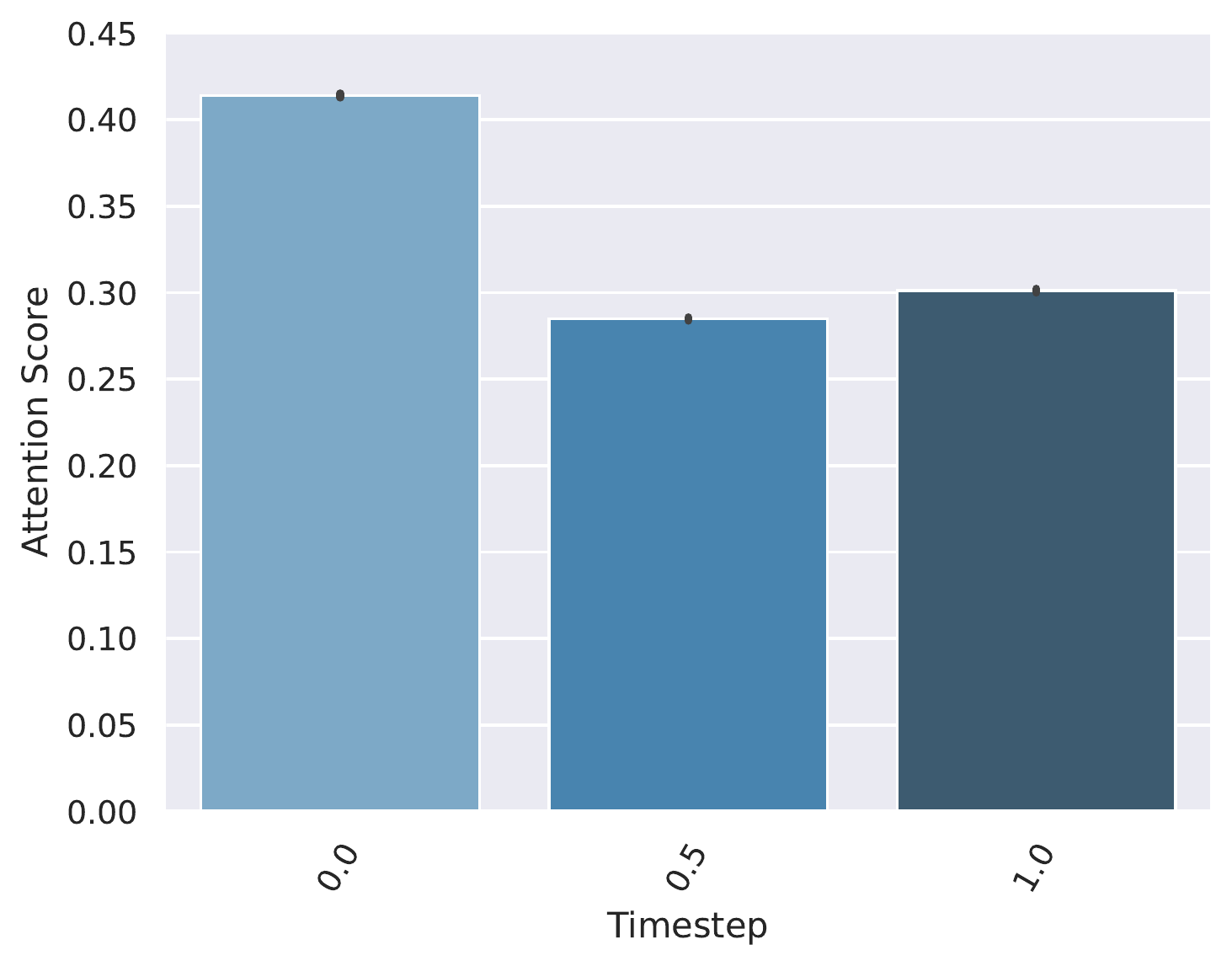}
\vspace{-6mm}
\subcaption*{\scriptsize Background Color}
\end{subfigure}
\begin{subfigure}[c]{0.32\textwidth}
\includegraphics[width=\textwidth]{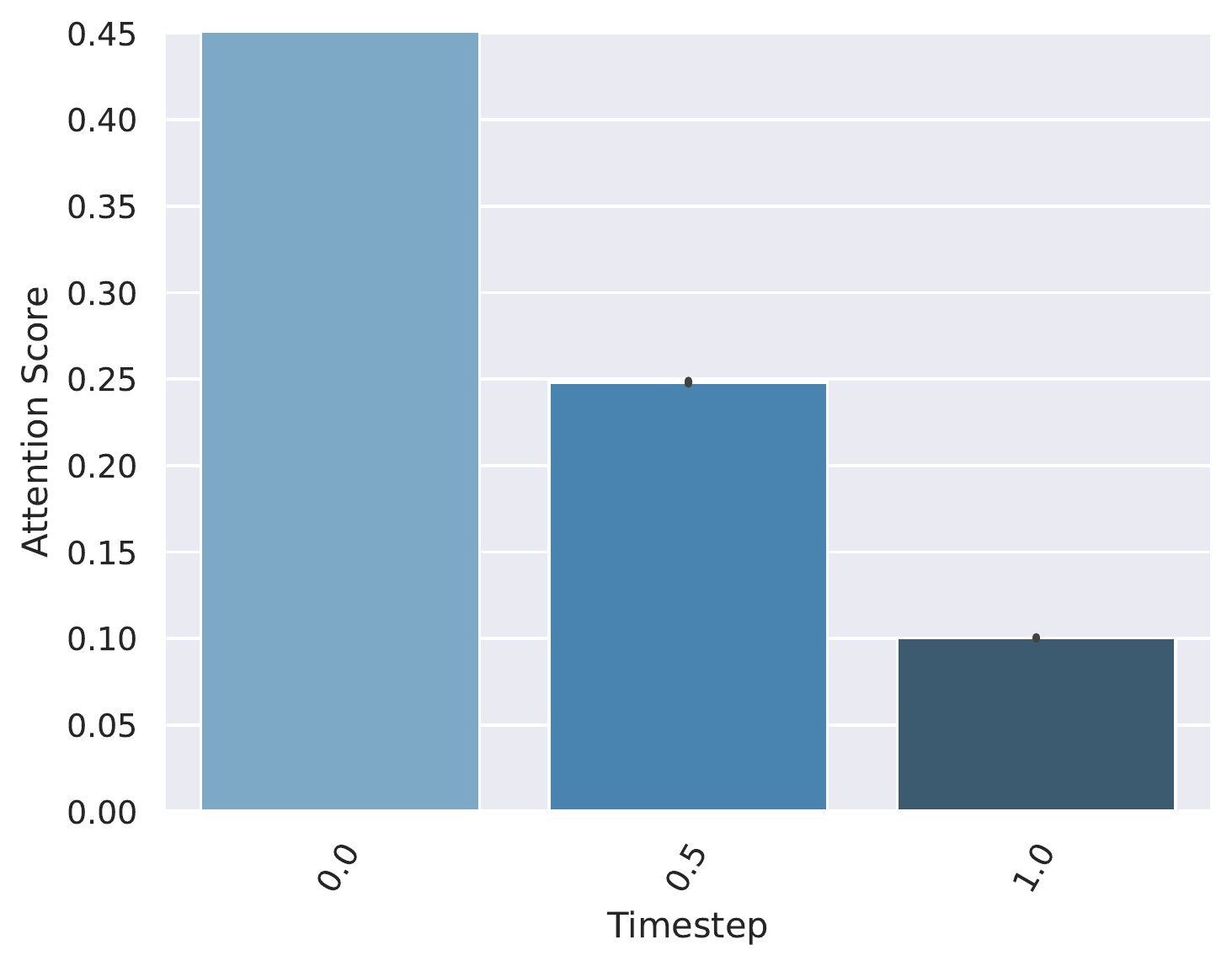}
\vspace{-6mm}
\subcaption*{\scriptsize Foreground Color}
\end{subfigure}
\begin{subfigure}[c]{0.32\textwidth}
\includegraphics[width=\textwidth]{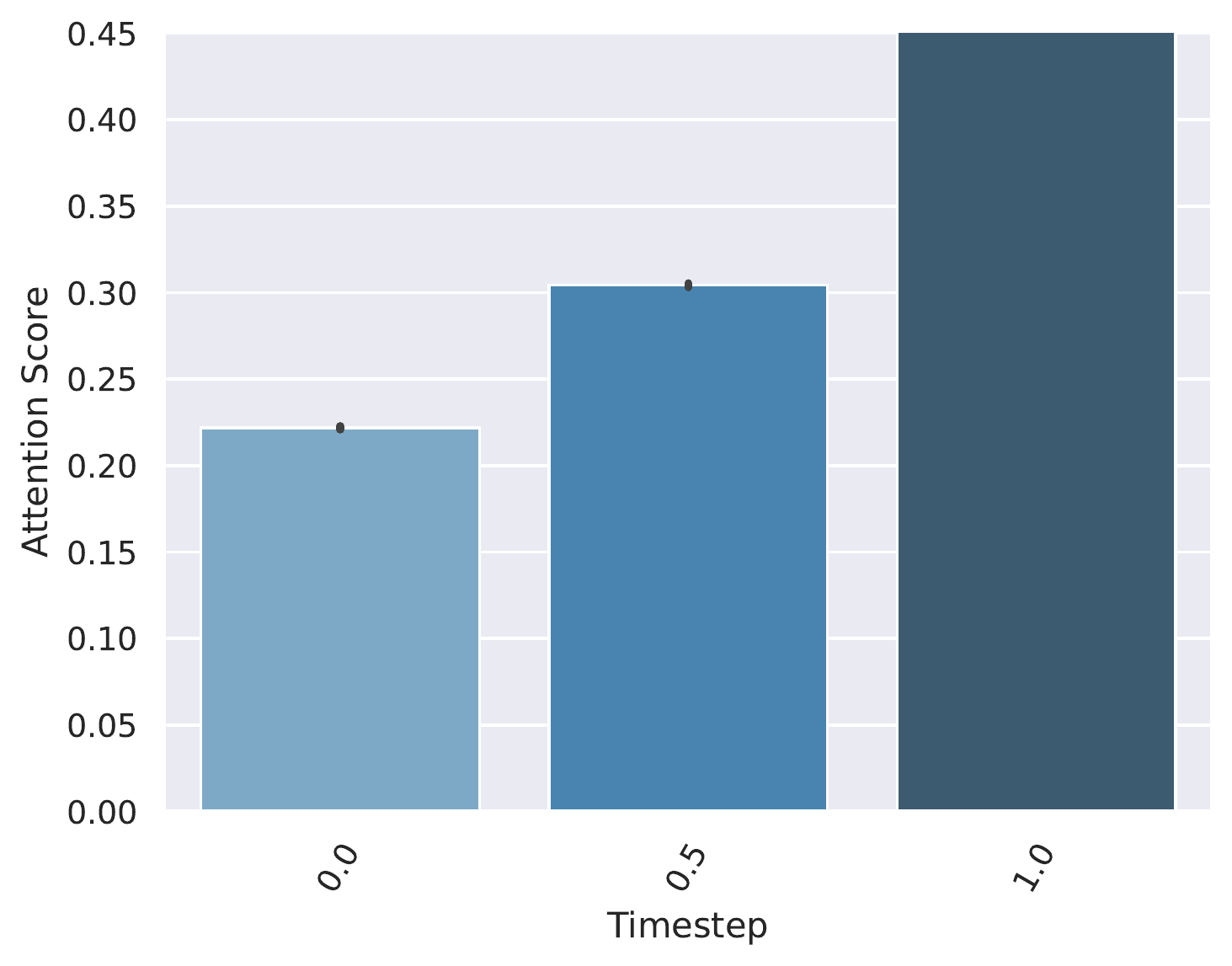}
\vspace{-6mm}
\subcaption*{\scriptsize Digit}
\end{subfigure}
\subcaption*{Granularity: 2}
\end{subfigure} \\
\begin{subfigure}[c]{\textwidth}
\begin{subfigure}[c]{0.32\textwidth}
\includegraphics[width=\textwidth]{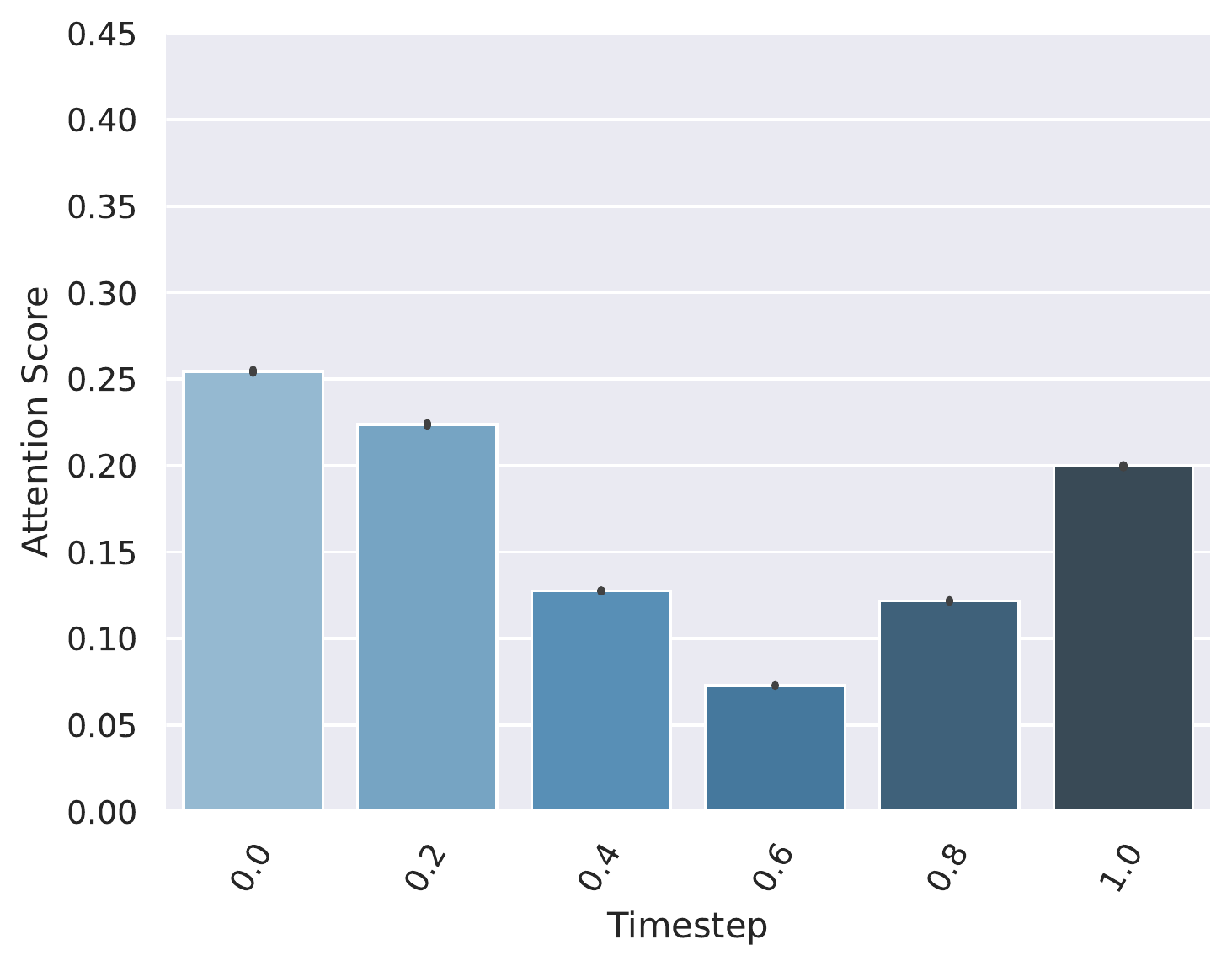}
\vspace{-6mm}
\subcaption*{\scriptsize Background Color}
\end{subfigure}
\begin{subfigure}[c]{0.32\textwidth}
\includegraphics[width=\textwidth]{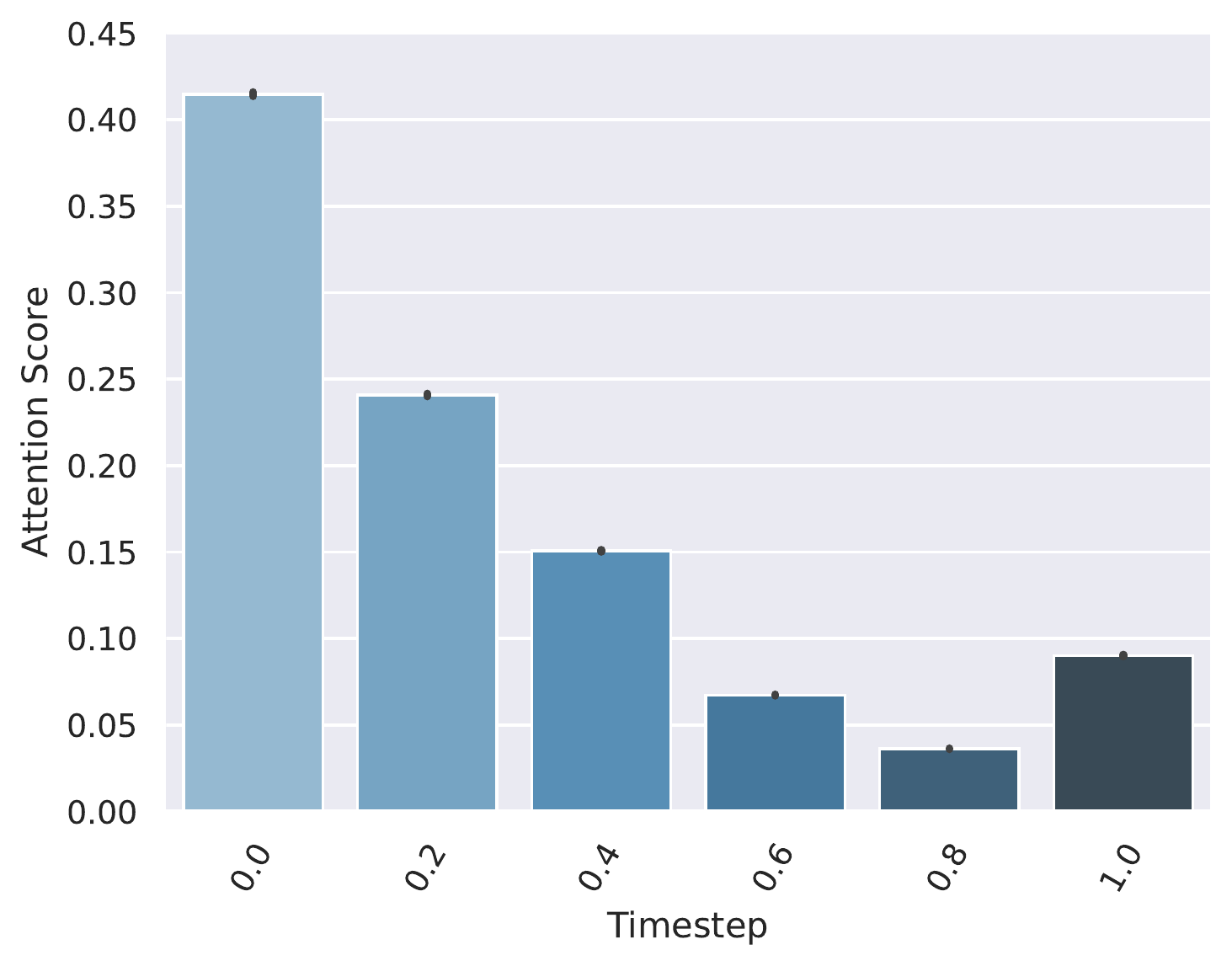}
\vspace{-6mm}
\subcaption*{\scriptsize Foreground Color}
\end{subfigure}
\begin{subfigure}[c]{0.32\textwidth}
\includegraphics[width=\textwidth]{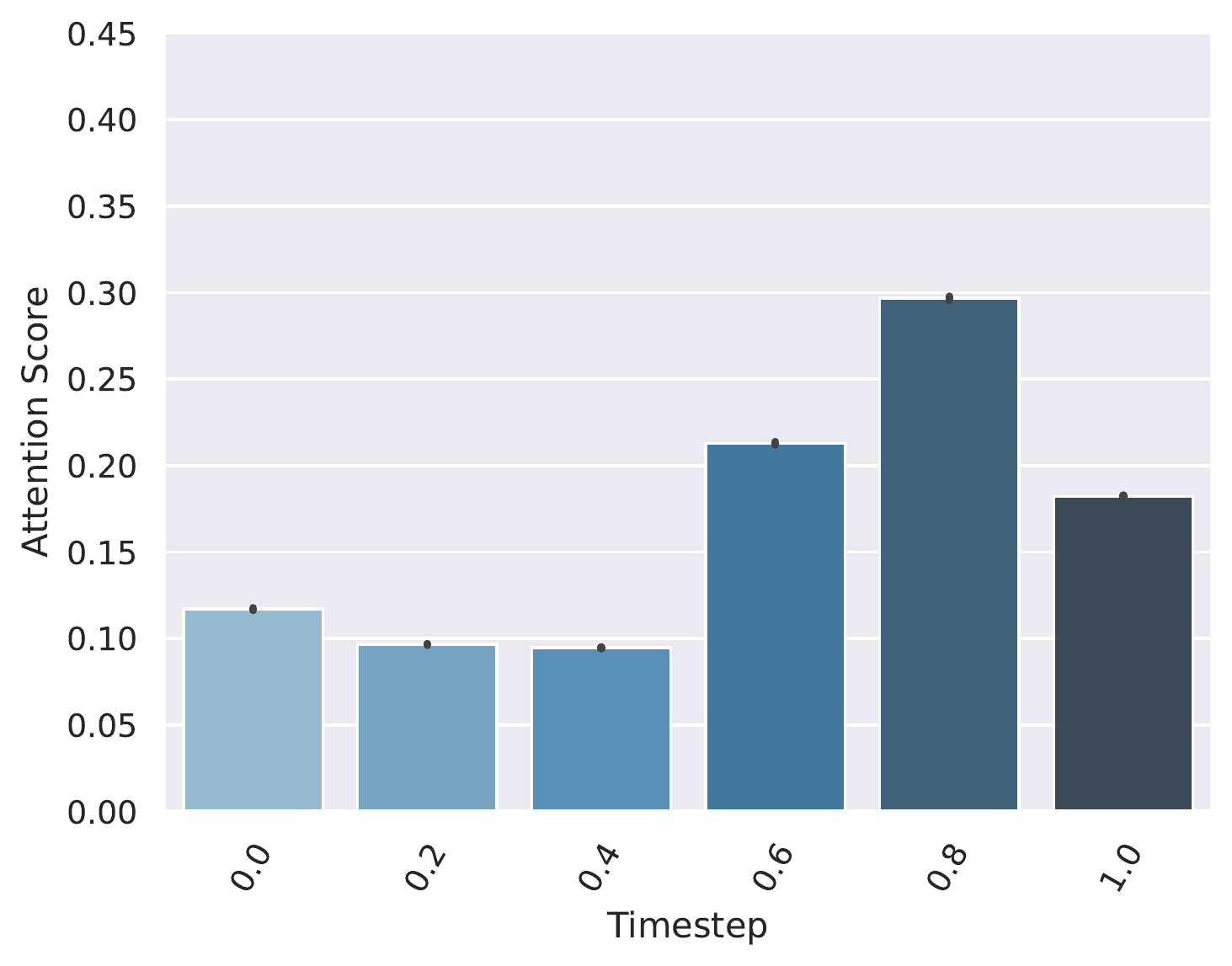}
\vspace{-6mm}
\subcaption*{\scriptsize Digit}
\end{subfigure}
\subcaption*{Granularity: 5}
\end{subfigure} \\
\begin{subfigure}[c]{\textwidth}
\begin{subfigure}[c]{0.32\textwidth}
\includegraphics[width=\textwidth]{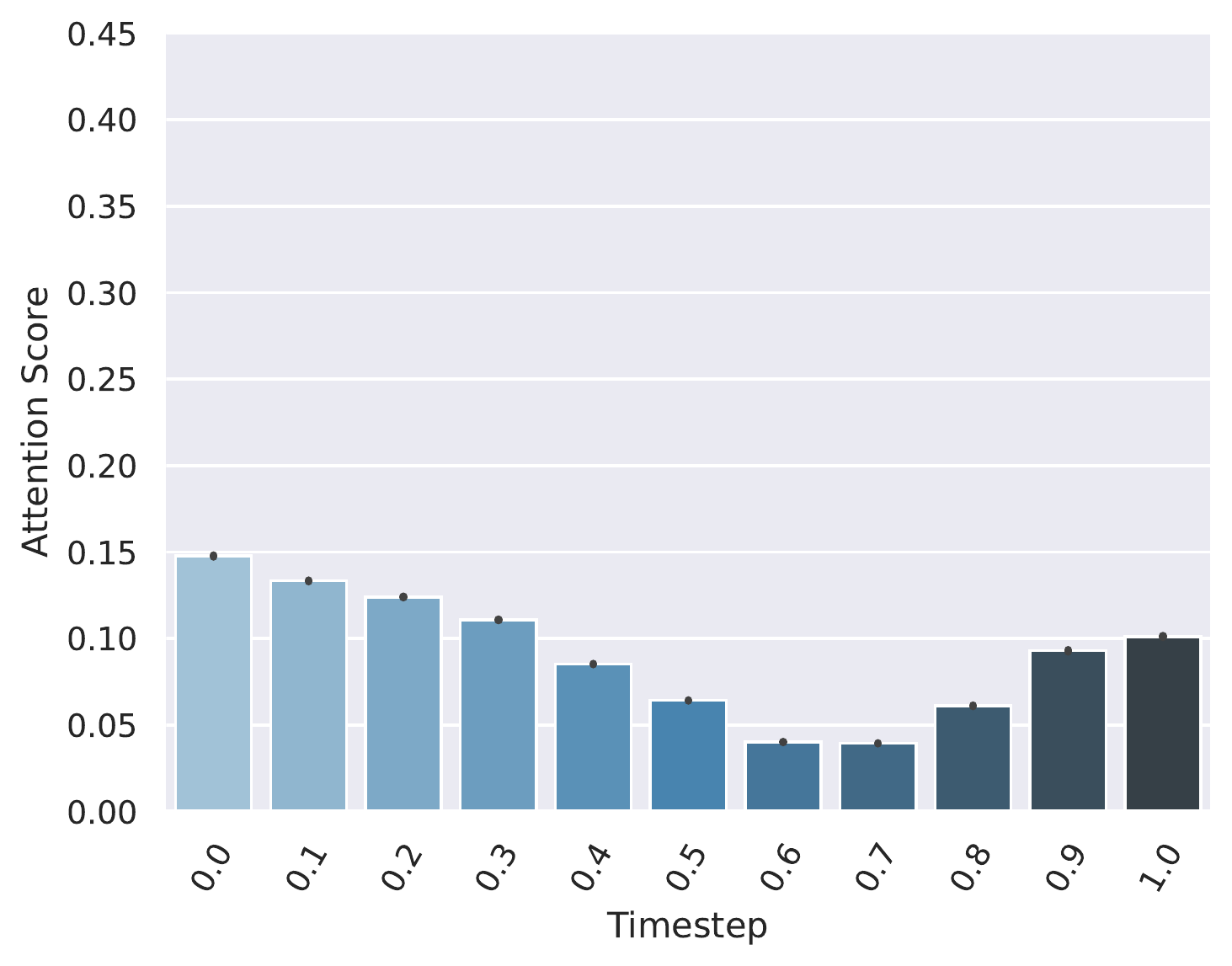}
\vspace{-6mm}
\subcaption*{\scriptsize Background Color}
\end{subfigure}
\begin{subfigure}[c]{0.32\textwidth}
\includegraphics[width=\textwidth]{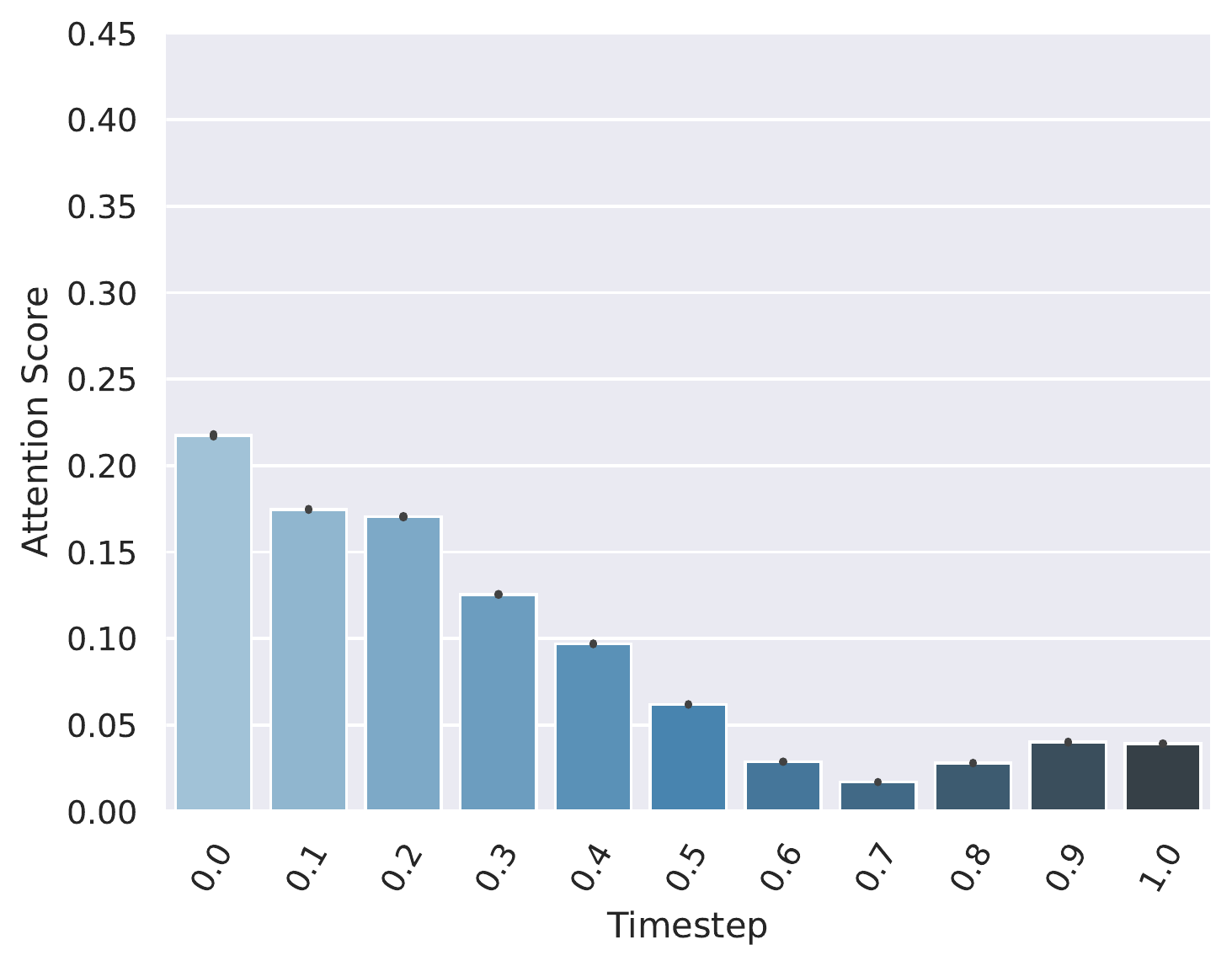}
\vspace{-6mm}
\subcaption*{\scriptsize Foreground Color}
\end{subfigure}
\begin{subfigure}[c]{0.32\textwidth}
\includegraphics[width=\textwidth]{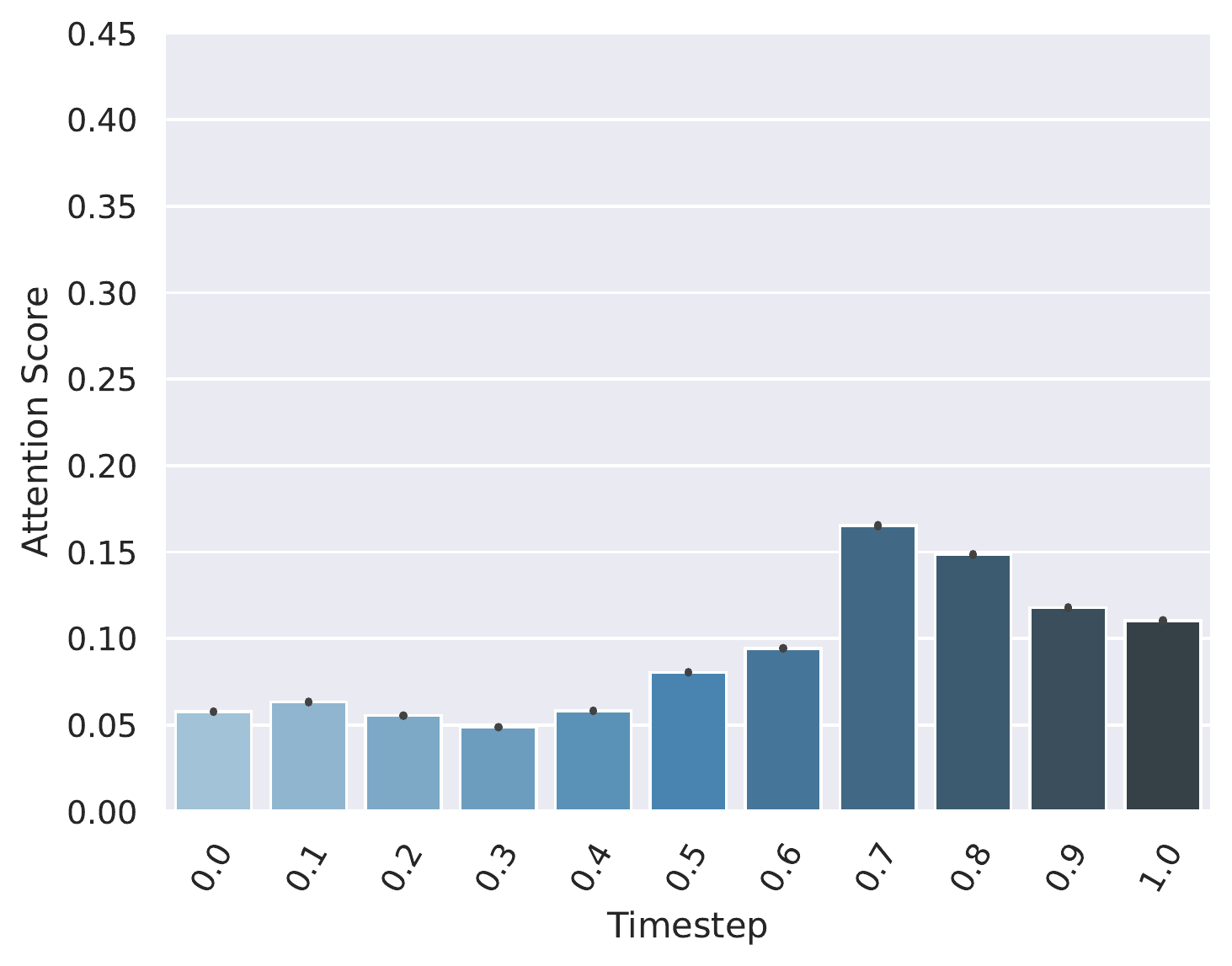}
\vspace{-6mm}
\subcaption*{\scriptsize Digit}
\end{subfigure}
\subcaption*{Granularity: 10}
\end{subfigure} \\
\begin{subfigure}[c]{\textwidth}
\begin{subfigure}[c]{0.32\textwidth}
\includegraphics[width=\textwidth]{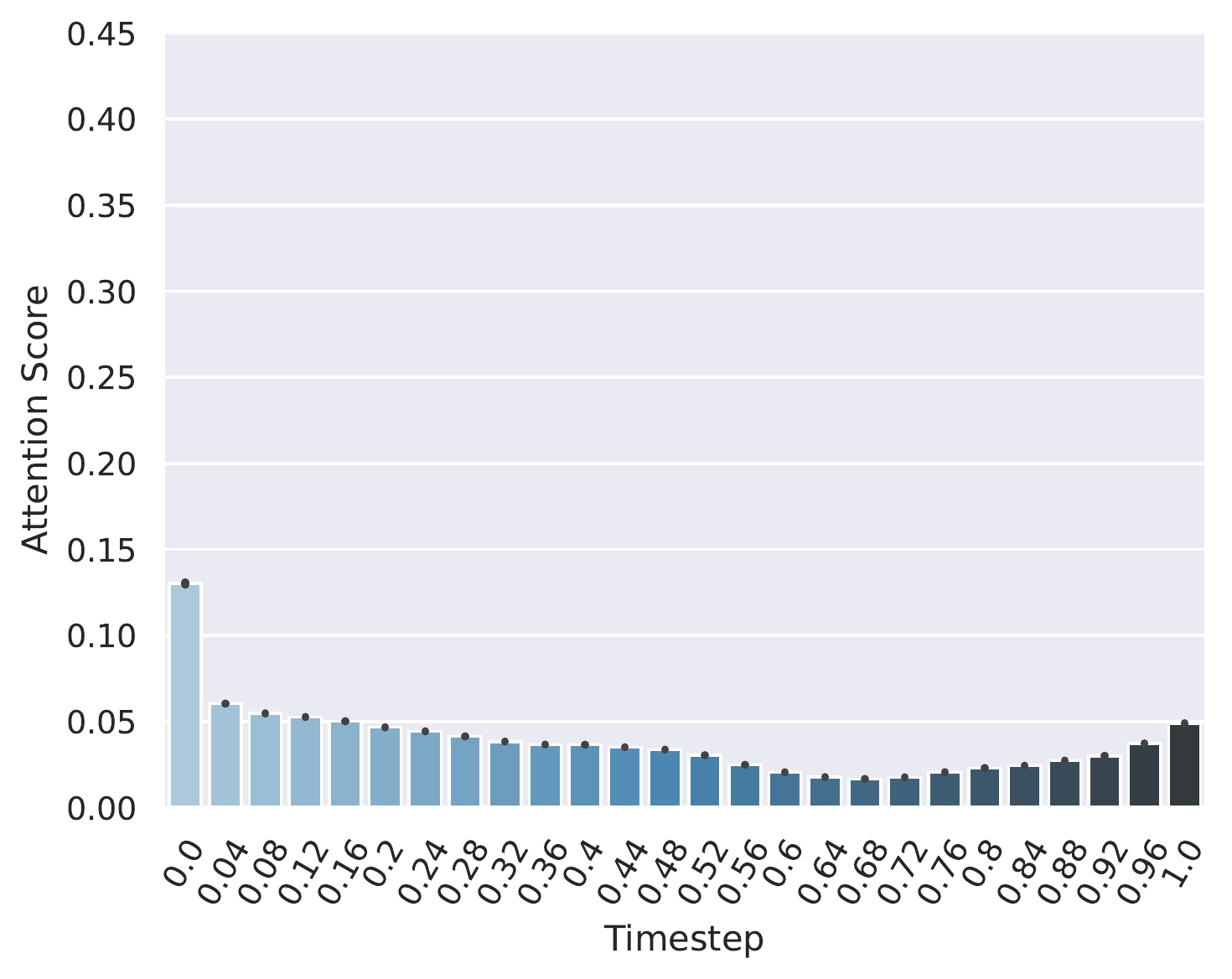}
\vspace{-6mm}
\subcaption*{\scriptsize Background Color}
\end{subfigure}
\begin{subfigure}[c]{0.32\textwidth}
\includegraphics[width=\textwidth]{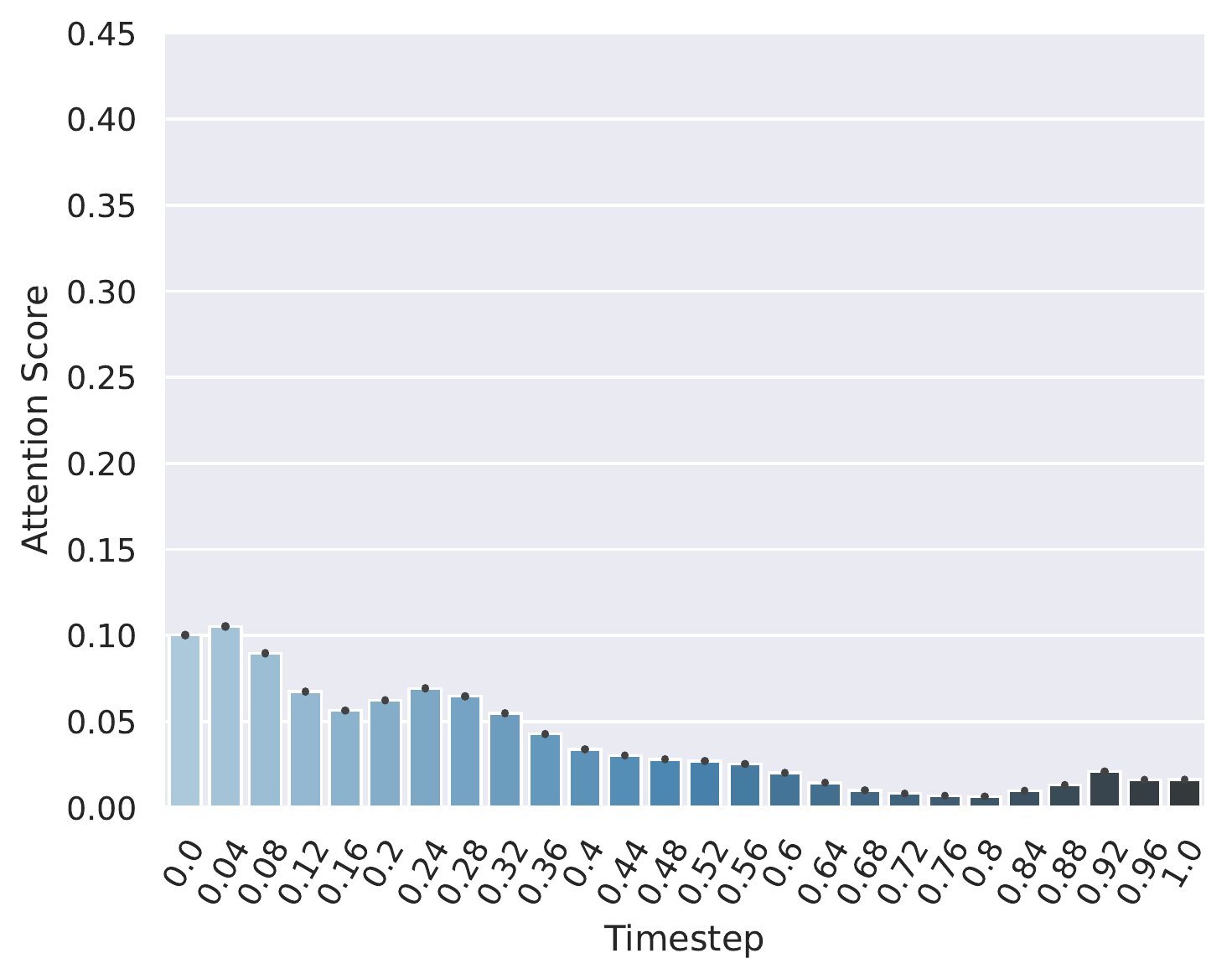}
\vspace{-6mm}
\subcaption*{\scriptsize Foreground Color}
\end{subfigure}
\begin{subfigure}[c]{0.32\textwidth}
\includegraphics[width=\textwidth]{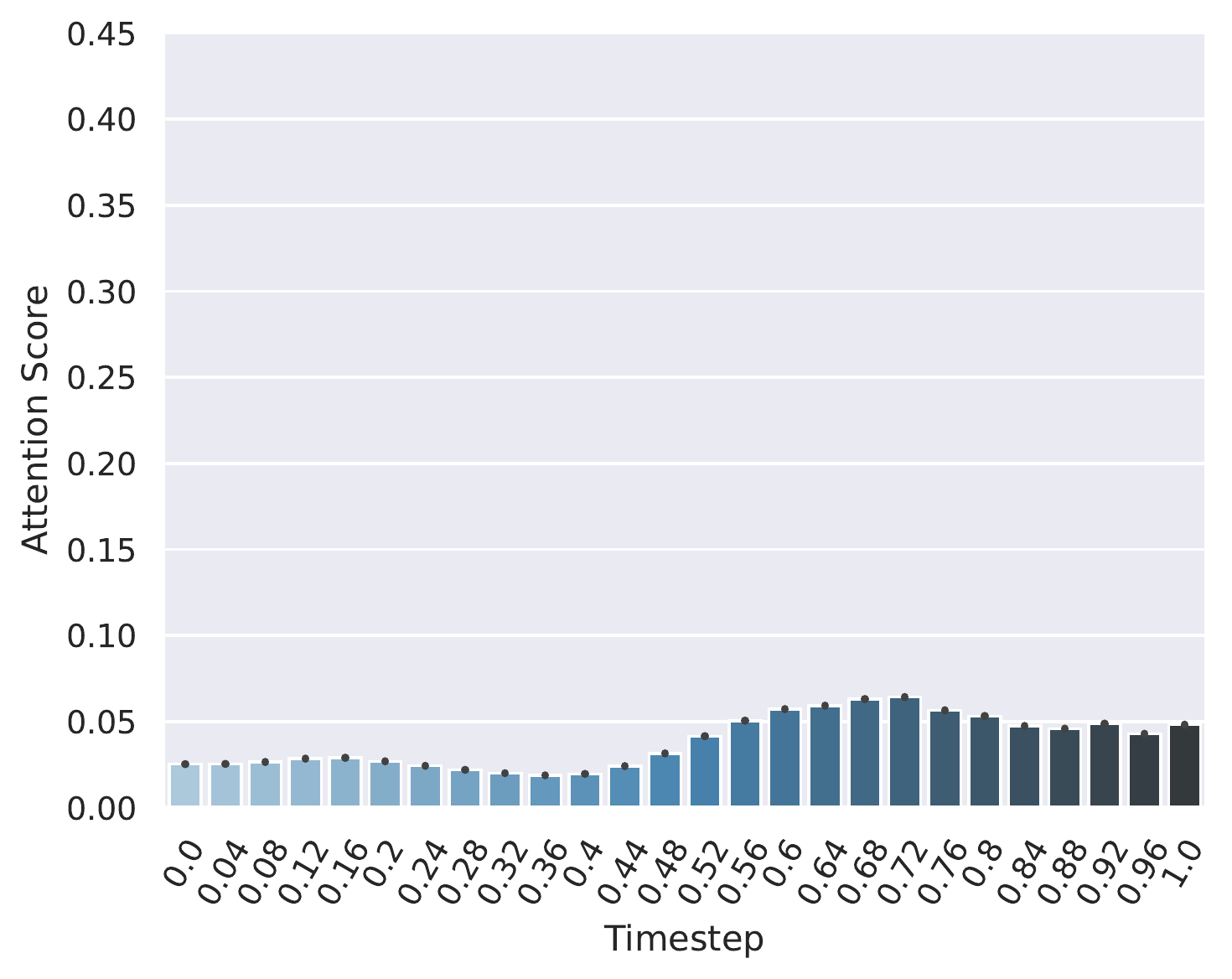}
\vspace{-6mm}
\subcaption*{\scriptsize Digit}
\end{subfigure}
\subcaption*{Granularity: 25}
\end{subfigure} \\
\caption{Attention score profiles for the Colored-MNIST dataset on the different features, using different granularities, with the dimensionality of the latent space as 16 and the VDRL encoder.}
\label{fig:cm_VDRL_16}
\end{figure}
\begin{figure}
\begin{subfigure}[c]{\textwidth}
\begin{subfigure}[c]{0.32\textwidth}
\includegraphics[width=\textwidth]{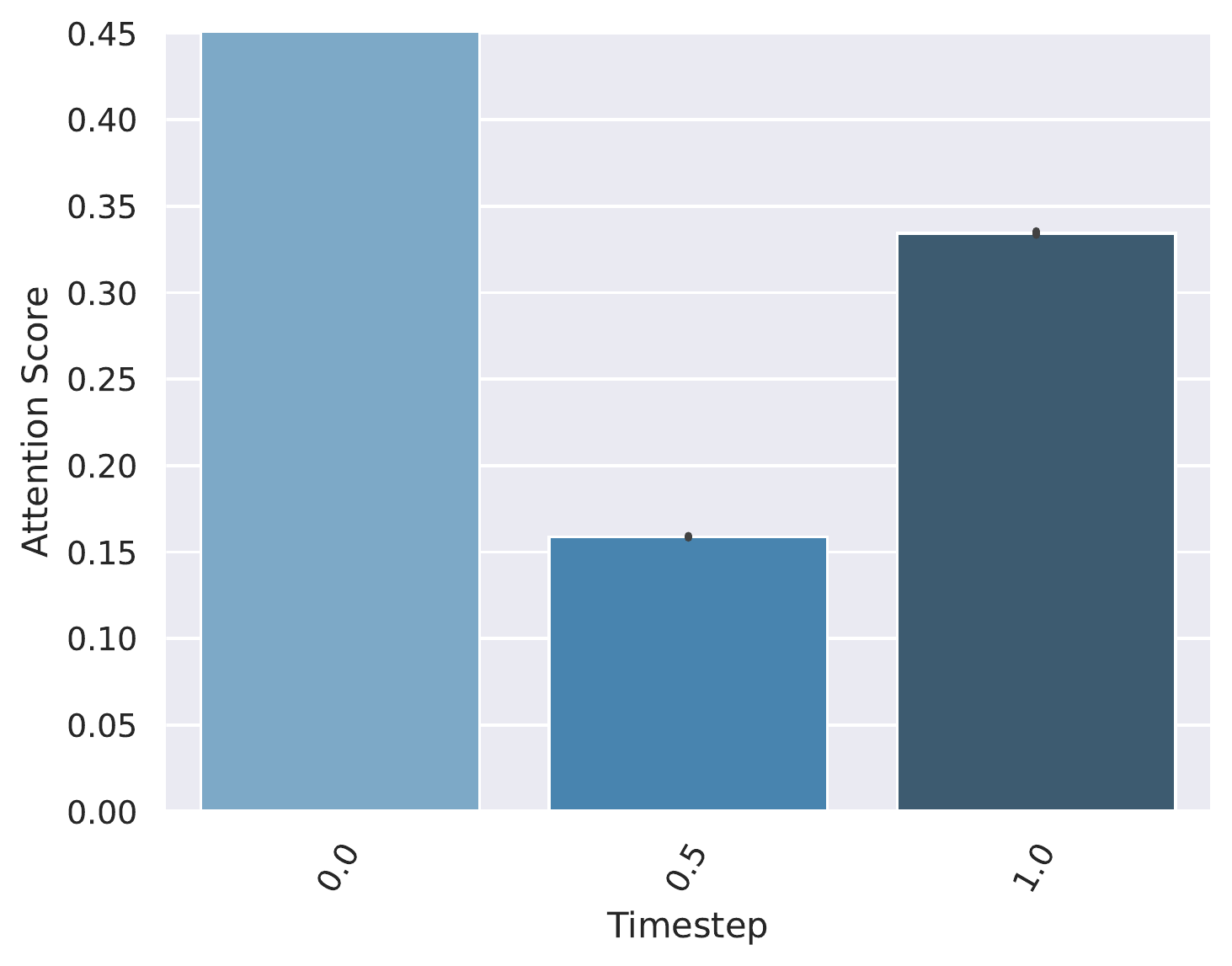}
\vspace{-6mm}
\subcaption*{\scriptsize Background Color}
\end{subfigure}
\begin{subfigure}[c]{0.32\textwidth}
\includegraphics[width=\textwidth]{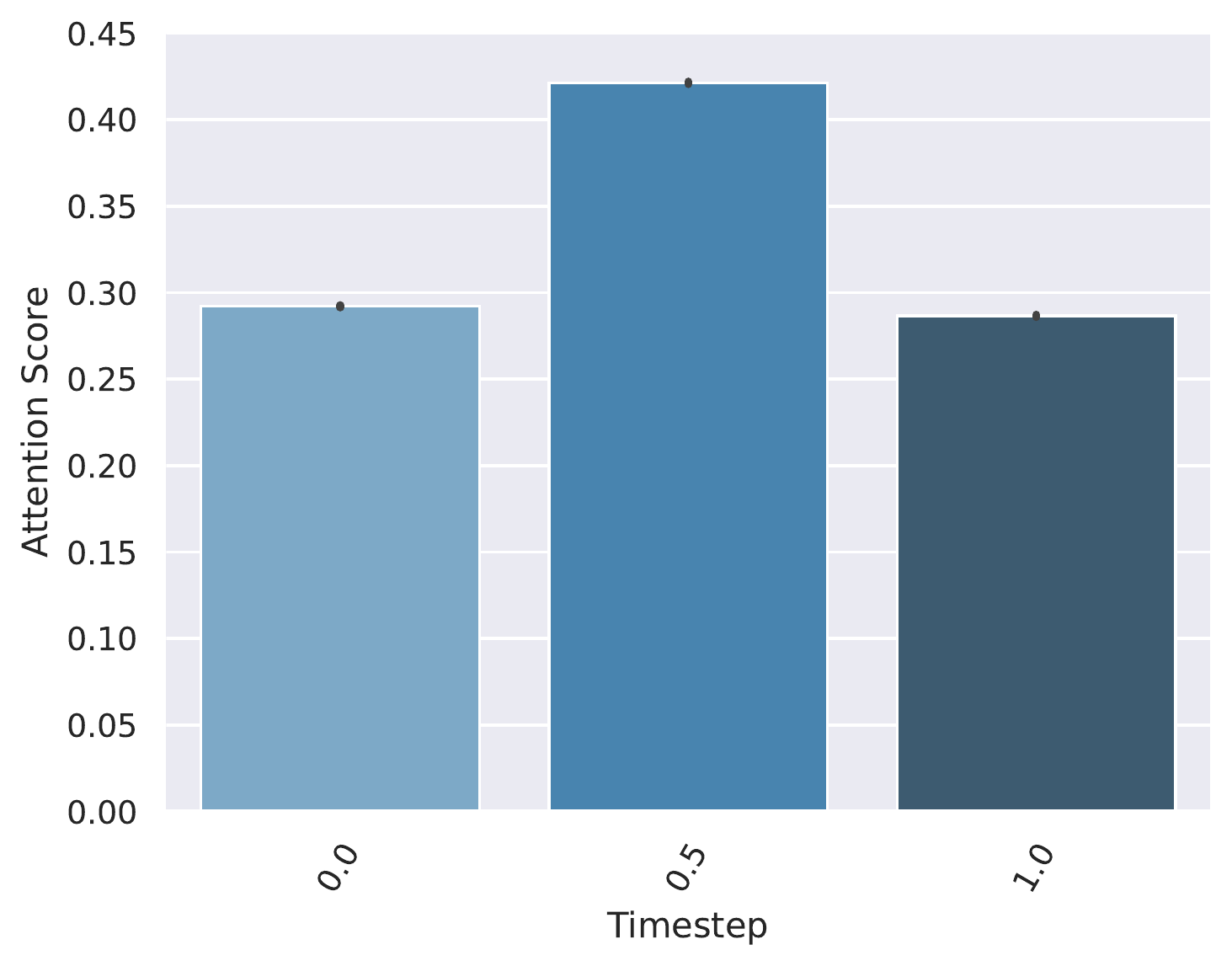}
\vspace{-6mm}
\subcaption*{\scriptsize Foreground Color}
\end{subfigure}
\begin{subfigure}[c]{0.32\textwidth}
\includegraphics[width=\textwidth]{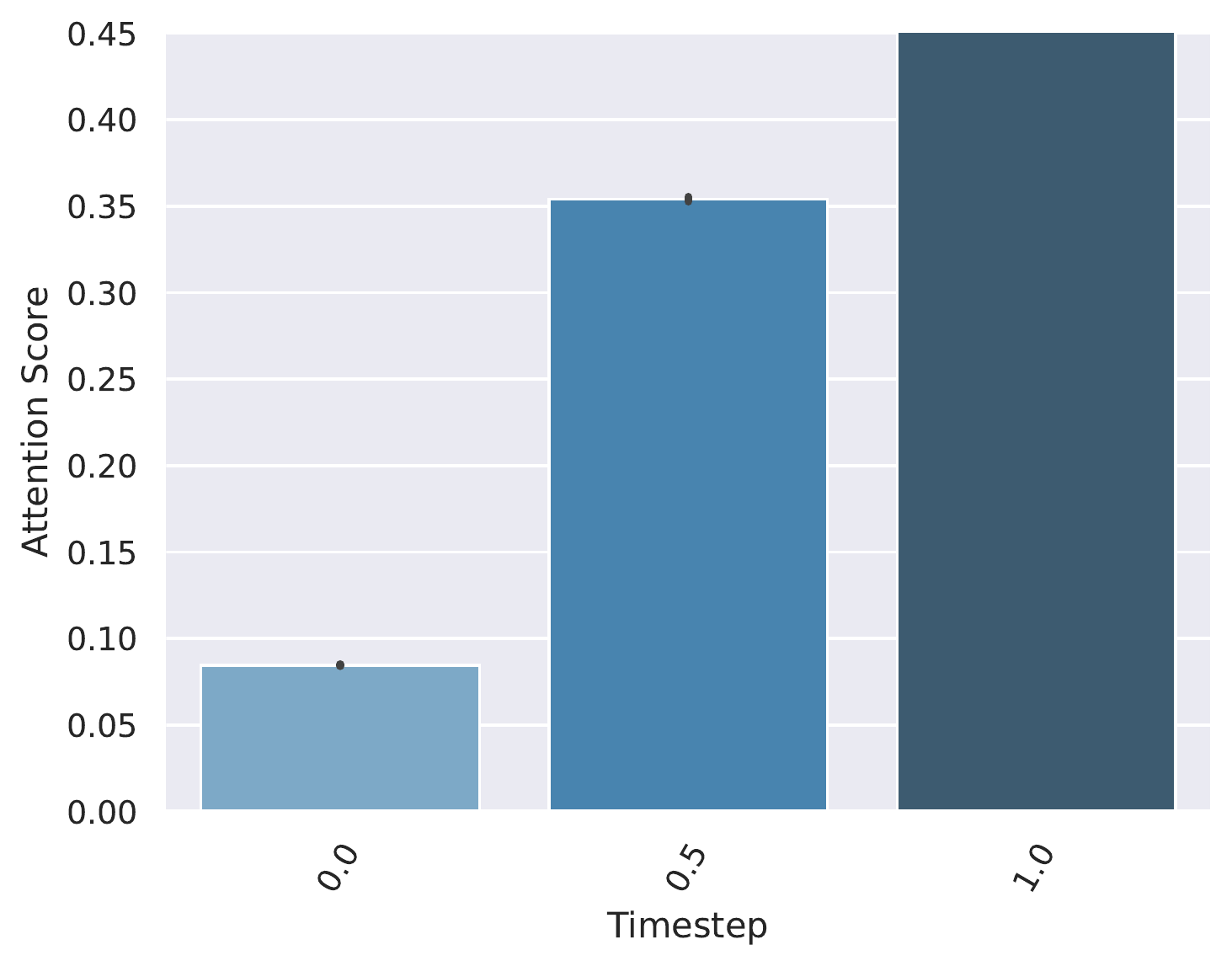}
\vspace{-6mm}
\subcaption*{\scriptsize Digit}
\end{subfigure}
\subcaption*{Granularity: 2}
\end{subfigure} \\
\begin{subfigure}[c]{\textwidth}
\begin{subfigure}[c]{0.32\textwidth}
\includegraphics[width=\textwidth]{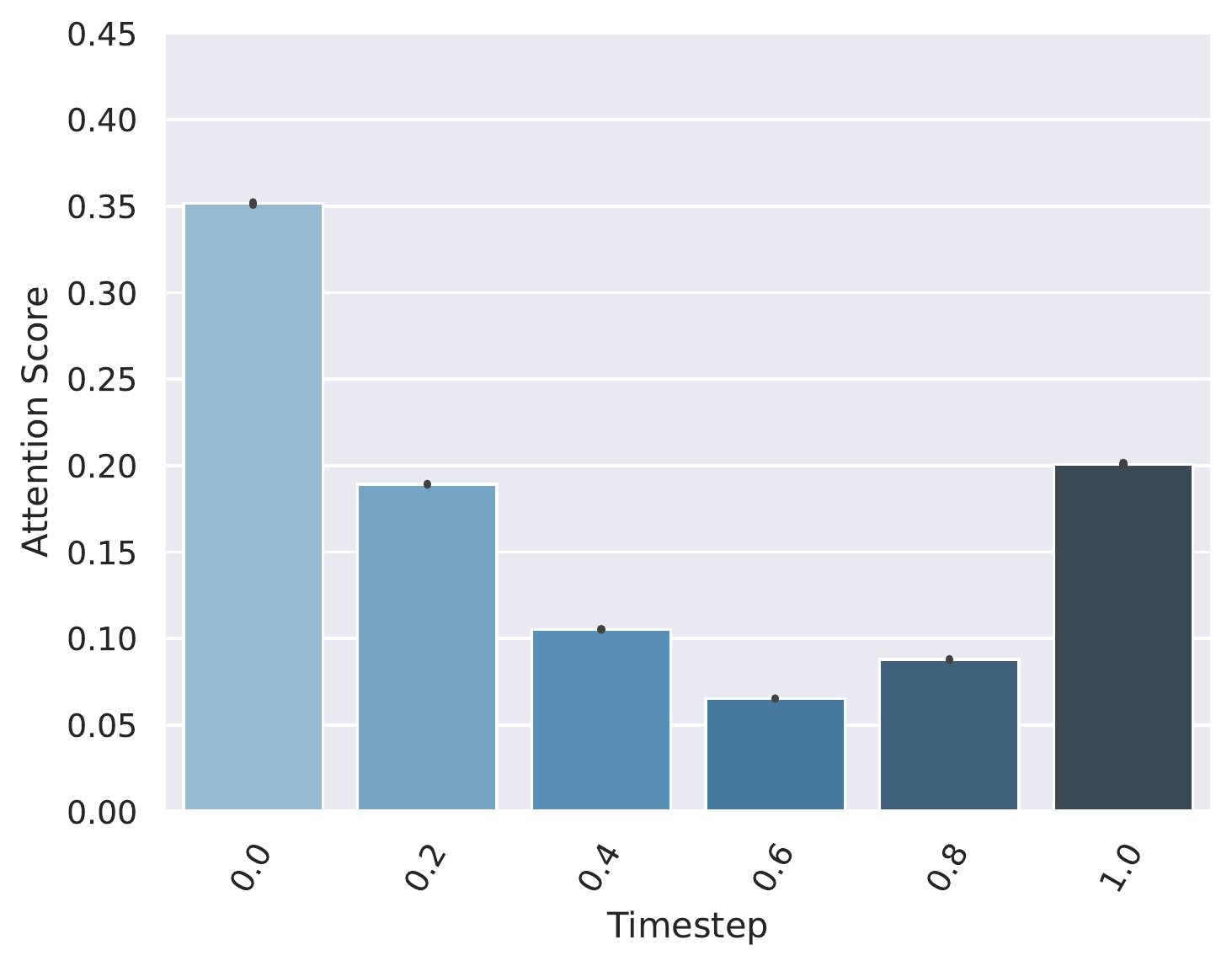}
\vspace{-6mm}
\subcaption*{\scriptsize Background Color}
\end{subfigure}
\begin{subfigure}[c]{0.32\textwidth}
\includegraphics[width=\textwidth]{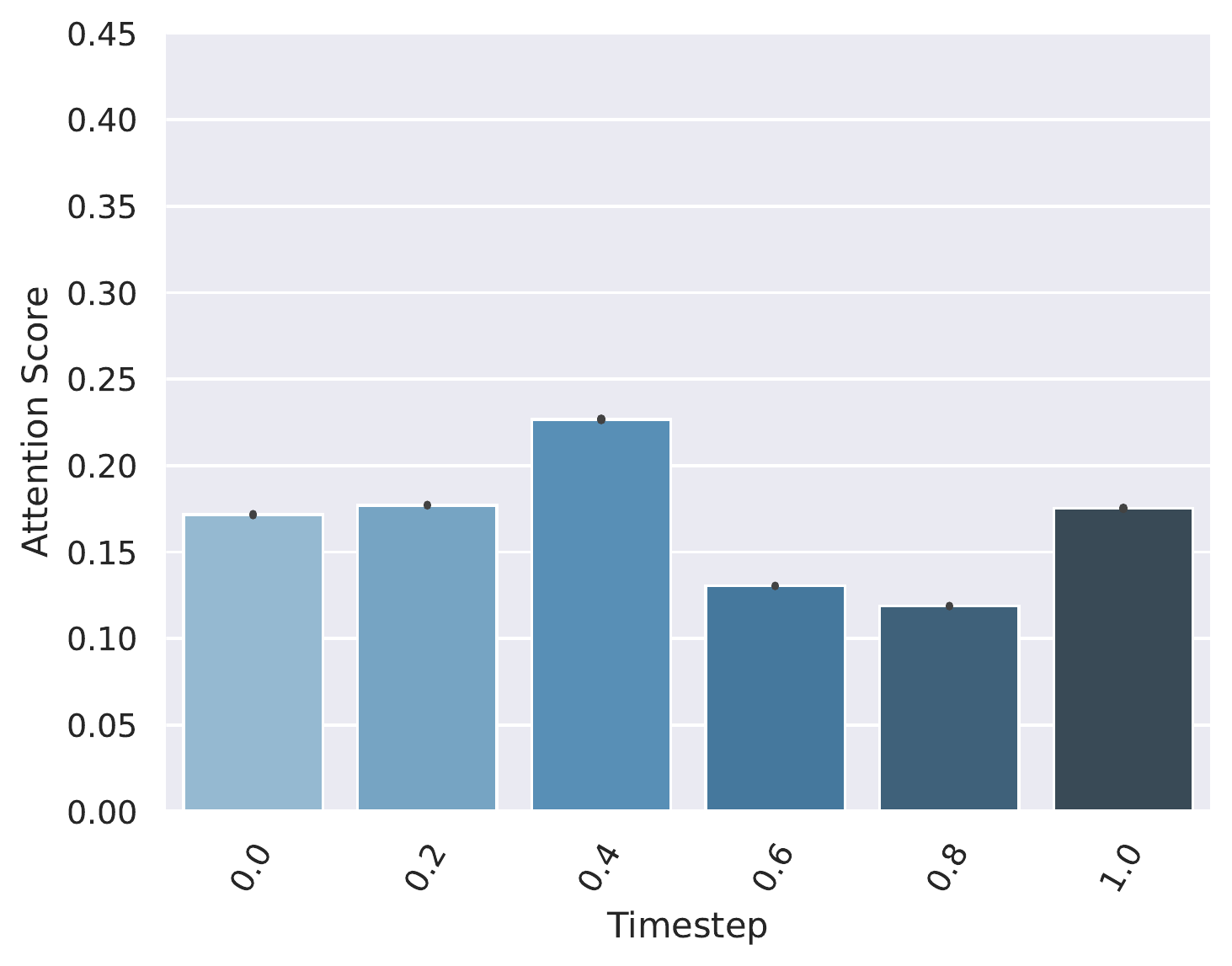}
\vspace{-6mm}
\subcaption*{\scriptsize Foreground Color}
\end{subfigure}
\begin{subfigure}[c]{0.32\textwidth}
\includegraphics[width=\textwidth]{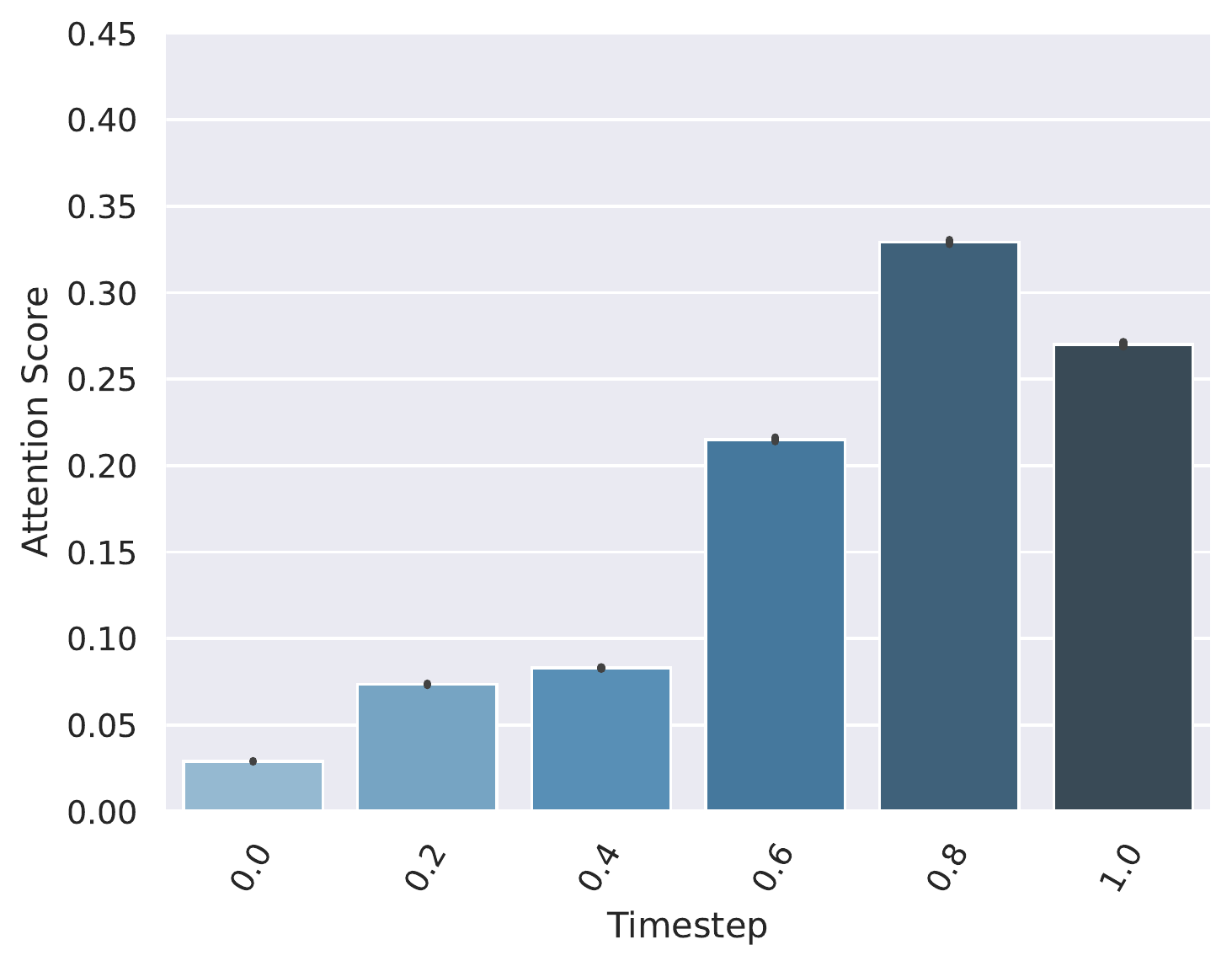}
\vspace{-6mm}
\subcaption*{\scriptsize Digit}
\end{subfigure}
\subcaption*{Granularity: 5}
\end{subfigure} \\
\begin{subfigure}[c]{\textwidth}
\begin{subfigure}[c]{0.32\textwidth}
\includegraphics[width=\textwidth]{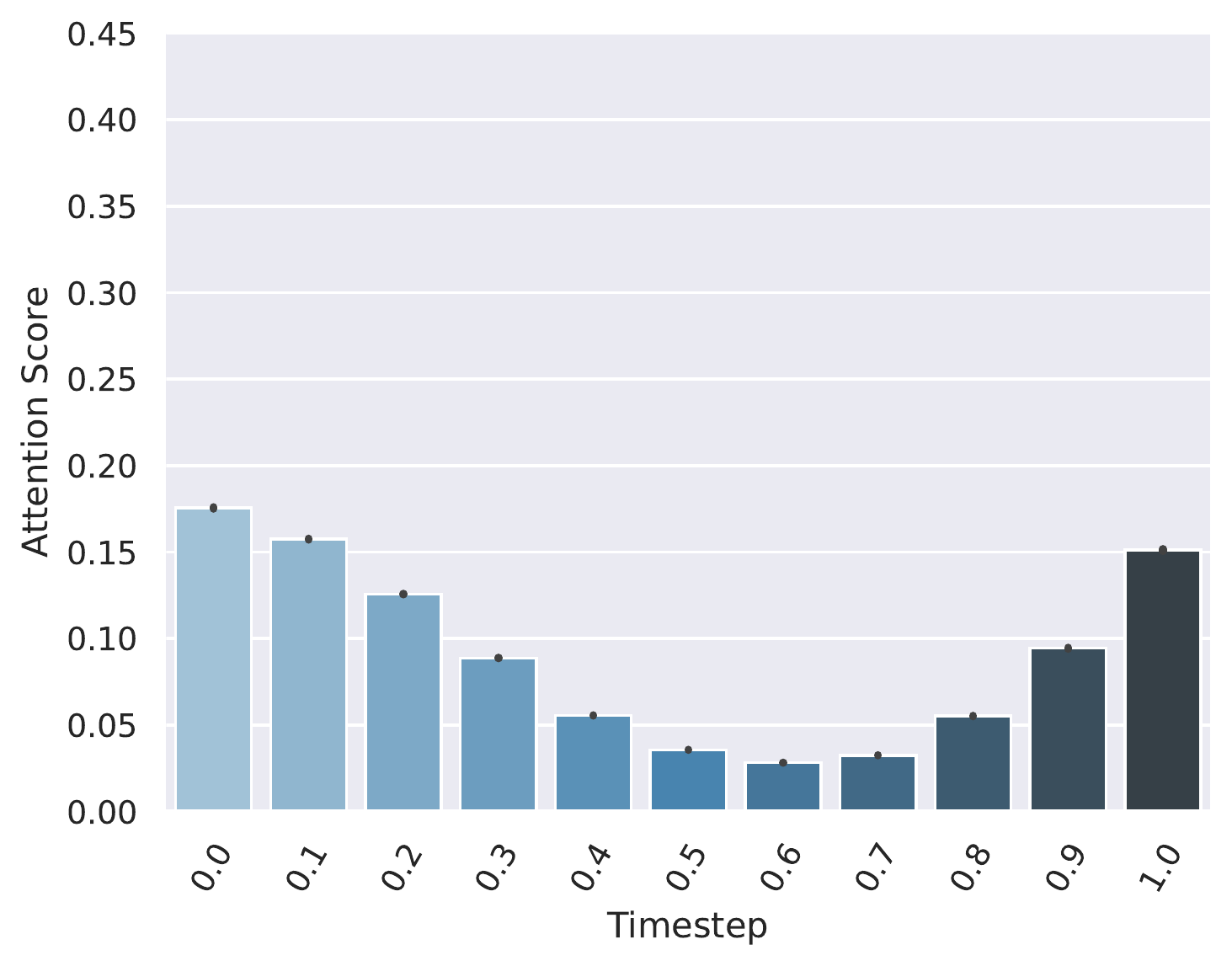}
\vspace{-6mm}
\subcaption*{\scriptsize Background Color}
\end{subfigure}
\begin{subfigure}[c]{0.32\textwidth}
\includegraphics[width=\textwidth]{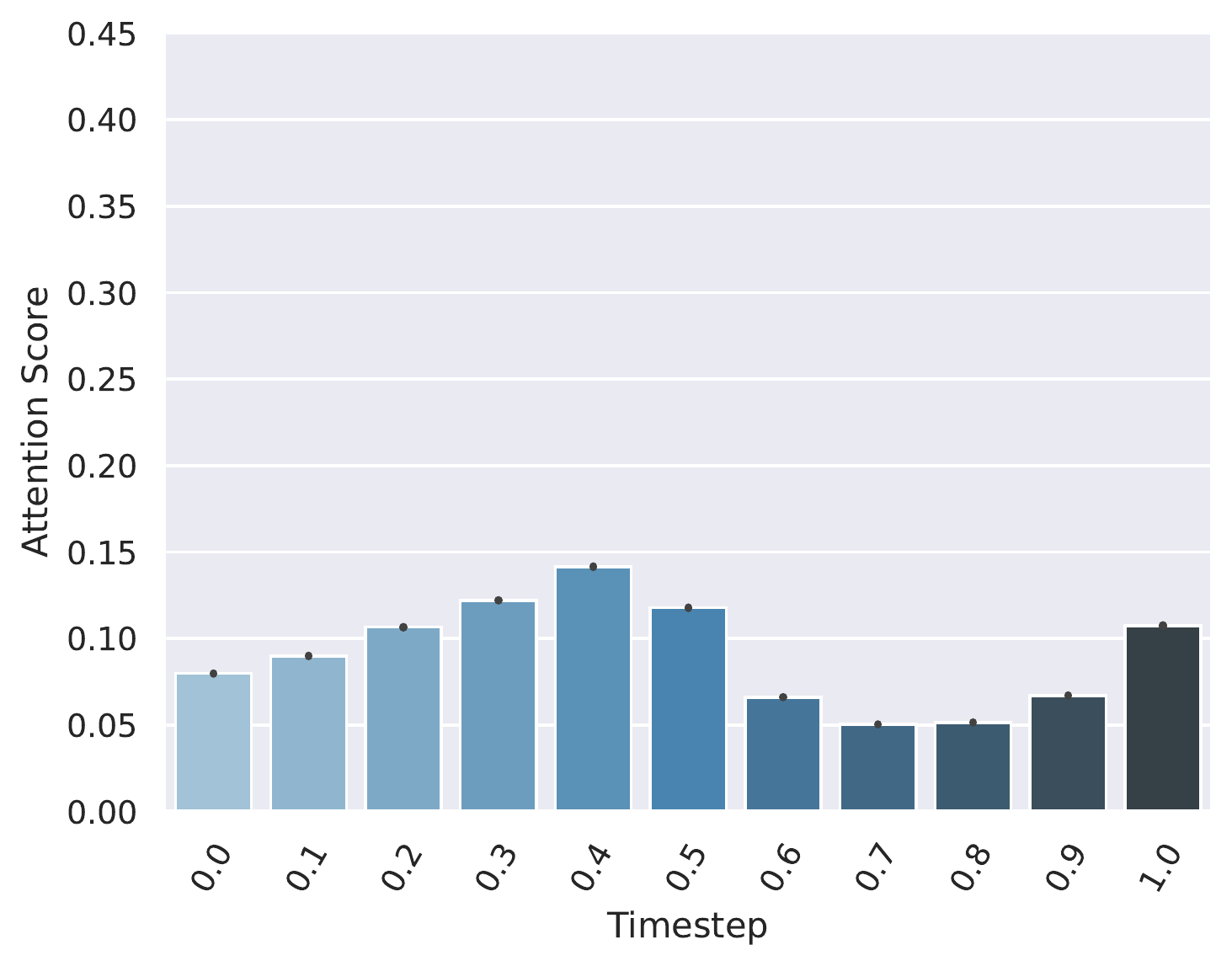}
\vspace{-6mm}
\subcaption*{\scriptsize Foreground Color}
\end{subfigure}
\begin{subfigure}[c]{0.32\textwidth}
\includegraphics[width=\textwidth]{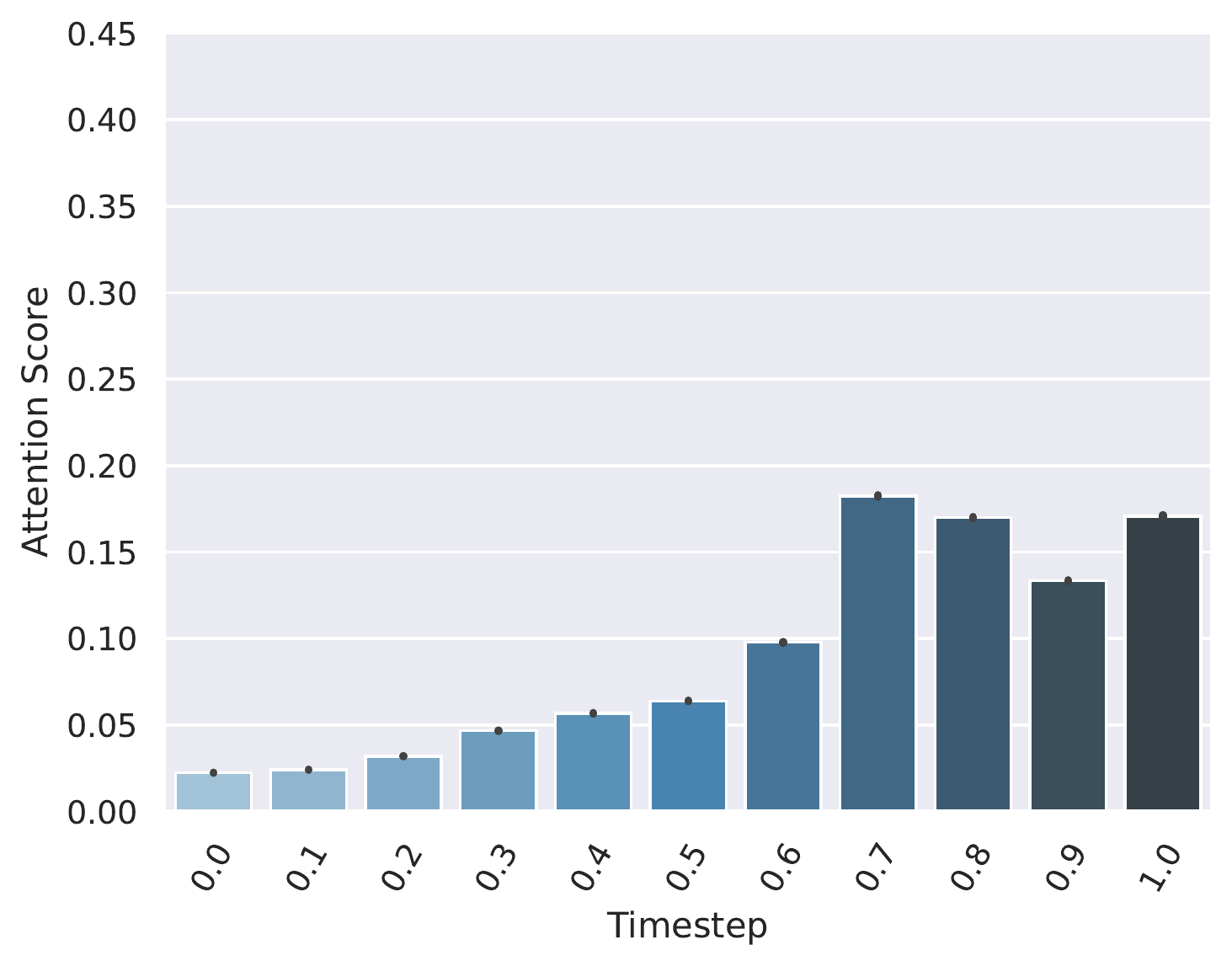}
\vspace{-6mm}
\subcaption*{\scriptsize Digit}
\end{subfigure}
\subcaption*{Granularity: 10}
\end{subfigure} \\
\begin{subfigure}[c]{\textwidth}
\begin{subfigure}[c]{0.32\textwidth}
\includegraphics[width=\textwidth]{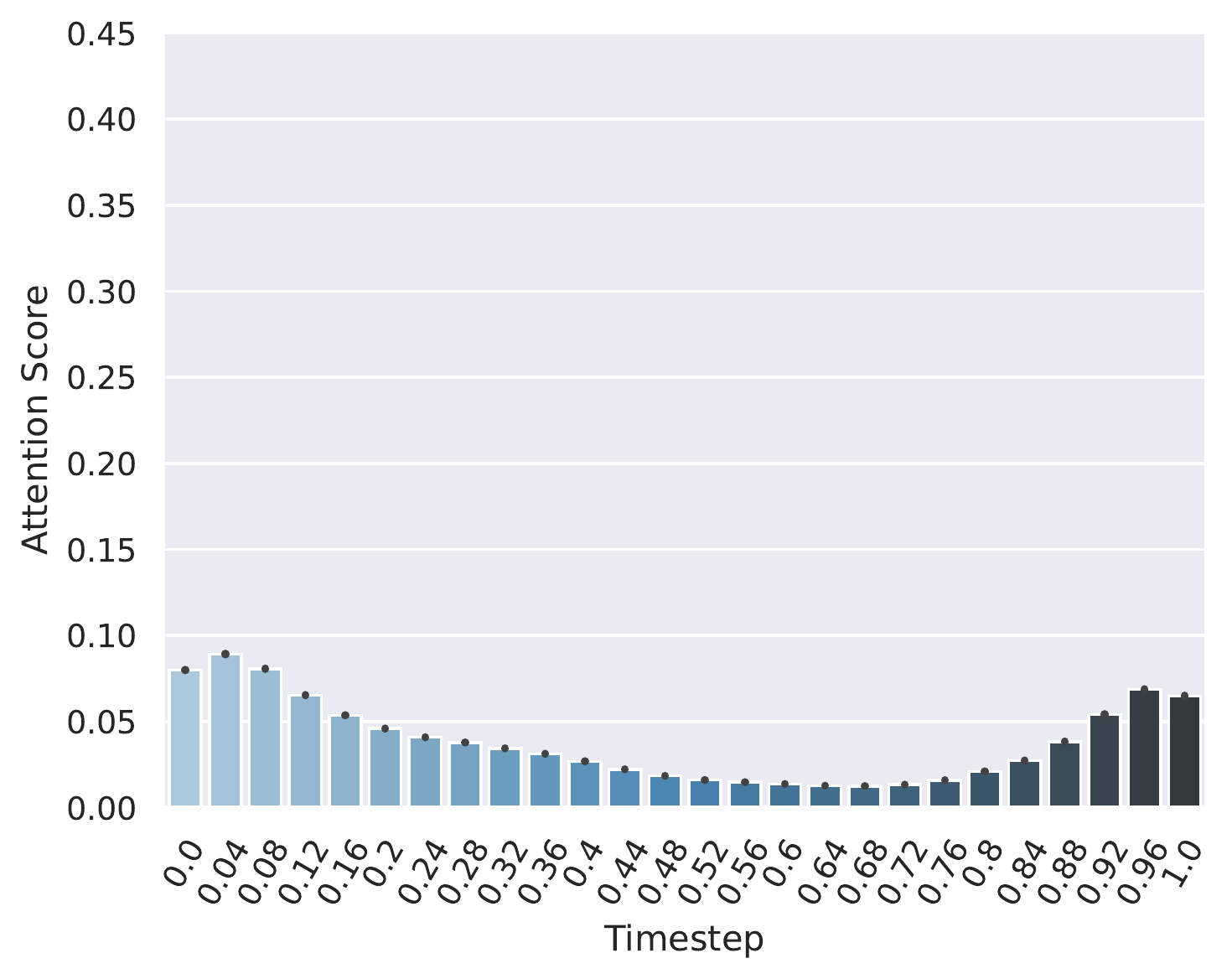}
\vspace{-6mm}
\subcaption*{\scriptsize Background Color}
\end{subfigure}
\begin{subfigure}[c]{0.32\textwidth}
\includegraphics[width=\textwidth]{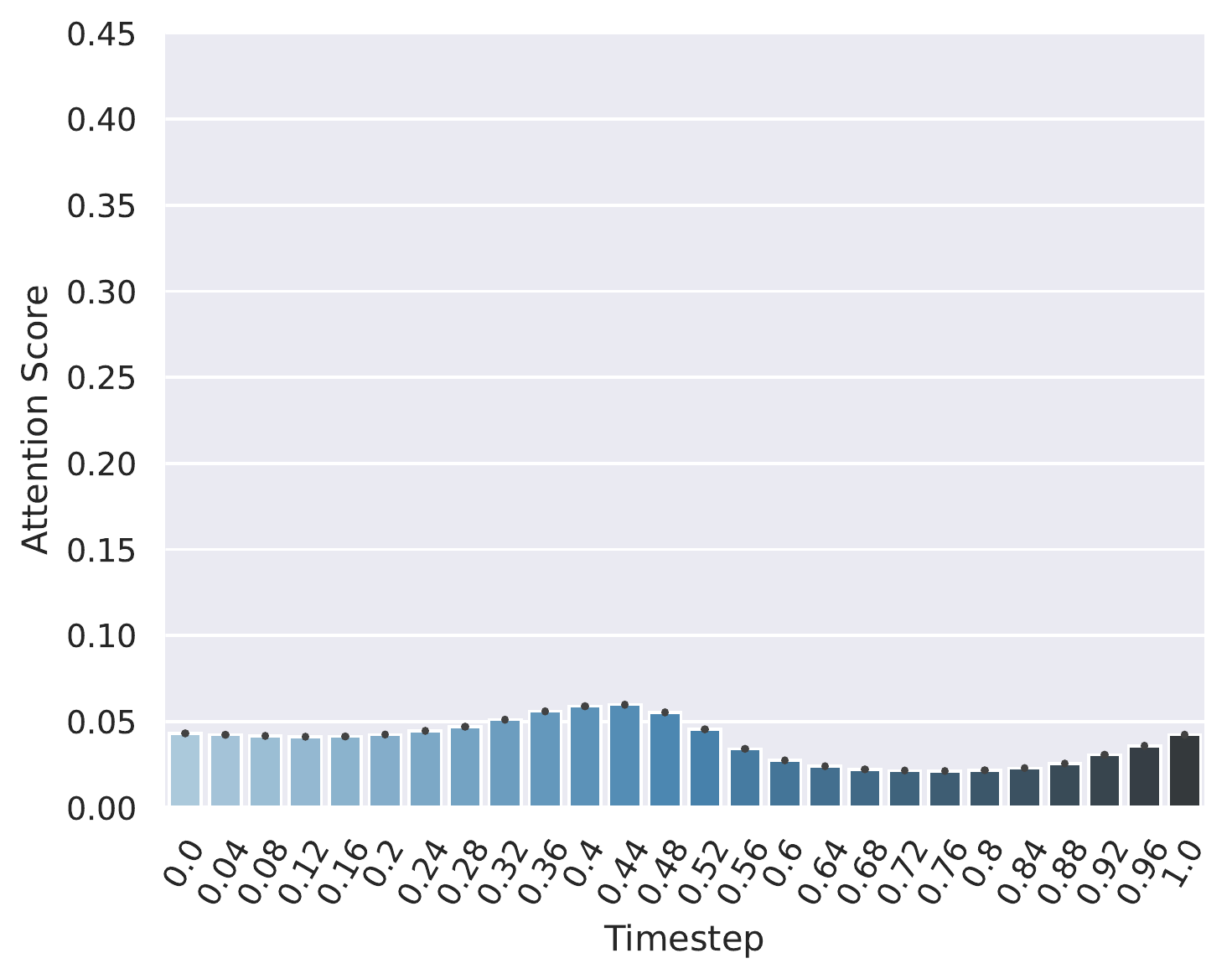}
\vspace{-6mm}
\subcaption*{\scriptsize Foreground Color}
\end{subfigure}
\begin{subfigure}[c]{0.32\textwidth}
\includegraphics[width=\textwidth]{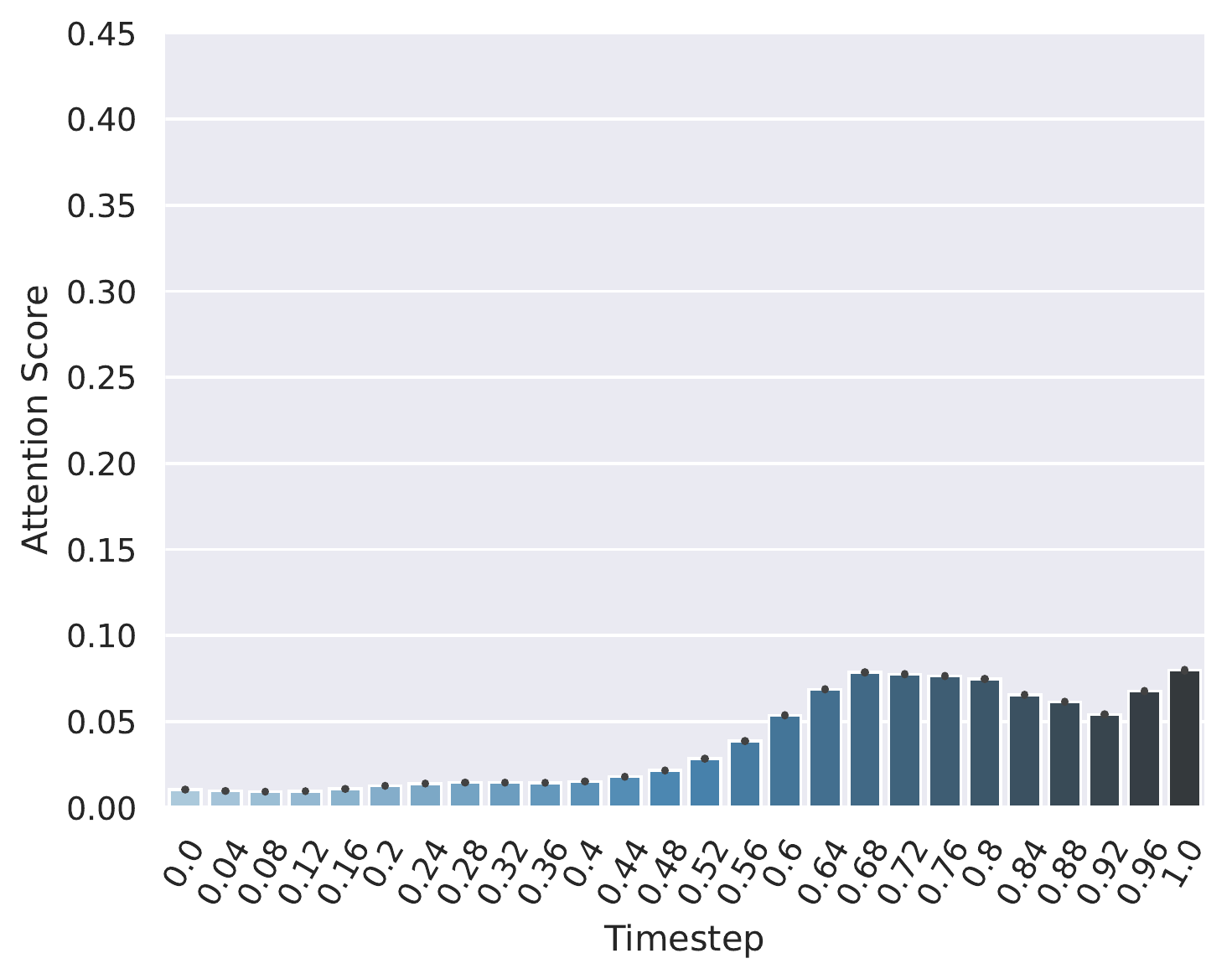}
\vspace{-6mm}
\subcaption*{\scriptsize Digit}
\end{subfigure}
\subcaption*{Granularity: 25}
\end{subfigure} \\
\caption{Attention score profiles for the Colored-MNIST dataset on the different features, using different granularities, with the dimensionality of the latent space as 32 and the VDRL encoder.}
\label{fig:cm_VDRL_32}
\end{figure}
\begin{figure}
\begin{subfigure}[c]{\textwidth}
\begin{subfigure}[c]{0.32\textwidth}
\includegraphics[width=\textwidth]{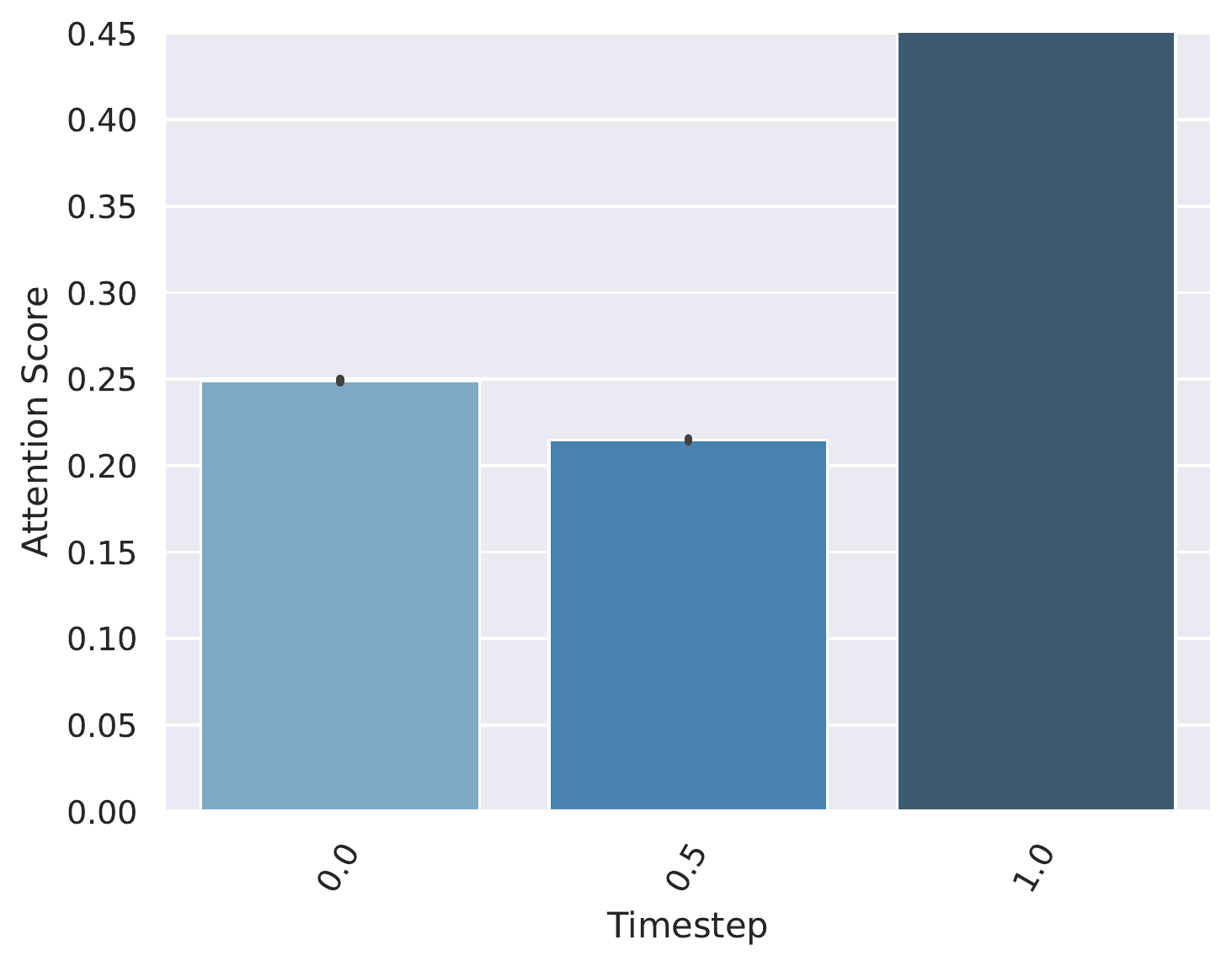}
\vspace{-6mm}
\subcaption*{\scriptsize Background Color}
\end{subfigure}
\begin{subfigure}[c]{0.32\textwidth}
\includegraphics[width=\textwidth]{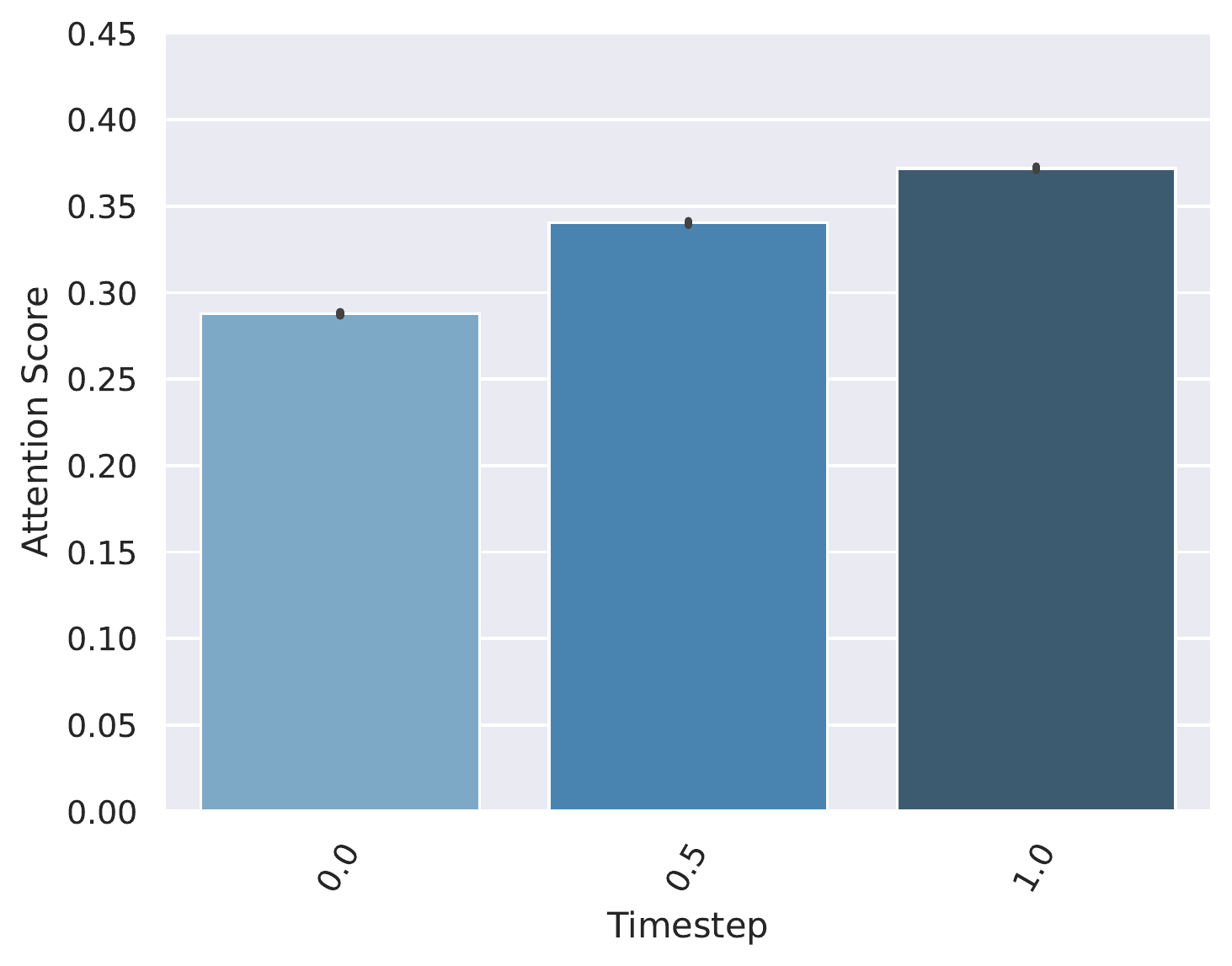}
\vspace{-6mm}
\subcaption*{\scriptsize Foreground Color}
\end{subfigure}
\begin{subfigure}[c]{0.32\textwidth}
\includegraphics[width=\textwidth]{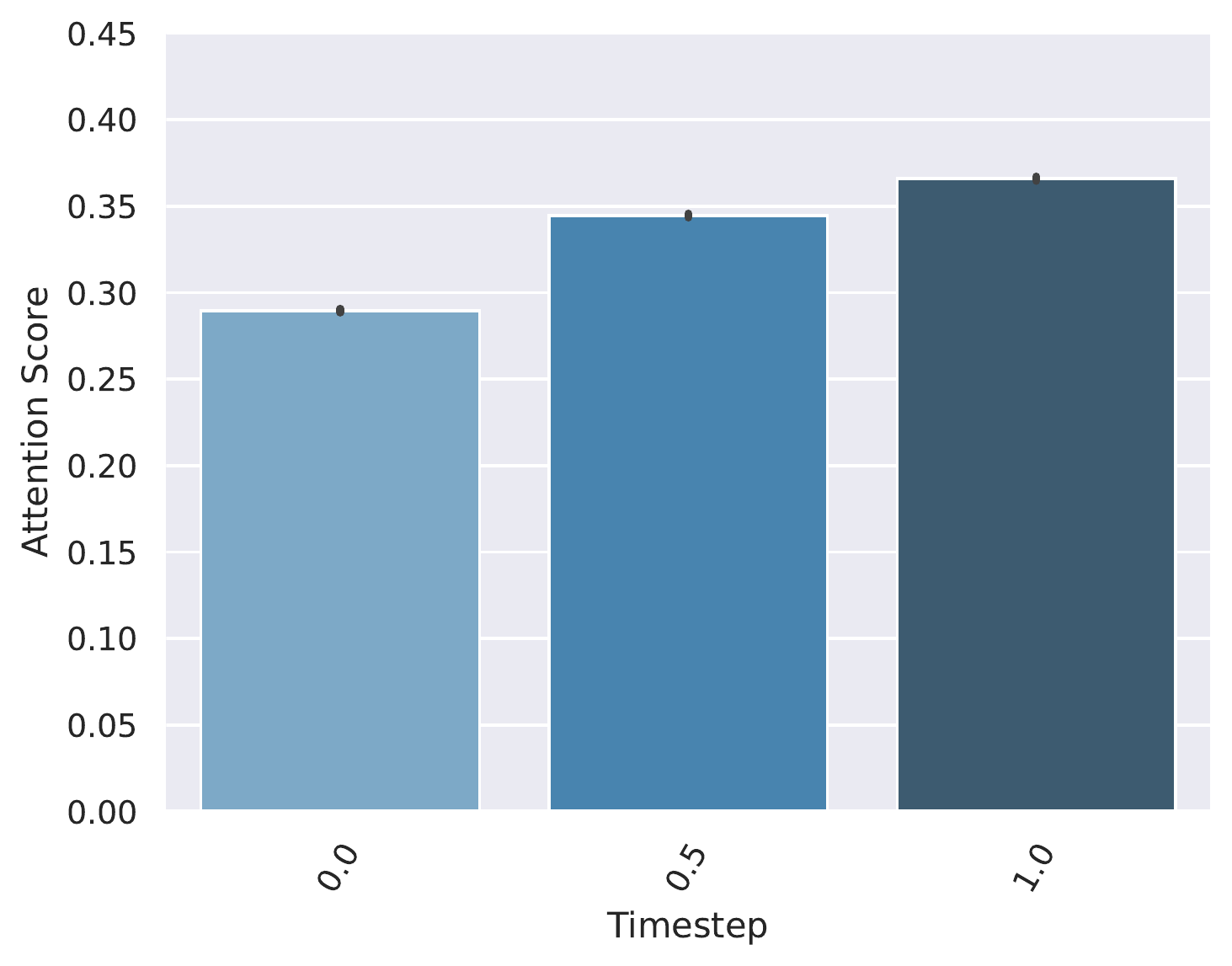}
\vspace{-6mm}
\subcaption*{\scriptsize Digit}
\end{subfigure}
\subcaption*{Granularity: 2}
\end{subfigure} \\
\begin{subfigure}[c]{\textwidth}
\begin{subfigure}[c]{0.32\textwidth}
\includegraphics[width=\textwidth]{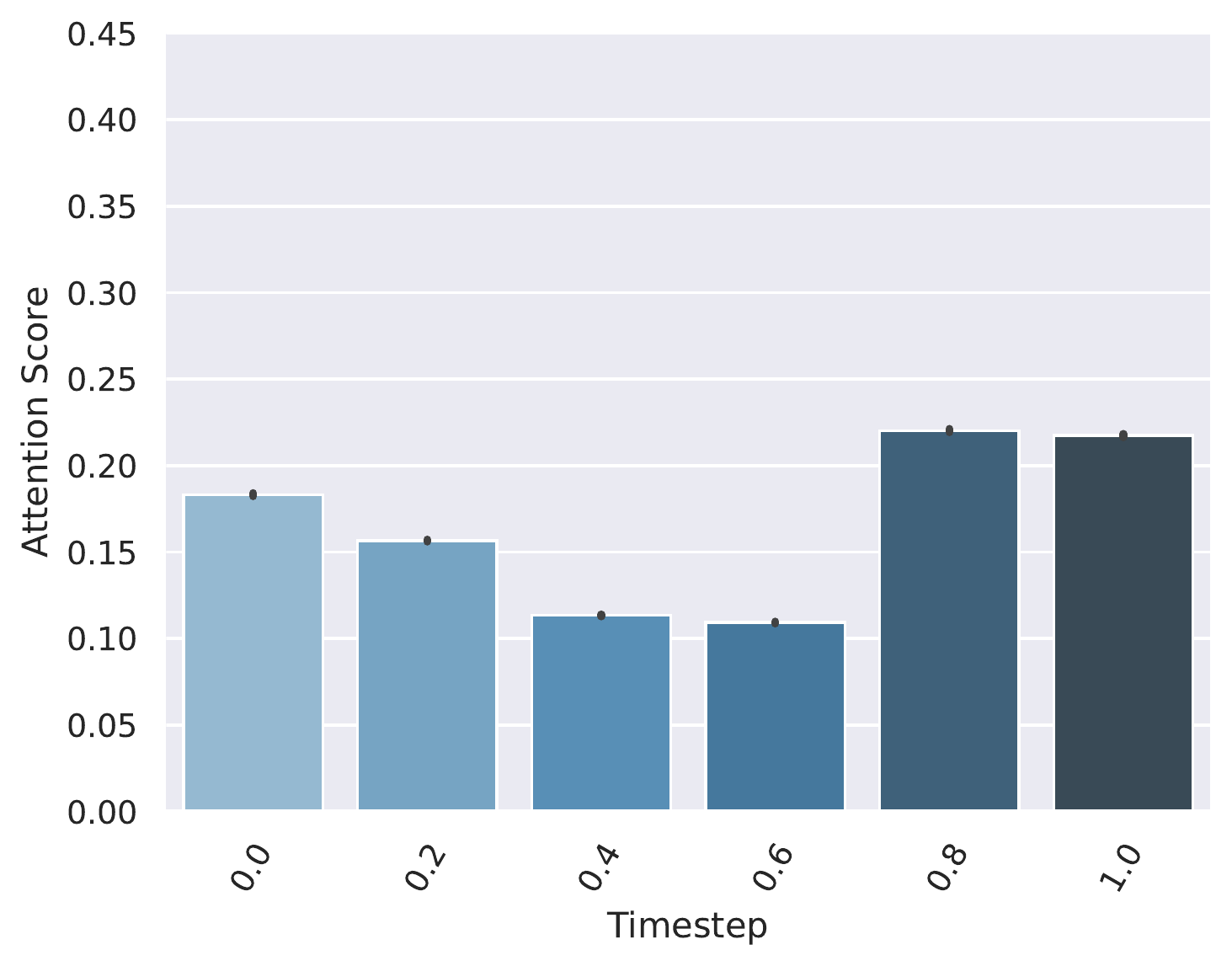}
\vspace{-6mm}
\subcaption*{\scriptsize Background Color}
\end{subfigure}
\begin{subfigure}[c]{0.32\textwidth}
\includegraphics[width=\textwidth]{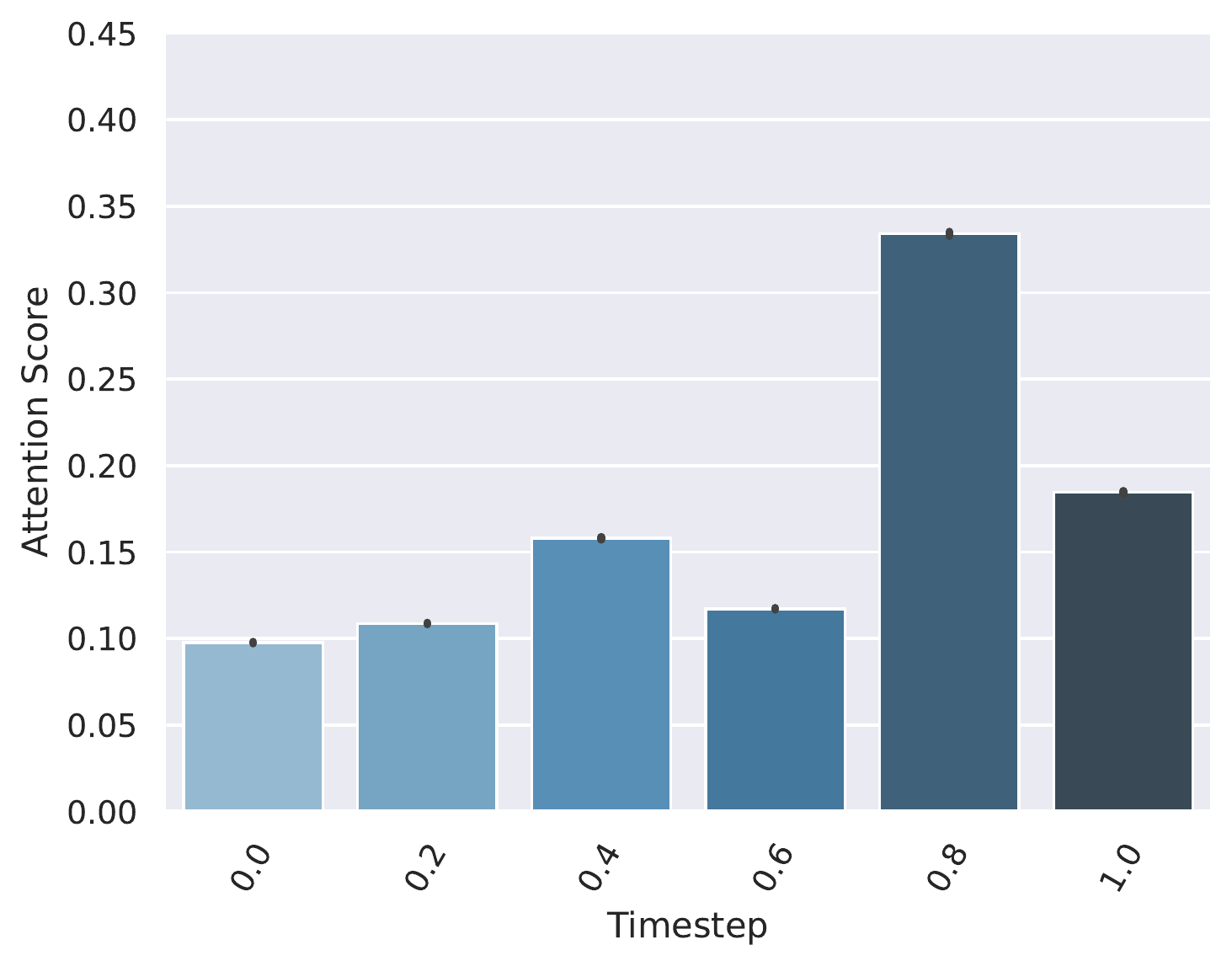}
\vspace{-6mm}
\subcaption*{\scriptsize Foreground Color}
\end{subfigure}
\begin{subfigure}[c]{0.32\textwidth}
\includegraphics[width=\textwidth]{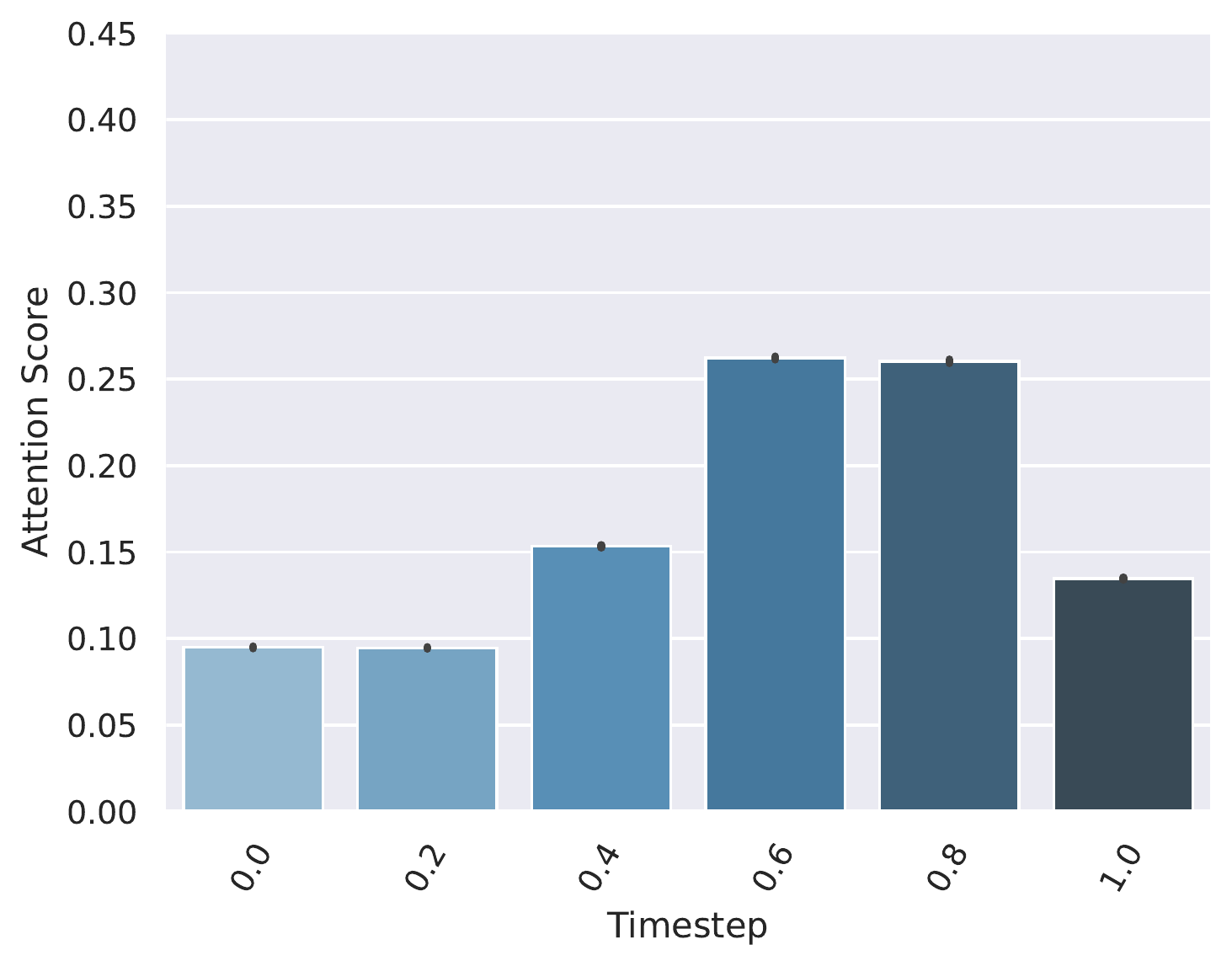}
\vspace{-6mm}
\subcaption*{\scriptsize Digit}
\end{subfigure}
\subcaption*{Granularity: 5}
\end{subfigure} \\
\begin{subfigure}[c]{\textwidth}
\begin{subfigure}[c]{0.32\textwidth}
\includegraphics[width=\textwidth]{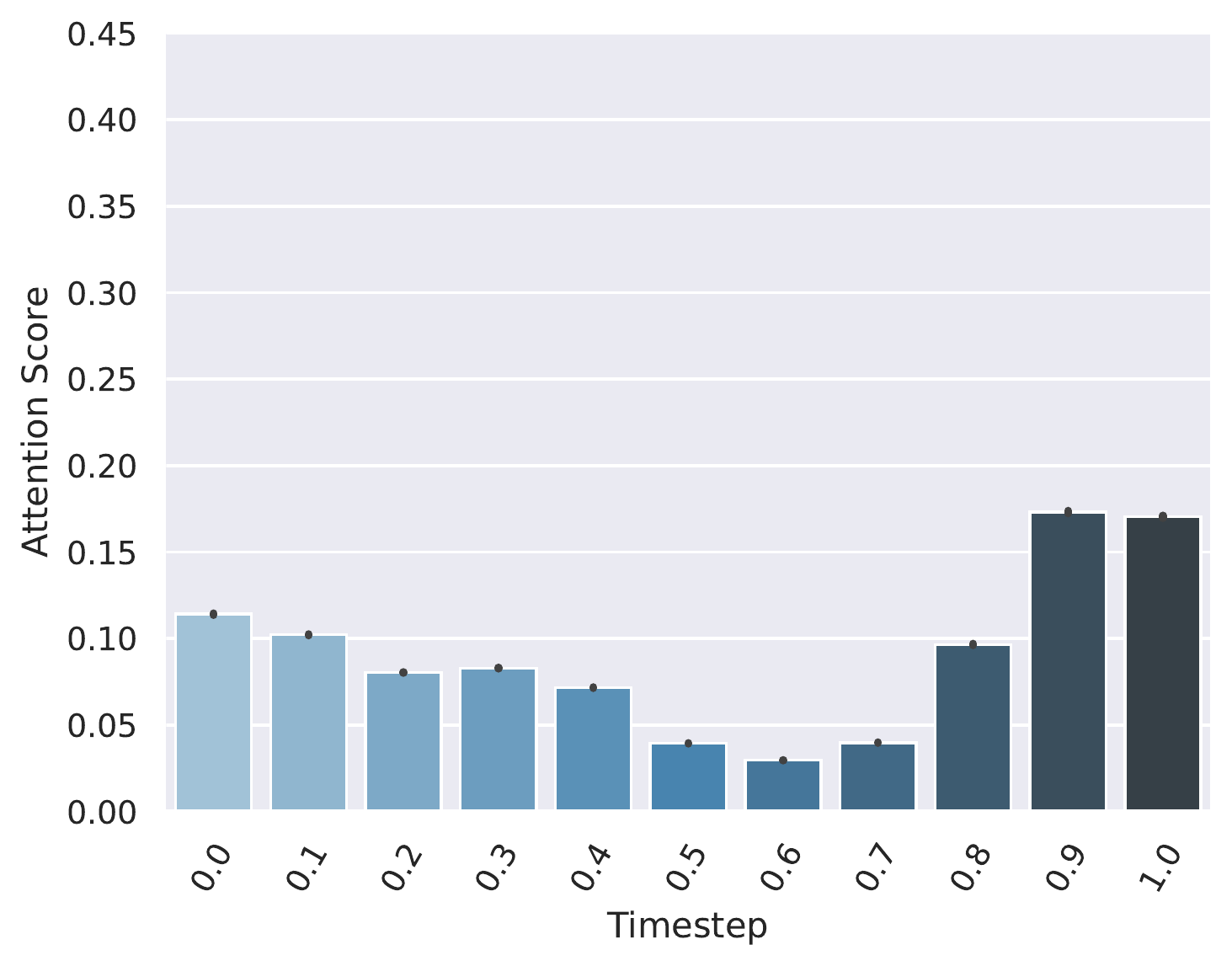}
\vspace{-6mm}
\subcaption*{\scriptsize Background Color}
\end{subfigure}
\begin{subfigure}[c]{0.32\textwidth}
\includegraphics[width=\textwidth]{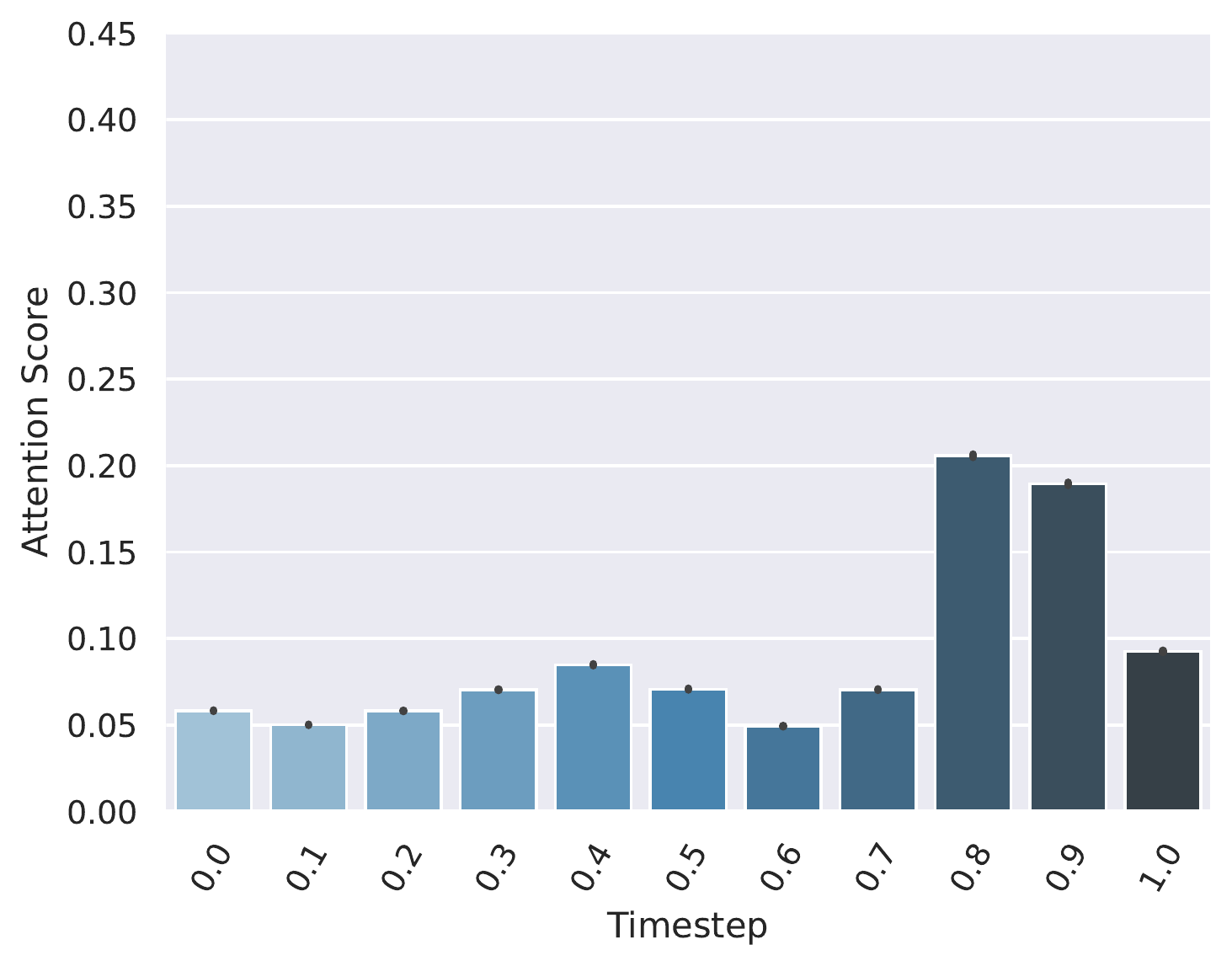}
\vspace{-6mm}
\subcaption*{\scriptsize Foreground Color}
\end{subfigure}
\begin{subfigure}[c]{0.32\textwidth}
\includegraphics[width=\textwidth]{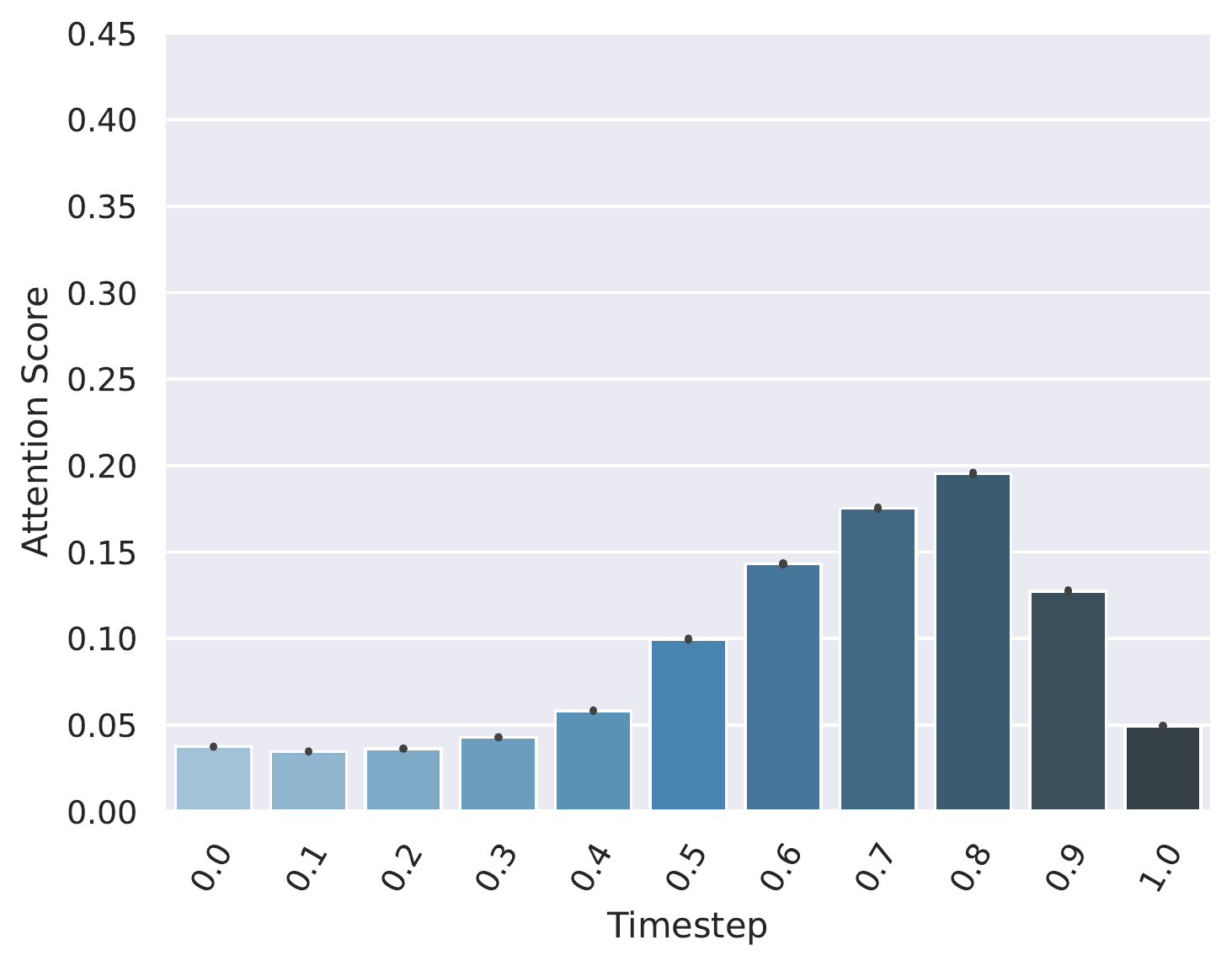}
\vspace{-6mm}
\subcaption*{\scriptsize Digit}
\end{subfigure}
\subcaption*{Granularity: 10}
\end{subfigure} \\
\begin{subfigure}[c]{\textwidth}
\begin{subfigure}[c]{0.32\textwidth}
\includegraphics[width=\textwidth]{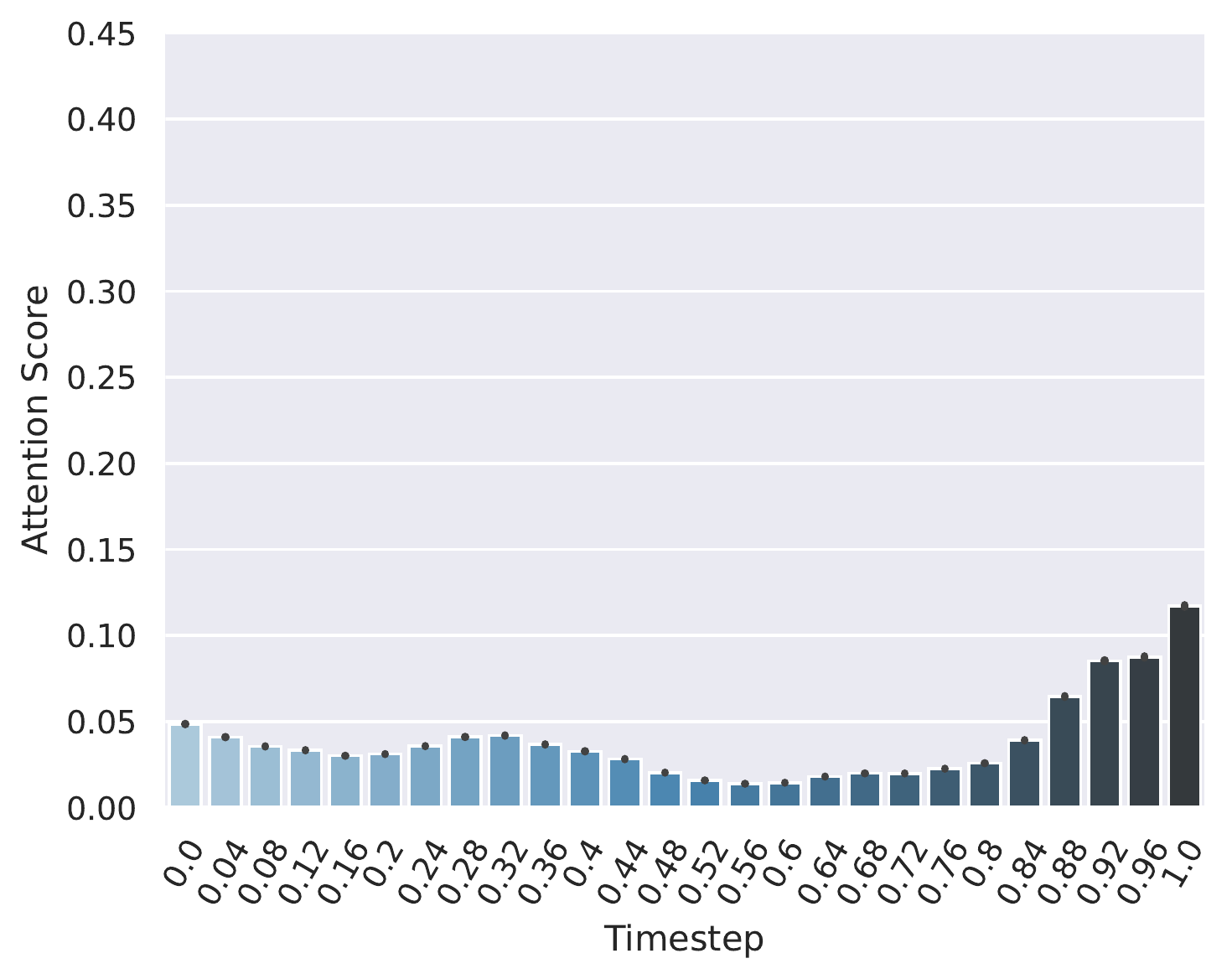}
\vspace{-6mm}
\subcaption*{\scriptsize Background Color}
\end{subfigure}
\begin{subfigure}[c]{0.32\textwidth}
\includegraphics[width=\textwidth]{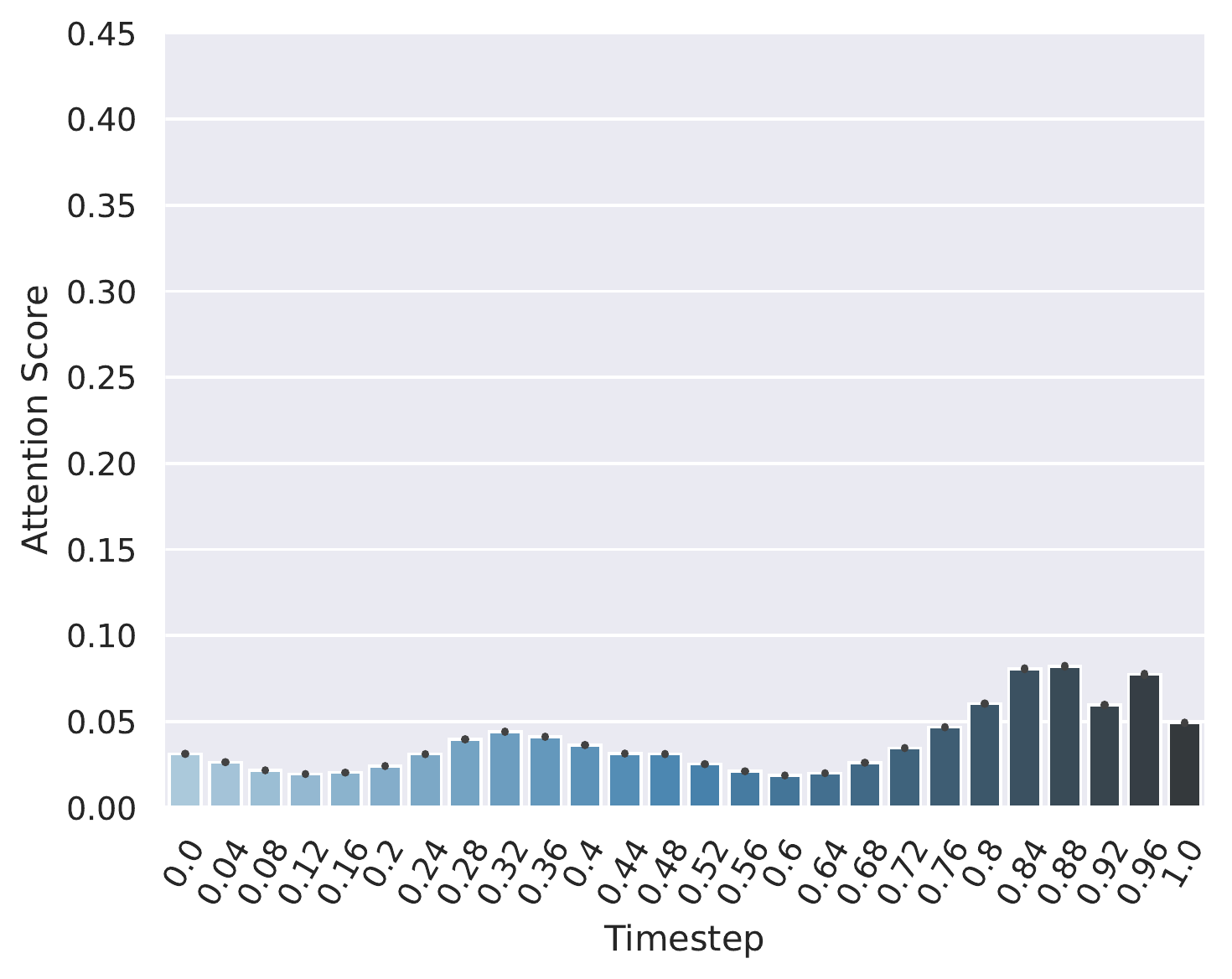}
\vspace{-6mm}
\subcaption*{\scriptsize Foreground Color}
\end{subfigure}
\begin{subfigure}[c]{0.32\textwidth}
\includegraphics[width=\textwidth]{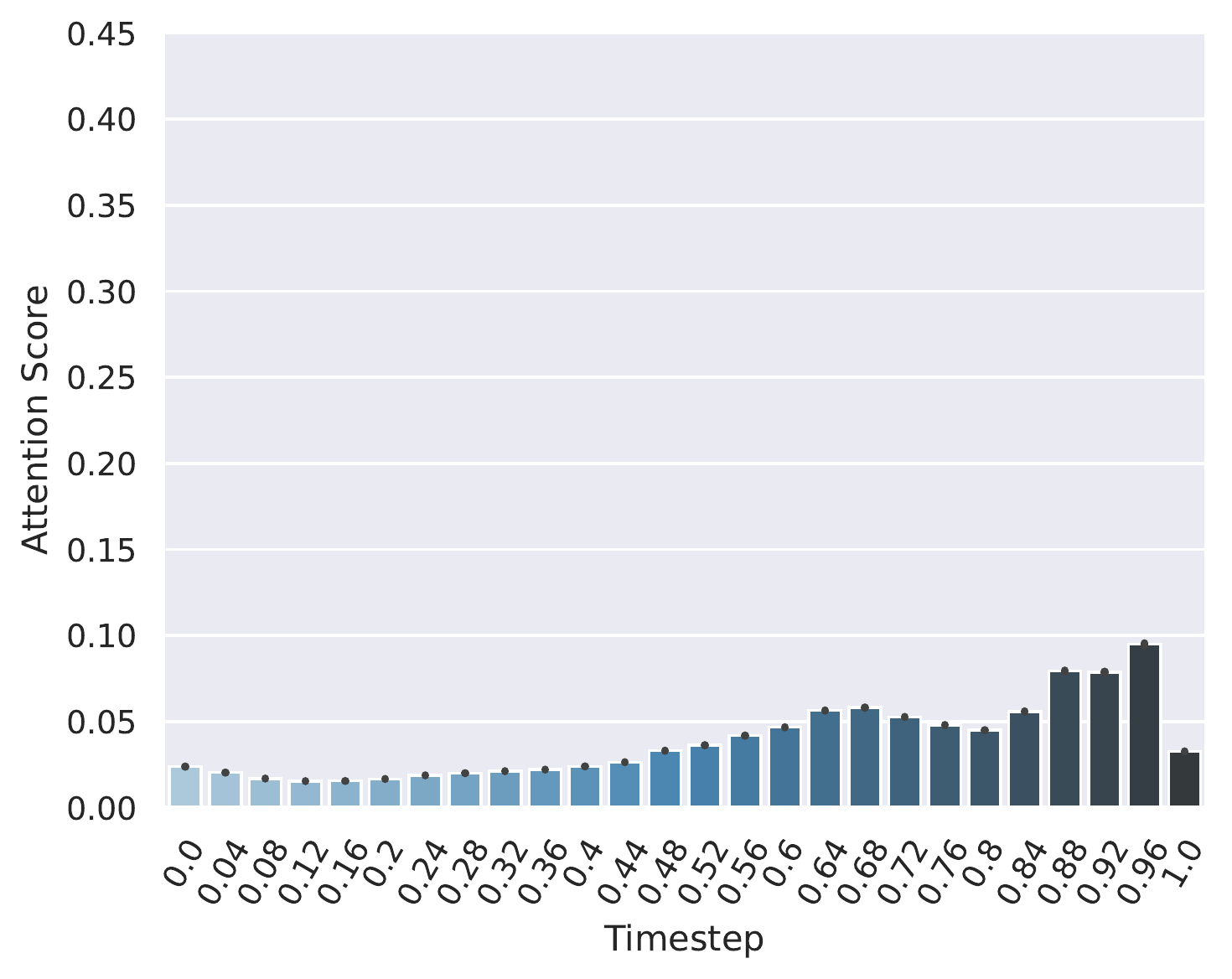}
\vspace{-6mm}
\subcaption*{\scriptsize Digit}
\end{subfigure}
\subcaption*{Granularity: 25}
\end{subfigure} \\
\caption{Attention score profiles for the Colored-MNIST dataset on the different features, using different granularities, with the dimensionality of the latent space as 2 and the DRL encoder.}
\label{fig:cm_DRL_2}
\end{figure}
\begin{figure}
\begin{subfigure}[c]{\textwidth}
\begin{subfigure}[c]{0.32\textwidth}
\includegraphics[width=\textwidth]{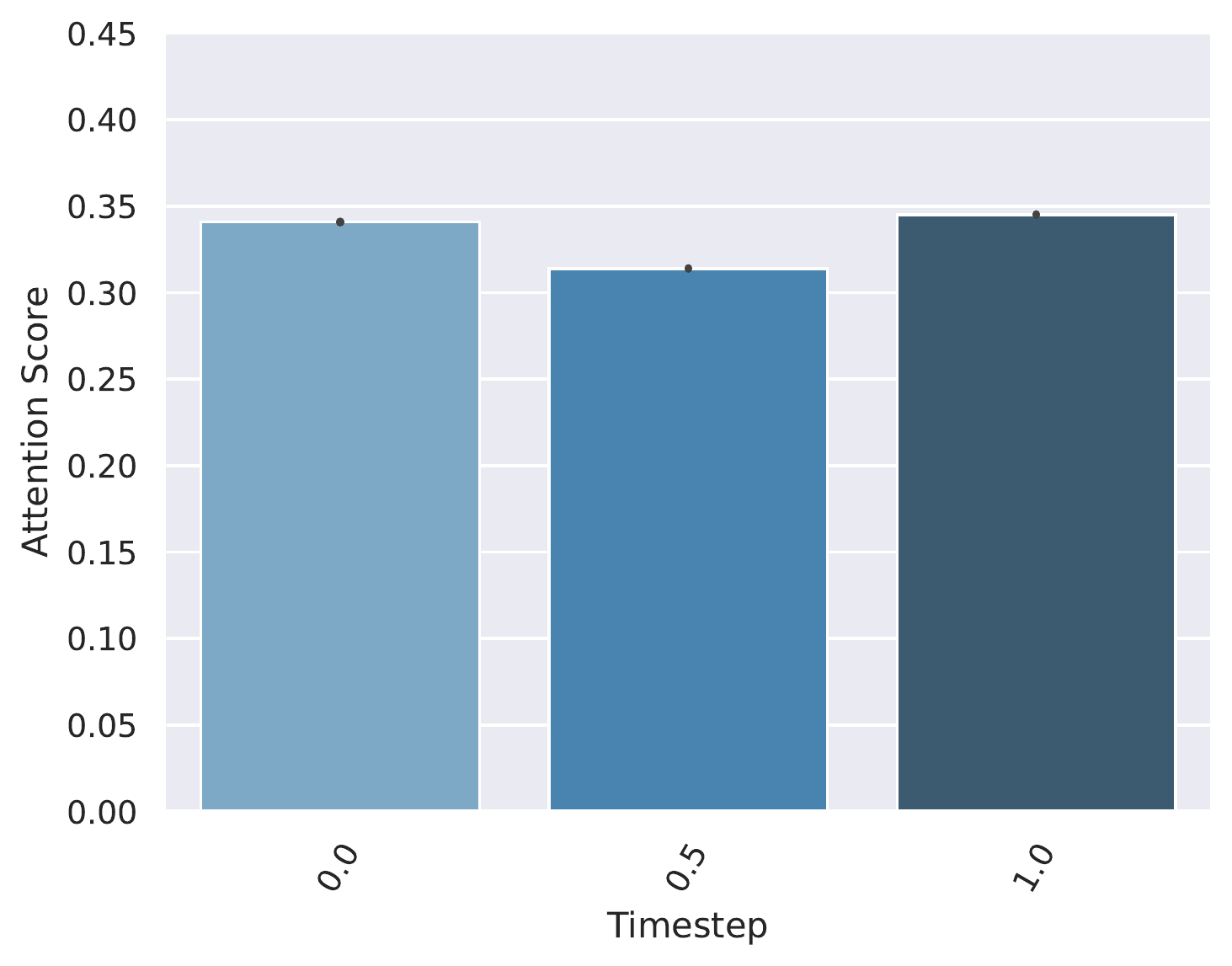}
\vspace{-6mm}
\subcaption*{\scriptsize Background Color}
\end{subfigure}
\begin{subfigure}[c]{0.32\textwidth}
\includegraphics[width=\textwidth]{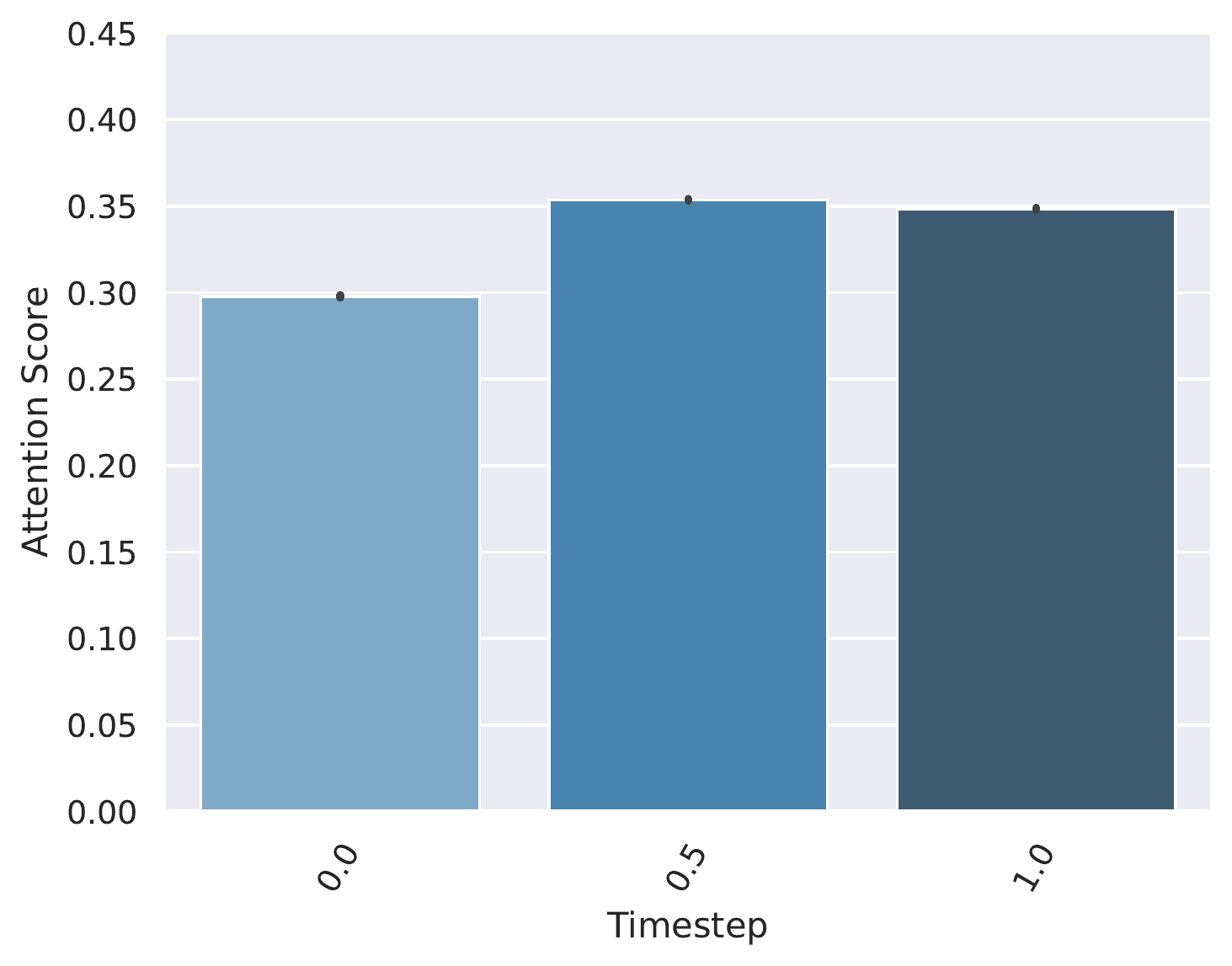}
\vspace{-6mm}
\subcaption*{\scriptsize Foreground Color}
\end{subfigure}
\begin{subfigure}[c]{0.32\textwidth}
\includegraphics[width=\textwidth]{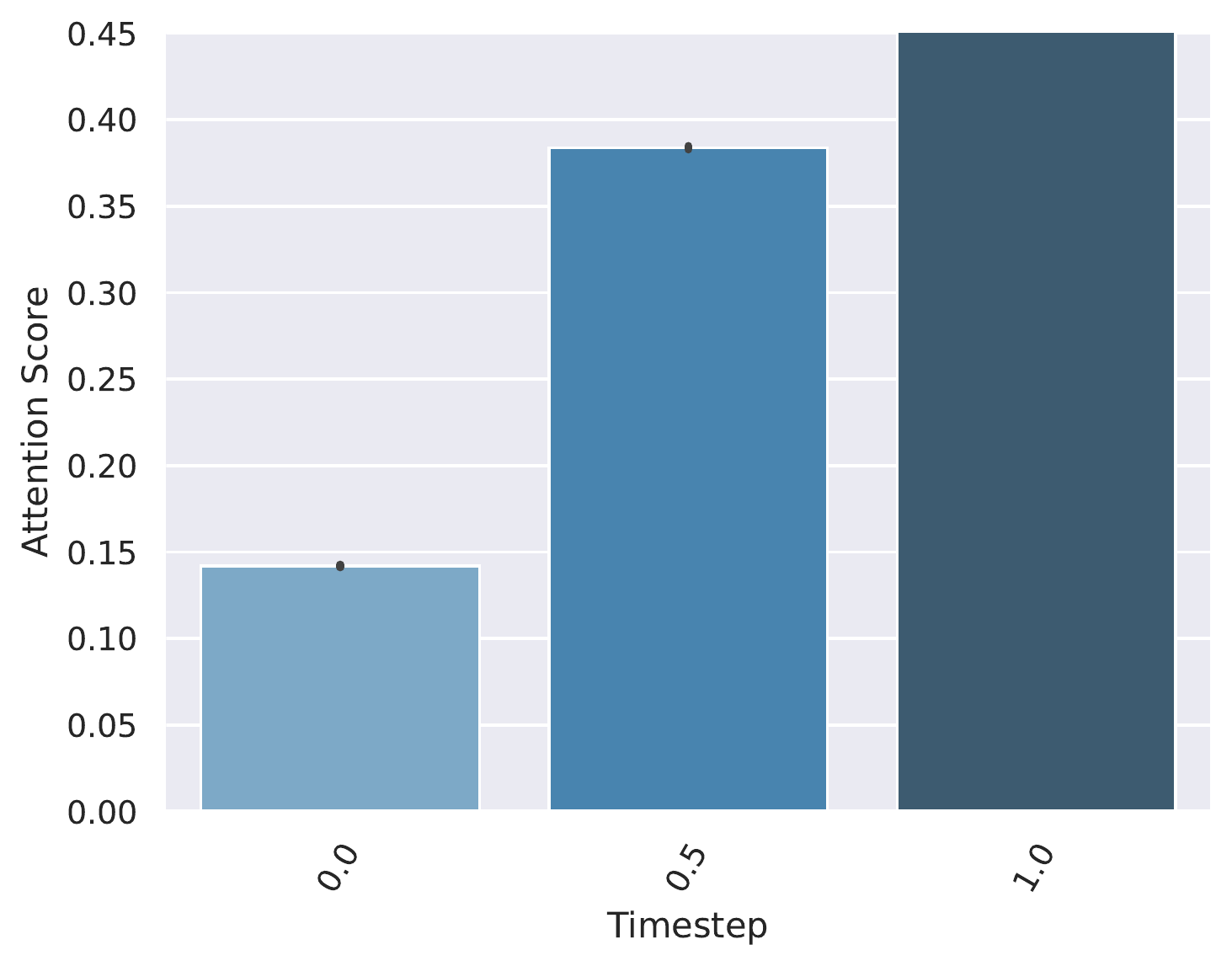}
\vspace{-6mm}
\subcaption*{\scriptsize Digit}
\end{subfigure}
\subcaption*{Granularity: 2}
\end{subfigure} \\
\begin{subfigure}[c]{\textwidth}
\begin{subfigure}[c]{0.32\textwidth}
\includegraphics[width=\textwidth]{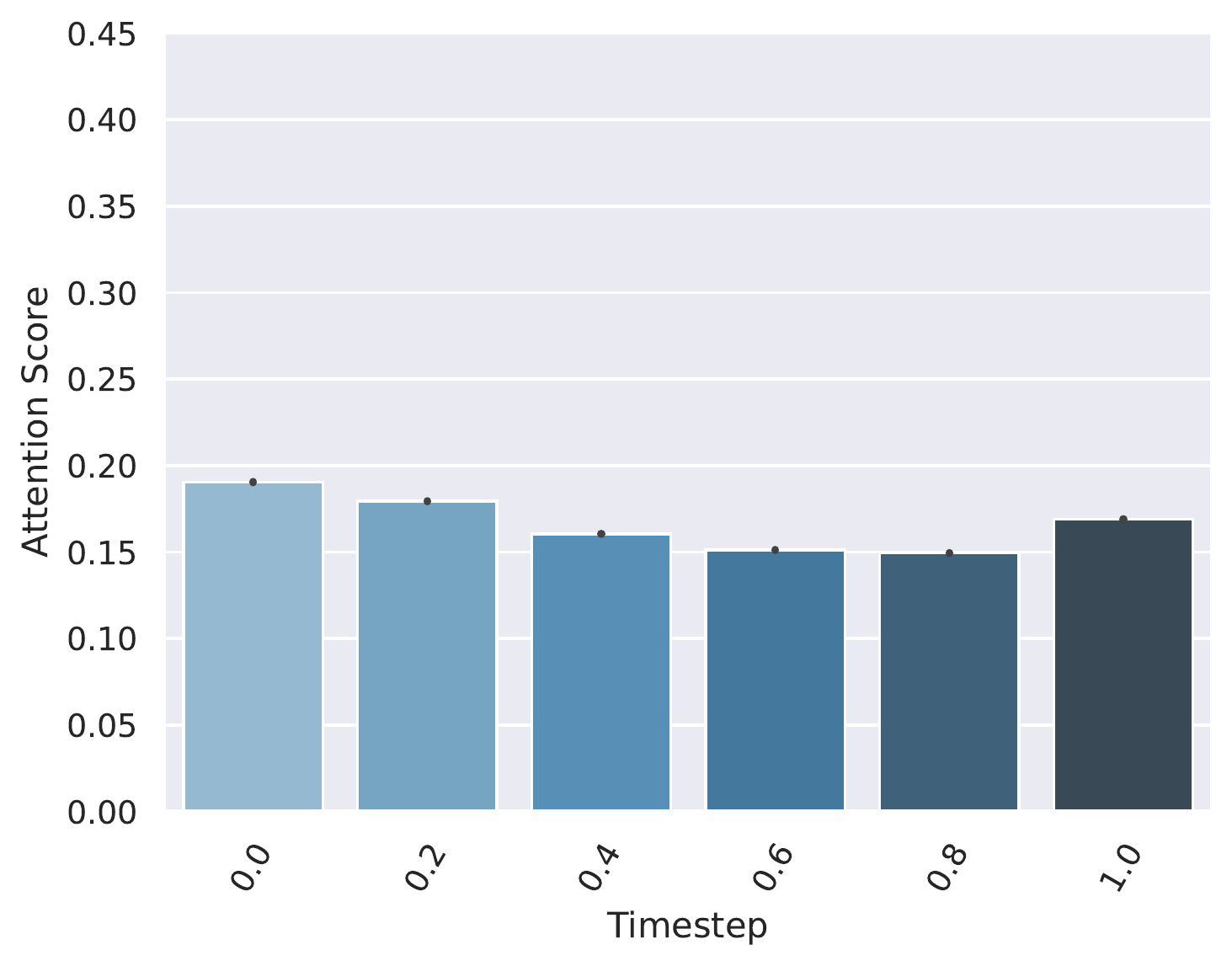}
\vspace{-6mm}
\subcaption*{\scriptsize Background Color}
\end{subfigure}
\begin{subfigure}[c]{0.32\textwidth}
\includegraphics[width=\textwidth]{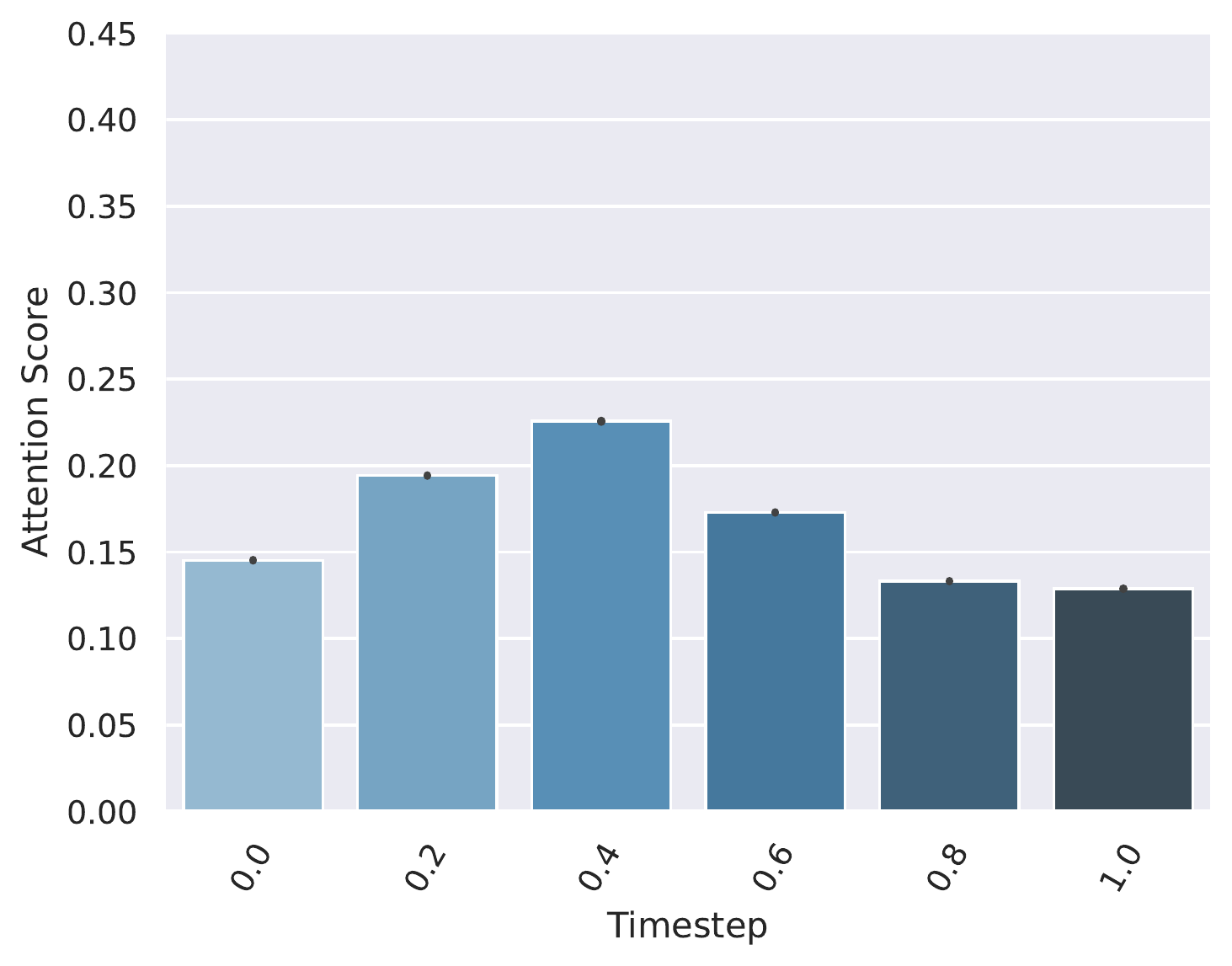}
\vspace{-6mm}
\subcaption*{\scriptsize Foreground Color}
\end{subfigure}
\begin{subfigure}[c]{0.32\textwidth}
\includegraphics[width=\textwidth]{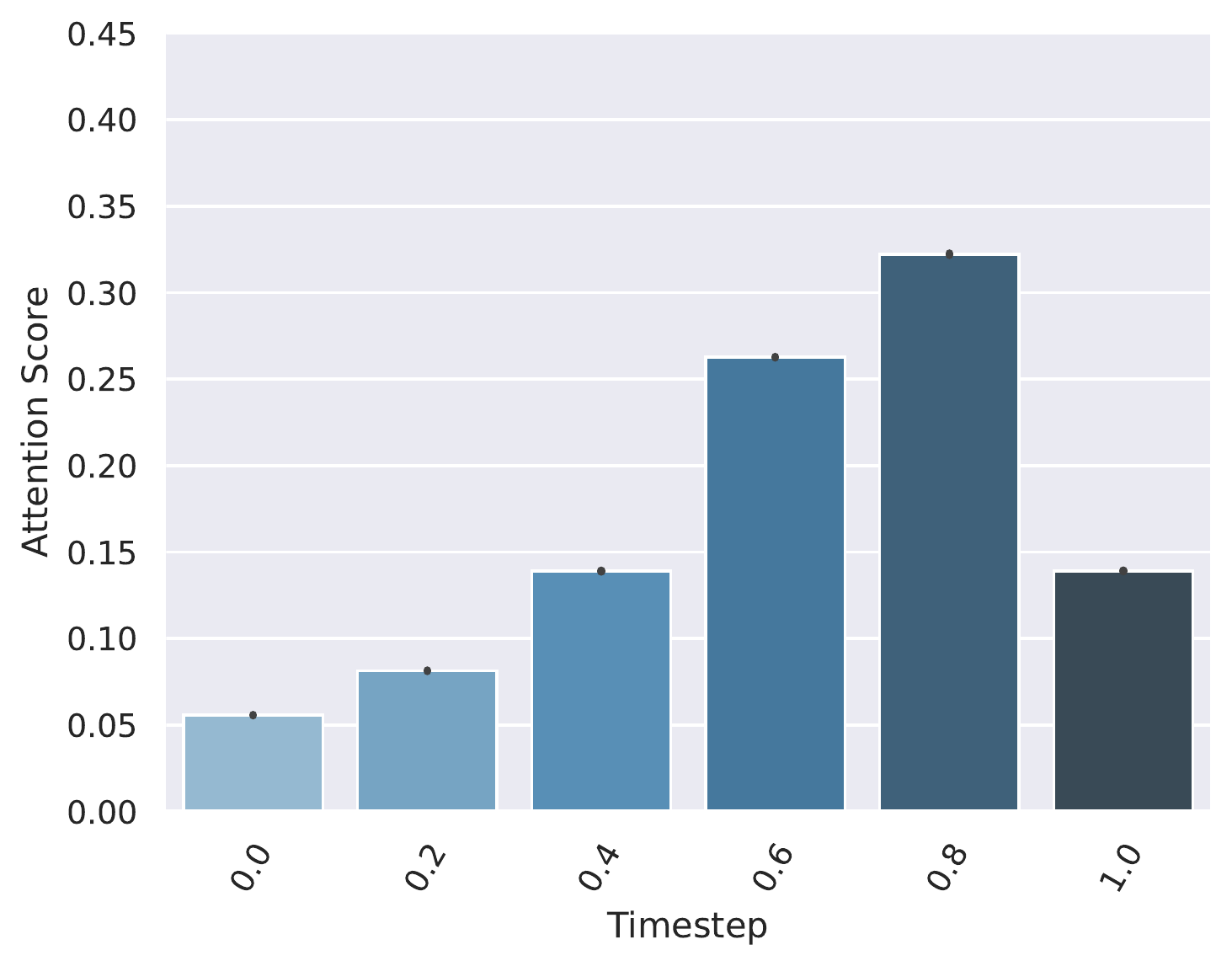}
\vspace{-6mm}
\subcaption*{\scriptsize Digit}
\end{subfigure}
\subcaption*{Granularity: 5}
\end{subfigure} \\
\begin{subfigure}[c]{\textwidth}
\begin{subfigure}[c]{0.32\textwidth}
\includegraphics[width=\textwidth]{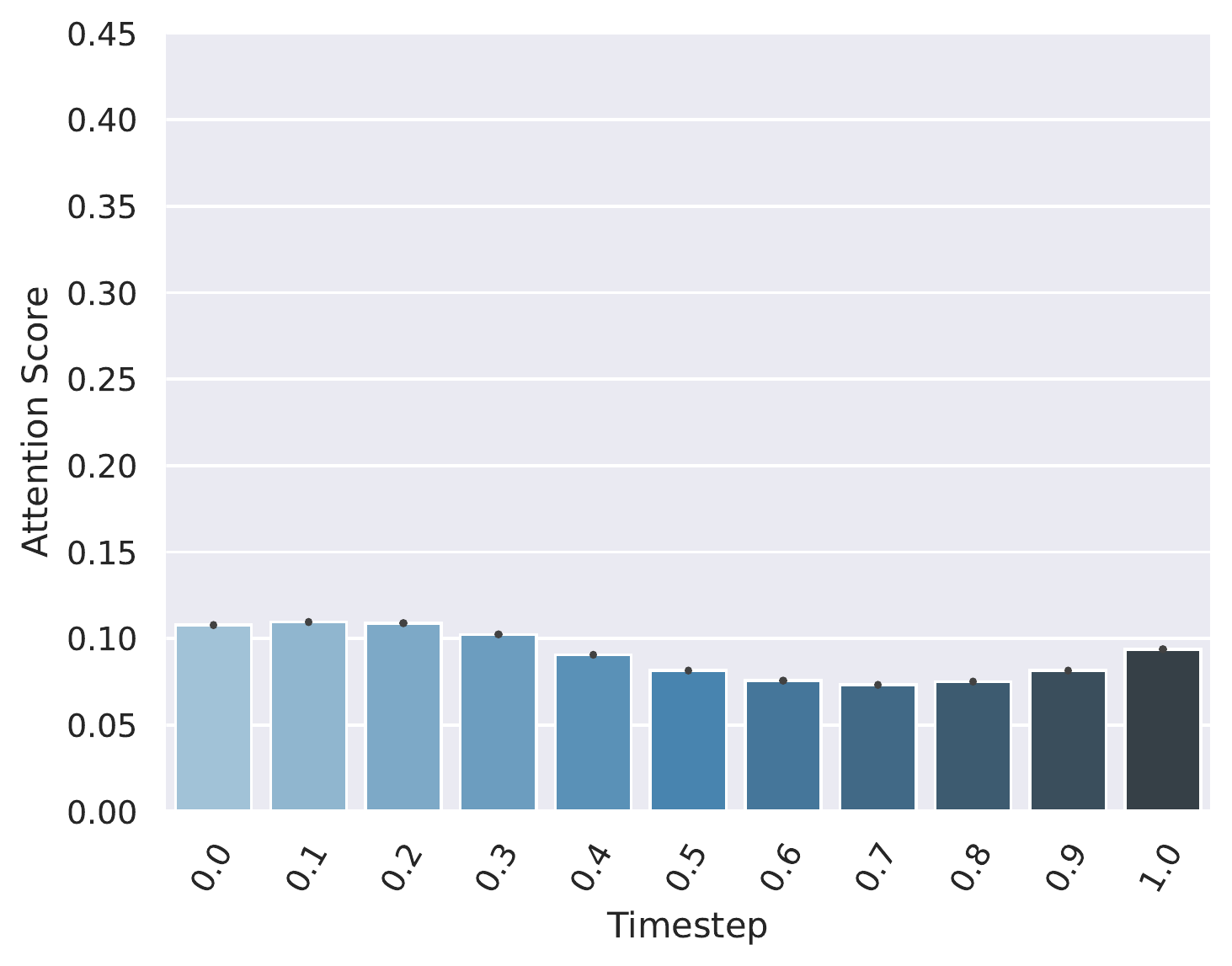}
\vspace{-6mm}
\subcaption*{\scriptsize Background Color}
\end{subfigure}
\begin{subfigure}[c]{0.32\textwidth}
\includegraphics[width=\textwidth]{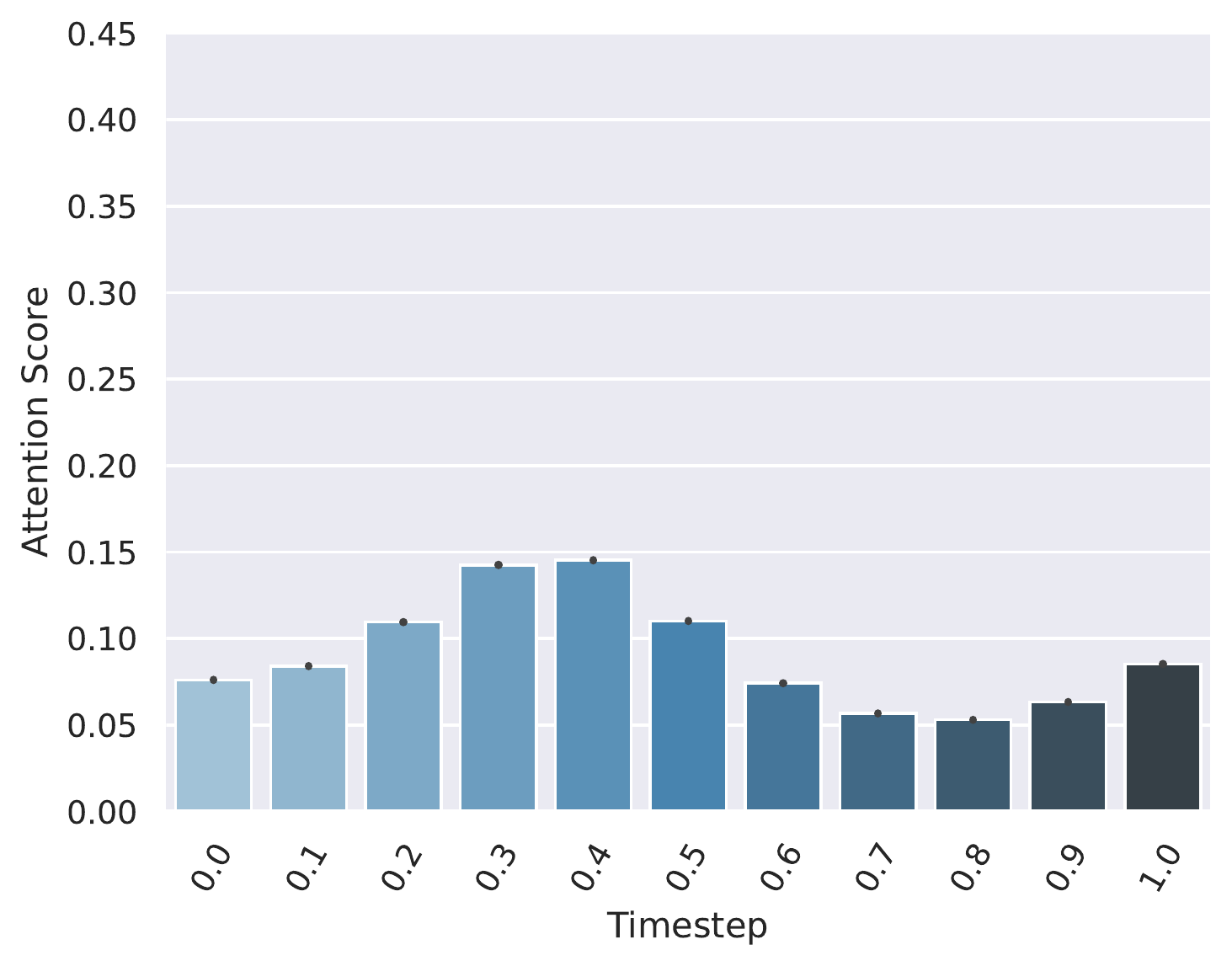}
\vspace{-6mm}
\subcaption*{\scriptsize Foreground Color}
\end{subfigure}
\begin{subfigure}[c]{0.32\textwidth}
\includegraphics[width=\textwidth]{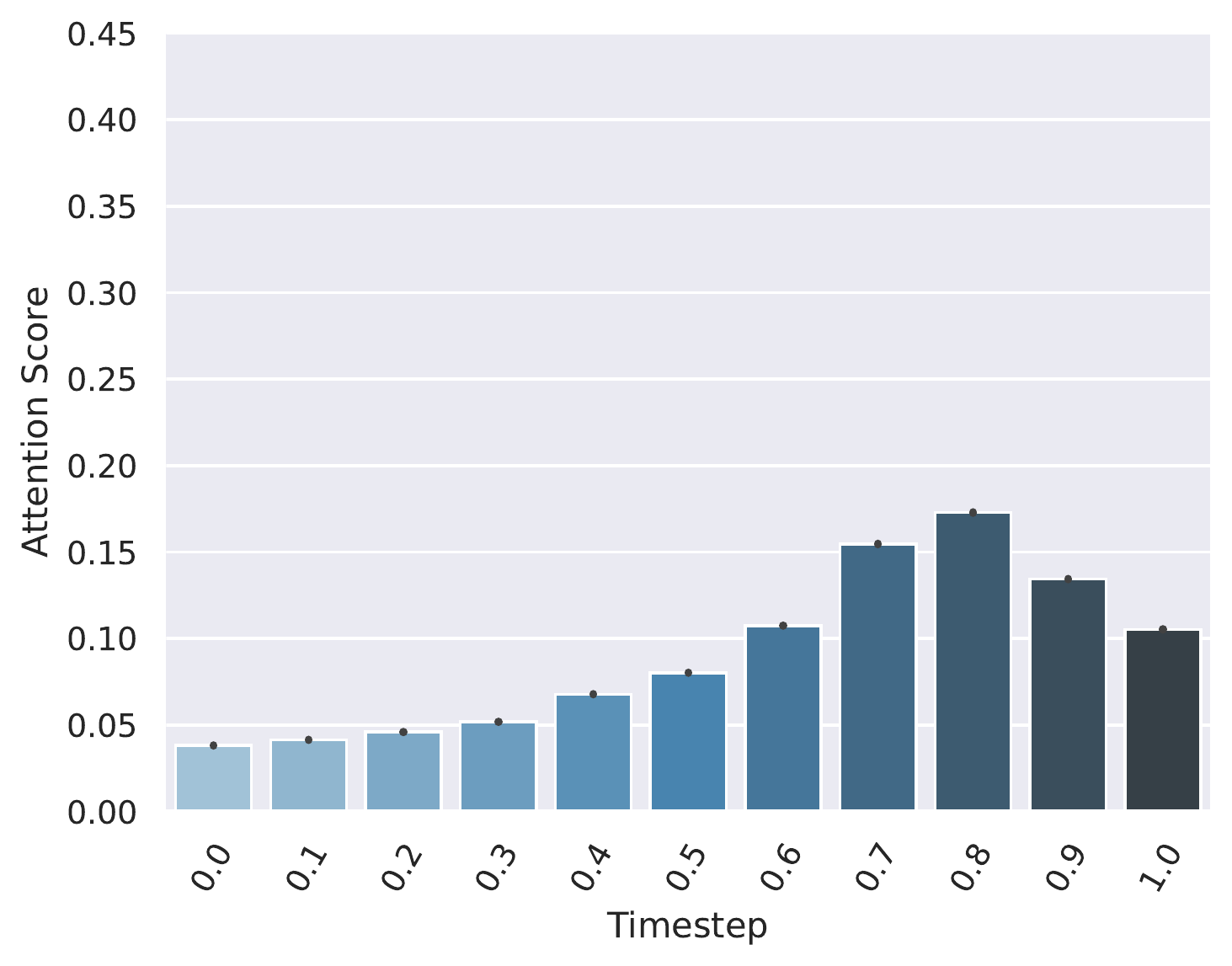}
\vspace{-6mm}
\subcaption*{\scriptsize Digit}
\end{subfigure}
\subcaption*{Granularity: 10}
\end{subfigure} \\
\begin{subfigure}[c]{\textwidth}
\begin{subfigure}[c]{0.32\textwidth}
\includegraphics[width=\textwidth]{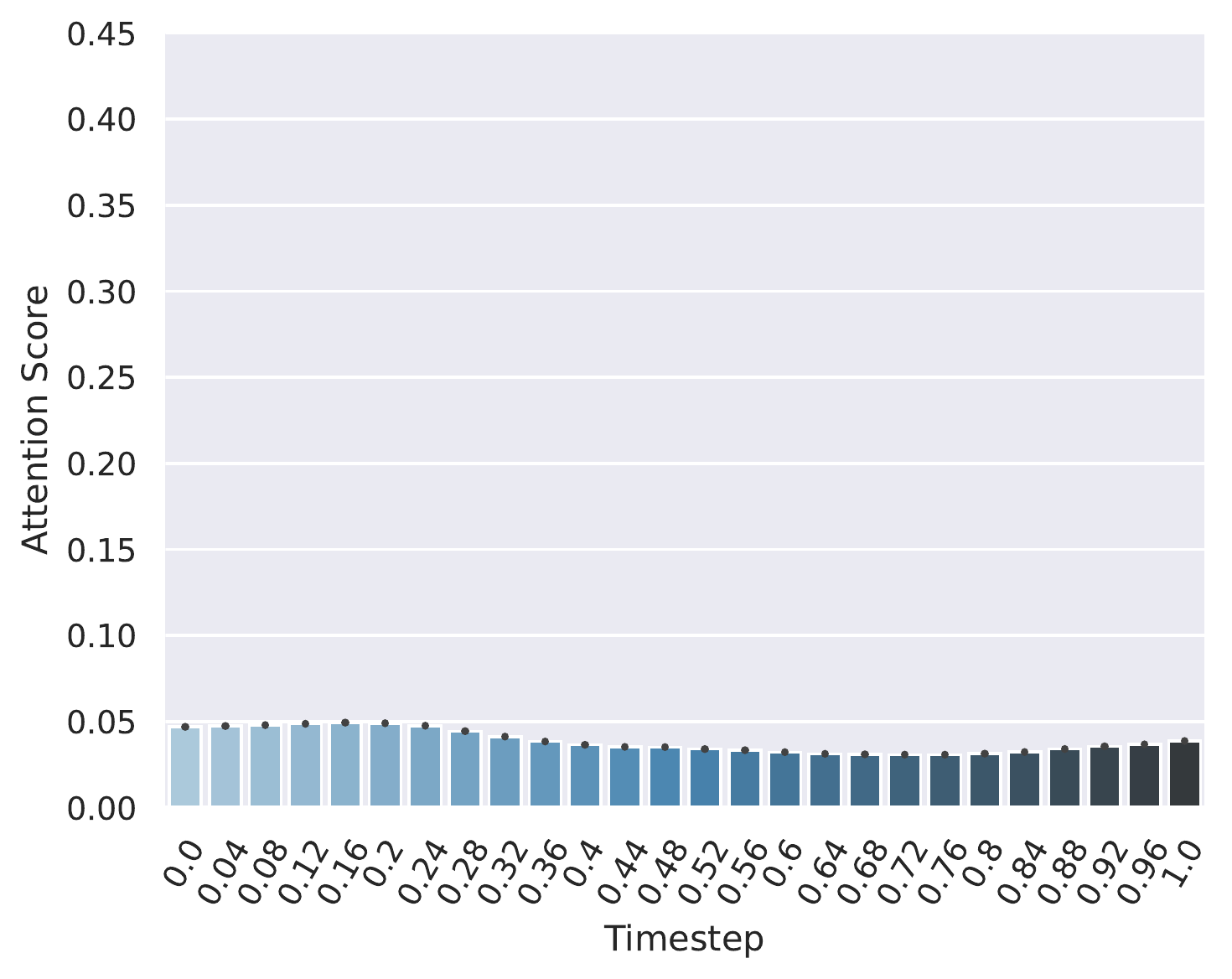}
\vspace{-6mm}
\subcaption*{\scriptsize Background Color}
\end{subfigure}
\begin{subfigure}[c]{0.32\textwidth}
\includegraphics[width=\textwidth]{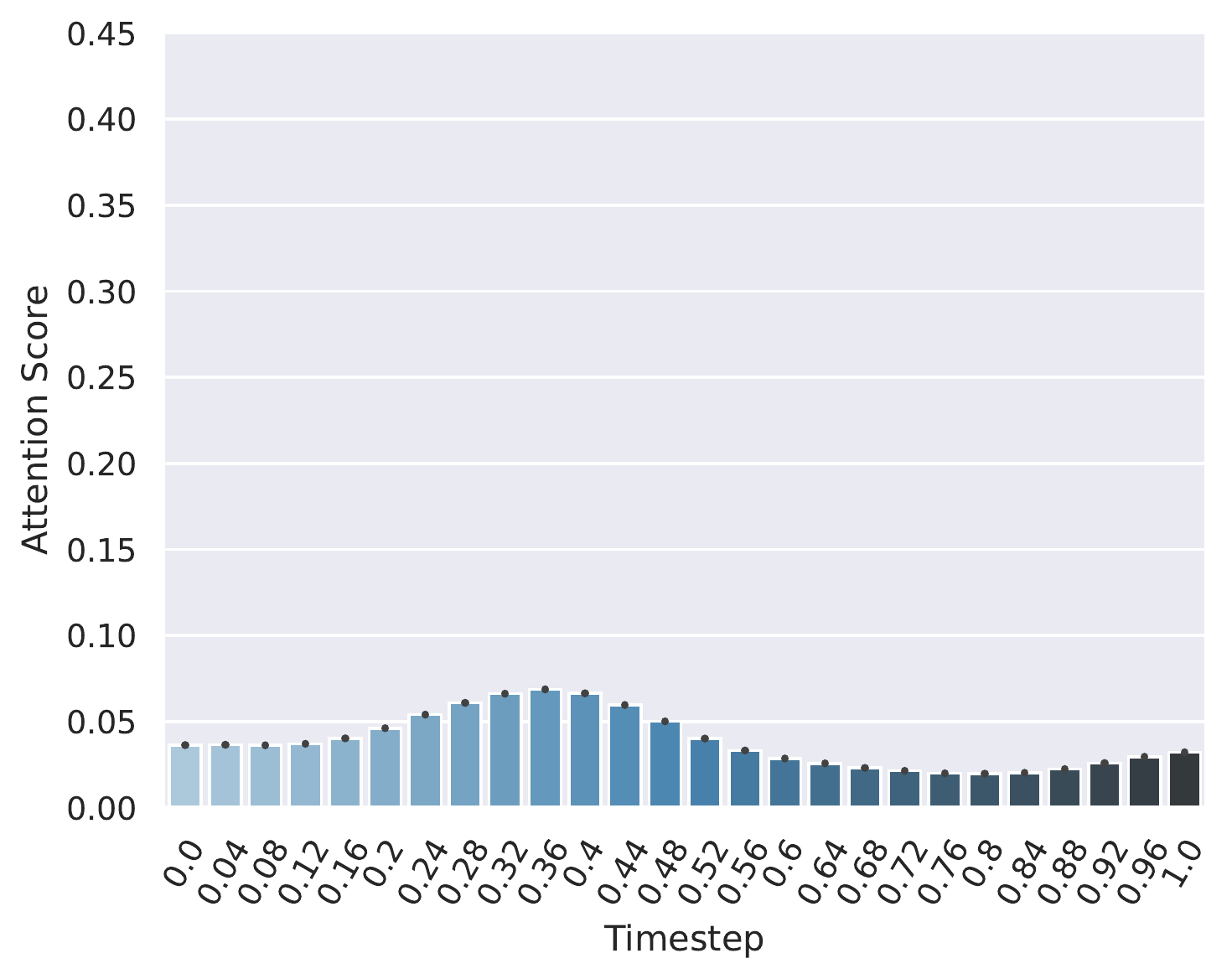}
\vspace{-6mm}
\subcaption*{\scriptsize Foreground Color}
\end{subfigure}
\begin{subfigure}[c]{0.32\textwidth}
\includegraphics[width=\textwidth]{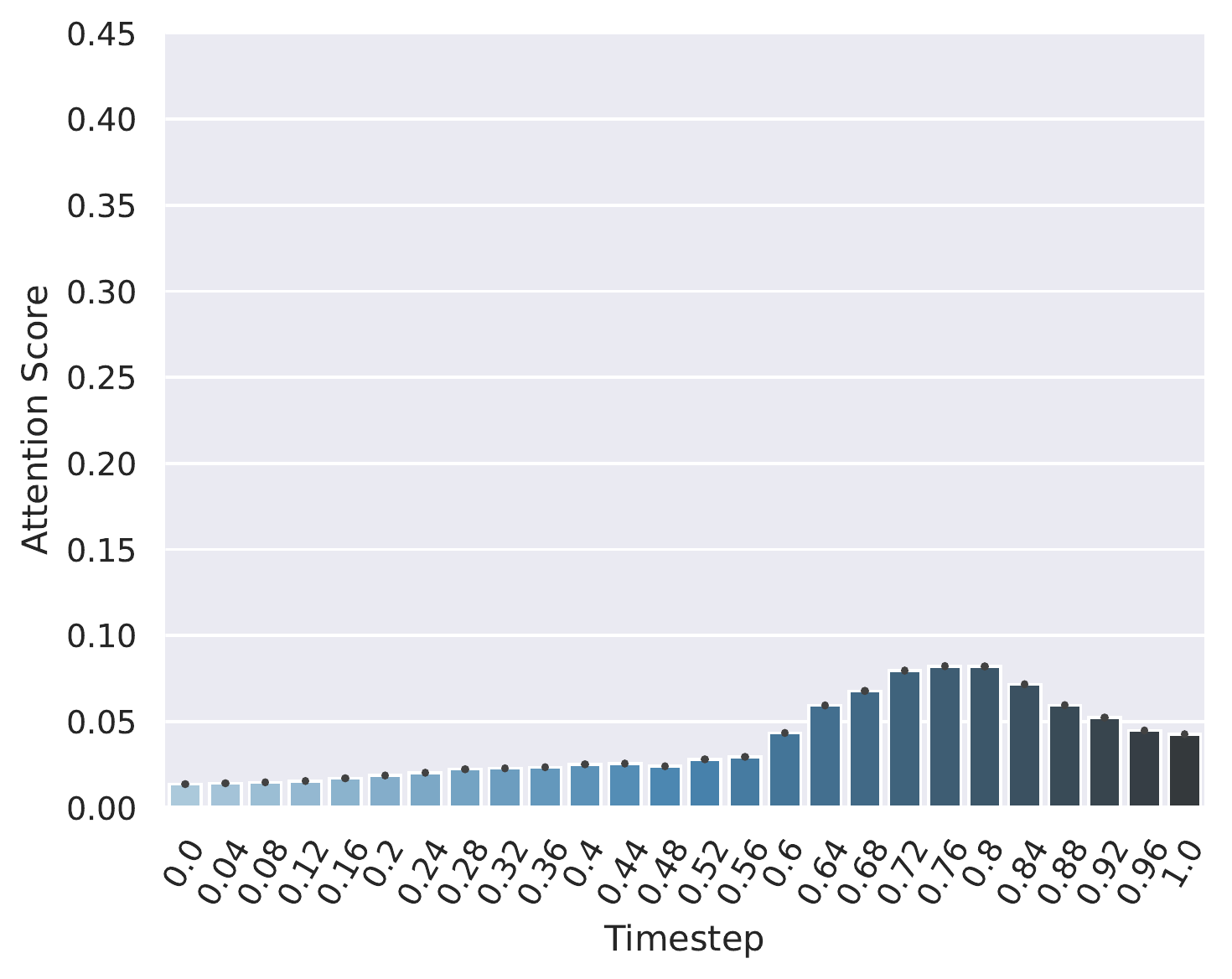}
\vspace{-6mm}
\subcaption*{\scriptsize Digit}
\end{subfigure}
\subcaption*{Granularity: 25}
\end{subfigure} \\
\caption{Attention score profiles for the Colored-MNIST dataset on the different features, using different granularities, with the dimensionality of the latent space as 16 and the DRL encoder.}
\label{fig:cm_DRL_16}
\end{figure}
\begin{figure}
\begin{subfigure}[c]{\textwidth}
\begin{subfigure}[c]{0.32\textwidth}
\includegraphics[width=\textwidth]{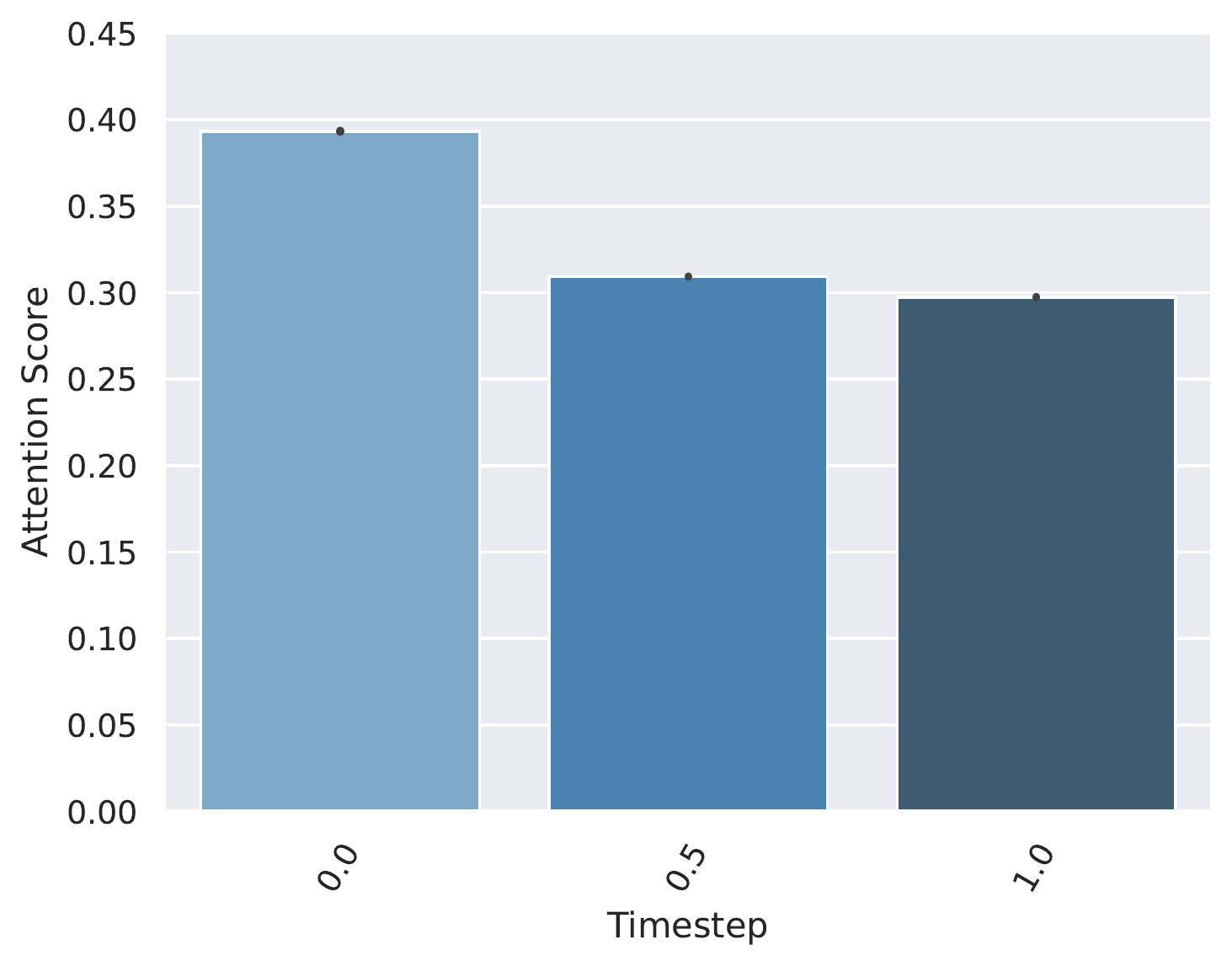}
\vspace{-6mm}
\subcaption*{\scriptsize Background Color}
\end{subfigure}
\begin{subfigure}[c]{0.32\textwidth}
\includegraphics[width=\textwidth]{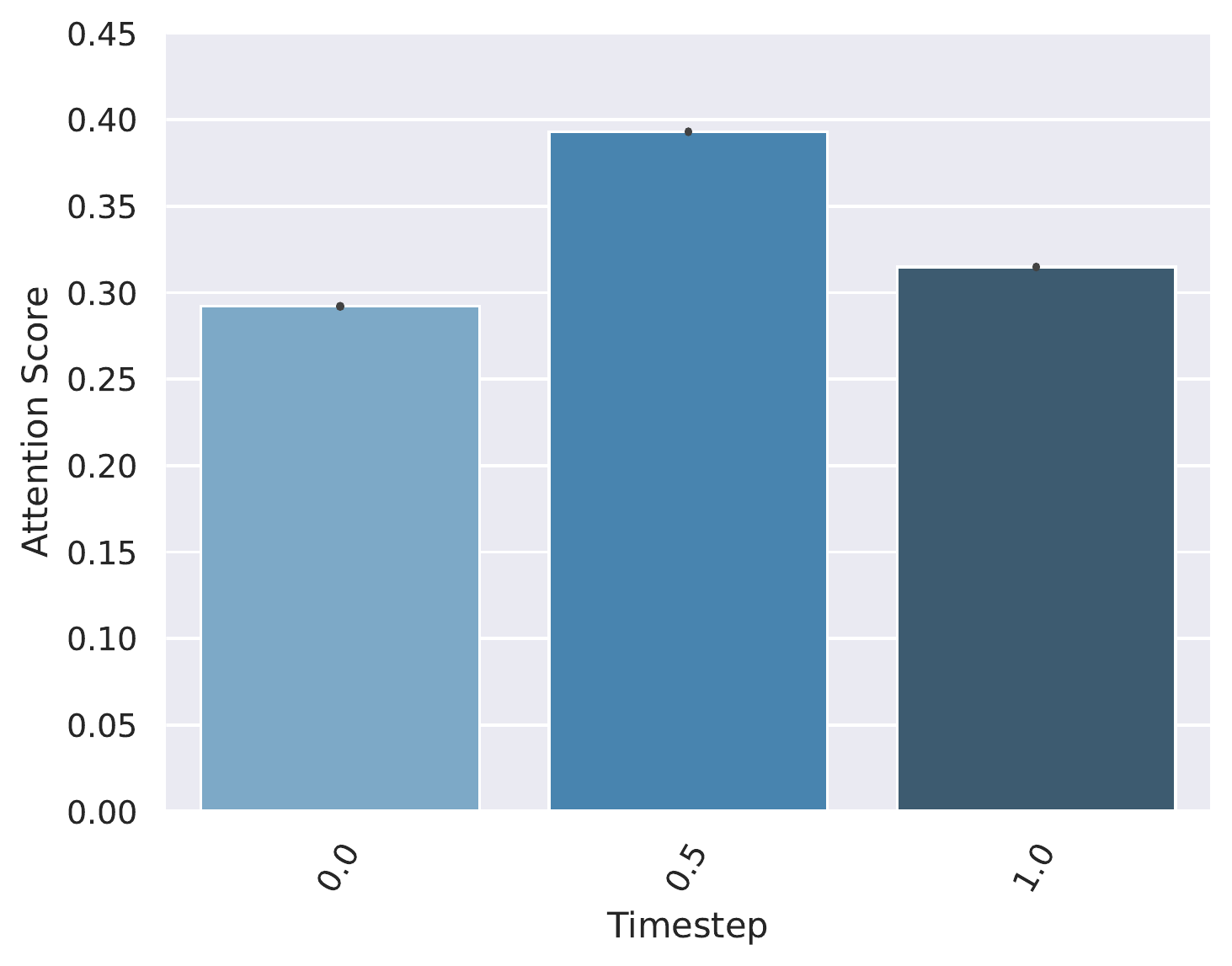}
\vspace{-6mm}
\subcaption*{\scriptsize Foreground Color}
\end{subfigure}
\begin{subfigure}[c]{0.32\textwidth}
\includegraphics[width=\textwidth]{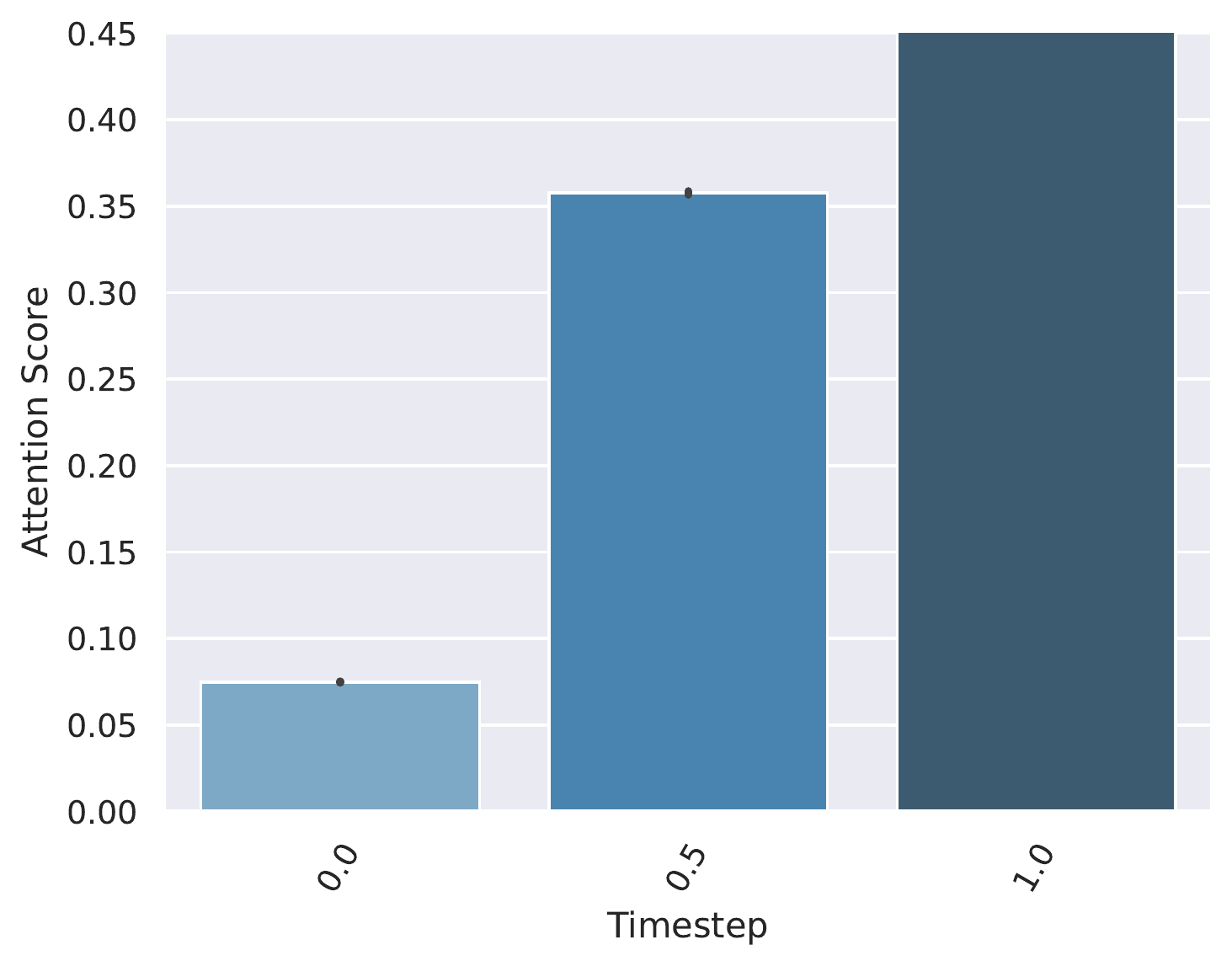}
\vspace{-6mm}
\subcaption*{\scriptsize Digit}
\end{subfigure}
\subcaption*{Granularity: 2}
\end{subfigure} \\
\begin{subfigure}[c]{\textwidth}
\begin{subfigure}[c]{0.32\textwidth}
\includegraphics[width=\textwidth]{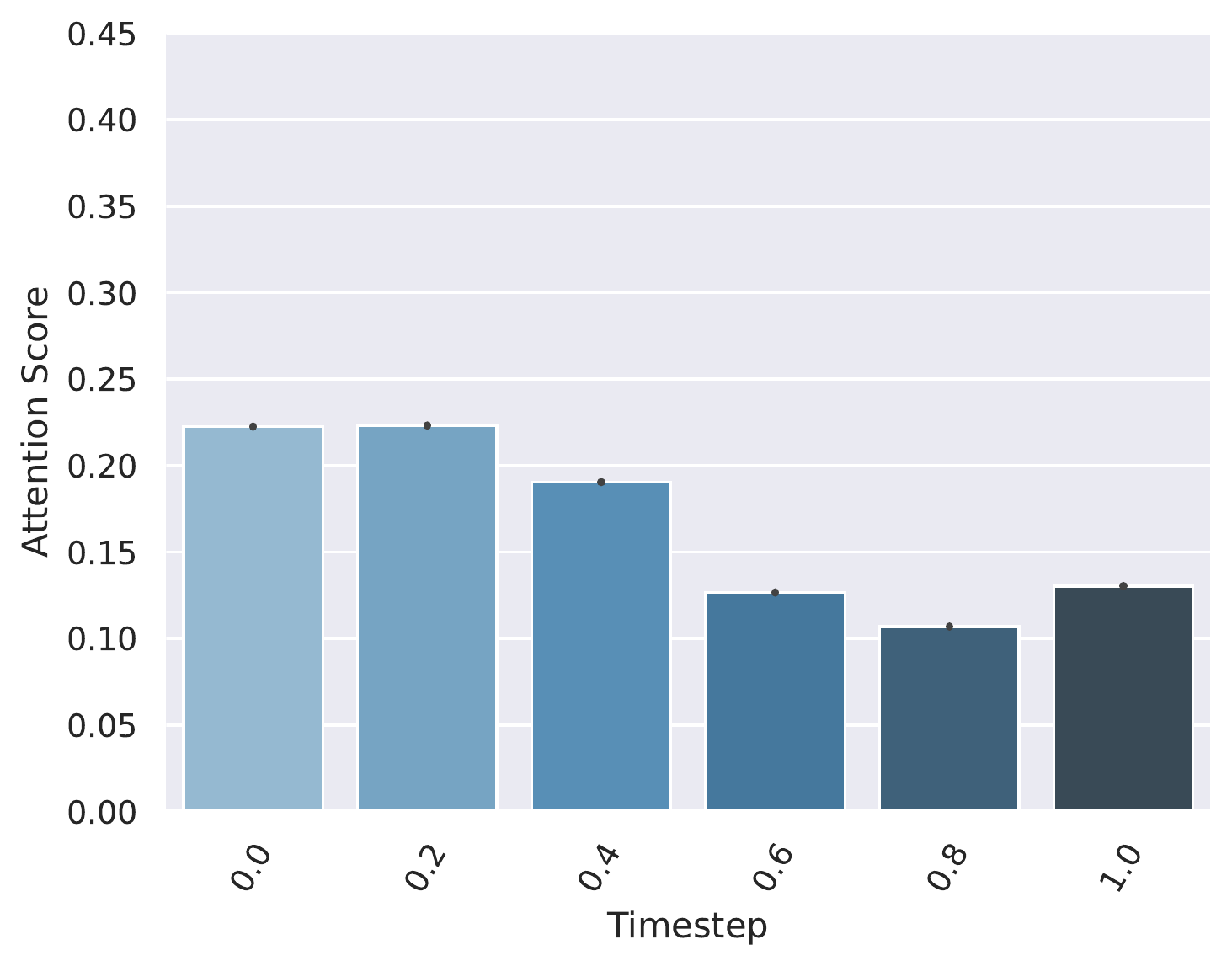}
\vspace{-6mm}
\subcaption*{\scriptsize Background Color}
\end{subfigure}
\begin{subfigure}[c]{0.32\textwidth}
\includegraphics[width=\textwidth]{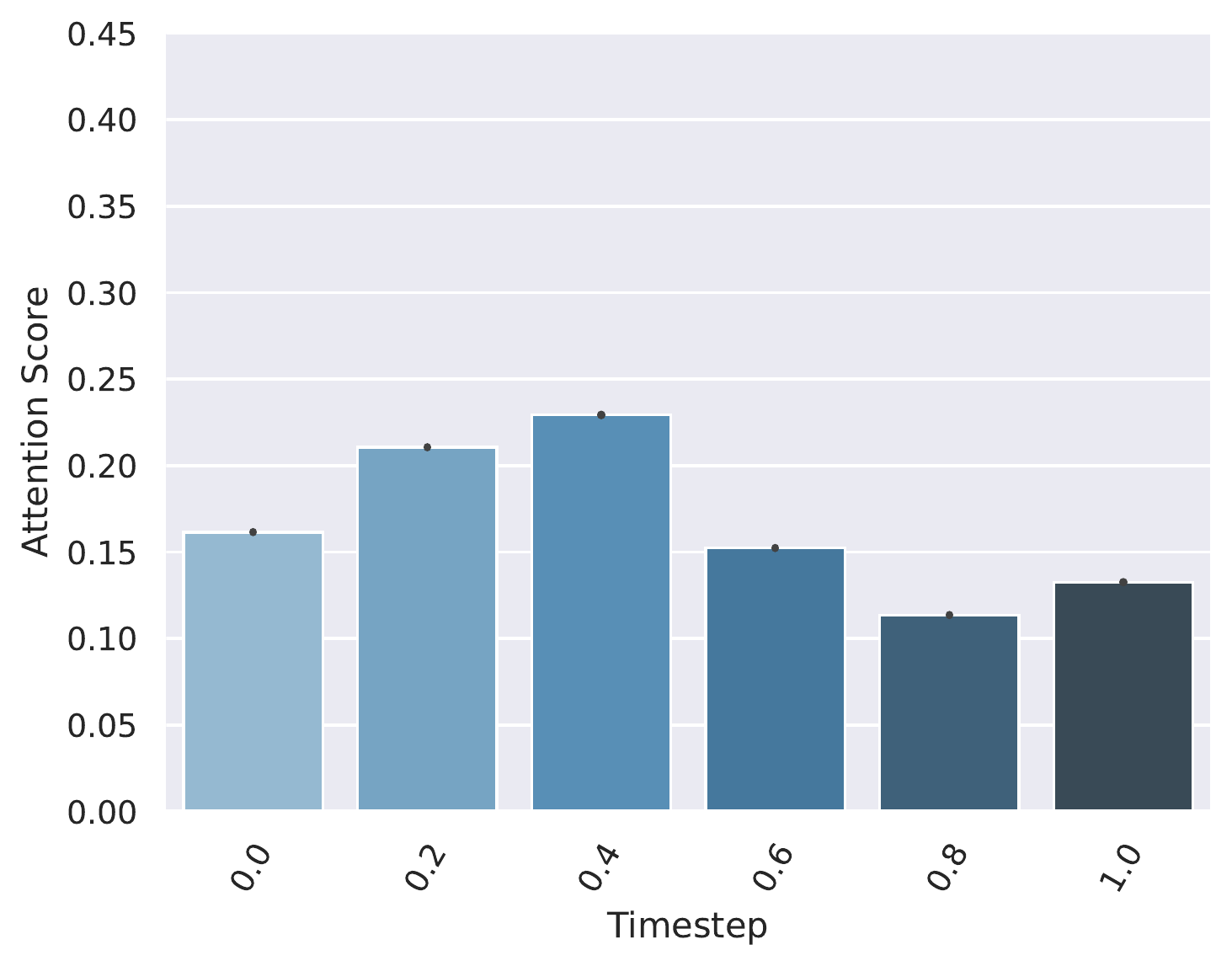}
\vspace{-6mm}
\subcaption*{\scriptsize Foreground Color}
\end{subfigure}
\begin{subfigure}[c]{0.32\textwidth}
\includegraphics[width=\textwidth]{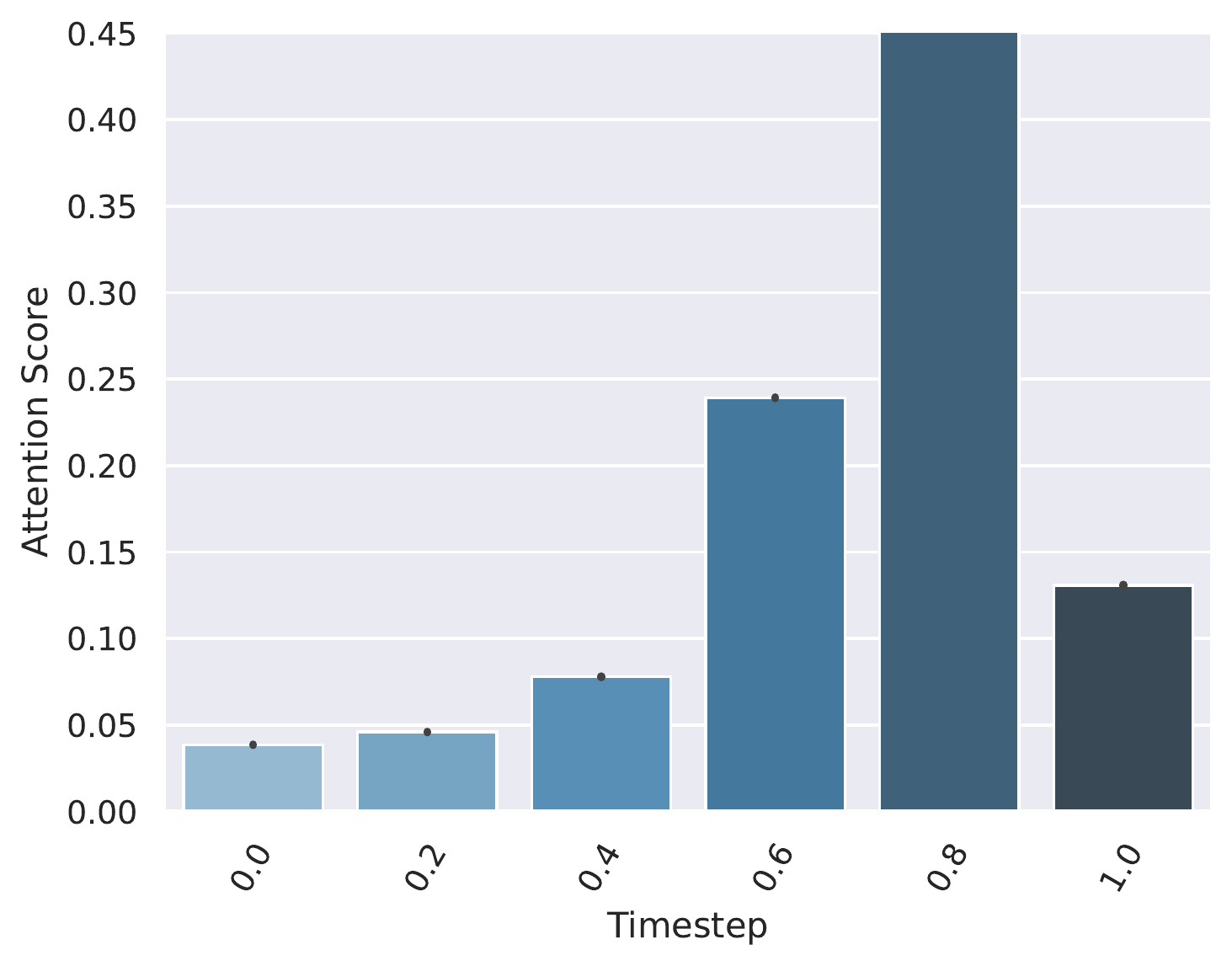}
\vspace{-6mm}
\subcaption*{\scriptsize Digit}
\end{subfigure}
\subcaption*{Granularity: 5}
\end{subfigure} \\
\begin{subfigure}[c]{\textwidth}
\begin{subfigure}[c]{0.32\textwidth}
\includegraphics[width=\textwidth]{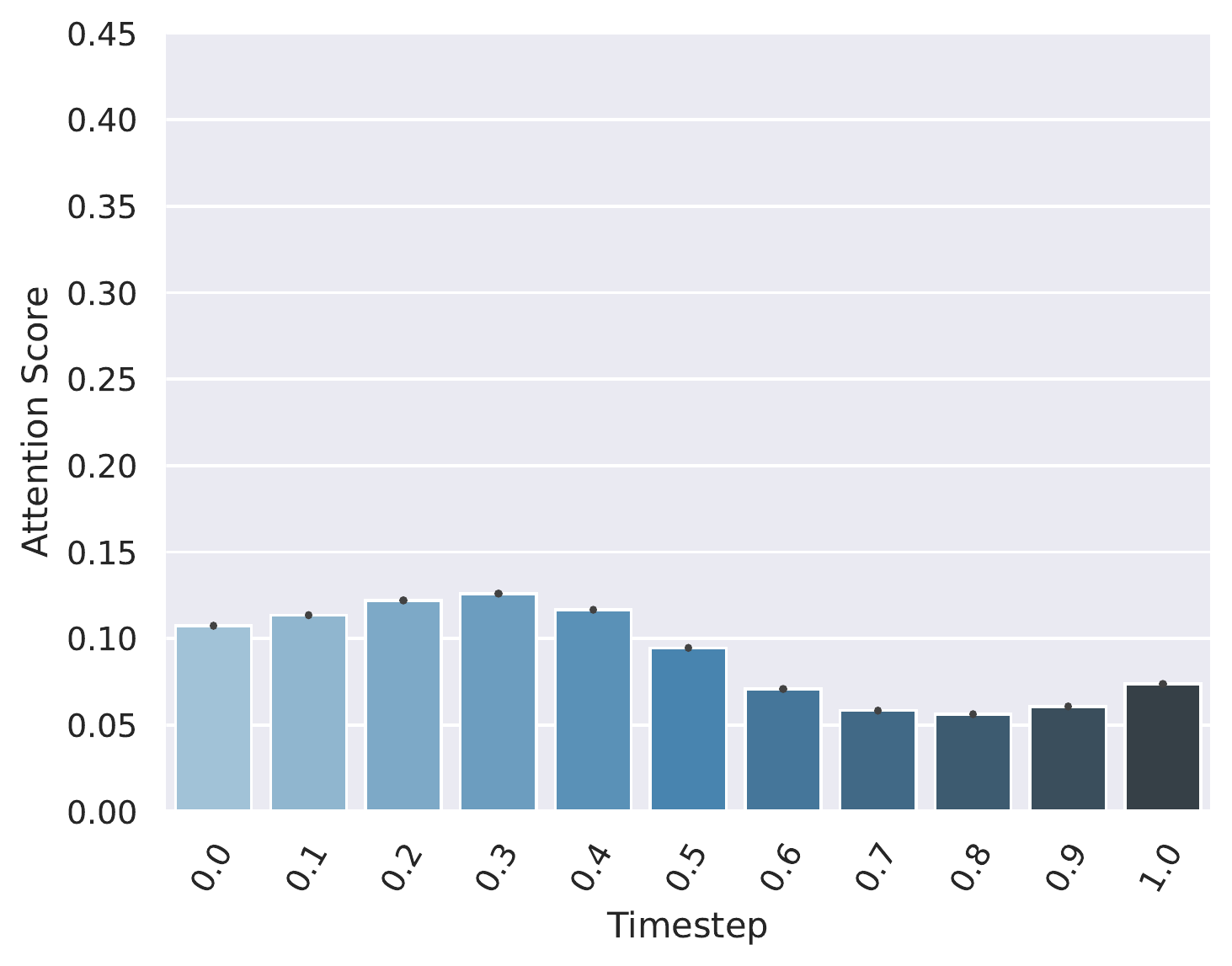}
\vspace{-6mm}
\subcaption*{\scriptsize Background Color}
\end{subfigure}
\begin{subfigure}[c]{0.32\textwidth}
\includegraphics[width=\textwidth]{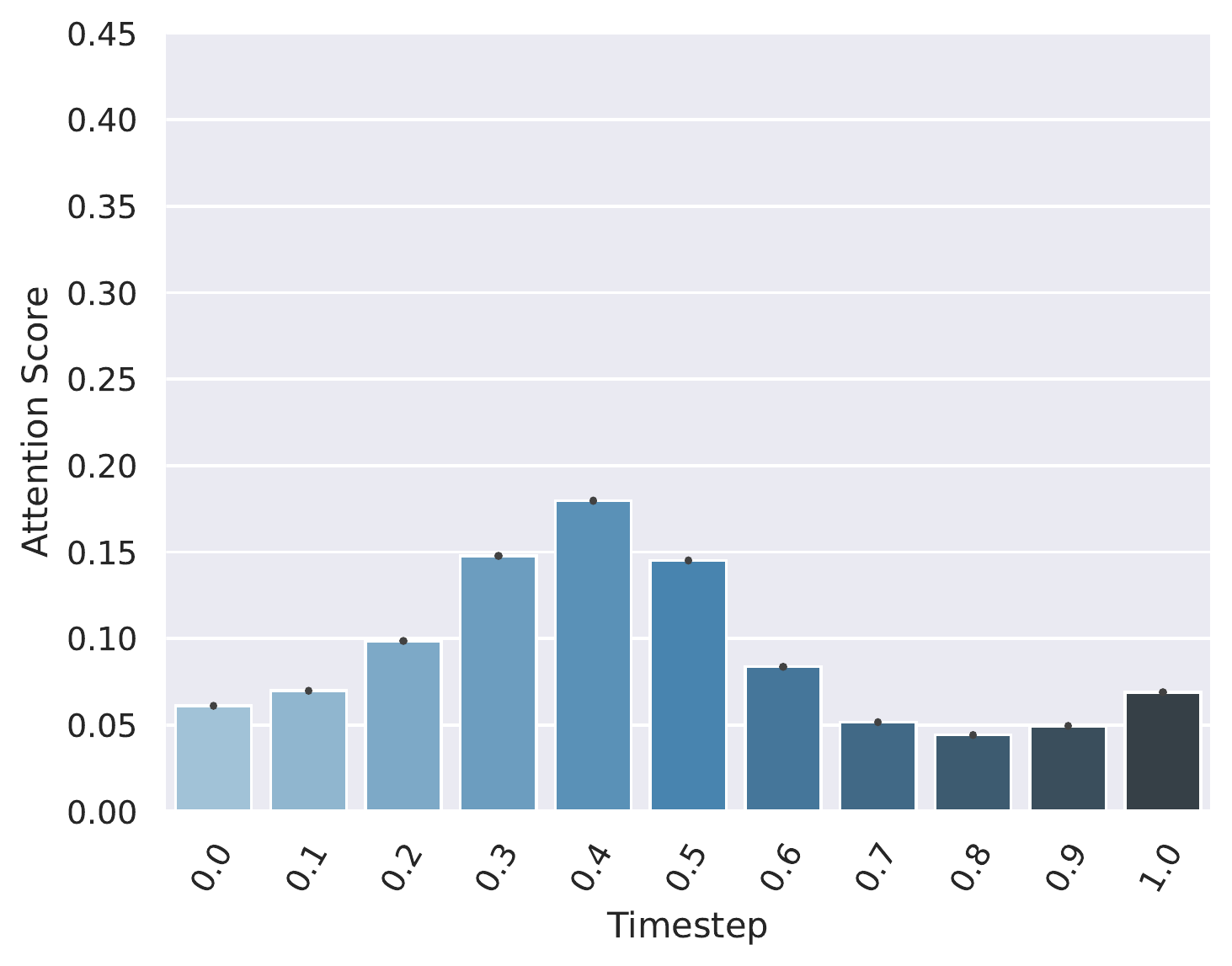}
\vspace{-6mm}
\subcaption*{\scriptsize Foreground Color}
\end{subfigure}
\begin{subfigure}[c]{0.32\textwidth}
\includegraphics[width=\textwidth]{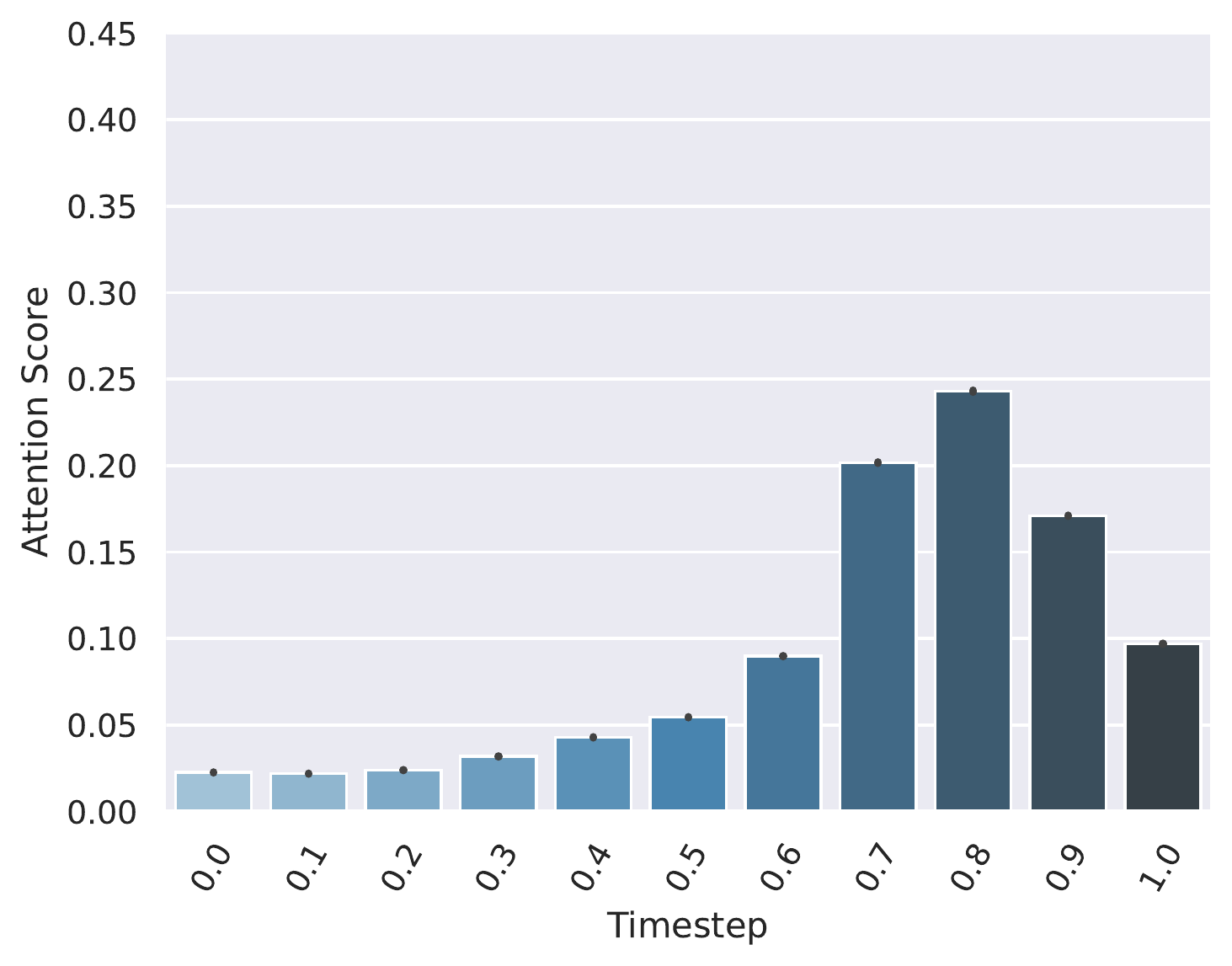}
\vspace{-6mm}
\subcaption*{\scriptsize Digit}
\end{subfigure}
\subcaption*{Granularity: 10}
\end{subfigure} \\
\begin{subfigure}[c]{\textwidth}
\begin{subfigure}[c]{0.32\textwidth}
\includegraphics[width=\textwidth]{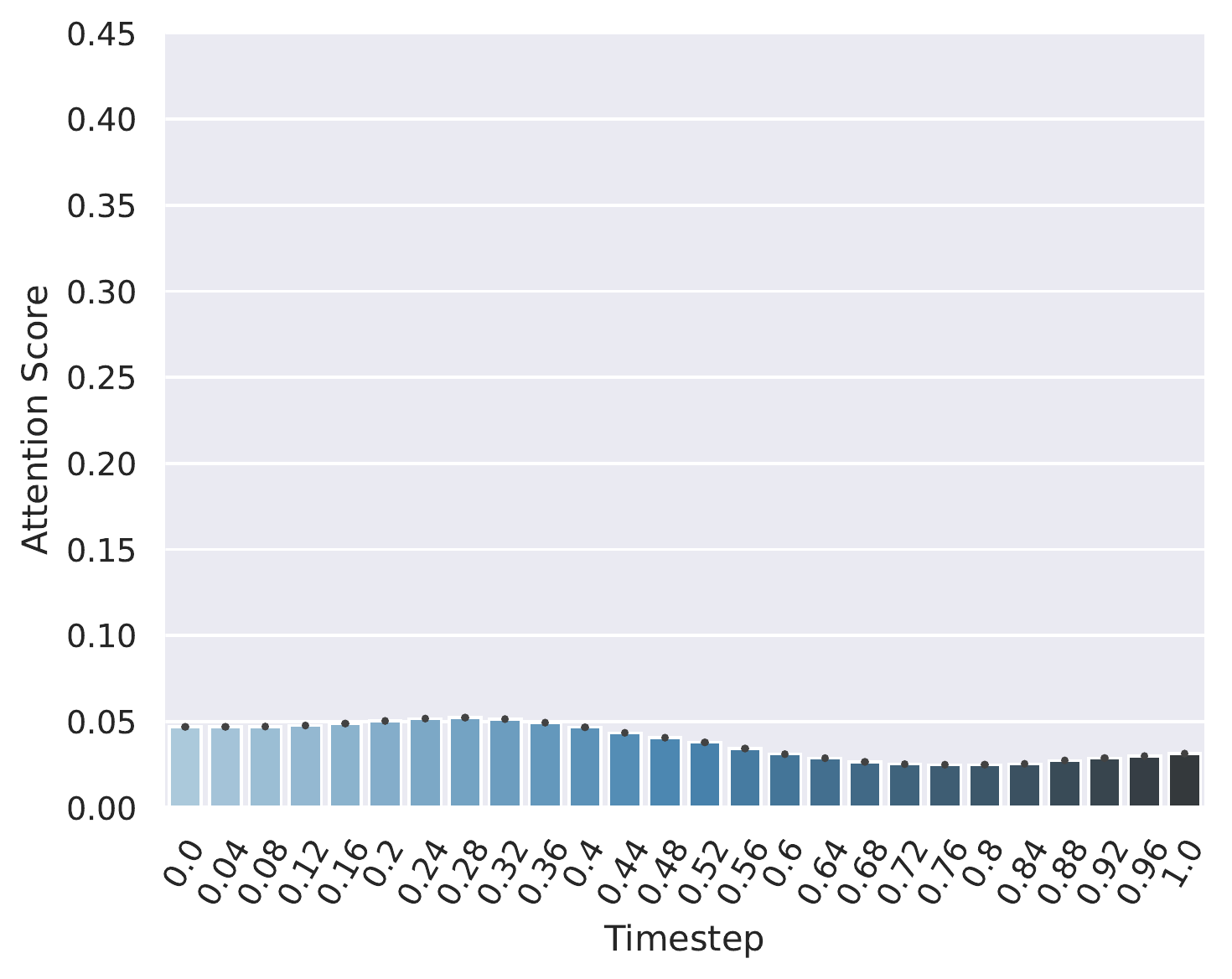}
\vspace{-6mm}
\subcaption*{\scriptsize Background Color}
\end{subfigure}
\begin{subfigure}[c]{0.32\textwidth}
\includegraphics[width=\textwidth]{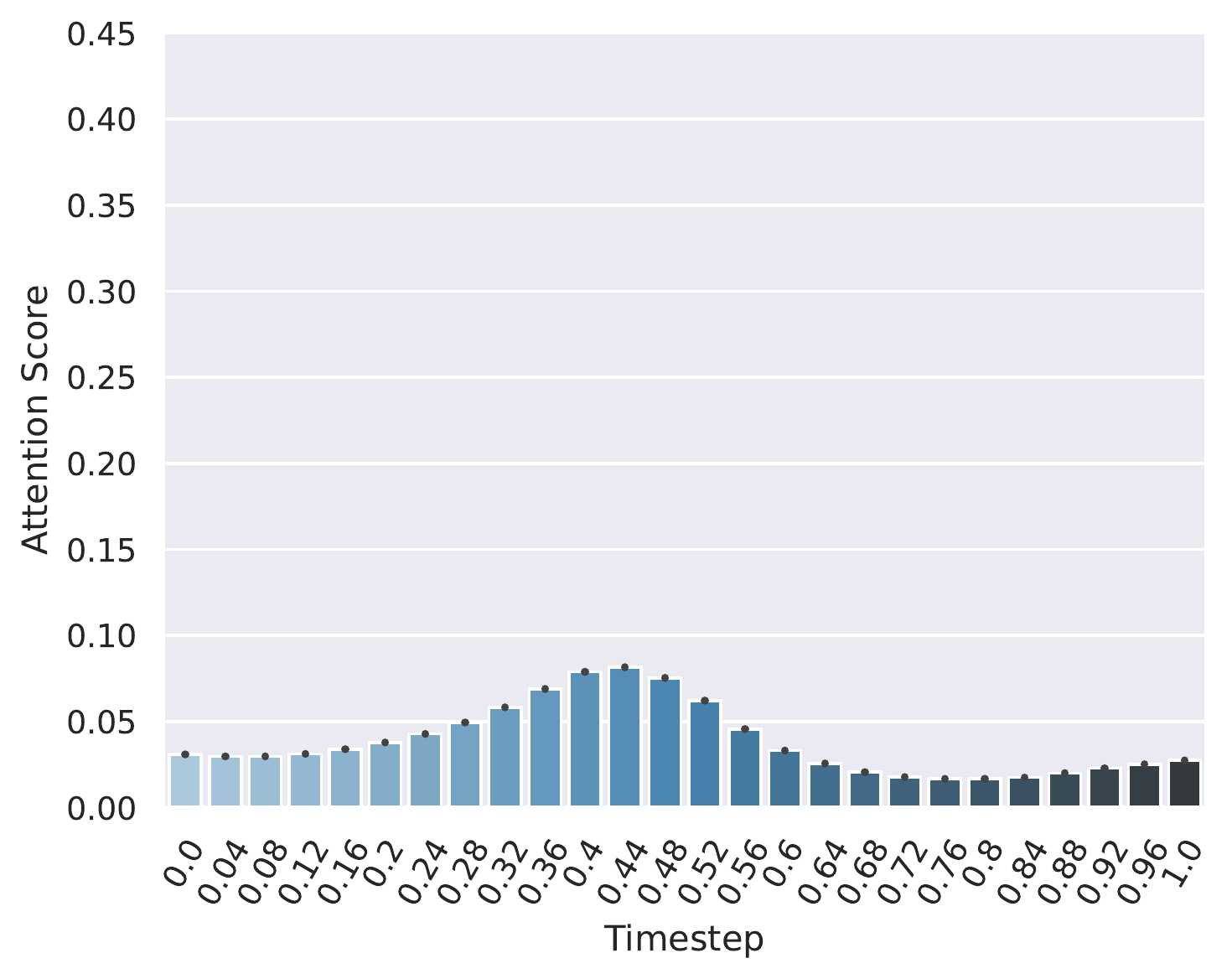}
\vspace{-6mm}
\subcaption*{\scriptsize Foreground Color}
\end{subfigure}
\begin{subfigure}[c]{0.32\textwidth}
\includegraphics[width=\textwidth]{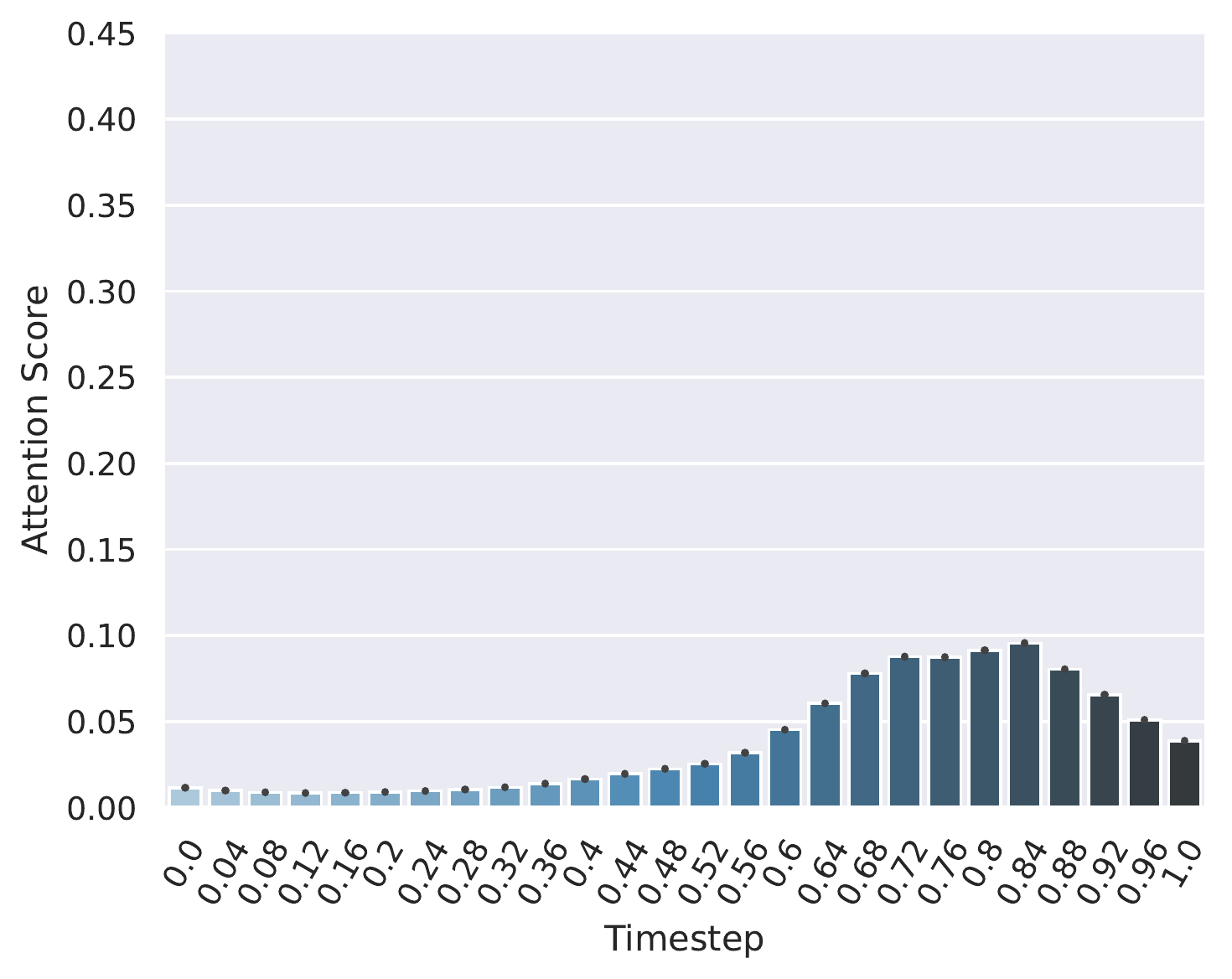}
\vspace{-6mm}
\subcaption*{\scriptsize Digit}
\end{subfigure}
\subcaption*{Granularity: 25}
\end{subfigure} \\
\caption{Attention score profiles for the Colored-MNIST dataset on the different features, using different granularities, with the dimensionality of the latent space as 32 and the DRL encoder.}
\label{fig:cm_DRL_32}
\end{figure}

\begin{figure}
    \centering
    \begin{subfigure}[c]{\linewidth}
    \includegraphics[width=\textwidth]{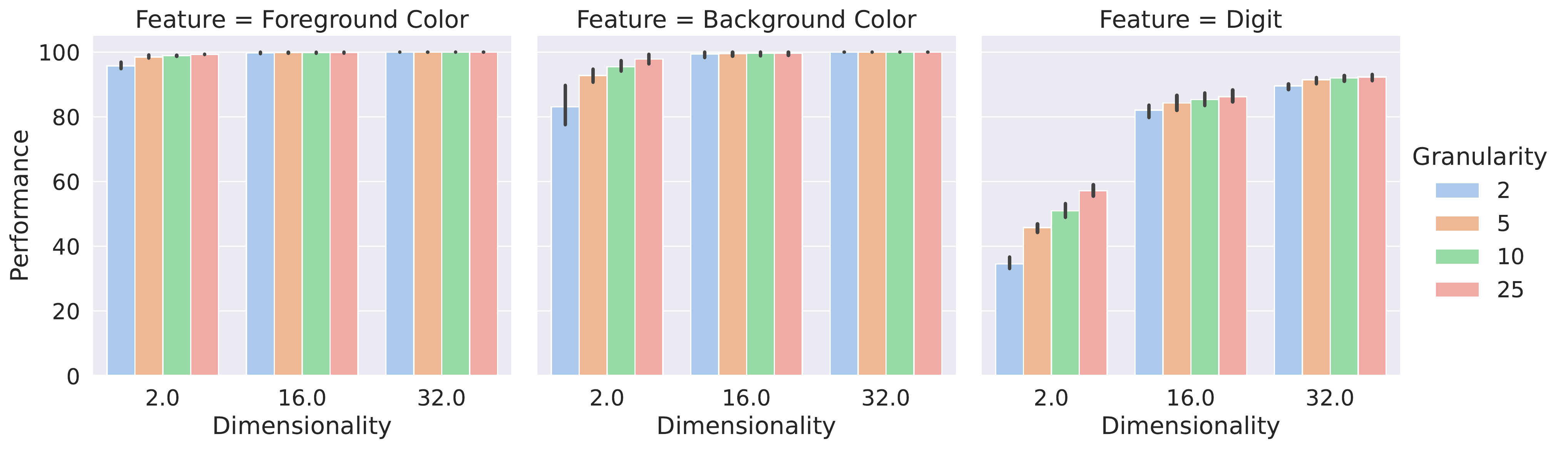}
    \subcaption{VDRL Encoder}
    \end{subfigure}
    \begin{subfigure}[c]{\linewidth}
    \includegraphics[width=\textwidth]{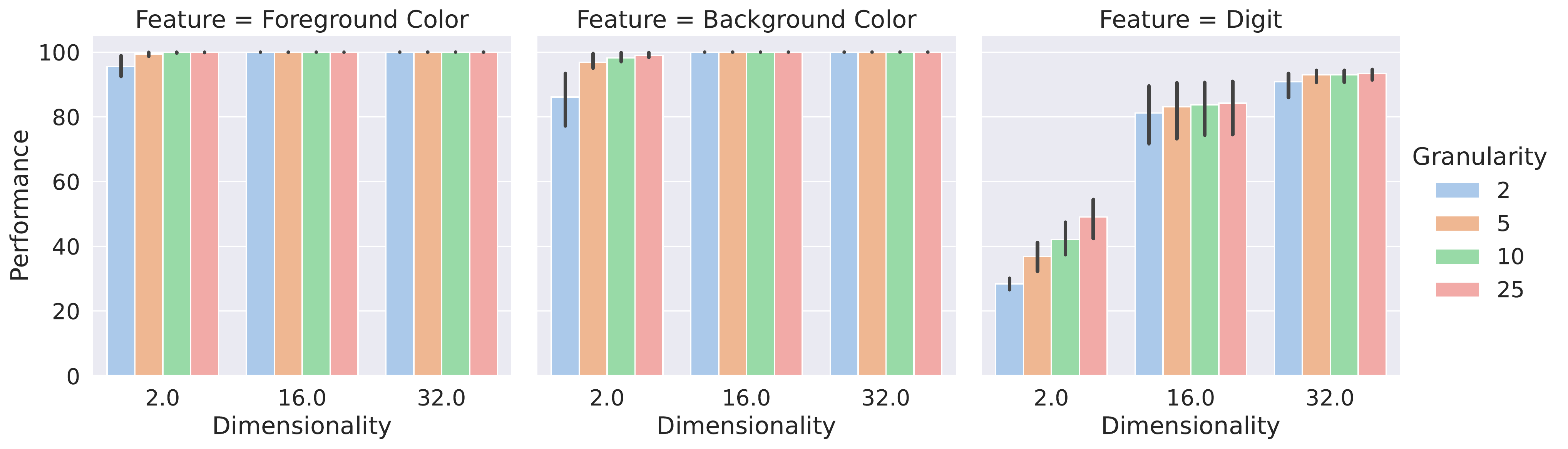}
    \subcaption{DRL Encoder}
    \end{subfigure}
    \caption{Downstream performance plots for the Colored-MNIST Dataset for different features, when the score model is trained with different latent dimensionality and the downstream models are trained with different granularities for discretization.}
    \label{fig:cm_abla}
\end{figure}
\section{Colored MNIST}
\label{apdx:cmnist}
\begin{wrapfigure}{r}{0.2\textwidth}
    \vspace{-8mm}
    \includegraphics[width=0.2\textwidth]{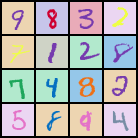}
    \caption{Samples from Colored-MNIST Dataset}
    \label{fig:cmnist_samples}
    \vspace{-10mm}
\end{wrapfigure}
We train the score model for 250,000 iterations and then the downstream models for 1500 epochs. Figure \ref{fig:cmnist_samples} shows some samples obtained from this dataset, showcasing the different features present as well as the diversity of these different features.

We additionally perform the Jensen-Shannon Divergence analysis between different features for different granularities, as well as visualize the attention score profiles for the different granularities as well. Furthermore, we do the same analysis with both the types of encoders; VDRL and DRL.

The corresponding plots for the attention score profiles are present in Figures \ref{fig:cm_VDRL_2} - \ref{fig:cm_DRL_32} for different latent space dimensionalities, different granularities and the different types of encoding schemes (VDRL and DRL). Further analysis into the performance on different features with different granularities and dimensionalities can be found in Figure \ref{fig:cm_abla}.

\section{CelebA}
\label{apdx:celeba}
\begin{figure}
\begin{minipage}[c]{0.49\textwidth}
    \includegraphics[width=\textwidth]{Plots/CelebA/JSD/probabilistic-drl/2/10/jsd.pdf}
\end{minipage}
\begin{minipage}[c]{0.49\textwidth}
    \includegraphics[width=\textwidth]{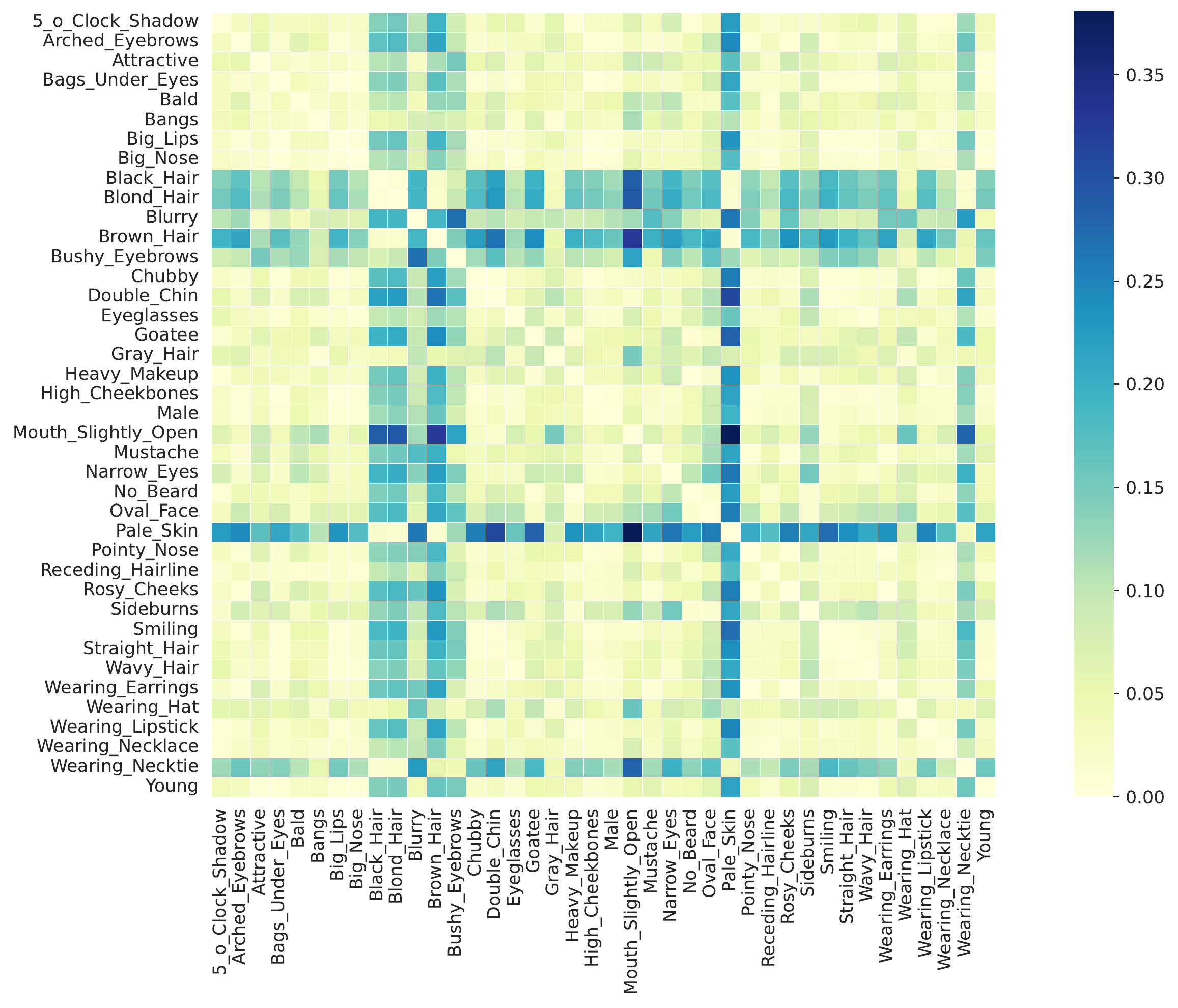}
\end{minipage}
\caption{Jensen Shannon Divergence plot for the attention profiles for any pair of features in the \textit{CelebA} dataset when using the \textit{Left:} VDRL Encoder, and \textit{Right:} DRL Encoder.}
\label{fig:celeba_ablation}
\end{figure}

\begin{figure}
\begin{minipage}[c]{0.49\textwidth}
    \includegraphics[width=\textwidth]{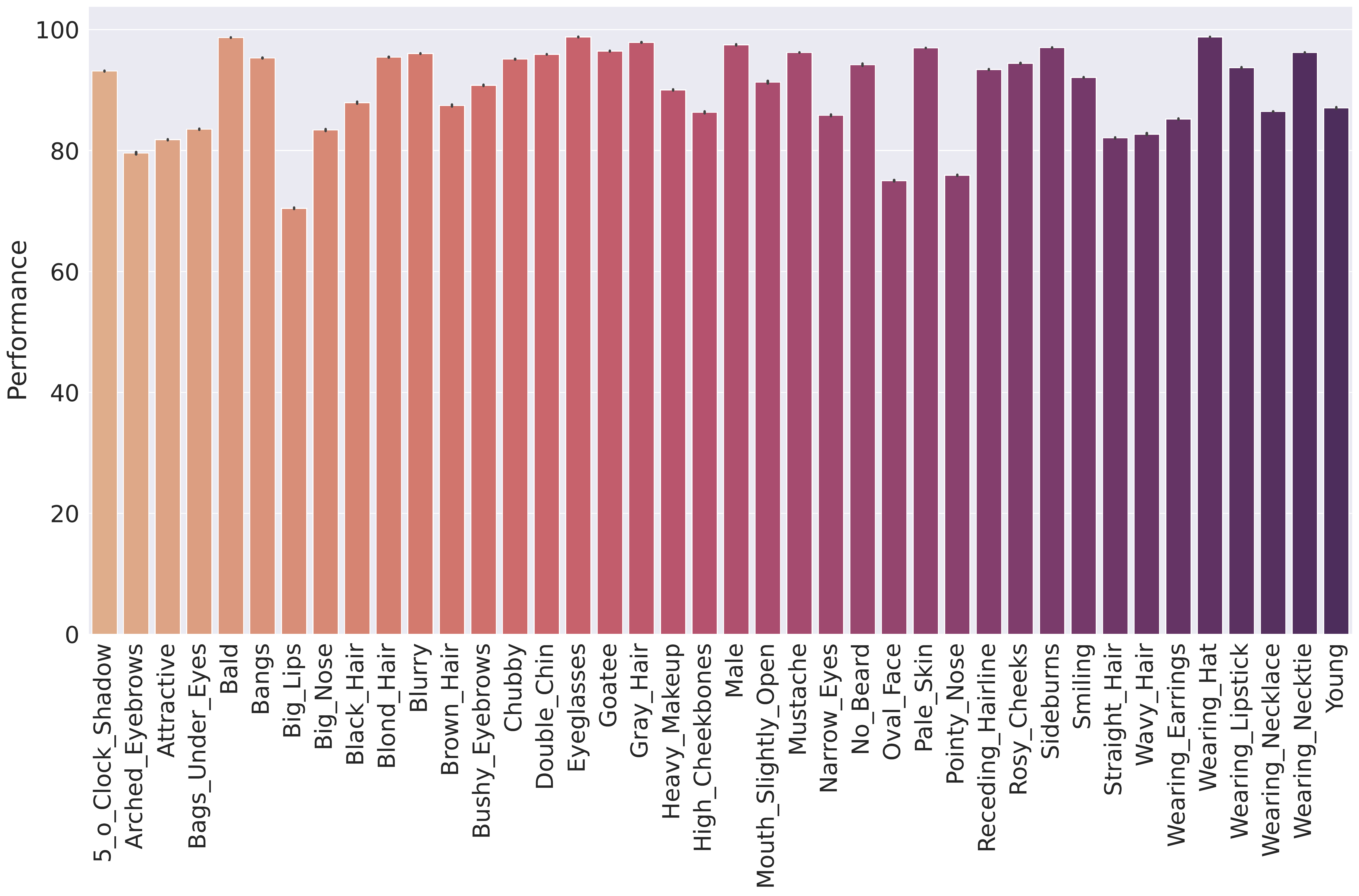}
\end{minipage}
\begin{minipage}[c]{0.49\textwidth}
    \includegraphics[width=\textwidth]{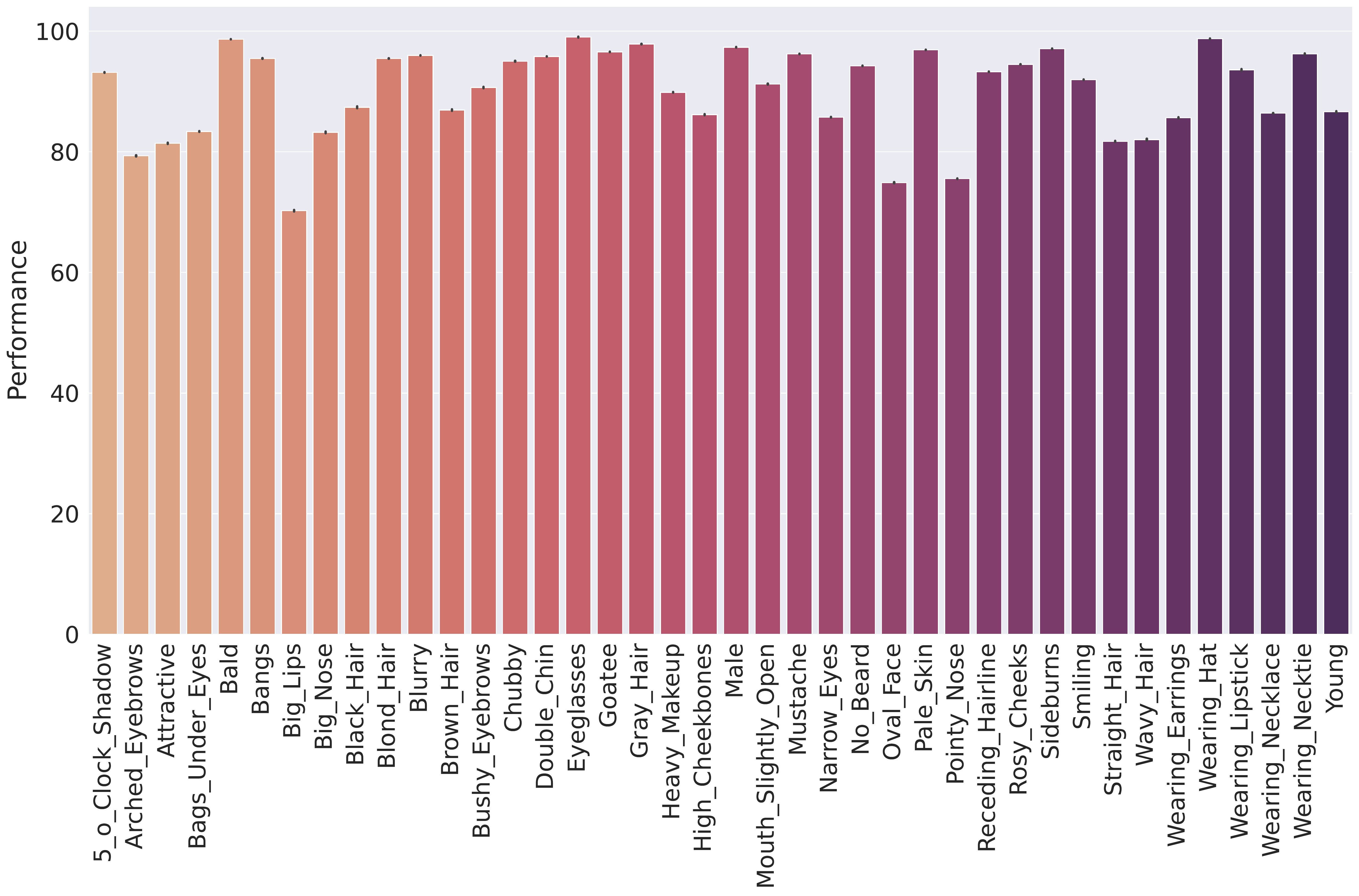}
\end{minipage}
\caption{Downstream performance plots for the different features in the \textit{CelebA} dataset when using the \textit{Left:} VDRL Encoder, and \textit{Right:} DRL Encoder.}
\label{fig:celeba_perf}
\end{figure}

\begin{figure}\begin{minipage}[c]{0.24\textwidth}
\includegraphics[width=\textwidth]{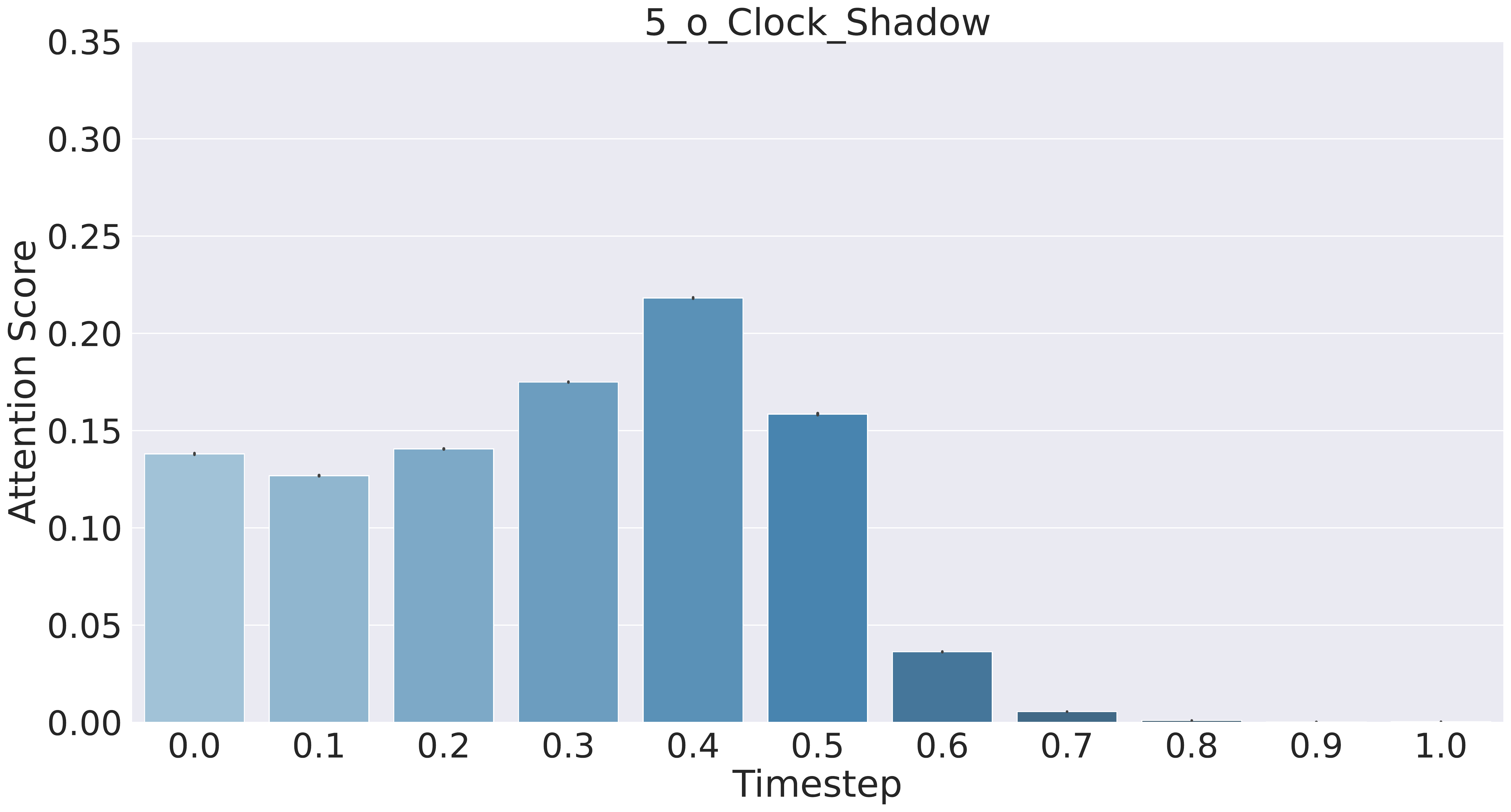}
\end{minipage}
\begin{minipage}[c]{0.24\textwidth}
\includegraphics[width=\textwidth]{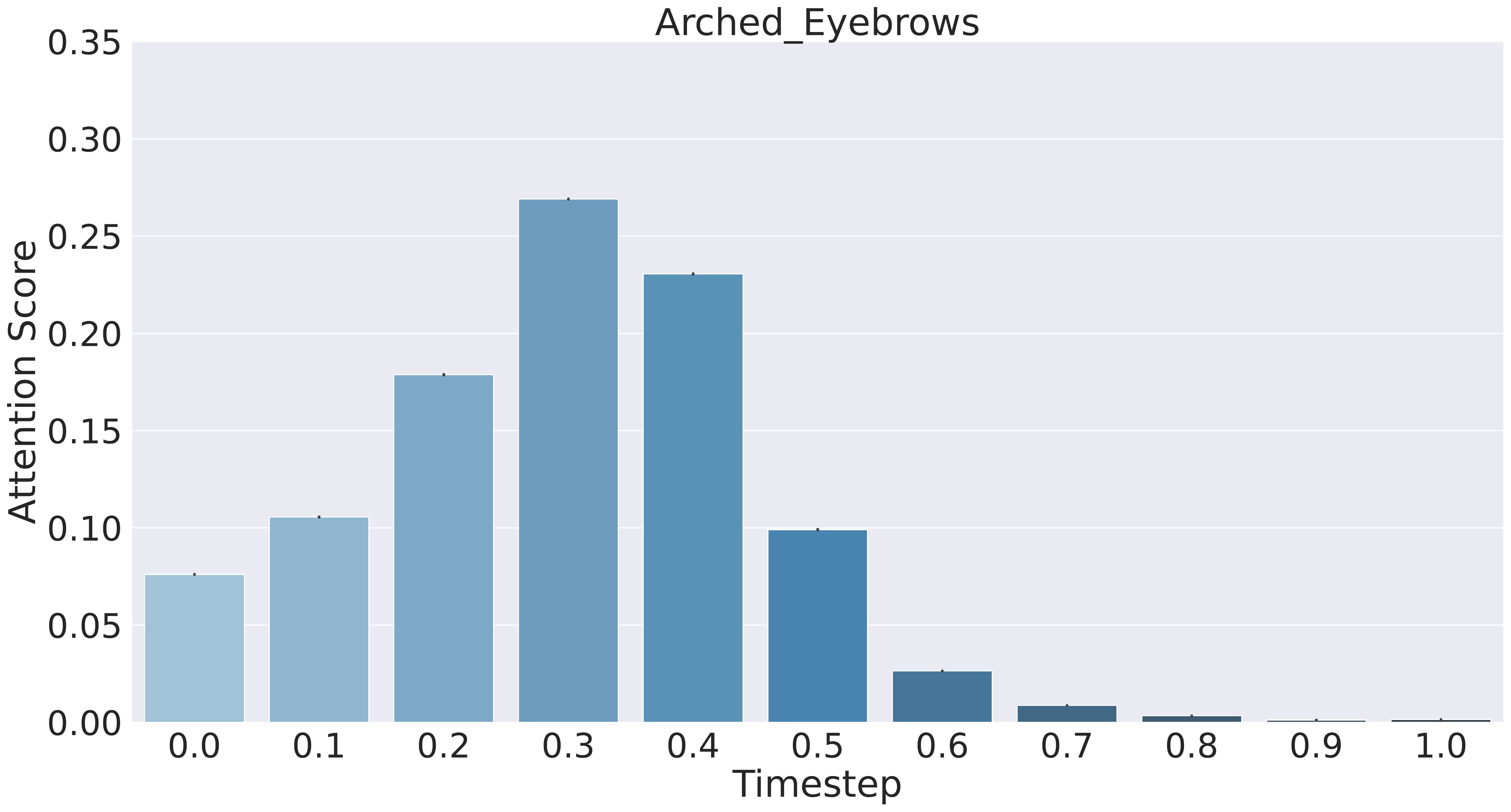}
\end{minipage}
\begin{minipage}[c]{0.24\textwidth}
\includegraphics[width=\textwidth]{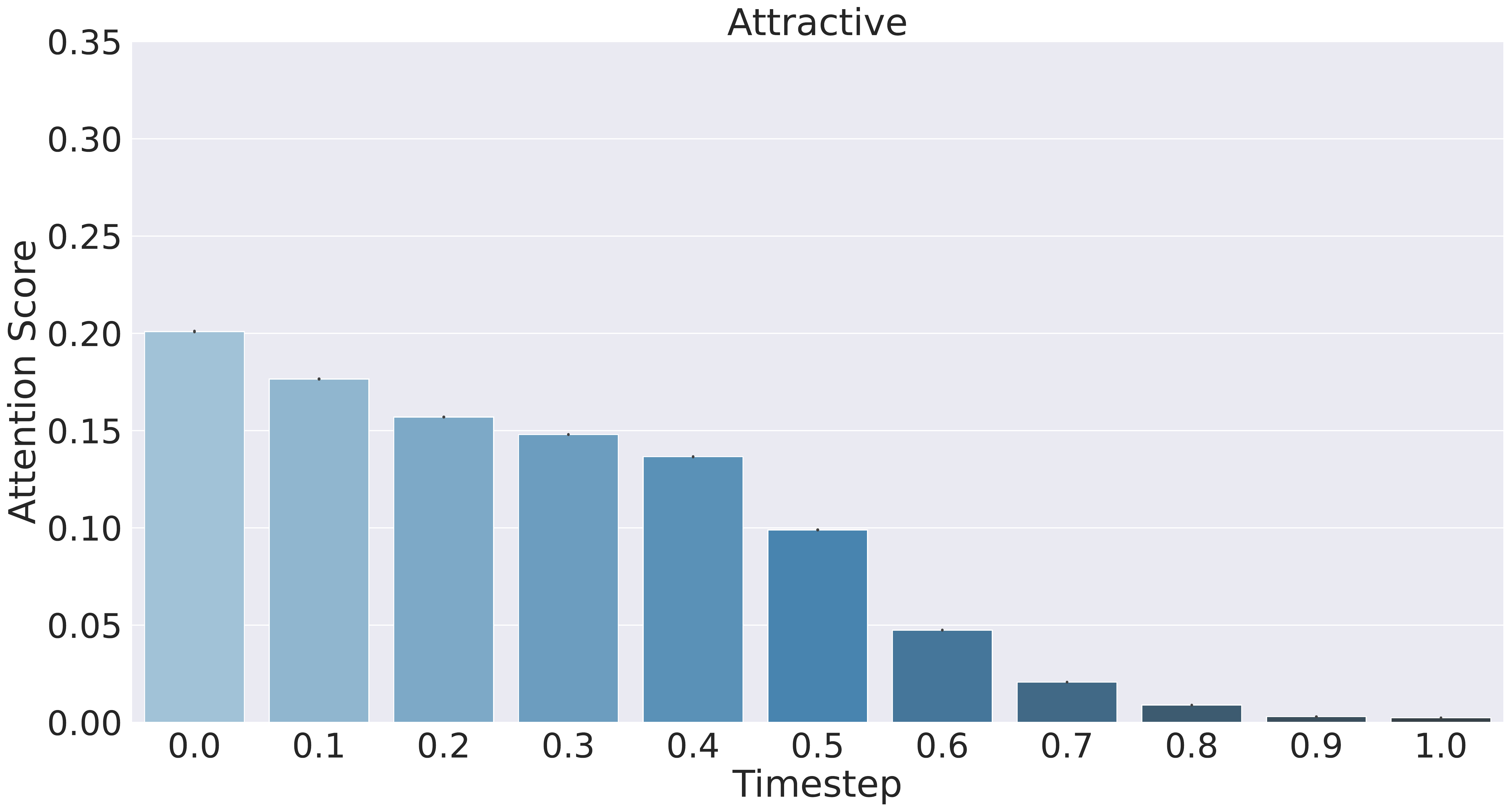}
\end{minipage}
\begin{minipage}[c]{0.24\textwidth}
\includegraphics[width=\textwidth]{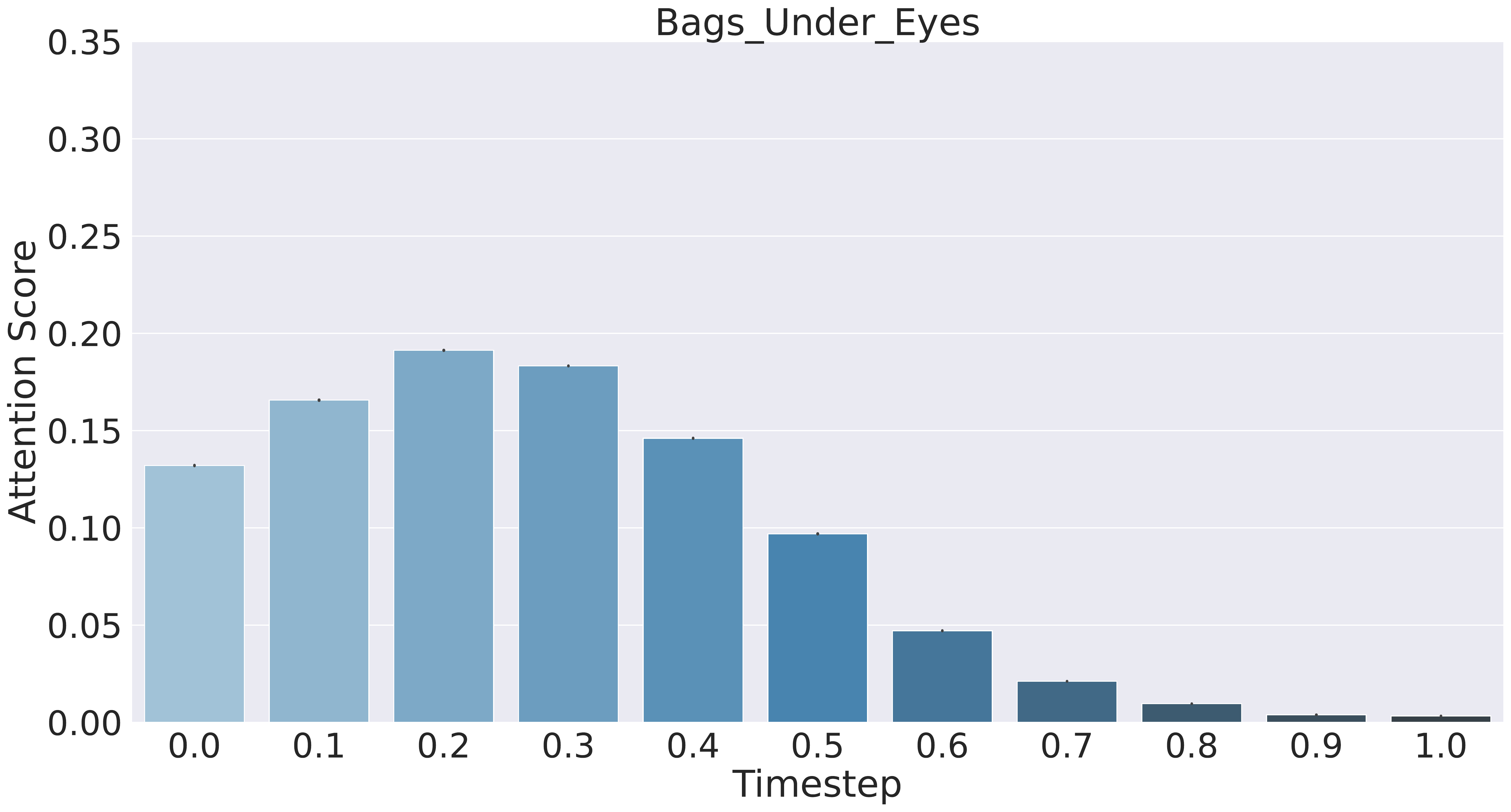}
\end{minipage}
\begin{minipage}[c]{0.24\textwidth}
\includegraphics[width=\textwidth]{Plots/CelebA/Activations/probabilistic-drl/2/10/Bald.pdf}
\end{minipage}
\begin{minipage}[c]{0.24\textwidth}
\includegraphics[width=\textwidth]{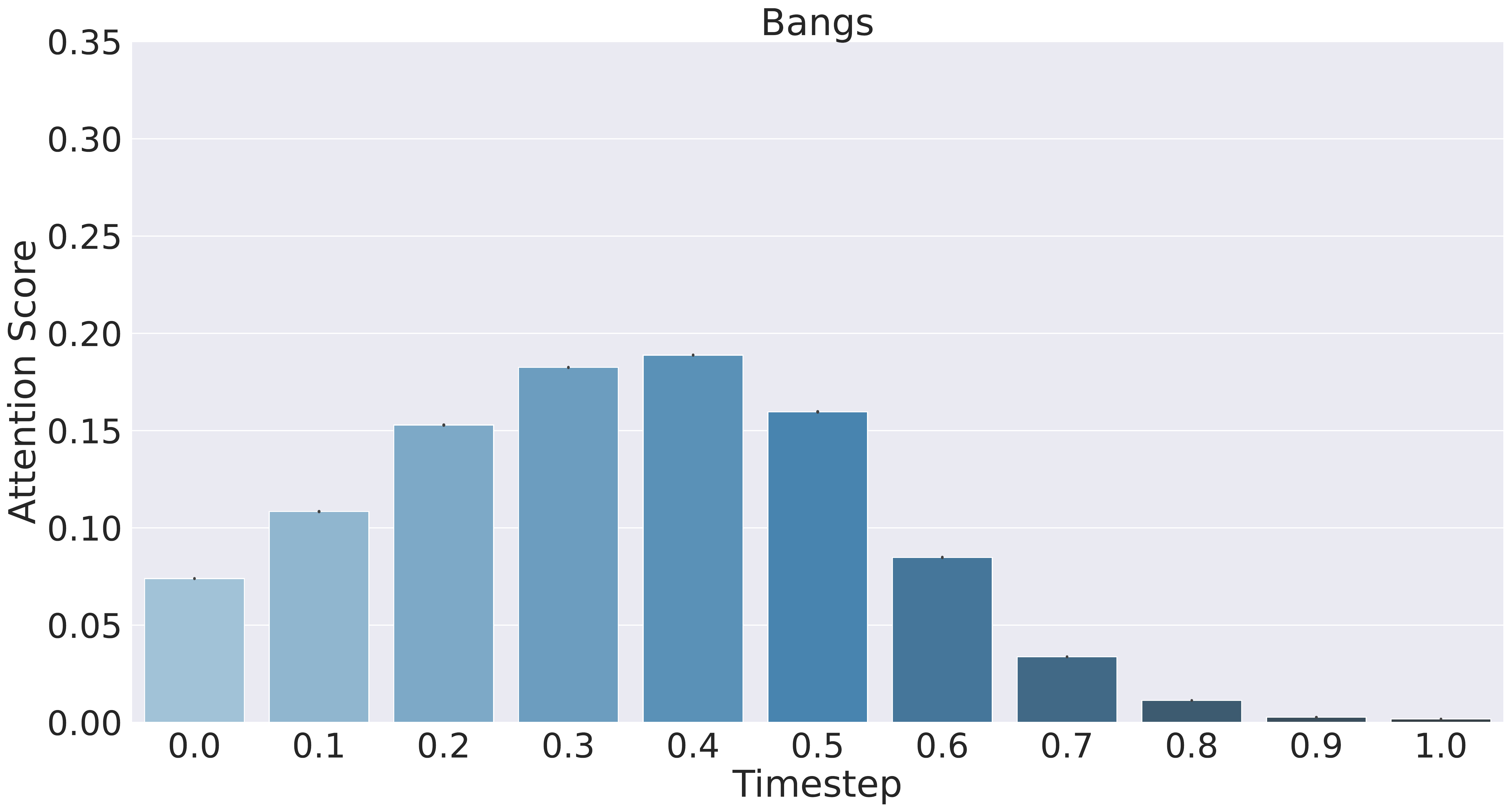}
\end{minipage}
\begin{minipage}[c]{0.24\textwidth}
\includegraphics[width=\textwidth]{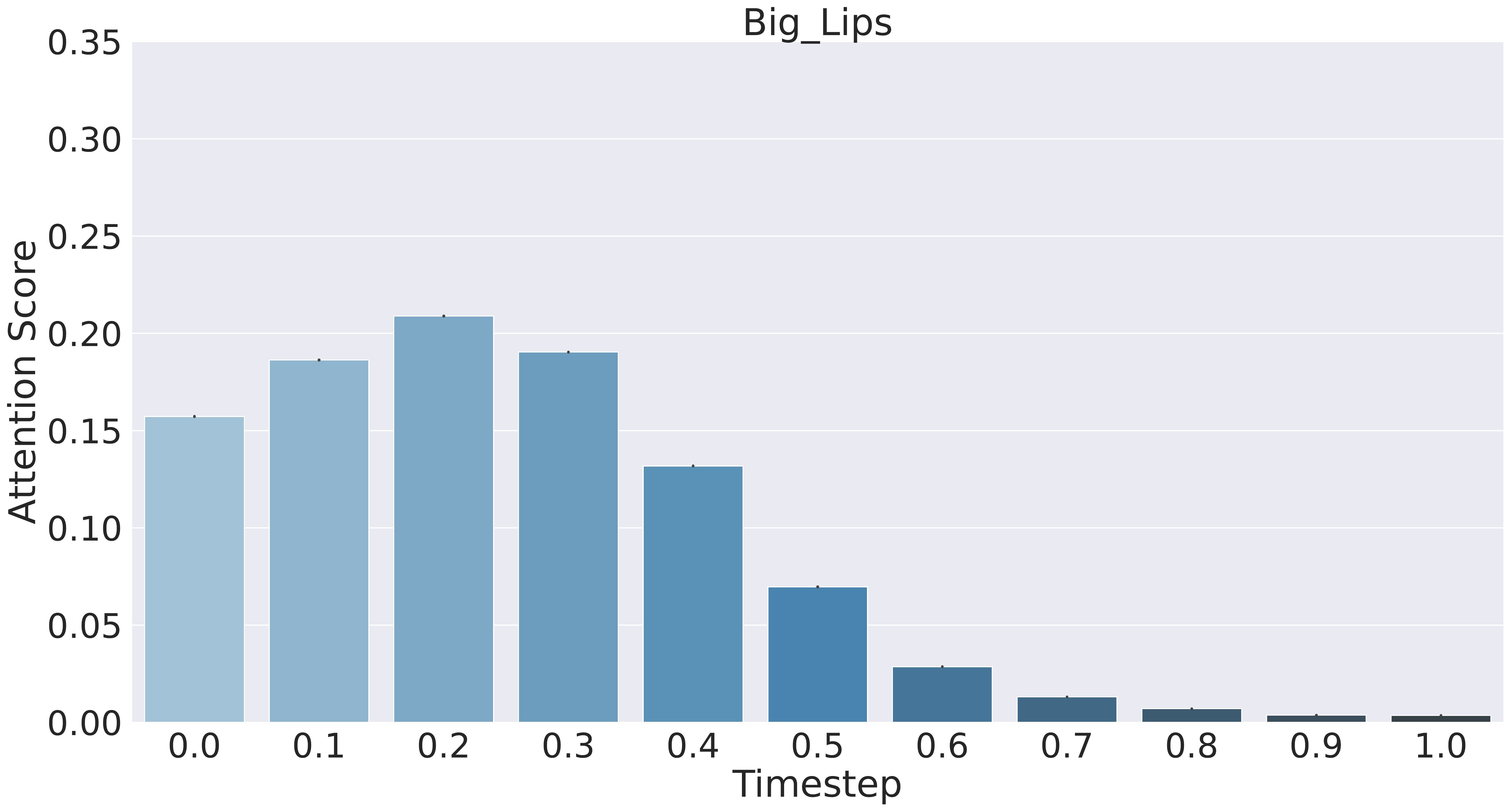}
\end{minipage}
\begin{minipage}[c]{0.24\textwidth}
\includegraphics[width=\textwidth]{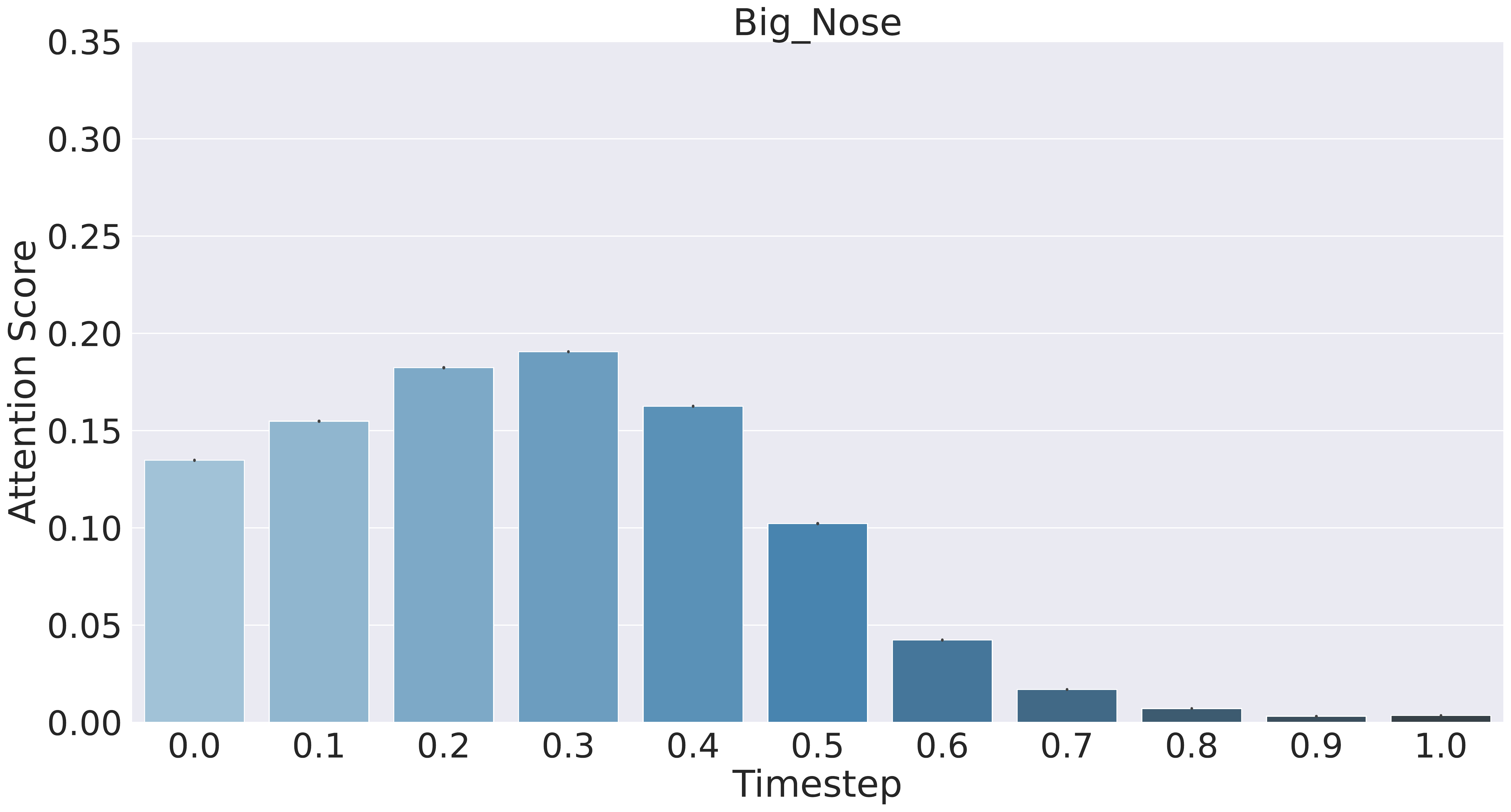}
\end{minipage}
\begin{minipage}[c]{0.24\textwidth}
\includegraphics[width=\textwidth]{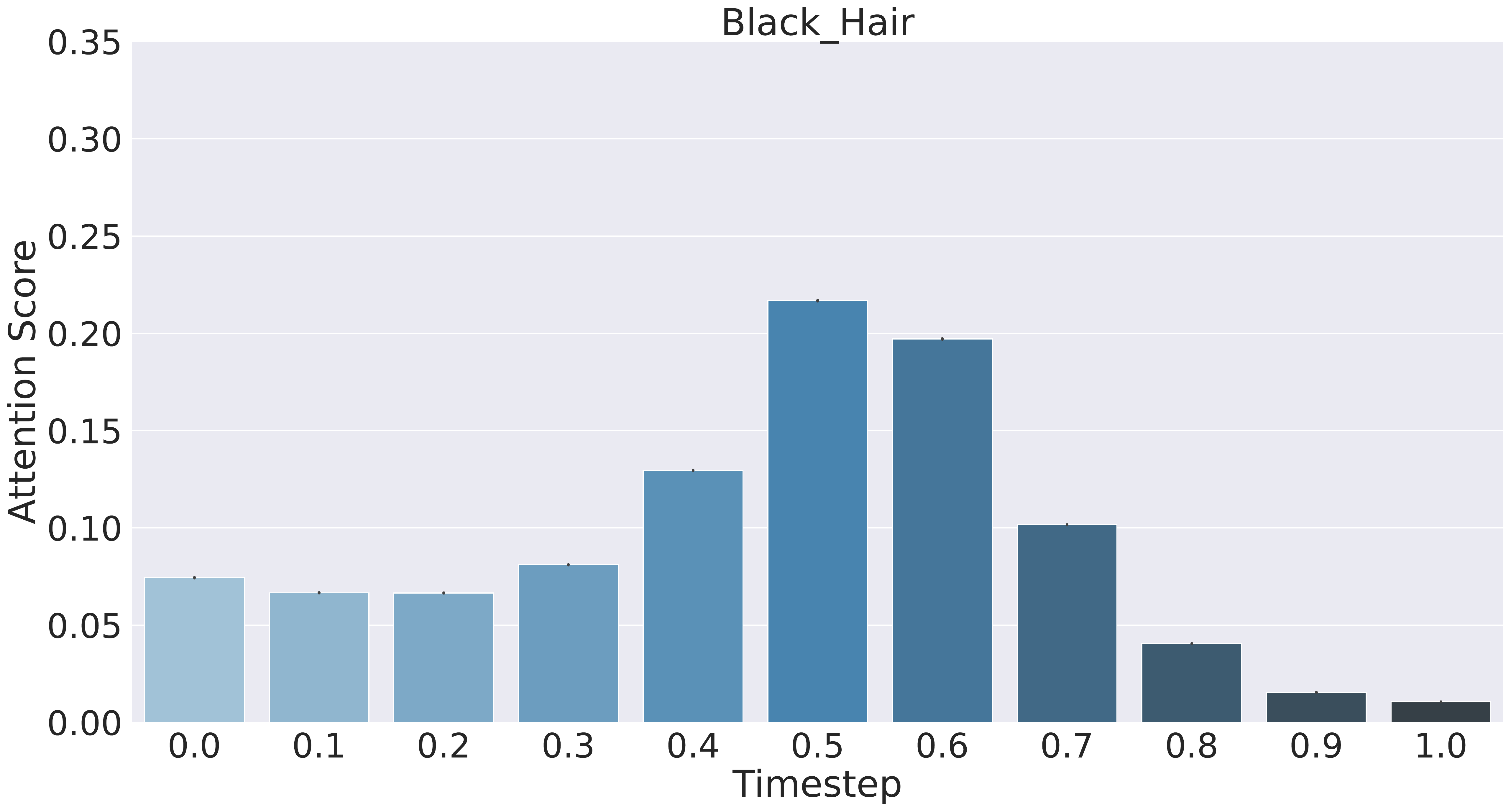}
\end{minipage}
\begin{minipage}[c]{0.24\textwidth}
\includegraphics[width=\textwidth]{Plots/CelebA/Activations/probabilistic-drl/2/10/Blond_Hair.pdf}
\end{minipage}
\begin{minipage}[c]{0.24\textwidth}
\includegraphics[width=\textwidth]{Plots/CelebA/Activations/probabilistic-drl/2/10/Blurry.pdf}
\end{minipage}
\begin{minipage}[c]{0.24\textwidth}
\includegraphics[width=\textwidth]{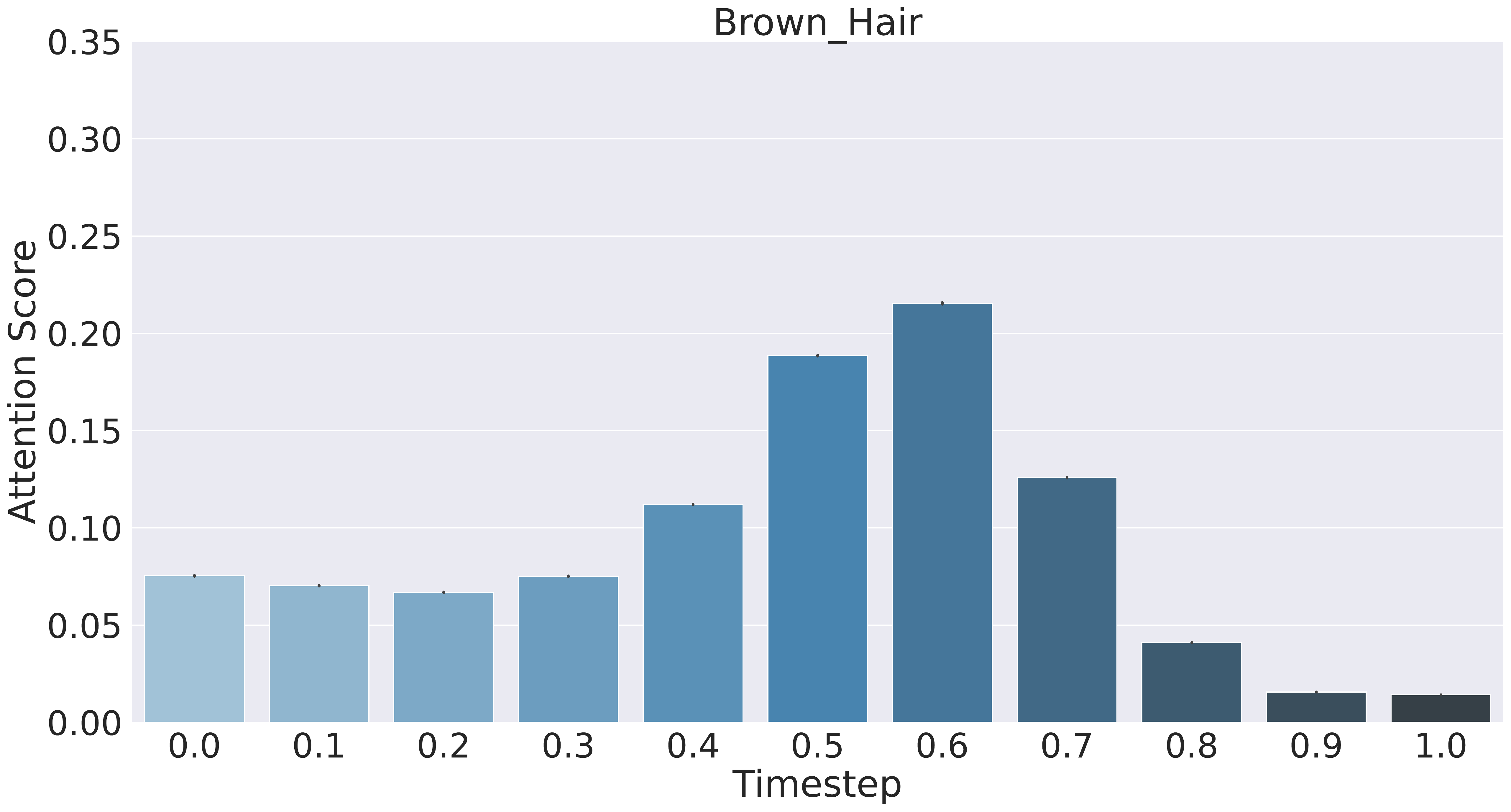}
\end{minipage}
\begin{minipage}[c]{0.24\textwidth}
\includegraphics[width=\textwidth]{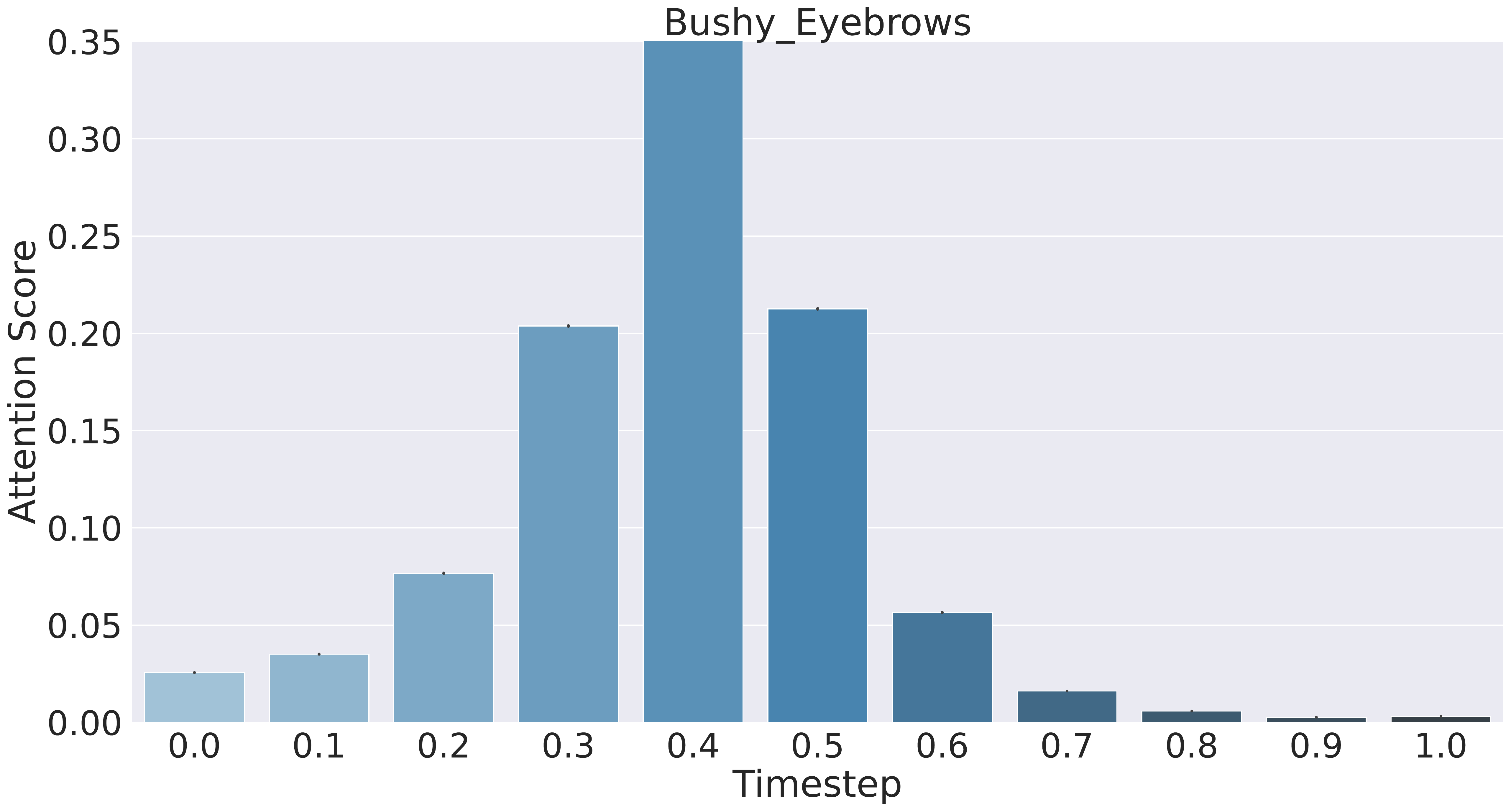}
\end{minipage}
\begin{minipage}[c]{0.24\textwidth}
\includegraphics[width=\textwidth]{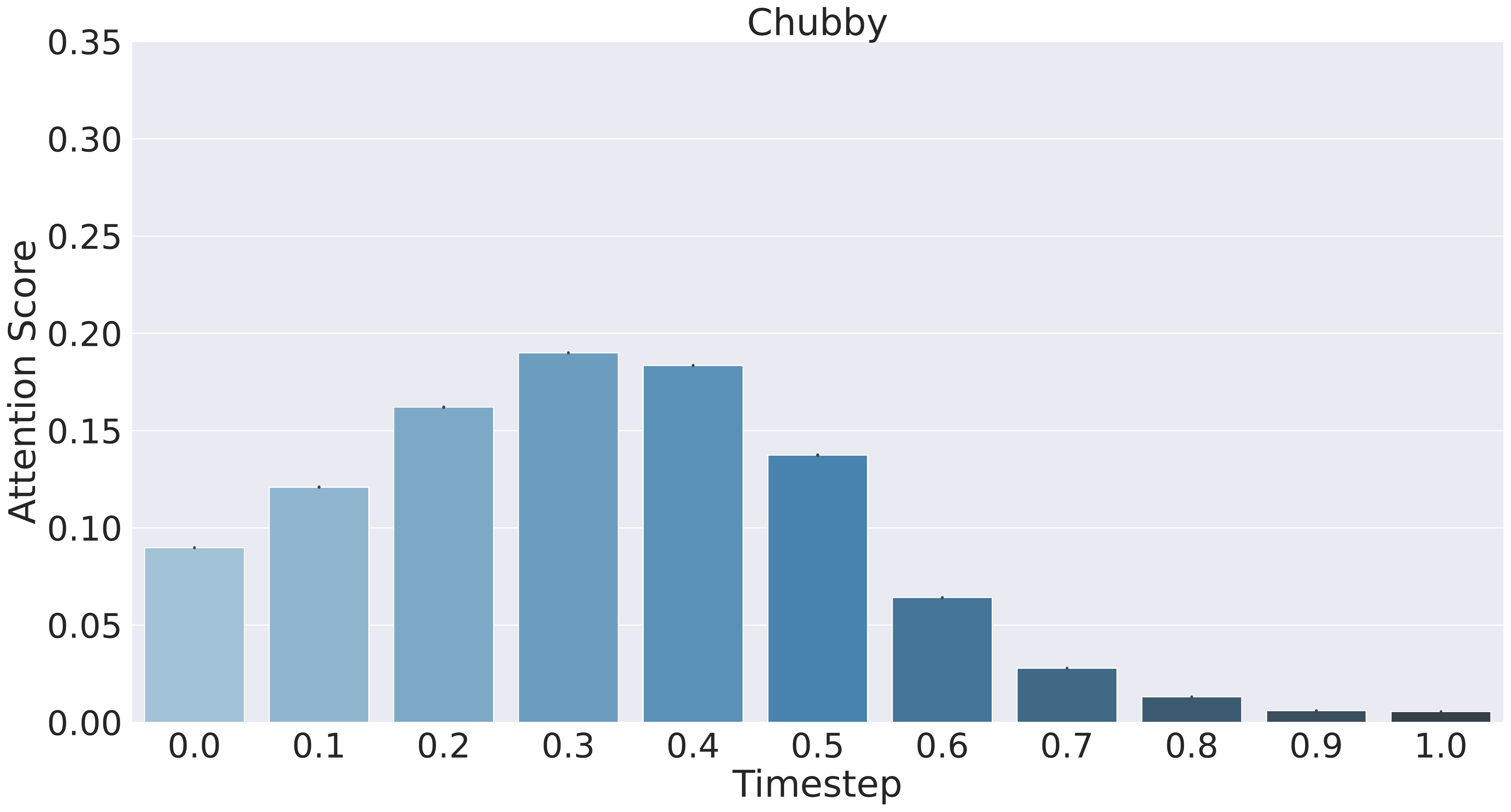}
\end{minipage}
\begin{minipage}[c]{0.24\textwidth}
\includegraphics[width=\textwidth]{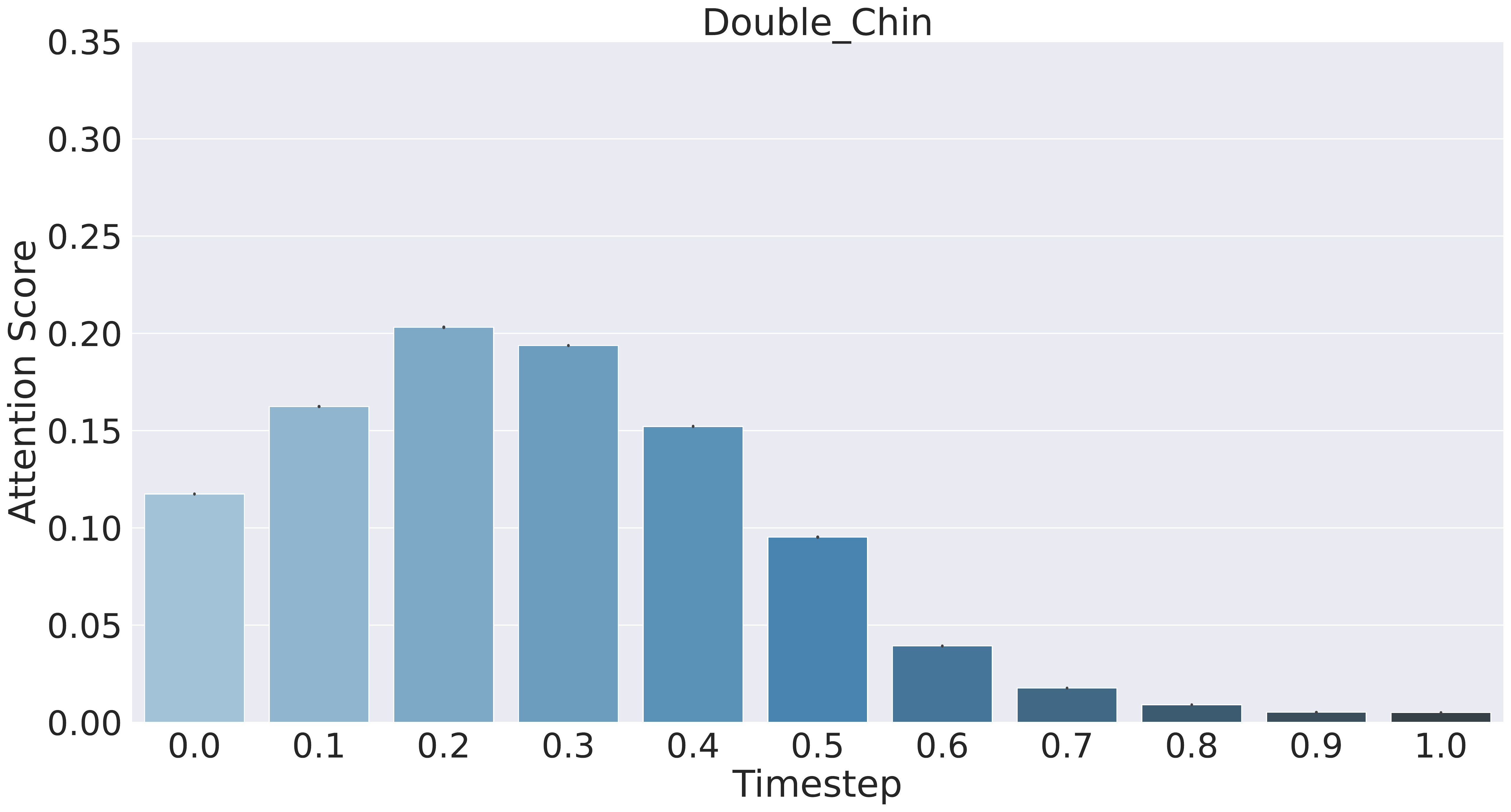}
\end{minipage}
\begin{minipage}[c]{0.24\textwidth}
\includegraphics[width=\textwidth]{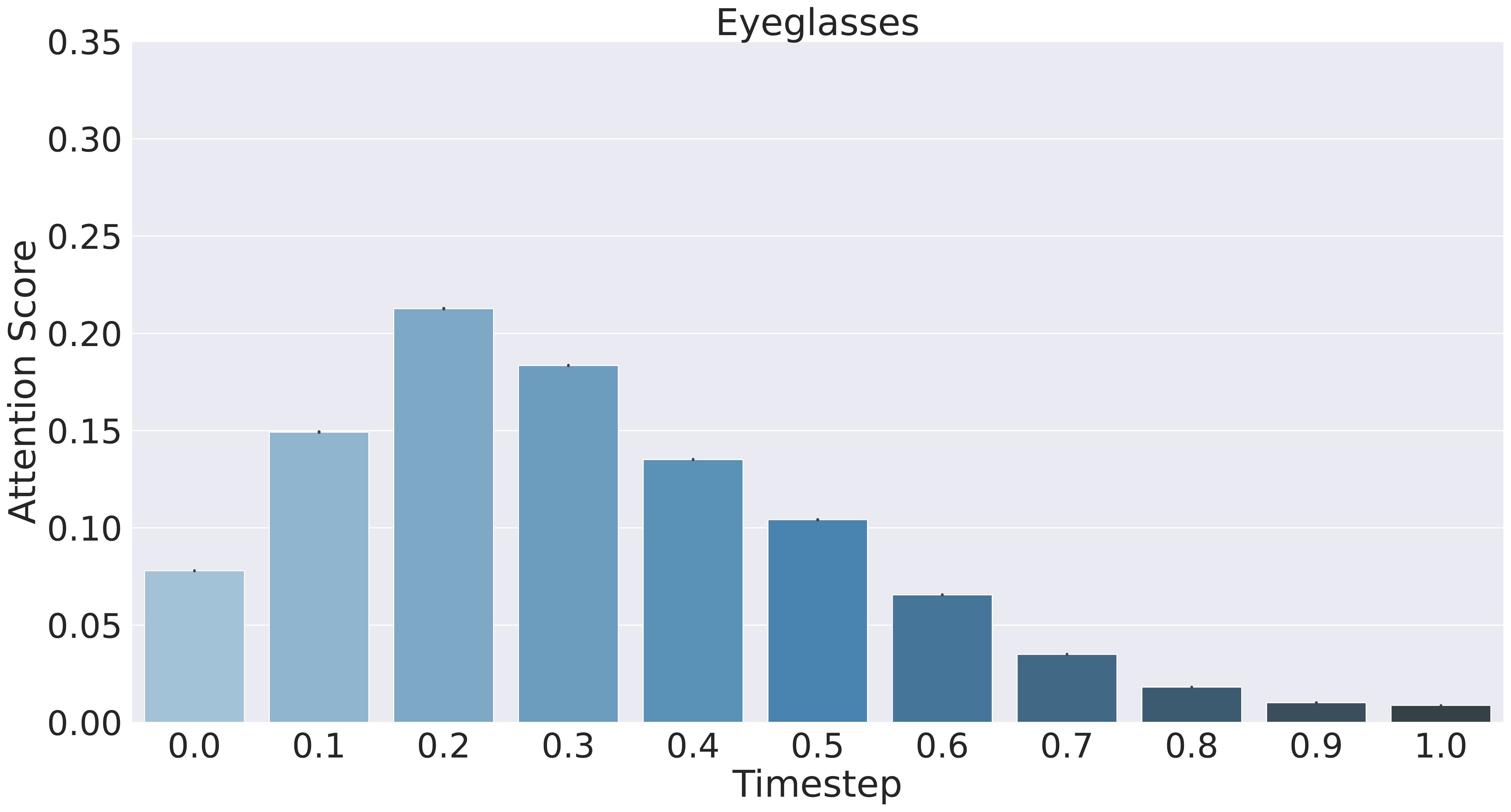}
\end{minipage}
\begin{minipage}[c]{0.24\textwidth}
\includegraphics[width=\textwidth]{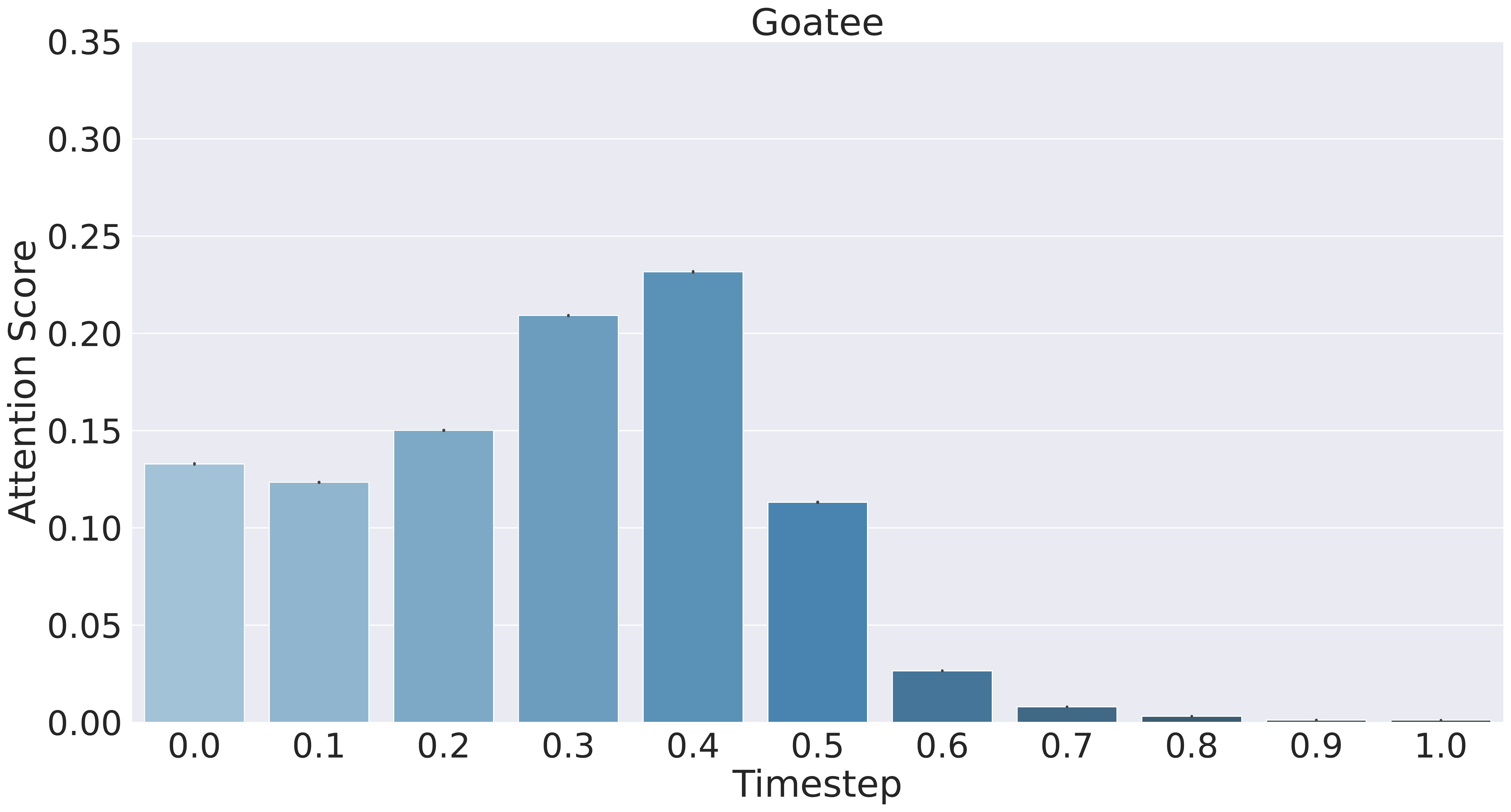}
\end{minipage}
\begin{minipage}[c]{0.24\textwidth}
\includegraphics[width=\textwidth]{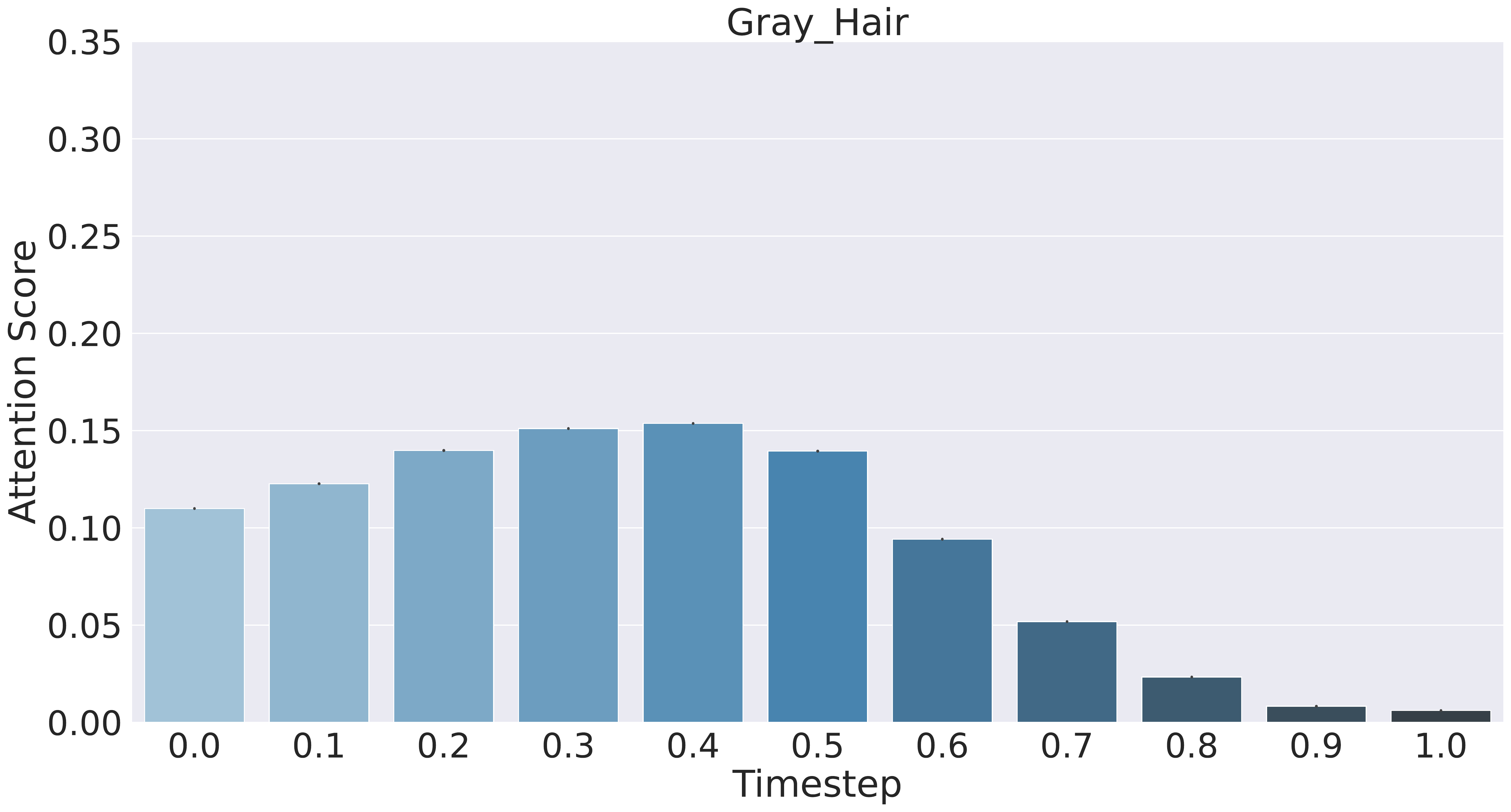}
\end{minipage}
\begin{minipage}[c]{0.24\textwidth}
\includegraphics[width=\textwidth]{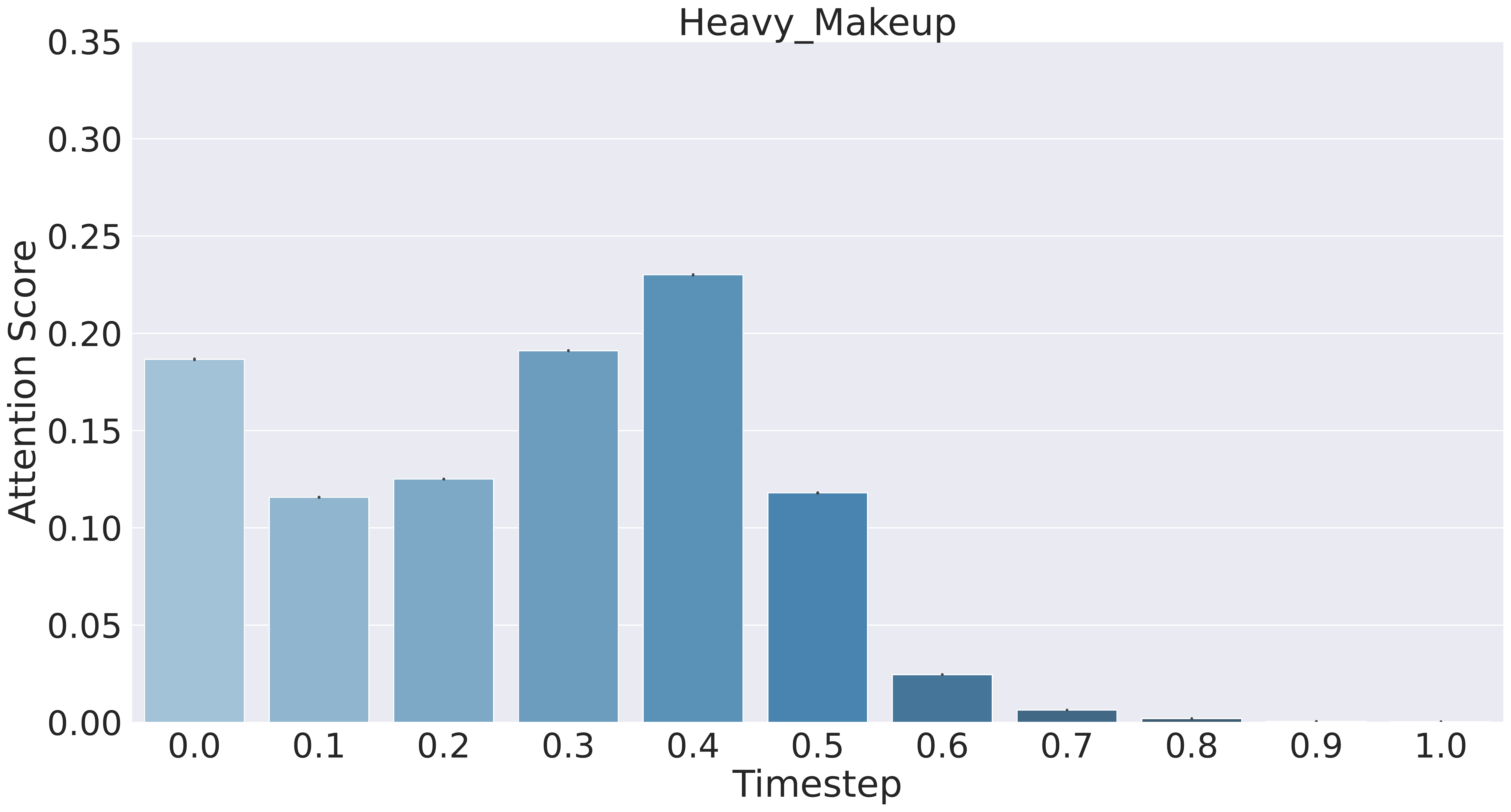}
\end{minipage}
\begin{minipage}[c]{0.24\textwidth}
\includegraphics[width=\textwidth]{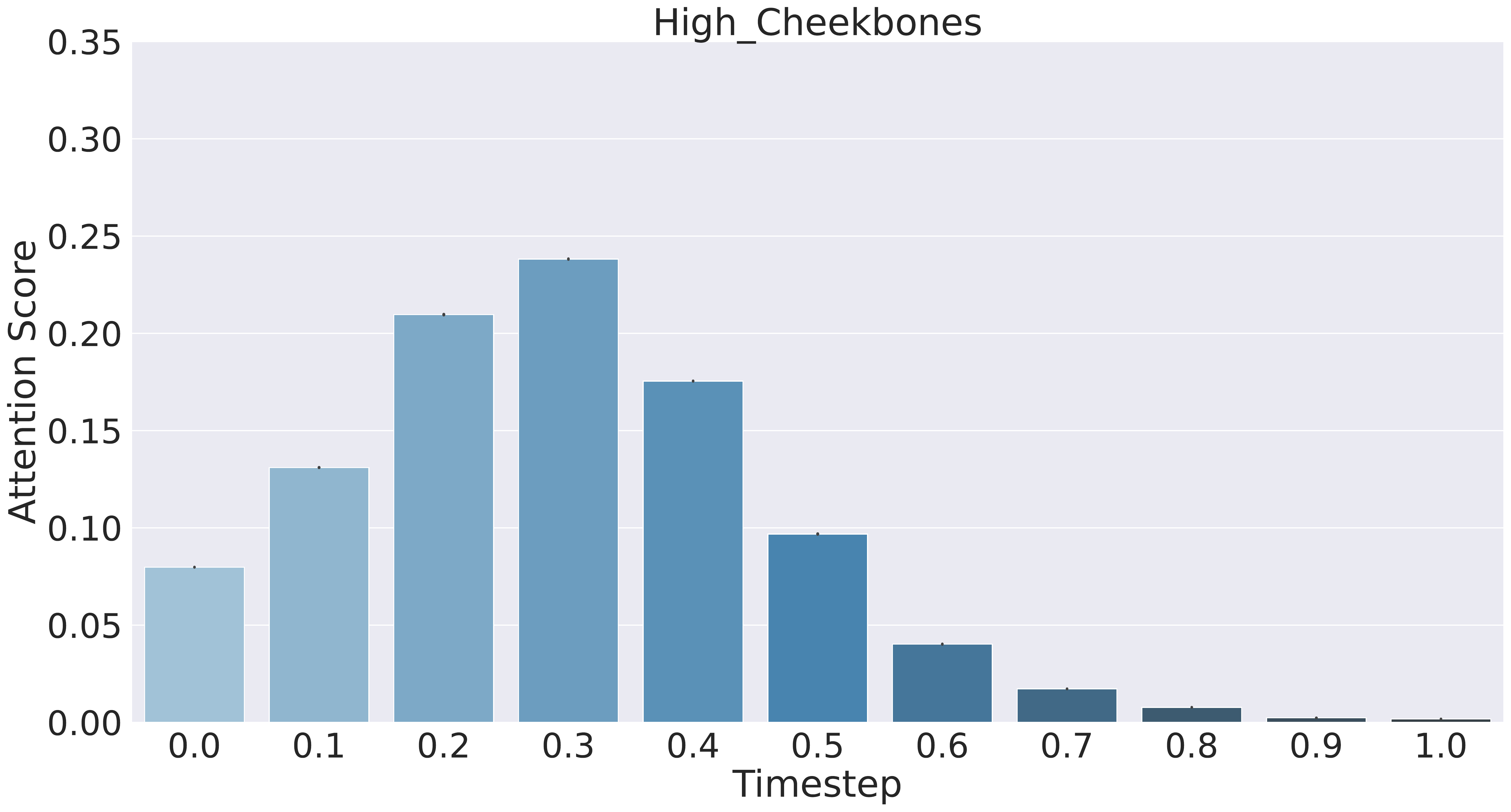}
\end{minipage}
\begin{minipage}[c]{0.24\textwidth}
\includegraphics[width=\textwidth]{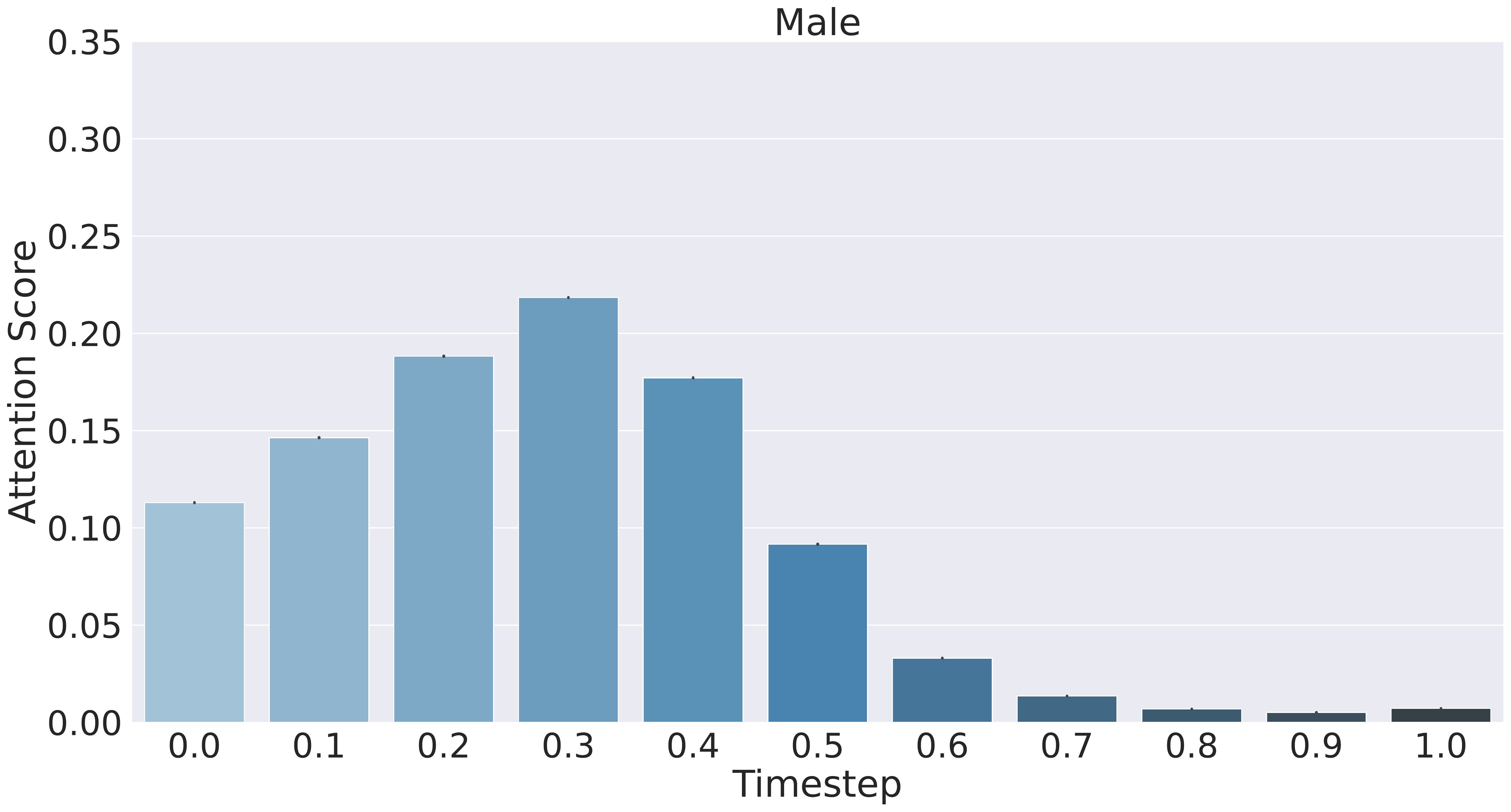}
\end{minipage}
\begin{minipage}[c]{0.24\textwidth}
\includegraphics[width=\textwidth]{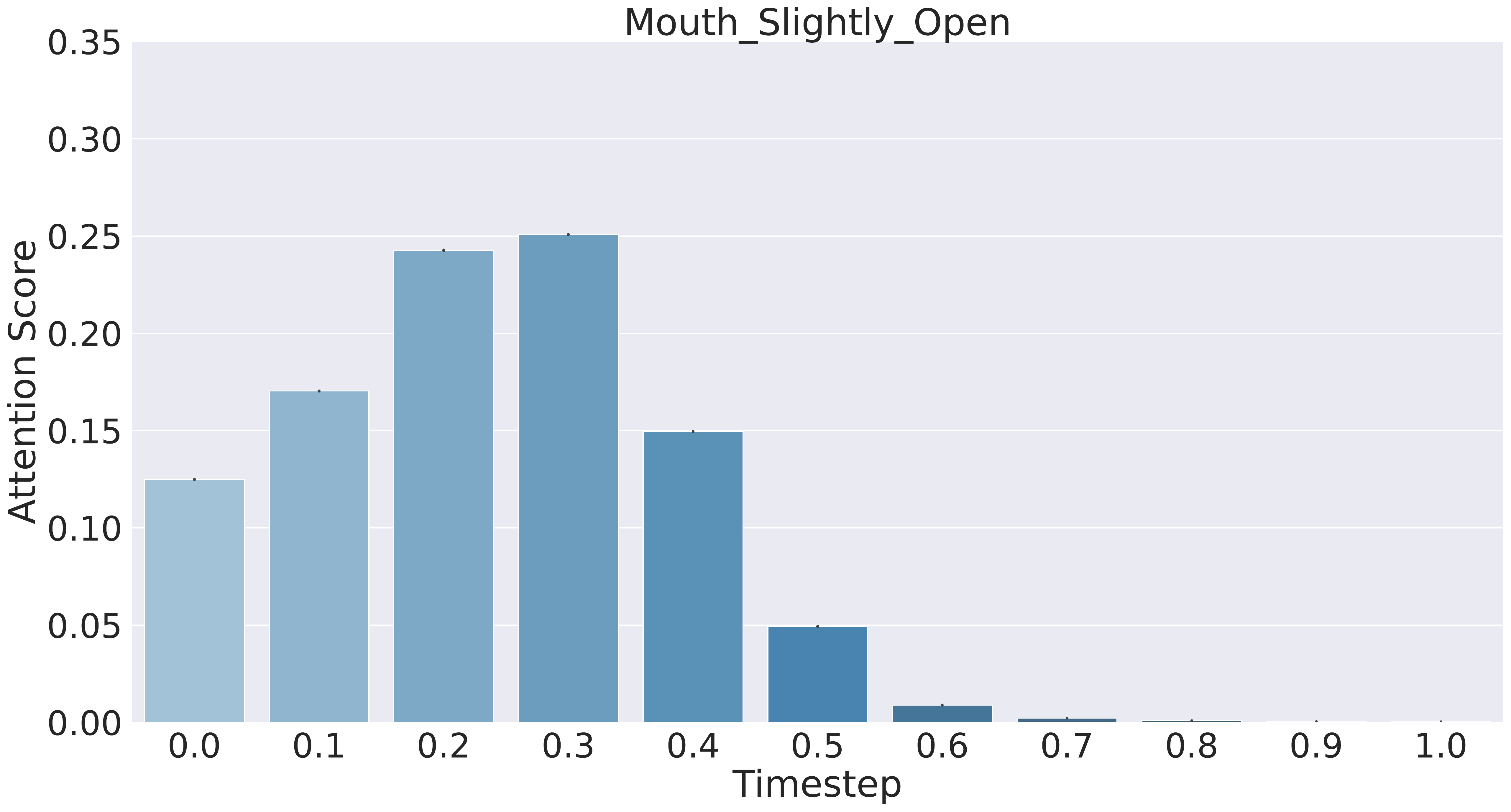}
\end{minipage}
\begin{minipage}[c]{0.24\textwidth}
\includegraphics[width=\textwidth]{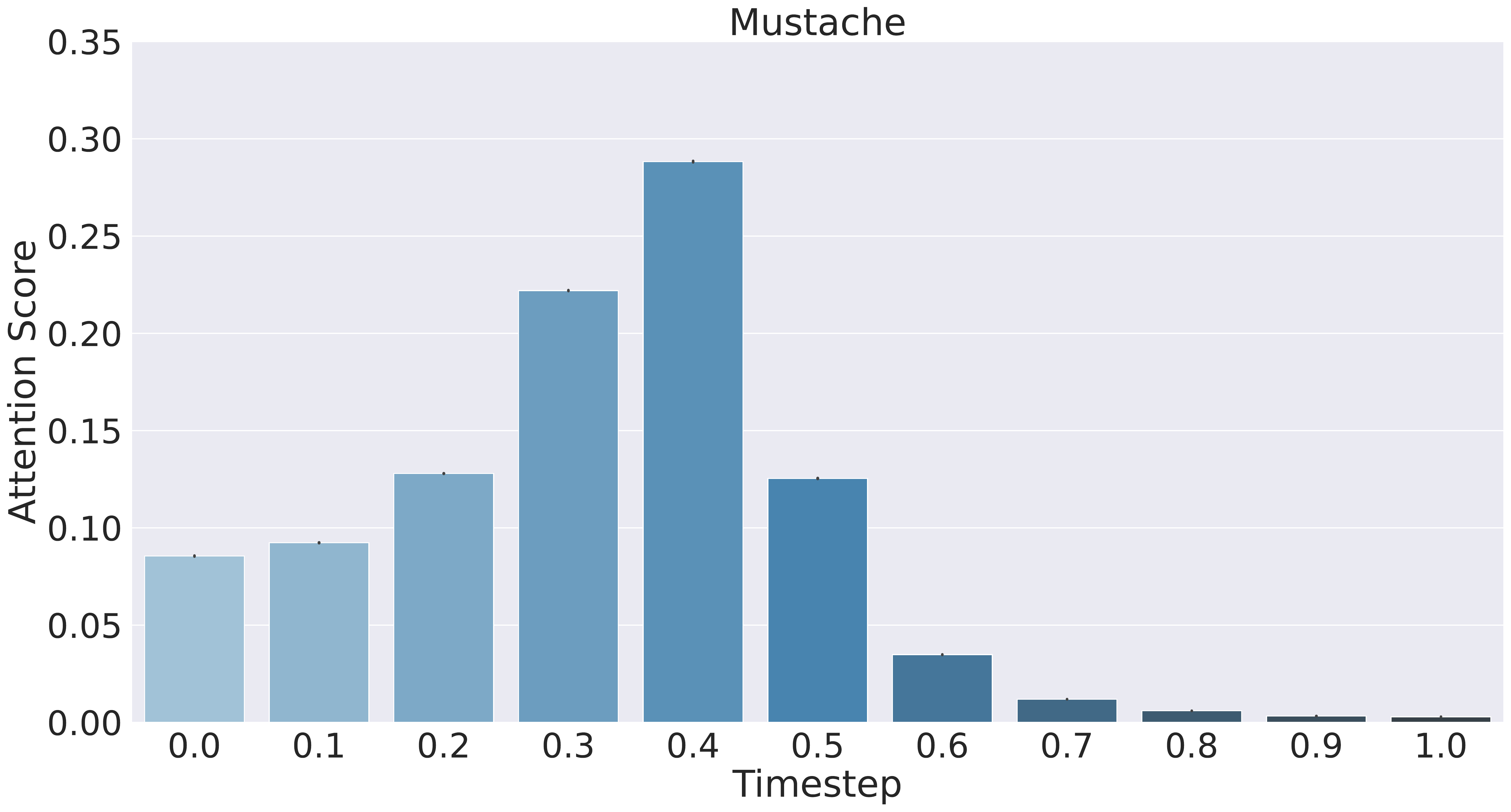}
\end{minipage}
\begin{minipage}[c]{0.24\textwidth}
\includegraphics[width=\textwidth]{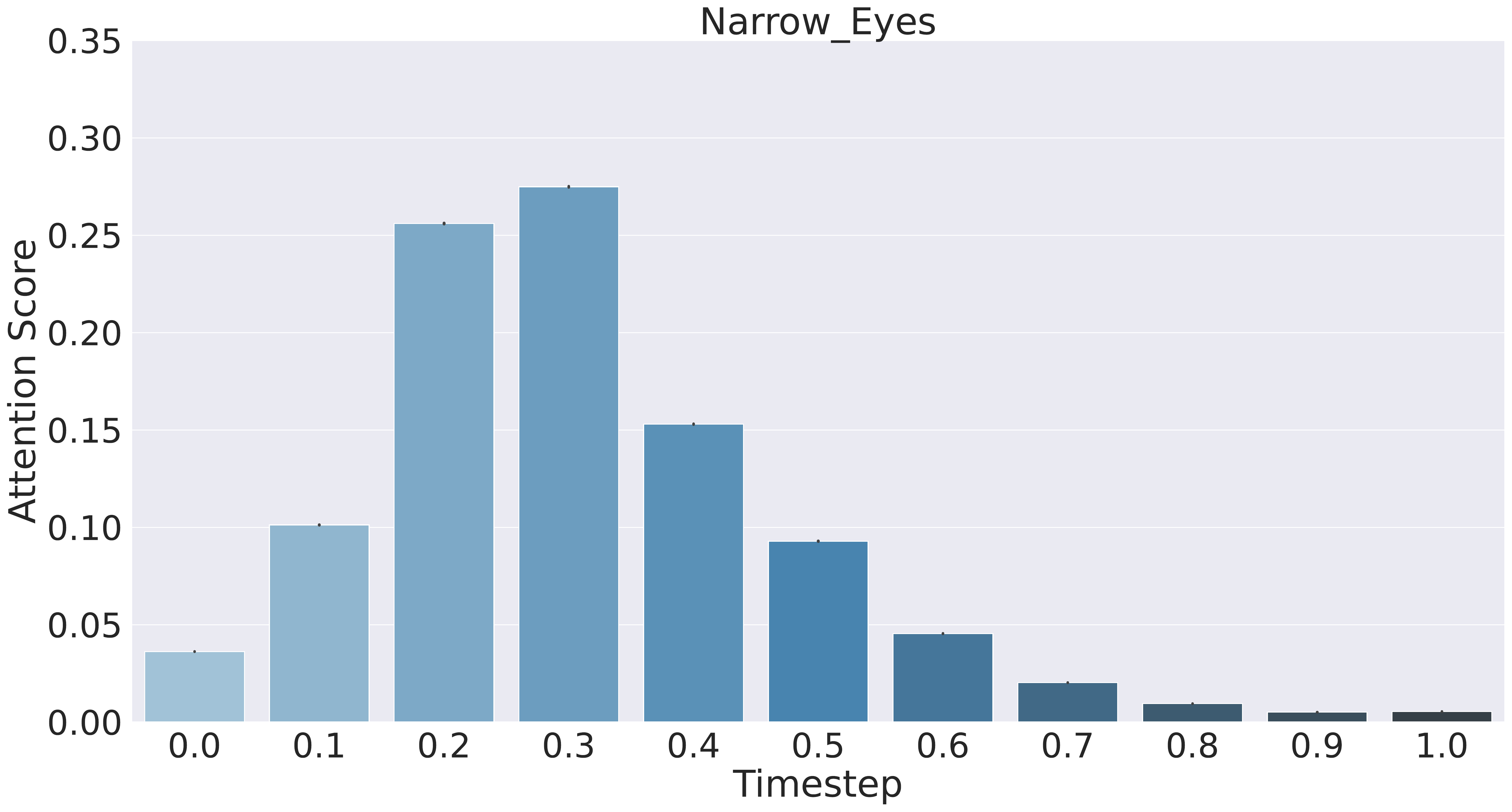}
\end{minipage}
\begin{minipage}[c]{0.24\textwidth}
\includegraphics[width=\textwidth]{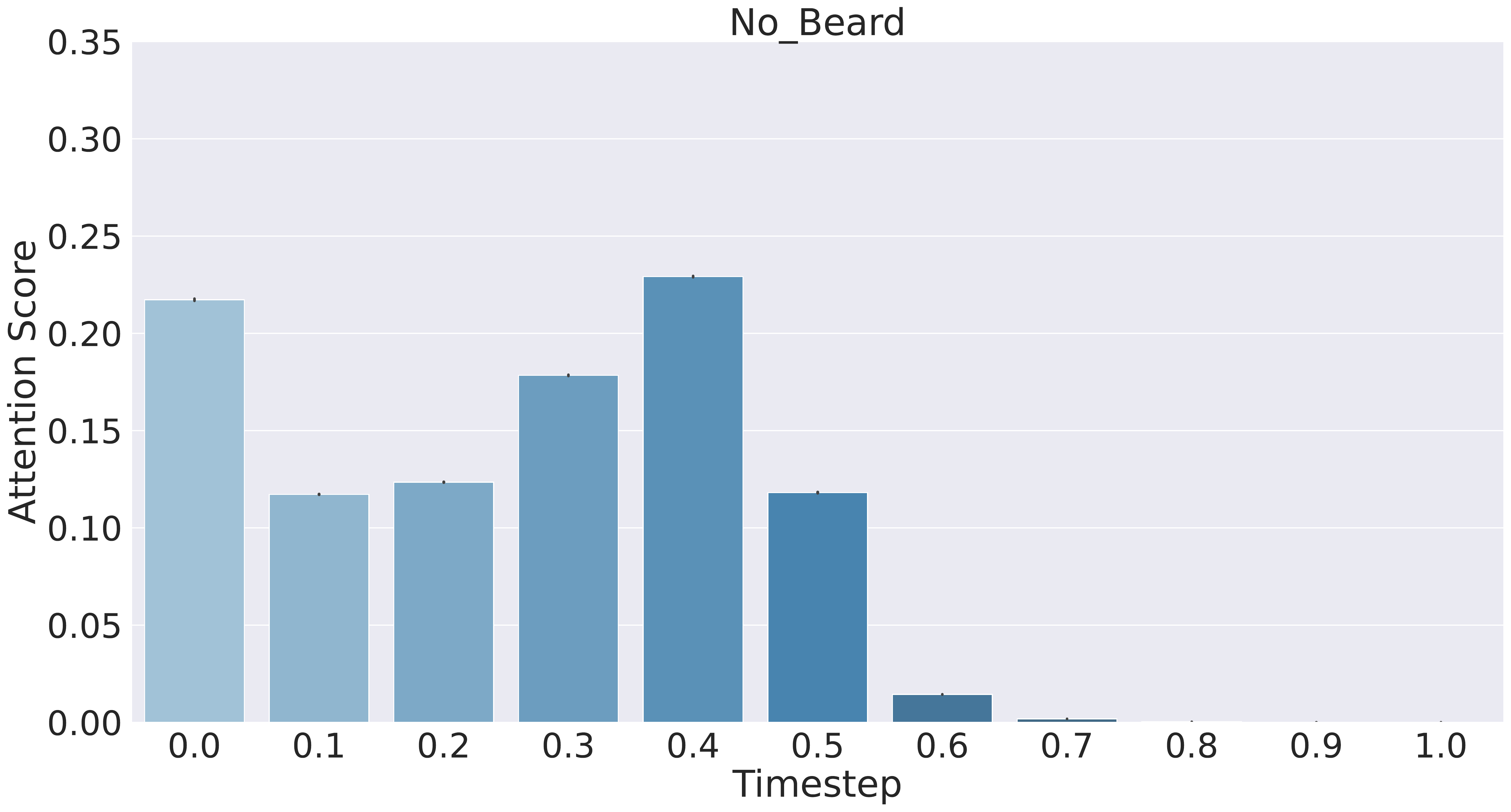}
\end{minipage}
\begin{minipage}[c]{0.24\textwidth}
\includegraphics[width=\textwidth]{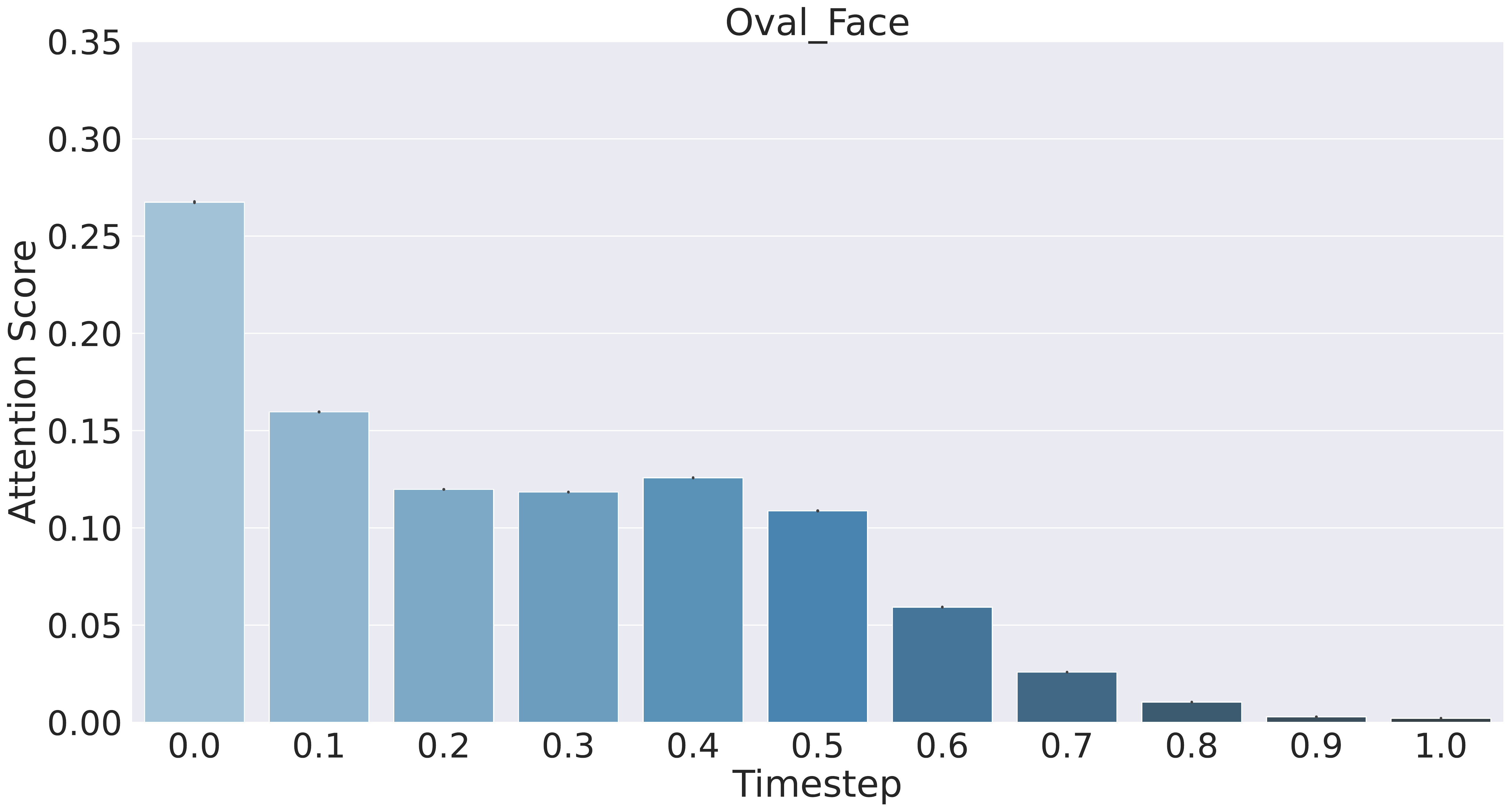}
\end{minipage}
\begin{minipage}[c]{0.24\textwidth}
\includegraphics[width=\textwidth]{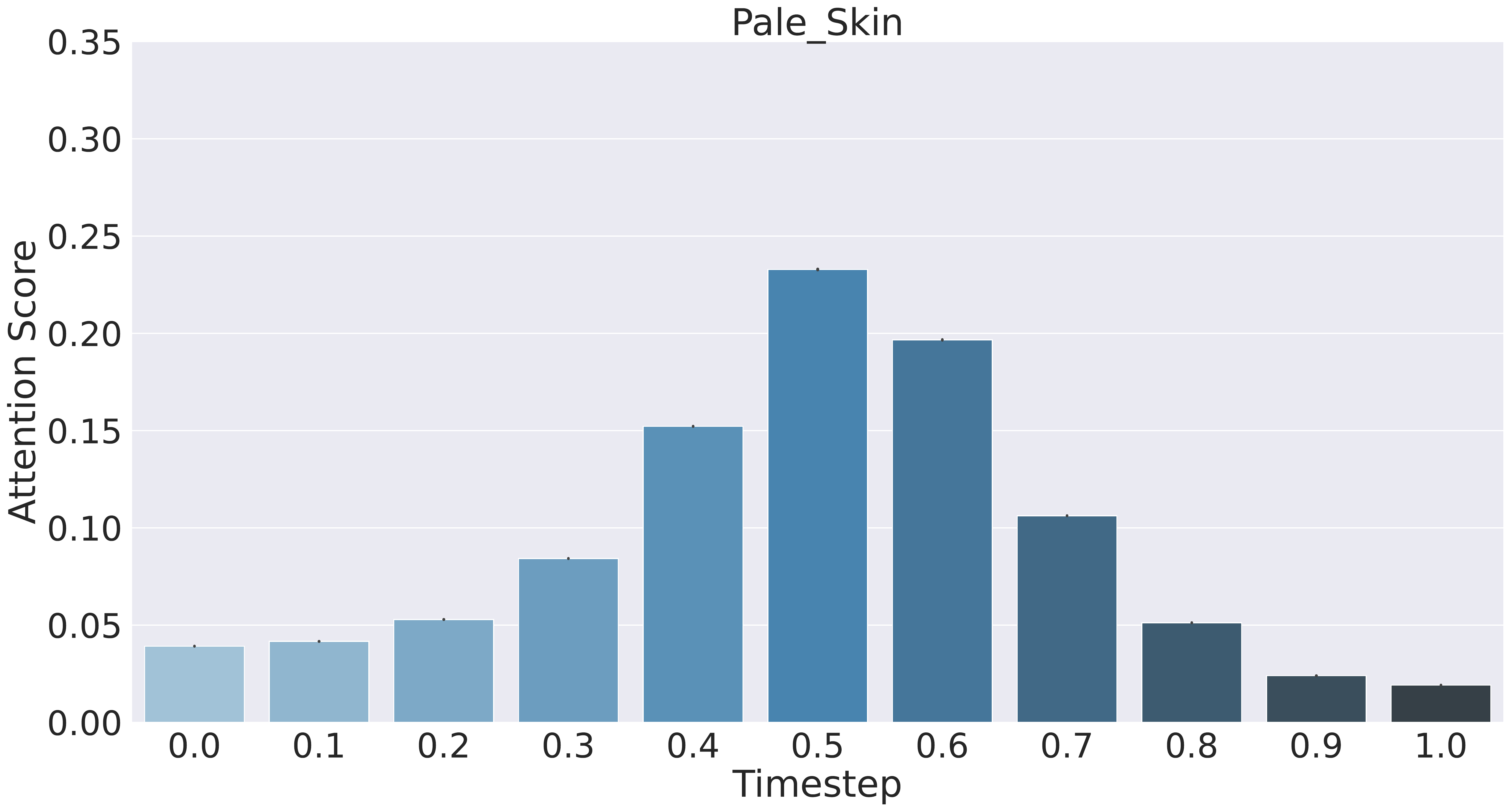}
\end{minipage}
\begin{minipage}[c]{0.24\textwidth}
\includegraphics[width=\textwidth]{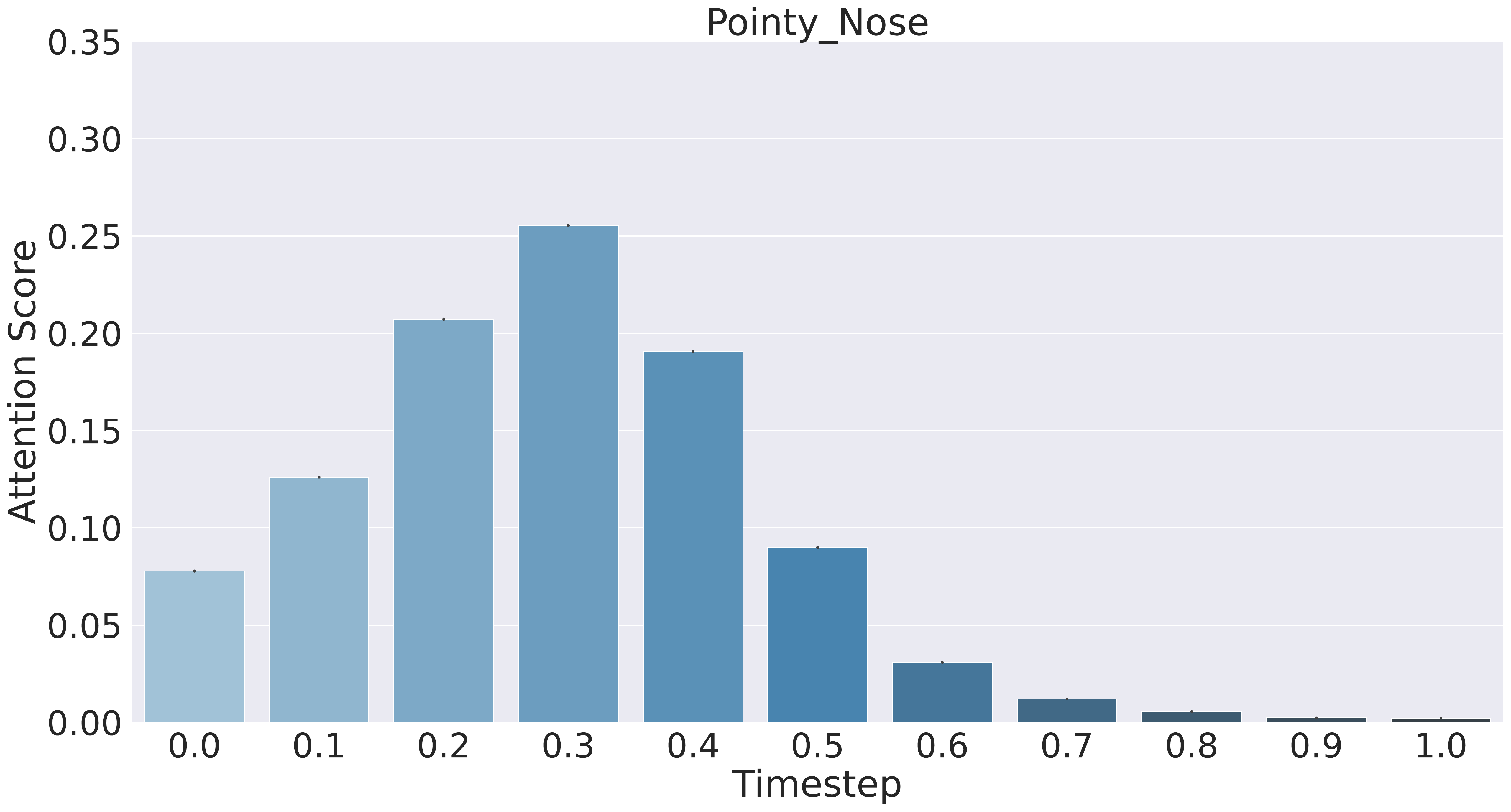}
\end{minipage}
\begin{minipage}[c]{0.24\textwidth}
\includegraphics[width=\textwidth]{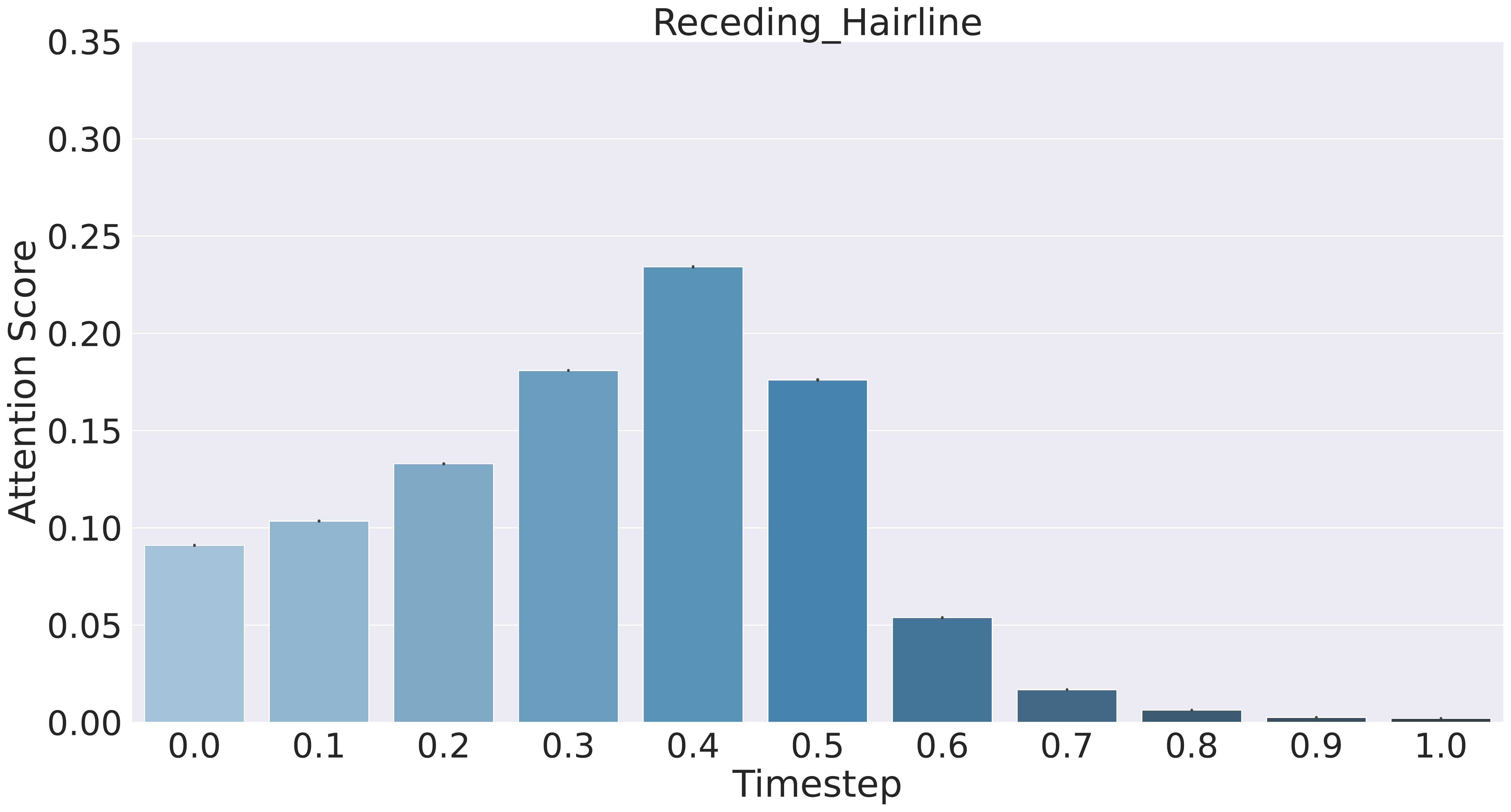}
\end{minipage}
\begin{minipage}[c]{0.24\textwidth}
\includegraphics[width=\textwidth]{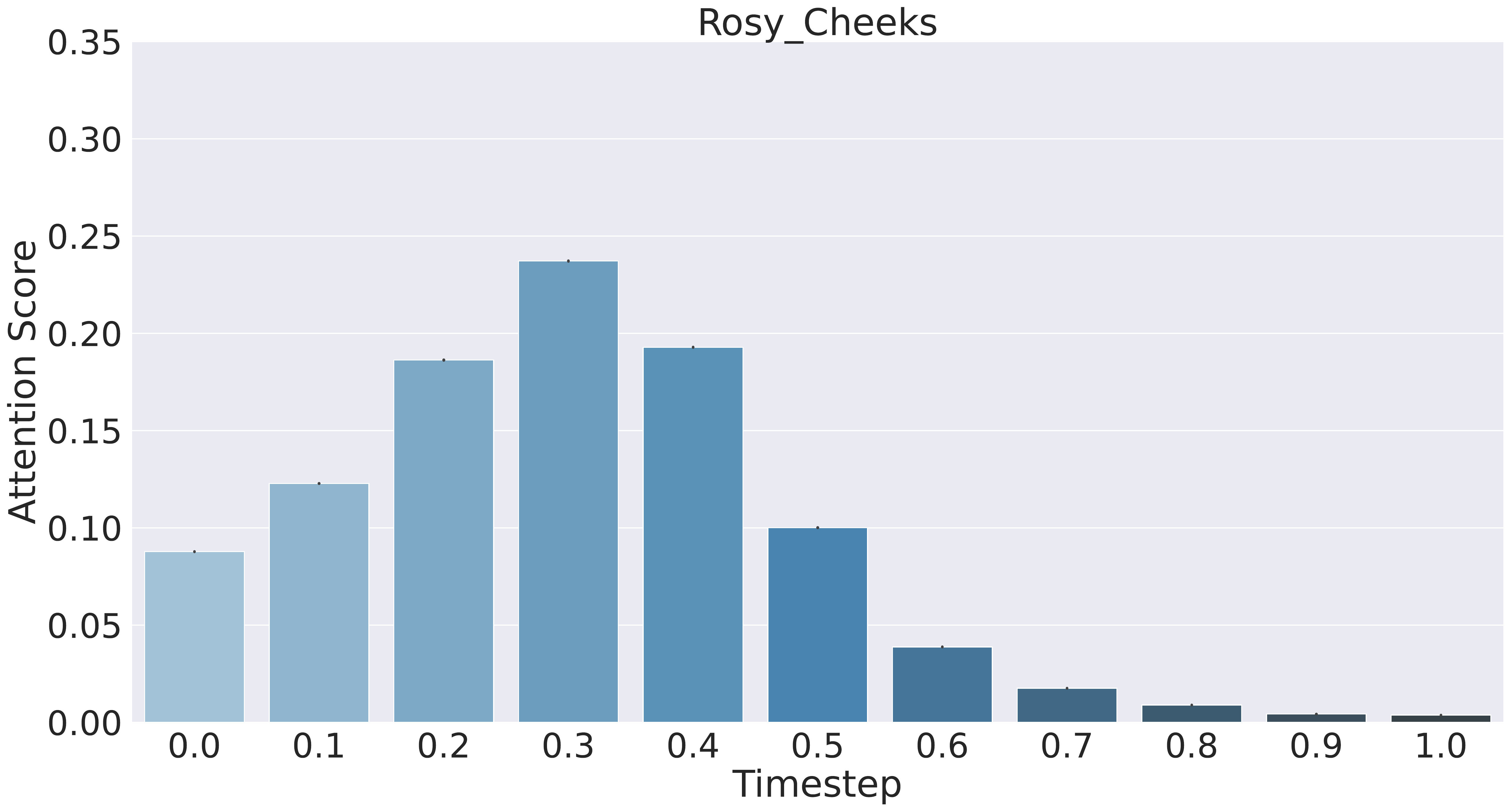}
\end{minipage}
\begin{minipage}[c]{0.24\textwidth}
\includegraphics[width=\textwidth]{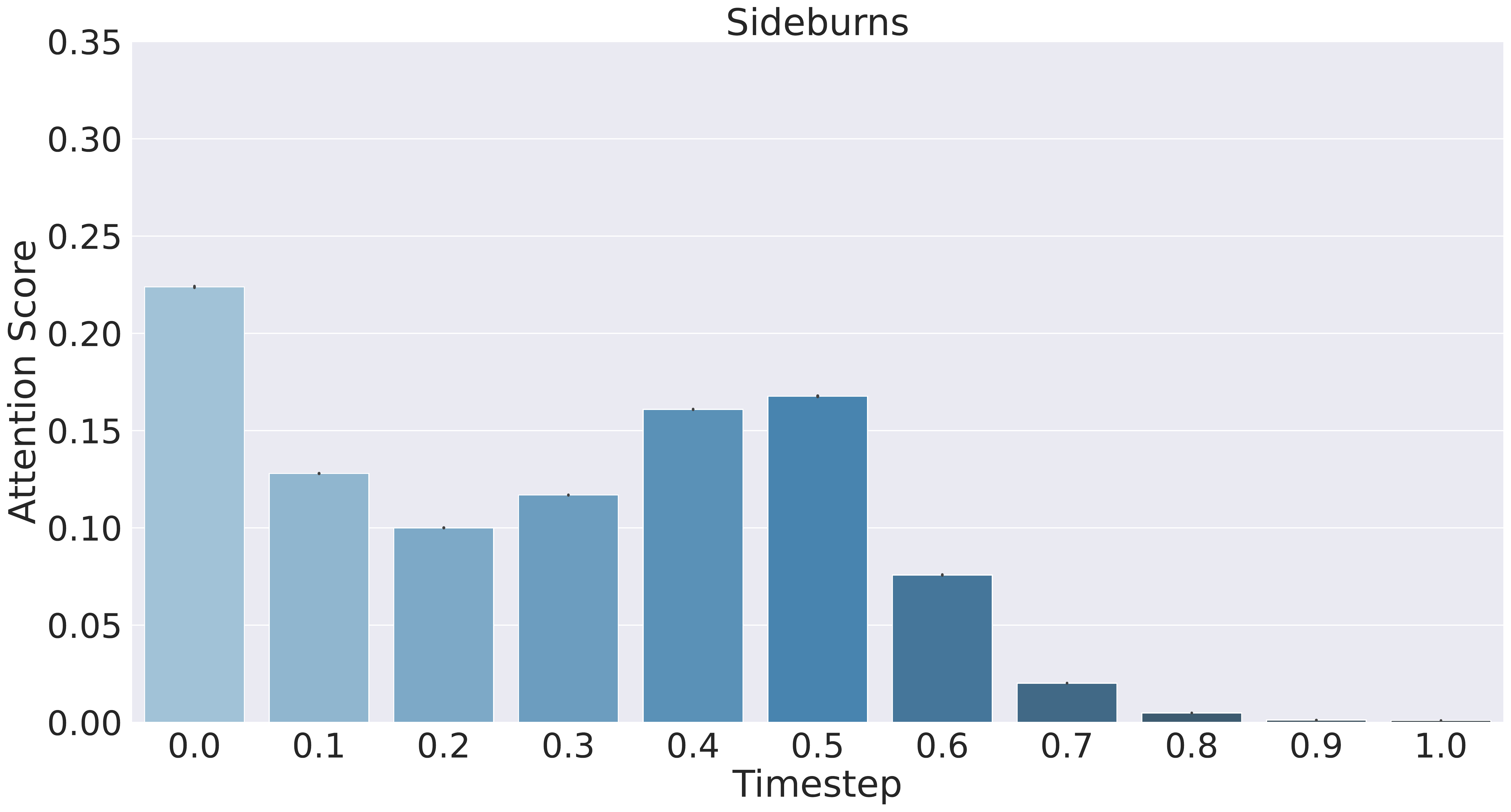}
\end{minipage}
\begin{minipage}[c]{0.24\textwidth}
\includegraphics[width=\textwidth]{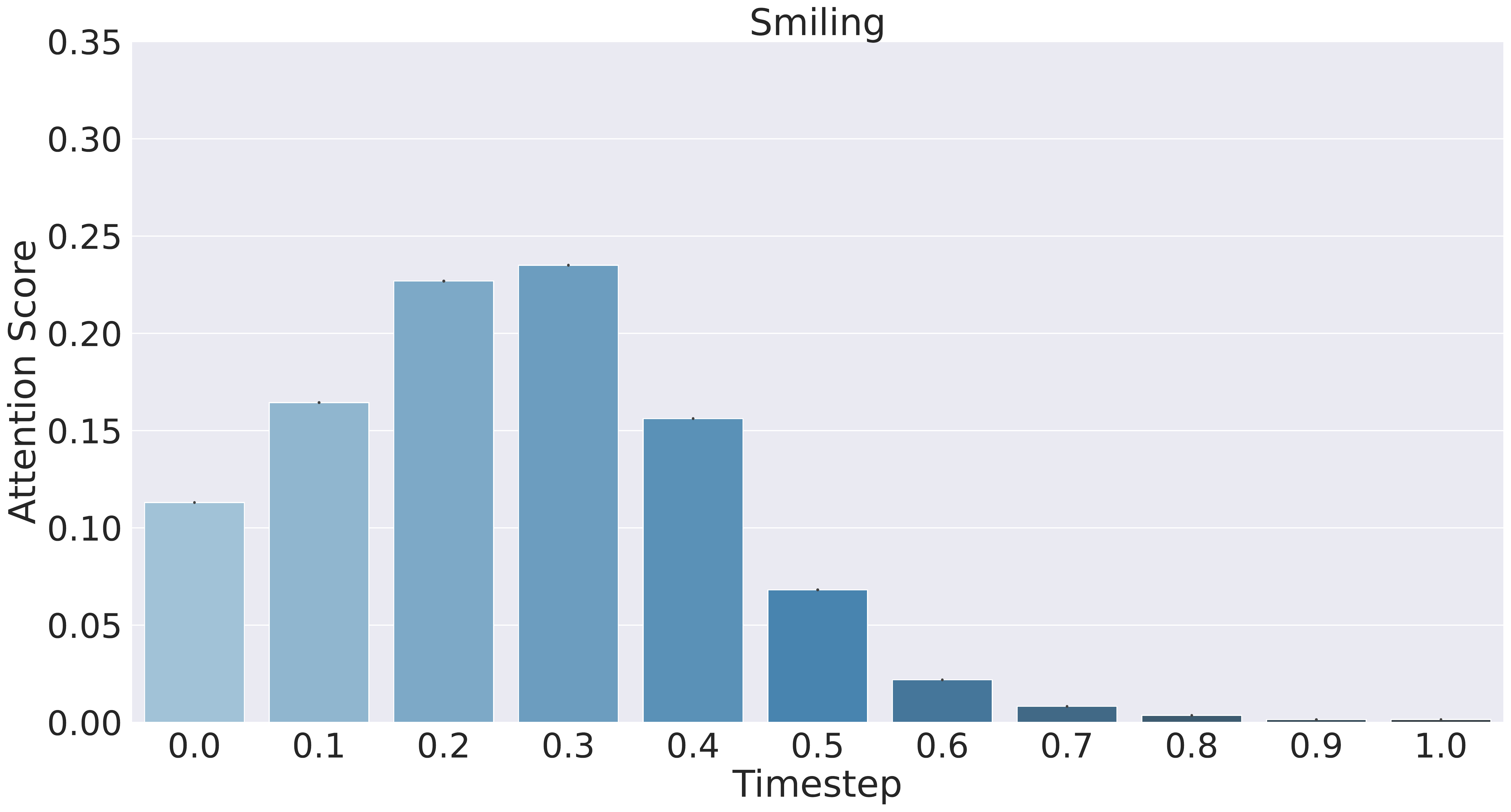}
\end{minipage}
\begin{minipage}[c]{0.24\textwidth}
\includegraphics[width=\textwidth]{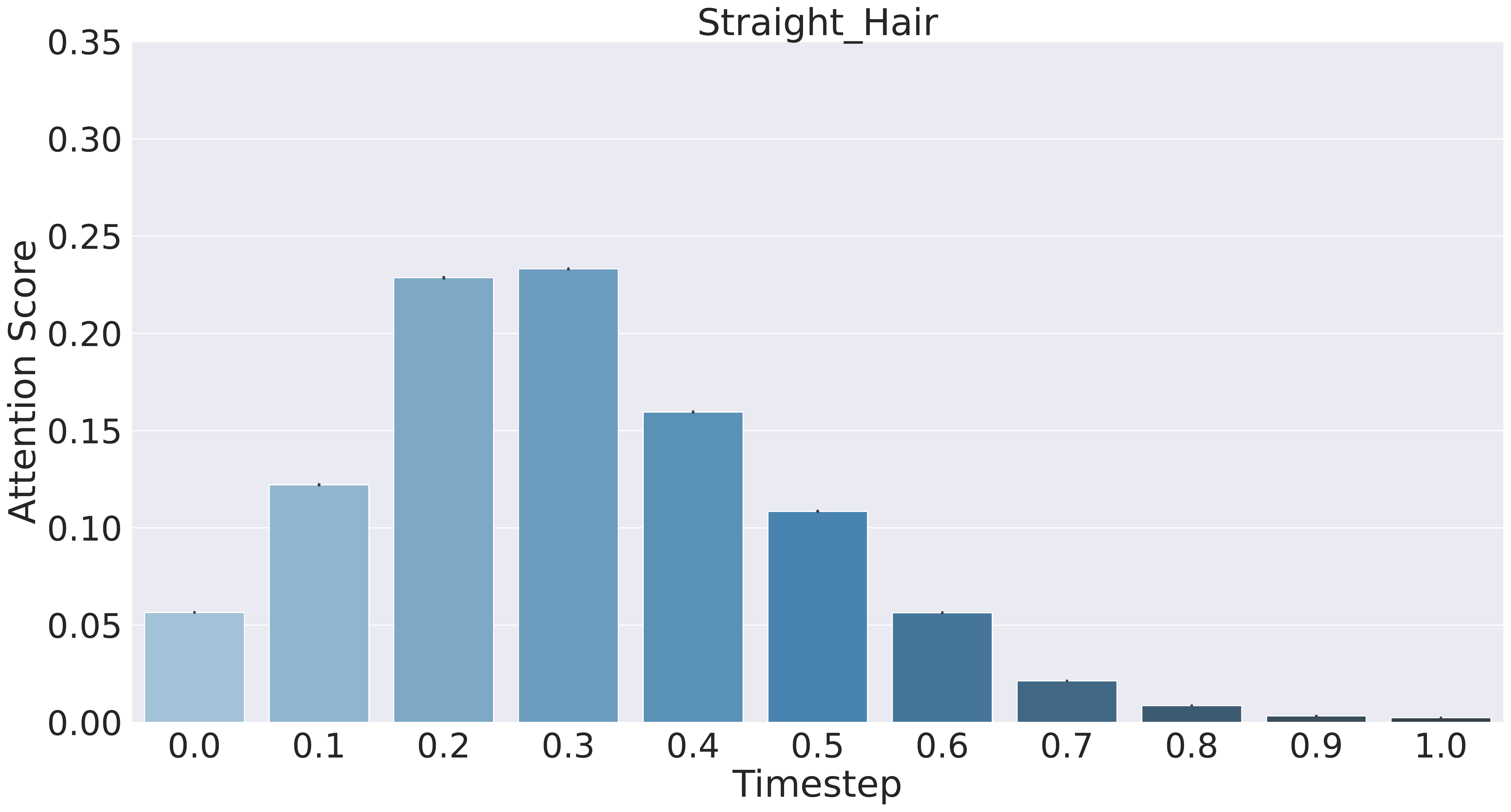}
\end{minipage}
\begin{minipage}[c]{0.24\textwidth}
\includegraphics[width=\textwidth]{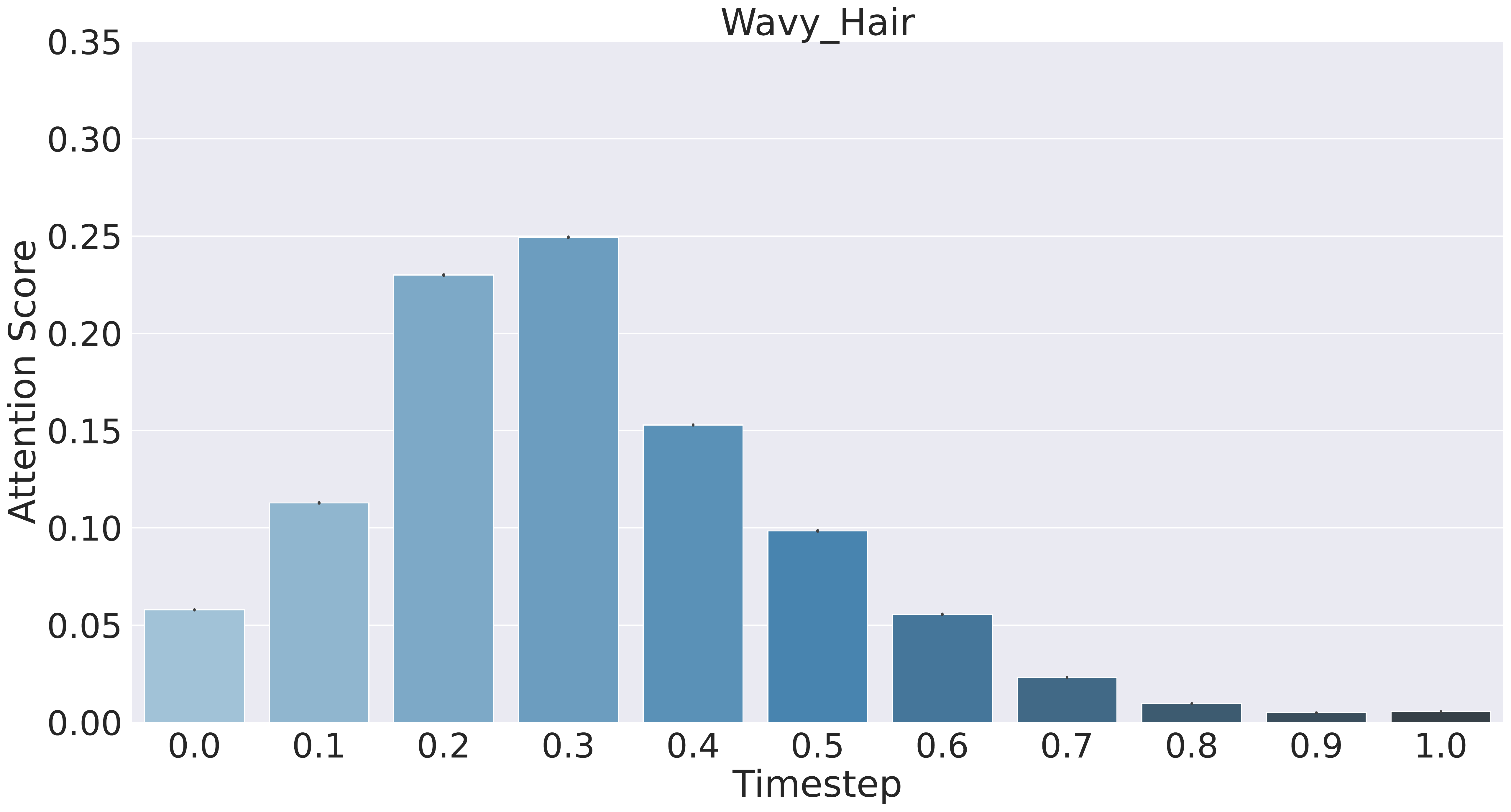}
\end{minipage}
\begin{minipage}[c]{0.24\textwidth}
\includegraphics[width=\textwidth]{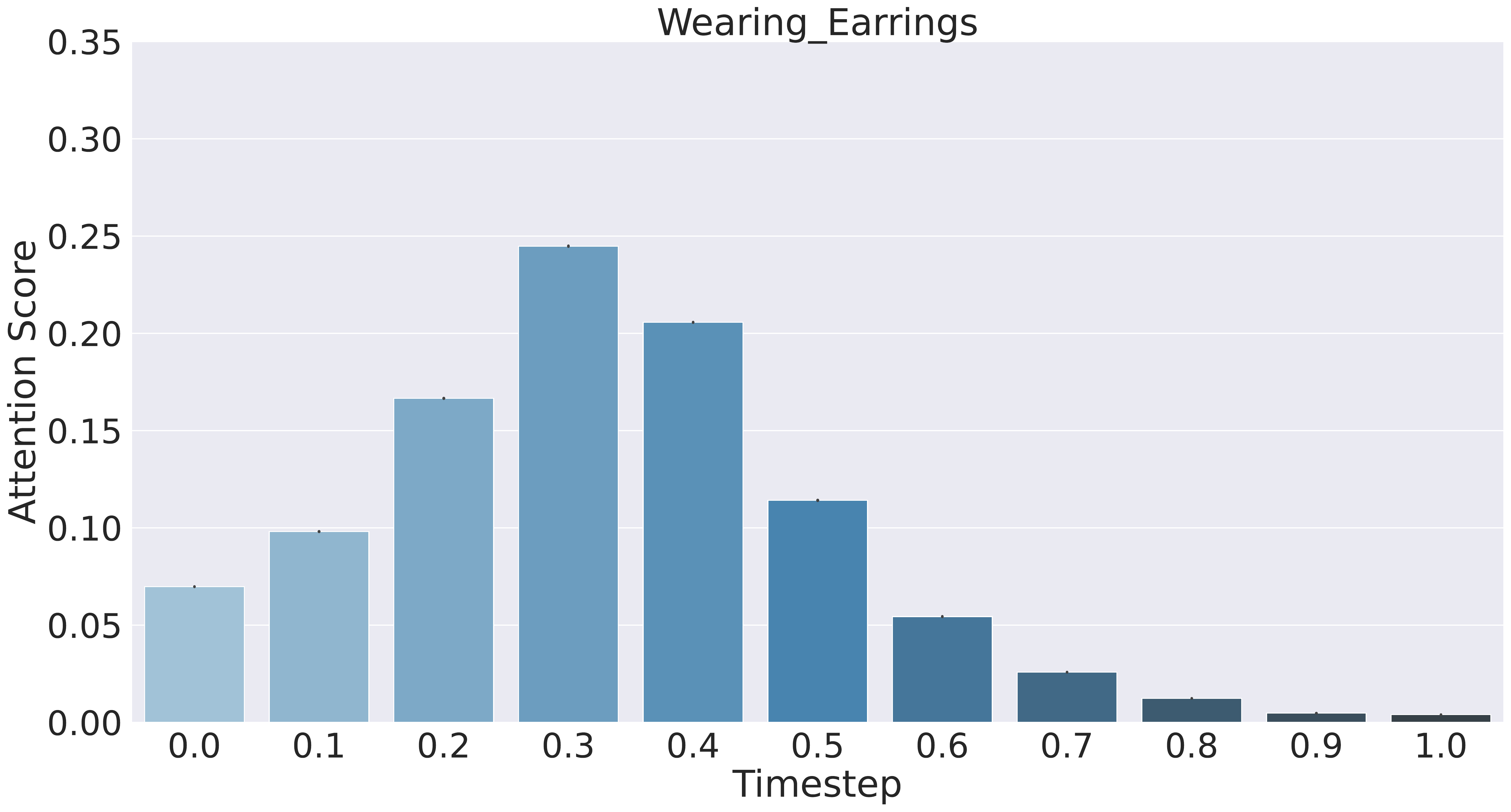}
\end{minipage}
\begin{minipage}[c]{0.24\textwidth}
\includegraphics[width=\textwidth]{Plots/CelebA/Activations/probabilistic-drl/2/10/Wearing_Hat.pdf}
\end{minipage}
\begin{minipage}[c]{0.24\textwidth}
\includegraphics[width=\textwidth]{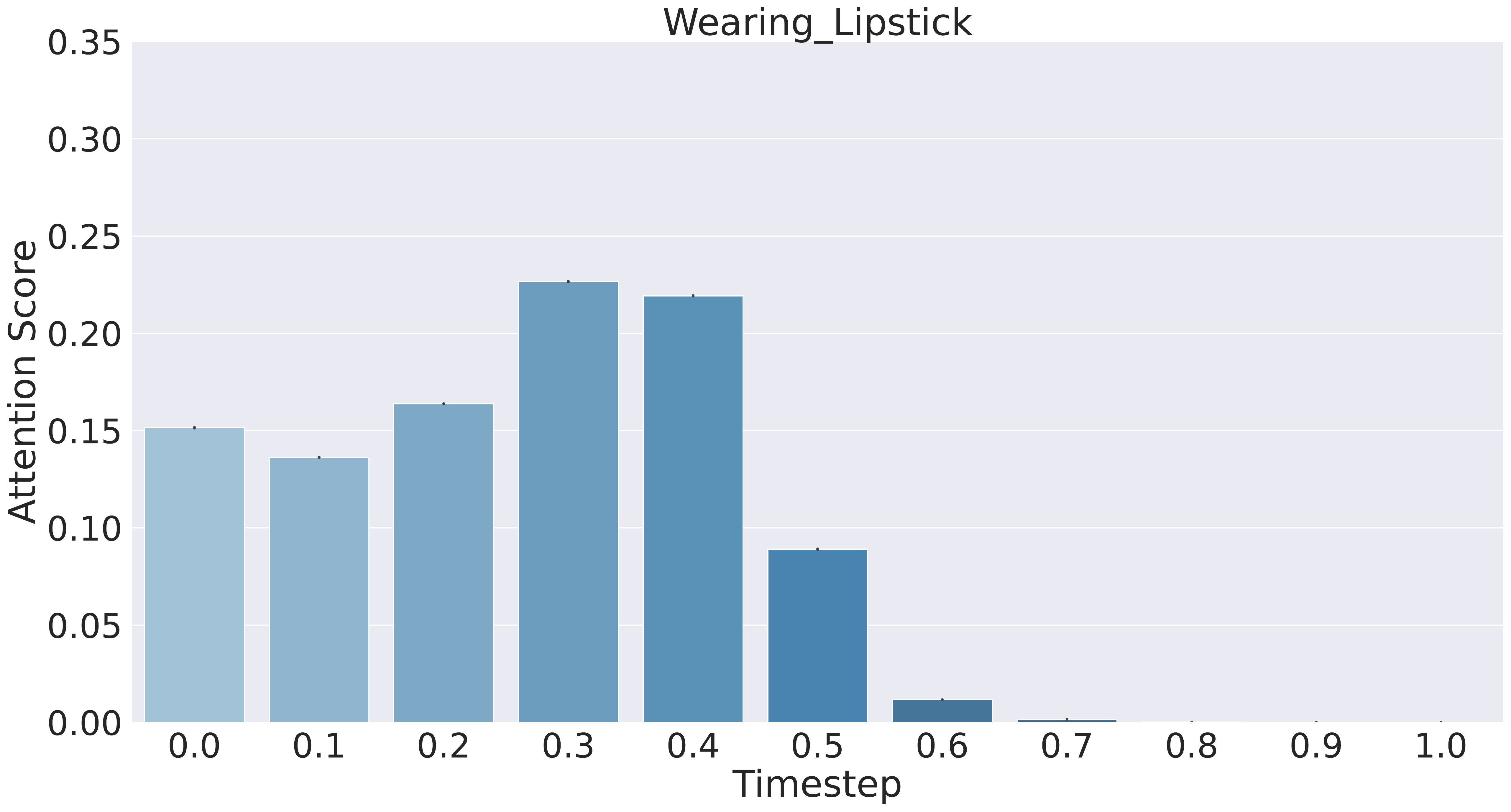}
\end{minipage}
\begin{minipage}[c]{0.24\textwidth}
\includegraphics[width=\textwidth]{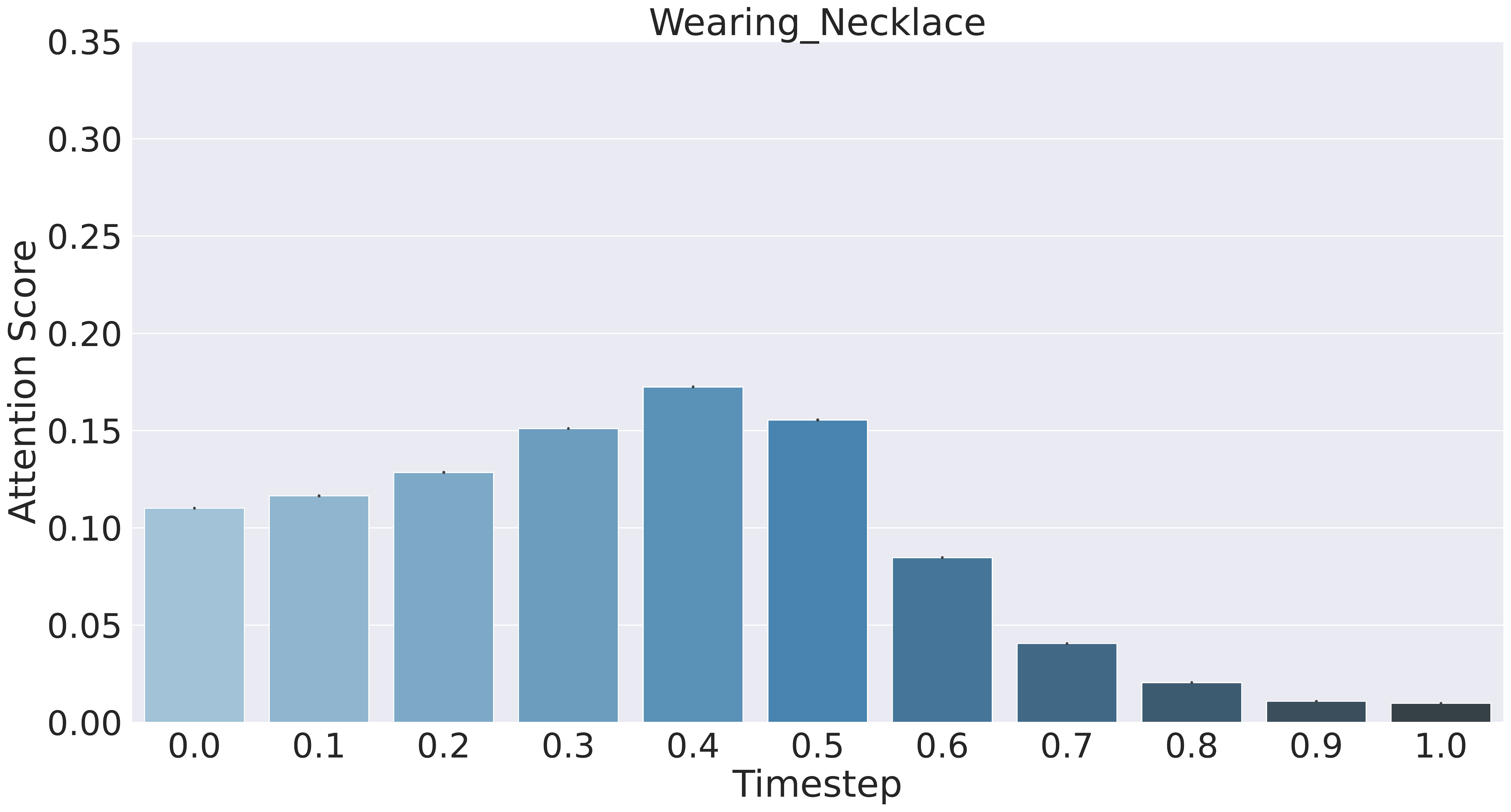}
\end{minipage}
\begin{minipage}[c]{0.24\textwidth}
\includegraphics[width=\textwidth]{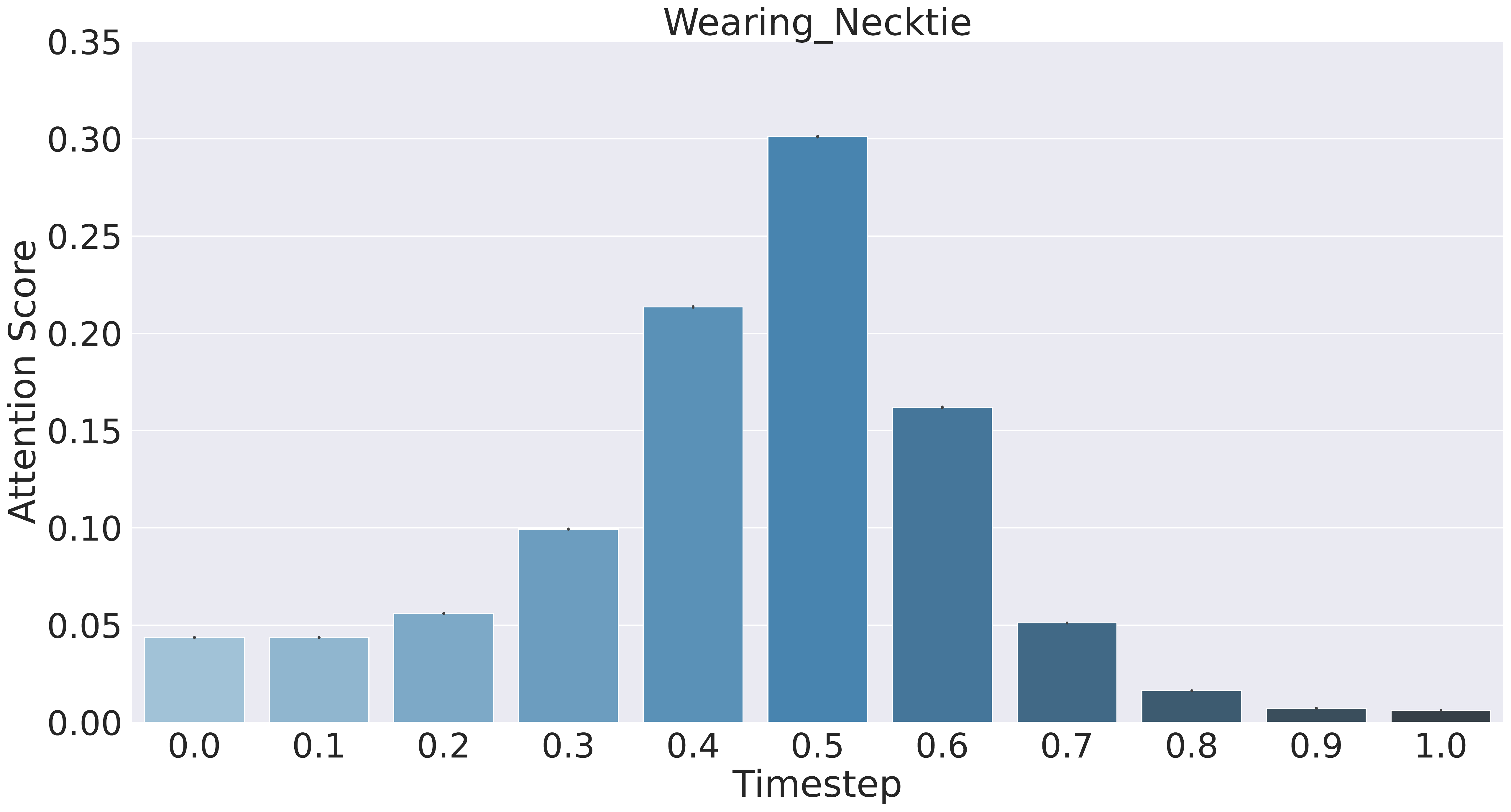}
\end{minipage}
\begin{minipage}[c]{0.24\textwidth}
\includegraphics[width=\textwidth]{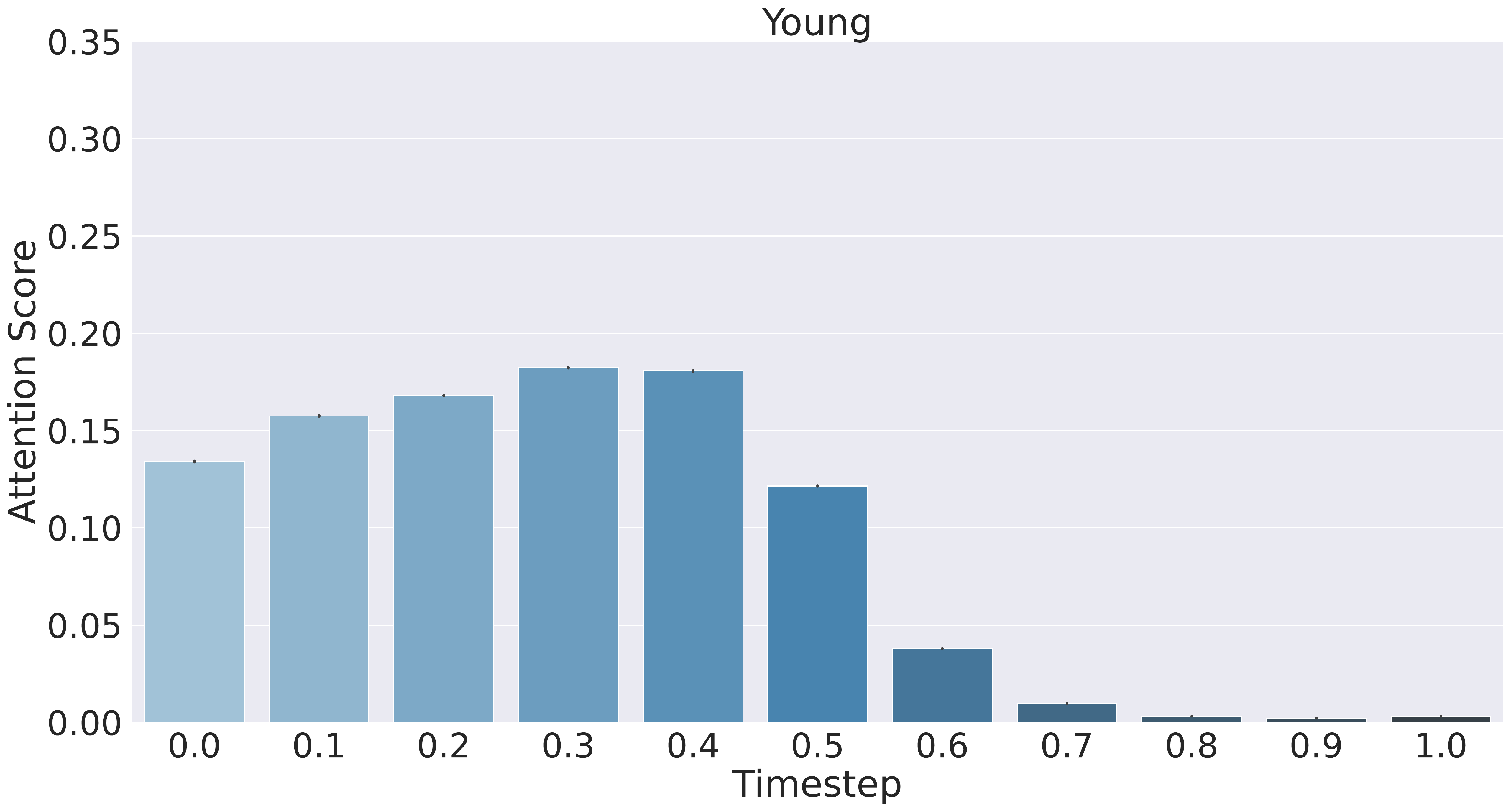}
\end{minipage}
\caption{Attention Score profiles for different parts of the trajectory-based representation on CelebA when using the VDRL stochastic encoder.}
\label{fig:celeba_activ_vdrl}
\end{figure}

\begin{figure}\begin{minipage}[c]{0.24\textwidth}
\includegraphics[width=\textwidth]{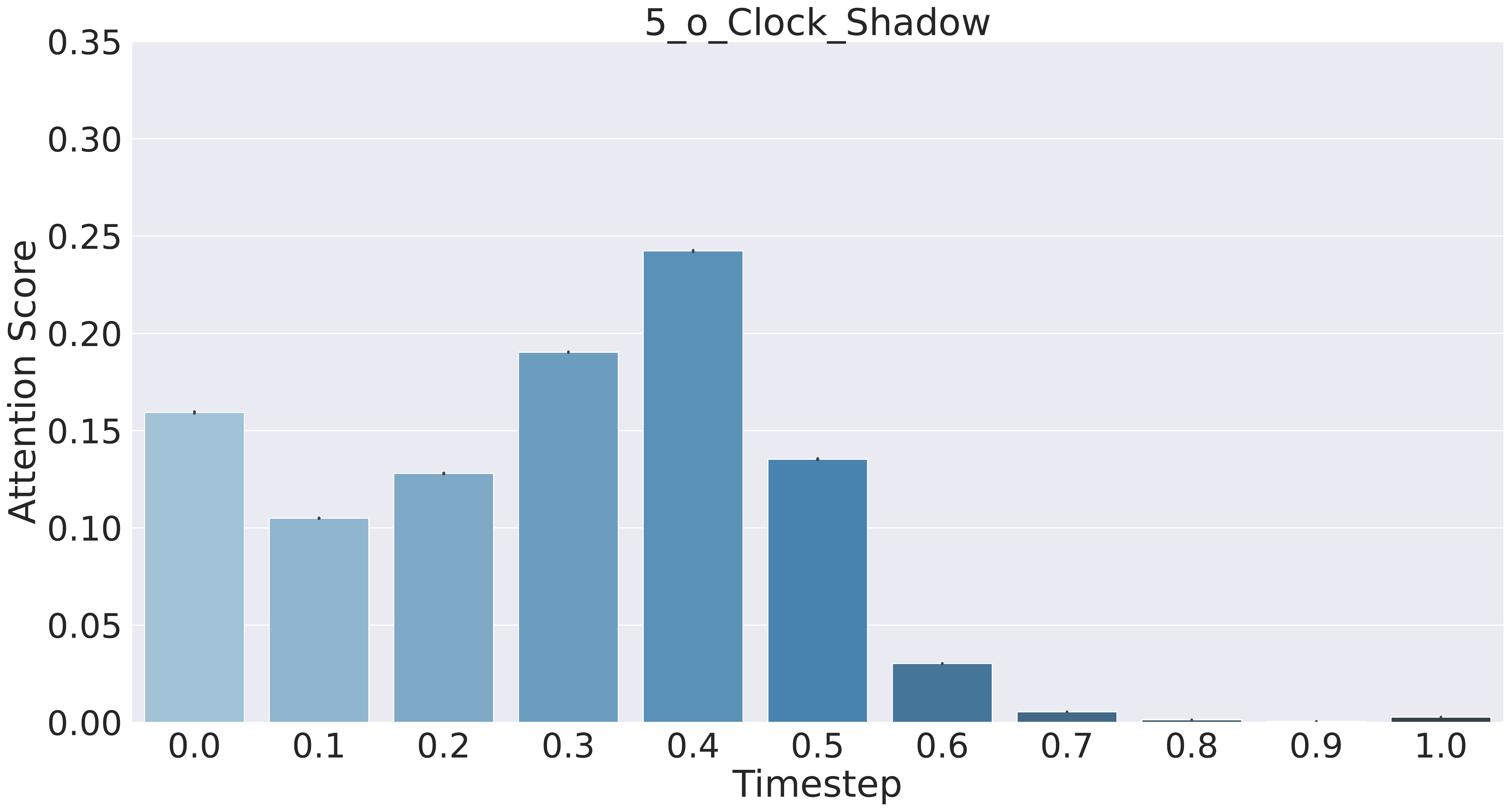}
\end{minipage}
\begin{minipage}[c]{0.24\textwidth}
\includegraphics[width=\textwidth]{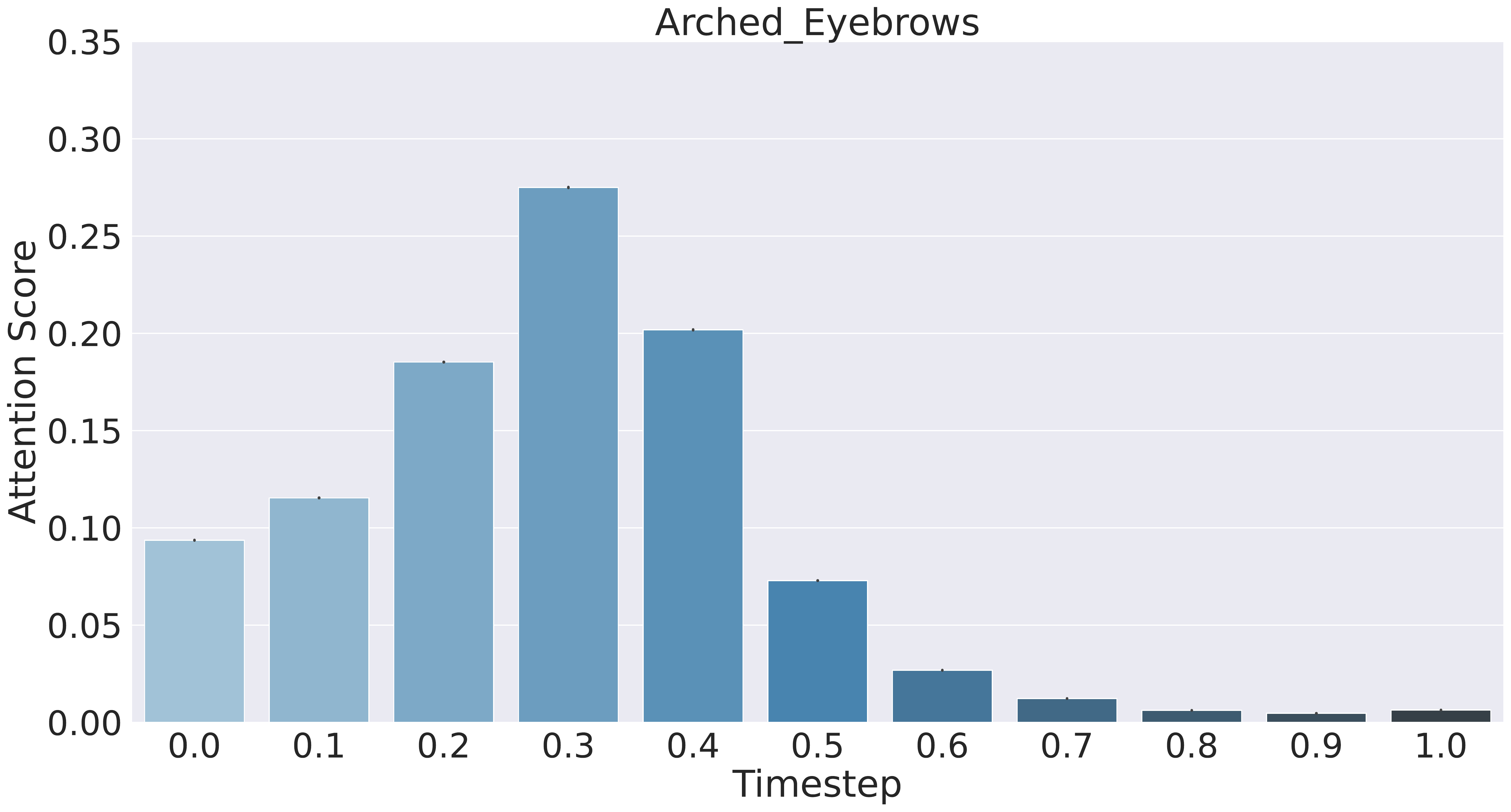}
\end{minipage}
\begin{minipage}[c]{0.24\textwidth}
\includegraphics[width=\textwidth]{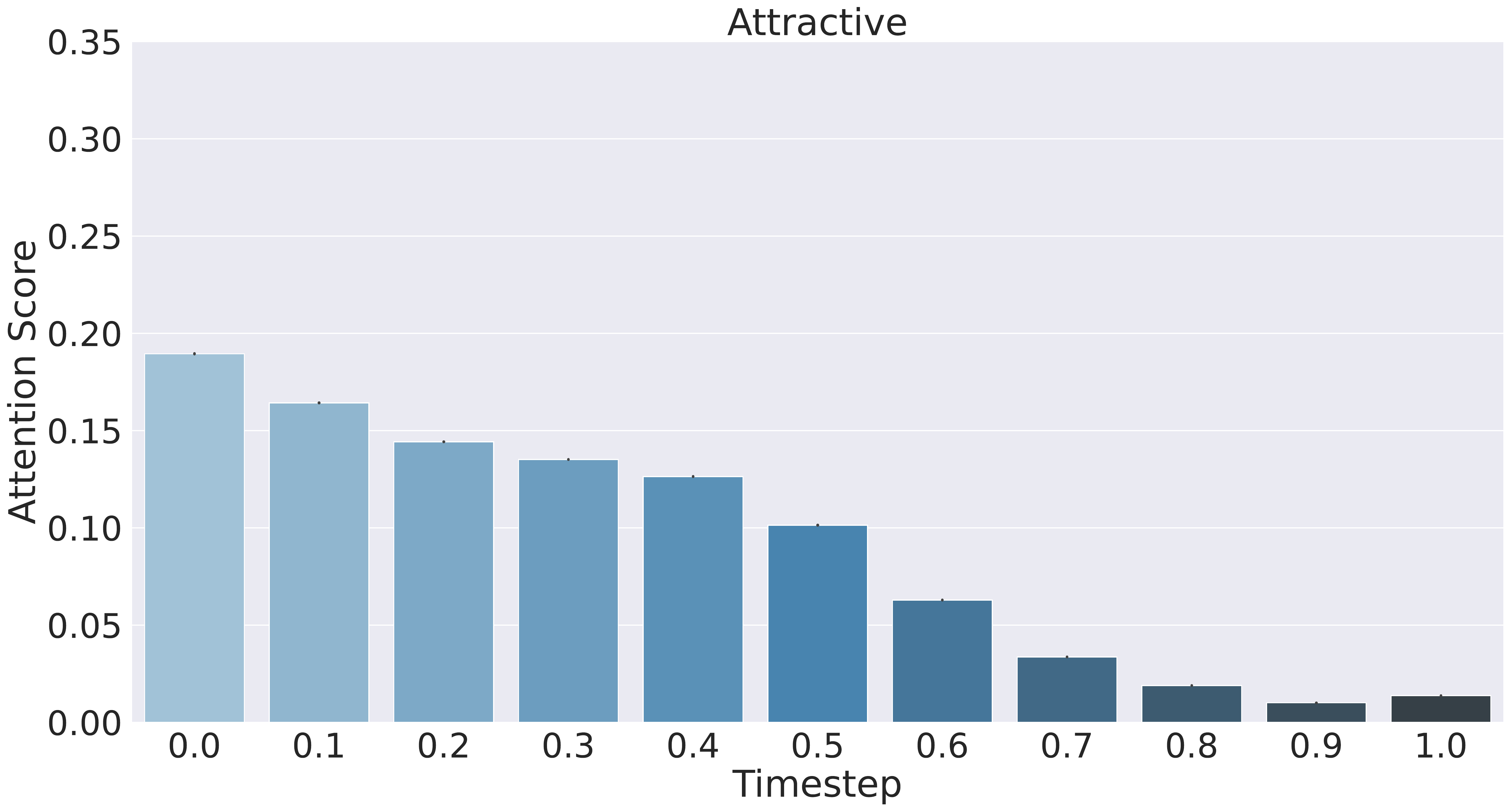}
\end{minipage}
\begin{minipage}[c]{0.24\textwidth}
\includegraphics[width=\textwidth]{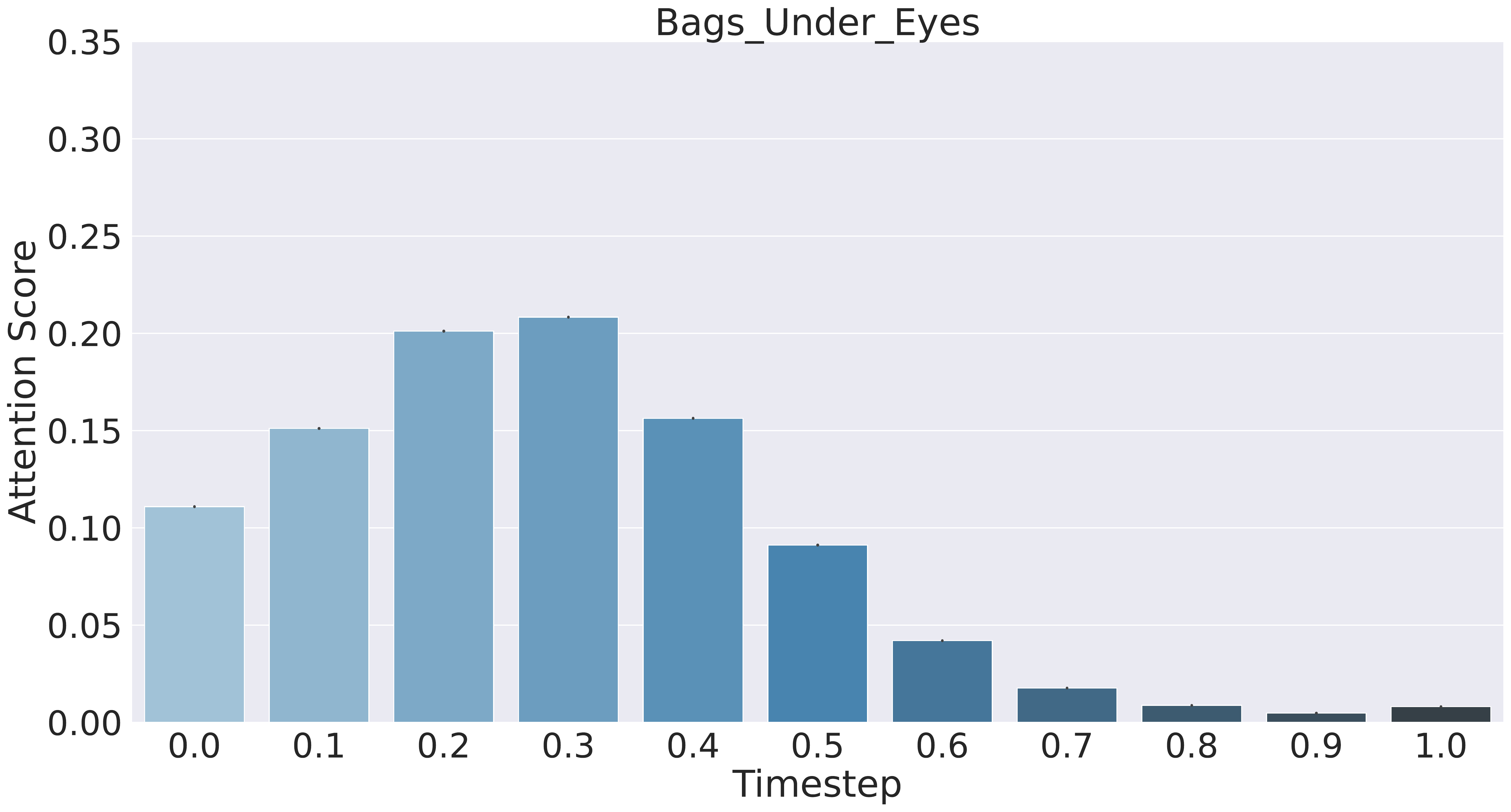}
\end{minipage}
\begin{minipage}[c]{0.24\textwidth}
\includegraphics[width=\textwidth]{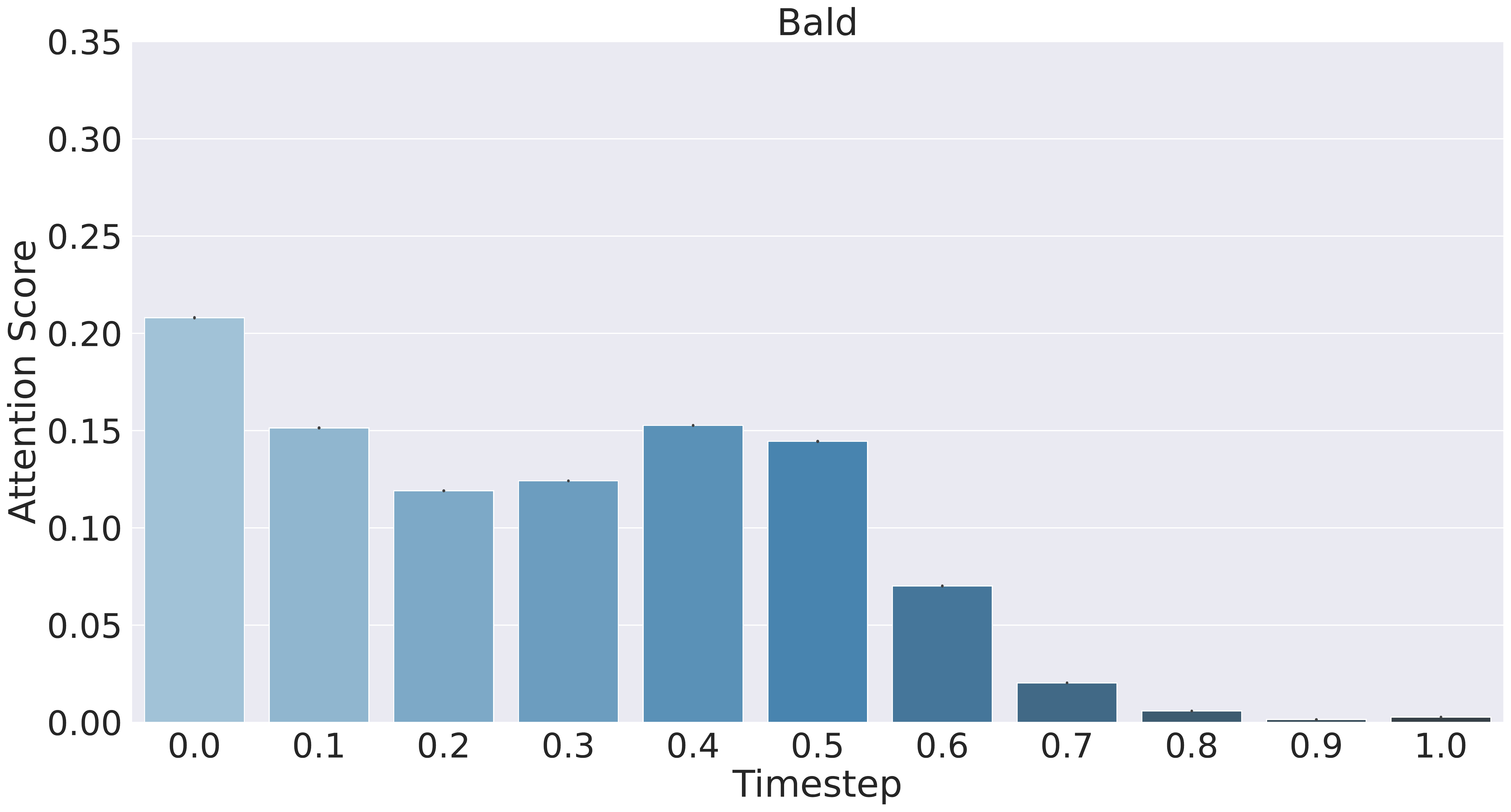}
\end{minipage}
\begin{minipage}[c]{0.24\textwidth}
\includegraphics[width=\textwidth]{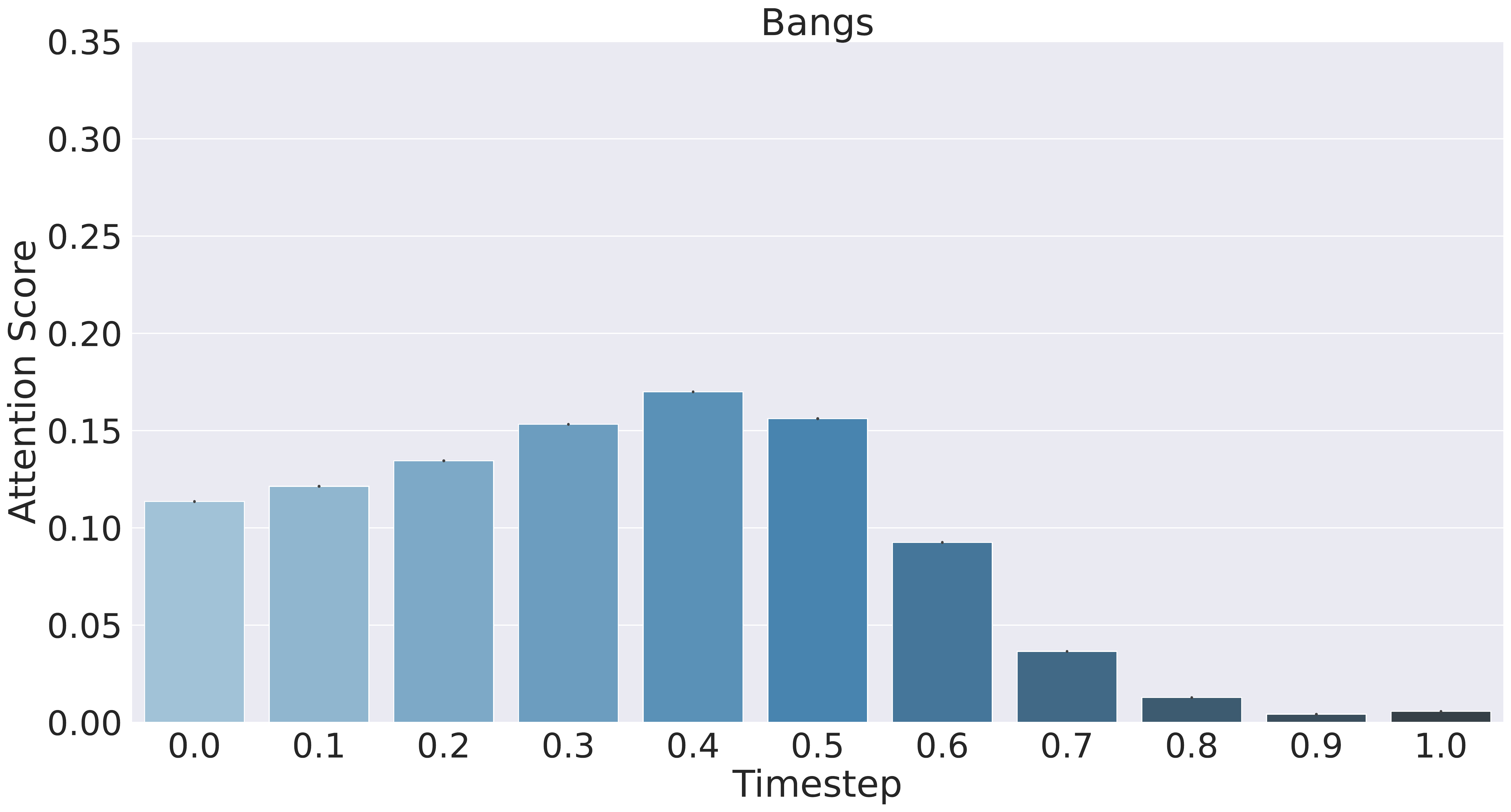}
\end{minipage}
\begin{minipage}[c]{0.24\textwidth}
\includegraphics[width=\textwidth]{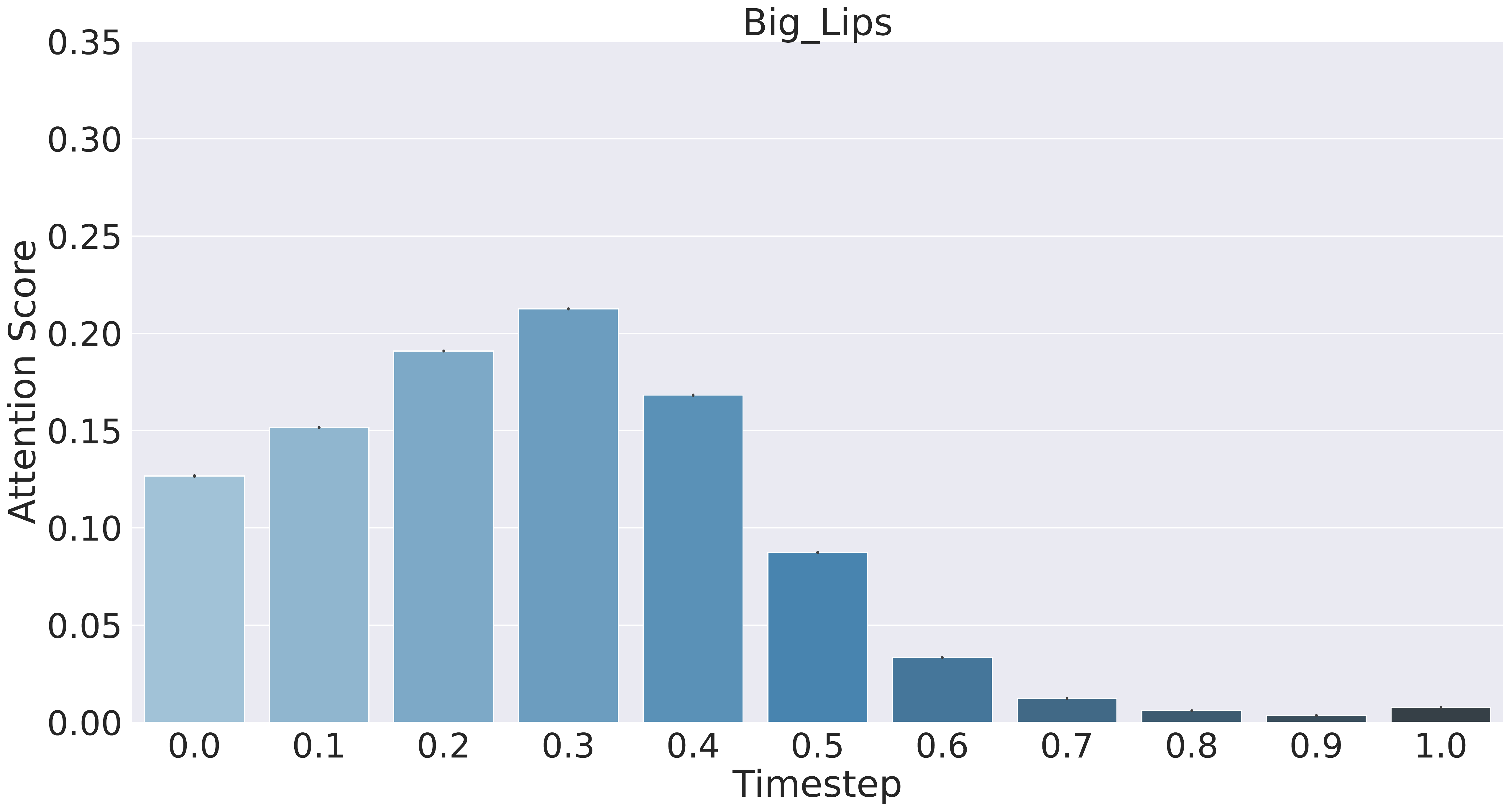}
\end{minipage}
\begin{minipage}[c]{0.24\textwidth}
\includegraphics[width=\textwidth]{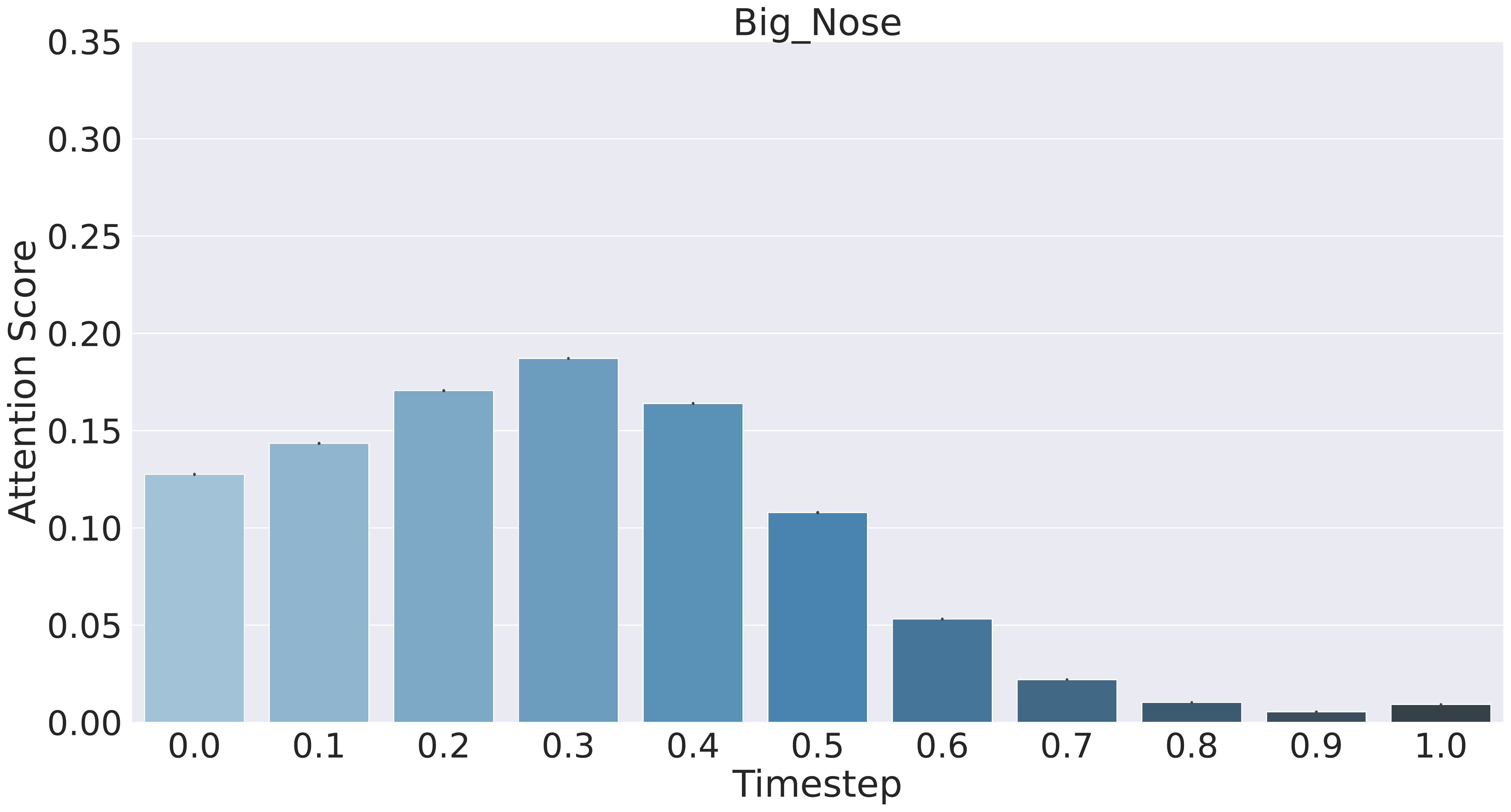}
\end{minipage}
\begin{minipage}[c]{0.24\textwidth}
\includegraphics[width=\textwidth]{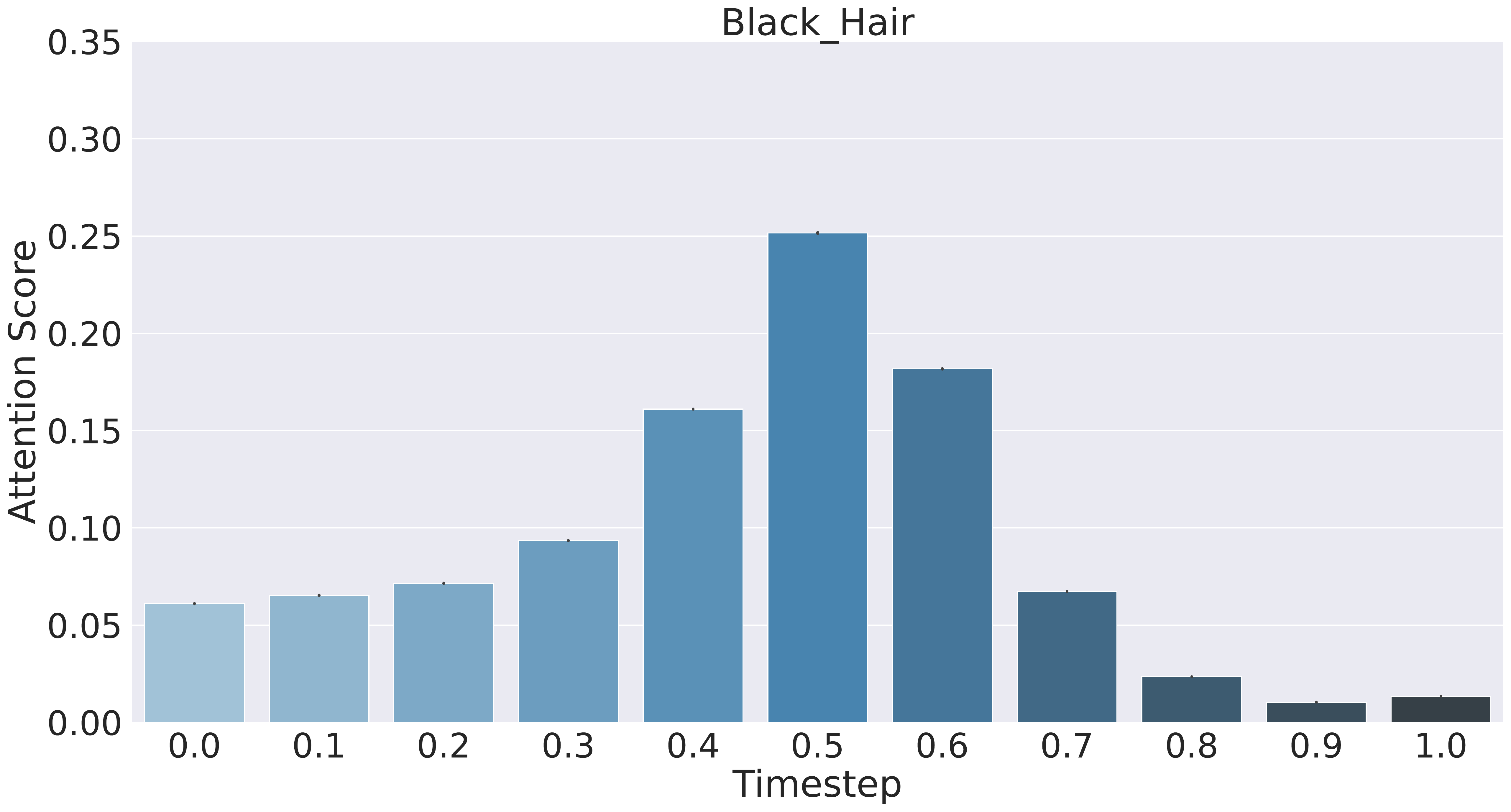}
\end{minipage}
\begin{minipage}[c]{0.24\textwidth}
\includegraphics[width=\textwidth]{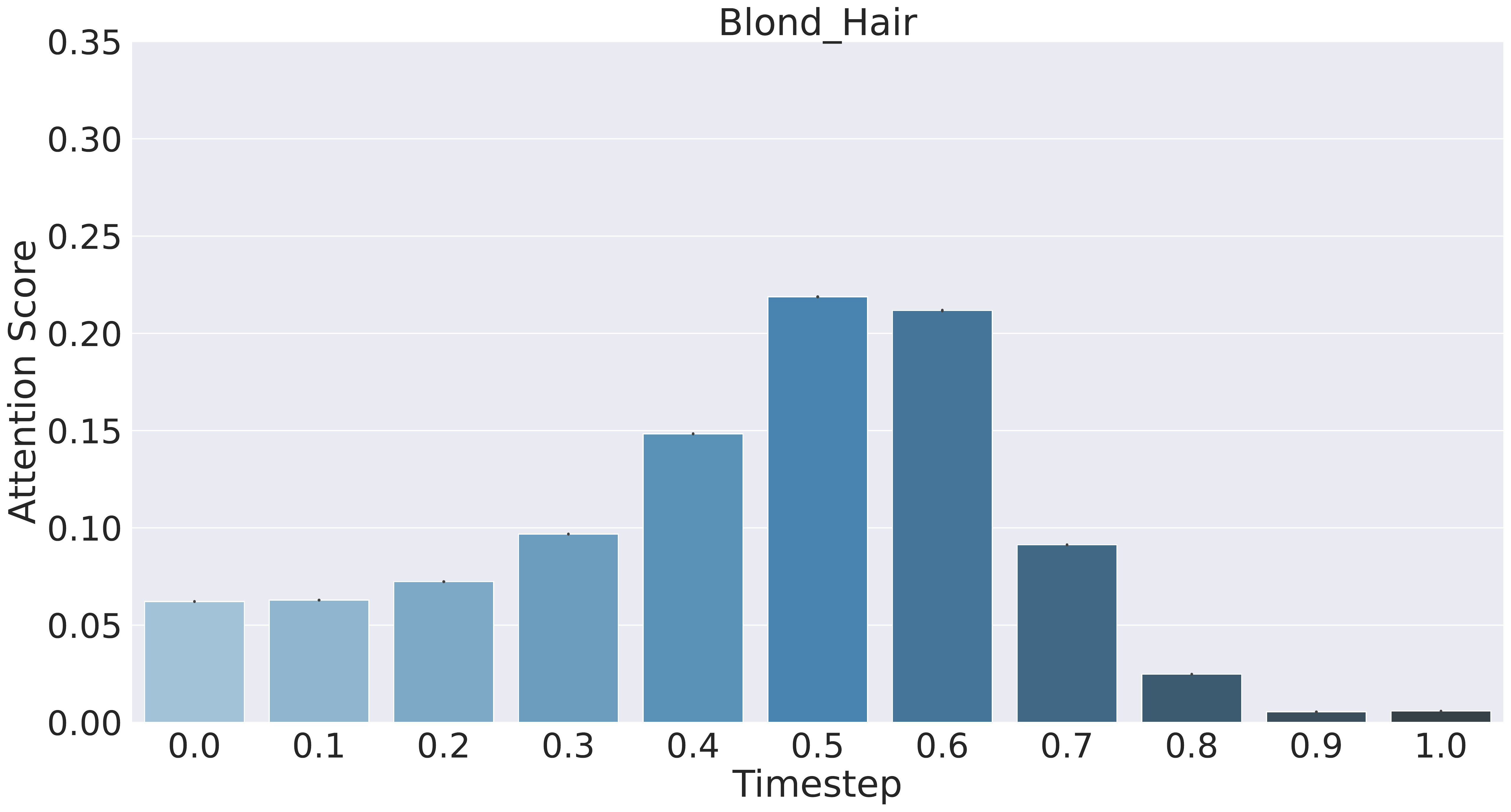}
\end{minipage}
\begin{minipage}[c]{0.24\textwidth}
\includegraphics[width=\textwidth]{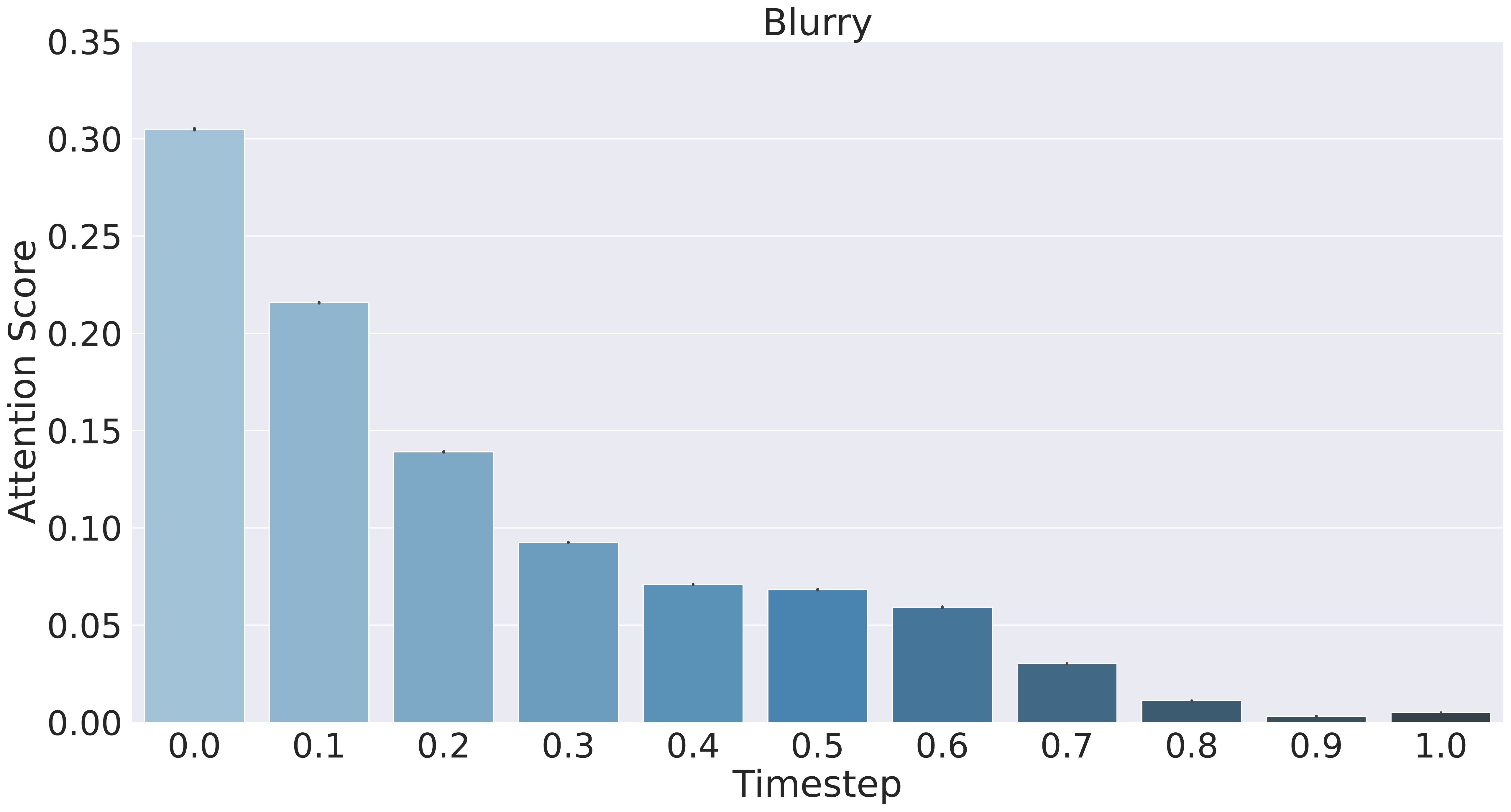}
\end{minipage}
\begin{minipage}[c]{0.24\textwidth}
\includegraphics[width=\textwidth]{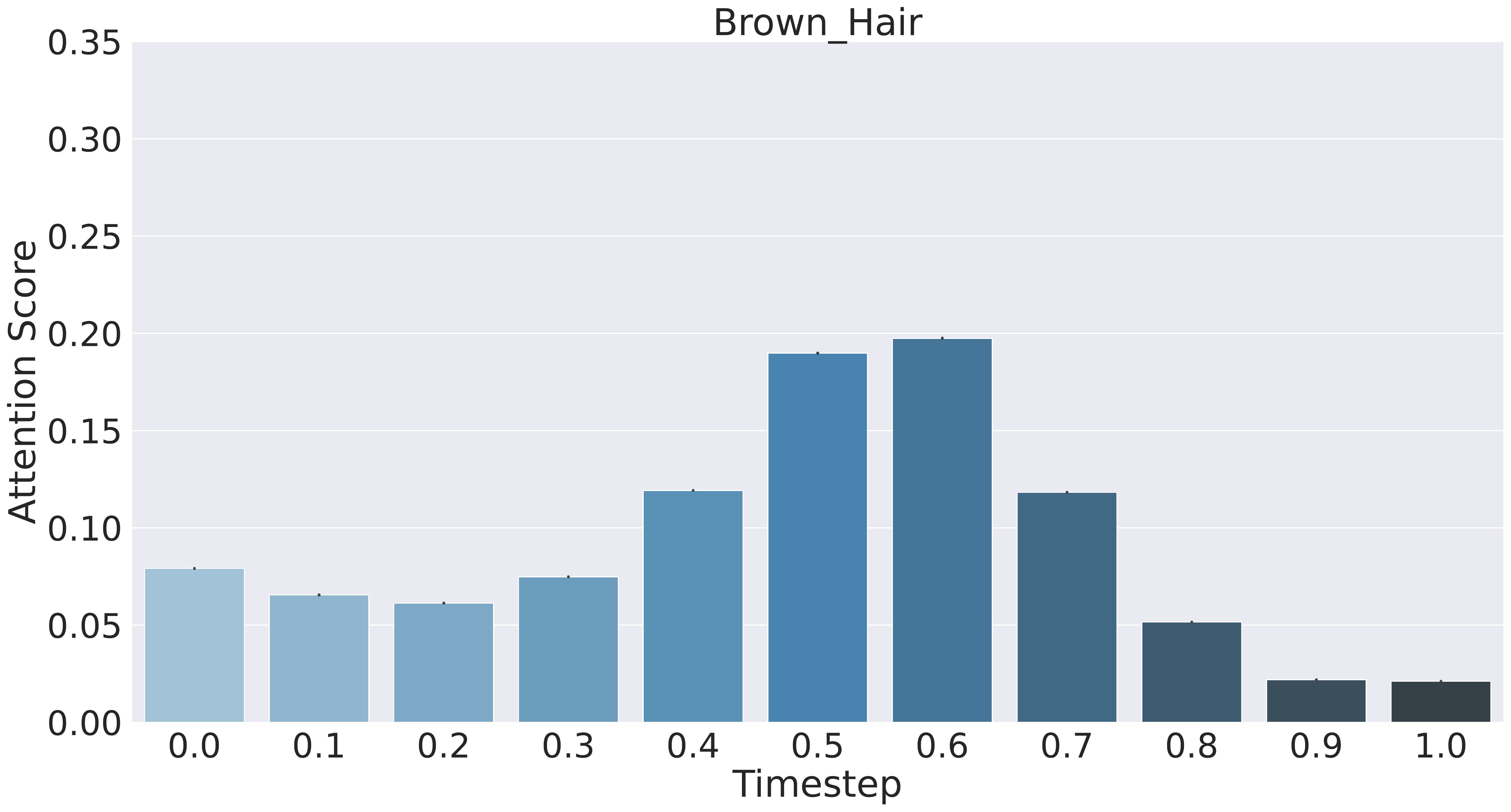}
\end{minipage}
\begin{minipage}[c]{0.24\textwidth}
\includegraphics[width=\textwidth]{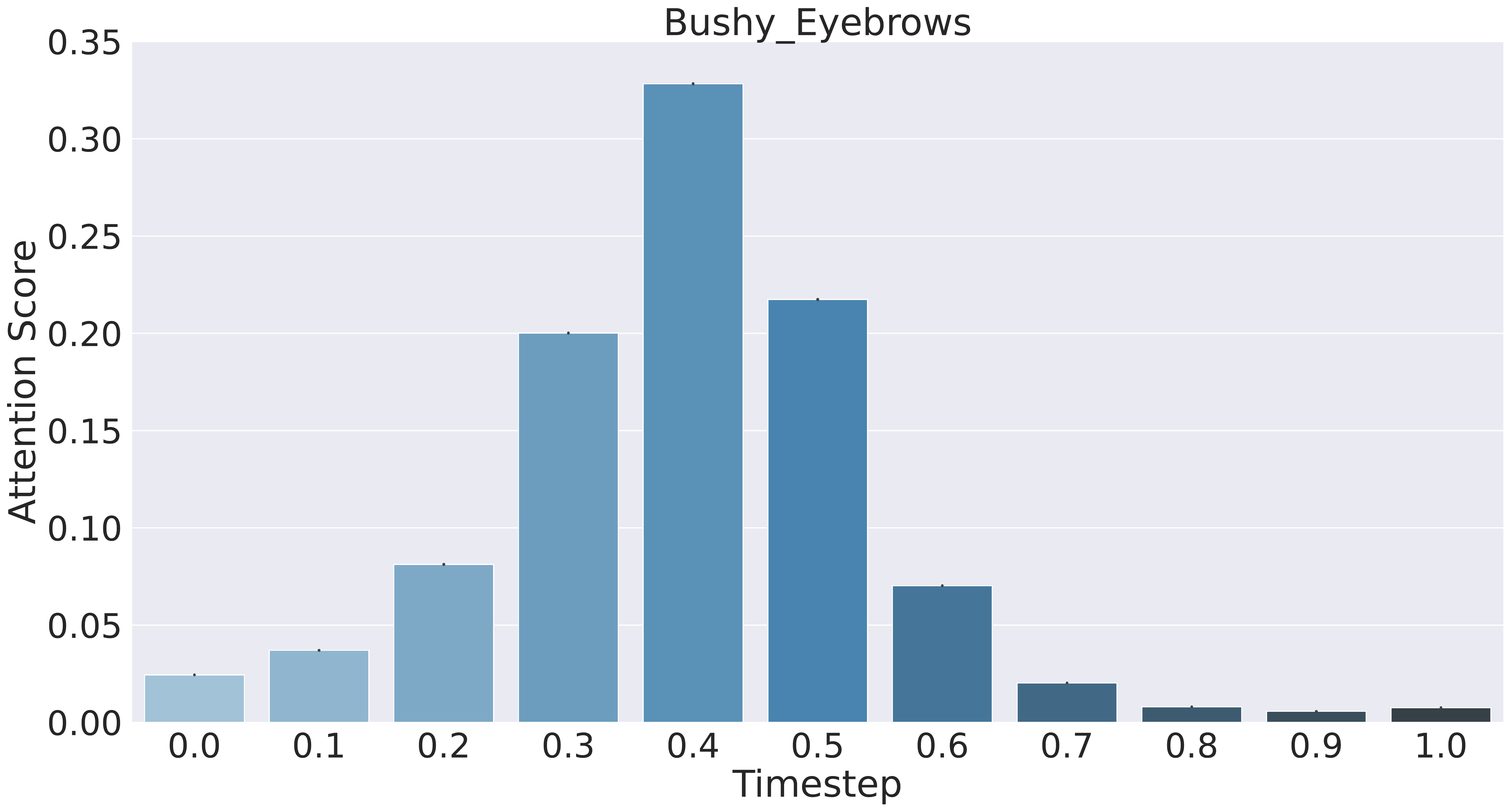}
\end{minipage}
\begin{minipage}[c]{0.24\textwidth}
\includegraphics[width=\textwidth]{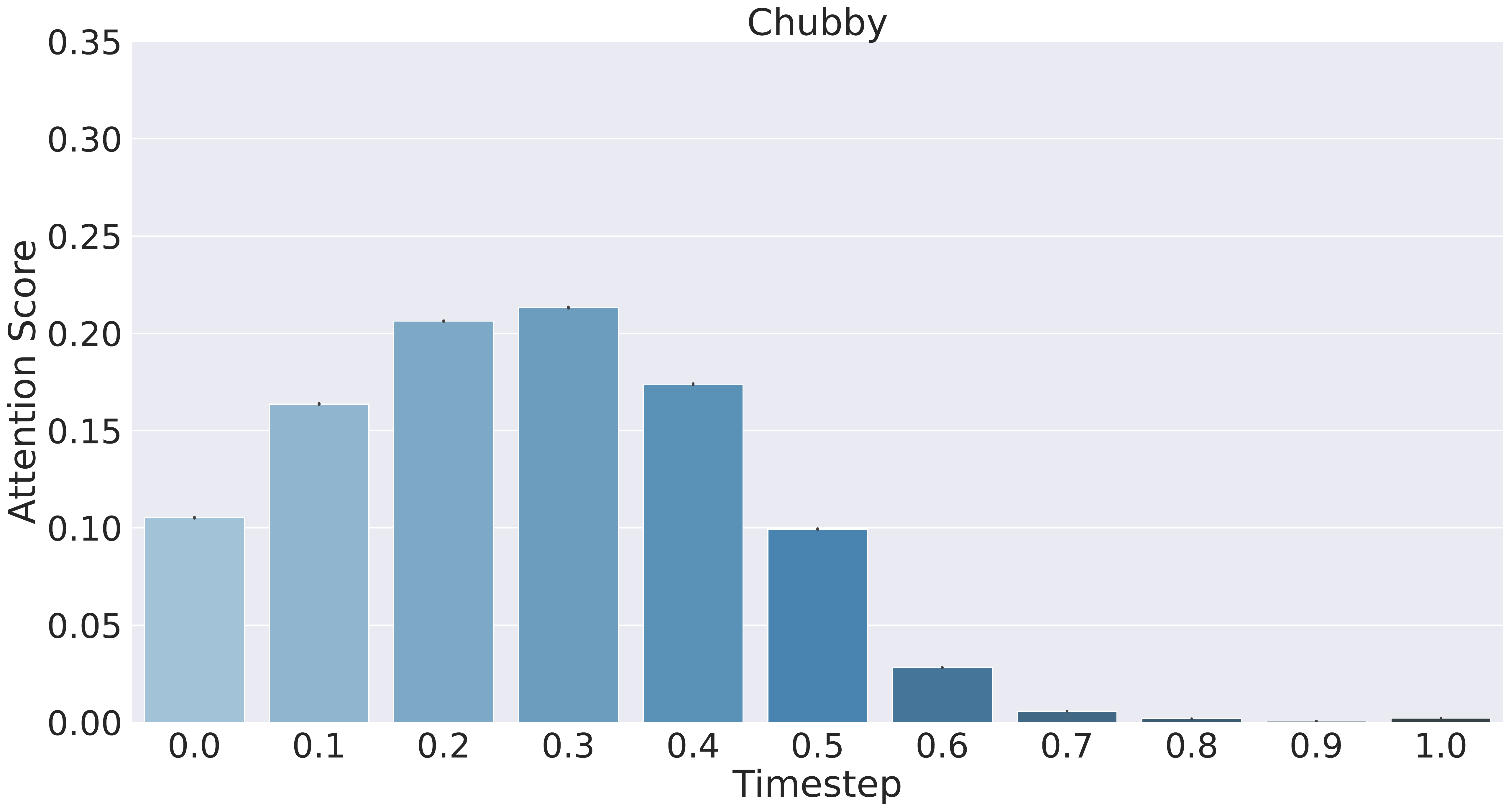}
\end{minipage}
\begin{minipage}[c]{0.24\textwidth}
\includegraphics[width=\textwidth]{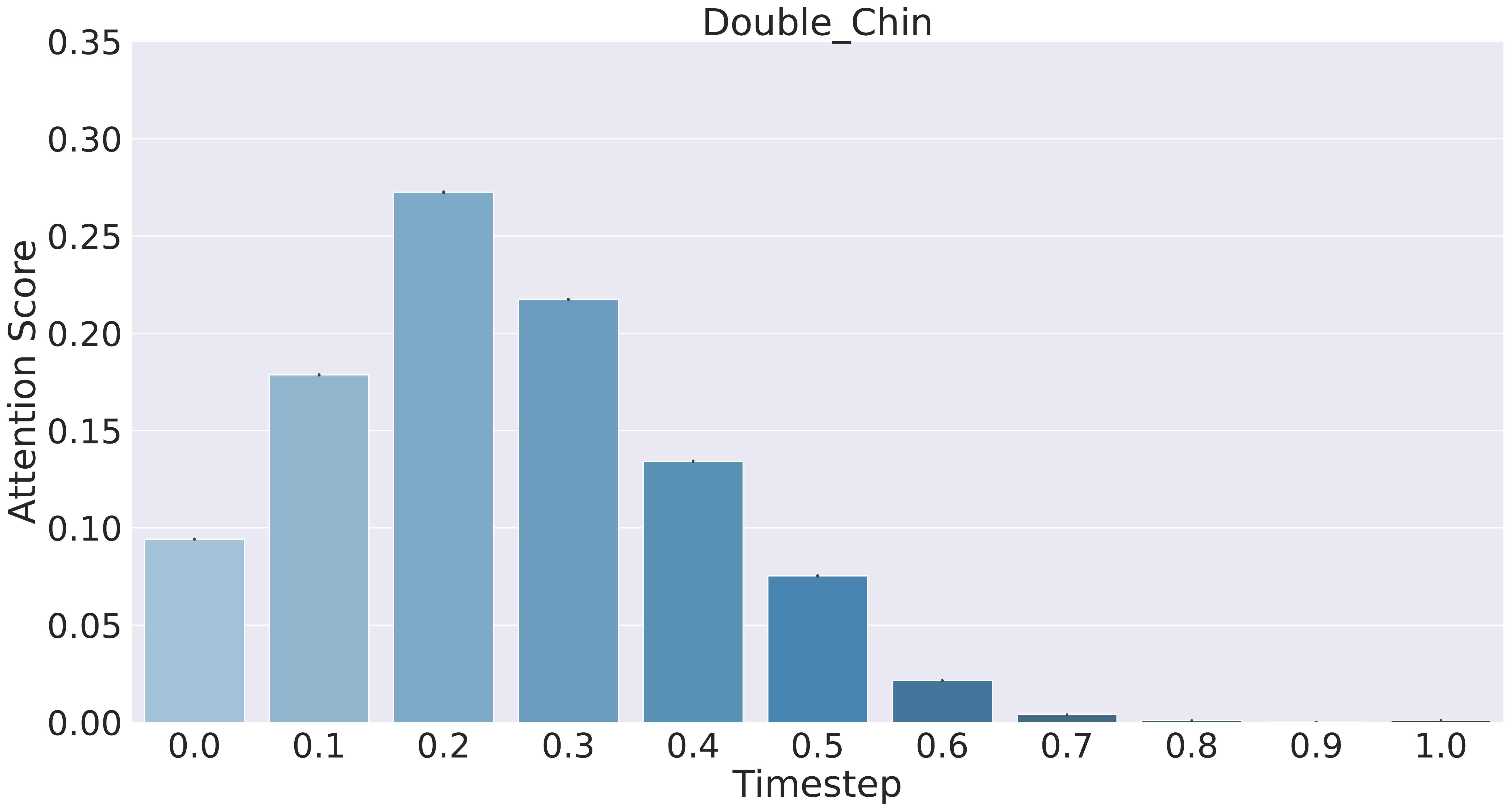}
\end{minipage}
\begin{minipage}[c]{0.24\textwidth}
\includegraphics[width=\textwidth]{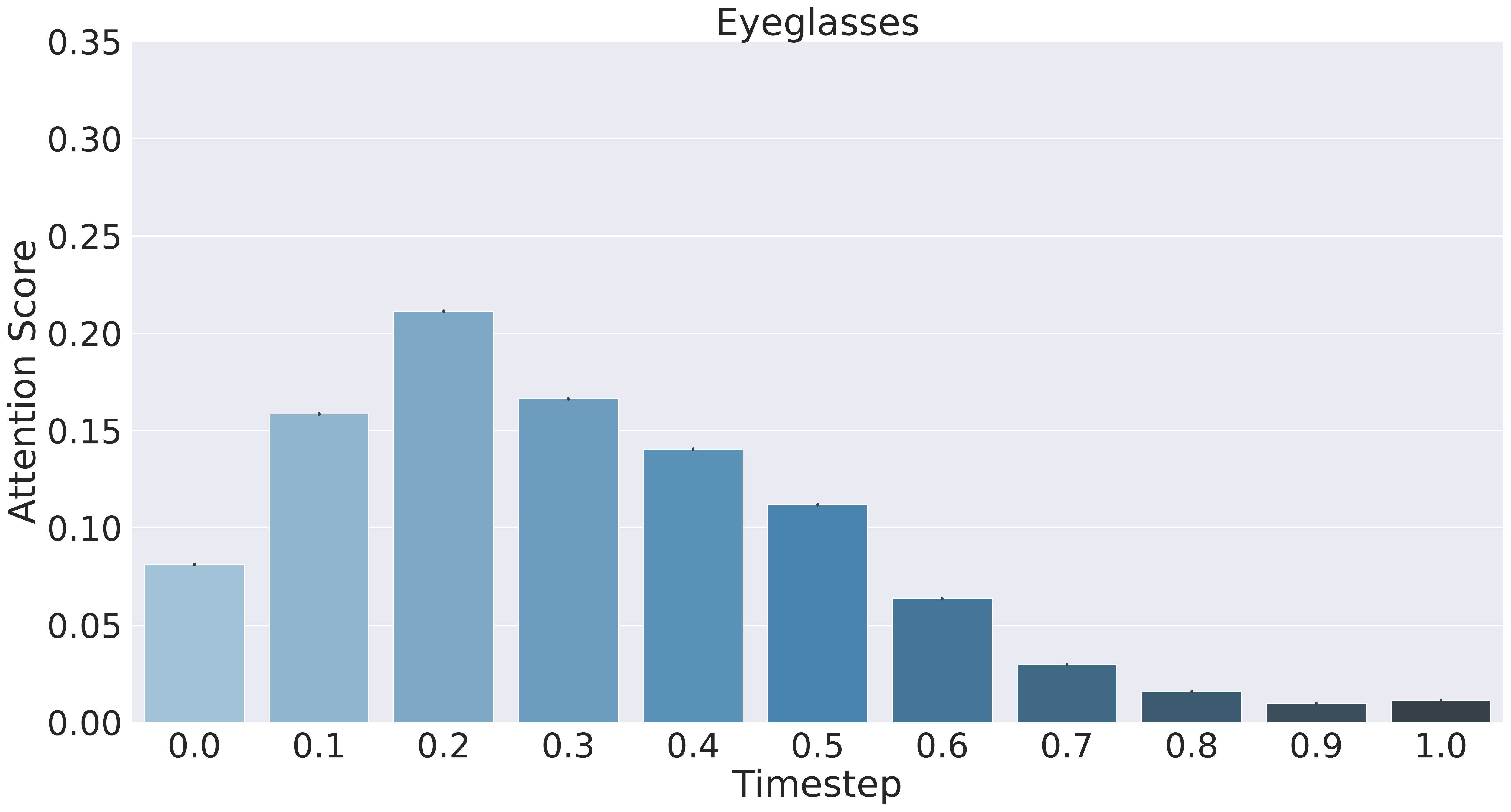}
\end{minipage}
\begin{minipage}[c]{0.24\textwidth}
\includegraphics[width=\textwidth]{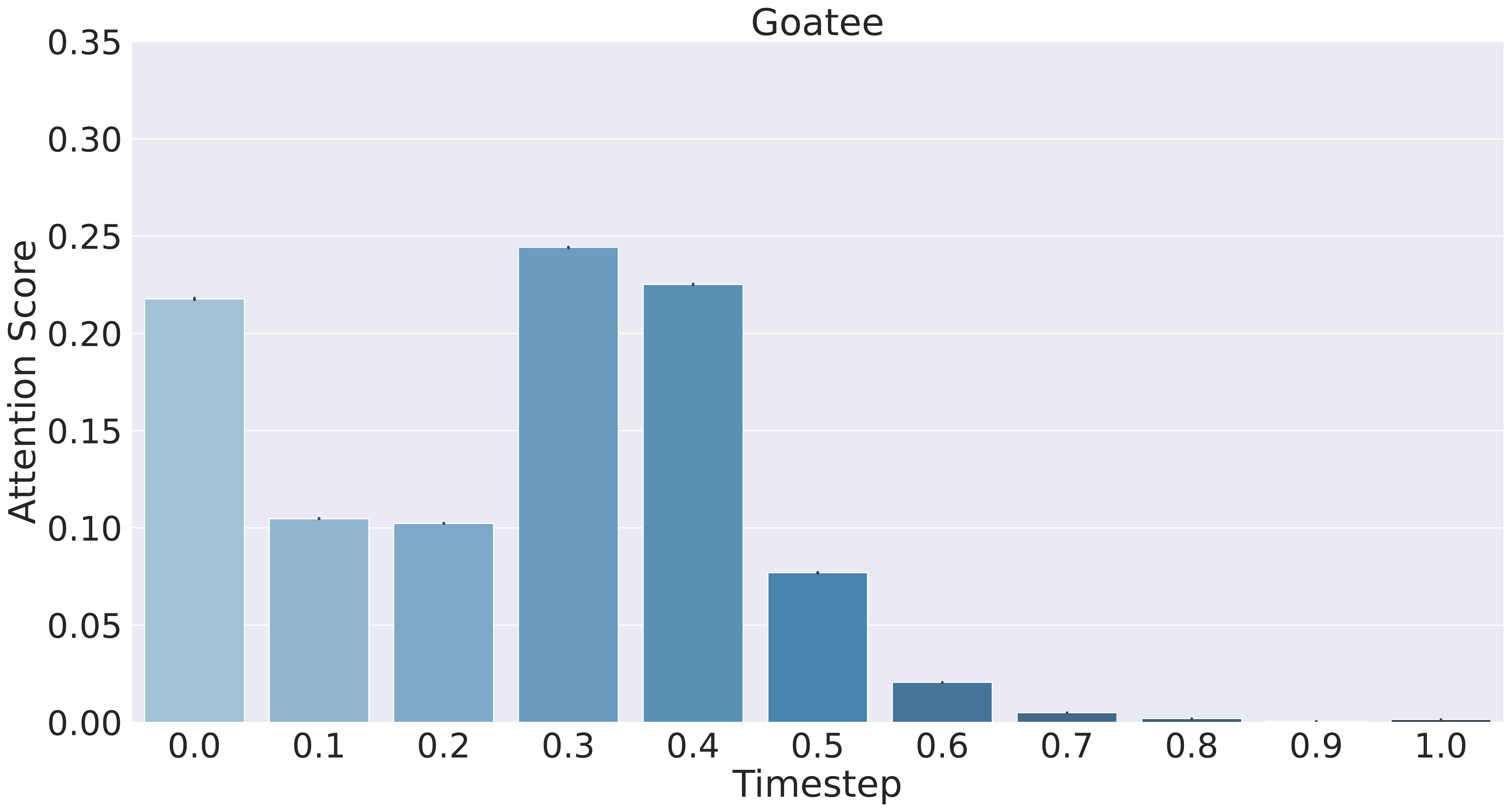}
\end{minipage}
\begin{minipage}[c]{0.24\textwidth}
\includegraphics[width=\textwidth]{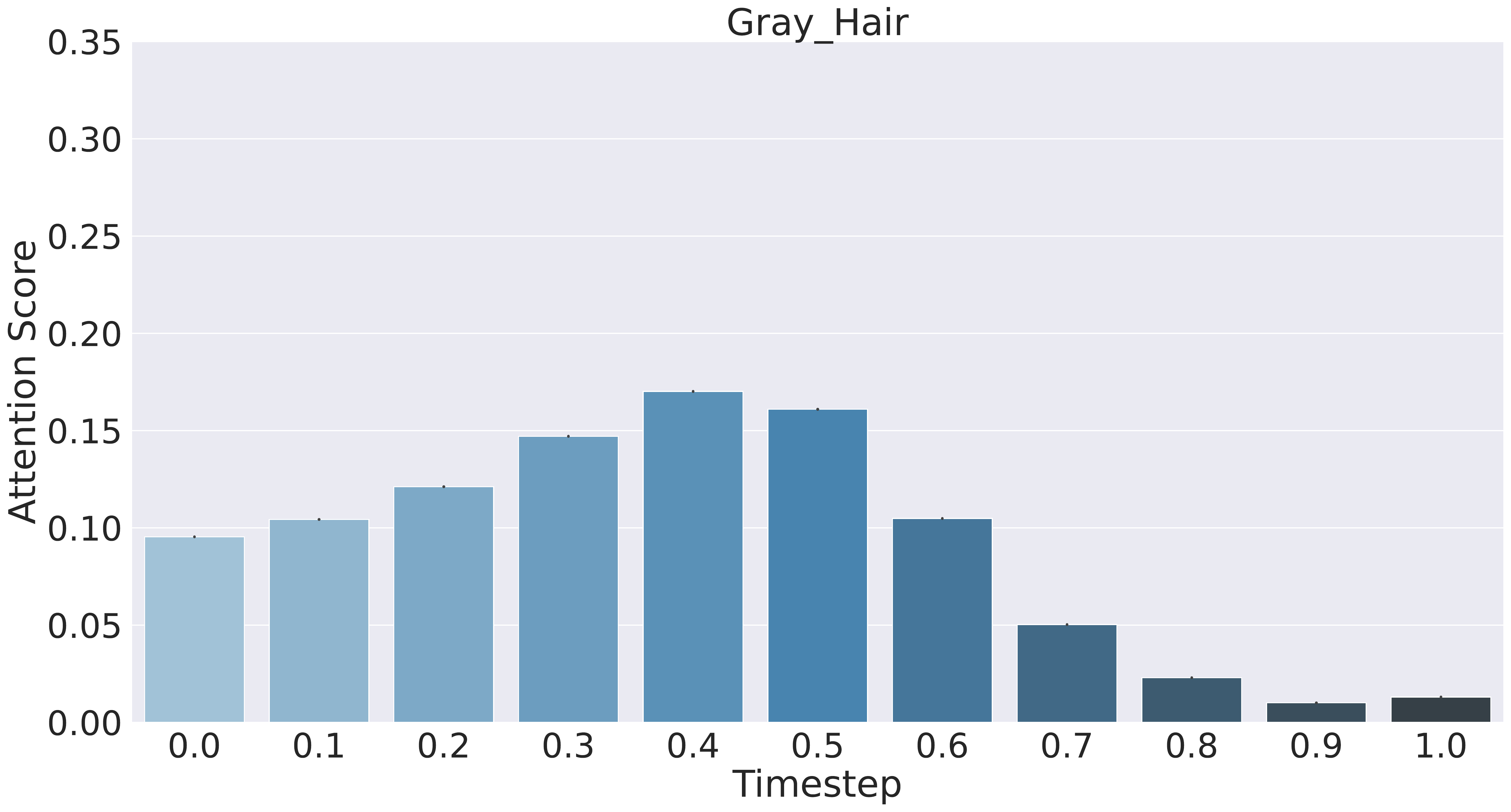}
\end{minipage}
\begin{minipage}[c]{0.24\textwidth}
\includegraphics[width=\textwidth]{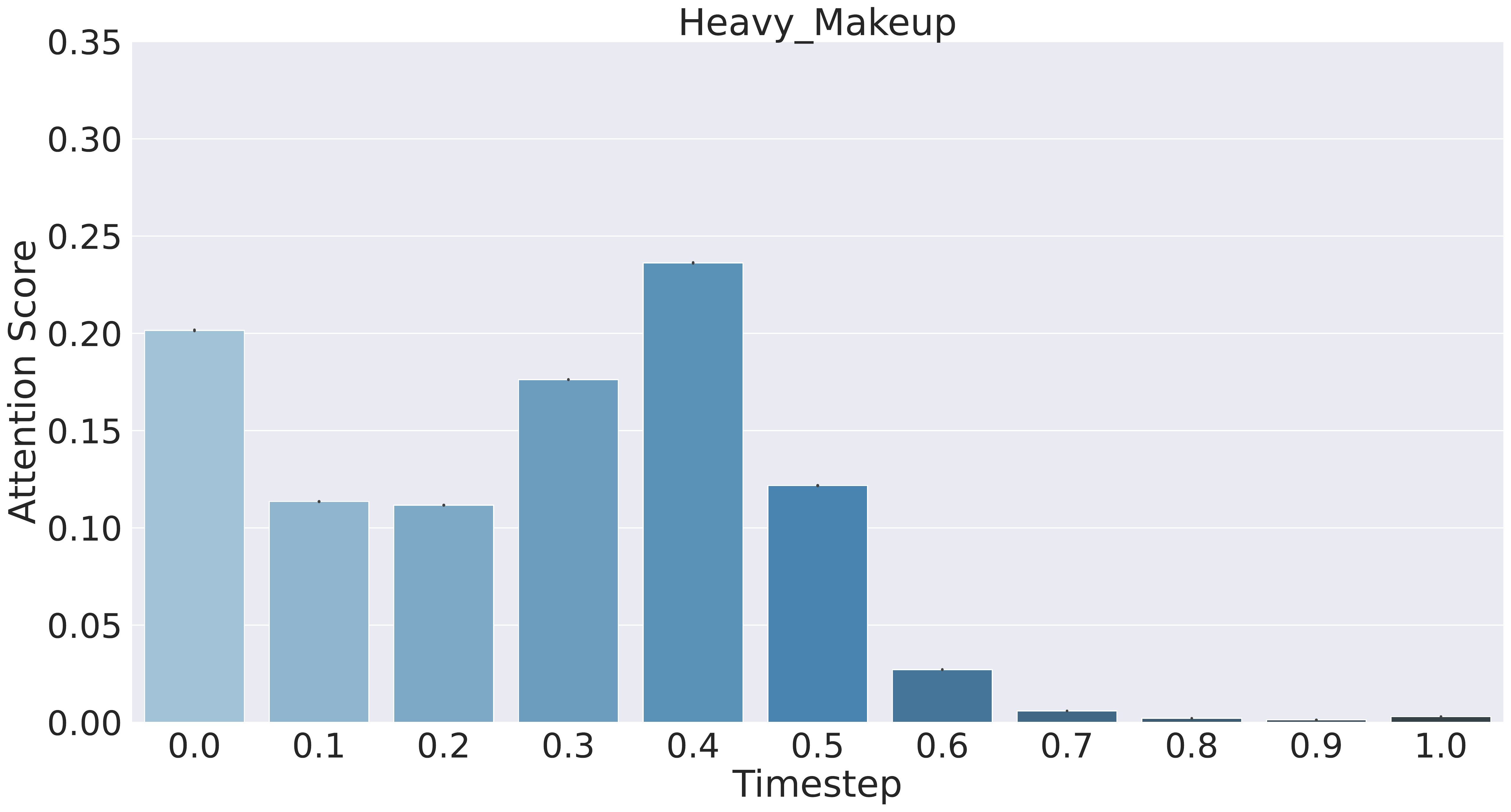}
\end{minipage}
\begin{minipage}[c]{0.24\textwidth}
\includegraphics[width=\textwidth]{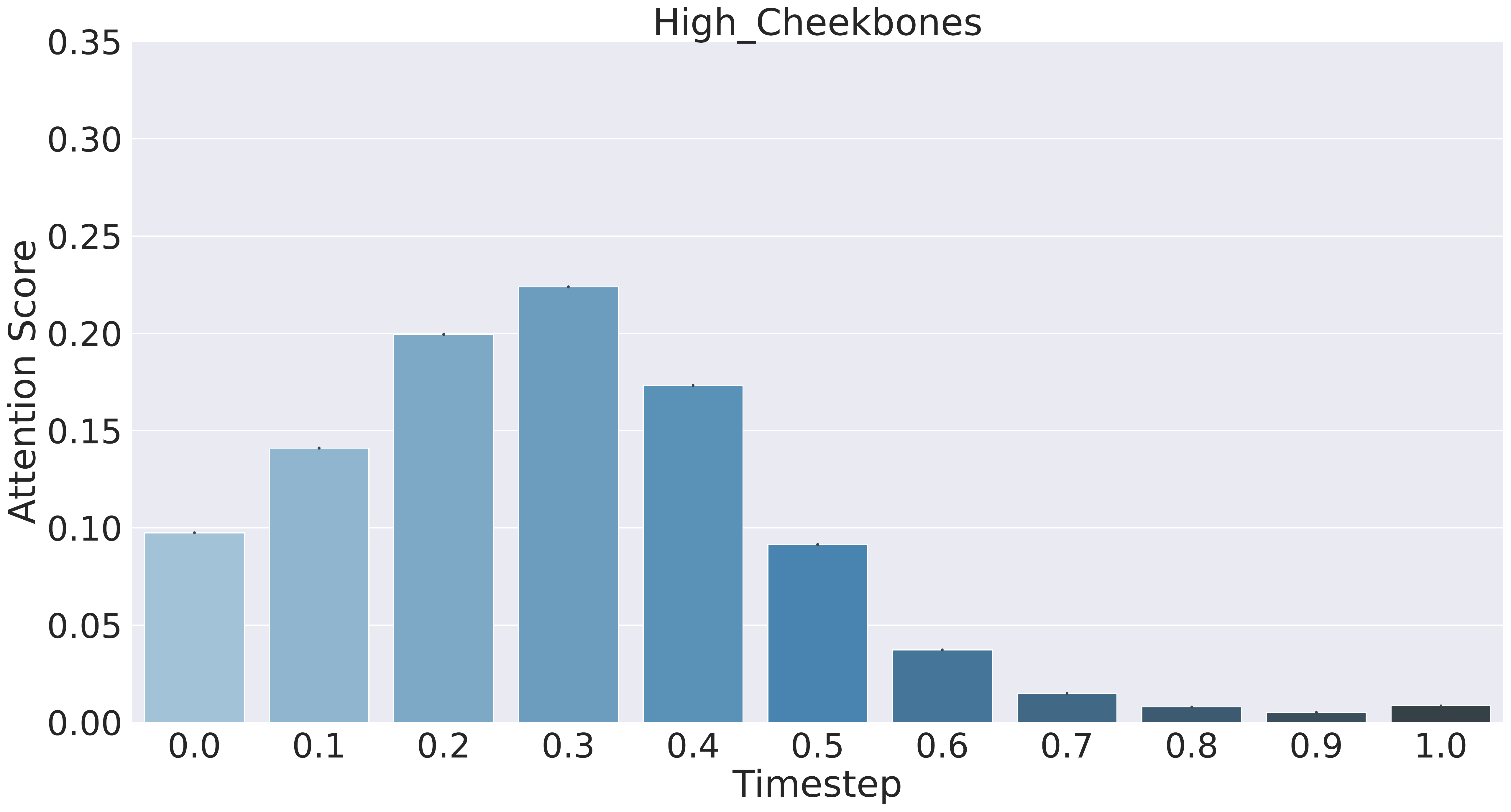}
\end{minipage}
\begin{minipage}[c]{0.24\textwidth}
\includegraphics[width=\textwidth]{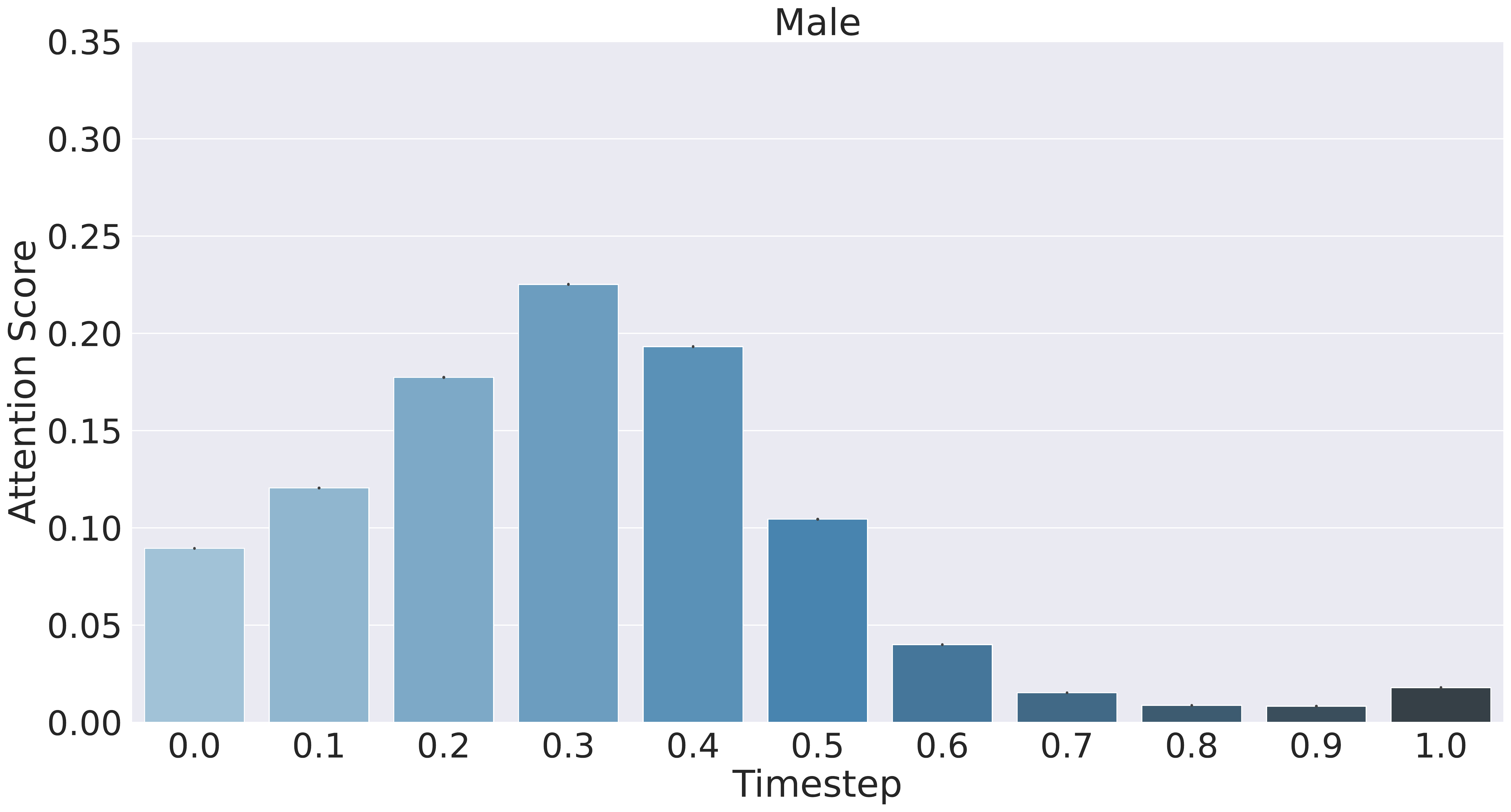}
\end{minipage}
\begin{minipage}[c]{0.24\textwidth}
\includegraphics[width=\textwidth]{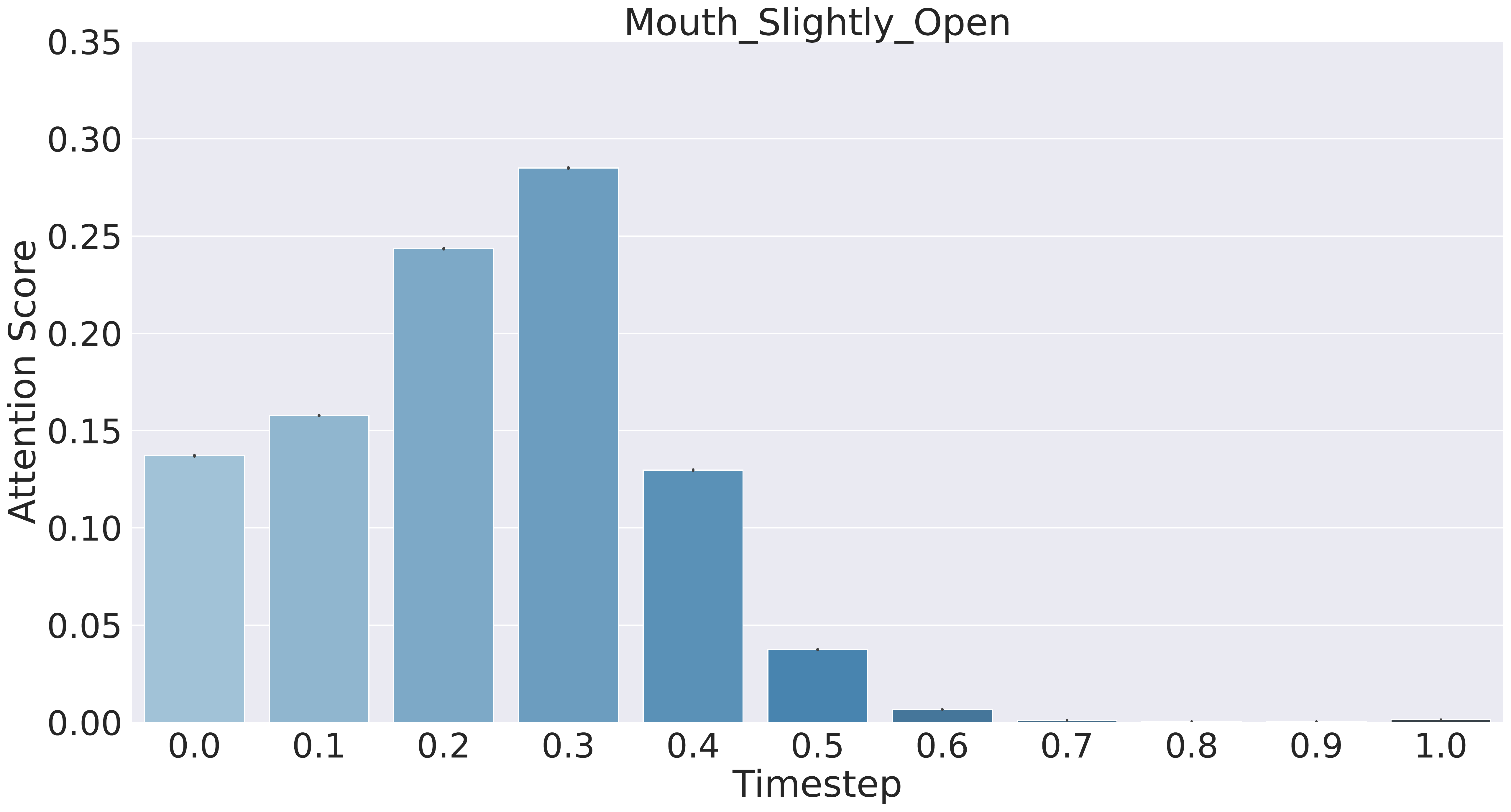}
\end{minipage}
\begin{minipage}[c]{0.24\textwidth}
\includegraphics[width=\textwidth]{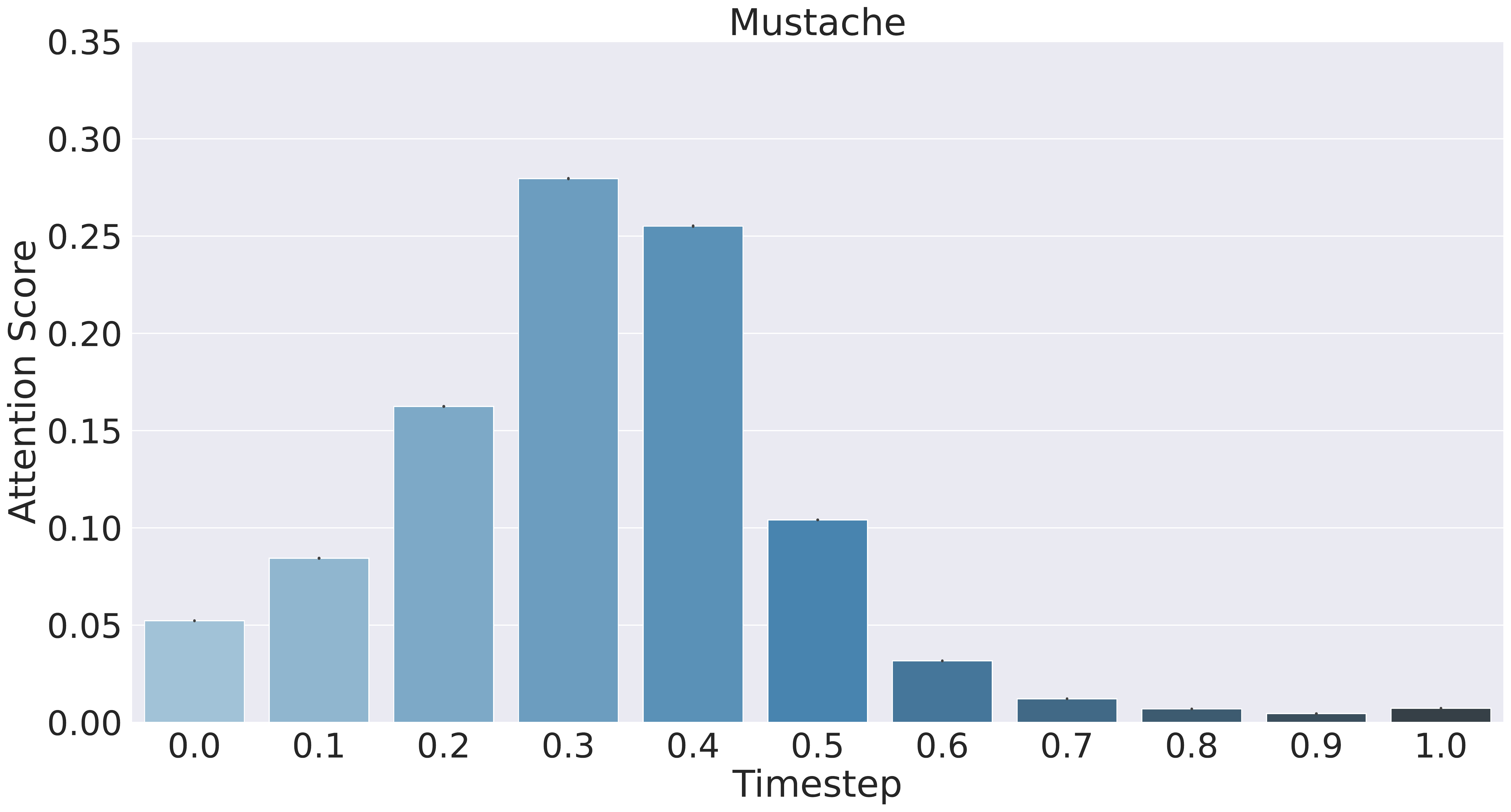}
\end{minipage}
\begin{minipage}[c]{0.24\textwidth}
\includegraphics[width=\textwidth]{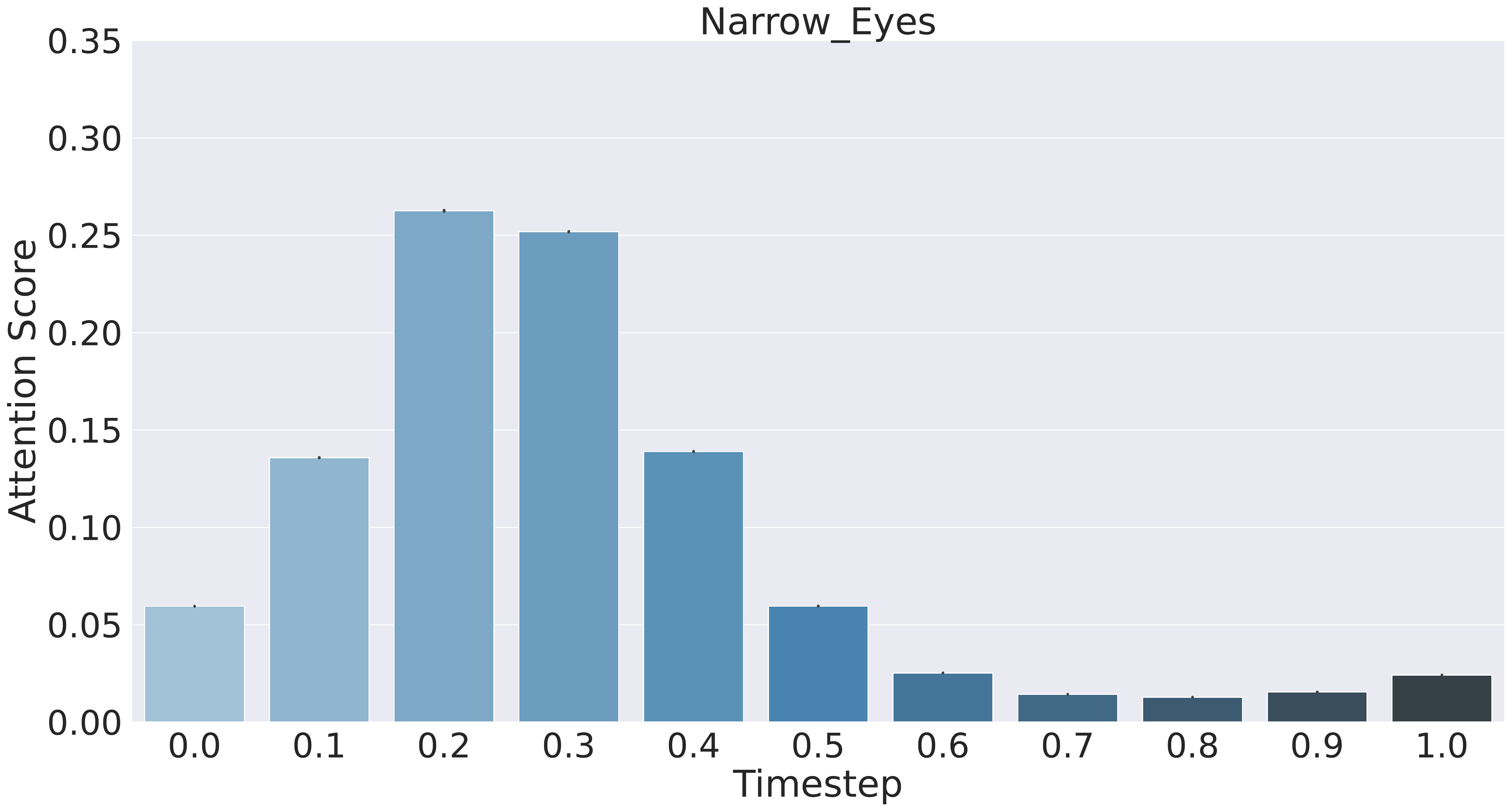}
\end{minipage}
\begin{minipage}[c]{0.24\textwidth}
\includegraphics[width=\textwidth]{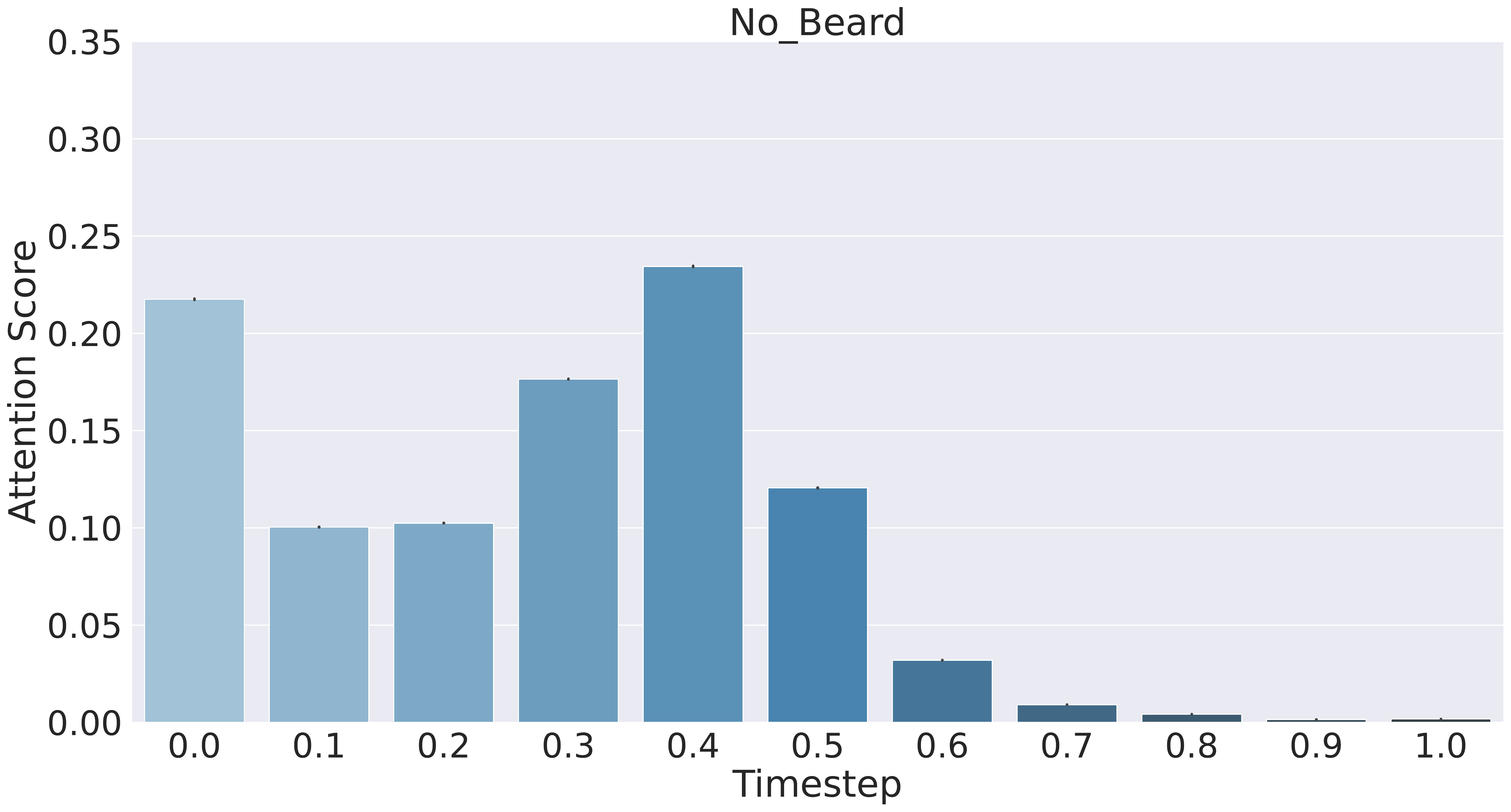}
\end{minipage}
\begin{minipage}[c]{0.24\textwidth}
\includegraphics[width=\textwidth]{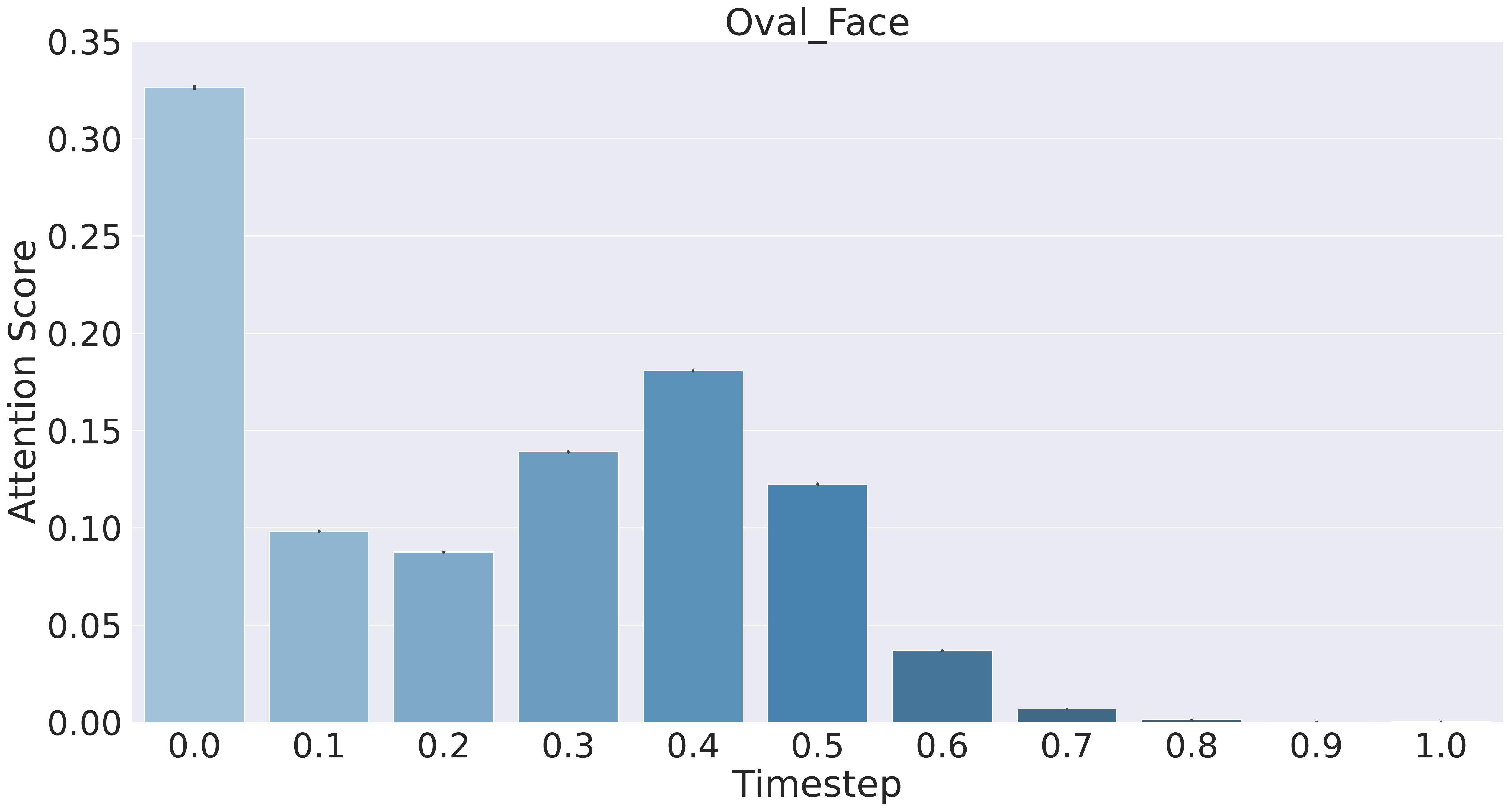}
\end{minipage}
\begin{minipage}[c]{0.24\textwidth}
\includegraphics[width=\textwidth]{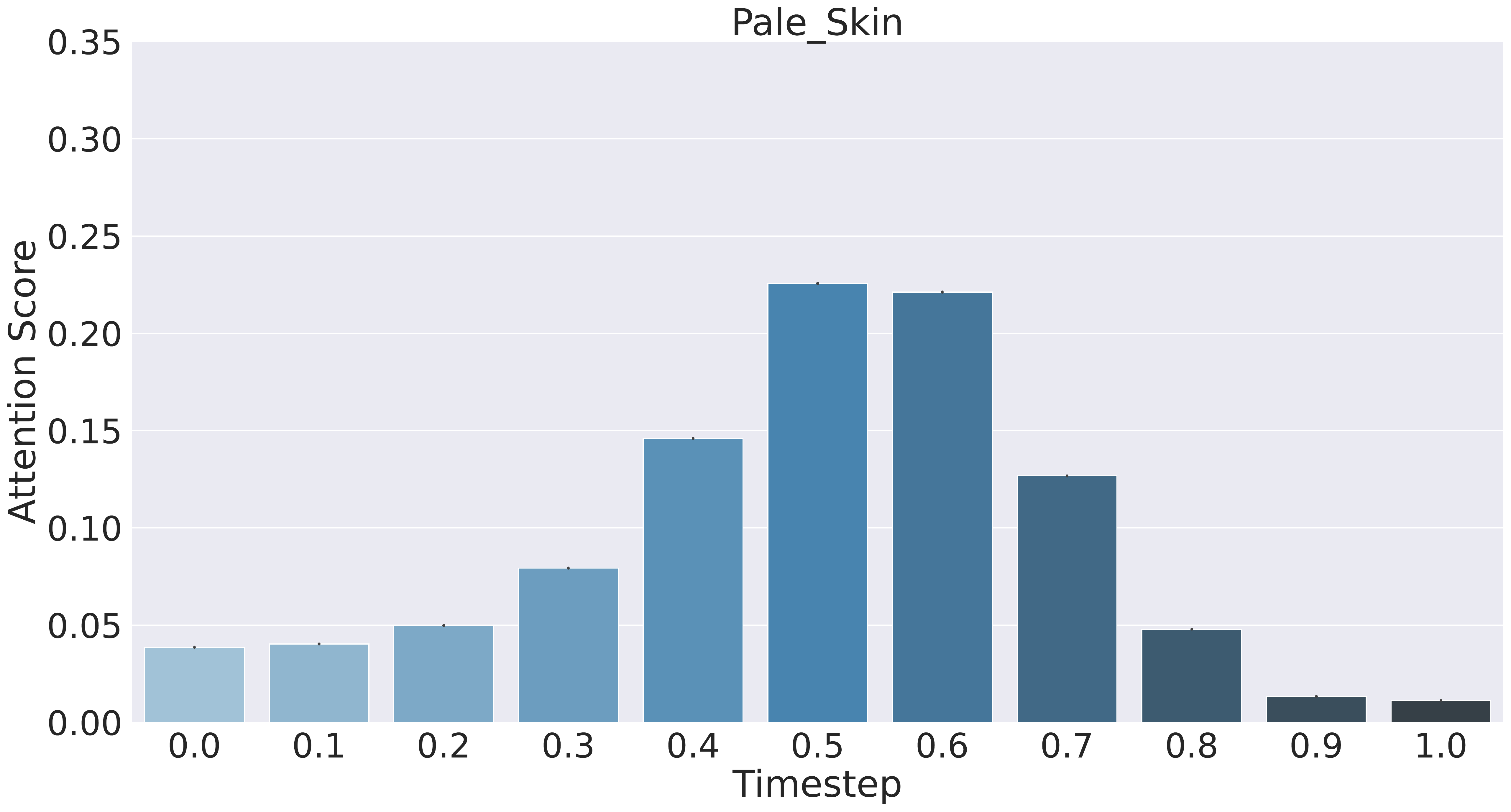}
\end{minipage}
\begin{minipage}[c]{0.24\textwidth}
\includegraphics[width=\textwidth]{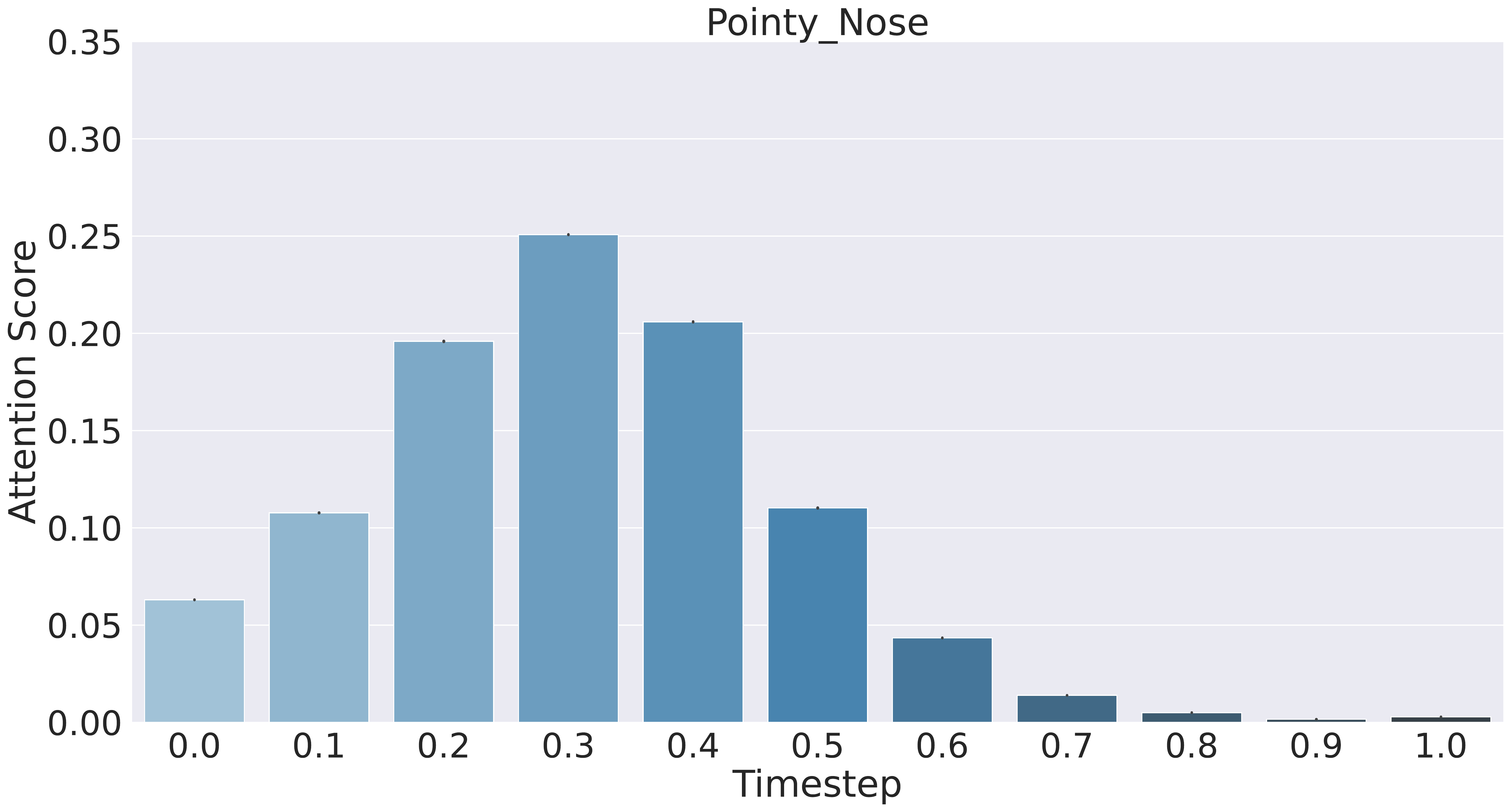}
\end{minipage}
\begin{minipage}[c]{0.24\textwidth}
\includegraphics[width=\textwidth]{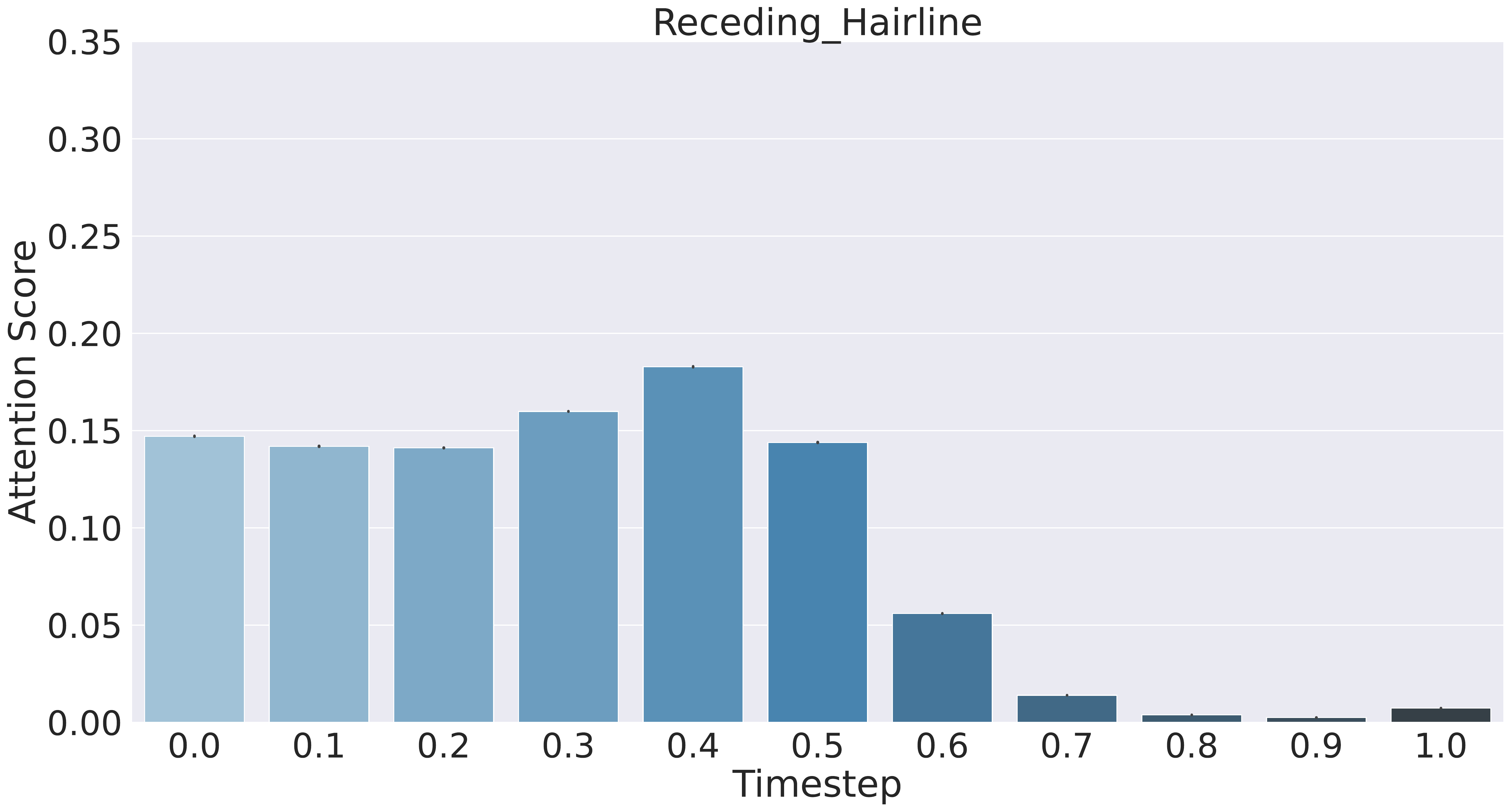}
\end{minipage}
\begin{minipage}[c]{0.24\textwidth}
\includegraphics[width=\textwidth]{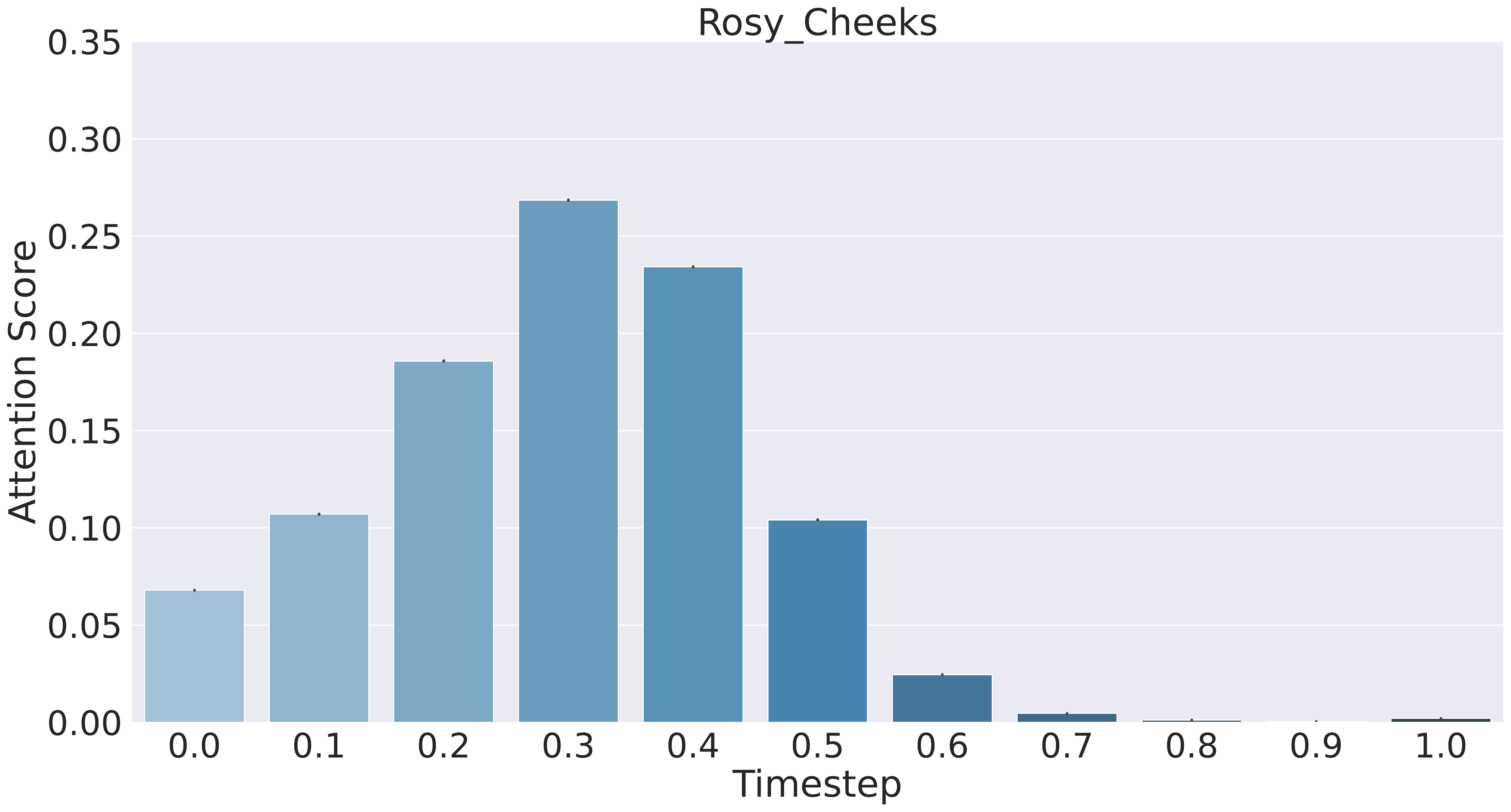}
\end{minipage}
\begin{minipage}[c]{0.24\textwidth}
\includegraphics[width=\textwidth]{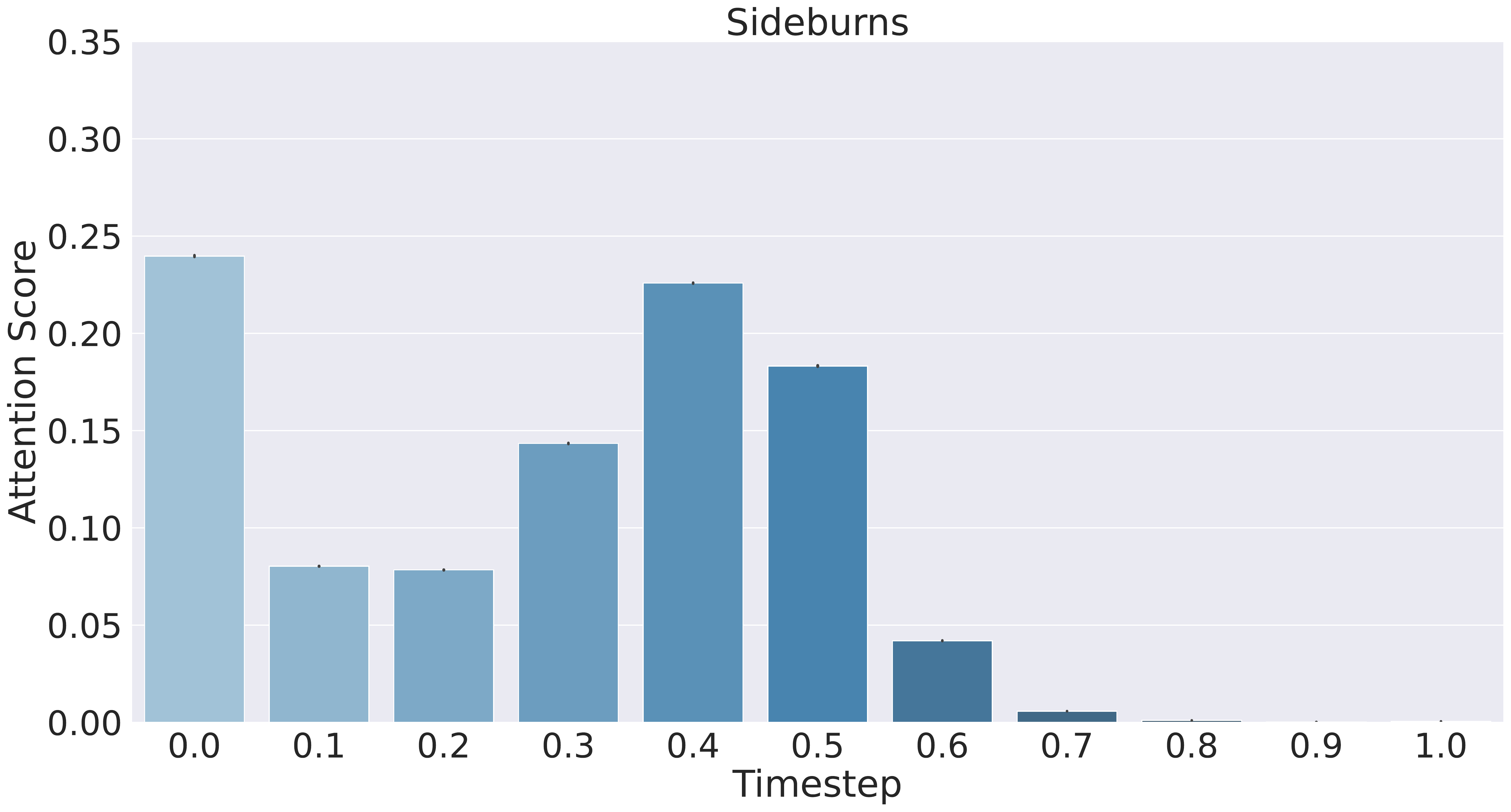}
\end{minipage}
\begin{minipage}[c]{0.24\textwidth}
\includegraphics[width=\textwidth]{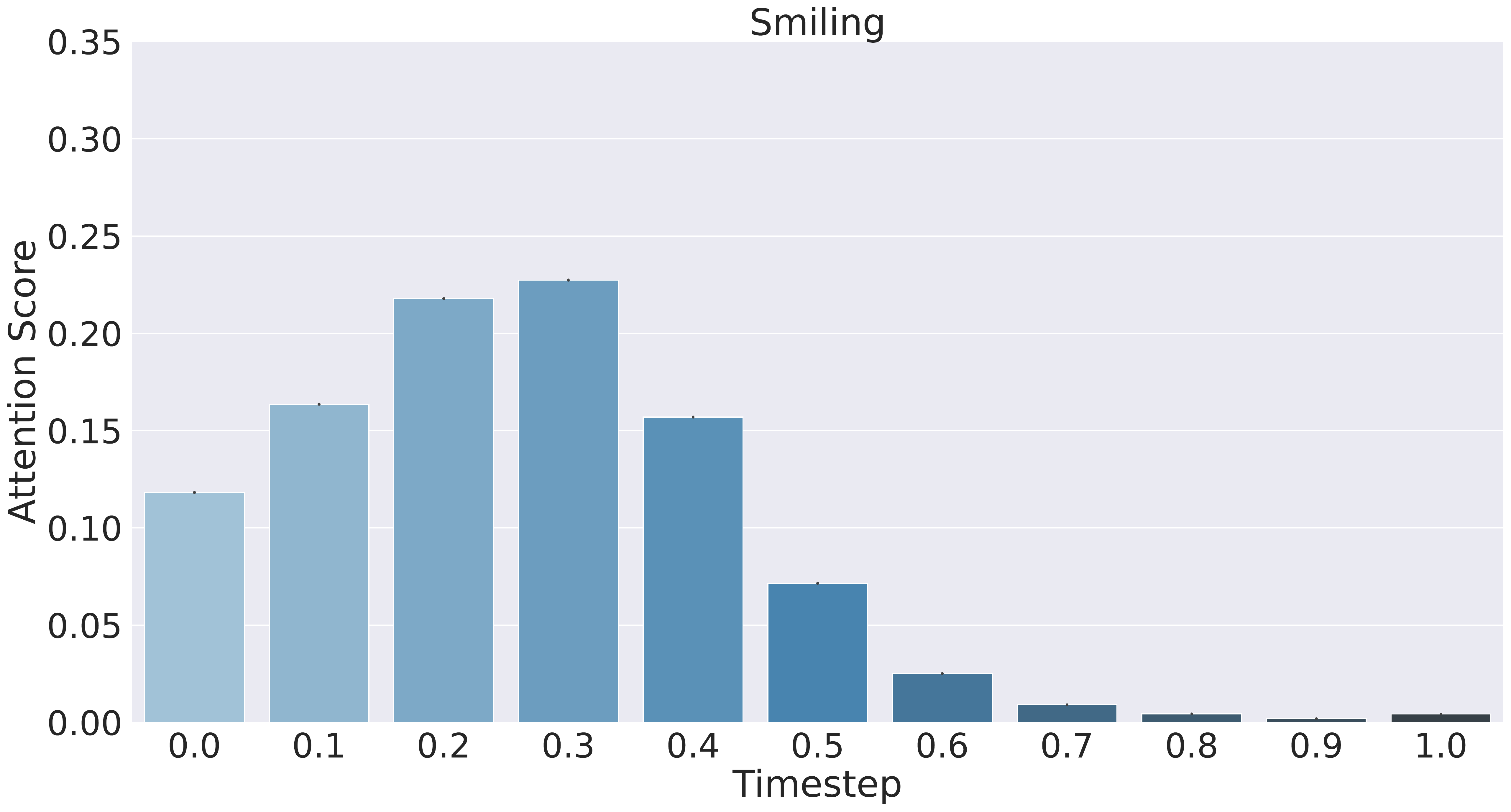}
\end{minipage}
\begin{minipage}[c]{0.24\textwidth}
\includegraphics[width=\textwidth]{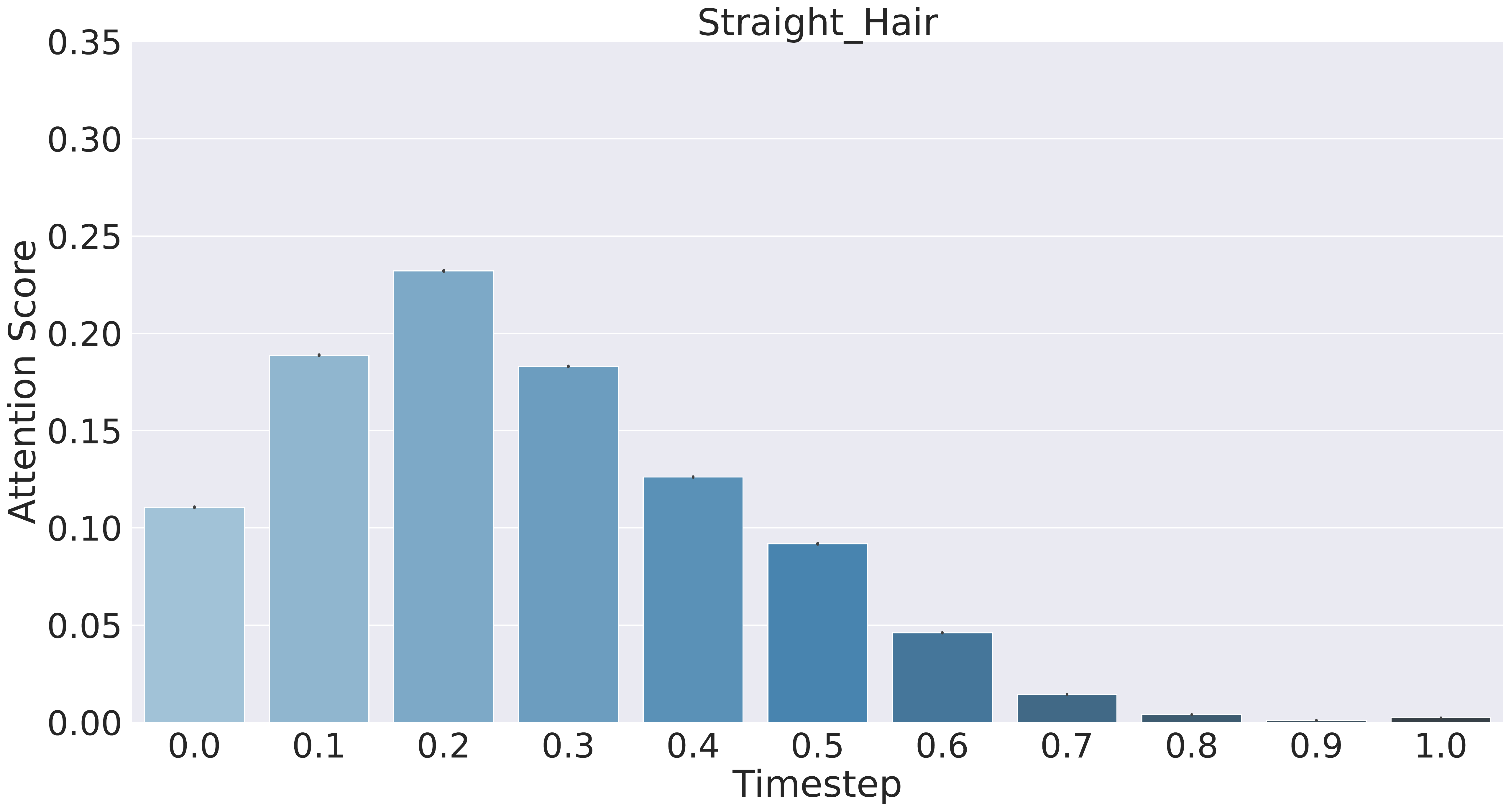}
\end{minipage}
\begin{minipage}[c]{0.24\textwidth}
\includegraphics[width=\textwidth]{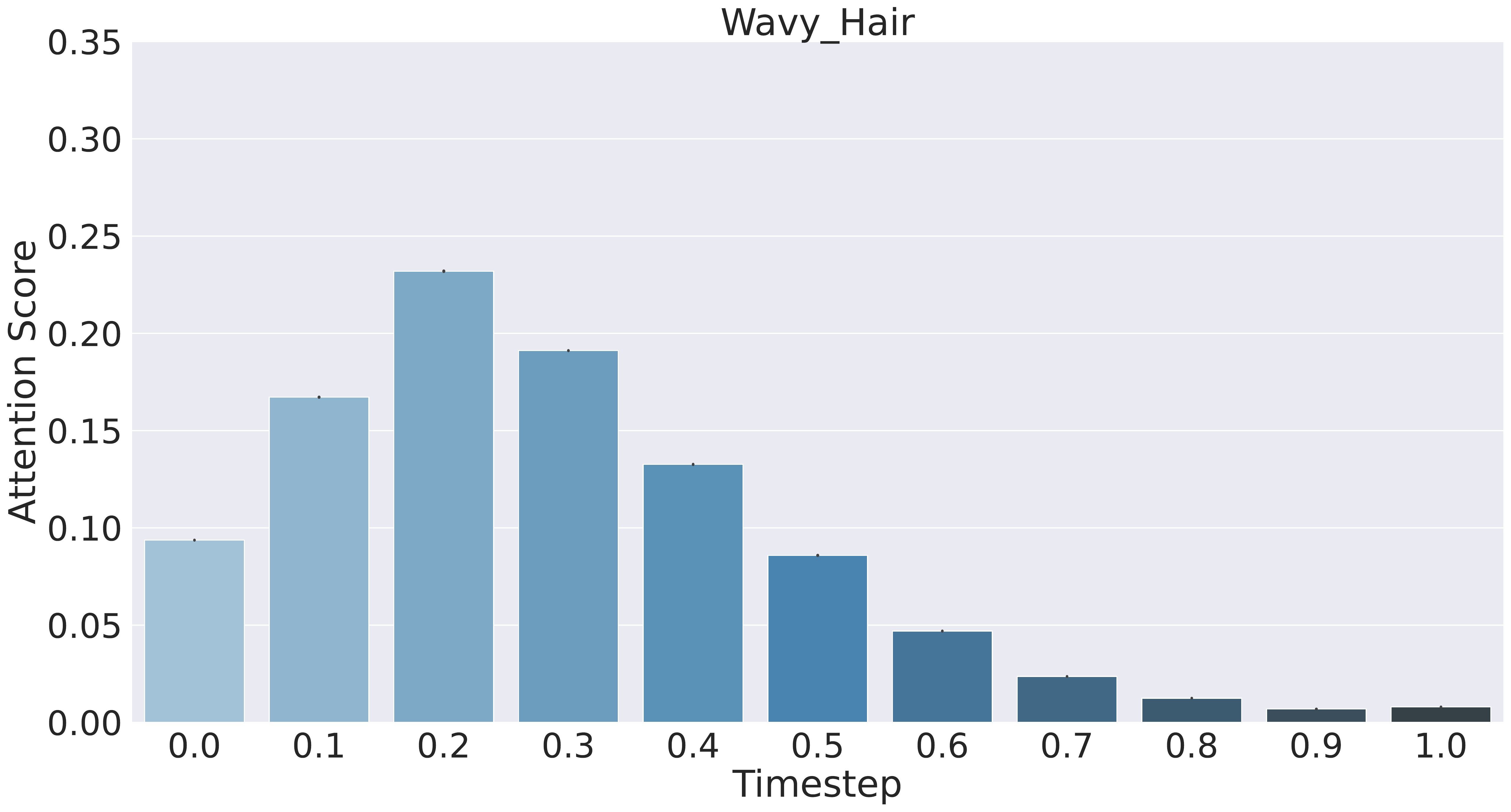}
\end{minipage}
\begin{minipage}[c]{0.24\textwidth}
\includegraphics[width=\textwidth]{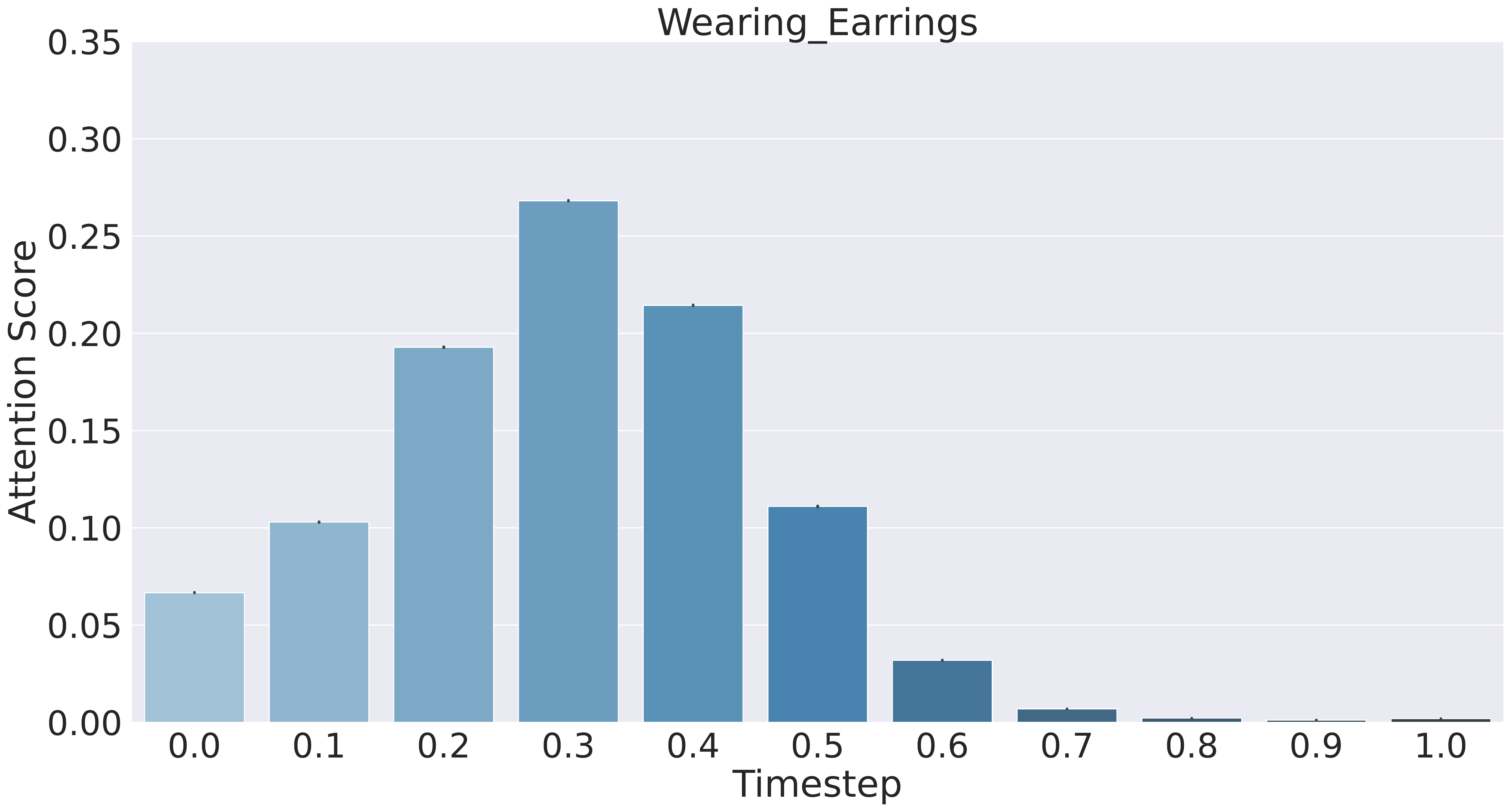}
\end{minipage}
\begin{minipage}[c]{0.24\textwidth}
\includegraphics[width=\textwidth]{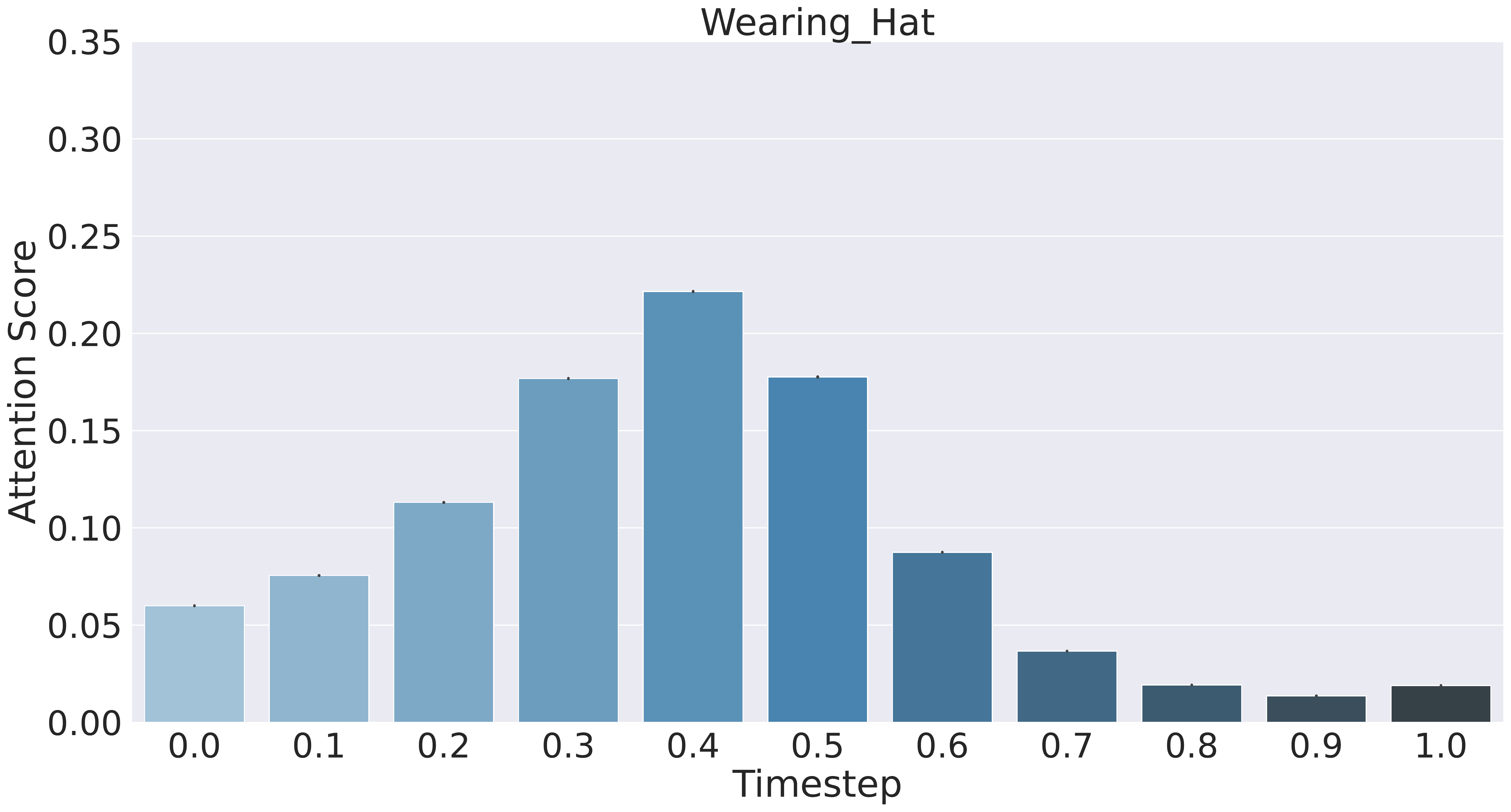}
\end{minipage}
\begin{minipage}[c]{0.24\textwidth}
\includegraphics[width=\textwidth]{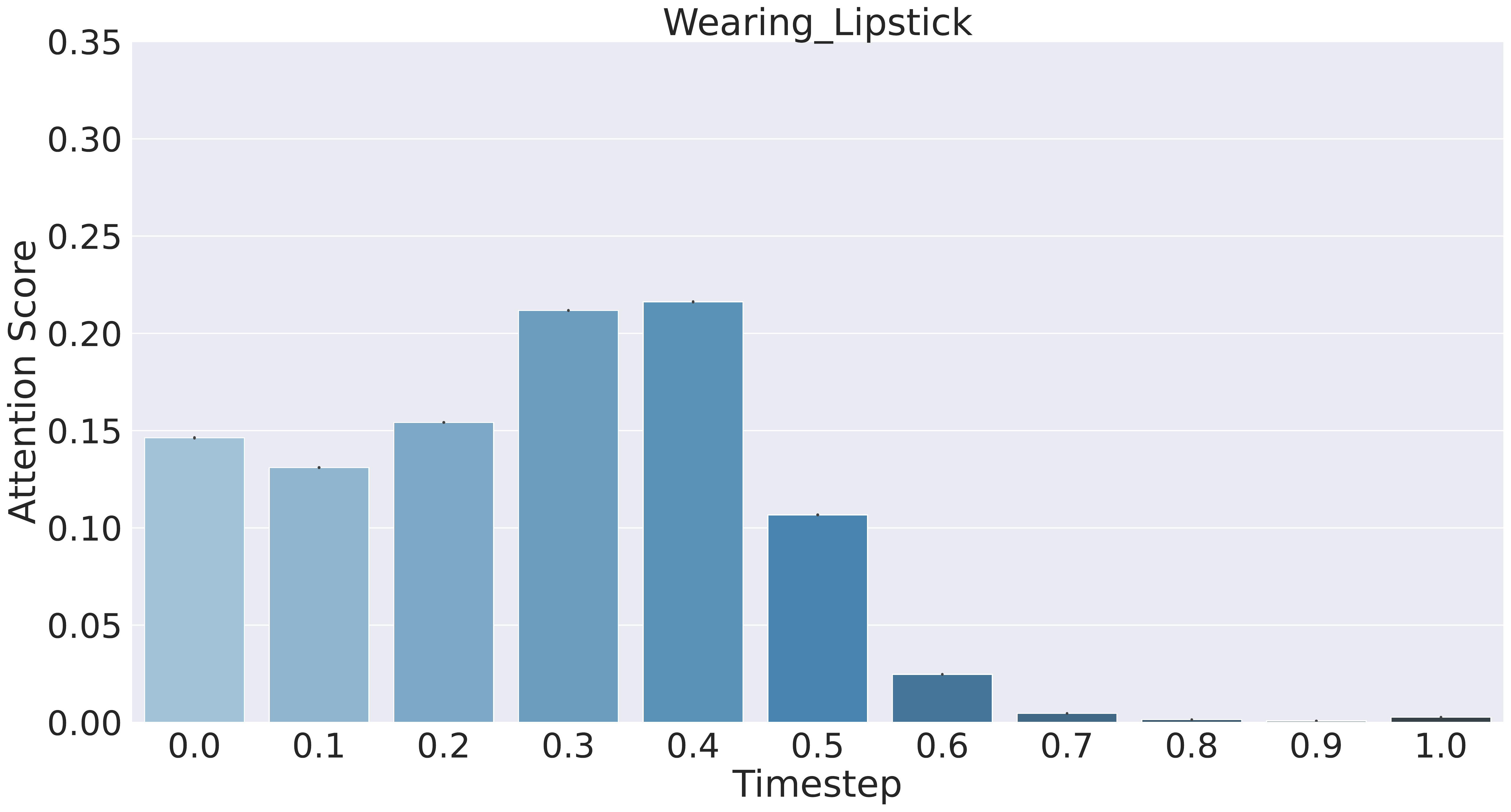}
\end{minipage}
\begin{minipage}[c]{0.24\textwidth}
\includegraphics[width=\textwidth]{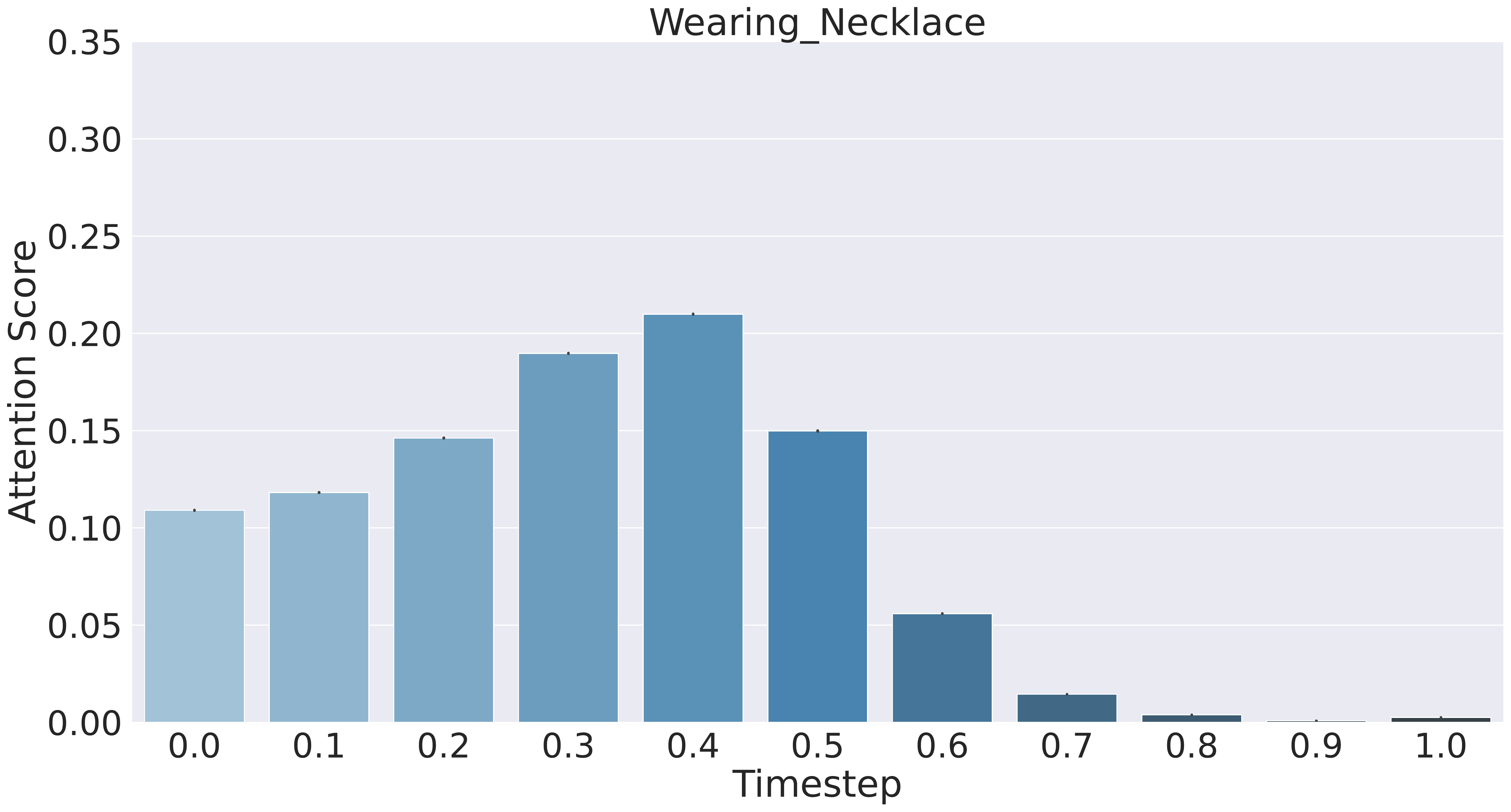}
\end{minipage}
\begin{minipage}[c]{0.24\textwidth}
\includegraphics[width=\textwidth]{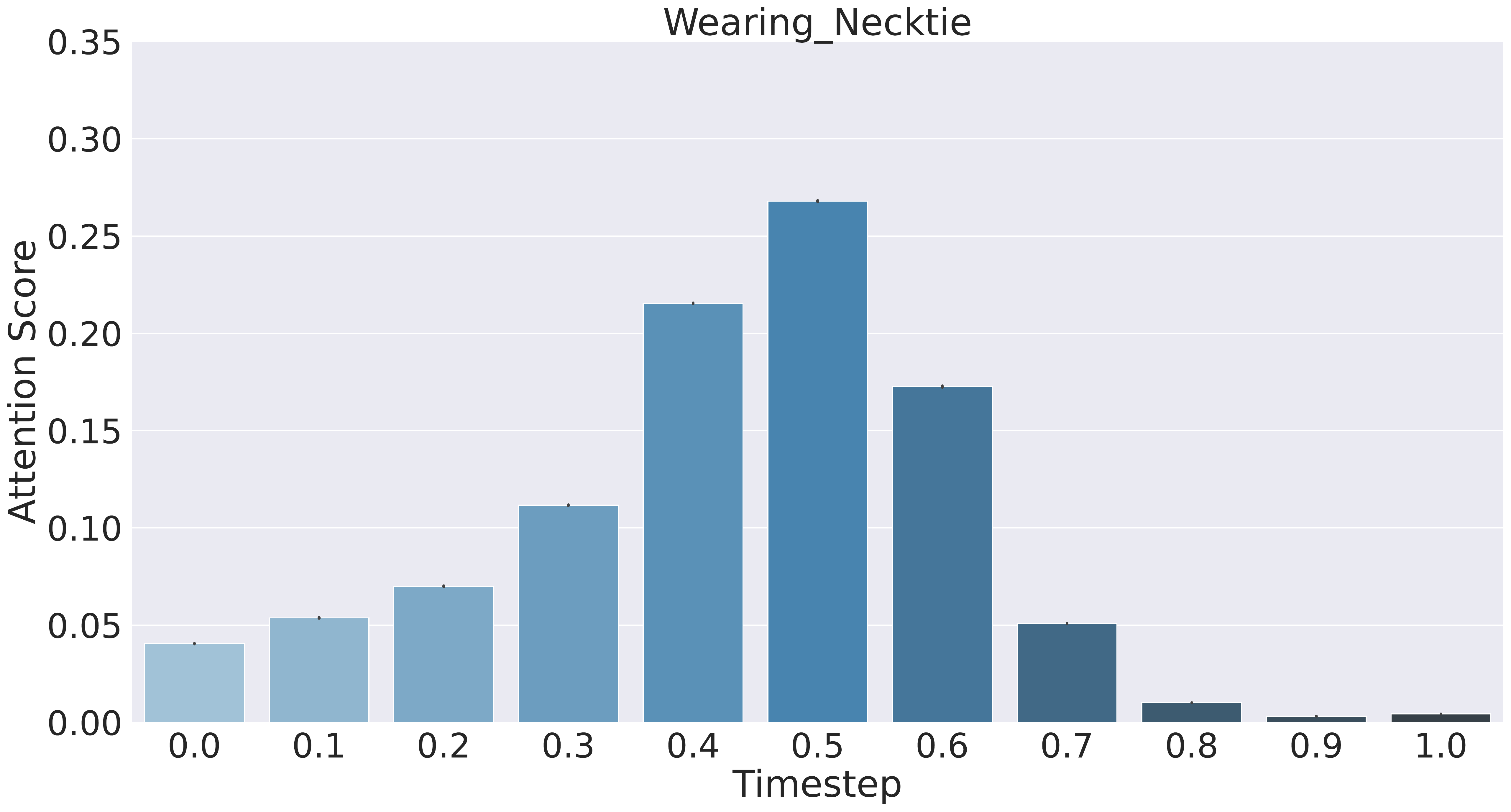}
\end{minipage}
\begin{minipage}[c]{0.24\textwidth}
\includegraphics[width=\textwidth]{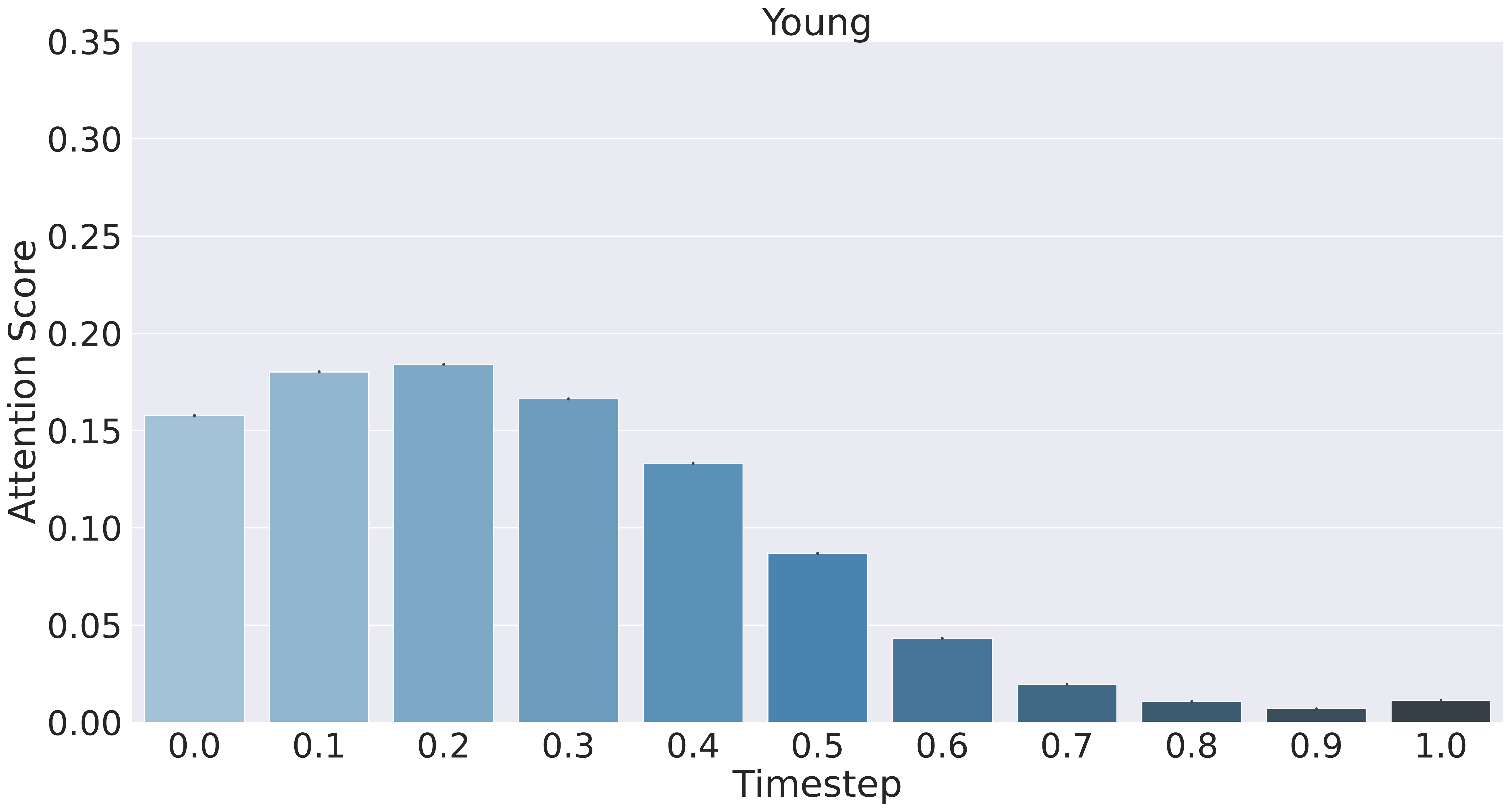}
\end{minipage}
\caption{Attention Score profiles for different parts of the trajectory-based representation on CelebA when using the DRL deterministic encoder.}
\label{fig:celeba_activ_drl}
\end{figure}

We train the score model for 250,000 iterations and then the downstream models for 100 epochs. We additionally perform the Jensen-Shannon Divergence analysis between different features for two different types of encoders; VDRL and DRL. The corresponding plots for these analysis, as well as for the attention score profiles and performances on different features, are present in Figures \ref{fig:celeba_ablation} - \ref{fig:celeba_activ_drl}. The figures also enumerate the different attributes present in the dataset.

\clearpage

\end{document}